\documentstyle[11pt,titlepage,aaai-named,twoside,lingmacros]{report}
\title{A  Lexicalized Tree Adjoining Grammar for English}
\author{The XTAG Research Group \\ \\
Institute for Research in Cognitive Science\\
University of Pennsylvania \\
3401 Walnut St., Suite 400A \\
Philadelphia, PA 19104-6228 \\ \\
{\tt http://www.cis.upenn.edu/\~{}xtag}
}
\date{31 August 1998}

\bigskip

\makeindex
\newcommand{\vertical}[1]{
\setlength{\unitlength}{0.012500in}%

\hspace*{2pt}
\begin{picture}(12,12)(0,0)
\put(0,0){\makebox(0,0)[lb]{\raisebox{0pt}[0pt][0pt]{
\rm #1
}}}
\end{picture}

\hspace*{-18pt}
}

\newcommand{\xtagdef}[1]{{\sc #1}}
\newcommand{\xtagcheck}{$\surd$}

\begin{document}
\setcounter{bottomnumber}{20}
\setcounter{topnumber}{20}
\renewcommand{\bottomfraction}{1}
\renewcommand{\topfraction}{1}
\setcounter{totalnumber}{30}
\renewcommand{\textfraction}{0}
\renewcommand{\floatpagefraction}{0}
\input{psfig}
\pagestyle{plain}

\maketitle
\pagenumbering{roman}
\newpage\setcounter{page}{0}\mbox{}\newpage
\tableofcontents
\cleardoublepage
\listoffigures
\cleardoublepage
\begin{abstract} 

This document describes a sizable grammar of English written in the
TAG formalism and implemented for use with the XTAG system. This
report and the grammar described herein supersedes the TAG grammar
described in \cite{tech-rept95}. The English grammar described in this
report is based on the TAG formalism developed in \cite{joshi75},
which has been extended to include lexicalization (\cite{schabes88}),
and unification-based feature structures (\cite{vijay91}).  The range
of syntactic phenomena that can be handled is large and includes
auxiliaries (including inversion), copula, raising and small clause
constructions, topicalization, relative clauses, infinitives, gerunds,
passives, adjuncts, it-clefts, wh-clefts, PRO constructions, noun-noun
modifications, extraposition, determiner sequences, genitives,
negation, noun-verb contractions, sentential adjuncts and imperatives.
This technical report corresponds to the XTAG Release 8/31/98. The
XTAG grammar is continuously updated with the addition of new analyses
and modification of old ones, and an online version of this report can
be found at the XTAG web page: {\tt
  http://www.cis.upenn.edu/\~{}xtag/}.

\end{abstract}

\newpage\setcounter{page}{0}\mbox{}\newpage
\pagestyle{plain}
\null\vfil
\begin{center}
{\bf Acknowledgements}
\end{center}
\setcounter{page}{0}

We are immensely grateful to Aravind Joshi for supporting this
project. 

The following people have contributed to the development of grammars
in the project: Anne Abeille, Jason Baldridge, Rajesh Bhatt, Kathleen
Bishop, Raman Chandrasekar, Sharon Cote, Beatrice Daille, Christine
Doran, Dania Egedi, Tim Farrington, Jason Frank, Caroline Heycock,
Beth Ann Hockey, Roumyana Izvorski, Karin Kipper, Daniel Karp, Seth
Kulick, Young-Suk Lee, Heather Matayek, Patrick Martin, Megan Moser,
Sabine Petillon, Rashmi Prasad, Laura Siegel, Yves Schabes, Victoria
Tredinnick and Raffaella Zanuttini.

The XTAG system has been developed by: Tilman Becker, Richard
Billington, Andrew Chalnick, Dania Egedi, Devtosh Khare, Albert Lee,
David Magerman, Alex Mallet, Patrick Paroubek, Rich Pito, Gilles
Prigent, Carlos Prolo, Anoop Sarkar, Yves Schabes, William Schuler,
B. Srinivas, Fei Xia, Yuji Yoshiie and Martin Zaidel.

We would also like to thank Michael Hegarty, Lauri Karttunen, Anthony
Kroch, Mitchell Marcus, Martha Palmer, Owen Rambow, Philip Resnik,
Beatrice Santorini and Mark Steedman.

In addition, Jeff Aaronson, Douglas DeCarlo, Mark-Jason Dominus, Mark
Foster, Gaylord Holder, David Magerman, Ken Noble, Steven Shapiro and
Ira Winston have provided technical support.  Adminstrative support
was provided by Susan Deysher, Carolyn Elken, Jodi Kerper, Christine
Sandy and Trisha Yannuzzi.

This work was partially supported by  NSF Grant SBR8920230 and ARO Grant
DAAH0404-94-G-0426. 

\newpage

\newpage\setcounter{page}{0}\mbox{}\newpage
\pagenumbering{arabic}
\pagestyle{headings}
\part{General Information}
\chapter{Getting Around}

This technical report presents the English XTAG grammar as implemented by the
XTAG Research Group at the University of Pennsylvania.  The technical report is
organized into four parts, plus a set of appendices.  Part 1 contains general
information about the XTAG system and some of the underlying mechanisms that
help shape the grammar.  Chapter~\ref{intro-FBLTAG} contains an introduction to
the formalism behind the grammar and parser, while Chapter~\ref{overview}
contains information about the entire XTAG system.  Linguists interested solely
in the grammar of the XTAG system may safely skip Chapters~\ref{intro-FBLTAG}
and \ref{overview}.  Chapter~\ref{underview} contains information on some of
the linguistic principles that underlie the XTAG grammar, including the
distinction between complements and adjuncts, and how case is handled.

The actual description of the grammar begins with Part 2, and is contained in
the following three parts.  Parts 2 and 3 contains information on the verb
classes and the types of trees allowed within the verb classes, respectively,
while Part 4 contains information on trees not included in the verb classes
(e.g.  NP's, PP's, various modifiers, etc).  Chapter~\ref{table-intro} of Part
2 contains a table that attempts to provide an overview of the verb classes and
tree types by providing a graphical indication of which tree types are allowed
in which verb classes.  This has been cross-indexed to tree figures shown in
the tech report.  Chapter~\ref{verb-classes} contains an overview of all of the
verb classes in the XTAG grammar.  The rest of Part 2 contains more details on
several of the more interesting verb classes, including ergatives, sentential
subjects, sentential complements, small classes, ditransitives, and it-clefts.

Part 3 contains information on some of the tree types that are available within
the verb classes.  These tree types correspond to what would be transformations
in a movement based approach.  Not all of these types of trees are contained in
all of the verb classes.  The table (previously mentioned) in Part 2 contains a
list of the tree types and indicates which verb classes each occurs in.  

Part 4 focuses on the non-verb class trees in the grammar.  NP's and
determiners are presented in Chapter~\ref{det-comparitives}, while the various
modifier trees are presented in Chapter~\ref{modifiers}.  Auxiliary verbs,
which are classed separate from the verb classes, are presented in
Chapter~\ref{auxiliaries}, while certain types of conjunction are shown in
Chapter~\ref{conjunction}.  The XTAG treatment of comparatives is
presented in Chapter~\ref{compars-chapter}, and our treatment of
punctuation is discussed in Chapter~\ref{punct-chapt}.

Throughout the technical report, mention is occasionally made of
changes or analyses that we hope to incorporate in the future.
Appendix~\ref{future-work} details a list of these and other future
work.  The appendices also contain information on some of the nitty
gritty details of the XTAG grammar, including a system of metarules
which can be used for grammar development and maintenance in
Appendix~\ref{metarules}, a system for the organization of the grammar
in terms of an inheritance hierarchy is in Appendix~\ref{lexorg}, the
tree naming conventions used in XTAG are explained in detail in
Appendix~\ref{tree-naming}, and a comprehensive list of the features
used in the grammar is given in Appendix~{\ref{features}.
  Appendix~\ref{evaluation} contains an evaluation of the XTAG
  grammar, including comparisons with other wide coverage grammars.

\chapter{Feature-Based, Lexicalized Tree Adjoining Grammars}
\label{intro-FBLTAG}

The English grammar described in this report is based on the TAG formalism
(\cite{joshi75}), which has been extended to include lexicalization
(\cite{schabes88}), and unification-based feature structures
(\cite{vijay91}). Tree Adjoining Languages (TALs) fall into the class of mildly
context-sensitive languages, and as such are more powerful than context free
languages.  The TAG formalism in general, and lexicalized TAGs in particular,
are well-suited for linguistic applications.  As first shown by \cite{joshi85}
and \cite{kj87}, the properties of TAGs permit us to encapsulate diverse
syntactic phenomena in a very natural way.  For example, TAG's extended domain
of locality and its factoring of recursion from local dependencies lead, among
other things, to a localization of so-called unbounded dependencies.

\section{TAG formalism}

The primitive elements of the standard TAG formalism are known as elementary
trees.  \xtagdef{Elementary trees} are of two types: initial trees and
auxiliary trees (see Figure \ref{elem-fig}).  In describing natural language,
\xtagdef{initial trees} are minimal linguistic structures that contain no
recursion, i.e. trees containing the phrasal structure of simple sentences,
NP's, PP's, and so forth.  Initial trees are characterized by the following: 1)
all internal nodes are labeled by non-terminals, 2) all leaf nodes are labeled
by terminals, or by non-terminal nodes marked for substitution. An initial tree
is called an X-type initial tree if its root is labeled with type X.

\begin{figure}[htb]
\centering
\psfig{figure=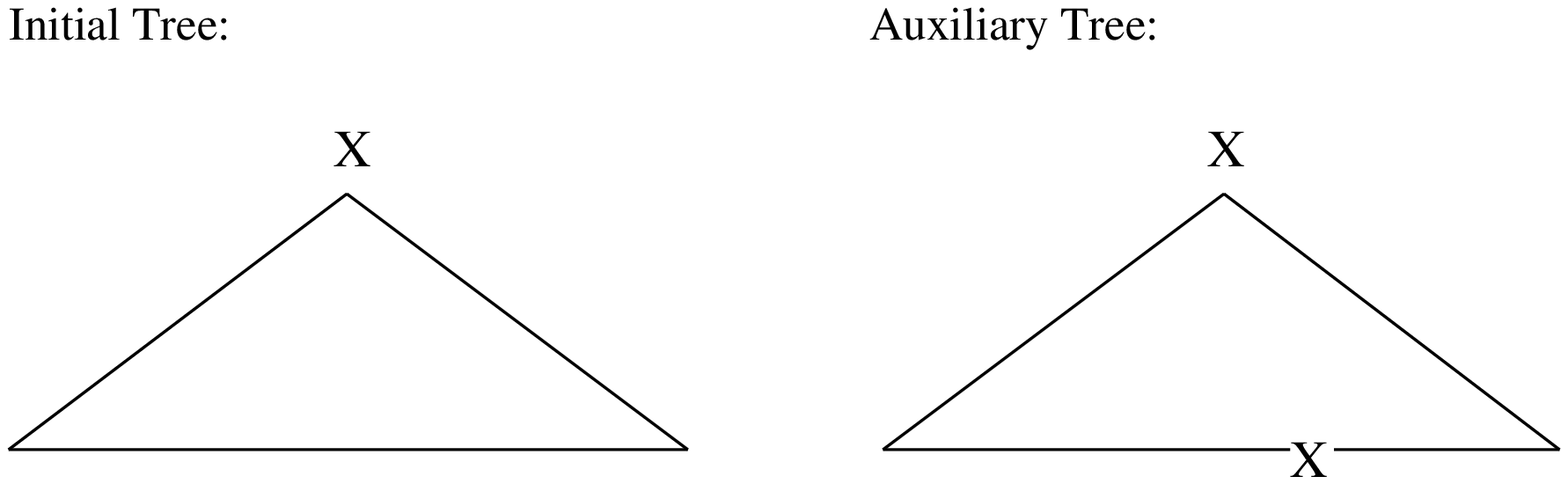,height=1.9in}
\caption{Elementary trees in TAG}
\label{elem-fig}
\end{figure}

Recursive structures are represented by \xtagdef{auxiliary trees}, which
represent constituents that are adjuncts to basic structures (e.g. adverbials).
Auxiliary trees are characterized as follows: 1) all internal nodes are labeled
by non-terminals, 2) all leaf nodes are labeled by terminals, or by
non-terminal nodes marked for substitution, except for exactly one non-terminal
node, called the foot node, which can only be used to adjoin the tree to
another node\footnote{A null adjunction constraint (NA) is systematically put
on the foot node of an auxiliary tree. This disallows adjunction of a tree onto
the foot node itself.}, 3) the foot node has the same label as the root node of
the tree.

There are two operations defined in the TAG formalism,
substitution\footnote{Technically, substitution is a specialized version of
adjunction, but it is useful to make a distinction between the two.} and
adjunction.  In the \xtagdef{substitution} operation, the root node on an
initial tree is merged into a non-terminal leaf node marked for substitution in
another initial tree, producing a new tree.  The root node and the substitution
node must have the same name.  Figure \ref{proto-subst} shows two initial trees
and the tree resulting from the substitution of one tree into the other.

\begin{figure}[htb]
\centering
\psfig{figure=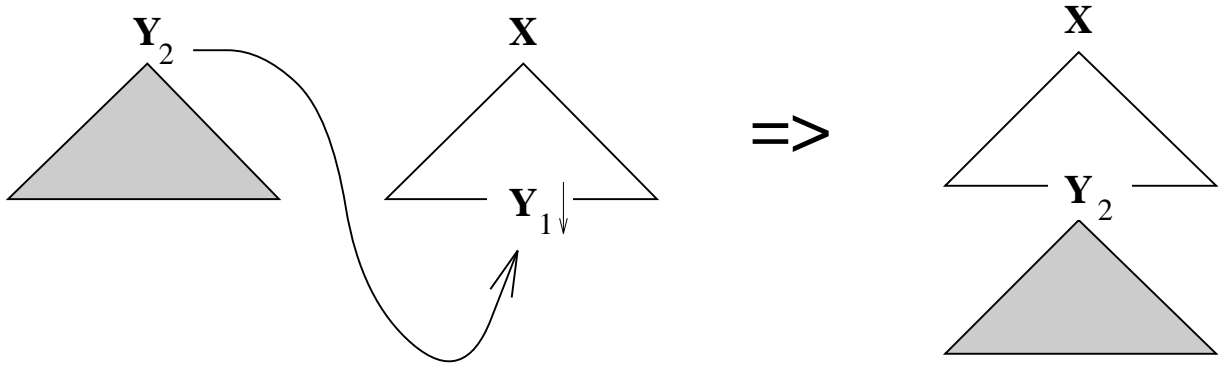,height=1.9in}
\caption{Substitution in TAG}
\label{proto-subst}
\end{figure}

In an \xtagdef{adjunction} operation, an auxiliary tree is grafted onto a
non-terminal node anywhere in an initial tree.  The root and foot nodes of the
auxiliary tree must match the node at which the auxiliary tree adjoins.  Figure
\ref{proto-adjunction} shows an auxiliary tree and an initial tree, and the
tree resulting from an adjunction operation.

\begin{figure}[htb]
\centering
\psfig{figure=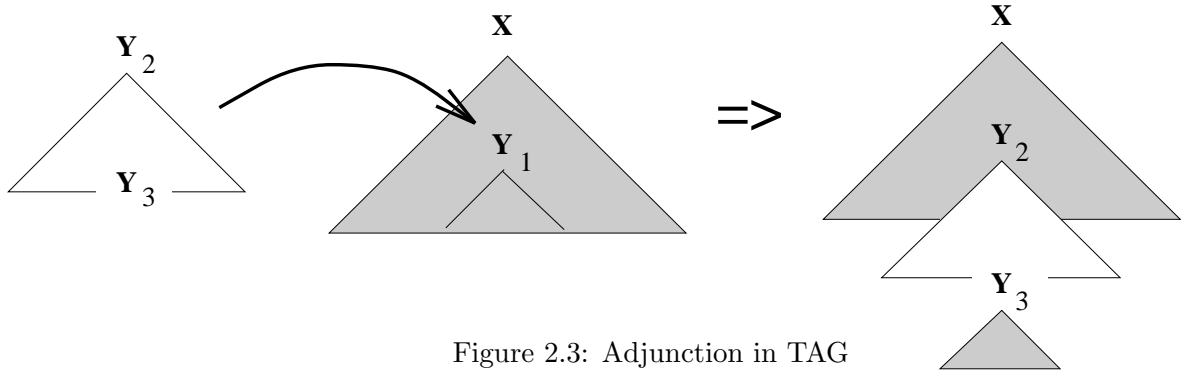,height=1.9in}
\caption{Adjunction in TAG}
\label{proto-adjunction}
\end{figure}

A TAG $G$ is a collection of finite initial trees, $I$, and auxiliary trees,
$A$.  The \xtagdef{tree set} of a TAG $G$, ${\cal T}(G)$ is defined to be the
set of all derived trees starting from S-type initial trees in $I$ whose
frontier consists of terminal nodes (all substitution nodes having been
filled). The \xtagdef{string language} generated by a TAG, ${\cal L}(G)$, is
defined to be the set of all terminal strings on the frontier of the trees in
${\cal T}(G)$.

\section{Lexicalization}

`Lexicalized' grammars systematically associate each elementary structure with
a lexical anchor. This means that in each structure there is a lexical item
that is realized.  It does not mean simply adding feature structures (such as
head) and unification equations to the rules of the formalism.  These resultant
elementary structures specify extended domains of locality (as compared to
CFGs) over which constraints can be stated.

Following \cite{schabes88} we say that a grammar is \xtagdef{lexicalized} if it
consists of 1) a finite set of structures each associated with a lexical item,
and 2) an operation or operations for composing the structures.  Each lexical
item will be called the \xtagdef{anchor} of the corresponding structure, which
defines the domain of locality over which constraints are specified.  Note
then, that constraints are local with respect to their anchor.

Not every grammar is in a lexicalized form.\footnote{Notice the similarity of
the definition of a lexicalized grammar with the off line parsability
constraint (\cite{kaplan83}). As consequences of our definition, each structure
has at least one lexical item (its anchor) attached to it and all sentences are
finitely ambiguous.} In the process of lexicalizing a grammar, the lexicalized
grammar is required to be strongly equivalent to the original grammar, i.e. it
must produce not only the same language, but the same structures or tree set as
well.

\begin{figure*}[htb]
\centering
\begin{tabular}{ccccccc}
{{\psfig{figure=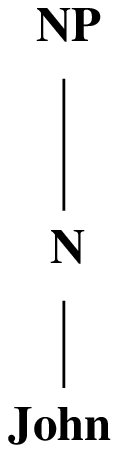,height=1.0in}}\label{fig1a}}  &
\hspace{0.1in} &
{{\psfig{figure=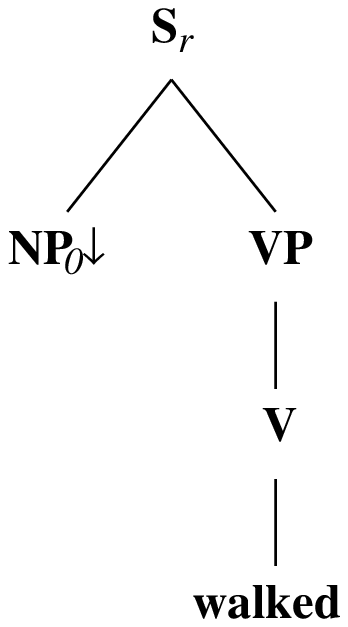,height=1.4in}}\label{fig1b}}  & 
\hspace{0.1in} &
{{\psfig{figure=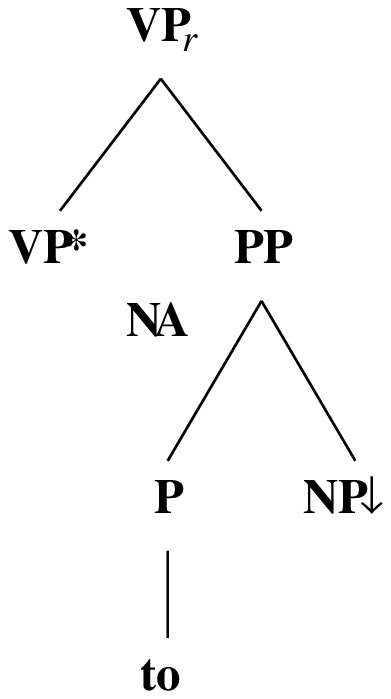,height=1.7in}} \label{fig1c} }  & 
\hspace{0.1in} &
{{\psfig{figure=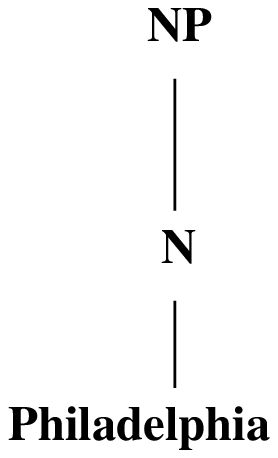,height=1.0in}} \label{fig1d}} \\
(a)&&(b)&&(c)&&(d)\\
\end{tabular}\\
\caption {Lexicalized Elementary trees}
\label {lex-elem-trees}
\end{figure*}

In Figure \ref{lex-elem-trees}, which shows sample initial and auxiliary trees,
substitution sites are marked by a $\downarrow$, and foot nodes are marked by
an $\ast$.  This notation is standard and is followed in the rest of this
report.

\section{Unification-based features}

In a unification framework, a feature structure is associated with each node in
an elementary tree.  This feature structure contains information about how the
node interacts with other nodes in the tree.  It consists of a top part, which
generally contains information relating to the supernode, and a bottom part,
which generally contains information relating to the subnode.  Substitution
nodes, however, have only the top features, since the tree substituting in
logically carries the bottom features.

\begin{figure}[htb]
\centering
\begin{tabular}{c}
\psfig{figure=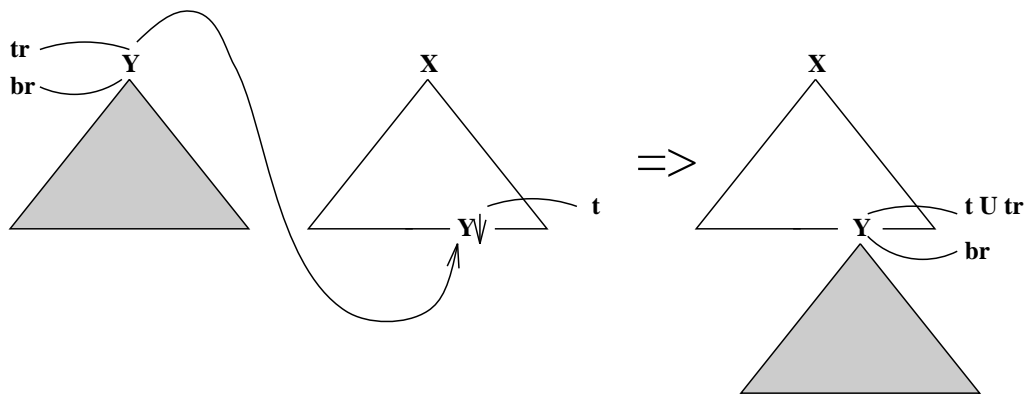,height=2.0in}
\end{tabular}
\caption{Substitution in FB-LTAG}
\label{subst-fig}
\end{figure}

The notions of substitution and adjunction must be augmented to fit
within this new framework.  The feature structure of a new node
created by substitution inherits the union of the features of the
original nodes.  The top feature of the new node is the union of the
top features of the two original nodes, while the bottom feature of
the new node is simply the bottom feature of the top node of the
substituting tree (since the substitution node has no bottom feature).
Figure \ref{subst-fig}\footnote{abbreviations in the figure:
t$=$top feature structure, tr$=$top feature structure of the root, br$=$bottom
feature structure of the root, U$=$unification} shows this more
clearly.

\begin{figure}[htb]
\centering
\begin{tabular}{c}
\hspace{0.65in}
\psfig{figure=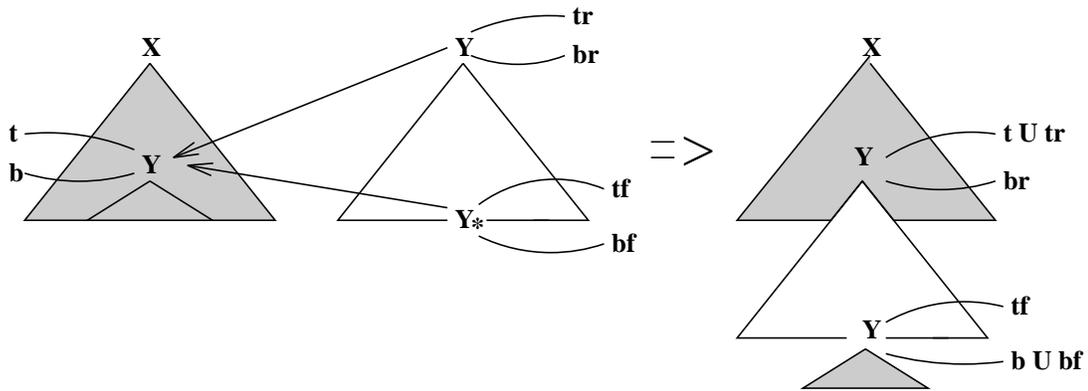,height=2.0in}
\end{tabular}
\caption{Adjunction in FB-LTAG}
\label{adjunct-fig}
\end{figure}

Adjunction is only slightly more complicated.  The node being adjoined into
splits, and its top feature unifies with the top feature of the root
adjoining node, while its bottom feature unifies with the bottom feature of the
foot adjoining node.  Again, this is easier shown graphically, as in Figure
\ref{adjunct-fig}\footnote{abbreviations in the figure: t$=$top
feature structure, b$=$bottom feature structure, tr$=$top feature
structure of the root, br$=$bottom feature structure of the root,
tf$=$top feature structure of the foot, bf$=$bottom feature structure
of the foot, U$=$unification}.

\begin{figure}[htbp]
\centering
\begin{tabular}{ccc}
{\psfig{figure=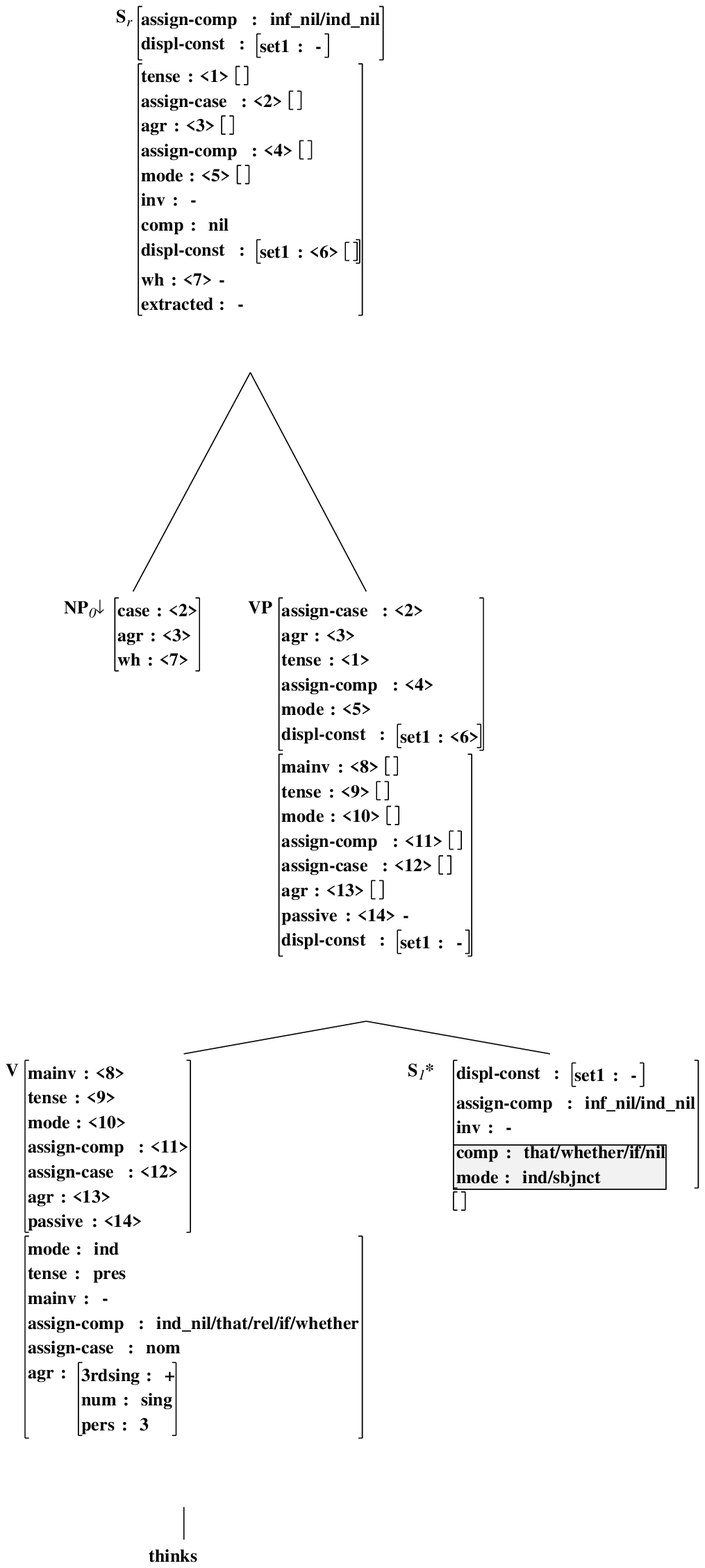,height=6.5in}}  &
\hspace{0.6in}
{\psfig{figure=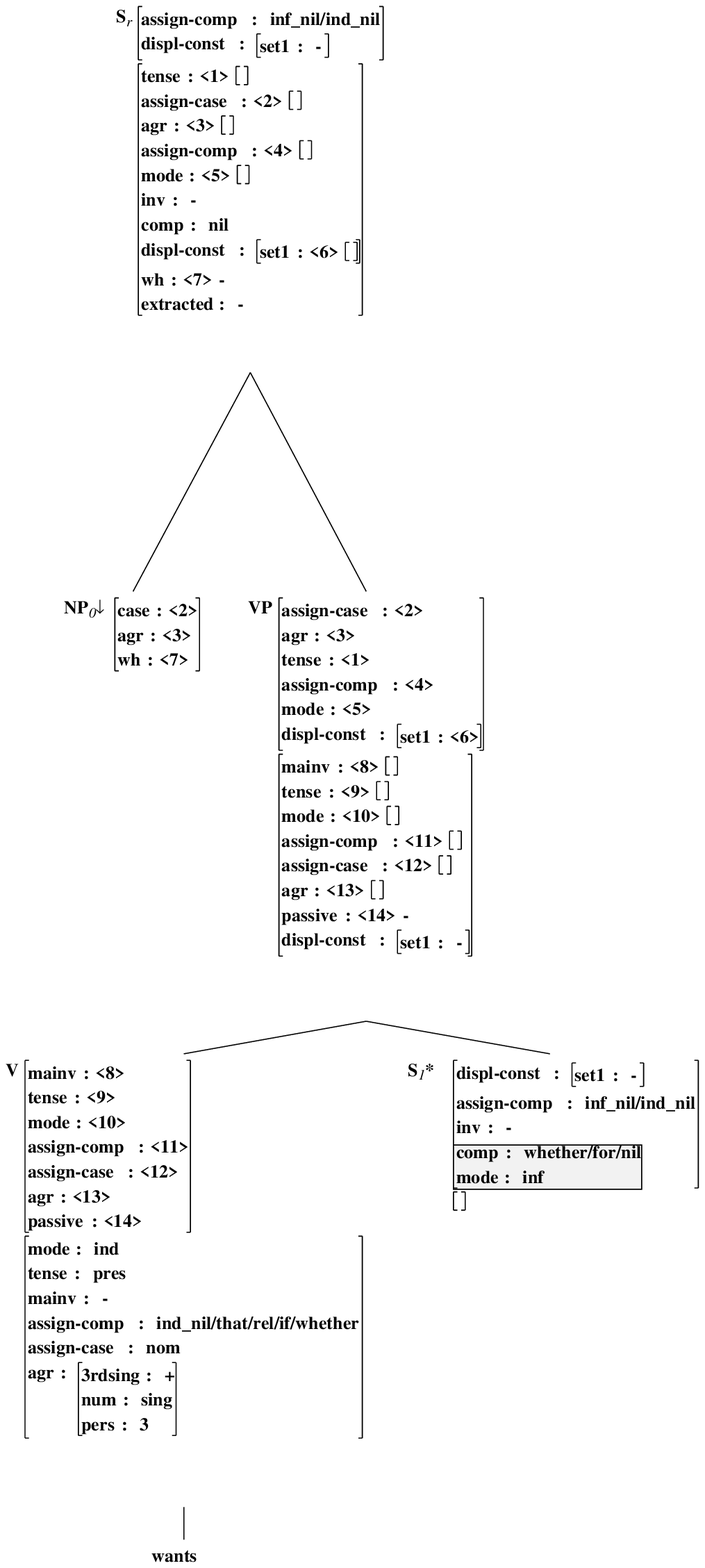,height=6.5in}} \\
{\it think} tree&{\it want} tree\\
\end{tabular}\\
\caption {Lexicalized Elementary Trees with Features}
\label {lex-with-features}
\label{2;Tnx0Vs1}
\end{figure}

The embedding of the TAG formalism in a unification framework allows
us to dynamically specify local constraints that would have otherwise
had to have been made statically within the trees.  Constraints that
verbs make on their complements, for instance, can be implemented
through the feature structures.  The notions of Obligatory and
Selective Adjunction, crucial to the formation of lexicalized
grammars, can also be handled through the use of
features.\footnote{The remaining constraint, Null Adjunction (NA),
must still be specified directly on a node.} Perhaps more important to
developing a grammar, though, is that the trees can serve as a
schemata to be instantiated with lexical-specific features when an
anchor is associated with the tree.  To illustrate this, Figure
\ref{lex-with-features} shows the same tree lexicalized with two
different verbs, each of which instantiates the features of the tree
according to its lexical selectional restrictions.

In Figure \ref{lex-with-features}, the lexical item {\it thinks} takes an
indicative sentential complement, as in the sentence {\it John thinks that Mary
loves Sally}.  {\it Want} takes a sentential complement as well, but an
infinitive one, as in {\it John wants to love Mary}.  This distinction is
easily captured in the features and passed to other nodes to constrain which
trees this tree can adjoin into, both cutting down the number of separate trees
needed and enforcing conceptual Selective Adjunctions (SA).

\chapter{Overview of the XTAG System}
\label{overview}

This section focuses on the various components that comprise the
parser and English grammar in the XTAG system.  Persons interested
only in the linguistic analyses in the grammar may skip this section
without loss of continuity, although a quick glance at the tagset used
in XTAG and the set of non-terminal labels used will be useful. We may
occasionally refer back to the various components mentioned in this
section.

\section{System Description}

Figure~{\ref{flowchart}} shows the overall flow of the system when
parsing a sentence; a summary of each component is presented in
Table~\ref{sys-table}. At the heart of the system is a parser for
lexicalized TAGs (\cite{schabesjoshi88,schabes90}) which produces all
legitimate parses for the sentence. The parser has two phases: {\bf
Tree Selection} and {\bf Tree Grafting}.

\begin{figure}[t]
\hspace{0.35in}
\centering
{\psfig{figure=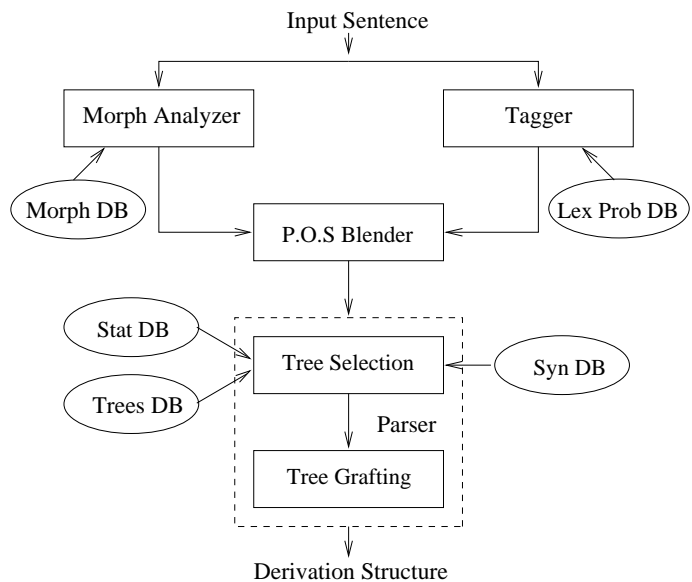,height=3.0in,angle=270}}
\caption[XTAG system diagram]{Overview of XTAG system }
\label{flowchart}
\end{figure}

\begin{table}[ht]
\small
\centering
\begin{tabular}{|l|l|} \hline
Component & Details \\ \hline
Morphological & Consists of approximately 317,000 inflected items \\ 
Analyzer and & derived from over 90000 stems. \\ 
Morph Database & Entries are indexed on the inflected form and return \\
& the root form, POS, and inflectional information.\\ \hline
POS Tagger & Wall Street Journal-trained
trigram tagger (\cite{kwc88})  \\ 
and  Lex Prob & extended to output N-best POS sequences  \\
Database & (\cite{soong90}). Decreases the time to parse \\
&a sentence by an average of 93\%. \\\hline
Syntactic &  More than 30,000 entries. \\
Database & Each entry consists of: the uninflected form of the word, \\
& its POS, the list of trees or tree-families associated with \\
& the word, and a list of feature equations that capture \\
&lexical idiosyncrasies. \\ \hline
Tree Database &  1094 trees, divided into 52 tree families and 218 individual \\
& trees. Tree families represent subcategorization frames; \\
& the trees in a tree family would be related to each other \\

& transformationally in a movement-based approach. \\ \hline
X-Interface & Menu-based facility for creating and modifying tree files. \\
&  User controlled parser parameters: parser's start category, \\ 
& enable/disable/retry on failure for POS tagger. \\
& Storage/retrieval facilities for elementary and parsed trees.\\
& Graphical displays of tree and feature data structures. \\
& Hand combination of trees by adjunction or substitution \\
& for grammar development. \\ 
& Ability to manually assign POS tag \\
& and/or Supertag before parsing \\ \hline
\end{tabular}
\caption{System Summary}
\label{sys-table}
\end{table}

\subsection{Tree Selection}

Since we are working with lexicalized TAGs, each word in the sentence
selects at least one tree. The advantage of a lexicalized formalism
like LTAGs is that rather than parsing with all the trees in the
grammar, we can parse with only the trees selected by the words in the
input sentence.

In the XTAG system, the selection of trees by the words is done in
several steps. Each step attempts to reduce ambiguity, i.e. reduce the
number of trees selected by the words in the sentence.

\begin{description}
\item[Morphological Analysis and POS Tagging] The input sentence is
  first submitted to the {\bf Morphological Analyzer} and the {\bf
    Tagger}. The morphological analyzer~(\cite{karp92}) consists of a
  disk-based database (a compiled version of the derivational rules)
  which is used to map an inflected word into its stem, part of speech
  and feature equations corresponding to inflectional information.
  These features are inserted at the anchor node of the tree
  eventually selected by the stem. The POS Tagger can be disabled in
  which case only information from the morphological analyzer is used.
  The morphology data was originally extracted from the Collins
  English Dictionary (\cite{ced79}) and Oxford Advanced Learner's
  Dictionary (\cite{oald74}) available through ACL-DCI
  (\cite{liberman89}), and then cleaned up and augmented by hand
  (\cite{karp92}).
    
\item[POS Blender] The output from the morphological analyzer and the
  POS tagger go into the {\bf POS Blender} which uses the output of
  the POS tagger as a filter on the output of the morphological
  analyzer. Any words that are not found in the morphological database
  are assigned the POS given by the tagger.
  
\item[Syntactic Database] The syntactic database contains the mapping
  between particular stem(s) and the tree templates or tree-families
  stored in the {\bf Tree Database} (see Table~\ref{sys-table}). The
  syntactic database also contains a list of feature equations that
  capture lexical idiosyncrasies. The output of the POS Blender is
  used to search the {\bf Syntactic Database} to produce a set of
  lexicalized trees with the feature equations associated with the
  word(s) in the syntactic database unified with the feature equations
  associated with the trees. Note that the features in the syntactic
  database can be assigned to any node in the tree and not just to the
  anchor node. The syntactic database entries were originally
  extracted from the Oxford Advanced Learner's Dictionary
  (\cite{oald74}) and Oxford Dictionary for Contemporary Idiomatic
  English (\cite{cie75}) available through ACL-DCI
  (\cite{liberman89}), and then modified and augmented by hand
  (\cite{EgediMartin94}).  There are more than 31,000 syntactic
  database entries.\footnote{This number does not include trees
    assigned by default based on the part-of-speech of the word.}
  Selected entries from this database are shown in
  Table~\ref{syn-entries}.
    
\item[Default Assignment] For words that are not found in the
  syntactic database, default trees and tree-families are assigned
  based on their POS tag.
  
\item[Filters] Some of the lexicalized trees chosen in previous stages
  can be eliminated in order to reduce ambiguity. Two methods are
  currently used: structural filters which eliminate trees which have
  impossible spans over the input sentence and a statistical filter
  based on unigram probabilities of non-lexicalized trees (from a hand
  corrected set of approximately 6000 parsed sentences). These methods
  speed the runtime by approximately 87\%.
  
\item[Supertagging] Before parsing, one can avail of an optional step
  of {\em supertagging} the sentence. This step uses statistical
  disambiguation to assign a unique elementary tree (or {\em
    supertag}) to each word in the sentence. These assignments can
  then be hand-corrected. These supertags are used as a filter on the
  tree assignments made so far. More information on supertagging can
  be found in (\cite{srini97diss,srini97iwpt}).

\end{description}

\begin{table}[htb]
\begin{verbatim}
<<INDEX>>porousness<<ENTRY>>porousness<<POS>>N
<<TREES>>^BNXN ^BN ^CNn
<<FEATURES>>#N_card- #N_const- #N_decreas- #N_definite- #N_gen- 
#N_quan- #N_refl-

<<INDEX>>coo<<ENTRY>>coo<<POS>>V<<FAMILY>>Tnx0V

<<INDEX>>engross<<ENTRY>>engross<<POS>>V<<FAMILY>>Tnx0Vnx1
<<FEATURES>>#TRANS+

<<INDEX>>forbear<<ENTRY>>forbear<<POS>>V<<FAMILY>>Tnx0Vs1
<<FEATURES>>#S1_WH- #S1_inf_for_nil

<<INDEX>>have<<ENTRY>>have<<POS>>V<<ENTRY>>out<<POS>>PL
<<FAMILY>>Tnx0Vplnx1
\end{verbatim}   
\caption{Example Syntactic Database Entries.}

\label{syn-entries}
\end{table}

\subsection{Tree Database}
\label{tree-db}

The {\bf Tree Database} contains the tree templates that are
lexicalized by following the various steps given above. The lexical
items are inserted into distinguished nodes in the tree template
called the {\em anchor nodes}.  The part of speech of each word in the
sentence corresponds to the label of the anchor node of the trees.
Hence the tagset used by the POS Tagger corresponds exactly to the
labels of the anchor nodes in the trees.  The tagset used in the XTAG
system is given in Table~\ref{tagset}. The tree templates are
subdivided into tree families (for verbs and other predicates), and
tree files which are simply collections of trees for lexical items
like prepositions, determiners, etc%
\footnote{ The nonterminals in the tree database are {\tt A, AP, Ad,
    AdvP, Comp, Conj, D, N, NP, P, PP, Punct, S, V, VP}.}%
.

\subsection{Tree Grafting}

Once a particular set of lexicalized trees for the sentence have been
selected, XTAG uses an Earley-style predictive left-to-right parsing
algorithm for LTAGs (\cite{schabesjoshi88,schabes90}) to find all
derivations for the sentence. The derivation trees and the associated
derived trees can be viewed using the X-interface (see
Table~\ref{sys-table}). The X-interface can also be used to save
particular derivations to disk.

The output of the parser for the sentence {\it I had a map yesterday} is
illustrated in Figure~\ref{sentence}. The parse tree\footnote{The feature
structures associated with each note of the parse tree are not shown here.}
represents the surface constituent structure, while the derivation tree
represents the derivation history of the parse. The nodes of the derivation
tree are the tree names anchored by the lexical items\footnote{Appendix
\ref{tree-naming} explains the conventions used in naming the trees.}.  The
composition operation is indicated by the nature of the arcs: a dashed line is
used for substitution and a bold line for adjunction.  The number beside each
tree name is the address of the node at which the operation took place.  The
derivation tree can also be interpreted as a dependency graph with unlabeled
arcs between words of the sentence.

\begin{figure}[htb]
\centering
\begin{tabular}{cc}
{{\psfig{figure=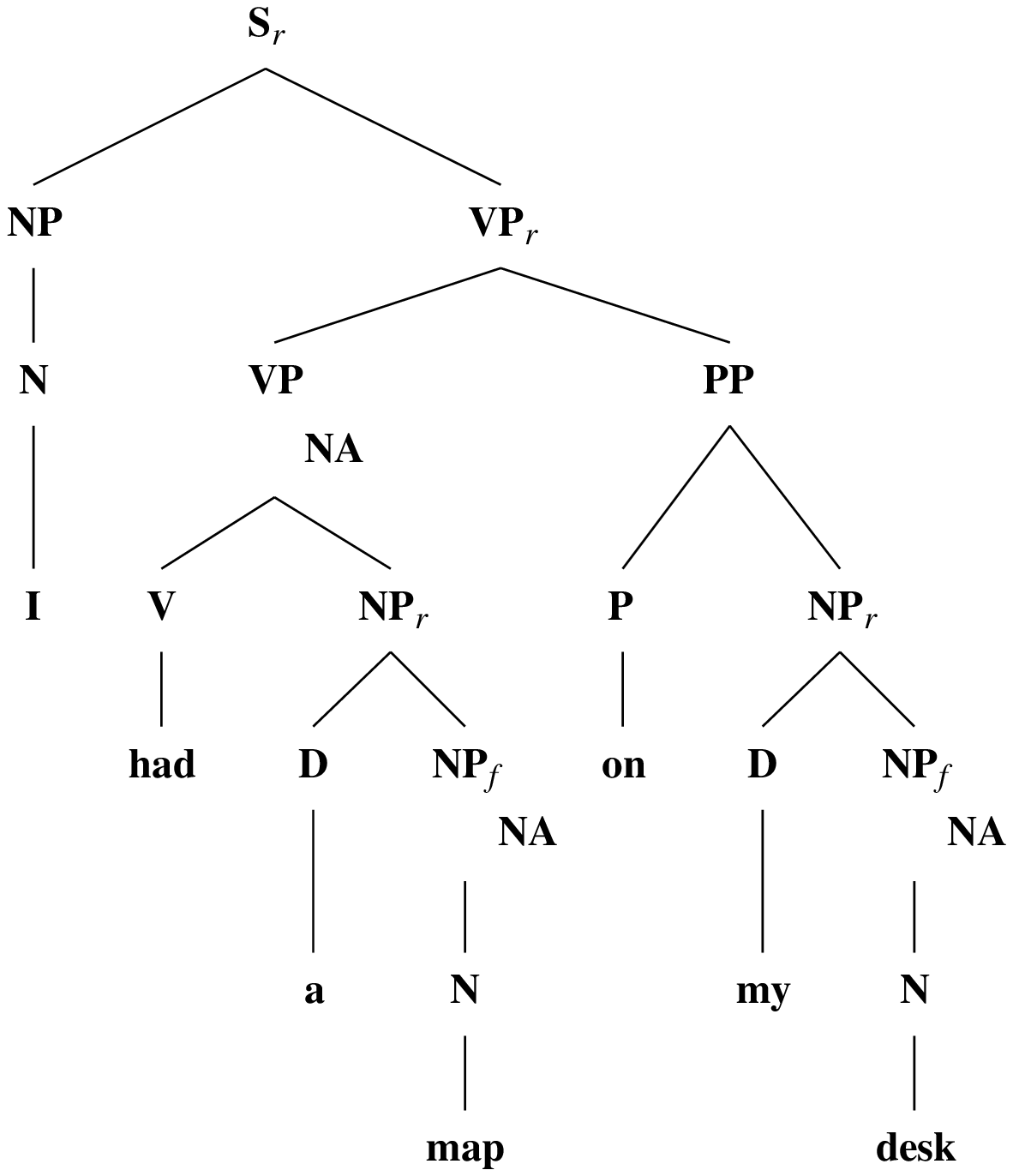,height=3.0in}}}  &
{{\psfig{figure=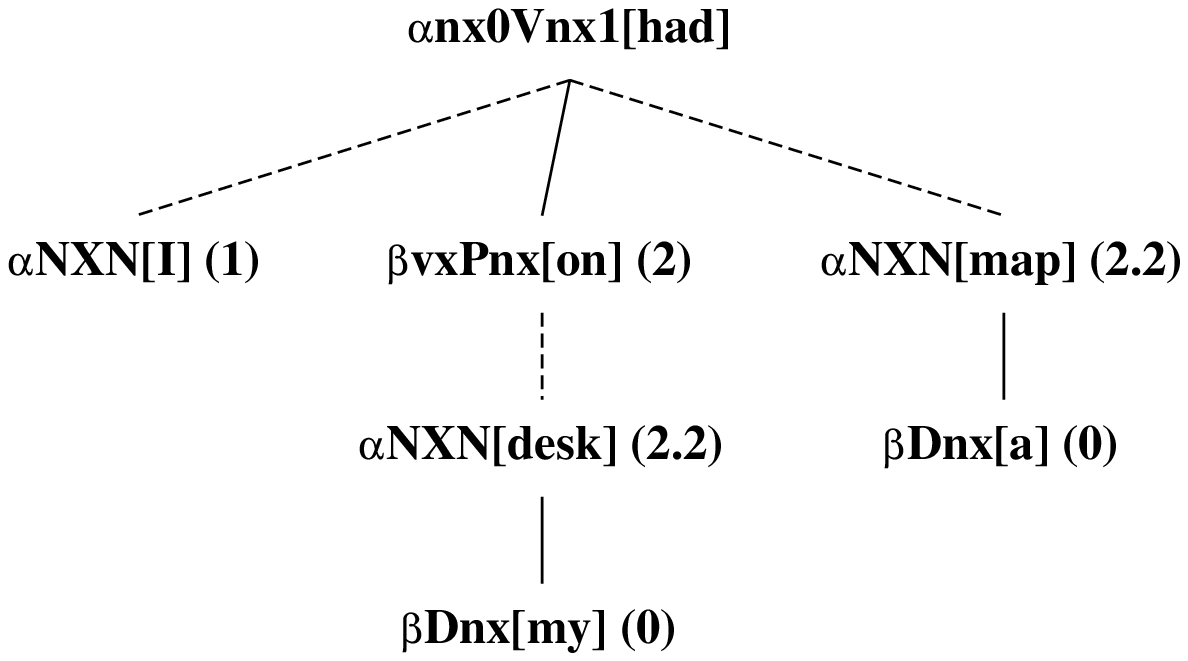,height=2.0in,width=2.7in}}} \\
Parse Tree  & Derivation Tree \\
\end{tabular}
\caption{Output Structures from the Parser}
\label{sentence}
\end{figure}

\begin{table*}[ht]
\small
\centering
\begin{tabular}{|l|l|} \hline
Part of Speech & Description \\ \hline
A & Adjective \\ \hline
Ad & Adverb \\ \hline
Comp & Complementizer \\ \hline
D & Determiner \\ \hline
G & Genitive Noun \\ \hline
I & Interjection \\ \hline
N & Noun \\ \hline
P & Preposition \\ \hline
PL & Particle \\ \hline
Punct & Punctuation \\ \hline
V & Verb \\ \hline
\end{tabular}
\caption{XTAG tagset}
\label{tagset}
\end{table*}

\subsection{The Grammar Development Environment}

Working with and developing a large grammar is a challenging process,
and the importance of having good visualization tools cannot be
over-emphasized. Currently the XTAG system has X-windows based tools
for viewing and updating the morphological and syntactic databases
(\cite{karp92,EgediMartin94}). These are available in both ASCII and
binary-encoded database format. The ASCII format is well-suited for
various UNIX utilities (awk, sed, grep) while the database format is
used for fast access during program execution.  However even the ASCII
formatted representation is not well-suited for human readability. An
X-windows interface for the databases allows users to easily examine
them.  Searching for specific information on certain fields of the
syntactic database is also available. Also, the interface allows a
user to insert, delete and update any information in the databases.
Figure~\ref{morphsyn-tool}(a) shows the interface for the morphology
database and Figure~\ref{morphsyn-tool}(b) shows the interface for the
syntactic database.

\begin{figure}[htb]
\begin{tabular}{cc}
{\psfig{figure=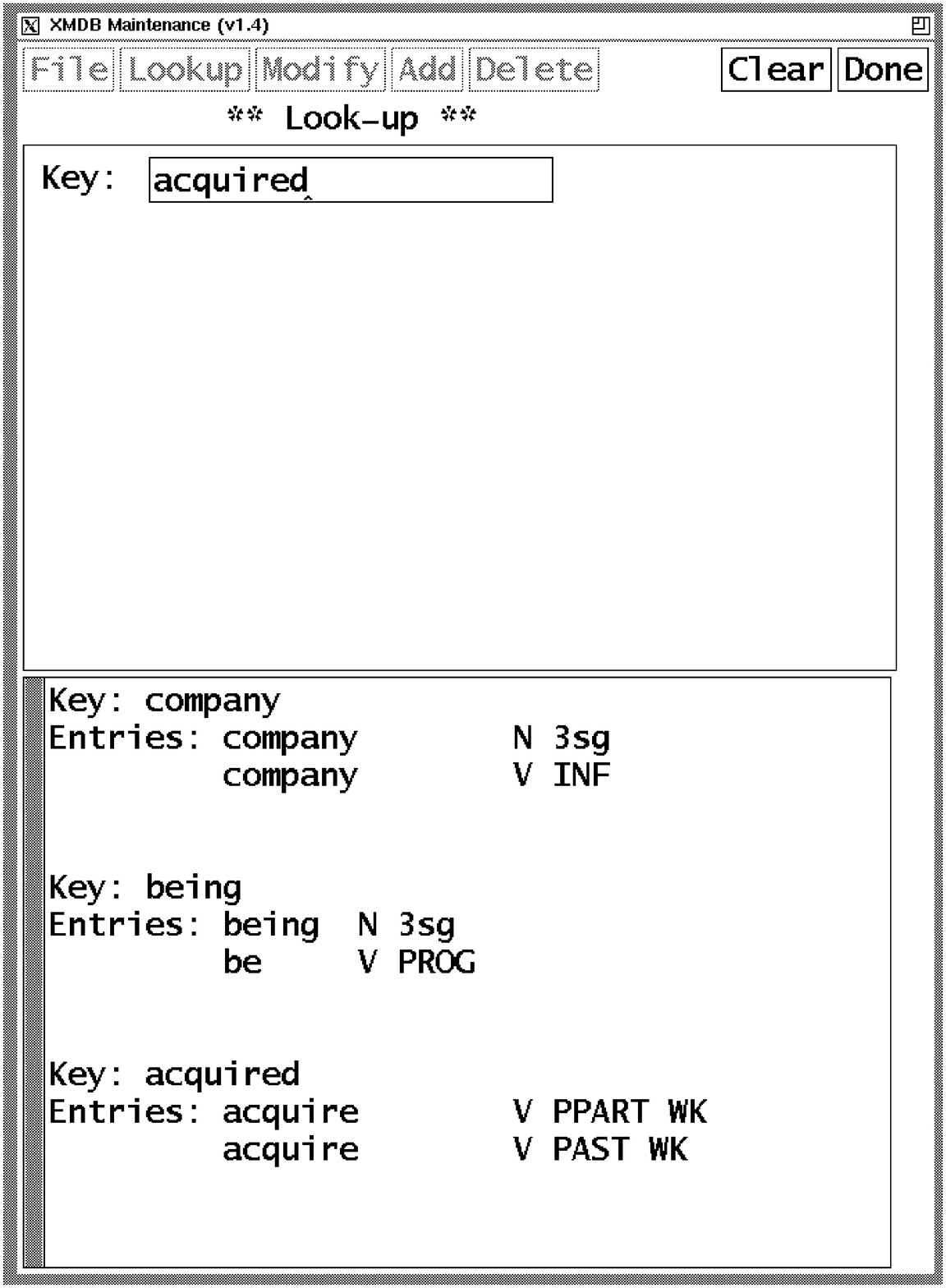,height=3.0in}}&
{\psfig{figure=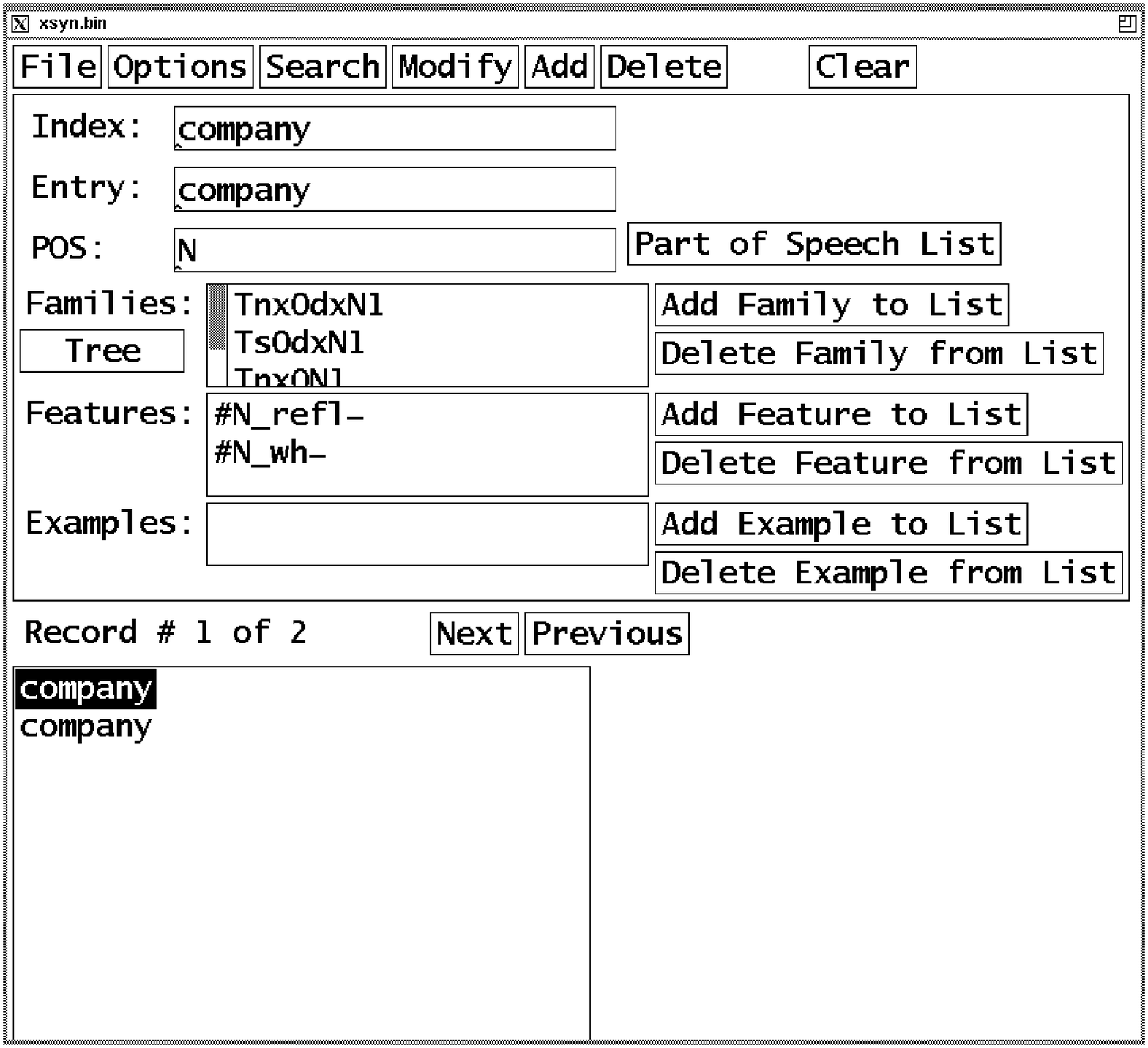,height=3.0in,width=2.0in}}\\
(a) Morphology database&(b) Syntactic database
\end{tabular}
\caption[Interfaces database]{Interfaces to the database maintenance tools}
\label{morphsyn-tool}
\end{figure}

XTAG also has a parsing and grammar development interface
(\cite{PSJ92}). This interface includes a tree editor, the ability to
vary parameters in the parser, work with multiple grammars and/or
parsers, and use metarules for more efficient tree editing and
construction (\cite{becker94}). The interface is shown in
Figure~\ref{xtag-interface}. It has the following features:

\begin{itemize}

\item Menu-based facility for creating and modifying tree files and 
loading grammar files.

\item User controlled parser parameters, including the root
category (main S, embedded S, NP, etc.), and the use of the tagger
(on/off/retry on failure).

\item Storage/retrieval facilities for elementary and parsed trees.

\item The production of postscript files corresponding to elementary
and parsed trees.

\item Graphical displays of tree and feature data structures,
including a scroll `web' for large tree structures.

\item Mouse-based tree editor for creating and modifying trees and
feature structures.

\item Hand combination of trees by adjunction or substitution for use
in diagnosing grammar problems.

\item Metarule tool for automatic aid to the generation of trees by using 
tree-based transformation rules
 
\end{itemize}

\begin{figure}[htb]
\centering
\mbox{}
{\psfig{figure=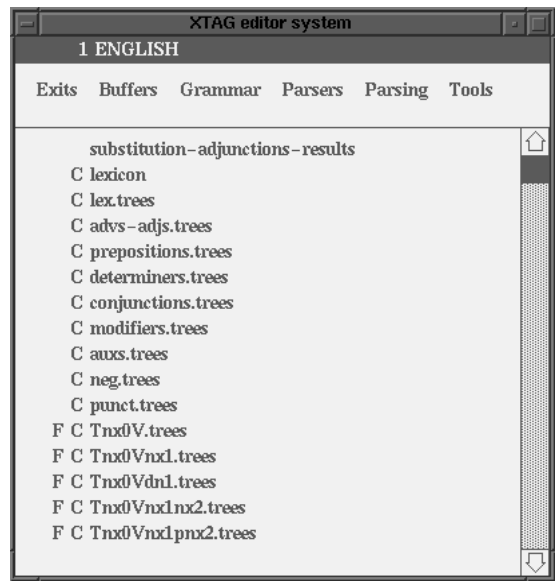,height=3.0in}}
\caption[XTAG Interface]{Interface to the XTAG system}
\label{xtag-interface}
\end{figure}

\section{Computer Platform}

XTAG was developed on the Sun SPARC station series. It has been tested
on various Sun platforms including Ultra-1, Ultra-Enterprise. XTAG is
freely available from the XTAG web page at {\tt
  http://www.cis.upenn.edu/\~{}xtag/}. It requires 75 MB of disk space
(once all binaries and databases are created after the install). XTAG
requires the following software to run:

\begin{itemize}
  
\item A machine running UNIX and X11R4 (or higher). Previous releases
  of X will not work. X11R4 is free software which usually comes
  bundled with your OS. It is also freely available for various
  platforms at {\tt http://www.xfree86.org/}
  
\item A Common Lisp compiler which supports the latest definition of
  Common Lisp (Steele's Common Lisp, second edition). XTAG has been
  tested on Lucid Common Lisp/SPARC Solaris, Version: 4.2.1. Allegro
  CL is no longer directly supported, however there have been third
  party ports to recent versions of Allegro CL.
  
\item CLX version 4 or higher. CLX is the Lisp equivalent to the Xlib
  package written in C.
  
\item Mark Kantrowitz's Lisp Utilities from CMU: logical-pathnames and
  defsystem.

\end{itemize}

A patched version of CLX (Version 5.02) for SunOS 5.5.1 and the CMU
Lisp Utilities are provided in our ftp directory for your convenience.
However, we ask that you refer to the appropriate sources for updates.

The morphology database component (\cite{karp92}), no longer under
licensing restrictions, is available as a separate download from the
XTAG web page (see above for URL).

The syntactic database component is also available as part of the XTAG
system (\cite{EgediMartin94}). 

More information can be obtained on the XTAG web page at \\
{\tt http://www.cis.upenn.edu/\~{}xtag/}.

\chapter{Underview}
\label{underview}

The morphology, syntactic, and tree databases together comprise the
English grammar.  A lexical item that is not in the databases receives
a default tree selection and features for its part of speech and
morphology.  In designing the grammar, a decision was made early on to
err on the side of acceptance whenever there are conflicting opinions
as to whether or not a construction is grammatical.  In this sense,
the XTAG English grammar is intended to function primarily as an
acceptor rather than a generator of English sentences.  The range of
syntactic phenomena that can be handled is large and includes
auxiliaries (including inversion), copula, raising and small clause
constructions, topicalization, relative clauses, infinitives, gerunds,
passives, adjuncts, it-clefts, wh-clefts, PRO constructions, noun-noun
modifications, extraposition, determiner sequences, genitives,
negation, noun-verb contractions, clausal adjuncts and imperatives.

\section{Subcategorization Frames}
\label{subcat-frames}

Elementary trees for non-auxiliary verbs are used to represent the linguistic
notion of subcategorization frames.  The anchor of the elementary tree
subcategorizes for the other elements that appear in the tree, forming a
clausal or sentential structure.  Tree families group together trees belonging
to the same subcategorization frame.  Consider the following uses of the verb
{\it buy}:

\enumsentence{Srini bought a book.}
\enumsentence{Srini bought Beth a book.}

In sentence (\ex{-1}), the verb {\it buy} subcategorizes for a direct
object NP.  The elementary tree anchored by {\it buy} is shown in
Figure~\ref{subcat-trees}(a) and includes nodes for the NP complement
of {\it buy} and for the NP subject.  In addition to this declarative
tree structure, the tree family also contains the trees that would be
related to each other transformationally in a movement based approach,
i.e passivization, imperatives, wh-questions, relative clauses, and so
forth.  Sentence (\ex{0}) shows that {\it buy} also subcategorizes for
a double NP object.  This means that {\it buy} also selects the double
NP object subcategorization frame, or tree family, with its own set of
transformationally related sentence structures.
Figure~\ref{subcat-trees}(b) shows the declarative structure for this
set of sentence structures.

\begin{figure}[ht]
\centering
\begin{tabular}{ccc}
{\psfig{figure=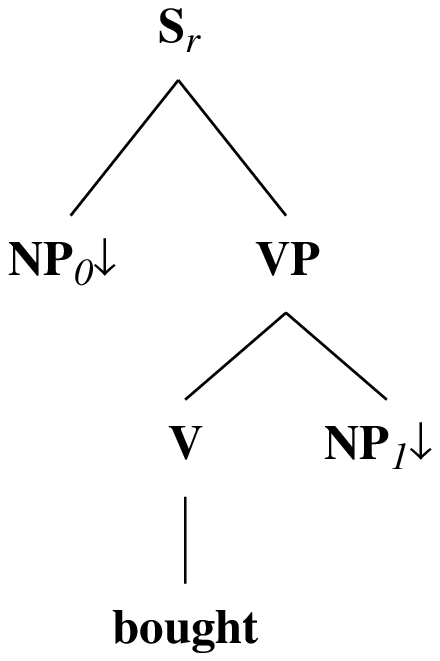,height=1.8in}} & 
\hspace*{0.5in} & 
{\psfig{figure=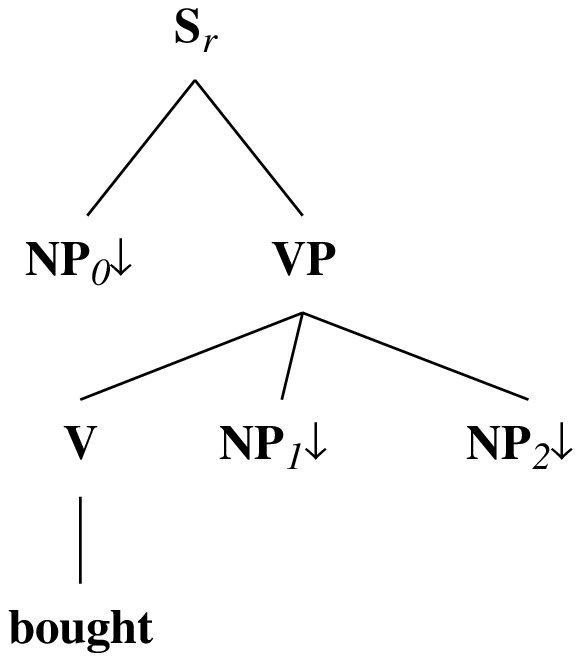,height=1.8in}}\\
(a) & \hspace*{0.5in} & (b) \\ 
\end{tabular}
\caption{Different subcategorization frames for the verb {\it buy}}
\label{subcat-trees}
\end{figure}

\section{Complements and Adjuncts}
\label{compl-adj}

Complements and adjuncts have very different structures in the XTAG grammar.
Complements are included in the elementary tree anchored by the verb that
selects them, while adjuncts do not originate in the same elementary tree as
the verb anchoring the sentence, but are instead added to a structure by
adjunction.  The contrasts between complements and adjuncts have been
extensively discussed in the linguistics literature and the classification of a
given element as one or the other remains a matter of debate (see
\cite{rizzi90},
\cite{larson88}, \cite{jackendoff90}, \cite{larson90}, \cite{cinque90}, 
\cite{obernauer84}, \cite{lasnik-saito84}, and \cite{chomsky86}).  The guiding
rule used in developing the XTAG grammar is whether or not the sentence is
ungrammatical without the questioned structure.\footnote{Iteration of a
structure can also be used as a diagnostic: {\it Srini bought a book at the
bookstore on Walnut Street for a friend}.} Consider the following
sentences:

\enumsentence{Srini bought a book.}
\enumsentence{Srini bought a book at the bookstore.}
\enumsentence{Srini arranged for a ride.}
\enumsentence{$\ast$Srini arranged.}

Prepositional phrases frequently occur as adjuncts, and when they are
used as adjuncts they have a tree structure such as that shown in
Figure~\ref{compl-adjunct}(a).  This adjunction tree would adjoin into
the tree shown in Figure~\ref{subcat-trees}(a) to generate sentence
(\ex{-2}).  There are verbs, however, such as {\it arrange}, {\it
hunger} and {\it differentiate}, that take prepositional phrases as
complements.  Sentences (\ex{-1}) and (\ex{0}) clearly show that the
prepositional phrase are not optional for {\it arrange}.  For these
sentences, the prepositional phrase will be an initial tree (as shown
in Figure~\ref{compl-adjunct}(b)) that substitutes into an elementary
tree, such as the one anchored by the verb {\it arrange} in
Figure~\ref{compl-adjunct}(c).

\begin{figure}[ht]
\centering
\begin{tabular}{ccccc}
{\psfig{figure=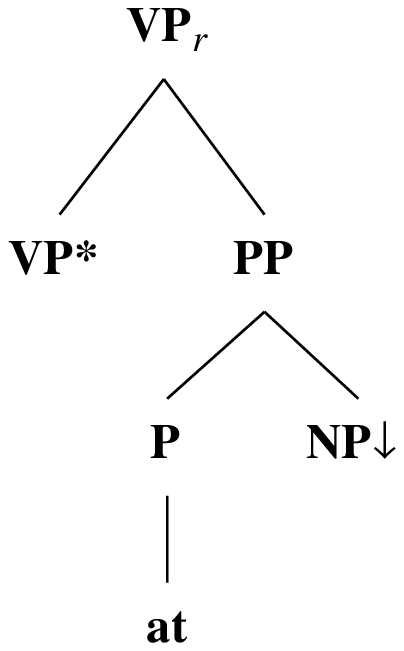,height=1.8in}} &
\hspace*{0.5in} &
{\psfig{figure=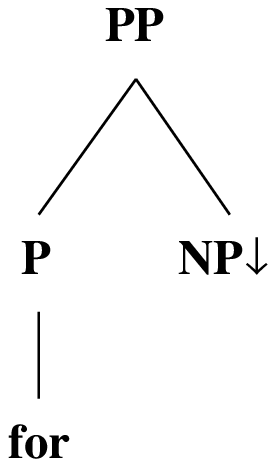,height=1.3in}} &
\hspace*{0.5in} & 
{\psfig{figure=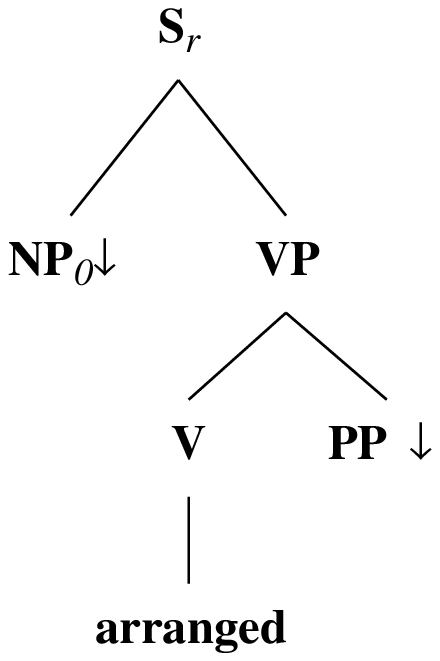,height=1.8in}}\\
(a) & \hspace*{0.5in} & (b) & \hspace*{0.5in} & (c) \\ 
\end{tabular}
\caption{Trees illustrating the difference between Complements and Adjuncts}
\label{compl-adjunct}
\label{2;1,9}
\end{figure}

Virtually all parts of speech, except for main verbs, function as both
complements and adjuncts in the grammar.  More information is available in this
report on various parts of speech as complements: adjectives (e.g. section
\ref{nx0Vax1-family}), nouns (e.g.  section~\ref{nx0Vnx1-family}), and
prepositions (e.g. section~\ref{nx0Vpnx1-family}); and as adjuncts: adjectives
(section~\ref{adj-modifier}), adverbs (section~\ref{adv-modifier}), nouns
(section~\ref{noun-modifier}), and prepositions (section~\ref{prep-modifier}).

\section{Non-S constituents}

Although sentential trees are generally considered to be special cases in any
grammar, insofar as they make up a `starting category', it is the case that any
initial tree constitutes a phrasal constituent.  These initial trees may have
substitution nodes that need to be filled (by other initial trees), and may be
modified by adjunct trees, exactly as the trees rooted in S.  Although grouping
is possible according to the heads or anchors of these trees, we have not found
any classification similar to the subcategorization frames for verbs that can
be used by a lexical entry to `group select' a set of trees.  These trees are
selected one by one by each lexical item, according to each lexical item's
idiosyncrasies.  The grammar described by this technical report places them
into several files for ease of use, but these files do not constitute tree
families in the way that the subcategorization frames do.

\section{Case Assignment}
\label{case-assignment}
\subsection{Approaches to Case}
\subsubsection{Case in GB theory}

GB (Government and Binding) theory proposes the following
`case filter' as a requirement on S-structure.\footnote{There are certain
problems with applying the case filter as a requirement at the level of
S-structure.  These issues are not crucial to the discussion of the English
XTAG implementation of case and so will not be discussed here.  Interested
readers are referred to
\cite{lasnik-uriagereka88}.}

\begin{verse}
\xtagdef{Case Filter}
Every overt NP must be assigned abstract case. \cite{haegeman91}
\end{verse}

Abstract case is taken to be universal.  Languages with rich morphological case
marking, such as Latin, and languages with very limited morphological case
marking, like English, are all presumed to have full systems of abstract case
that differ only in the extent of morphological realization.

In GB, abstract case is argued to be assigned to NP's by various case
assigners, namely verbs, prepositions, and INFL.  Verbs and
prepositions are said to assign accusative case to NP's that they
govern, and INFL assigns nominative case to NP's that it governs.
These governing categories are constrained as to where they can assign
case by means of `barriers' based on `minimality conditions', although
these are relaxed in `exceptional case marking' situations.  The
details of the GB analysis are beyond the scope of this technical
report, but see \cite{chomsky86} for the original analysis or
\cite{haegeman91} for an overview.  Let it suffice for us to say that
the notion of abstract case and the case filter are useful in
accounting for a number of phenomena including the distribution of
nominative and accusative case, and the distribution of overt NP's and
empty categories (such as PRO).

\subsubsection{Minimalism and Case} 

A major conceptual difference between GB theories and Minimalism is that in
Minimalism, lexical items carry their features with them rather than being
assigned their features based on the nodes that they end up at.  For nouns,
this means that they carry case with them, and that their case is `checked'
when they are in SPEC position of AGR$_s$ or AGR$_o$, which subsequently
disappears \cite{chomsky92}.

\subsection{Case in XTAG}

The English XTAG grammar adopts the notion of case and the case filter for many
of the same reasons argued in the GB literature.  However, in some respects the
English XTAG grammar's implementation of case more closely resembles the
treatment in Chomsky's Minimalism framework \cite{chomsky92} than the system
outlined in the GB literature \cite{chomsky86}.  As in Minimalism, nouns in
the XTAG grammar carry case with them, which is eventually `checked'. However
in the XTAG grammar, noun cases are checked against the case values assigned
by the verb during the unification of the feature structures.  Unlike Chomsky's
Minimalism, there are no separate AGR nodes; the case checking comes from the
verbs directly. Case assignment from the verb is more like the GB approach than
the requirement of a SPEC-head relationship in Minimalism.

Most nouns in English do not have separate forms for nominative and accusative
case, and so they are ambiguous between the two.  Pronouns, of course, are
morphologically marked for case, and each carries the appropriate case in its
feature.  Figures~\ref{nouns-with-case}(a) and \ref{nouns-with-case}(b) show
the NP tree anchored by a noun and a pronoun, respectively, along with the
feature values associated with each word.  Note that {\it books} simply gets
the default case {\bf nom/acc}, while {\it she} restricts the case to be {\bf
nom}.

\begin{figure}[htb]
\centering
\begin{tabular}{ccc}
{\psfig{figure=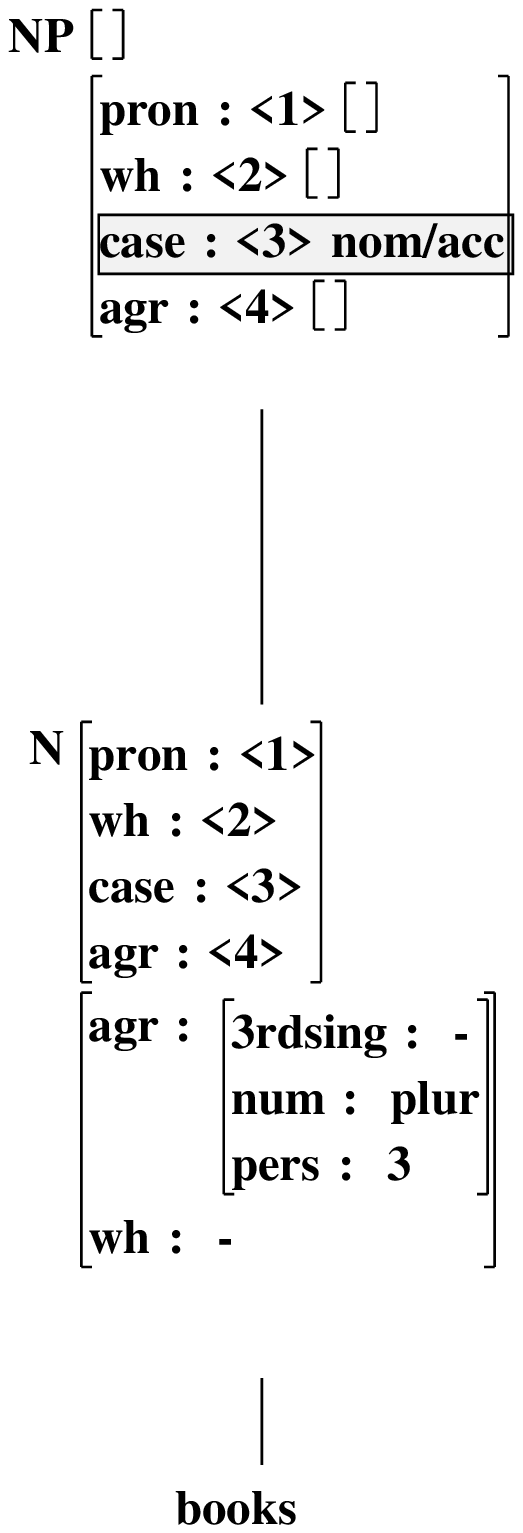,height=3.0in}}  &
\hspace*{0.5in} &
{\psfig{figure=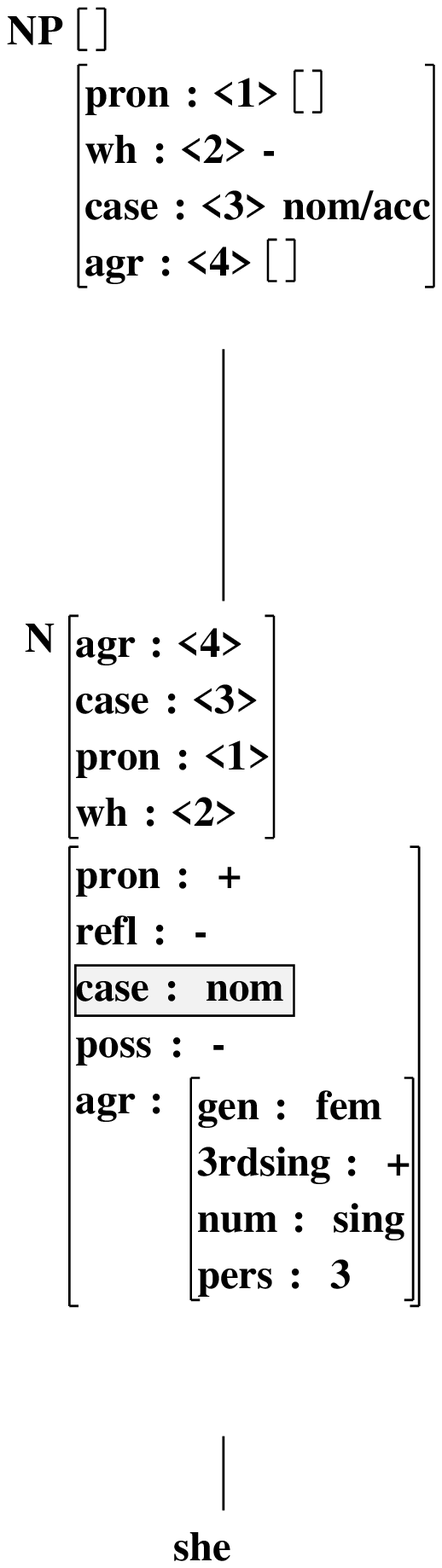,height=3.2in}} \\
(a)& \hspace*{0.5in}&(b)\\
\end{tabular}\\
\caption{Lexicalized NP trees with case markings}
\label {nouns-with-case}
\end{figure}

\subsection{Case Assigners}

\subsubsection{Prepositions}
\label{prep-case}

Case is assigned in the XTAG English grammar by two lexical categories
- verbs and prepositions.\footnote{{\it For} also assigns case as a
complementizer.  See section \ref{for-complementizer} for more
details.}  Prepositions assign accusative case ({\bf acc}) through
their {\bf $<$assign-case$>$} feature, which is linked directly to the
{\bf $<$case$>$} feature of their objects.
Figure~\ref{PXPnx-with-case}(a) shows a lexicalized preposition tree,
while Figure~\ref{PXPnx-with-case}(b) shows the same tree with the NP
tree from Figure~\ref{nouns-with-case}(a) substituted into the NP
position.  Figure~\ref{PXPnx-with-case}(c) is the tree in
Figure~\ref{PXPnx-with-case}(b) after unification has taken place.
Note that the case ambiguity of {\it books} has been resolved to
accusative case.

\begin{figure}[htb]
\centering
\begin{tabular}{ccccc}
{\psfig{figure=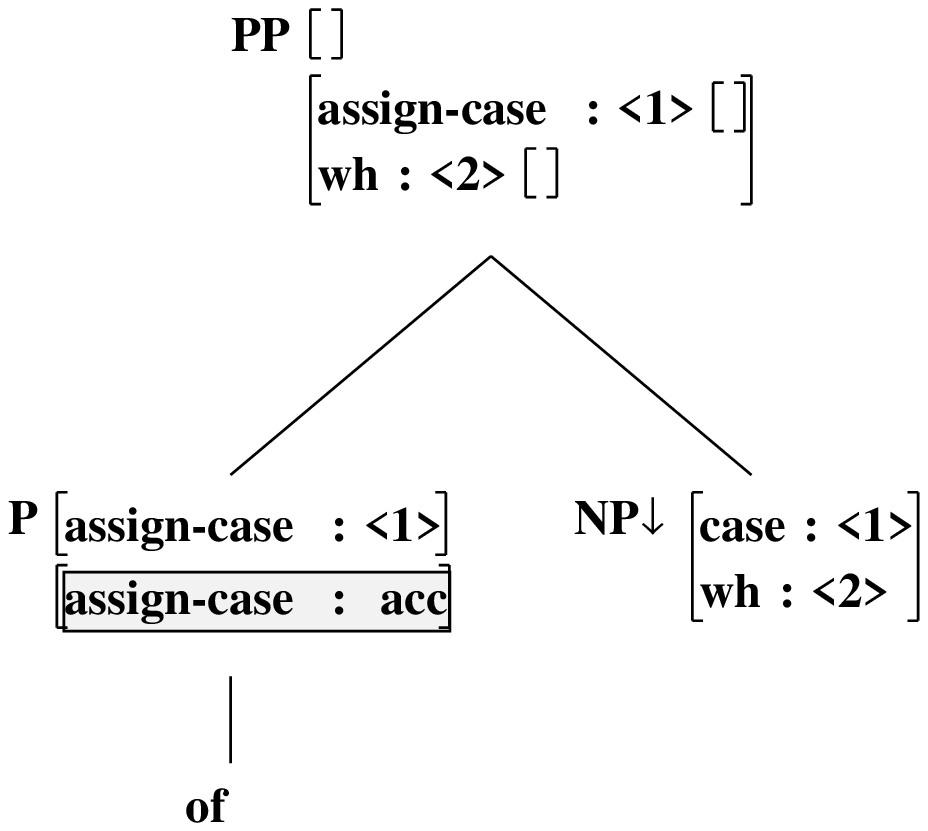,height=1.7in}}  &
&
{\psfig{figure=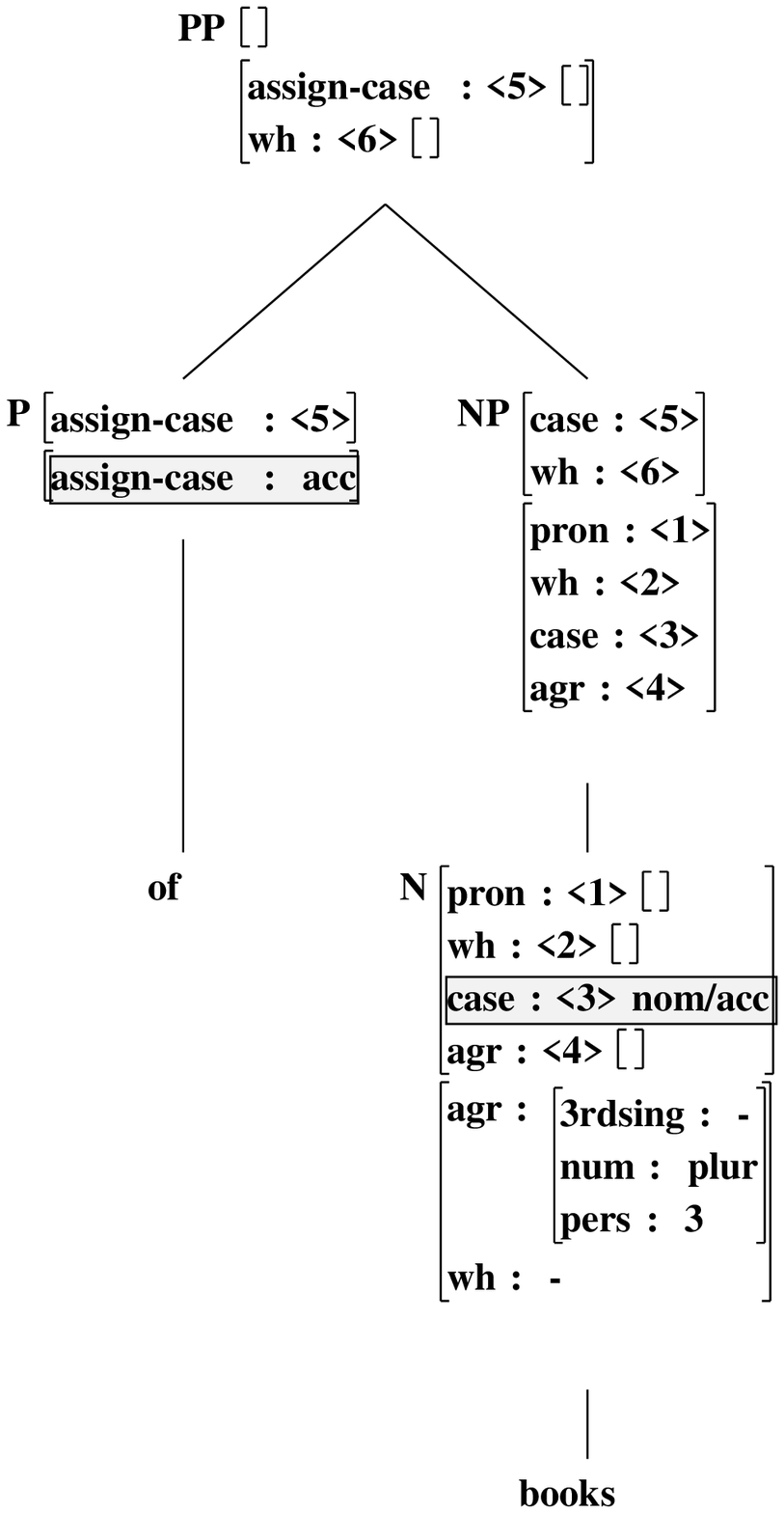,height=3.5in}} &
&
{\psfig{figure=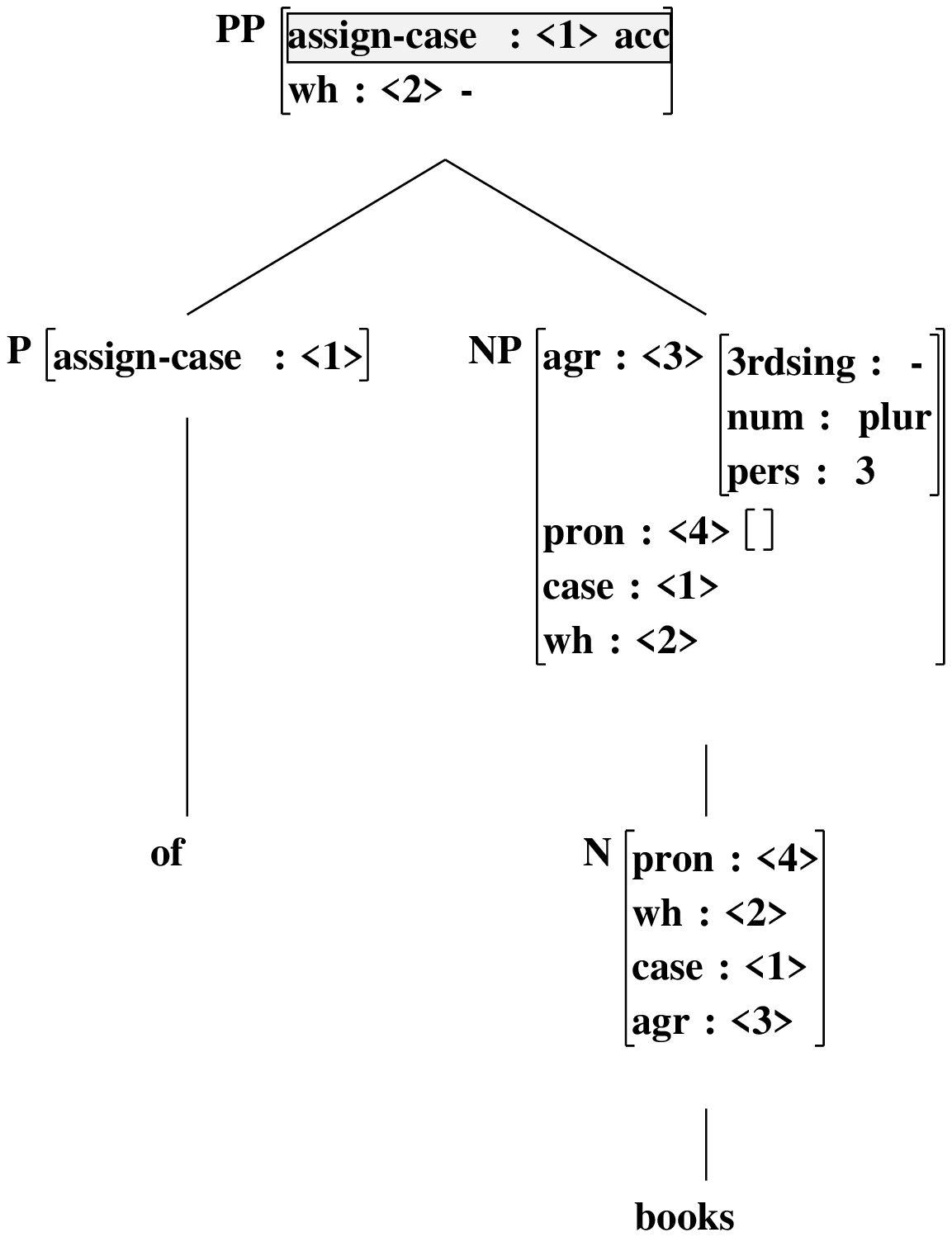,height=2.8in}} \\
(a)& \hspace*{0.05in}&(b)& \hspace*{0.05in}&(c)\\
\end{tabular}\\
\caption {Assigning case in prepositional phrases}
\label{PXPnx-with-case}
\end{figure}

\subsubsection{Verbs}
\label{case-for-verbs}
Verbs are the other part of speech in the XTAG grammar that can assign case.
Because XTAG does not distinguish INFL and VP nodes, verbs must provide case
assignment on the subject position in addition to the case assigned to their NP
complements.

Assigning case to NP complements is handled by building the case values of the
complements directly into the tree that the case assigner (the verb) anchors.
Figures~\ref{S-tree-with-case}(a) and \ref{S-tree-with-case}(b) show an S
tree\footnote{Features not pertaining to this discussion have been taken out to
improve readability and to make the trees easier to fit onto the page.} that
would be anchored\footnote{The diamond marker ($\diamond$) indicates the
anchor(s) of a structure if the tree has not yet been lexicalized.} by a
transitive and ditransitive verb, respectively.  Note that the case assignments
for the NP complements are already in the tree, even though there is not yet a
lexical item anchoring the tree.  Since every verb that selects these trees
(and other trees in each respective subcategorization frame) assigns the same
case to the complements, building case features into the tree has exactly the
same result as putting the case feature value in each verb's lexical entry.

\begin{figure}[htb]
\centering
\begin{tabular}{ccc}
{\psfig{figure=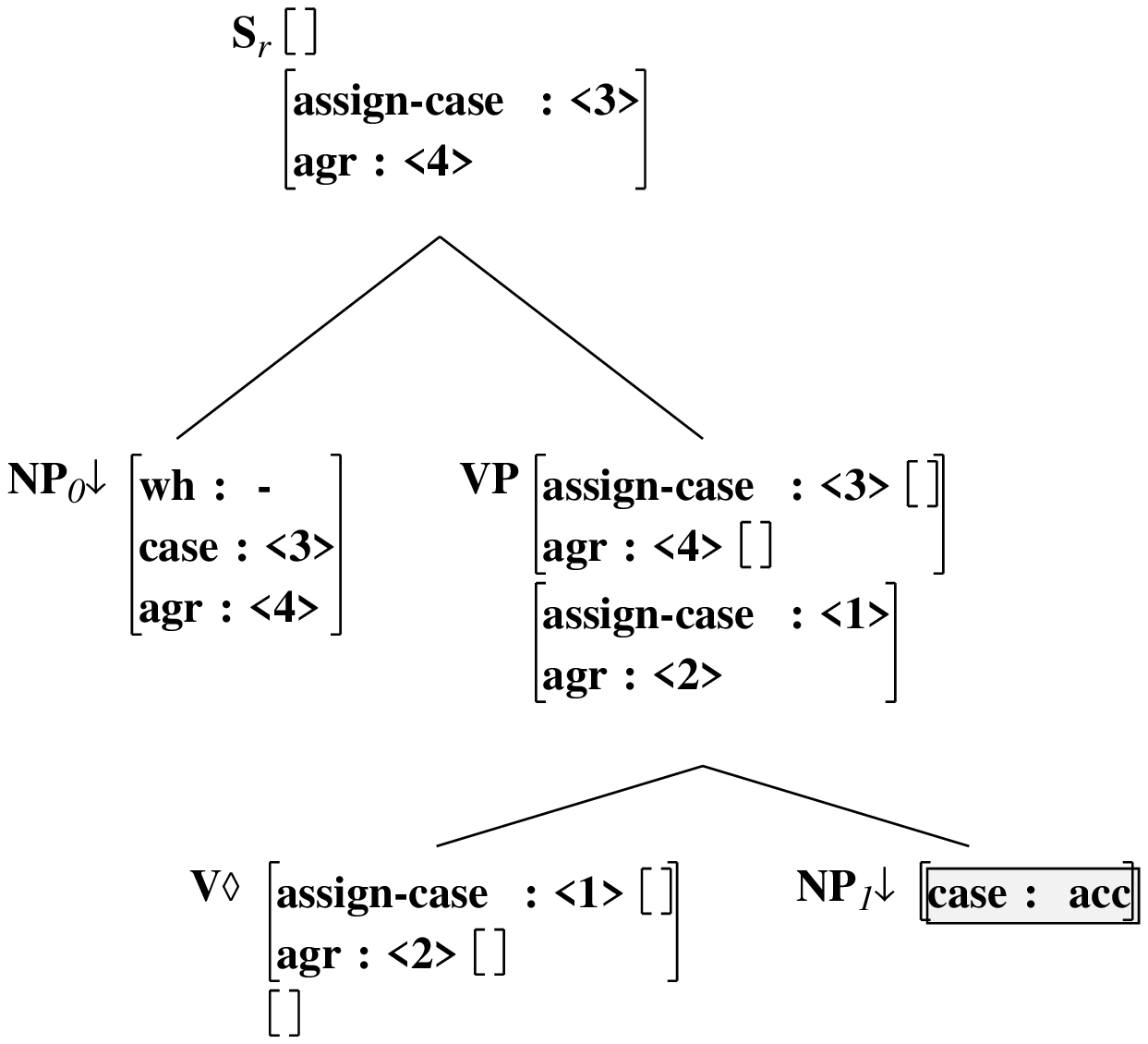,height=2.0in}}
& \hspace*{0.5in} &
{\psfig{figure=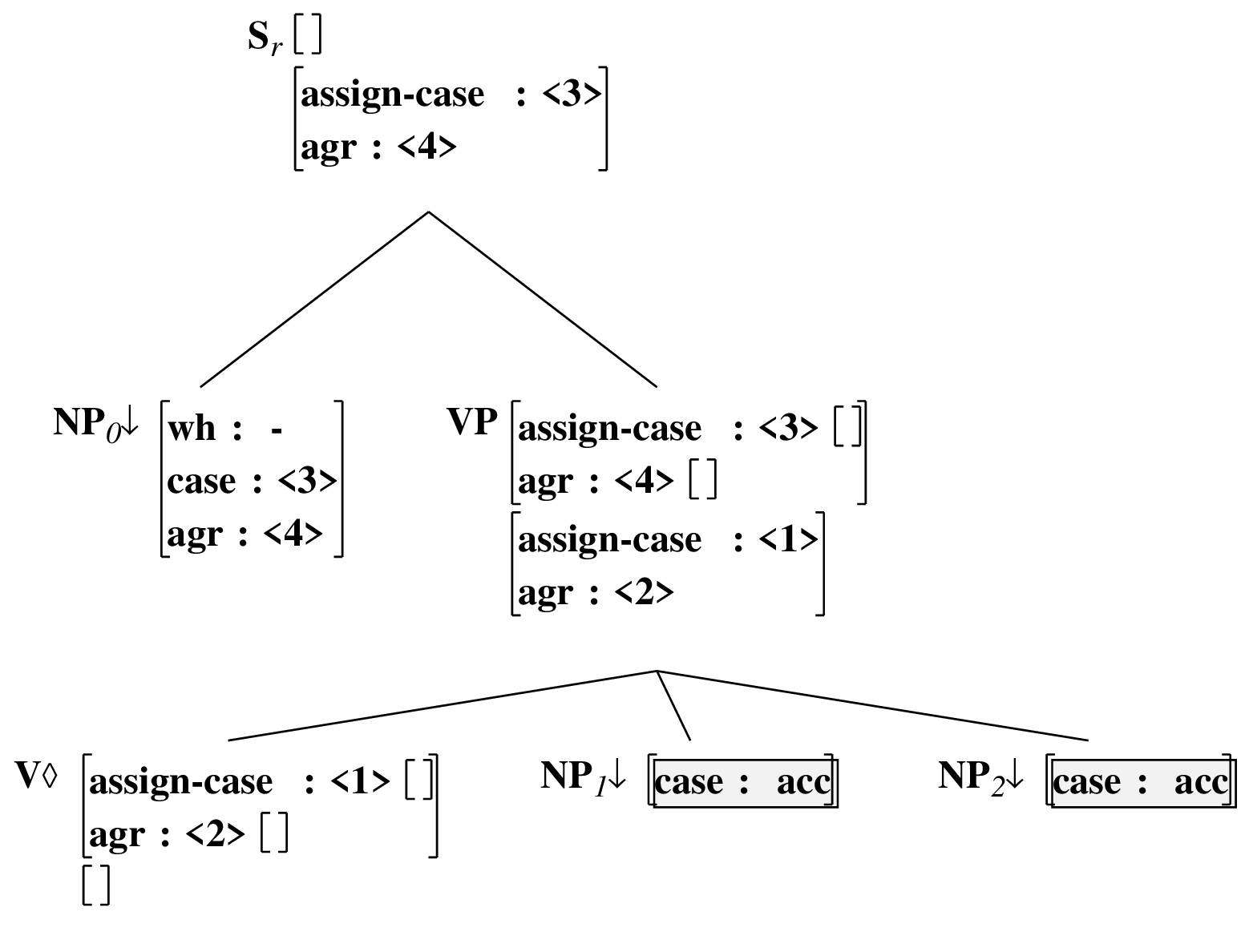,height=2.0in}} \\
(a)& \hspace*{0.5in}&(b)\\
\end{tabular}\\
\caption {Case assignment to NP arguments}
\label{S-tree-with-case}
\label{2;1,1}
\label{2;1,3}
\end{figure}

The case assigned to the subject position varies with verb form.  Since the
XTAG grammar treats the inflected verb as a single unit rather than dividing it
into INFL and V nodes, case, along with tense and agreement, is expressed in
the features of verbs, and must be passed in the appropriate manner.  The trees
in Figure~\ref{lexicalized-S-tree-with-case} show the path of linkages that
joins the {\bf$<$assign-case$>$} feature of the V to the {\bf $<$case$>$}
feature of the subject NP.  The morphological form of the verb determines the
value of the {\bf $<$assign-case$>$} feature.
Figures~\ref{lexicalized-S-tree-with-case}(a) and
\ref{lexicalized-S-tree-with-case}(b) show the same tree\footnote{Again, the 
feature structures shown have been restricted to those that pertain to the V/NP
interaction.} anchored by different morphological forms of the verb {\it sing},
which give different values for the {\bf $<$assign-case$>$} feature.

\begin{figure}[htbp]
\centering
\begin{tabular}{ccc}
{\psfig{figure=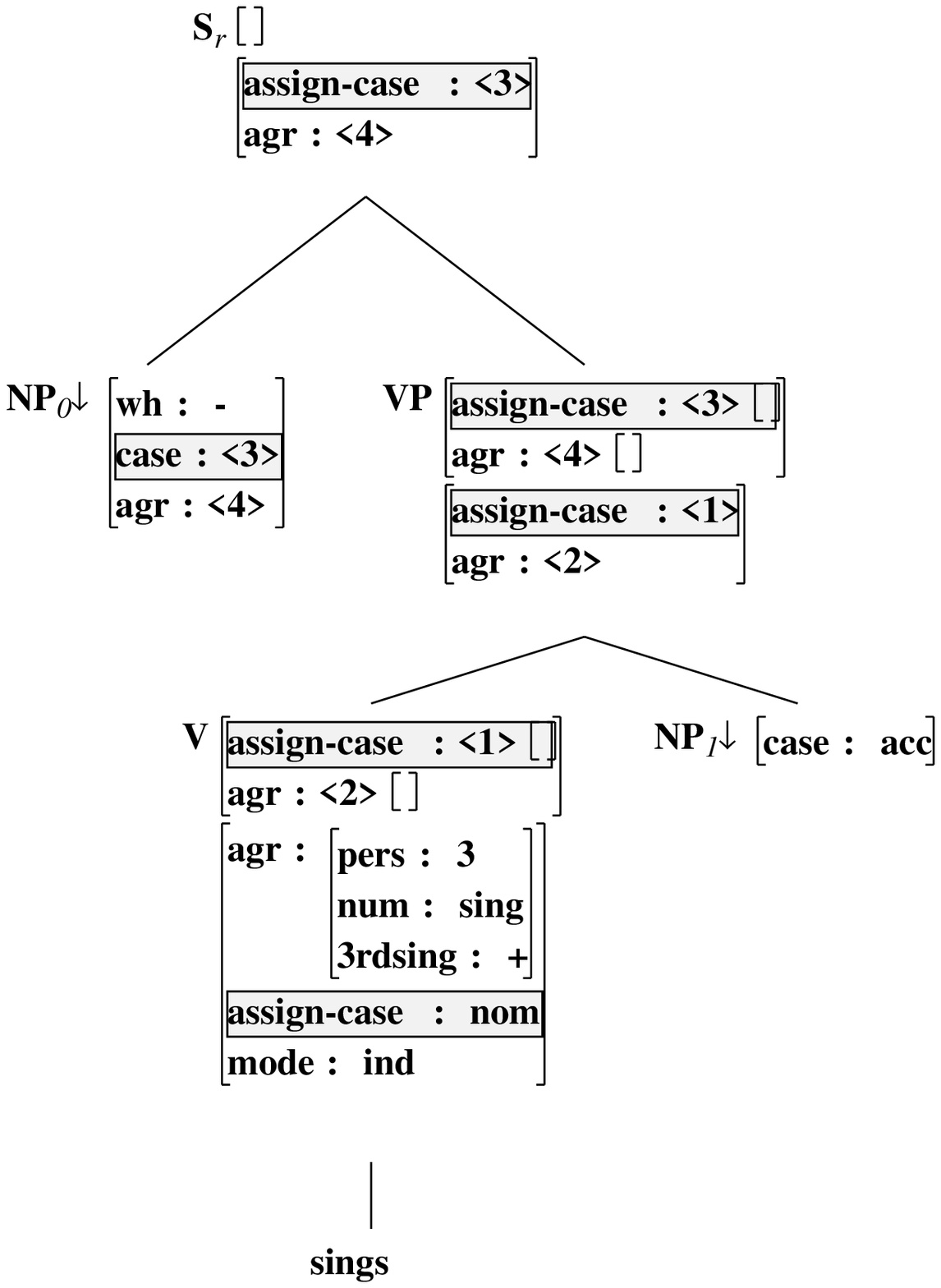,height=3.3in}}  & \hspace*{0.5in}&
{\psfig{figure=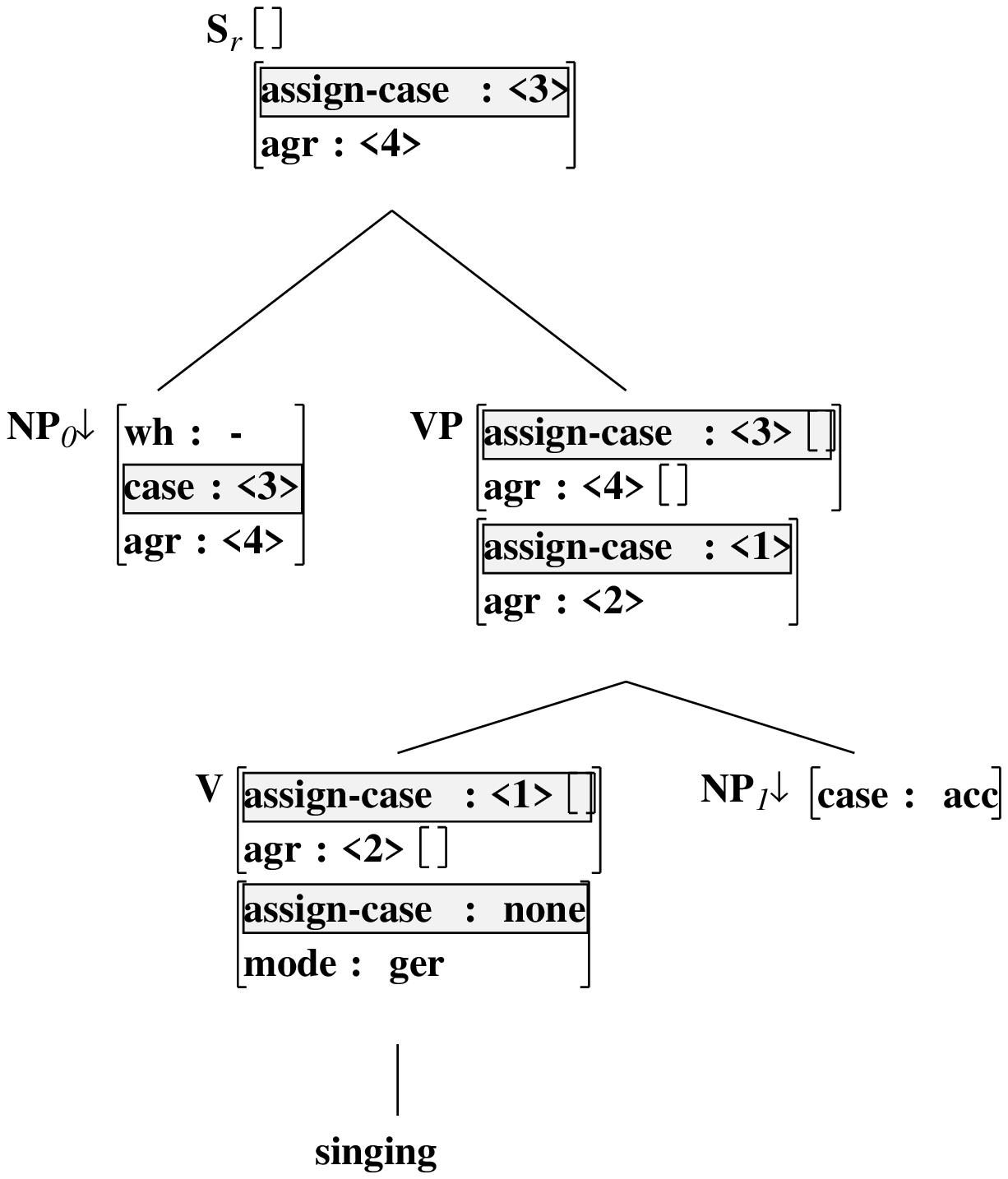,height=3.0in}} \\
(a)& \hspace*{0.5in}&(b)\\
\end{tabular}\\
\caption {Assigning case according to verb form}
\label {lexicalized-S-tree-with-case}
\end{figure}

\begin{figure}[htbp]
\centering
\begin{tabular}{ccc}
{\psfig{figure=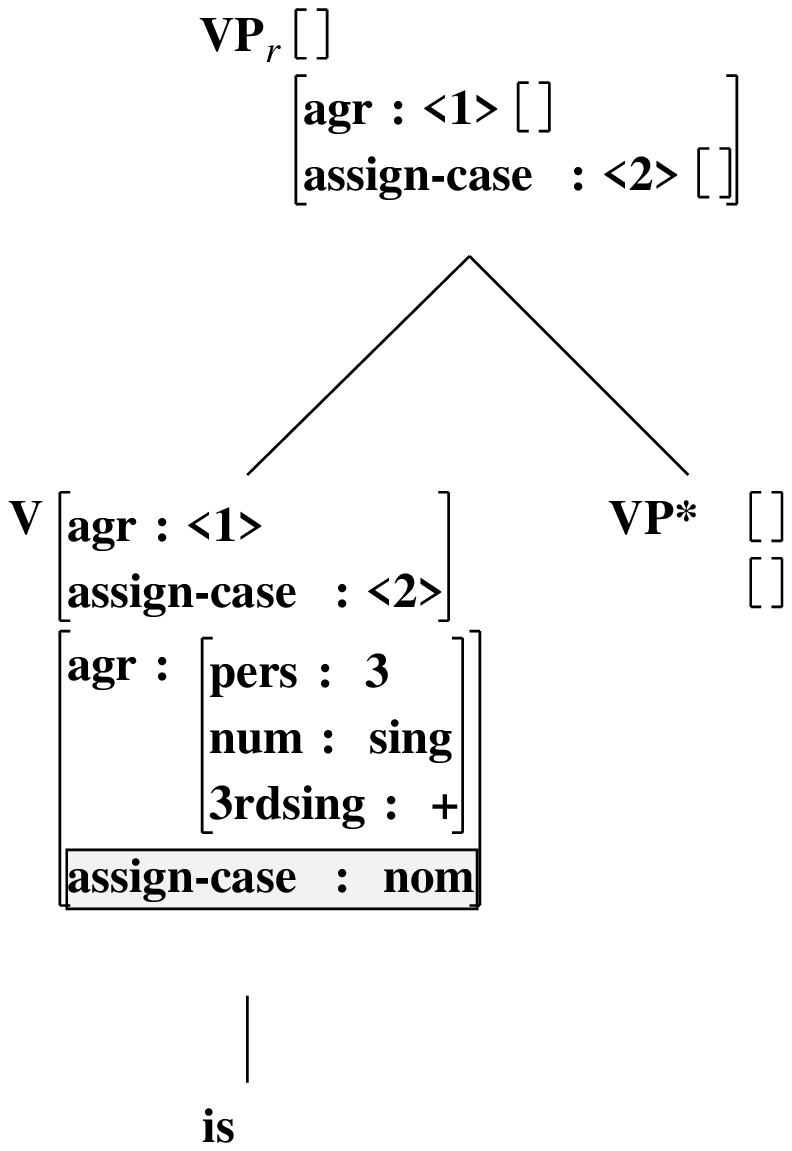,height=2.6in}}  &
\hspace*{0.5in} &
{\psfig{figure=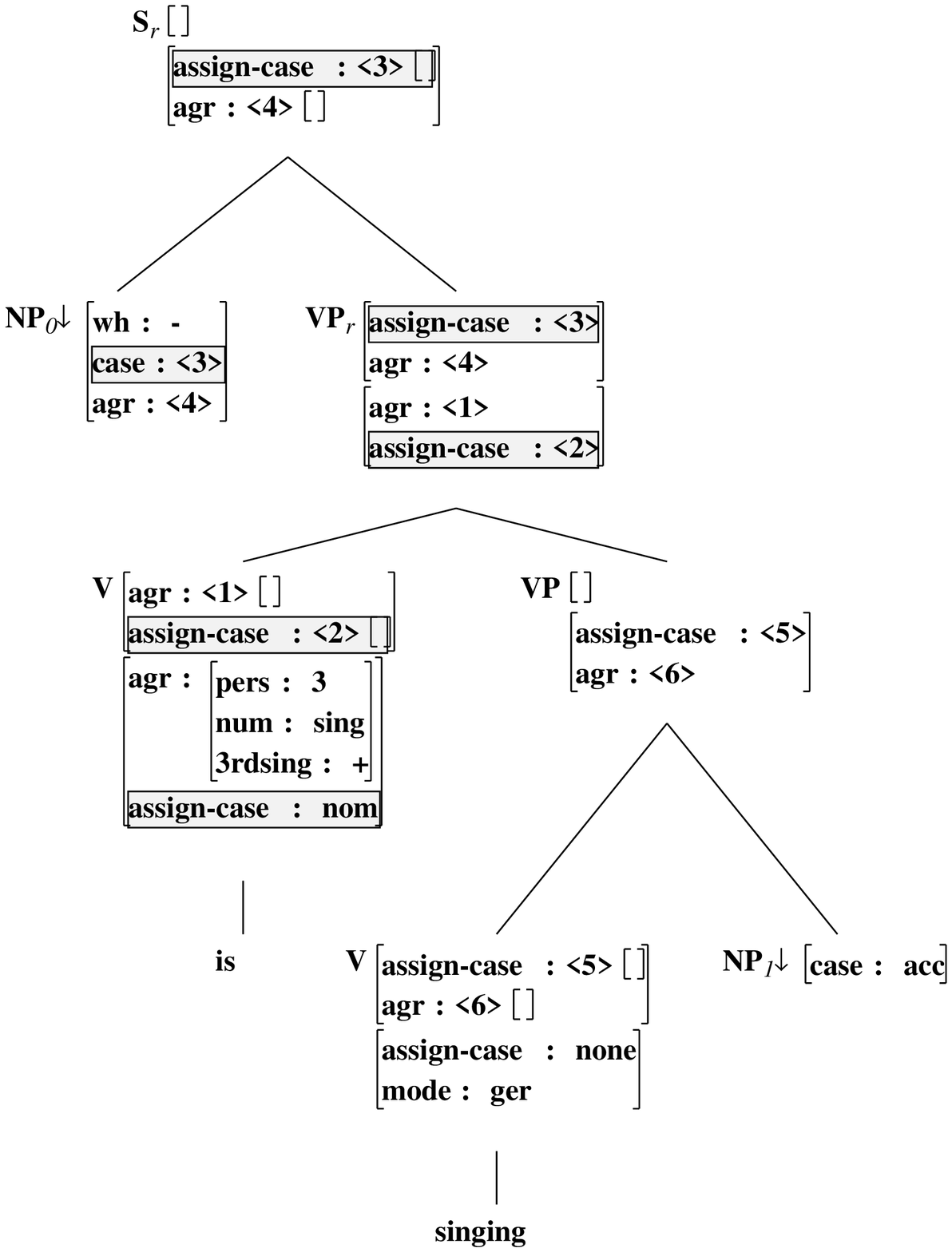,height=3.7in}} \\
(a)&\hspace*{0.5in} &(b)\\
\end{tabular}\\
\caption {Proper case assignment with auxiliary verbs}
\label{Vvx-with-case}
\end{figure}

The adjunction of an auxiliary verb onto the VP node breaks the {\bf
$<$assign-case$>$} link from the main V, replacing it with a link from the
auxiliary verb instead.\footnote{See section \ref{aux-non-inverted} for a more
complete explanation of how this relinking occurs.} The progressive form of the
verb in Figure~\ref{lexicalized-S-tree-with-case}(b) has the feature-value {\bf
$<$assign-case$>$=none}, but this is overridden by the adjunction of the
appropriate form of the auxiliary word {\it be}.  Figure~\ref{Vvx-with-case}(a)
shows the lexicalized auxiliary tree, while Figure~\ref{Vvx-with-case}(b) shows
it adjoined into the transitive tree shown in
Figure~\ref{lexicalized-S-tree-with-case}(b).  The case value passed to the
subject NP is now {\bf nom} (nominative).

\subsection{PRO in a unification based framework}

Tensed forms of a verb assign nominative case, and untensed forms
assign case {\bf none}, as the progressive form of the verb {\it sing}
does in Figure~\ref{lexicalized-S-tree-with-case}(b). This is
different than assigning no case at all, as one form of the infinitive
marker {\it to} does. See Section~\ref{for-complementizer} for more
discussion of this special case.) The distinction of a case {\bf none}
from no case is indicative of a divergence from the standard GB
theory.  In GB theory, the absence of case on an NP means that only
PRO can fill that NP.  With feature unification as is used in the
FB-LTAG grammar, the absence of case on an NP means that {\em any\/}
NP can fill it, regardless of its case.  This is due to the mechanism
of unification, in which if something is unspecified, it can unify
with anything.  Thus we have a specific case {\bf none} to handle verb
forms that in GB theory do not assign case.  PRO is the only NP with
case {\bf none}.  Note that although we are drawn to this treatment by
our use of unification for feature manipulation, our treatment is very
similar to the assignment of null case to PRO in
\cite{ChomskyLasnik93}.  \cite{watanabe93} also proposes a very similar 
approach within Chomsky's Minimalist framework.\footnote{See Sections~
\ref{PRO} and \ref{PRO-control}
for additional discussion of PRO.}

\part{Verb Classes}
\chapter{Where to Find What}
\label{table-intro}

The two page table that follows gives an overview of what types of
trees occur in various tree families with pointers to discussion in
this report.  An entry in a cell of the table indicates that the
tree(s) for the construction named in the row header are included in
the tree family named in the column header. Entries are of two types.
If the particular tree(s) are displayed and/or discussed in this
report the entry gives a page number reference to the relevant
discussion or figure.\footnote{Since Chapter~\ref{verb-classes} has a
  brief discussion and a declarative tree for every tree family, page
  references are given only for other sections in which discussion or
  tree diagrams appear.}  Otherwise, a \xtagcheck \space indicates
inclusion in the tree family but no figure or discussion related
specifically to that tree in this report.  Blank cells indicate that
there are no trees for the construction named in the row header in the
tree family named in the column header.  Two tables are given below.
The first one gives the expansion of abbreviations in the table
headers. The second table gives the name given to each tree family in
the actual XTAG grammar. This makes it easier to find the description
of each tree family in Chapter~\ref{verb-classes} and to compare the
description with the online XTAG grammar.

\vspace{0.3in}

\small
\begin{tabular}{ll}
Abbreviation&Full Name\\
\hline
Sent. Subj. w. {\it to} & Sentential Subject with {\it to} PP complement \\
Pred. Mult-wd. ARB, P & Predicative Multi-word PP with Adv, Prep anchors\\
Pred. Mult-wd. A, P & Predicative Multi-word PP with Adj, Prep anchors\\
Pred. Mult-wd. N, P & Predicative Multi-word PP with Noun, Prep
anchors\\
Pred. Mult-wd. P, P & Predicative Multi-word PP with two Prep
anchors\\
Pred. Mult-wd. no int. mod. & Predicative Multi-word PP with no internal
modification\\
Pred. Sent. Subj., ARB, P & Predicative PP with Sentential Subject, and
Adv, Prep anchors\\
Pred. Sent. Subj., A, P & Predicative PP with Sentential Subject, and
Adj, Prep anchors\\
Pred. Sent. Subj., Conj, P & Predicative PP with Sentential Subject, and
Conj, Prep anchors\\
Pred. Sent. Subj., N, P & Predicative PP with Sentential Subject, and
Noun, Prep anchors\\
Pred. Sent. Subj., P, P & Predicative PP with Sentential Subject, and two
Prep anchors\\
Pred. Sent. Subj., no int-mod & Predicative PP with Sentential Subject,
no internal modification\\
Pred. Locative & Predicative anchored by a Locative Adverb\\
Pred. A Sent. Subj., Comp. & Predicative Adjective with Sentential
Subject and Complement\\
Sentential Comp. with NP&Sentential Complement with NP\\
Pred. Mult wd. V, P & Predicative Multi-word with Verb, Prep anchors \\
Adj. Sm. Cl. w. Sentential Subj.&Adjective Small Clause with Sentential Subject\\
NP Sm. Clause w. Sentential Subj.&NP Small Clause with Sentential Subject\\
PP Sm. Clause w. Sentential Subj.&PP Small Clause with Sentential Subject\\
NP Sm. Cl. w. Sent. Comp.&NP Small Clause with Sentential Complement\\
Adj. Sm. Cl. w. Sent. Comp.&Adjective Small Clause with Sentential
Complement\\
Exhaustive PP Sm. Cl.&Exhaustive PP Small Clause\\
Ditrans. Light Verbs w. PP Shift&Ditransitive Light Verbs with PP Shift\\
Ditrans. Light Verbs w/o PP Shift&Ditransitive Light Verbs without PP Shift\\
Y/N question&Yes/No question \\
Wh-mov. NP complement&Wh-moved NP complement \\
Wh-mov. S comp.&Wh-moved S complement \\
Wh-mov. Adj comp.&Wh-moved Adjective complement \\
Wh-mov. object of a P&Wh-moved object of a P \\
Wh-mov. PP&Wh-moved PP \\
Topic. NP complement&Topicalized NP complement \\
Det. gerund&Determiner gerund \\
Rel. cl. on NP comp.&Relative clause on NP complement \\
Rel. cl. on PP comp.& Relative clause on PP complement\\
Rel. cl. on NP object of P& Relative clause on NP object of P\\
Pass. with wh-moved subj.&Passive with wh-moved subject (with and without {\it by} phrase) \\
Pass. w. wh-mov. ind. obj.&Passive with wh-moved indirect object (with and without {\it by} phrase) \\
Pass. w. wh-mov. obj. of the {\it {\it by} phrase}&Passive with wh-moved object of the {\it by} phrase \\
Pass. w. wh-mov. {\it by} phrase&Passive with wh-moved {\it by} phrase \\
Trans. Idiom with V, D and N & Transitive Idiom with Verb, Det and
Noun anchors\\
Idiom with V, D, N & Idiom with V, D, and N anchors \\
Idiom with V, D, A, N & Idiom with V, D, A, and N anchors \\
Idiom with V, N & Idiom with V, and N anchor \\
Idiom with V, A, N & Idiom with V, A, and N anchors \\
Idiom with V, D, N, P & Idiom with V, D, N, and Prep anchors \\
Idiom with V, D, A, N, P & Idiom with V, D, A, N, and Prep anchors \\
Idiom with V, N, P & Idiom with V, N, and Prep anchors \\
Idiom with V, A, N, P & Idiom with V, A, N, and Prep anchors 
\end{tabular}
\normalsize

\small
\begin{tabular}{ll}
Full Name&XTAG Name\\
\hline
Intransitive Sentential Subject &  Ts0V\\
Sentential Subject with `to' complement &  Ts0Vtonx1\\
PP Small Clause, with Adv and Prep anchors & Tnx0ARBPnx1\\
PP Small Clause, with Adj and Prep anchors & Tnx0APnx1\\
PP Small Clause, with Noun and Prep anchors & Tnx0NPnx1\\
PP Small Clause, with Prep anchors & Tnx0PPnx1\\
PP Small Clause, with Prep and Noun anchors & Tnx0PNaPnx1\\
PP Small Clause with Sentential Subject, and Adv and Prep anchors & Ts0ARBPnx1\\
PP Small Clause with Sentential Subject, and Adj and Prep anchors & Ts0APnx1\\
PP Small Clause with Sentential Subject, and Noun and Prep anchors & Ts0NPnx1\\
PP Small Clause with Sentential Subject, and Prep anchors & Ts0PPnx1\\
PP Small Clause with Sentential Subject, and Prep and Noun anchors & Ts0PNaPnx1\\
Exceptional Case Marking & TXnx0Vs1\\
Locative Small Clause with Ad anchor & Tnx0nx1ARB\\
Predicative Adjective with Sentential Subject and Complement & Ts0A1s1\\
Transitive & Tnx0Vnx1\\
Ditransitive with PP shift & Tnx0Vnx1tonx2\\
Ditransitive & Tnx0Vnx1nx2\\
Ditransitive with PP & Tnx0Vnx1pnx2\\
Sentential Complement with NP & Tnx0Vnx1s2\\
Intransitive Verb Particle & Tnx0Vpl\\
Transitive Verb Particle & Tnx0Vplnx1\\
Ditransitive Verb Particle & Tnx0Vplnx1nx2\\
Intransitive with PP & Tnx0Vpnx1\\
Sentential Complement & Tnx0Vs1\\
Light Verbs & Tnx0lVN1\\
Ditransitive Light Verbs with PP Shift & Tnx0lVN1Pnx2\\
Adjective Small Clause with Sentential Subject & Ts0Ax1\\
NP Small Clause with Sentential Subject &  Ts0N1\\
PP Small Clause with Sentential Subject & Ts0Pnx1\\
Predicative Multi-word with Verb, Prep anchors & Tnx0VPnx1\\
Adverb It-Cleft & TItVad1s2\\
NP It-Cleft & TItVnx1s2\\
PP It-Cleft & TItVpnx1s2\\
Adjective Small Clause Tree & Tnx0Ax1\\
Adjective Small Clause with Sentential Complement & Tnx0A1s1\\
Equative {\it BE} & Tnx0BEnx1\\
NP Small Clause & Tnx0N1\\
NP with Sentential Complement Small Clause & Tnx0N1s1\\
PP Small Clause & Tnx0Pnx1\\
Exhaustive PP Small Clause & Tnx0Px1\\
Intransitive & Tnx0V\\
Intransitive with Adjective & Tnx0Vax1\\
Transitive Sentential Subject &  Ts0Vnx1\\
Idiom with V, D and N & Tnx0VDN1\\
Idiom with V, D, A, and N anchors & Tnx0VDAN1\\
Idiom with V and N anchors & Tnx0VN1\\
Idiom with V, A, and N anchors & Tnx0VAN1\\
Idiom with V, D, N, and Prep anchors & Tnx0VDN1Pnx2\\
Idiom with V, D, A, N, and Prep anchors & Tnx0VDAN1Pnx2\\
Idiom with V, N, and Prep anchors & Tnx0VN1Pnx2\\
Idiom with V, A, N, and Prep anchors & Tnx0VAN1Pnx2
\end{tabular}
\normalsize

\clearpage

\begin{center}
\hspace*{-0.75in}  %%% the table is too wide and center doesn't work
\begin{tabular}{|p{2.4in}||*{15}{c|}}
\cline{2-16}
\multicolumn{1}{c||}{} & \multicolumn{15}{c|}{Tree families}\\
\hline
\vspace*{10em}
& & & & & & & & & & & & & & & \\
 &
\vertical{Intransitive Sentential Subj } &
\vertical{Sent. Subj. w. to } & %s0Vtonx1 
\vertical{Pred. Mult-wd. ARB, P } &
\vertical{Pred. Mult-wd. A, P } &
\vertical{Pred. Mult-wd. N, P } &
\vertical{Pred. Mult-wd. P, P } &
\vertical{Pred. Mult-wd. no int. mod. } &
\vertical{Pred. Sent. Subj., ARB, P } &
\vertical{Pred. Sent. Subj., A, P } &
\vertical{Pred. Sent. Subj., N, P } &
\vertical{Pred. Sent. Subj., P, P } &
\vertical{Pred. Sent. Subj., no int-mod } &
\vertical{ECM}  & %TXnx0Vs1 
\vertical{Pred. Locative} &  %Tnx0nx1ARB
\vertical{Pred. A Sent. Subj., Comp.} \\

\hline\hline
\vspace*{-2.3em} \centerline{Constructions} \vspace*{0.5em}
Declarative & \xtagcheck & \xtagcheck &\xtagcheck &\xtagcheck
&\xtagcheck & \xtagcheck& \xtagcheck& \xtagcheck& \xtagcheck&
\xtagcheck &\xtagcheck &\xtagcheck & {\tiny \pageref{3;1,15}}  & {\tiny \pageref{3;nx0nx1ARB}} &\\
\hline
Passive w/ \& w/o {\it by} phrase & & & & & & & & &  & & & & {\tiny \pageref{3;2,15}} & &\\
\hline
Y/N quest. & & &  &  &  & &  & & & & & & & & \\
\hline
Wh-moved subject & \xtagcheck &  \xtagcheck & \xtagcheck & \xtagcheck
& \xtagcheck &  \xtagcheck & \xtagcheck& \xtagcheck& \xtagcheck &
\xtagcheck & \xtagcheck & \xtagcheck  & \xtagcheck & \xtagcheck &
\xtagcheck \\
\hline
Wh-mov.\ NP complement, DO or IO & & & & & & & & & & & & & & & \\
\hline
Wh-mov.\ S comp. & & & & & & & & & & & & & & & \\
\hline
Wh-mov.\ Adj. or Adv.\ comp. & & & & & & & & & & & & & & {\tiny \pageref{3;W1nx0nx1ARB}} & \\
        \hline
Wh-mov.\ object of a P & & & \xtagcheck & \xtagcheck & \xtagcheck & \xtagcheck & \xtagcheck & & & & & & & & \\
\hline
Wh-mov.\ PP & & & \xtagcheck & &  & \xtagcheck & \xtagcheck & & & & & & & & \\
\hline
Topic.\ NP comp.& & & & & & & & & & & & & & & \\
\hline
Imperative & & & \xtagcheck & & & \xtagcheck & \xtagcheck  & & & & & &
\xtagcheck  & \xtagcheck  & \\
\hline
Det.\ gerund & & & & & & & & & & & & & & & \\
\hline
NP gerund & & & \xtagcheck & & & & \xtagcheck  & & & & & & \xtagcheck & \xtagcheck &\\
\hline
Ergative & & & & & & & & & & & & & & & \\
\hline
Rel.\ cl.\ on subj. w/ NP  & & & \xtagcheck & \xtagcheck & \xtagcheck & \xtagcheck & \xtagcheck & & & & & & \xtagcheck & \xtagcheck &\\
\hline
Rel.\ cl.\ on subj. w/ Comp  & & & \xtagcheck & \xtagcheck & \xtagcheck & \xtagcheck & \xtagcheck & & & & & & \xtagcheck & \xtagcheck &\\
\hline
Rel.\ cl.\ on NP comp., DO, IO w/ NP & & & & & & & & & & & & & & & \\
\hline
Rel.\ cl.\ on NP comp., DO, IO w/ Comp & & & & & & & & & & & & & & & \\
\hline
Rel.\ cl.\ on PP comp. w/ pied-piping  & & & & & & \xtagcheck & \xtagcheck &  &  & & & & & & \\
\hline
Rel.\ cl.\ on NP object of P w/ NP & & & \xtagcheck & \xtagcheck &  \xtagcheck & \xtagcheck  & \xtagcheck & \xtagcheck & \xtagcheck & \xtagcheck & \xtagcheck & \xtagcheck & & &\\
\hline
Rel.\ cl.\ on NP object of P w/ Comp & & & \xtagcheck & \xtagcheck & & \xtagcheck  & \xtagcheck & \xtagcheck & \xtagcheck & \xtagcheck & \xtagcheck & \xtagcheck & & &\\
\hline
Rel.\ cl.\ on adjunct w/ PP & \xtagcheck & \xtagcheck & \xtagcheck &
\xtagcheck &  \xtagcheck & \xtagcheck  & \xtagcheck & \xtagcheck &
\xtagcheck & \xtagcheck & \xtagcheck & \xtagcheck & \xtagcheck & &
\xtagcheck \\
\hline
Rel.\ cl.\ on adjunct w/ Comp & \xtagcheck & \xtagcheck & \xtagcheck &
\xtagcheck &  \xtagcheck & \xtagcheck  & \xtagcheck & \xtagcheck &
\xtagcheck & \xtagcheck & \xtagcheck & \xtagcheck & \xtagcheck & &
\xtagcheck \\
\hline
Pass.\ w.\ wh-mov.\ subj.\ & & & & & & & & & & & & & & & \\
\hline
Pass.\ w.\ wh-mov.\ ind.\ obj.\ & & & & & & & & & & & & & & & \\
\hline
Pass.\ w.\ wh-mov.\ obj. of  {\it by} phrase  & & & & & & & & & & & & & & & \\
\hline
Pass.\ w.\ wh-mov.\ {\it by} phrase  & & & & & & & & & & & & & & & \\
\hline
\end{tabular}
\end{center}

\clearpage

\vspace*{-0.5in}

\begin{center}
\hspace*{-0.75in}  %%% the table is too wide and center doesn't work
\begin{tabular}{|p{2.4in}||*{16}{c|}}
\cline{2-17}
\multicolumn{1}{c||}{} & \multicolumn{16}{c|}{Tree families}\\
\hline
\vspace*{12em} & & & & & & & & & & & & & & & & \\
 &
\vertical{Transitive} &
\vertical{Ditransitive with PP shift} &
\vertical{Ditransitive} &
\vertical{Ditransitive with PP} &
\vertical{Sentential Comp.\ with NP} &
\vertical{Intransitive Verb Particle} &
\vertical{Transitive Verb Particle} &
\vertical{Ditransitive Verb Particle} &
\vertical{Intransitive with PP} &
\vertical{Sentential Complement} &
\vertical{Trans. Light Vs} &
\vertical{Ditrans.\ Light Vs} &
\vertical{Adj.\ Sm.\ Cl.\ w.\ Sentential Subj.} &
\vertical{NP Sm.\ Cl.\ w.\ Sentential Subj.} &
\vertical{PP Sm.\ Cl.\ w.\ Sentential Subj.} &
\vertical{Pred.\ Mult.\ wd. V, P} \\

\hline\hline
\vspace*{-2.3em} \centerline{Constructions} \vspace*{0.5em}
Declarative &{\tiny \pageref{2;1,1}} & {\tiny \pageref{2;1,2}} & {\tiny \pageref{2;1,3}}& \xtagcheck & \xtagcheck & \xtagcheck & \xtagcheck & \xtagcheck &{\tiny \pageref{2;1,9}}&{\tiny \pageref{2;Tnx0Vs1},\pageref{2;1,10}} & \xtagcheck & \xtagcheck & \xtagcheck & \xtagcheck & \xtagcheck &\xtagcheck \\
\hline
Passive w/ \& w/o {\it by} phrase &\xtagcheck & \xtagcheck & \xtagcheck & \xtagcheck & {\tiny \pageref{2;2,5}} & & \xtagcheck & \xtagcheck & & & & \xtagcheck & & & &\xtagcheck\\
\hline
Y/N quest.\ & & & & & & & & & & & & & & & &\\
\hline
Wh-moved subject & \xtagcheck& \xtagcheck& \xtagcheck& \xtagcheck& \xtagcheck&\xtagcheck &\xtagcheck &\xtagcheck &\xtagcheck &\xtagcheck  &\xtagcheck & &\xtagcheck & \xtagcheck& \xtagcheck &\xtagcheck \\
\hline
Wh-mov.\ NP complement, DO or IO  &{\tiny \pageref{2;5,1}}&\xtagcheck &{\tiny \pageref{2;5,3}}&\xtagcheck &\xtagcheck & &\xtagcheck &\xtagcheck & & & & & & & & \\
\hline
Wh-mov.\ S comp.\ & & & & & \xtagcheck & & & & & \xtagcheck & & & & & & \\
\hline
Wh-mov.\ Adj. or Adv.\ comp.  & & & & & & & & & & & & & \xtagcheck & & & \\
\hline
Wh-mov.\ object of a P  & &\xtagcheck & &{\tiny \pageref{2;8,4}}& & & & &\xtagcheck & & &\xtagcheck & & & &\xtagcheck  \\
\hline
Wh-mov.\ PP  & &\xtagcheck & &{\tiny \pageref{2;9,4}}& & & & &\xtagcheck & & &\xtagcheck & & & & \\
\hline
Topic.\ NP comp.  &\xtagcheck &\xtagcheck &\xtagcheck &\xtagcheck &\xtagcheck & &\xtagcheck &\xtagcheck & & & & & & & & \\
\hline
Imperative &{\tiny \pageref{2;11,1}}&\xtagcheck &\xtagcheck & \xtagcheck&\xtagcheck &\xtagcheck &\xtagcheck & \xtagcheck&\xtagcheck &\xtagcheck &\xtagcheck &\xtagcheck & & &  &\xtagcheck \\
\hline
Det.\ gerund &{\tiny \pageref{2;12,1}}&\xtagcheck &\xtagcheck &\xtagcheck &\xtagcheck &\xtagcheck &\xtagcheck &\xtagcheck &\xtagcheck &\xtagcheck &\xtagcheck &\xtagcheck & & & &\xtagcheck  \\
\hline
NP gerund &{\tiny \pageref{2;13,1}}&\xtagcheck &\xtagcheck &\xtagcheck &\xtagcheck & \xtagcheck& \xtagcheck& \xtagcheck& \xtagcheck& \xtagcheck &\xtagcheck &\xtagcheck & & &  &\xtagcheck \\
\hline
Ergative &{\tiny \pageref{2;14,1}}& & & & & & & & & & & & & & & \\
\hline
Rel.\ cl.\ on subj. w/ NP  & \xtagcheck & \xtagcheck & \xtagcheck & \xtagcheck & \xtagcheck & \xtagcheck & \xtagcheck & \xtagcheck & \xtagcheck & \xtagcheck & \xtagcheck & \xtagcheck & & & &\xtagcheck \\
\hline
Rel.\ cl.\ on subj. w/ Comp  & \xtagcheck & \xtagcheck & \xtagcheck & \xtagcheck & \xtagcheck & \xtagcheck & \xtagcheck & \xtagcheck & \xtagcheck & \xtagcheck & \xtagcheck & \xtagcheck &  & & &\xtagcheck \\
\hline
Rel.\ cl.\ on NP comp., DO, IO w/ NP & \xtagcheck & \xtagcheck & \xtagcheck & \xtagcheck & \xtagcheck & & \xtagcheck & \xtagcheck & \xtagcheck & & & \xtagcheck & & &  & \\
\hline
Rel.\ cl.\ on NP comp., DO, IO w/ Comp & \xtagcheck & \xtagcheck & \xtagcheck & \xtagcheck & \xtagcheck & & \xtagcheck & \xtagcheck & \xtagcheck & & & \xtagcheck & & &  & \\
\hline
Rel.\ cl.\ on PP comp. w/ pied-piping  & \xtagcheck & \xtagcheck & \xtagcheck & \xtagcheck & \xtagcheck & & \xtagcheck & & \xtagcheck & & & \xtagcheck & & & \xtagcheck &\\
\hline
Rel.\ cl.\ on NP object of P w/ NP & \xtagcheck & \xtagcheck & \xtagcheck & \xtagcheck &  \xtagcheck & & \xtagcheck & & \xtagcheck & & & \xtagcheck & & & \xtagcheck  &\xtagcheck \\
\hline
Rel.\ cl.\ on NP object of P w/ Comp & \xtagcheck & \xtagcheck & \xtagcheck & \xtagcheck &  \xtagcheck & & \xtagcheck & & \xtagcheck & & & \xtagcheck & & &  \xtagcheck  &\xtagcheck \\
\hline
Rel.\ cl.\ on adjunct w/ PP &  \xtagcheck & \xtagcheck & \xtagcheck & \xtagcheck &  \xtagcheck &\xtagcheck  & \xtagcheck & \xtagcheck & \xtagcheck & \xtagcheck & \xtagcheck & \xtagcheck & \xtagcheck & \xtagcheck & \xtagcheck  &\xtagcheck  \\
\hline
Rel.\ cl.\ on adjunct w/ Comp &  \xtagcheck & \xtagcheck & \xtagcheck & \xtagcheck &  \xtagcheck &\xtagcheck  & \xtagcheck & \xtagcheck & \xtagcheck & \xtagcheck & \xtagcheck & \xtagcheck & \xtagcheck & \xtagcheck &  \xtagcheck  &\xtagcheck \\
\hline
Parenthetical quoting clause &   &   &   &   & \xtagcheck & &   &   &
& \xtagcheck & &   & & & & \\
\hline %alpha_AV
Past-participal as arg Adj & \xtagcheck   &   &   &   &  & &   &   & & & &   & & &  &\xtagcheck  \\
\hline %betaVtransn
Past-participial NP pre-mod  & \xtagcheck   &   &   &   & & &   &   & & & &   & & &  &\xtagcheck  \\
\hline
Pass.\ w.\ wh-mov.\ subj. &\xtagcheck &\xtagcheck &\xtagcheck &\xtagcheck &\xtagcheck & & \xtagcheck&\xtagcheck & & & &\xtagcheck & & & &\xtagcheck  \\
\hline
Pass.\ w.\ wh-mov.\ ind.\ obj. & & \xtagcheck& \xtagcheck& \xtagcheck&
\xtagcheck& & &\xtagcheck & & & & \xtagcheck& & & & \\
\hline
Pass.\ w.\ wh-mov.\ obj. of {\it by} phrase & \xtagcheck & \xtagcheck & \xtagcheck & \xtagcheck & \xtagcheck & & \xtagcheck & \xtagcheck & & & & \xtagcheck & & & &\xtagcheck  \\
\hline
Pass.\ w.\ wh-mov.\ {\it by} phrase &\xtagcheck &\xtagcheck & \xtagcheck&\xtagcheck & \xtagcheck& &\xtagcheck &\xtagcheck & & & & & & &  &\xtagcheck \\
\hline
\end{tabular}
\end{center}

\clearpage

\begin{center}
\hspace*{-0.75in}  %%% the table is too wide and center doesn't work
\begin{tabular}{|p{2.4in}||*{13}{c|}}
\cline{2-14}
\multicolumn{1}{c||}{} & \multicolumn{13}{c|}{Tree families}\\
\hline
\vspace*{10em}
& & & & & & & & & & & & & \\
 &
\vertical{Adverb It-Cleft } &
\vertical{NP It-Cleft } &
\vertical{PP It-Cleft } &
\vertical{Adj. Small Clause } &
\vertical{Adj.\ Sm.\ Cl.\ w.\ Sent.\ Comp.} &
\vertical{Equative {\it BE} } &
\vertical{NP Small Clause } &
\vertical{NP Sm.\ Cl.\ w.\ Sent.\ Comp.} &
\vertical{PP Small Clause} &
\vertical{Exhaustive PP Sm. Cl. } &
\vertical{Intransitive} &
\vertical{Intransitive with Adjective} &
\vertical{Transitive Sentential Subj} \\

\hline\hline
\vspace*{-2.3em} \centerline{Constructions} \vspace*{0.5em}
Declarative &\xtagcheck & \xtagcheck &{\tiny \pageref{1;1,3}}&{\tiny \pageref{1;1,4}}& \xtagcheck & {\tiny \pageref{1;1,6}} &{\tiny \pageref{1;1,7}}& \xtagcheck &{\tiny \pageref{1;1,9}}  & \xtagcheck & \xtagcheck & \xtagcheck & {\tiny \pageref{1;1,16}} \\
\hline
Passive w/ \& w/o {\it by} phrase & & & & & & & & &  & & & & \\
\hline
Y/N quest. & \xtagcheck & \xtagcheck & {\tiny \pageref{1;3,3}} & & & \xtagcheck & & &  & & & & \\
\hline
Wh-moved subject & & & &{\tiny \pageref{1;4,4}} & \xtagcheck& &\xtagcheck &\xtagcheck &\xtagcheck  &\xtagcheck &{\tiny \pageref{1;4,13}}& {\tiny \pageref{1;4,14}} &\xtagcheck \\
\hline
Wh-mov.\ NP complement, DO or IO & &\xtagcheck & & & & &\xtagcheck & &
 & & & & \xtagcheck \\
\hline
Wh-mov.\ S comp. & & & & & & & & &  & & & & \\
\hline
Wh-mov.\ Adj. or Adv.\ comp. &\xtagcheck & & &\xtagcheck & & & & & 
& &  & {\tiny \pageref{1;7,14}} & \\
\hline
Wh-mov.\ object of a P & & & & & & & & & \xtagcheck  & & & & \\
\hline
Wh-mov.\ PP & & &\xtagcheck & & & & & &\xtagcheck  & & & & \\
\hline
Topic.\ NP comp. & &\xtagcheck & & & & & \xtagcheck& &  & & & & \\
\hline
Imperative & & & &\xtagcheck &\xtagcheck & &\xtagcheck &\xtagcheck &\xtagcheck  &\xtagcheck &\xtagcheck &\xtagcheck & \\
\hline
Det.\ gerund & & & & & & & & &  & & & & \\
\hline
NP gerund & & & &\xtagcheck &\xtagcheck & &\xtagcheck &\xtagcheck &\xtagcheck  &\xtagcheck &\xtagcheck &\xtagcheck &  \\
\hline
Ergative & & & & & & & & &  & & & & \\
\hline
Rel.\ cl.\ on subj. w/ NP  & & & & \xtagcheck & \xtagcheck &  & \xtagcheck & \xtagcheck & \xtagcheck   & \xtagcheck & \xtagcheck & \xtagcheck & \\
\hline
Rel.\ cl.\ on subj. w/ Comp  & & & & \xtagcheck & \xtagcheck & & \xtagcheck & \xtagcheck & \xtagcheck  & \xtagcheck & \xtagcheck & \xtagcheck & \\
\hline
Rel.\ cl.\ on NP comp., DO, IO w/ NP & & & & & & & & &  & & & & \\
\hline
Rel.\ cl.\ on NP comp., DO, IO w/ Comp & & & & & & & & &  & & & & \\
\hline
Rel.\ cl.\ on PP comp. w/ pied-piping  & & & & & & & & & \xtagcheck  & & & & \\
\hline
Rel.\ cl.\ on NP object of P w/ NP & & & & & &  & & & \xtagcheck  & & & &\\
\hline
Rel.\ cl.\ on NP object of P w/ Comp & & & & & &  & & & \xtagcheck  & & & &\\
\hline
Rel.\ cl.\ on adjunct w/ PP & \xtagcheck & \xtagcheck  & \xtagcheck & \xtagcheck &  \xtagcheck &  & \xtagcheck & \xtagcheck & \xtagcheck   &  \xtagcheck & \xtagcheck  & \xtagcheck &  \xtagcheck \\
\hline
Rel.\ cl.\ on adjunct w/ Comp & \xtagcheck & \xtagcheck  & \xtagcheck & \xtagcheck &  \xtagcheck &  & \xtagcheck & \xtagcheck & \xtagcheck   &  \xtagcheck & \xtagcheck  & \xtagcheck &  \xtagcheck\\
\hline %beta_Vintransn
Participial  NP pre-mod  & & & & & & & & &  & & \xtagcheck & & \\
\hline
Pass.\ w.\ wh-mov.\ subj.\ & & & & & & & & &  & & & & \\
\hline
Pass.\ w.\ wh-mov.\ ind.\ obj.\ & & & & & & & & &  & & & & \\
\hline
Pass.\ w.\ wh-mov.\ obj. of  {\it by} phrase  & & & & & & & & &  & & & & \\
\hline
Pass.\ w.\ wh-mov.\ {\it by} phrase  & & & & & & & & &  & & & & \\
\hline
\end{tabular}
\end{center}

\begin{center}
\hspace*{-0.75in} % the table is too wide and center doesn't work
\begin{tabular}{|p{2.4in}||*{8}{c|}}
\cline{2-9}
\multicolumn{1}{c||}{} & \multicolumn{8}{c|}{Tree families}\\
\hline
\vspace*{10em}
& & & & & & & & \\
 &
\vertical{Idiom with V, D, N } &
\vertical{Idiom with V, D, A, N } &
\vertical{Idiom with V, N } &
\vertical{Idiom with V, A, N} &
\vertical{Idiom with V, D, N, P } &
\vertical{Idiom with V, D, A, N, P } &
\vertical{Idiom with V, N, P } &
\vertical{Idiom with V, A, N, P} \\

\hline\hline
\vspace*{-2.3em} \centerline{Constructions} \vspace*{0.5em}
Declarative & \xtagcheck & \xtagcheck &\xtagcheck &\xtagcheck
&\xtagcheck & \xtagcheck& \xtagcheck& \xtagcheck \\
\hline
Passive w/ \& w/o {\it by} phrase &\xtagcheck &\xtagcheck &\xtagcheck &\xtagcheck &\xtagcheck &\xtagcheck &\xtagcheck &\xtagcheck \\
\hline
Y/N quest. & & & & & & & & \\
\hline
Wh-moved subject & \xtagcheck & \xtagcheck & \xtagcheck & \xtagcheck & \xtagcheck & \xtagcheck & \xtagcheck& \xtagcheck \\
\hline
Wh-mov.\ NP complement, DO or IO & & & & & & & & \\
\hline
Wh-mov.\ S comp. & & & & & & & & \\
\hline
Wh-mov.\ Adj. or Adv.\ comp. & & & & & & & & \\
\hline
Wh-mov.\ object of a P & & & & & & & & \\
\hline
Wh-mov.\ PP & & & & & & & & \\
\hline
Topic.\ NP comp. & & & & & & & & \\
\hline
Imperative &\xtagcheck &\xtagcheck &\xtagcheck &\xtagcheck &\xtagcheck &\xtagcheck &\xtagcheck &\xtagcheck \\
\hline
Det.\ gerund & & & & & & & & \\
\hline
NP gerund &\xtagcheck &\xtagcheck &\xtagcheck &\xtagcheck &\xtagcheck &\xtagcheck &\xtagcheck &\xtagcheck \\
\hline
Ergative & & & & & & & & \\
\hline
Rel.\ cl.\ on subj. w/ NP & \xtagcheck & \xtagcheck &\xtagcheck &\xtagcheck &\xtagcheck &\xtagcheck &\xtagcheck &\xtagcheck \\
\hline
Rel.\ cl.\ on subj. w/ Comp  &\xtagcheck &\xtagcheck &\xtagcheck &\xtagcheck &\xtagcheck &\xtagcheck &\xtagcheck &\xtagcheck \\
\hline
Rel.\ cl.\ on NP comp., DO, IO w/ NP & & & & & & & & \\
\hline
Rel.\ cl.\ on NP comp., DO, IO w/ Comp & & & & & & & & \\
\hline
Rel.\ cl.\ on PP comp. w/ pied-piping  & & & & & & & & \\
\hline
Rel.\ cl.\ on NP object of P w/ NP & & & & & & & & \\
\hline
Rel.\ cl.\ on NP object of P w/ Comp & & & & & & & & \\
\hline
Rel.\ cl.\ on adjunct w/ PP & \xtagcheck & \xtagcheck & \xtagcheck & \xtagcheck &  \xtagcheck & \xtagcheck  & \xtagcheck & \xtagcheck \\
\hline
Rel.\ cl.\ on adjunct w/ Comp & \xtagcheck & \xtagcheck & \xtagcheck & \xtagcheck &  \xtagcheck & \xtagcheck  & \xtagcheck & \xtagcheck \\
\hline
Pass.\ w.\ wh-mov.\ subj.\ & & & & & & & & \\
\hline
Pass.\ w.\ wh-mov.\ ind.\ obj.\ & & & & & & & & \\
\hline
Pass.\ w.\ wh-mov.\ obj. of  {\it by} phrase & \xtagcheck & \xtagcheck &\xtagcheck &\xtagcheck &\xtagcheck &\xtagcheck &\xtagcheck &\xtagcheck \\
\hline
Pass.\ w.\ wh-mov.\ {\it by} phrase & \xtagcheck & \xtagcheck &\xtagcheck &\xtagcheck &\xtagcheck &\xtagcheck &\xtagcheck &\xtagcheck \\
\hline
Outer Pass.\ w.\ and wo.\ {\it by} phrase & & & & & \xtagcheck & \xtagcheck & \xtagcheck & \xtagcheck \\
\hline
Outer Pass.\ w.\ Rel.\ cl.\ on subj.\ w.\ Comp  & \xtagcheck & & & & \xtagcheck & \xtagcheck & \xtagcheck & \xtagcheck \\
\hline
Outer Pass.\ w.\ Rel.\ cl.\ on subj.\ w.\ NP  & \xtagcheck & & & & \xtagcheck & \xtagcheck & \xtagcheck & \xtagcheck \\
\hline
\end{tabular}
\end{center}

\clearpage

\chapter{Verb Classes}
\label{verb-classes}

Each main\footnote{Auxiliary verbs are handled under a different
mechanism.  See Chapter~\ref{auxiliaries} for details.} verb in the
syntactic lexicon selects at least one tree family\footnote{See
section \ref{tree-db} for explanation of tree families.}
(subcategorization frame).  Since the tree database and syntactic
lexicon are already separated for space efficiency (see
Chapter~\ref{overview}), each verb can efficiently select a large
number of trees by specifying a tree family, as opposed to each of the
individual trees.  This approach allows for a considerable reduction
in the number of trees that must be specified for any given verb or
form of a verb.

There are currently 52 tree families in the system.\footnote{An
  explanation of the naming convention used in naming the trees and
  tree families is available in Appendix~\ref{tree-naming}.}  This
chapter gives a brief description of each tree family and shows the
corresponding declarative tree\footnote{Before lexicalization, the
  $\diamond$ indicates the anchor of the tree.}, along with any
peculiar characteristics or trees.  It also indicates which
transformations are in each tree family, and gives the number of verbs
that select that family.\footnote{Numbers given are as of August 1998
  and are subject to some change with further development of the
  grammar.} A few sample verbs are given, along with example
sentences.

\section{Intransitive: Tnx0V}\index{verbs, intransitive}
\label{nx0V-family}

\begin{description}
  
\item[Description:] This tree family is selected by verbs that do not
  require an object complement of any type.  Adverbs, prepositional
  phrases and other adjuncts may adjoin on, but are not required for
  the sentences to be grammatical.  1,878 verbs select this family.

\item[Examples:]  {\it eat}, {\it sleep}, {\it dance} \\
{\it Al ate .} \\ 
{\it Seth slept .} \\ 
{\it Hyun danced .}

\item[Declarative tree:]  See Figure~\ref{nx0V-tree}.

\begin{figure}[htb]
\centering
\begin{tabular}{c}
\psfig{figure=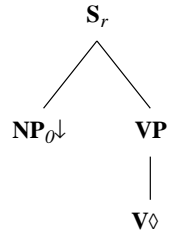,height=3.4cm}
\end{tabular}
\caption{Declarative Intransitive Tree:  $\alpha$nx0V}
\label{nx0V-tree}
\end{figure}

\item[Other available trees:] wh-moved subject, 
subject relative clause with and without comp, 
adjunct (gap-less) relative clause with comp,
adjunct (gap-less) relative clause with PP pied-piping,
imperative, determiner
gerund, NP gerund, pre-nominal participal.

\end{description}

\section{Transitive: Tnx0Vnx1}\index{verbs,transitive}
\label{nx0Vnx1-family}

\begin{description}
  
\item[Description:] This tree family is selected by verbs that require
  only an NP object complement.  The NP's may be complex structures,
  including gerund NP's and NP's that take sentential complements.
  This does not include light verb constructions (see
  sections~\ref{nx0lVN1-family} and \ref{nx0lVN1Pnx2-family}).  4,343
  verbs select the transitive tree family.

\item[Examples:] {\it eat}, {\it dance}, {\it take}, {\it like}\\
{\it Al ate an apple .} \\ 
{\it Seth danced the tango .} \\ 
{\it Hyun is taking an algorithms course .} \\
{\it Anoop likes the fact that the semester is finished .}

\item[Declarative tree:] See Figure~\ref{nx0Vnx1-tree}.

\begin{figure}[htb]
\centering
\begin{tabular}{c}
\psfig{figure=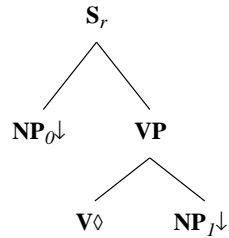,height=3.4cm}
\end{tabular}
\caption{Declarative Transitive Tree:  $\alpha$nx0Vnx1}
\label{nx0Vnx1-tree}
\end{figure}

\item[Other available trees:] wh-moved subject, wh-moved object, subject
relative clause with and without comp, adjunct (gap-less) relative clause
with comp/with PP pied-piping, object relative
clause with and without comp, imperative, determiner gerund, NP gerund, passive with {\it
by} phrase, passive without {\it by} phrase, passive with wh-moved
subject and {\it by} phrase, passive with wh-moved subject and no {\it
by} phrase, passive with wh-moved object out of the {\it by} phrase,
passive with wh-moved {\it by} phrase, passive with relative clause on
subject and {\it by} phrase with and without comp, passive with relative clause on subject
and no {\it by} phrase with and without comp, passive with relative clause on object on the
{\it by} phrase with and without comp/with PP pied-piping, 
gerund passive with {\it by} phrase, gerund passive
without {\it by} phrase, ergative, ergative with wh-moved subject,
ergative with subject relative clause with and without comp,
ergative with adjunct (gap-less) relative clause 
with comp/with PP pied-piping.  In addition, two other trees
that allow transitive verbs to function as adjectives (e.g. {\it the
stopped truck}) are also in the family.

\end{description}

\section{Ditransitive: Tnx0Vnx1nx2}\index{verbs,ditransitive}
\label{nx0Vnx1nx2-family}

\begin{description}

\item[Description:]  This tree family is selected by verbs that take exactly 
two NP complements.  It does {\bf not} include verbs that undergo the
ditransitive verb shift (see section~\ref{nx0Vnx1Pnx2-family}).  The apparent
ditransitive alternates involving verbs in this class and benefactive PP's
(e.g. {\it John baked a cake for Mary}) are analyzed as transitives (see
section~\ref{nx0Vnx1-family}) with a PP adjunct. Benefactives are
taken to be adjunct PP's because they are optional (e.g. {\it John baked a
cake} vs. {\it John baked a cake for Mary}).  122 verbs select the ditransitive
tree family.

\item[Examples:] {\it ask}, {\it cook}, {\it win} \\
{\it Christy asked Mike a question .} \\ 
{\it Doug cooked his father dinner .} \\
{\it Dania won her sister a stuffed animal .}

\item[Declarative tree:]  See Figure~\ref{nx0Vnx1nx2-tree}.

\begin{figure}[htb]
\centering
\begin{tabular}{c}
\psfig{figure=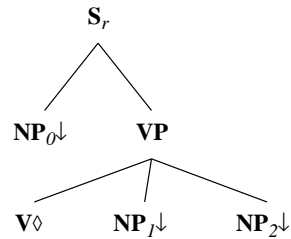,height=3.4cm}
\end{tabular}
\caption{Declarative Ditransitive Tree:  $\alpha$nx0Vnx1nx2}
\label{nx0Vnx1nx2-tree}
\end{figure}

\item[Other available trees:] wh-moved subject, wh-moved direct object,
wh-moved indirect object, subject relative clause with and without comp, 
adjunct (gap-less) relative clause with comp/with PP pied-piping, direct object relative
clause with and without comp, indirect object relative clause with and without comp, 
imperative, determiner gerund, NP
gerund, passive with {\it by} phrase, passive without {\it by} phrase,
passive with wh-moved subject and {\it by} phrase, passive with wh-moved
subject and no {\it by} phrase, passive with wh-moved object out of the
{\it by} phrase, passive with wh-moved {\it by} phrase, passive with
wh-moved indirect object and {\it by} phrase, passive with wh-moved
indirect object and no {\it by} phrase, passive with relative clause on
subject and {\it by} phrase with and without comp, 
passive with relative clause on subject and no
{\it by} phrase with and without comp, passive with relative clause on object of the {\it by}
phrase with and without comp/with PP pied-piping, 
passive with relative clause on the indirect object and {\it by}
phrase with and without comp, passive with relative clause on the indirect object and no {\it by}
phrase with and without comp, passive with/without {\it by}-phrase with adjunct 
(gap-less) relative clause with comp/with PP pied-piping, 
gerund passive with {\it by} phrase, gerund passive without {\it by} phrase.

\end{description}

\section{Ditransitive with PP: Tnx0Vnx1pnx2}\index{verbs, NP with VP verbs}
\label{nx0Vnx1pnx2-family}

\begin{description}

\item[Description:]  This tree family is selected by ditransitive verbs that
take a noun phrase followed by a prepositional phrase.  The
preposition is not constrained in the syntactic lexicon.  The
preposition must be required and not optional - that is, the sentence
must be ungrammatical with just the noun phrase (e.g. {\it $\ast$John
put the table}).  No verbs, therefore, should select both this tree
family and the transitive tree family (see
section~\ref{nx0Vnx1-family}).  This tree family is also distinguished
from the ditransitive verbs, such as {\it give}, that undergo verb
shifting (see section~\ref{nx0Vnx1Pnx2-family}).  There are 62 verbs
that select this tree family.

\item[Examples:] {\it associate}, {\it put}, {\it refer} \\
{\it Rostenkowski associated money with power .}   \\
{\it He put his reputation on the line .}  \\
{\it He referred all questions to his attorney .}

\item[Declarative tree:]  See Figure~\ref{nx0Vnx1pnx2-tree}.

\begin{figure}[htb]
\centering
\begin{tabular}{c}
\psfig{figure=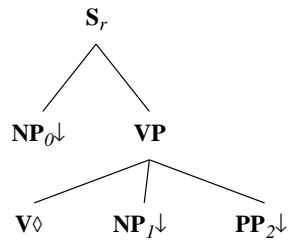,height=3.4cm}
\end{tabular}
\caption{Declarative Ditransitive with PP Tree:  $\alpha$nx0Vnx1pnx2}
\label{nx0Vnx1pnx2-tree}
\end{figure}

\item[Other available trees:] wh-moved subject, wh-moved direct object,
wh-moved object of PP, wh-moved PP, subject relative clause with and without comp, 
adjunct (gap-less) relative clause with comp/with PP pied-piping, direct object
relative clause with and without comp, object of PP relative clause with and 
without comp/with PP pied-piping, imperative, determiner
gerund, NP gerund, passive with {\it by} phrase, passive without {\it by}
phrase, passive with wh-moved subject and {\it by} phrase, passive with
wh-moved subject and no {\it by} phrase, passive with wh-moved object out
of the {\it by} phrase, passive with wh-moved {\it by} phrase, passive with
wh-moved object out of the PP and {\it by} phrase, passive with wh-moved
object out of the PP and no {\it by} phrase, passive with wh-moved PP and
{\it by} phrase, passive with wh-moved PP and no {\it by} phrase, passive
with relative clause on subject and {\it by} phrase with and without comp, 
passive with relative
clause on subject and no {\it by} phrase with and without comp, passive with relative clause on
object of the {\it by} phrase with and without comp/with PP pied-piping, 
passive with relative clause on the object
of the PP and {\it by} phrase with and without comp/with PP pied-piping, 
passive with relative clause on the object
of the PP and no {\it by} phrase with and without comp/with PP pied-piping, 
passive with and without {\it by} phrase with adjunct (gap-less) relative clause
with comp/with PP pied-piping,
gerund passive with {\it by} phrase,
gerund passive without {\it by} phrase.

\end{description}

\section{Ditransitive with PP shift: Tnx0Vnx1tonx2}\index{verbs,ditransitive
with PP shift}
\label{nx0Vnx1Pnx2-family}

\begin{description}

\item[Description:]  This tree family is selected by ditransitive verbs that
undergo a shift to a {\it to} prepositional phrase.  These ditransitive verbs
are clearly constrained so that when they shift, the prepositional phrase must
start with {\it to}.  This is in contrast to the Ditransitives with PP in
section~\ref{nx0Vnx1pnx2-family}, in which verbs may appear in [NP V NP PP]
constructions with a variety of prepositions.  Both the dative shifted and
non-shifted PP complement trees are included.  56 verbs select this family.

\item[Examples:] {\it give}, {\it promise}, {\it tell} \\
{\it Bill gave Hillary flowers .} \\ 
{\it Bill gave flowers to Hillary .} \\
{\it Whitman promised the voters a tax cut .} \\
{\it Whitman promised a tax cut to the voters .} \\
{\it Pinnochino told Gepetto a lie .} \\
{\it Pinnochino told a lie to Gepetto .}

\item[Declarative tree:]  See Figure~\ref{nx0Vnx1Pnx2-tree}.

\begin{figure}[htb]
\centering
\begin{tabular}{ccc}
\psfig{figure=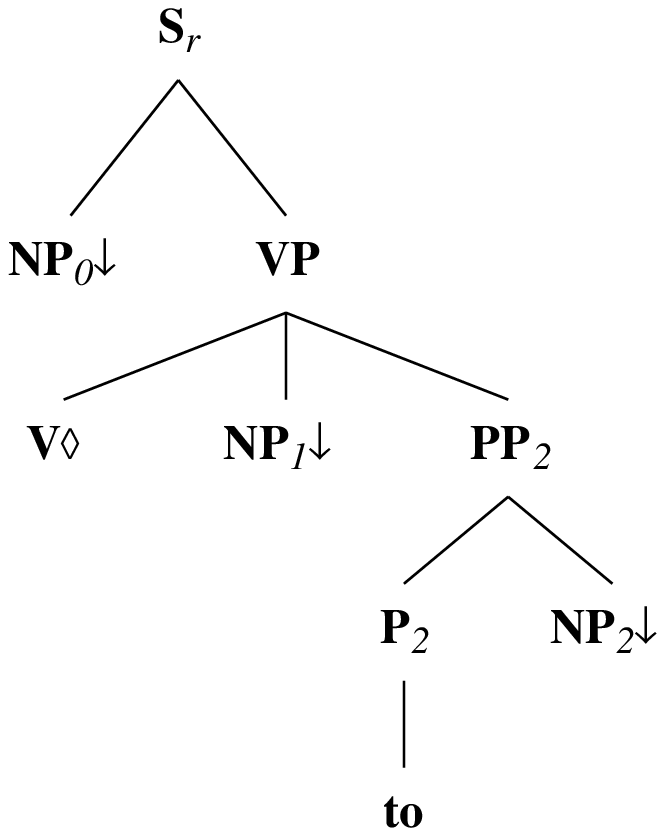,height=5.2cm} &
\hspace{1.0in}&
\psfig{figure=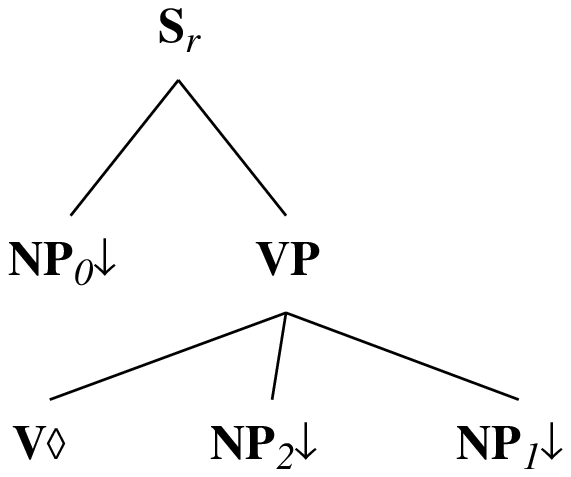,height=3.3cm} \\
(a) && (b)
\end{tabular}
\caption{Declarative Ditransitive with PP shift Trees: $\alpha$nx0Vnx1Pnx2~(a)
and $\alpha$nx0Vnx2nx1~(b)}
\label{nx0Vnx1Pnx2-tree}
\end{figure}

\item[Other available trees:] {\bf Non-shifted:} wh-moved subject, wh-moved
direct object, wh-moved indirect object, subject relative clause with and without comp, 
adjunct (gap-less) relative clause with comp/with PP pied-piping, direct
object relative clause with comp/with PP pied-piping, indirect object relative clause
with and without comp/with PP pied-piping, imperative, NP
gerund, passive with {\it by} phrase, passive without {\it by} phrase,
passive with wh-moved subject and {\it by} phrase, passive with wh-moved
subject and no {\it by} phrase, passive with wh-moved object out of the
{\it by} phrase, passive with wh-moved {\it by} phrase, passive with
wh-moved indirect object and {\it by} phrase, passive with wh-moved
indirect object and no {\it by} phrase, passive with relative clause on
subject and {\it by} phrase with and without comp, 
passive with relative clause on subject and no
{\it by} phrase with and without comp, passive with relative clause on object of the {\it by}
phrase with and without comp/with PP pied-piping, 
passive with relative clause on the indirect object and {\it by}
phrase with and without comp/with PP pied-piping, 
passive with relative clause on the indirect object and no {\it by}
phrase with and without comp/with PP pied-piping, 
passive with/without {\it by}-phrase with adjunct (gap-less) relative clause
with comp/with PP pied-piping,
gerund passive with {\it by} phrase, gerund passive without {\it
by} phrase;\\ 
{\bf Shifted:} wh-moved subject, wh-moved direct object,
wh-moved object of PP, wh-moved PP, subject relative clause with and without comp, 
adjunct (gap-less) relative clause with comp/with PP pied-piping, direct object
relative clause with comp/with PP pied-piping, object of PP relative clause with and without 
comp/with PP pied-piping, imperative, determiner
gerund, NP gerund, passive with {\it by} phrase, passive without {\it by}
phrase, passive with wh-moved subject and {\it by} phrase, passive with
wh-moved subject and no {\it by} phrase, passive with wh-moved object out
of the {\it by} phrase, passive with wh-moved {\it by} phrase, passive with
wh-moved object out of the PP and {\it by} phrase, passive with wh-moved
object out of the PP and no {\it by} phrase, passive with wh-moved PP and
{\it by} phrase, passive with wh-moved PP and no {\it by} phrase, passive
with relative clause on subject and {\it by} phrase with and without comp, passive with relative
clause on subject and no {\it by} phrase with and without comp, passive with relative clause on
object of the {\it by} phrase with and without comp/with PP pied-piping, 
passive with relative clause on the object
of the PP and {\it by} phrase with and without comp/with PP pied-piping, 
passive with relative clause on the object
of the PP and no {\it by} phrase with and without comp/with PP pied-piping, 
passive with/without {\it by}-phrase with adjunct (gap-less) relative clause
with comp/with PP pied-piping, gerund passive with {\it by} phrase,
gerund passive without {\it by} phrase.

\end{description}

\section{Sentential Complement with NP: Tnx0Vnx1s2}\index{verbs,Sentential
Complement with NP} 
\label{nx0Vnx1s2-family}

\begin{description}
  
\item[Description:] This tree family is selected by verbs that take
  both an NP and a sentential complement.  The sentential complement
  may be infinitive or indicative.  The type of clause is specified by
  each individual verb in its syntactic lexicon entry.  A given verb
  may select more than one type of sentential complement.  The
  declarative tree, and many other trees in this family, are auxiliary
  trees, as opposed to the more common initial trees.  These auxiliary
  trees adjoin onto an S node in an existing tree of the type
  specified by the sentential complement.  This is the mechanism by
  which TAGs are able to maintain long-distance dependencies (see
  Chapter~\ref{extraction}), even over multiple embeddings (e.g. {\it
    What did Bill tell Mary that John said?}).  79 verbs select this
  tree family.

\item[Examples:] {\it beg}, {\it expect}, {\it tell} \\
{\it Srini begged Mark to increase his disk quota .} \\
{\it Beth told Jim that it was his turn .}

\item[Declarative tree:]  See Figure~\ref{nx0Vnx1s2-tree}.

\begin{figure}[htb]
\centering
\begin{tabular}{c}
\psfig{figure=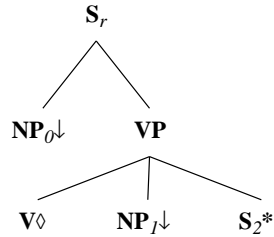,height=3.4cm}
\end{tabular}
\caption{Declarative Sentential Complement with NP Tree:  $\beta$nx0Vnx1s2}
\label{nx0Vnx1s2-tree}
\end{figure}

\item[Other available trees:] wh-moved subject, wh-moved object, wh-moved
sentential complement, subject relative clause with and without comp, 
adjunct (gap-less) relative clause with comp/with PP pied-piping, 
object relative clause with and without comp,
imperative, determiner gerund, NP gerund, passive with {\it by} phrase
before sentential complement, passive with {\it by} phrase after sentential
complement, passive without {\it by} phrase, passive with wh-moved subject
and {\it by} phrase before sentential complement, passive with wh-moved
subject and {\it by} phrase after sentential complement, passive with
wh-moved subject and no {\it by} phrase, passive with wh-moved object out
of the {\it by} phrase, passive with wh-moved {\it by} phrase, passive with
relative clause on subject and {\it by} phrase before sentential
complement with and without comp, passive with relative clause on subject and {\it by} phrase
after sentential complement with and without comp, 
passive with relative clause on subject and no {\it by} phrase with and without comp,
passive with/without {\it by}-phrase with adjunct (gap-less) relative clause
with comp/with PP pied-piping, gerund passive with {\it by} phrase befor sentential
complement, gerund passive with {\it by} phrase after the sentential
complement, gerund passive without {\it by} phrase, parenthetical
reporting clause.

\end{description}

\section{Intransitive Verb Particle: Tnx0Vpl}\index{verbs,verb-particle,intransitive}
\label{nx0Vpl}

\begin{description}

\item[Description:]  The trees in this tree family are anchored by both the
verb and the verb particle.  Both appear in the syntactic lexicon and together
select this tree family.  Intransitive verb particles can be difficult to
distinguish from intransitive verbs with adverbs adjoined on. The main
diagnostics for including verbs in this class are whether the meaning is
compositional or not, and whether there is a transitive version of the
verb/verb particle combination with the same or similar meaning.  The existence
of an alternate compositional meaning is a strong indication for a separate
verb particle construction.  There are 159 verb/verb particle combinations.

\item[Examples:] {\it add up}, {\it come out}, {\it sign off} \\
{\it The numbers never quite added up .} \\
{\it John finally came out (of the closet) .} \\
{\it I think that I will sign off now .}

\item[Declarative tree:]  See Figure~\ref{nx0Vpl-tree}.

\begin{figure}[htb]
\centering
\begin{tabular}{c}
\psfig{figure=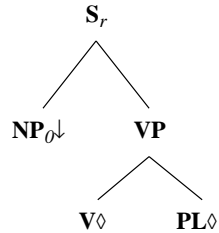,height=3.4cm}
\end{tabular}
\caption{Declarative Intransitive Verb Particle Tree:  $\alpha$nx0Vpl}
\label{nx0Vpl-tree}
\end{figure}

\item[Other available trees:] wh-moved subject, subject relative clause with and without comp, 
adjunct (gap-less) relative clause with comp/with PP pied-piping, 
imperative, determiner gerund, NP gerund.

\end{description}

\section{Transitive Verb Particle: Tnx0Vplnx1}\index{verbs,particle,transitive}
\label{nx0Vplnx1-family}

\begin{description}
  
\item[Description:] Verb/verb particle combinations that take an NP
  complement select this tree family.  Both the verb and the verb
  particle are anchors of the trees. Particle movement has been taken
  as the diagnostic to distinguish verb particle constructions from
  intransitives with adjoined PP's.  If the alleged particle is able
  to undergo particle movement, in other words appear both before and
  after the direct object, then it is judged to be a particle.  Items
  that do not undergo particle movement are taken to be prepositions.
  In many, but not all, of the verb particle cases, there is also an
  alternate prepositional meaning in which the lexical item did not
  move.  (e.g. {\it He looked up the number (in the phonebook).  He
    looked the number up. Srini looked up the road (for Purnima's
    car).  $\ast$He looked the road up.})  There are 489 verb/verb
  particle combinations.

\item[Examples:] {\it blow off}, {\it make up}, {\it pick out} \\
{\it He blew off his linguistics class for the third time .} \\
{\it He blew his linguistics class off for the third time .} \\
{\it The dyslexic leprechaun made up the syntactic lexicon .} \\
{\it The dyslexic leprechaun made the syntactic lexicon up .} \\
{\it I would like to pick out a new computer .} \\
{\it I would like to pick a new computer out .} 

\item[Declarative tree:]  See Figure~\ref{nx0Vplnx1-tree}.

\begin{figure}[htb]
\centering
\begin{tabular}{ccc}
\psfig{figure=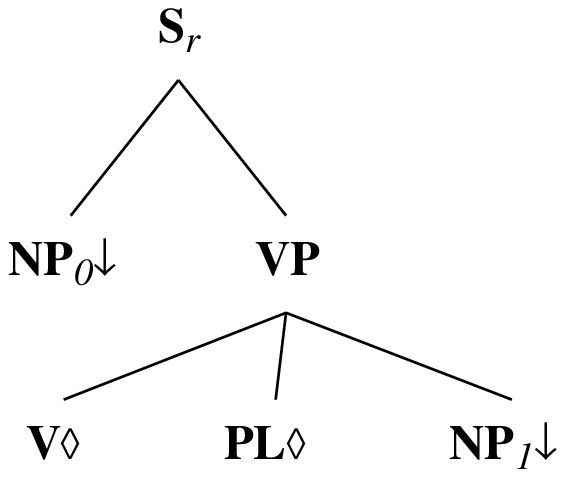,height=3.4cm} &
\hspace{1.0in}&
\psfig{figure=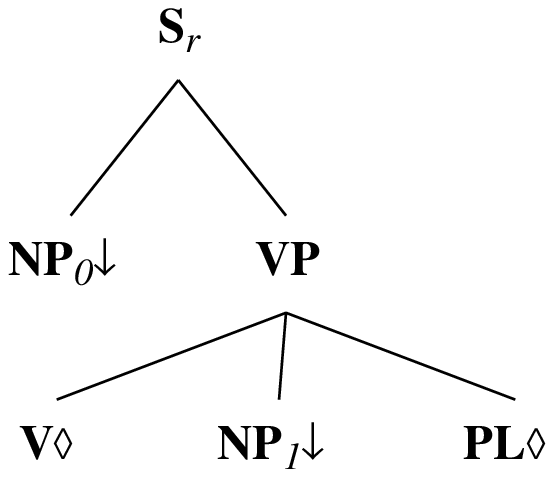,height=3.4cm} \\
(a)&&(b)
\end{tabular}
\caption{Declarative Transitive Verb Particle Tree: $\alpha$nx0Vplnx1~(a) and
$\alpha$nx0Vnx1pl~(b)}
\label{nx0Vplnx1-tree}
\end{figure}

\item[Other available trees:] wh-moved subject with particle before the NP,
wh-moved subject with particle after the NP, wh-moved object, subject
relative clause with particle before the NP with and without comp, subject relative clause with
particle after the NP with and without comp, object relative clause with and without comp, 
adjunct (gap-less) relative clause with particle before the NP with comp/with PP pied-piping,
adjunct (gap-less) relative clause with particle after the NP with comp/with PP pied-piping,
imperative with particle before the NP, imperative with particle after the NP, determiner gerund
with particle before the NP, NP gerund with particle before the NP, NP
gerund with particle after the NP, passive with {\it by} phrase, passive
without {\it by} phrase, passive with wh-moved subject and {\it by} phrase,
passive with wh-moved subject and no {\it by} phrase, passive with wh-moved
object out of the {\it by} phrase, passive with wh-moved {\it by} phrase,
passive with relative clause on subject and {\it by} phrase with and without comp, passive with
relative clause on subject and no {\it by} phrase with and without comp, passive with relative
clause on object of the {\it by} phrase with and without comp/with PP pied-piping, 
passive with/without {\it by}-phrase with adjunct (gap-less) relative clause
with comp/with PP pied-piping, gerund passive with {\it by}
phrase, gerund passive without {\it by} phrase.

\end{description}

\section{Ditransitive Verb Particle: Tnx0Vplnx1nx2}\index{verbs,particle,ditransitive}
\label{nx0Vplnx1nx2}

\begin{description}

\item[Description:]  Verb/verb particle combinations that select this tree
family take 2 NP complements.  Both the verb and the verb particle anchor the
trees, and the verb particle can occur before, between, or after the noun
phrases.  Perhaps because of the complexity of the sentence, these verbs do not
seem to have passive alternations ({\it $\ast$A new bank account was opened up
Michelle by me}).  There are 4 verb/verb particle combinations that select
this tree family.  The exhaustive list is given in the examples.

\item[Examples:] {\it dish out}, {\it open up}, {\it pay off}, {\it rustle up}
\\
{\it I opened up Michelle a new bank account .} \\
{\it I opened Michelle up a new bank account .} \\
{\it I opened Michelle a new bank account up .}

\item[Declarative tree:]  See Figure~\ref{nx0Vplnx1nx2-tree}.

\begin{figure}[htb]
\centering
\begin{tabular}{ccc}
\psfig{figure=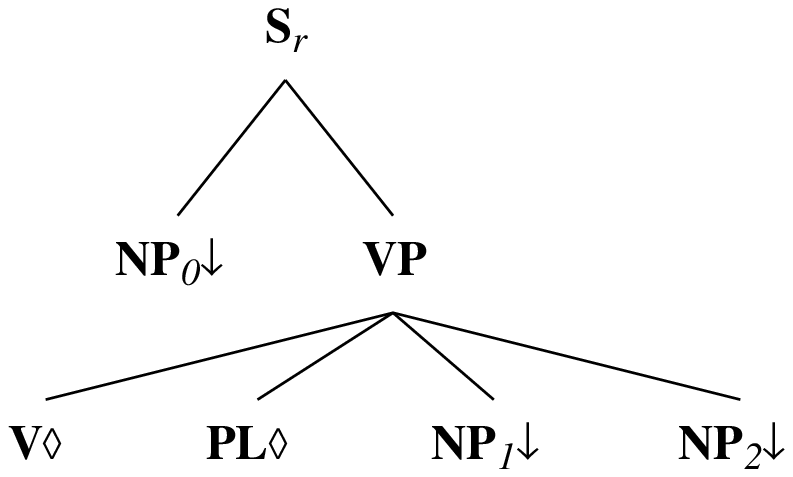,height=3.0cm} &
\psfig{figure=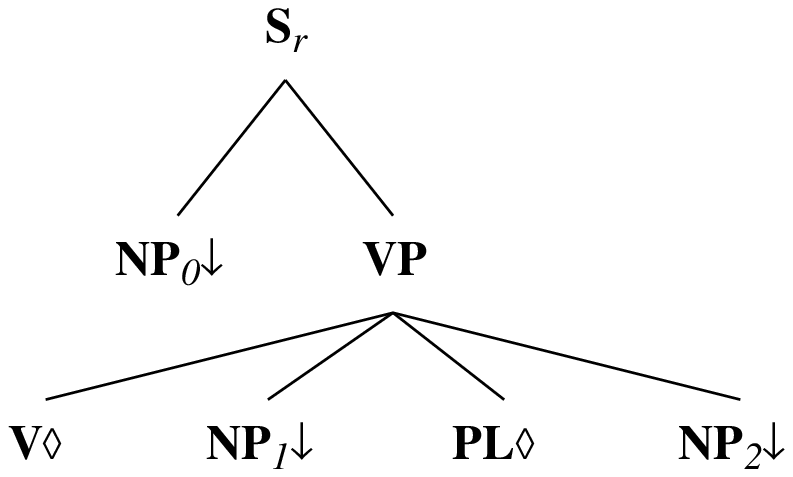,height=3.0cm} &
\psfig{figure=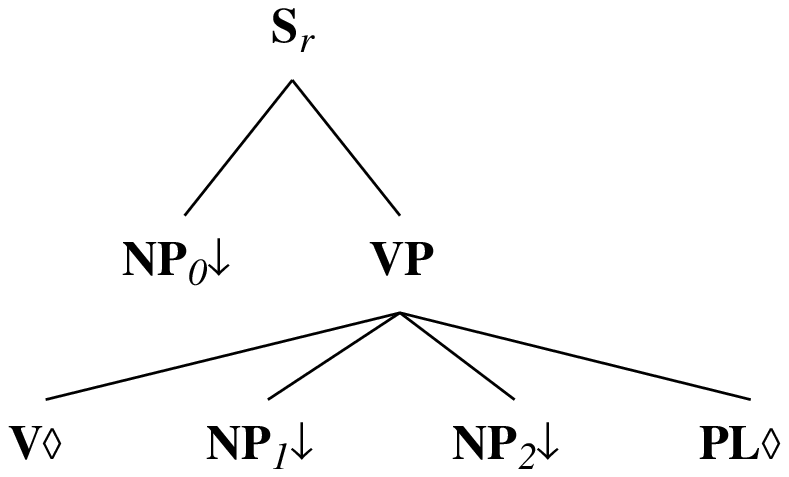,height=3.0cm} \\
(a) & (b)  & (c)
\end{tabular}
\caption{Declarative Ditransitive Verb Particle Tree: $\alpha$nx0Vplnx1nx2~(a),
$\alpha$nx0Vnx1plnx2~(b) and $\alpha$nx0Vnx1nx2pl~(c)}
\label{nx0Vplnx1nx2-tree}
\end{figure}

\item[Other available trees:] wh-moved subject with particle before the NP's,
wh-moved subject with particle between the NP's, wh-moved subject with particle
after the NP's, wh-moved indirect object with particle before the NP's,
wh-moved indirect object with particle after the NP's, wh-moved direct object
with particle before the NP's, wh-moved direct object with particle between the
NP's, subject relative clause with particle before the NP's with and without comp, 
subject relative clause with particle between the NP's with and without comp, 
subject relative clause with particle after the NP's with and without comp, 
indirect object relative clause with particle before the NP's with and without comp,
indirect object relative clause with particle after the NP's with and without comp, 
direct object relative clause with particle before the NP's with and without comp, 
direct object relative clause with particle between the NP's with and without comp, 
adjunct (gap-less) relative clause with comp/with PP pied-piping,
imperative with particle before the NP's,
imperative with particle between the NP's, imperative with particle after the
NP's, determiner gerund with particle before the NP's, NP gerund with particle
before the NP's, NP gerund with particle between the NP's, NP gerund with
particle after the NP's.

\end{description}

\section{Intransitive with PP: Tnx0Vpnx1}\index{verbs,intransitive with PP}
\label{nx0Vpnx1-family}
\begin{description}

\item[Description:]  The verbs that select this tree family are not strictly 
intransitive, in that they {\bf must} be followed by a prepositional phrase.
Verbs that are intransitive and simply {\bf can} be followed by a prepositional
phrase do not select this family, but instead have the PP adjoin onto the
intransitive sentence.  Accordingly, there should be no verbs in both this
class and the intransitive tree family (see section~\ref{nx0V-family}).  The
prepositional phrase is not restricted to being headed by any particular
lexical item.  Note that these are not transitive verb particles (see
section~\ref{nx0Vplnx1-family}), since the head of the PP does not move.  169
verbs select this tree family.

\item[Examples:] {\it grab}, {\it impinge}, {\it provide} \\
{\it Seth grabbed for the brass ring .} \\
{\it The noise gradually impinged on Dania's thoughts .} \\
{\it A good host provides for everyone's needs .}

\item[Declarative tree:]  See Figure~\ref{nx0Vpnx1-tree}.

\begin{figure}[htb]
\centering
\begin{tabular}{c}
\psfig{figure=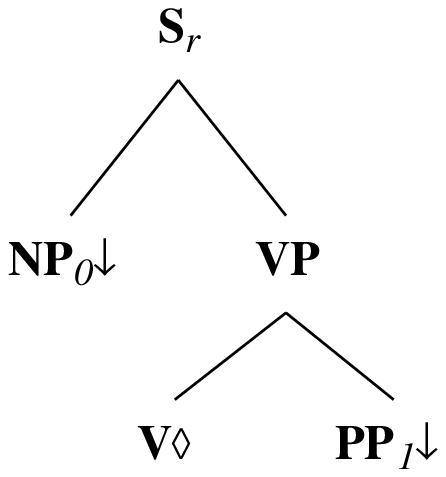,height=1.6in}
\end{tabular}
\caption{Declarative Intransitive with PP Tree:  $\alpha$nx0Vpnx1}
\label{nx0Vpnx1-tree}
\end{figure}

\item[Other available trees:] wh-moved subject, wh-moved object of the PP,
wh-moved PP, subject relative clause with and without comp, 
adjunct (gap-less) relative clause with comp/with PP pied-piping, 
object of the PP relative clause with and without comp/with PP pied-piping,
imperative, determiner gerund, NP gerund, passive with {\it by} phrase,
passive without {\it by} phrase, passive with wh-moved subject and {\it by}
phrase, passive with wh-moved subject and no {\it by} phrase, passive with
wh-moved {\it by} phrase, passive with relative clause on subject and {\it
by} phrase with and without comp, passive with relative clause on subject and no 
{\it by} phrase with and without comp,
passive with relative clause on object of the {\it by} phrase with and without comp/with PP 
pied-piping, passive with/without {\it by}-phrase with adjunct (gap-less) relative clause
with comp/with PP pied-piping, gerund
passive with {\it by} phrase, gerund passive without {\it by} phrase.

\end{description}

\section{Predicative Multi-word with Verb, Prep anchors: Tnx0VPnx1}\label{verbs,prepositional complement} 
\label{nx0VPnx1-family}

\begin{description}

\item[Description:] This tree family is selected by multiple anchor
verb/preposition pairs which together have a non-compositional
interpretation.  For example, {\it think of} has the non-compositional
interpretion involving the inception of a notion or mental entity in
addition to the interpretion in which the agent is thinking about
someone or something.  Anchors for this tree must be able to take both
gerunds and regular NP's in the second noun position.  To allow
adverbs to appear between the verb and the preposition, the trees
contain an extra VP level.  Several of the verbs which select the
Tnx0Vpnx1 family, but which should not have quite the freedom it
allows, will be moving to this family for the next release.  28
verb/preposition pairs select this tree family.

\item[Examples:] {\it think of}, {\it believe in}, {\it depend on} \\
{\it Calvin thought of a new idea .}\\
{\it Hobbes believes in sleeping all day .}\\
{\it Bill depends on drinking coffee for stimulation .}\\

\item[Declarative tree:] See Figure~\ref{nx0VPnx1-tree}.

\begin{figure}[htb]
\centering
\begin{tabular}{c}
\psfig{figure=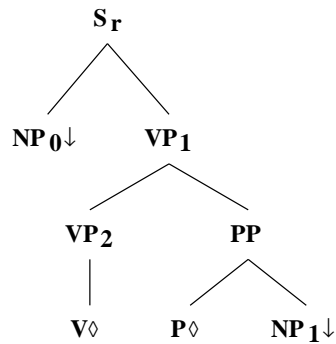,height=4.8cm}
\end{tabular}
\caption{Declarative PP Complement Tree:  $\alpha$nx0VPnx1}
\label{nx0VPnx1-tree}
\end{figure}

\item[Other available trees:] wh-moved subject, wh-moved object, subject
relative clause with and without comp, adjunct (gap-less) relative clause
with comp/with PP pied-piping, object relative
clause with and without comp, imperative, determiner gerund, NP gerund, passive with {\it
by} phrase, passive without {\it by} phrase, passive with wh-moved
subject and {\it by} phrase, passive with wh-moved subject and no {\it
by} phrase, passive with wh-moved object out of the {\it by} phrase,
passive with wh-moved {\it by} phrase, passive with relative clause on
subject and {\it by} phrase with and without comp, passive with relative clause on subject
and no {\it by} phrase with and without comp, passive with relative clause on object on the
{\it by} phrase with and without comp/with PP pied-piping, 
passive with/without {\it by}-phrase with adjunct (gap-less) relative clause
with comp/with PP pied-piping, gerund passive with {\it by} phrase, gerund passive
without {\it by} phrase.  In addition, two other trees
that allow transitive verbs to function as adjectives (e.g. {\it the
thought of idea}) are also in the family.

\end{description}

\section{Sentential Complement: Tnx0Vs1}\label{verbs,sentential complement}
\label{nx0Vs1-family}

\begin{description}
  
\item[Description:] This tree family is selected by verbs that take
  just a sentential complement.  The sentential complement may be of
  type infinitive, indicative, or small clause (see
  Chapter~\ref{small-clauses}).  The type of clause is specified by
  each individual verb in its syntactic lexicon entry, and a given
  verb may select more than one type of sentential complement.  The
  declarative tree, and many other trees in this family, are auxiliary
  trees, as opposed to the more common initial trees.  These auxiliary
  trees adjoin onto an S node in an existing tree of the type
  specified by the sentential complement.  This is the mechanism by
  which TAGs are able to maintain long-distance dependencies (see
  Chapter~\ref{extraction}), even over multiple embeddings (e.g. {\it
    What did Bill think that John said?}). 338 verbs select this tree
  family.

\item[Examples:]  {\it consider}, {\it think} \\
{\it Dania considered the algorithm unworkable .}\\
{\it Srini thought that the program was working .} \\

\item[Declarative tree:]  See Figure~\ref{nx0Vs1-tree}.

\begin{figure}[htb]
\centering
\begin{tabular}{c}
\psfig{figure=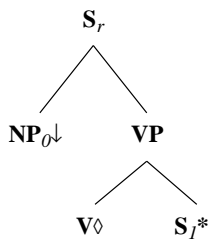,height=3.4cm}
\end{tabular}
\caption{Declarative Sentential Complement Tree:  $\beta$nx0Vs1}
\label{nx0Vs1-tree}
\end{figure}

\item[Other available trees:]  wh-moved subject, wh-moved sentential
complement, subject relative clause with and without comp, adjunct (gap-less) relative
clause with comp/with PP pied-piping, imperative, determiner gerund, NP gerund, parenthetical
reporting clause.

\end{description}

\section{Intransitive with Adjective: Tnx0Vax1}\index{verbs,intransitive with adjective}
\label{nx0Vax1-family}

\begin{description}

\item[Description:]  The verbs that select this tree family take an adjective
as a complement.  The adjective may be regular, comparative, or superlative.
It may also be formed from the special class of adjectives derived from the
transitive verbs (e.g. {\it agitated, broken}).  See
section~\ref{nx0Vnx1-family}).  Unlike the Intransitive with PP verbs (see
section~\ref{nx0Vpnx1-family}), some of these verbs may also occur as bare
intransitives as well.  This distinction is drawn because adjectives do not
normally adjoin onto sentences, as prepositional phrases do.  Other
intransitive verbs can only occur with the adjective, and these select only
this family.  The verb class is also distinguished from the adjective small
clauses (see section~\ref{nx0Ax1-family}) because these verbs are not raising
verbs.  34 verbs select this tree family.

\item[Examples:] {\it become}, {\it grow}, {\it smell} \\
{\it The greenhouse became hotter .} \\
{\it The plants grew tall and strong .} \\
{\it The flowers smelled wonderful .}

\item[Declarative tree:]  See Figure~\ref{nx0Vax1-tree}.

\begin{figure}[htb]
\centering
\begin{tabular}{c}
\psfig{figure=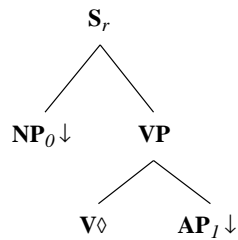,height=3.4cm}
\end{tabular}
\caption{Declarative Intransitive with Adjective Tree:  $\alpha$nx0Vax1}
\label{nx0Vax1-tree}
\end{figure}

\item[Other available trees:]  wh-moved subject, wh-moved adjective 
({\it how}), subject relative clause with and without comp, 
adjunct (gap-less) relative clause with comp/with PP pied-piping, imperative, NP gerund.

\end{description}

\section{Transitive Sentential Subject:  Ts0Vnx1}\index{verbs,sentential subject}
\label{s0Vnx1-family}

\begin{description}

\item[Description:] The verbs that select this tree family all take sentential
subjects, and are often referred to as `psych' verbs, since they all refer to
some psychological state of mind.  The sentential subject can be indicative
(complementizer required) or infinitive (complementizer optional).
 100 verbs that select this tree family.

\item[Examples:] {\it delight}, {\it impress}, {\it surprise} \\
{\it that the tea had rosehips in it delighted Christy .} \\
{\it to even attempt a marathon impressed Dania .} \\
{\it For Jim to have walked the dogs surprised Beth .}

\item[Declarative tree:]  See Figure~\ref{s0Vnx1-tree}.

\begin{figure}[htb]
\centering
\begin{tabular}{c}
\psfig{figure=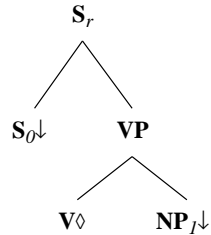,height=3.4cm}
\end{tabular}
\caption{Declarative Sentential Subject Tree:  $\alpha$s0Vnx1}
\label{s0Vnx1-tree}
\end{figure}

\item[Other available trees:]  wh-moved subject, wh-moved object,
subject relative clause with and without comp,
adjunct (gap-less) relative clause with comp/with PP pied-piping.

\end{description}

\section{Light Verbs: Tnx0lVN1}\index{verbs, light}
\label{nx0lVN1-family}

\begin{description}
  
\item[Description:] The verb/noun pairs that select this tree families
  are pairs in which the interpretation is non-compositional and the
  noun contributes argument structure to the predicate (e.g. {\it The
    man took a walk.} vs. {\it The man took a radio}).  The verb and
  the noun occur together in the syntactic database, and both anchor
  the trees.  The verbs in the light verb constructions are {\it do},
  {\it give}, {\it have}, {\it make} and {\it take}.  The noun
  following the light verb is (usually) in a bare infinitive form
  ({\it have a good cry}) and usually occurs with {\it a(n)}.
  However, we include deverbal nominals ({\it take a bath}, {\it give
    a demonstration}) as well.  Constructions with nouns that do not
  contribute an argument structure ({\it have a cigarette}, {\it give}
  NP {\it a black eye}) are excluded.  In addition to semantic
  considerations of light verbs, they differ syntactically from
  Transitive verbs (section~\ref{nx0Vnx1-family}) as well in that the
  noun in the light verb construction does not extract.  Some of the
  verb-noun anchors for this family, like {\it take aim} and {\it take
    hold} disallow determiners, while others require particular
  determiners.  For example, {\it have think} must be indefinite and
  singular, as attested by the ungrammaticality of *{\it John had the
    think/some thinks}.  Another anchor, {\it take leave} can occur
  either bare or with a possesive pronoun (e.g., {\it John took his
    leave}, but not *{\it John took the leave}).  This is accomplished
  through feature specification on the lexical entries.  There are 259
  verb/noun pairs that select the light verb tree.

\item[Examples:] {\it give groan}, {\it have discussion}, {\it make comment} \\
{\it The audience gave a collective groan .} \\
{\it We had a big discussion about closing the libraries .} \\
{\it The professors made comments on the paper .}

\item[Declarative tree:]  See Figure~\ref{nx0lVN1-tree}.

\begin{figure}[htb]
\centering
\begin{tabular}{c}
\psfig{figure=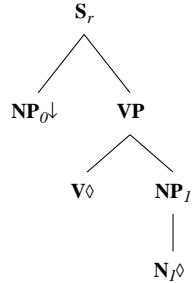,height=4.0cm}\\
\end{tabular}
\caption{Declarative Light Verb Tree: $\alpha$nx0lVN1}
\label{nx0lVN1-tree}
\end{figure}

\item[Other available trees:] wh-moved subject, subject relative clause with and without comp, 
adjunct (gap-less) relative clause with comp/with PP pied-piping,
imperative, determiner gerund, NP gerund.

\end{description}

\section{Ditransitive Light Verbs with PP Shift: Tnx0lVN1Pnx2}\index{verbs,ditransitive light verbs with PP shift}
\label{nx0lVN1Pnx2-family}

\begin{description}

\item[Description:]  The verb/noun pairs that select this tree family are 
pairs in which the interpretation is non-compositional and the noun
contributes argument structure to the predicate (e.g. {\it Dania made
Srini a cake.} vs.  {\it Dania made Srini a loan}).  The verb and the
noun occur together in the syntactic database, and both anchor the
trees.  The verbs in these light verb constructions are {\it give} and
{\it make}.  The noun following the light verb is (usually) a bare
infinitive form (e.g. {\it make a promise to Anoop}).  However, we
include deverbal nominals (e.g. {\it make a payment to Anoop}) as
well.  Constructions with nouns that do not contribute an argument
structure are excluded.  In addition to semantic considerations of
light verbs, they differ syntactically from the Ditransitive with PP
Shift verbs (see section~\ref{nx0Vnx1Pnx2-family}) as well in that the
noun in the light verb construction does not extract.  Also,
passivization is severely restricted.  Special determiner requirments
and restrictions are handled in the same manner as for the Tnx0lVN1
family.  There are 18 verb/noun pairs that select this family.

\item[Examples:] {\it give look}, {\it give wave}, {\it make promise} \\
{\it Dania gave Carl a murderous look .} \\
{\it Amanda gave us a little wave as she left .} \\
{\it Dania made Doug a promise .} 

\item[Declarative tree:]  See Figure~\ref{nx0lVN1Pnx2-tree}.

\begin{figure}[htb]
\centering
\mbox{}
\begin{tabular}{cc}
\psfig{figure=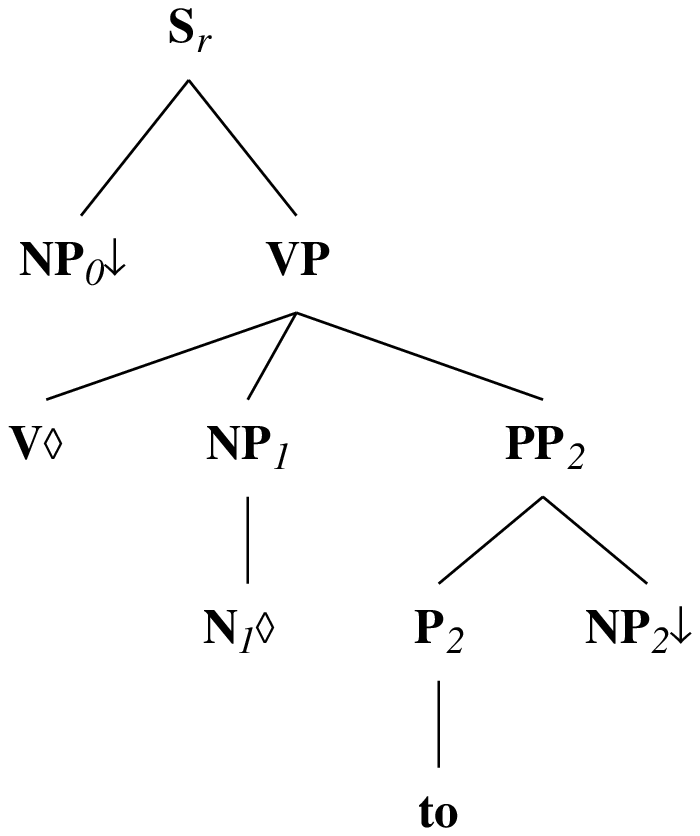,height=5.1cm}
\psfig{figure=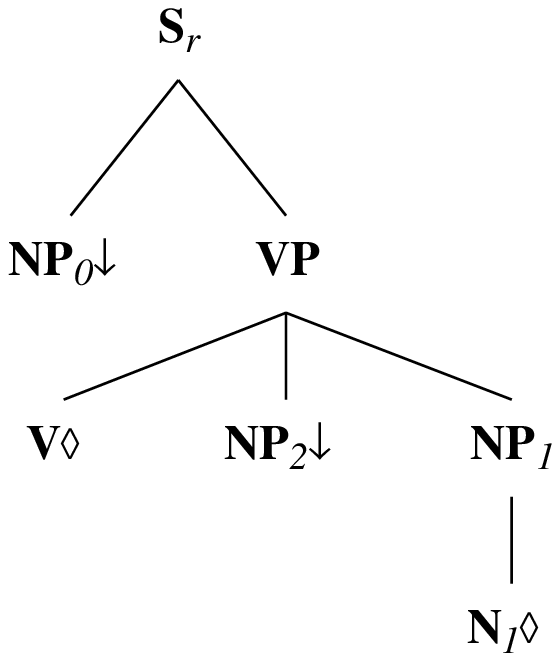,height=5.1cm} \\
(a) & (b) \vspace*{1.2cm}\\
\end{tabular}
\caption{Declarative Light Verbs with PP Tree: $\alpha$nx0lVN1Pnx2~(a),
$\alpha$nx0lVnx2N1~(b)}
\label{nx0lVN1Pnx2-tree}
\end{figure}

\item[Other available trees:] {\bf Non-shifted:} wh-moved subject, wh-moved
indirect object, subject relative clause with and without comp, 
adjunct (gap-less) relative clause with comp/with PP pied-piping, 
indirect object relative clause with and without comp/with PP pied-piping,
imperative, NP gerund, passive with {\it by} phrase, passive with {\it by}-phrase with
adjunct (gap-less) relative clause with comp/with PP pied-piping, gerund passive with
{\it by} phrase, gerund passive without {\it by} phrase \\ 
{\bf Shifted:}
wh-moved subject, wh-moved object of PP, wh-moved PP, subject relative
clause with and without comp, object of PP relative clause with and without comp/with PP 
pied-piping, imperative, determiner gerund, NP
gerund, passive with {\it by} phrase with adjunct (gap-less) relative clause
with comp/with PP pied-piping, gerund passive with {\it by} phrase,
gerund passive without {\it by} phrase.
\end{description}

\section{NP It-Cleft: TItVnx1s2}
\label{ItVnx1s2-family}

\begin{description}

\item[Description:] This tree family is selected by {\it be} as the
main verb and  {\it it} as the subject. Together these two items serve
as a multi-component anchor for the tree family.  This tree family is
used for it-clefts in which the clefted element is an NP and there are
no gaps in the clause which follows the NP.  The NP is interpreted as
an adjunct of the following clause. See Chapter~\ref{it-clefts} for
additional discussion.

\item[Examples:] {\it it be} \\
{\it it was yesterday that we had the meeting .}

\item[Declarative tree:]  See Figure~\ref{ItVnx1s2-tree}.

\begin{figure}[htb]
\centering
\begin{tabular}{c}
\psfig{figure=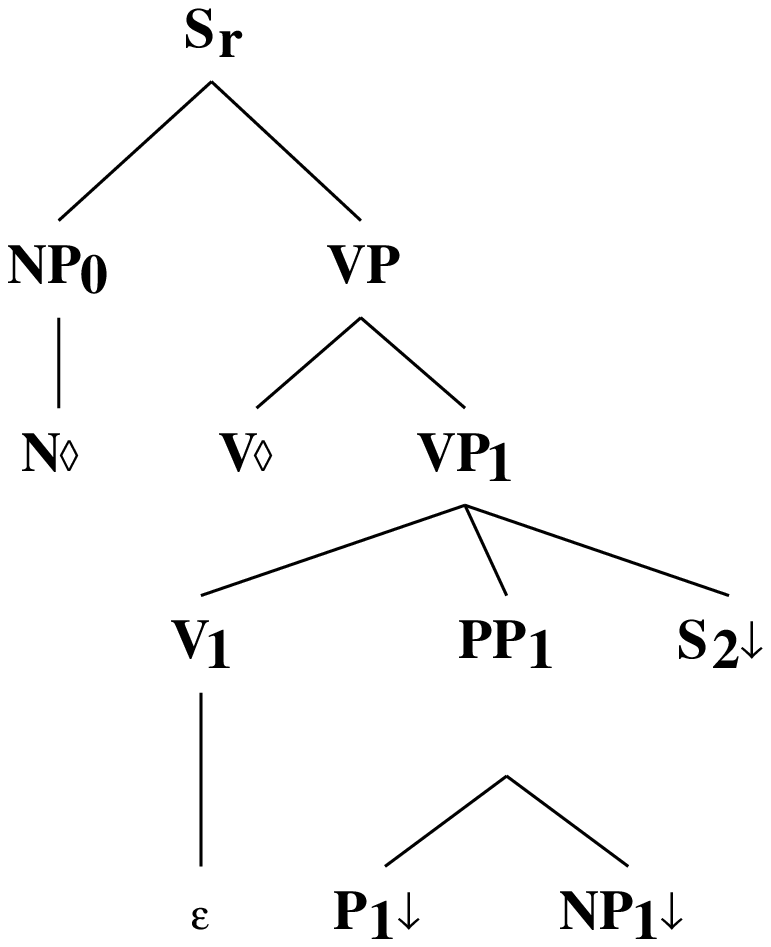,height=4.9cm}
\end{tabular}
\caption{Declarative NP It-Cleft Tree:  $\alpha$ItVpnx1s2}
\label{ItVnx1s2-tree}
\end{figure}

\item[Other available trees:]  inverted question, wh-moved object with
{\it be} inverted, wh-moved object with {\it be} not inverted, 
adjunct (gap-less) relative clause with comp/with PP pied-piping.

\end{description}

\section{PP It-Cleft: TItVpnx1s2}
\label{ItVpnx1s2-family}

\begin{description}
  
\item[Description:] This tree family is selected by {\it be} as the
  main verb and {\it it} as the subject. Together these two items
  serve as a multi-component anchor for the tree family.  This tree
  family is used for it-clefts in which the clefted element is a PP
  and there are no gaps in the clause which follows the PP.  The PP is
  interpreted as an adjunct of the following clause. See
  Chapter~\ref{it-clefts} for additional discussion.

\item[Examples:] {\it it be} \\
{\it it was at Kent State that the police shot all those students .}

\item[Declarative tree:]  See Figure~\ref{ItVpnx1s2-tree}.

\begin{figure}[htb]
\centering
\begin{tabular}{c}
\psfig{figure=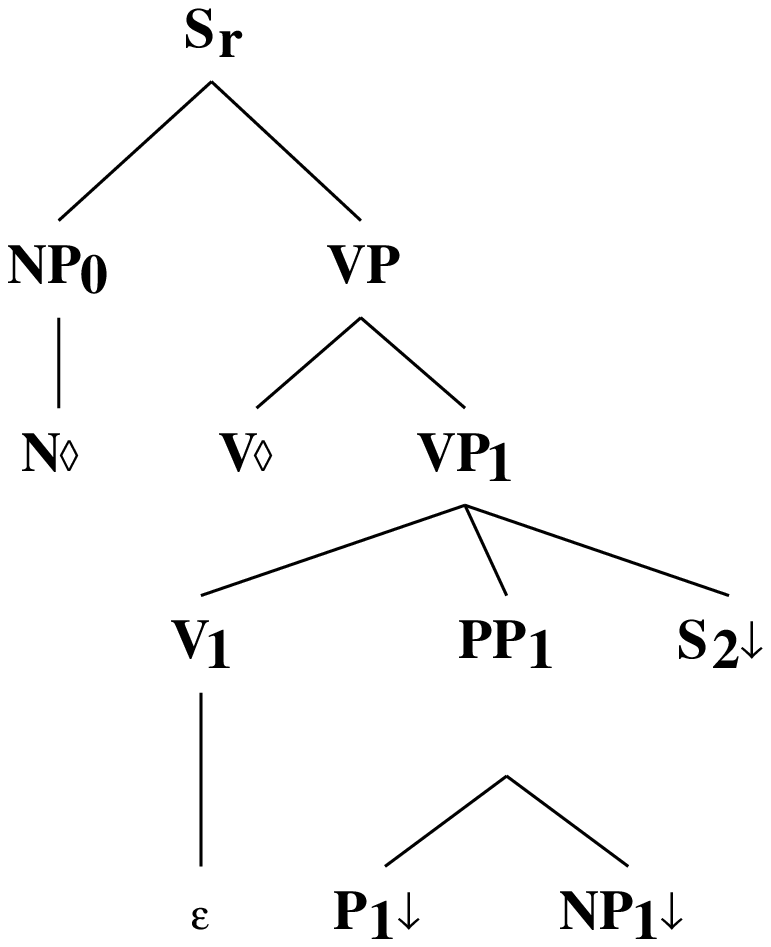,height=5.0cm}
\end{tabular}
\caption{Declarative PP It-Cleft Tree:  $\alpha$ItVnx1s2}
\label{ItVpnx1s2-tree}
\end{figure}

\item[Other available trees:] inverted question, wh-moved prepositional phrase
with {\it be} inverted, wh-moved prepositional phrase with {\it be} not
inverted, adjunct (gap-less) relative clause with comp/with PP pied-piping.

\end{description}

\section{Adverb It-Cleft: TItVad1s2}
\label{ItVad1s2-family}

\begin{description}

\item[Description:]   This tree family is selected by {\it be} as the
main verb and  {\it it} as the subject. Together these two items serve
as a multi-component anchor for the tree family.  This tree family is
used for it-clefts in which the clefted element is an adverb and there are
no gaps in the clause which follows the adverb.  The adverb is interpreted as
an adjunct of the following clause. See Chapter~\ref{it-clefts} for
additional discussion.

\item[Examples:] {\it it be} \\
{\it it was reluctantly that Dania agreed to do the tech report .}

\item[Declarative tree:]  See Figure~\ref{ItVad1s2-tree}.

\begin{figure}[htb]
\centering
\begin{tabular}{c}
\psfig{figure=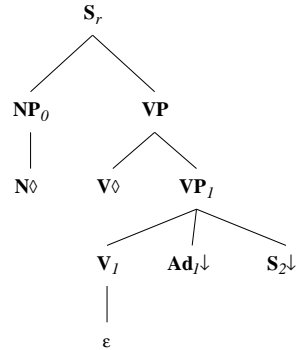,height=4.9cm}
\end{tabular}
\caption{Declarative Adverb It-Cleft Tree:  $\alpha$ItVad1s2}
\label{ItVad1s2-tree}
\end{figure}

\item[Other available trees:]  inverted question, wh-moved adverb {\it how}
with {\it be} inverted, wh-moved adverb {\it how} with {\it be} not
inverted, adjunct (gap-less) relative clause with comp/with PP pied-piping.

\end{description}

\section{Adjective Small Clause Tree: Tnx0Ax1}\index{verbs,small-clause}
\label{nx0Ax1-family}

\begin{description}
  
\item[Description:] These trees are not anchored by verbs, but by
  adjectives.  They are explained in much greater detail in the
  section on small clauses (see
  section~\ref{sm-clause-xtag-analysis}).  This section is presented
  here for completeness.  3244 adjectives select this tree family.

\item[Examples:] {\it addictive}, {\it dangerous}, {\it wary}\\
{\it cigarettes are addictive .} \\
{\it smoking cigarettes is dangerous .} \\
{\it John seems wary of the Surgeon General's warnings .}

\item[Declarative tree:]  See Figure~\ref{nx0Ax1-tree}.

\begin{figure}[htb]
\centering
\begin{tabular}{c}
\psfig{figure=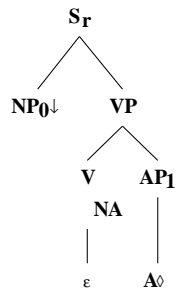,height=4.0cm}
\end{tabular}
\caption{Declarative Adjective Small Clause Tree:  $\alpha$nx0Ax1}
\label{nx0Ax1-tree}
\end{figure}

\item[Other available trees:]  wh-moved subject, wh-moved adjective {\it how},
relative clause on subject with and without comp, imperative, NP gerund,
adjunct (gap-less) relative clause with comp/with PP pied-piping.

\end{description}

\section{Adjective Small Clause with Sentential Complement: Tnx0A1s1}
\label{nx0A1s1-family}

\begin{description}
  
\item[Description:] This tree family is selected by adjectives that
  take sentential complements.  The sentential complements can be
  indicative or infinitive.  Note that these trees are anchored by
  adjectives, not verbs.  Small clauses are explained in much greater
  detail in section~\ref{sm-clause-xtag-analysis}.  This section is
  presented here for completeness.  669 adjectives select this tree
  family.

\item[Examples:] {\it able}, {\it curious}, {\it disappointed} \\
{\it Christy was able to find the problem .} \\
{\it Christy was curious whether the new analysis was working .} \\
{\it Christy was sad that the old analysis failed .} 

\item[Declarative tree:]  See Figure~\ref{nx0A1s1-tree}.

\begin{figure}[htb]
\centering
\begin{tabular}{c}
\psfig{figure=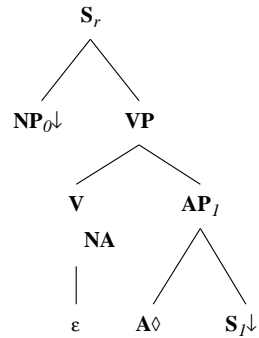,height=4.7cm}
\end{tabular}
\caption{Declarative  Adjective Small Clause with Sentential Complement Tree:  $\alpha$nx0A1s1}
\label{nx0A1s1-tree}
\end{figure}

\item[Other available trees:] wh-moved subject, wh-moved adjective {\it how},
relative clause on subject with and without comp, imperative, NP gerund,
adjunct (gap-less) relative clause with comp/with PP pied-piping.

\end{description}

\section{Adjective Small Clause with Sentential Subject: Ts0Ax1}
\label{s0Ax1-family}

\begin{description}
  
\item[Description:] This tree family is selected by adjectives that
  take sentential subjects.  The sentential subjects can be indicative
  or infinitive.  Note that these trees are anchored by adjectives,
  not verbs.  Most adjectives that take the Adjective Small Clause
  tree family (see section~\ref{nx0Ax1-family}) take this family as
  well.\footnote{No great attempt has been made to go through and
    decide which adjectives should actually take this family and which
    should not.}  Small clauses are explained in much greater detail
  in section~\ref{sm-clause-xtag-analysis}.  This section is presented
  here for completeness.  3,185 adjectives select this tree family.

\item[Examples:] {\it decadent}, {\it incredible}, {\it uncertain} \\
{\it to eat raspberry chocolate truffle ice cream is decadent .} \\
{\it that Carl could eat a large bowl of it is incredible .} \\
{\it whether he will actually survive the experience is uncertain .}

\item[Declarative tree:]  See Figure~\ref{s0Ax1-tree}.

\begin{figure}[htb]
\centering
\begin{tabular}{c}
\psfig{figure=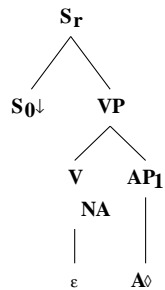,height=4.0cm}
\end{tabular}
\caption{Declarative Adjective Small Clause with Sentential Subject Tree:  $\alpha$s0Ax1}
\label{s0Ax1-tree}
\end{figure}

\item[Other available trees:]  wh-moved subject, wh-moved adjective,
adjunct (gap-less) relative clause with comp/with PP pied-piping.

\end{description}

\section{Equative {\it BE}: Tnx0BEnx1}
\label{nx0BEnx1-family}

\begin{description}

\item[Description:]  This tree family is selected only by the verb {\it be}.
It is distinguished from the predicative NP's (see section~\ref{nx0N1-family})
in that two NP's are equated, and hence interchangeable (see
Chapter~\ref{small-clauses} for more discussion on the English copula and
predicative sentences).  The XTAG analysis for equative {\it be} is explained
in greater detail in section~\ref{equative-be-xtag-analysis}.

\item[Examples:] {\it be} \\
{\it That man is my uncle.}

\item[Declarative tree:]  See Figure~\ref{nx0BEnx1-tree}.

\begin{figure}[htb]
\centering
\begin{tabular}{c}
\psfig{figure=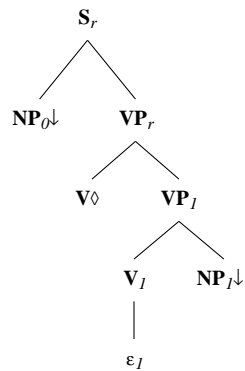,height=5.1cm}
\end{tabular}
\caption{Declarative Equative {\it BE} Tree:  $\alpha$nx0BEnx1}
\label{nx0BEnx1-tree}
\end{figure}

\item[Other available trees:] inverted-question.

\end{description}

\section{NP Small Clause: Tnx0N1}
\label{nx0N1-family}

\begin{description}
  
\item[Description:] The trees in this tree family are not anchored by
  verbs, but by nouns.  Small clauses are explained in much greater
  detail in section~\ref{sm-clause-xtag-analysis}.  This section is
  presented here for completeness.  5,595 nouns select this tree
  family.

\item[Examples:] {\it author}, {\it chair}, {\it dish} \\
{\it Dania is an author .} \\
{\it that blue, warped-looking thing is a chair .} \\
{\it those broken pieces were dishes .}

\item[Declarative tree:]  See Figure~\ref{nx0N1-tree}.

\begin{figure}[htb]
\centering
\begin{tabular}{c}
\psfig{figure=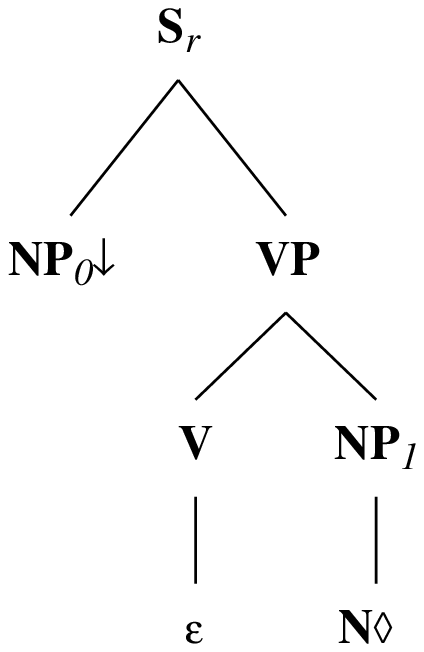,height=4.8cm}
\end{tabular}
\caption{Declarative NP Small Clause Trees: $\alpha$nx0N1}
\label{nx0N1-tree}
\end{figure}

\item[Other available trees:] wh-moved subject, wh-moved object, relative
clause on subject with and without comp, imperative, NP gerund,
adjunct (gap-less) relative clause with comp/with PP pied-piping.

\end{description}

\section{NP Small Clause with Sentential Complement: Tnx0N1s1}
\label{nx0N1s1-family}

\begin{description}

\item[Description:]  This tree family is selected by the small group of nouns
that take sentential complements by themselves (see section~\ref{NPA}).  The
sentential complements can be indicative or infinitive, depending on the noun.
Small clauses in general are explained in much greater detail in the
section~\ref{sm-clause-xtag-analysis}.  This section is presented here for
completeness.  141 nouns select this family.

\item[Examples:] {\it admission}, {\it claim}, {\it vow} \\
{\it The affidavits are admissions that they killed the sheep .} \\
{\it there is always the claim that they were insane .} \\
{\it this is his vow to fight the charges .}

\item[Declarative tree:]  See Figure~\ref{nx0N1s1-tree}.

\begin{figure}[htb]
\centering
\begin{tabular}{c}
\psfig{figure=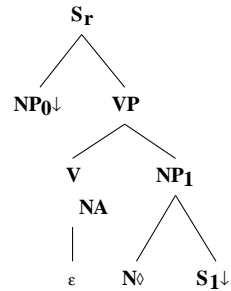,height=4.0cm} 
\end{tabular}
\caption{Declarative NP with Sentential Complement Small Clause Tree:
$\alpha$nx0N1s1}
\label{nx0N1s1-tree}
\end{figure}

\item[Other available trees:] wh-moved subject, wh-moved object, relative
clause on subject with and without comp, imperative, NP gerund,
adjunct (gap-less) relative clause with comp/with PP pied-piping.

\end{description}

\section{NP Small Clause with Sentential Subject:  Ts0N1}
\label{s0N1-family}

\begin{description}

\item[Description:]  This tree family is selected by nouns that take 
sentential subjects.  The sentential subjects can be indicative or infinitive.
Note that these trees are anchored by nouns, not verbs.  Most nouns that take
the NP Small Clause tree family (see section~\ref{nx0N1-family}) take this
family as well.\footnote{No great attempt has been made to go through and
decide which nouns should actually take this family and which should not.}
Small clauses are explained in much greater detail in
section~\ref{sm-clause-xtag-analysis}.  This section is presented here for
completeness.  5,519 nouns select this tree family.

\item[Examples:] {\it dilemma}, {\it insanity}, {\it tragedy} \\
{\it whether to keep the job he hates is a dilemma .} \\
{\it to invest all of your money in worms is insanity .} \\
{\it that the worms died is a tragedy .}

\item[Declarative tree:]  See Figure~\ref{s0N1-tree}.

\begin{figure}[htb]
\centering
\begin{tabular}{c}
\psfig{figure=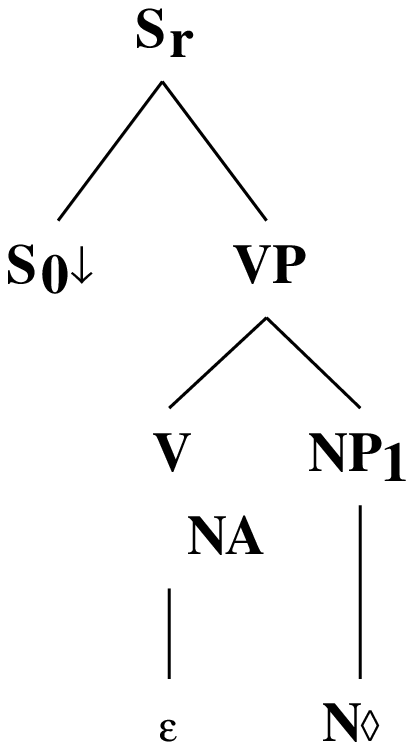,height=4.0cm} 
\end{tabular}
\caption{Declarative NP Small Clause with Sentential Subject Tree: $\alpha$s0N1}
\label{s0N1-tree}
\end{figure}

\item[Other available trees:]  wh-moved subject, adjunct (gap-less) relative clause 
with comp/with PP pied-piping.

\end{description}

\section{PP Small Clause: Tnx0Pnx1}
\label{nx0Pnx1-family}

\begin{description}

\item[Description:]  This family is selected by prepositions that can occur in
small clause constructions.  For more information on small clause
constructions, see section~\ref{sm-clause-xtag-analysis}.  This section is
presented here for completeness.  39 prepositions select this tree family.

\item[Examples:] {\it around}, {\it in}, {\it underneath} \\
{\it Chris is around the corner .} \\
{\it Trisha is in big trouble .} \\
{\it The dog is underneath the table .}

\item[Declarative tree:]  See Figure~\ref{nx0Pnx1-tree}.

\begin{figure}[htb]
\centering
\begin{tabular}{c}
\psfig{figure=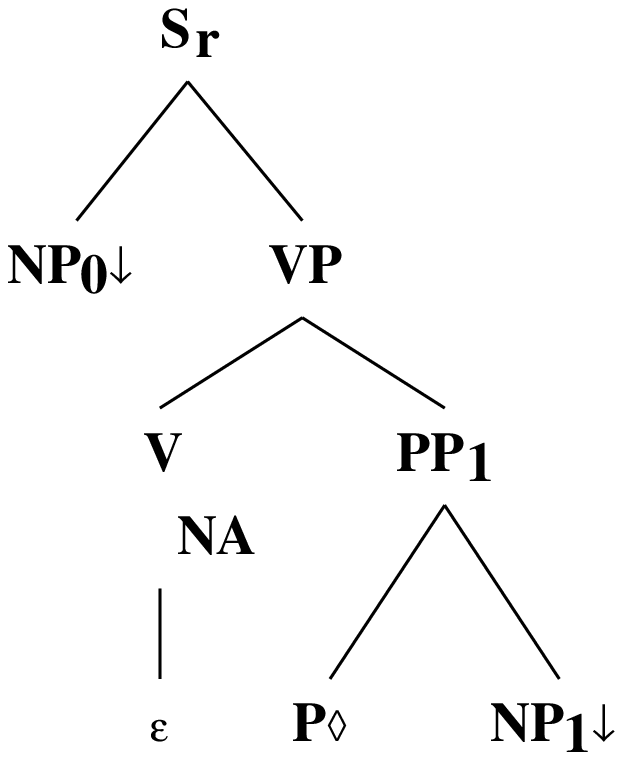,height=4.0cm}
\end{tabular}
\caption{Declarative PP Small Clause  Tree:  $\alpha$nx0Pnx1}
\label{nx0Pnx1-tree}
\end{figure}

\item[Other available trees:]  wh-moved subject, wh-moved object of PP, 
relative clause on subject with and without comp, relative clause on object of PP
with and without comp/with PP pied-piping, imperative, NP gerund,
adjunct (gap-less) relative clause with comp/with PP pied-piping.

\end{description}

\section{Exhaustive PP Small Clause: Tnx0Px1}
\label{nx0Px1-family}

\begin{description}

\item[Description:] This family is selected by {\bf exhaustive} prepositions
that can occur in small clauses.  Exhaustive prepositions are prepositions that
function as prepositional phrases by themselves.  For more information on small
clause constructions, please see section~\ref{sm-clause-xtag-analysis}.  The
section is included here for completeness.  33 exhaustive prepositions select
this tree family.

\item[Examples:] {\it abroad}, {\it below}, {\it outside} \\
{\it Dr. Joshi is abroad .} \\
{\it The workers are all below .} \\
{\it Clove is outside .}

\item[Declarative tree:]  See Figure~\ref{nx0Px1-tree}.

\begin{figure}[htb]
\centering
\begin{tabular}{c}
\psfig{figure=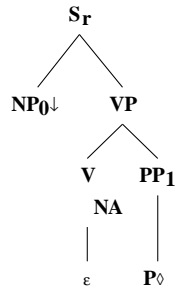,height=4.0cm}
\end{tabular}
\caption{Declarative Exhaustive PP Small Clause Tree:  $\alpha$nx0Px1}
\label{nx0Px1-tree}
\end{figure}

\item[Other available trees:] wh-moved subject, wh-moved PP, relative clause 
on subject with and without comp, imperative, NP gerund,
adjunct (gap-less) relative clause with comp/with PP pied-piping.

\end{description}

\section{PP Small Clause with Sentential Subject: Ts0Pnx1}
\label{s0Pnx1-family}

\begin{description}

\item[Description:]  This tree family is selected by prepositions that take
sentential subjects.  The sentential subject can be indicative or infinitive.
Small clauses are explained in much greater detail in
section~\ref{sm-clause-xtag-analysis}.  This section is presented here for
completeness.  39 prepositions select this tree family.

\item[Examples:] {\it beyond}, {\it unlike} \\
{\it that Ken could forget to pay the taxes is beyond belief .} \\
{\it to explain how this happened is outside the scope of this discussion .} \\
{\it for Ken to do something right is unlike him .}

\item[Declarative tree:]  See Figure~\ref{s0Pnx1-tree}.

\begin{figure}[htb]
\centering
\begin{tabular}{c}
\psfig{figure=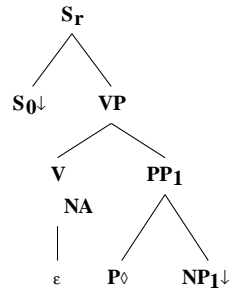,height=4.0cm}
\end{tabular}
\caption{Declarative PP Small Clause with Sentential Subject Tree:  $\alpha$s0Pnx1}
\label{s0Pnx1-tree}
\end{figure}

\item[Other available trees:] wh-moved subject, relative clause on object of
the PP with and without comp/with PP pied-piping, 
adjunct (gap-less) relative clause with comp/with PP pied-piping.

\end{description}

\section{Intransitive Sentential Subject:  Ts0V}\index{verbs,sentential subject}
\label{s0V-family}

\begin{description}

\item[Description:] Only the verb {\it matter} selects this tree
family.  The sentential subject can be indicative (complementizer
required) or infinitive (complementizer optional).

\item[Examples:] {\it matter} \\
{\it to arrive on time matters considerably .} \\
{\it that Joshi attends the meetings matters to everyone .}

\item[Declarative tree:]  See Figure~\ref{s0V-tree}.

\begin{figure}[htb]
\centering
\begin{tabular}{c}
\psfig{figure=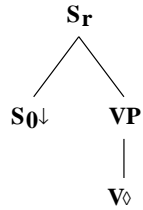,height=3.0cm}
\end{tabular}
\caption{Declarative Intransitive Sentential Subject Tree:  $\alpha$s0V}
\label{s0V-tree}
\end{figure}

\item[Other available trees:]  wh-moved subject, 
adjunct (gap-less) relative clause with comp/with PP pied-piping.

\end{description}

\section{Sentential Subject with `to' complement:  Ts0Vtonx1}\index{verbs,sentential
subject, PP complement}
\label{s0Vtonx1-family}

\begin{description}

\item[Description:] The verbs that select this tree family are {\it
fall}, {\it occur} and {\it leak}.  The sentential subject can be indicative
(complementizer required) or infinitive (complementizer optional).

\item[Examples:]  {\it fall}, {\it occur}, {\it leak}\\
{\it to wash the car fell to the children .} \\
{\it that he should leave occurred to the party crasher .} \\
{\it whether the princess divorced the prince leaked to the press .}

\item[Declarative tree:]  See Figure~\ref{s0Vtonx1-tree}.

\begin{figure}[htb]
\centering
\begin{tabular}{c}
\psfig{figure=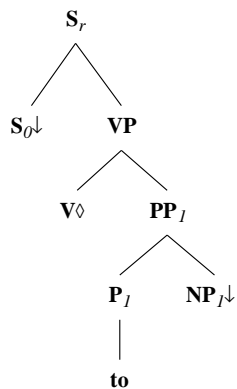,height=5.4cm}
\end{tabular}
\caption{Sentential Subject Tree with `to' complement:  $\alpha$s0Vtonx1}
\label{s0Vtonx1-tree}
\end{figure}

\item[Other available trees:]  wh-moved subject,
adjunct (gap-less) relative clause with comp/with PP pied-piping.

\end{description}

\section{PP Small Clause, with Adv and Prep anchors: Tnx0ARBPnx1}
\label{nx0ARBPnx1-family}

\begin{description}

\item[Description:]  This family is selected by multi-word prepositions that 
can occur in small clause constructions.  In particular, this family is 
selected by two-word prepositions, where the first word is an adverb, the 
second word a preposition.  Both components of the multi-word preposition
are anchors.  For more information on small clause constructions, see
section~\ref{sm-clause-xtag-analysis}.  8 multi-word prepositions select this
tree family.

\item[Examples:] {\it ahead of}, {\it close to} \\
{\it The little girl is ahead of everyone else in the race .} \\
{\it The project is close to completion .} \\

\item[Declarative tree:]  See Figure~\ref{nx0ARBPnx1-tree}.

\begin{figure}[htb]
\centering
\begin{tabular}{c}
\psfig{figure=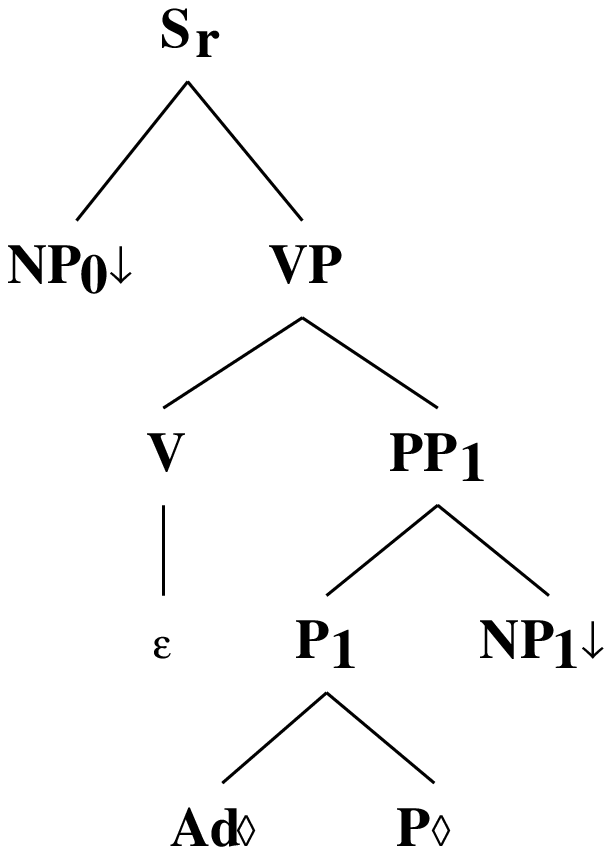,height=4.9cm}
\end{tabular}
\caption{Declarative PP Small Clause tree with two-word preposition, where the 
first word is an adverb, and the second word is a preposition:  $\alpha$nx0ARBPnx1}
\label{nx0ARBPnx1-tree}
\end{figure}

\item[Other available trees:] wh-moved subject, wh-moved object of PP,
  relative clause on subject with and without comp, relative clause on object of PP
with and without comp, adjunct (gap-less) relative clause
with comp/with PP pied-piping, imperative, NP Gerund.

\end{description}

\section{PP Small Clause, with Adj and Prep anchors: Tnx0APnx1}
\label{nx0APnx1-family}

\begin{description}

\item[Description:]  This family is selected by multi-word prepositions that
can occur in small clause constructions.  In particular, this family is 
selected by two-word prepositions, where the first word is an adjective, the 
second word a preposition.  Both components of the multi-word preposition
are anchors.  For more information on small clause constructions, see 
section~\ref{sm-clause-xtag-analysis}.  8 multi-word prepositions select this 
tree family.

\item[Examples:] {\it according to}, {\it void of} \\
{\it The operation we performed was according to standard procedure .} \\
{\it He is void of all feeling .} \\

\item[Declarative tree:]  See Figure~\ref{nx0APnx1-tree}.

\begin{figure}[htb]
\centering
\begin{tabular}{c}
\psfig{figure=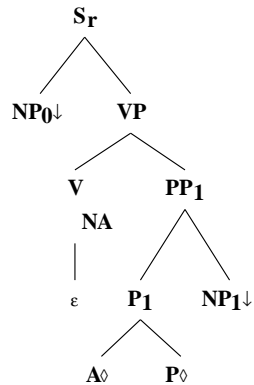,height=5.3cm}
\end{tabular}
\caption{Declarative PP Small Clause tree with two-word preposition, where the 
first word is an adjective, and the second word is a preposition: $\alpha$nx0APnx1}
\label{nx0APnx1-tree}
\end{figure}

\item[Other available trees:]  wh-moved subject, relative clause on subject 
with and without comp,
relative clause on object of PP with and without comp, wh-moved object of PP,
adjunct (gap-less) relative clause with comp/with PP pied-piping.

\end{description}

\section{PP Small Clause, with Noun and Prep anchors: Tnx0NPnx1}
\label{nx0NPnx1-family}

\begin{description}

\item[Description:] This family is selected by multi-word prepositions that 
can occur in small clause constructions.  In particular, this family is 
selected by two-word prepositions, where the first word is a noun, the 
second word a preposition.  Both components of the multi-word preposition are 
anchors.  For more information on small clause constructions, see 
section~\ref{sm-clause-xtag-analysis}. 1 multi-word preposition selects this 
tree family.

\item[Examples:] {\it thanks to} \\
{\it The fact that we are here tonight is thanks to the valiant efforts of our 
staff .} \\

\item[Declarative tree:] See Figure~\ref{nx0NPnx1-tree}.

\begin{figure}[htb]
\centering
\begin{tabular}{c}
\psfig{figure=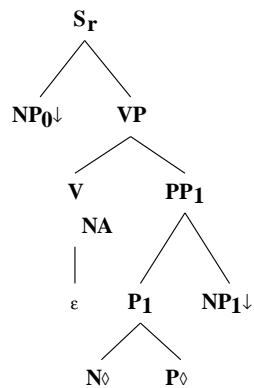,height=5.3cm}
\end{tabular}
\caption{Declarative PP Small Clause tree with two-word preposition, where the
first word is a noun, and the second word is a preposition:  $\alpha$nx0NPnx1}
\label{nx0NPnx1-tree}
\end{figure}

\item[Other available trees:]  wh-moved subject, wh-moved object of PP,
relative clause on subject with and without comp, relative clause on object with comp,
adjunct (gap-less) relative clause with comp/with PP pied-piping.

\end{description}

\section{PP Small Clause, with Prep anchors: Tnx0PPnx1}
\label{nx0PPnx1-family}

\begin{description}

\item[Description:]  This family is selected by multi-word prepositions that 
can occur in small clause constructions.  In particular, this family is 
selected by two-word prepositions, where both words are prepositions. Both 
components of the multi-word preposition are anchors.  For more information on
small clause constructions, see section~\ref{sm-clause-xtag-analysis}.  9 
multi-word prepositions select this tree family.

\item[Examples:] {\it on to}, {\it inside of} \\
{\it that detective is on to you .} \\
{\it The red box is inside of the blue box .} \\

\item[Declarative tree:] See Figure~\ref{nx0PPnx1-tree}.

\begin{figure}[htb]
\centering
\begin{tabular}{c}
\psfig{figure=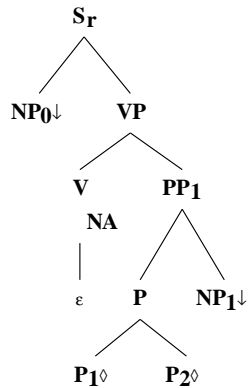,height=5.3cm}
\end{tabular}
\caption{Declarative PP Small Clause tree with two-word preposition, where both
words are prepositions:  $\alpha$nx0PPnx1}
\label{nx0PPnx1-tree}
\end{figure}    

\item[Other available trees:] wh-moved subject, wh-moved object of PP, relative
clause on subject with and without comp, relative clause on object of PP with 
and without comp/with PP pied-piping, 
imperative, wh-moved object of PP, adjunct (gap-less) relative clause
with comp/with PP pied-piping.

\end{description}

\section{PP Small Clause, with Prep and Noun anchors: Tnx0PNaPnx1}
\label{nx0PNaPnx1-family}

\begin{description}

\item[Description:]  This family is selected by multi-word prepositions that 
can occur in small clause constructions.  In particular, this family is 
selected by three-word prepositions.  The first and third words are always
prepositions, and the middle word is a noun.  The noun is marked for null 
adjunction since it cannot be modified by noun modifiers.  All three components
of the multi-word preposition are anchors.  For more information on small 
clause constructions, see section~\ref{sm-clause-xtag-analysis}.  24 multi-word
preposition select this tree family.

\item[Examples:] {\it in back of}, {\it in line with}, {\it on top of} \\
{\it The red plaid box should be in back of the plain black box .} \\
{\it The evidence is in line with my newly concocted theory .} \\
{\it She is on top of the world .} \\
{\it *She is on direct top of the world .} \\

\item[Declarative tree:] See Figure~\ref{nx0PNaPnx1-tree}.

\begin{figure}[htb]
\centering
\begin{tabular}{c}
\psfig{figure=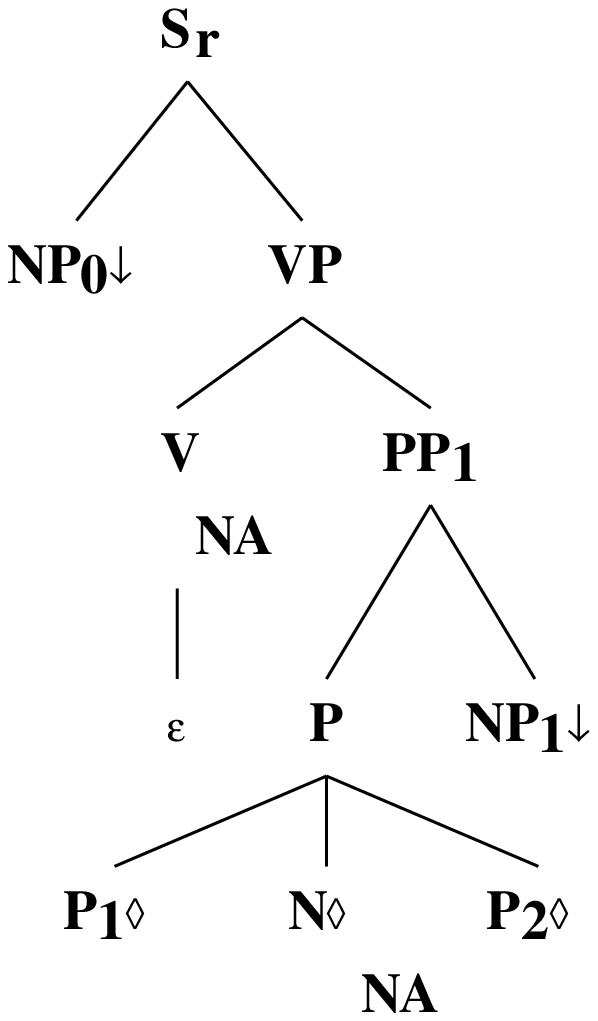,height=5.5cm}
\end{tabular}
\caption{Declarative PP Small Clause tree with three-word preposition,
where the middle noun is marked for null adjunction:  $\alpha$nx0PNaPnx1}
\label{nx0PNaPnx1-tree}
\end{figure}

\item[Other available trees:] wh-moved subject, wh-moved object of PP,
  relative clause on subject with and without comp, relative clause on object of PP
with and without comp/with PP pied-piping, adjunct (gap-less) relative clause
with comp/with PP pied-piping, imperative, NP Gerund.

\end{description}

\section{PP Small Clause with Sentential Subject, and Adv and Prep anchors: Ts0ARBPnx1}
\label{s0ARBPnx1-family}

\begin{description}

\item[Description:]  This tree family is selected by multi-word prepositions 
that take sentential subjects. In particular, this family is selected by
two-word prepositions, where the first word is an adverb, the second word a 
preposition.  Both components of the multi-word preposition are anchors. The 
sentential subject can be indicative or infinitive.  Small clauses are 
explained in much greater detail in section~\ref{sm-clause-xtag-analysis}.  
2 prepositions select this tree family.

\item[Examples:]  {\it due to}, {\it contrary to} \\
{\it that David slept until noon is due to the fact that he never sleeps during
the week .} \\
{\it that Michael's joke was funny is contrary to the usual status of his comic
attempts .} \\

\item[Declarative tree:]  See Figure~\ref{s0ARBPnx1-tree}.
 
\begin{figure}[htb]
\centering
\begin{tabular}{c}
\psfig{figure=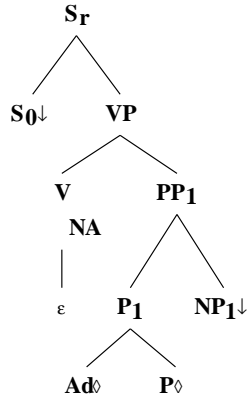,height=5.5cm}
\end{tabular}
\caption{Declarative PP Small Clause with Sentential Subject Tree, with 
two-word preposition, where the first word is an adverb, and the second word is
a preposition:  $\alpha$s0ARBPnx1}
\label{s0ARBPnx1-tree}
\end{figure}

\item[Other available trees:]  wh-moved subject, relative clause on object of 
the PP with and without comp, adjunct (gap-less) relative clause
with comp/with PP pied-piping.

\end{description}

\section{PP Small Clause with Sentential Subject, and Adj and Prep anchors: Ts0APnx1}
\label{s0APnx1-family}

\begin{description}

\item[Description:]  This tree family is selected by multi-word prepositions 
that take sentential subjects. In particular, this family is selected by
two-word prepositions, where the first word is an adjective, the second word a
preposition.  Both components of the multi-word preposition are anchors. The 
sentential subject can be indicative or infinitive.  Small clauses are 
explained in much greater detail in section~\ref{sm-clause-xtag-analysis}.  
5 prepositions select this tree family.

\item[Examples:] {\it devoid of}, {\it according to} \\ 
{\it that he could walk out on her is devoid of all reason .} \\
{\it that the conversation erupted precisely at that moment was according to my
theory .} \\

\item[Declarative tree:]  See Figure~\ref{s0APnx1-tree}.
        
\begin{figure}[htb]
\centering
\begin{tabular}{c}
\psfig{figure=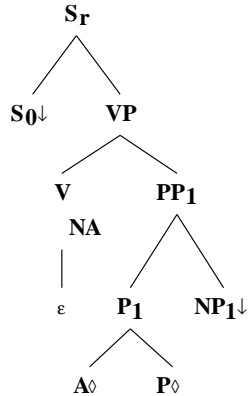,height=5.5cm}
\end{tabular}
\caption{Declarative PP Small Clause with Sentential Subject Tree, with 
two-word preposition, where the first word is an adjective, and the second word
is a preposition:  $\alpha$s0APnx1}
\label{s0APnx1-tree}
\end{figure}

\item[Other available trees:] wh-moved subject, relative clause on object of 
the PP with and without comp, adjunct (gap-less) relative clause
with comp/with PP pied-piping.

\end{description}

\section{PP Small Clause with Sentential Subject, and Noun and Prep anchors: Ts0NPnx1}
\label{s0NPnx1-family}

\begin{description}

\item[Description:]  This tree family is selected by multi-word prepositions 
that take sentential subjects. In particular, this family is selected by
two-word prepositions, where the first word is a noun, the second word a 
preposition.  Both components of the multi-word preposition are anchors. The 
sentential subject can be indicative or infinitive.  Small clauses are 
explained in much greater detail in section~\ref{sm-clause-xtag-analysis}.  
1 preposition selects this tree family.

\item[Examples:] {\it thanks to} \\
{\it that she is worn out is thanks to a long day in front of the computer
terminal .} \\ 

\item[Declarative tree:]  See Figure~\ref{s0NPnx1-tree}.

\begin{figure}[htb]
\centering
\begin{tabular}{c}
\psfig{figure=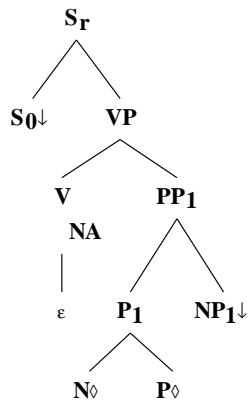,height=5.5cm}
\end{tabular}
\caption{Declarative PP Small Clause with Sentential Subject Tree, with 
two-word preposition, where the first word is a noun, and the second word is a preposition:  $\alpha$s0NPnx1}
\label{s0NPnx1-tree}
\end{figure}

\item[Other available trees:] wh-moved subject, relative clause on object of 
the PP with and without comp, adjunct (gap-less) relative clause
with comp/with PP pied-piping.

\end{description}

\section{PP Small Clause with Sentential Subject, and Prep anchors: Ts0PPnx1}
\label{s0PPnx1-family}

\begin{description}

\item[Description:]  This tree family is selected by multi-word prepositions
that take sentential subjects. In particular, this family is selected by
two-word prepositions, where both words are prepositions.  Both components of 
the multi-word preposition are anchors. The sentential subject can be 
indicative or infinitive.  Small clauses are explained in much greater detail 
in section~\ref{sm-clause-xtag-analysis}.  3 prepositions select this tree
family.

\item[Examples:] {\it outside of} \\
{\it that Mary did not complete the task on time is outside of the scope of 
this discussion .} \\

\item[Declarative tree:]  See Figure~\ref{s0PPnx1-tree}. 

\begin{figure}[htb]
\centering
\begin{tabular}{c}
\psfig{figure=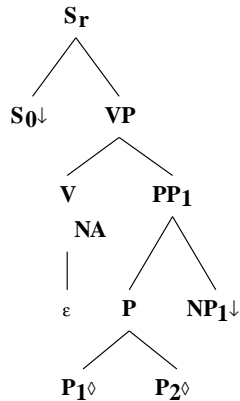,height=5.5cm}
\end{tabular}
\caption{Declarative PP Small Clause with Sentential Subject Tree, with 
two-word preposition, where both words are prepositions:  $\alpha$s0PPnx1}
\label{s0PPnx1-tree}
\end{figure}

\item[Other available trees:]  wh-moved subject, relative clause on object of 
the PP with and without comp, adjunct (gap-less) relative clause
with comp/with PP pied-piping.

\end{description}

\section{PP Small Clause with Sentential Subject, and Prep and Noun anchors: Ts0PNaPnx1}
\label{s0PNaPnx1-family}

\begin{description}
  
\item{Description:} This tree family is selected by multi-word
  prepositions that take sentential subjects. In particular, this
  family is selected by three-word prepositions.  The first and third
  words are always prepositions, and the middle word is a noun.  The
  noun is marked for null adjunction since it cannot be modified by
  noun modifiers.  All three components of the multi-word preposition
  are anchors.  Small clauses are explained in much greater detail in
  section~\ref{sm-clause-xtag-analysis}.  9 prepositions select this
  tree family.

\item[Examples:]  {\it on account of}, {\it in support of} \\
{\it that Joe had to leave the beach was on account of the hurricane .} \\
{\it that Maria could not come is in support of my theory about her .} \\
{\it *that Maria could not come is in direct/strict/desparate support of my
theory about her .} \\

\item[Declarative tree:]  See Figure~\ref{s0PNaPnx1-tree}.

\begin{figure}[htb]
\centering
\begin{tabular}{c}
\psfig{figure=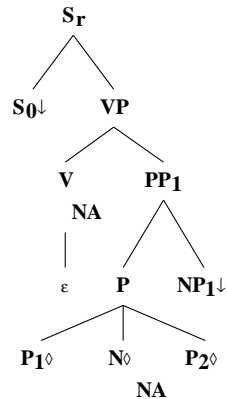,height=5.5cm}
\end{tabular}
\caption{Declarative PP Small Clause with Sentential Subject Tree, with 
three-word preposition, where the middle noun is marked for null adjunction:
$\alpha$s0PNaPnx1} 
\label{s0PNaPnx1-tree}
\end{figure}
        
\item[Other available trees:] wh-moved subject, relative clause on object of 
the PP with and without comp, adjunct (gap-less) relative clause
with comp/with PP pied-piping.

\end{description}

\section{Predicative Adjective with Sentential Subject and Complement: Ts0A1s1}
\label{s0A1s1-family}

\begin{description}
  
\item{Description:} This tree family is selected by predicative
  adjectives that take sentential subjects and a sentential
  complement. This tree family is selected by {\it likely} and {\it
    certain}.

\item[Examples:]  {\it likely}, {\it certain} \\
{\it that Max continues to drive a Jaguar is certain to make Bill jealous .} \\
{\it for the Jaguar to be towed seems likely to make Max very angry .} \\

\item[Declarative tree:]  See Figure~\ref{s0A1s1-tree}.

\begin{figure}[htb]
\centering
\begin{tabular}{c}
\psfig{figure=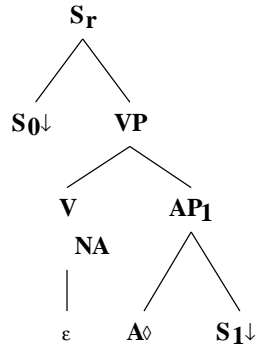,height=4.8cm}
\end{tabular}
\caption{Predicative Adjective with Sentential Subject and Complement:
$\alpha$s0A1s1} 
\label{s0A1s1-tree}
\end{figure}
        
\item[Other available trees:] wh-moved subject, 
adjunct (gap-less) relative clause with comp/with PP pied-piping.

\end{description}

\section{Locative Small Clause with Ad anchor: Tnx0nx1ARB}
\label{nx0nx1ARB-family}

\begin{description}

\item[Description:]  These trees are not anchored by verbs, but by adverbs 
that are part of locative adverbial phrases. Locatives are explained 
in much greater detail in the section on the locative modifier trees
(see section~\ref{locatives}). The only remarkable aspect of this tree 
family is the wh-moved locative tree, $\alpha$W1nx0nx1ARB, shown in 
Figure~\ref{W1nx0nx1ARB-tree}. This is the only tree family with this type of 
transformation, in which the entire adverbial phrase is wh-moved but not all 
elements are replaced by wh items (as in {\it how many city blocks away 
is the record store?}). Locatives that consist of just the locative adverb 
or the locative adverb and a degree adverb (see Section \ref{locatives} for 
details) are treated as exhaustive PPs and therefore select that tree 
family (Section~\ref{nx0Px1-family}) when used predicatively. For an 
extensive description of small clauses, see 
Section~\ref{sm-clause-xtag-analysis}. 26 adverbs select this tree family.

\item[Examples:] {\it ahead}, {\it offshore}, {\it behind} \\
{\it the crash is three blocks ahead} \\
{\it the naval battle was many kilometers offshore} \\
{\it how many blocks behind was Max?} \\

\item[Declarative tree:]  See Figure~\ref{nx0nx1ARB-tree}.

\begin{figure}[htb]
\centering
\begin{tabular}{c}
\psfig{figure=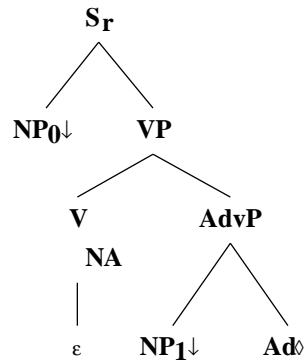,height=5.0cm}
\end{tabular}
\caption{Declarative Locative Adverbial Small Clause Tree:  $\alpha$nx0nx1ARB}
\label{nx0nx1ARB-tree}
\label{3;nx0nx1ARB}
\end{figure}

\begin{figure}[htb]
\centering
\begin{tabular}{c}
\psfig{figure=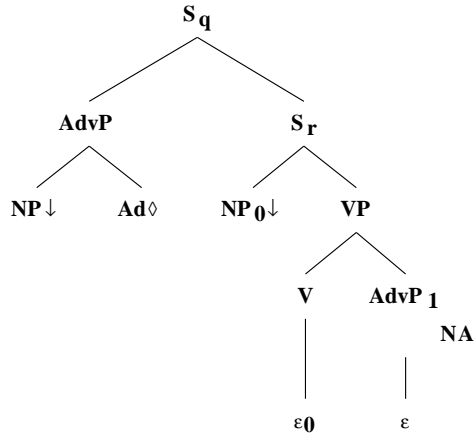,height=6.0cm}
\end{tabular}
\caption{Wh-moved Locative Small Clause Tree:  $\alpha$W1nx0nx1ARB}
\label{W1nx0nx1ARB-tree}
\label{3;W1nx0nx1ARB}
\end{figure}

\item[Other available trees:]  wh-moved subject, relative clause on subject
with and without comp, wh-moved locative, imperative, NP gerund.

\end{description}

\section{Exceptional Case Marking: TXnx0Vs1}\index{verbs,ecm}
\label{Xnx0Vs1-family}

\begin{description}

\item[Description:]  This tree family is selected by verbs that are classified
as exceptional case marking, meaning that the verb asssigns accusative case
to the subject of the sentential complement.  This is in contrast to verbs
in the Tnx0Vnx1s2 family (section~\ref{nx0Vnx1s2-family}), which assign 
accusative
case to a NP which is not part of the sentential complement.  ECM verbs
take sentential complements which are either an infinitive or a ``bare''
infinitive.  As with the Tnx0Vs1 family (section~\ref{nx0Vs1-family}), the
declarative and other trees in the Xnx0Vs1 family are auxiliary trees, as
opposed to the more common initial trees.  These auxiliary trees adjoin onto an
S node in an existing tree of the type specified by the sentential complement.
This is the mechanism by which TAGs are able to maintain long-distance
dependencies (see Chapter~\ref{extraction}), even over multiple embeddings
(e.g. {\it Who did Bill expect to eat beans?}) or 
{\it who did Bill expect Mary to like?}  See section~\ref{ecm-verbs} 
for details on this family.  20 verbs select this tree family.

\item[Examples:]  {\it expect}, {\it see} \\
{\it Van expects Bob to talk .}
{\it Bob sees the harmonica fall .}

\item[Declarative tree:]  See Figure~\ref{Xnx0Vs1-tree}.

\begin{figure}[htb]
\centering
\begin{tabular}{c}
\psfig{figure=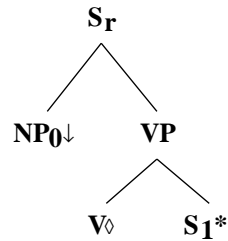,height=3.4cm}
\end{tabular}
\caption{ECM Tree:  $\beta$Xnx0Vs1}
\label{Xnx0Vs1-tree}
\end{figure}

\item[Other available trees:]  wh-moved subject, subject relative clause with and without comp, 
adjunct (gap-less) relative clause with and without comp/with PP pied-piping, 
imperative,  NP gerund. 

\end{description}

\section{Idiom with V, D, and N anchors: Tnx0VDN1}\index{verbs,idiomatic}
\label{nx0VDN1-family}

\begin{description}

\item[Description:]  This tree family is selected by idiomatic phrases
in which the verb, determiner, and NP are all frozen (as in {\it He
kicked the bucket.}).  Only a limited number of transformations are
allowed, as compared to the normal transitive tree family (see
section~\ref{nx0Vnx1-family}).  Other idioms that have the same
structure as {\it kick the bucket}, and that are limited to the same
transformations would select this tree, while different tree families
are used to handle other idioms.  Note that {\it John kicked the
bucket} is actually ambiguous, and would result in two parses - an
idiomatic one (meaning that John died), and a compositional transitive
one (meaning that there is an physical bucket that John hit with
his foot). 1 idiom selects this family.

\item[Examples:] {\it kick the bucket} \\
{\it Nixon kicked the bucket .}

\item[Declarative tree:]  See Figure~\ref{nx0VDN1-tree}.

\begin{figure}[htb]
\centering
\begin{tabular}{c}
\psfig{figure=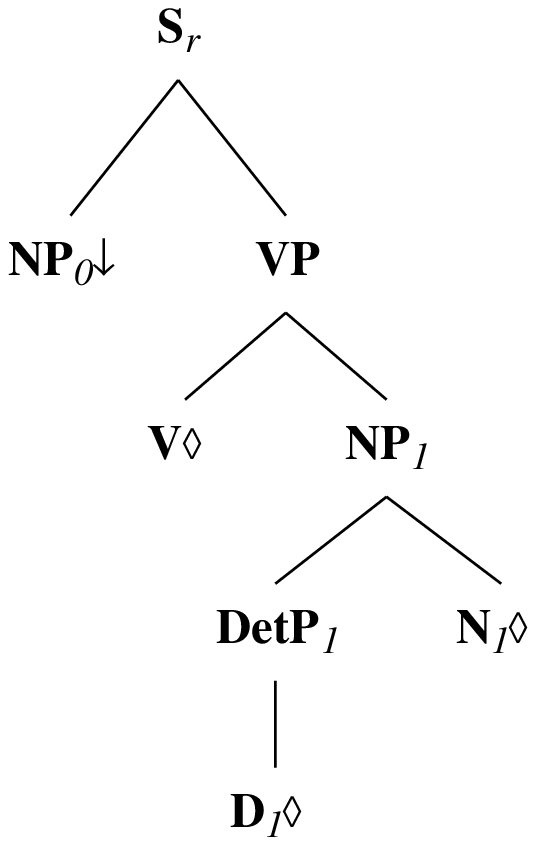,height=5.2cm}
\end{tabular}
\caption{Declarative Transitive Idiom Tree:  $\alpha$nx0VDN1}
\label{nx0VDN1-tree}
\end{figure}

\item[Other available trees:] subject relative clause with and without
  comp, declarative, wh-moved subject, imperative, NP gerund, adjunct gapless
  relative with comp/with PP pied-piping, passive, w/wo by-phrase, wh-moved object of by-phrase,
  wh-moved by-phrase, relative (with and without comp) on subject of
  passive, PP relative.

\end{description}

\section{Idiom with V, D, A, and N anchors: Tnx0VDAN1}\index{verbs,idiomatic}
\label{nx0VDAN1-family}

\begin{description}

\item[Description:]
This tree family is selected by transitive idioms that are anchored by a 
verb, determiner, adjective, and noun. 19 idioms select this family.

\item[Examples:] {\it have a green thumb}, {\it sing a different tune} \\
{\it Martha might have a green thumb, but it's uncertain after the death of all the plants.} \\
{\it After his conversion John sang a different tune.} \\

\item[Declarative tree:]  See Figure~\ref{nx0VDAN1-tree}.

\begin{figure}[htb]
\centering
\begin{tabular}{c}
\psfig{figure=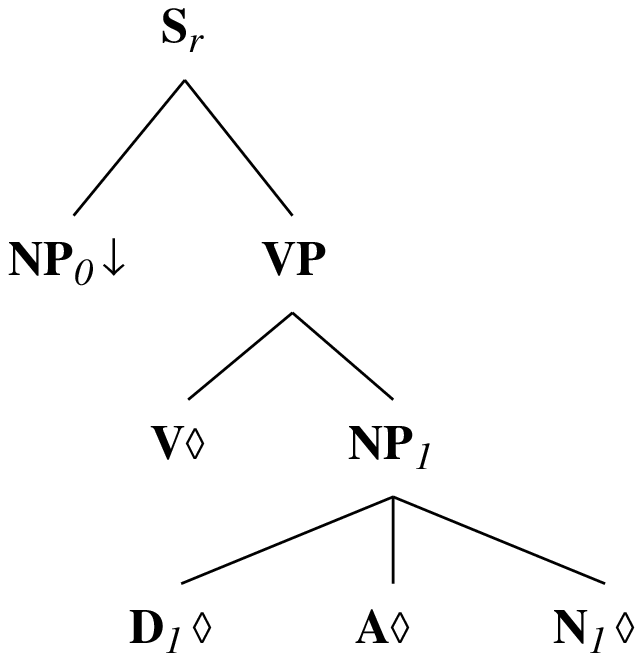,height=5.0cm}
\end{tabular}
\caption{Declarative Idiom with V, D, A, and N Anchors Tree: $\alpha$nx0VDAN1}
\label{nx0VDAN1-tree}
\label{3;nx0VDAN1}
\end{figure}

\item[Other available trees:] Subject relative clause with and without comp, 
adjunct relative clause with comp/with PP pied-piping,
wh-moved subject, imperative, NP gerund, passive without {\it by} phrase, passive with 
{\it by} phrase, passive with wh-moved object of {\it by} phrase, passive with 
wh-moved {\it by phrase}, passive with relative on object of {\it by} phrase
with and without comp.

\end{description}

\section{Idiom with V and N anchors: Tnx0VN1}\index{verbs,idiomatic}
\label{nx0VN1-family}

\begin{description}

\item[Description:]
This tree family is selected by transitive idioms that are anchored by a 
verb and noun. 15 idioms select this family.

\item[Examples:] {\it draw blood}, {\it cry wolf} \\
{\it Graham's retort drew blood.} \\
{\it The neglected boy cried wolf.} \\

\item[Declarative tree:]  See Figure~\ref{nx0VN1-tree}.

\begin{figure}[htb]
\centering
\begin{tabular}{c}
\psfig{figure=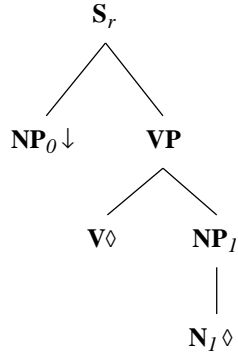,height=5.0cm}
\end{tabular}
\caption{Declarative Idiom with V and N Anchors Tree: $\alpha$nx0VN1}
\label{nx0VN1-tree}
\label{3;nx0VN1}
\end{figure}

\item[Other available trees:] Subject relative clause with and without comp, 
adjunct relative clause with comp/with PP pied-piping,
wh-moved subject, imperative, NP gerund, passive without {\it by} phrase, passive with 
{\it by} phrase, passive with wh-moved object of {\it by} phrase, passive with 
wh-moved {\it by phrase}, passive with relative on object of {\it by} phrase
with and without comp.

\end{description}

\section{Idiom with V, A, and N anchors: Tnx0VAN1}\index{verbs,idiomatic}
\label{nx0VAN1-family}

\begin{description}

\item[Description:]
This tree family is selected by transitive idioms that are anchored by a 
verb, adjective, and noun. 4 idioms select this family.

\item[Examples:] {\it break new ground}, {\it cry bloody murder} \\
{\it The avant-garde film breaks new ground.} \\
{\it The investors cried bloody murder after the suspicious takeover.} \\

\item[Declarative tree:]  See Figure~\ref{nx0VAN1-tree}.

\begin{figure}[htb]
\centering
\begin{tabular}{c}
\psfig{figure=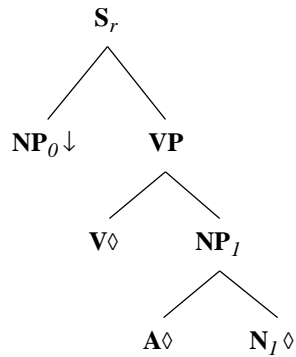,height=5.0cm}
\end{tabular}
\caption{Declarative Idiom with V, A, and N Anchors Tree: $\alpha$nx0VAN1}
\label{nx0VAN1-tree}
\label{3;nx0VAN1}
\end{figure}

\item[Other available trees:] Subject relative clause with and without comp, 
adjunct relative clause with comp/with PP pied-piping,
wh-moved subject, imperative, NP gerund, passive without {\it by} phrase, passive with 
{\it by} phrase, passive with wh-moved object of {\it by} phrase, passive with 
wh-moved {\it by phrase}, passive with relative on object of {\it by} phrase
with and without comp.

\end{description}

\section{Idiom with V, D, A, N, and Prep anchors: Tnx0VDAN1Pnx2}\index{verbs,idiomatic}
\label{nx0VDAN1Pnx2-family}

\begin{description}

\item[Description:]
This tree family is selected by transitive idioms that are anchored by a 
verb, determiner, adjective, noun, and preposition. 6 idioms select this family.

\item[Examples:] {\it make a big deal about}, {\it make a great show of} \\
{\it John made a big deal about a miniscule dent in his car.} \\
{\it The company made a big show of paying generous dividends.} \\

\item[Declarative tree:]  See Figure~\ref{nx0VDAN1Pnx2-tree}.

\begin{figure}[htb]
\centering
\begin{tabular}{c}
\psfig{figure=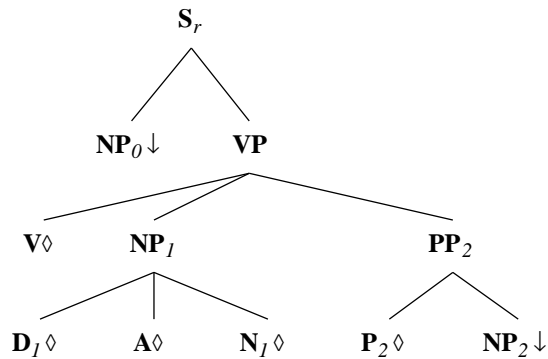,height=5.0cm}
\end{tabular}
\caption{Declarative Idiom with V, D, A, N, and Prep Anchors Tree: $\alpha$nx0VDAN1Pnx2}
\label{nx0VDAN1Pnx2-tree}
\label{3;nx0VDAN1Pnx2}
\end{figure}

\item[Other available trees:] Subject relative clause with and without comp, 
adjunct relative clause with comp/with PP pied-piping,
wh-moved subject, imperative, NP gerund, passive without {\it by} phrase, passive with 
{\it by} phrase, passive with wh-moved object of {\it by} phrase, passive with 
wh-moved {\it by phrase},

outer passive without {\it by} phrase, outer passive with {\it by} phrase, 
outer passive with wh-moved {\it by} phrase, outer passive with wh-moved 
object of {\it by} phrase, 
outer passive without {\it by} phrase with relative on the subject with and without comp, 
outer passive with {\it by} phrase with relative on subject with and without comp.

\end{description}

\section{Idiom with V, A, N, and Prep anchors: Tnx0VAN1Pnx2}\index{verbs,idiomatic}
\label{nx0VAN1Pnx2-family}

\begin{description}

\item[Description:]
This tree family is selected by transitive idioms that are anchored by a 
verb, adjective, noun, and preposition. 3 idioms select this family.

\item[Examples:] {\it make short work of} \\
{\it John made short work of the glazed ham.} \\

\item[Declarative tree:]  See Figure~\ref{nx0VAN1Pnx2-tree}.

\begin{figure}[htb]
\centering
\begin{tabular}{c}
\psfig{figure=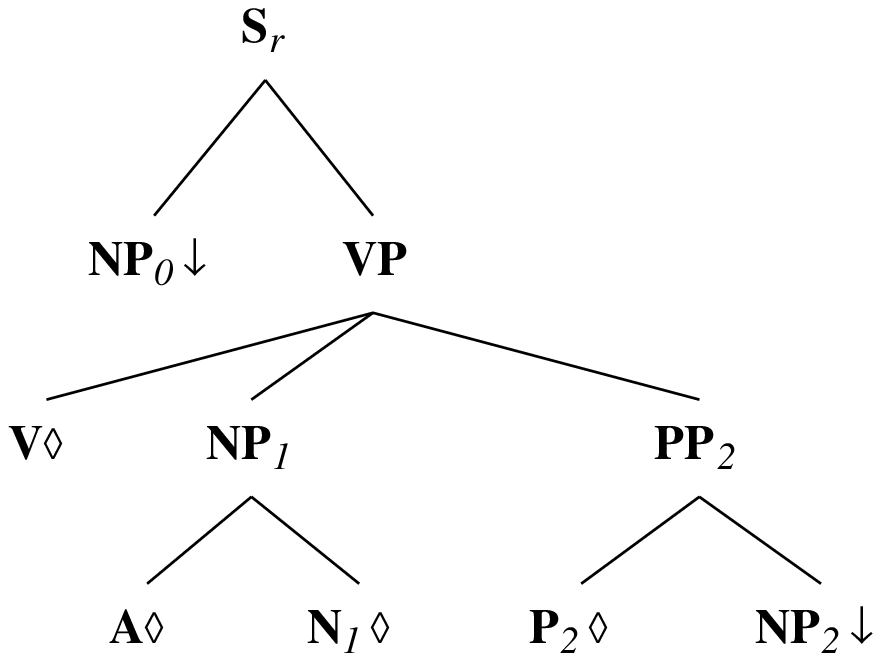,height=5.0cm}
\end{tabular}
\caption{Declarative Idiom with V, A, N, and Prep Anchors Tree: $\alpha$nx0VAN1Pnx2}
\label{nx0VAN1Pnx2-tree}
\label{3;nx0VAN1Pnx2}
\end{figure}

\item[Other available trees:] Subject relative clause with and without comp, 
adjunct relative clause with comp/with PP pied-piping,
wh-moved subject, imperative, NP gerund, passive without {\it by} phrase, passive with 
{\it by} phrase, passive with wh-moved object of {\it by} phrase, passive with 
wh-moved {\it by phrase},

outer passive without {\it by} phrase, outer passive with {\it by} phrase, 
outer passive with wh-moved {\it by} phrase, outer passive with wh-moved 
object of {\it by} phrase, 
outer passive without {\it by} phrase with relative on the subject with and without comp, 
outer passive with {\it by} phrase with relative on subject with and without comp.

\end{description}

\section{Idiom with V, N, and Prep anchors: Tnx0VN1Pnx2}\index{verbs,idiomatic}
\label{nx0VN1Pnx2-family}

\begin{description}

\item[Description:]
This tree family is selected by transitive idioms that are anchored by a 
verb, noun, and preposition. 6 idioms select this family.

\item[Examples:] {\it look daggers at}, {\it keep track of} \\
{\it Maria looked daggers at her ex-husband across the courtroom.} \\
{\it The company kept track of its inventory.} \\

\item[Declarative tree:]  See Figure~\ref{nx0VN1Pnx2-tree}.

\begin{figure}[htb]
\centering
\begin{tabular}{c}
\psfig{figure=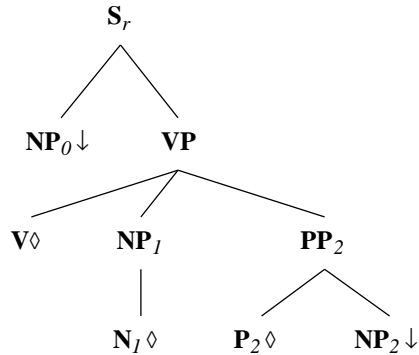,height=5.0cm}
\end{tabular}
\caption{Declarative Idiom with V, N, and Prep Anchors Tree: $\alpha$nx0VN1Pnx2}
\label{nx0VN1Pnx2-tree}
\label{3;nx0VN1Pnx2}
\end{figure}

\item[Other available trees:] Subject relative clause with and without comp, 
adjunct relative clause with comp/with PP pied-piping,
wh-moved subject, imperative, NP gerund, passive without {\it by} phrase, passive with 
{\it by} phrase, passive with wh-moved object of {\it by} phrase, passive with 
wh-moved {\it by phrase}, 

outer passive without {\it by} phrase, outer passive with {\it by} phrase, 
outer passive with wh-moved {\it by} phrase, outer passive with wh-moved 
object of {\it by} phrase, 
outer passive without {\it by} phrase with relative on the subject with and without comp, 
outer passive with {\it by} phrase with relative on subject with and without comp.

\end{description}

\section{Idiom with V, D, N, and Prep anchors: Tnx0VDN1Pnx2}\index{verbs,idiomatic}
\label{nx0VDN1Pnx2-family}

\begin{description}

\item[Description:]
This tree family is selected by transitive idioms that are anchored by a 
verb, determiner, noun, and preposition. 17 idioms select this family.

\item[Examples:] {\it make a mess of}, {\it keep the lid on} \\
{\it John made a mess out of his new suit.} \\
{\it The tabloid didn't keep a lid on the imminent celebrity nuptials.} \\

\item[Declarative tree:]  See Figure~\ref{nx0VDN1Pnx2-tree}.

\begin{figure}[htb]
\centering
\begin{tabular}{c}
\psfig{figure=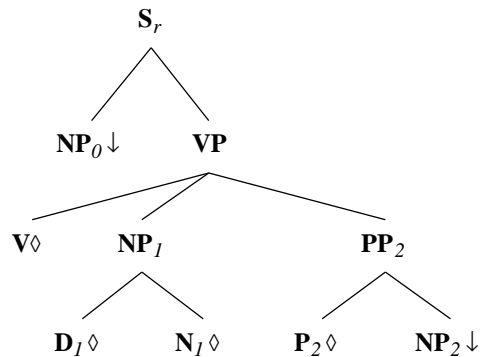,height=5.0cm}
\end{tabular}
\caption{Declarative Idiom with V, D, N, and Prep Anchors Tree: $\alpha$nx0VDN1Pnx2}
\label{nx0VDN1Pnx2-tree}
\label{3;nx0VDN1Pnx2}
\end{figure}

\item[Other available trees:] Subject relative clause with and without comp, 
adjunct relative clause with comp/with PP pied-piping,
wh-moved subject, imperative, NP gerund, passive without {\it by} phrase, passive with 
{\it by} phrase, passive with wh-moved object of {\it by} phrase, passive with 
wh-moved {\it by phrase}, 

outer passive without {\it by} phrase, outer passive with {\it by} phrase, 
outer passive with wh-moved {\it by} phrase, outer passive with wh-moved 
object of {\it by} phrase, 
outer passive without {\it by} phrase with relative on the subject with and without comp, 
outer passive with {\it by} phrase with relative on subject with and without comp.

\end{description}

\chapter{Ergatives}
\label{ergatives}

Verbs in English that are termed ergative display the kind of
alternation shown in the sentences in (\ex{1}) below.

\enumsentence{The sun melted the ice .\\
The ice melted .}

The pattern of ergative pairs as seen in (\ex{0}) is for the object of the
transitive sentence to be the subject of the intransitive sentence.
The literature discussing such pairs is based largely on syntactic
models that involve movement, particularly GB.  Within that framework
two basic approaches are discussed:

\begin{itemize}
\item {\bf Derived Intransitive}\\ The intransitive member of the
ergative pair is derived through processes of movement and deletion from:
\begin{itemize}
\item a transitive D-structure \cite{Burzio86}; or	
\item transitive lexical structure \cite{HaleKeyser86,HaleKeyser87}
\end{itemize}

\item {\bf Pure Intransitive}\\ The intransitive member is intransitive at all levels of the
syntax and the lexicon and is not related to the transitive member
syntactically or lexically \cite{Napoli88}.
\end{itemize}

The Derived Intransitive approach's notions of movement in the
lexicon or in the grammar are not represented as such in the XTAG grammar.
However, distinctions drawn in these arguments can be translated to the FB-LTAG
framework.  In the XTAG grammar the difference between these two approaches is
not a matter of movement but rather a question of tree family membership.  The
relation between sentences represented in terms of movement in other frameworks
is represented in XTAG by membership in the same tree family. Wh-questions and
their indicative counterparts are one example of this.  Adopting the Pure
Intransitive approach suggested by \cite{Napoli88} would mean placing the
intransitive ergatives in a tree family with other intransitive verbs and
separate from the transitive variants of the same verbs.  This would result in
a grammar that represented intransitive ergatives as more closely related to
other intransitives than to their transitive counterparts.  The only hint of
the relation between the intransitive ergatives and the transitive ergatives
would be that ergative verbs would select both tree families. While
this is a workable solution, it is an unattractive one for the English XTAG
grammar because semantic coherence is implicitly associated with tree families
in our analysis of other constructions.  In particular, constancy in thematic
role is represented by constancy in node names across sentence types within a
tree family. For example, if the object of a declarative tree is NP$_{1}$ the
subject of the passive tree(s) in that family will also be NP$_{1}$.

The analysis that has been implemented in the English XTAG grammar is an
adaptation of the Derived Intransitive approach. The ergative verbs select one
family, Tnx0Vnx1, that contains both transitive and intransitive trees.  The
{\bf$<$trans$>$} feature appears on the intransitive ergative trees with the
value {\bf --} and on the transitive trees with the value {\bf +}.  This
creates the two possibilities needed to account for the data.

\begin{itemize}

\item {\bf intransitive ergative/transitive alternation.}  These verbs
have transitive and intransitive variants as shown in sentences~(\ex{1}) and
(\ex{2}).

\enumsentence{The sun melted the ice cream .}
\enumsentence{The ice cream melted .}

In the English XTAG grammar, verbs with this behavior are left unspecified as
to value for the {\bf$<$trans$>$} feature.  This lack of specification allows
these verbs to anchor either type of tree in the Tnx0Vnx1 tree family because
the unspecified {\bf$<$trans$>$} value of the verb can unify with either {\bf
+} or {\bf --} values in the trees.

\item {\bf transitive only.}  Verbs of this type select only the
transitive trees and do not allow intransitive ergative variants as in
the pattern show in sentences~(\ex{1}) and (\ex{2}).

\enumsentence{Elmo borrowed a book .}
\enumsentence{$\ast$A book borrowed .}

The restriction to selecting only transitive trees is accomplished by
setting the {\bf$<$trans$>$} feature value to {\bf +} for these verbs.
\end{itemize}

\begin{figure}[htb]
\centering
\mbox{}
\psfig{figure=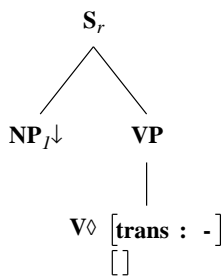,height=4.0cm}
\caption{Ergative Tree: $\alpha$Enx1V}
\label{decl-erg-tree}
\label{2;14,1}
\end{figure}

The declarative ergative tree is shown in Figure~\ref{decl-erg-tree} with the
{\bf $<$trans$>$} feature displayed.  Note that the index of the subject NP
indicates that it originated as the object of the verb.

\chapter{Sentential Subjects and Sentential Complements}
\label{scomps-section}

In the XTAG grammar, arguments of a lexical item, including
subjects, appear in the initial tree anchored by that lexical item.  A
sentential argument appears as an S node in the appropriate position
within an elementary tree anchored by the lexical item that selects
it. This is the case for sentential complements of verbs, prepositions
and nouns and for sentential subjects. The distribution of
complementizers in English is intertwined with the distribution of
embedded sentences.  A successful analysis of complementizers in
English must handle both the cooccurrence restrictions between
complementizers and various types of clauses, and the distribution of
the clauses themselves, in both subject and complement positions.

\section{S or VP complements?}
 
Two comparable grammatical formalisms, Generalized Phrase Structure
Grammar (GPSG) \cite{gazdar85} and Head-driven Phrase Structure
Grammar (HPSG) \cite{PollardSag94:BK}, have rather different
treatments of sentential complements (S-comps).  They both treat
embedded sentences as VP's with subjects, which generates the correct
structures but misses the generalization that S's behave similarly in
both matrix and embedded environments, and VP's behave quite
differently.  Neither account has PRO\label{PRO} subjects of
infinitival clauses-- they have subjectless VP's instead.  GPSG has a
complete complementizer system, which appears to cover the same range
of data as our analysis.  It is not clear what sort of complementizer
analysis could be implemented in HPSG.

Following standard GB approach, the English XTAG grammar does not
allow VP complements but treats verb-anchored structures without overt
subjects as having PRO subjects. Thus, indicative clauses, infinitives
and gerunds all have a uniform treatment as embedded clauses using the
same trees under this approach. Furthermore, our analysis is able to
preserve the selectional and distributional distinction between S's and
VP's, in the spirit of GB theories, without having to posit `extra'
empty categories.\footnote{i.e. empty complementizers. We do have PRO
and NP traces in the grammar.} Consider the alternation between {\it
that} and the null complementizer\footnote{Although we will continue
to refer to `null' complementizers, in our analysis this is actually
the absence of a complementizer.}, shown in sentences~(\ex{1}) and (\ex{2}).

\enumsentence{He hopes $\emptyset$ Muriel wins.}
\enumsentence{He hopes that Muriel wins.}

 In GB both {\it Muriel wins} in (\ex{-1}) and {\it that Muriel wins} in
(\ex{0}) are CPs even though there is no overt complementizer to head the
phrase in (\ex{-1}).  Our grammar does not distinguish by category label
between the phrases that would be labeled in GB as IP and CP.  We label
both of these phrases S.  The difference between these two levels is the
presence or absence of the complementizer (or extracted WH constituent), and is
represented in our system as a difference in feature values (here, of the {\bf
$<$comp$>$} feature), and the presence of the additional structure contributed
by the complementizer or extracted constituent.  This illustrates an important
distinction in XTAG, that between features and node labels.  Because we have a
sophisticated feature system, we are able to make fine-grained distinctions
between nodes with the same label which in another system might have to be
realized by using distinguishing node labels.
 
\section{Complementizers and Embedded Clauses in English:  The
Data}
\label{data}

Verbs selecting sentential complements (or subjects) place restrictions on
their complements, in particular, on the form of the embedded verb
phrase.\footnote{Other considerations, such as the relationship between the
tense/aspect of the matrix clause and the tense/aspect of a complement clause
are also important but are not currently addressed in the current English XTAG
grammar.}  Furthermore, complementizers are constrained to appear with certain
types of clauses, again, based primarily on the form of the embedded VP.  For
example, {\it hope\/} selects both indicative and infinitival complements. With
an indicative complement, it may only have {\it that\/} or null as possible
complementizers; with an infinitival complement, it may only have a null
complementizer.  Verbs that allow wh+ complementizers, such as {\it ask}, can
take {\it whether} and {\it if} as complementizers.  The possible combinations
of complementizers and clause types is summarized in Table \ref{facts}.

As can be seen in Table \ref{facts}, sentential subjects differ from
sentential complements in requiring the complementizer {\it that\/}
for all indicative and subjunctive clauses.  In sentential complements,
{\it that\/} often varies freely with a null complementizer, as
illustrated in (\ex{1})-(\ex{6}).

\enumsentence{Christy hopes that Mike wins.}
\enumsentence{Christy hopes Mike wins.}
\enumsentence{Dania thinks that Newt is a liar.}
\enumsentence{Dania thinks Newt is a liar.}
\enumsentence{That Helms won so easily annoyed me.}
\enumsentence{$\ast$Helms won so easily annoyed me.}

\begin{table}[ht]
\centering
\begin{tabular}{|l|llllll|} \hline
Complementizer:&&that&whether&if&for&null\\
\hline
Clause type&&&&&&\\
\hline
indicative&subject&Yes&Yes&No&No&No\\
&complement&Yes&Yes&Yes&No&Yes\\
\hline
infinitive&subject&No&Yes&No&Yes&Yes\\
&complement&No&Yes&No&Yes&Yes\\
\hline
subjunctive&subject&Yes&No&No&No&No\\
&complement&Yes&No&No&No&Yes\\
\hline
gerundive\footnotemark\ &complement&No&No&No&No&Yes\\
\hline
base & complement & No & No & No & No & Yes \\
\hline
small clause & complement & No & No & No & No & Yes \\
\hline
\end{tabular}
\vspace{.2in}
\caption{Summary of Complementizer and Clause Combinations}
\label{facts}
\end{table}
\footnotetext{Most gerundive phrases are treated as NP's.  In
fact, all gerundive subjects are treated as NP's, and the only gerundive
complements which receive a sentential parse are those for which there is no
corresponding NP parse.  This was done to reduce duplication of parses. See
Chapter~\ref{gerunds-chapter} for further discussion of
gerunds.\label{gerund-footnote}}

Another fact which must be accounted for in the analysis is that in infinitival
clauses, the complementizer {\it for} must appear with an overt subject NP,
whereas a complementizer-less infinitival clause never has an overt subject, as
shown in (\ex{1})-(\ex{4}). (See section~\ref{for-complementizer} for more
discussion of the case assignment issues relating to this construction.)

\enumsentence{To lose would be awful.}
\enumsentence{For Penn to lose would be awful.}
\enumsentence{$\ast$For to lose would be awful.}
\enumsentence{$\ast$Penn to lose would be awful.}

In addition, some verbs select {\bf $<$wh$>$=+} complements (either questions
or clauses with {\it whether} or {\it if}) \cite{grimshaw90}:

\enumsentence{Jesse wondered who left.}
\enumsentence{Jesse wondered if Barry left.}
\enumsentence{Jesse wondered whether to leave.}
\enumsentence{Jesse wondered whether Barry left.}
\enumsentence{$\ast$Jesse thought who left.}
\enumsentence{$\ast$Jesse thought if Barry left.}
\enumsentence{$\ast$Jesse thought whether to leave.}
\enumsentence{$\ast$Jesse thought whether Barry left.}

\section{Features Required}
\label{s-features}

As we have seen above, clauses may be {\bf $<$wh$>$=+} or {\bf $<$wh$>$=--},
may have one of several complementizers or no complementizer, and can be of
various clause types.  The XTAG analysis uses three features to capture these
possibilities: {\bf $<$comp$>$} for the variation in complementizers,
{\bf$<$wh$>$} for the question vs.  non-question alternation and {\bf
$<$mode$>$}\footnote{{\bf $<$mode$>$} actually conflates several types of
information, in particular verb form and mood.} for clause types.  In addition
to these three features, the {\bf $<$assign-comp$>$} feature represents
complementizer requirements of the embedded verb.  More detailed discussion of
the {\bf $<$assign-comp$>$} feature appears below in the discussions of
sentential subjects and of infinitives.  The four features and their possible
values are shown in Table \ref{feat}.

\begin{table}[th]
\centering
\begin{tabular}{|l|c|} \hline
Feature&Values\\
\hline
{\bf $<$comp$>$}&that, if, whether, for, rel, nil\\
\hline
{\bf$<$mode$>$}&ind, inf, subjnt, ger, base, ppart, nom/prep\\
\hline
{\bf$<$assign-comp$>$}&that, if, whether, for, rel, ind\underline{~}nil, inf\underline{~}nil\\
\hline
{\bf$<$wh$>$}&+,--\\
\hline
\end{tabular}
\caption{Summary of Relevant Features}
\label{feat}
\end{table}

\section{Distribution of Complementizers}
\label{comp-distr}

Like other non-arguments, complementizers anchor an auxiliary tree (shown in
Figure \ref{comp-tree}) and adjoin to elementary clausal trees.  The auxiliary
tree for complementizers is the only alternative to having a complementizer
position `built into' every sentential tree.  The latter choice would mean
having an empty complementizer substitute into every matrix sentence and a
complementizerless embedded sentence to fill the substitution node.  Our choice
follows the XTAG principle that initial trees consist only of the arguments of
the anchor\footnote{See section~\ref{compl-adj} for a discussion of the
difference between complements and adjuncts in the XTAG grammar.} -- the S tree
does not contain a slot for a complementizer, and the $\beta$COMP tree has only
one argument, an S with particular features determined by the complementizer.
Complementizers select the type of clause to which they adjoin through
constraints on the {\bf $<$mode$>$} feature of the S foot node in the tree
shown in Figure~\ref{comp-tree}.  These features also pass up to the root node,
so that they are `visible' to the tree where the embedded sentence
adjoins/substitutes.

\begin{figure}[hbt]
\centering
\hspace{0.0in}
\psfig{figure=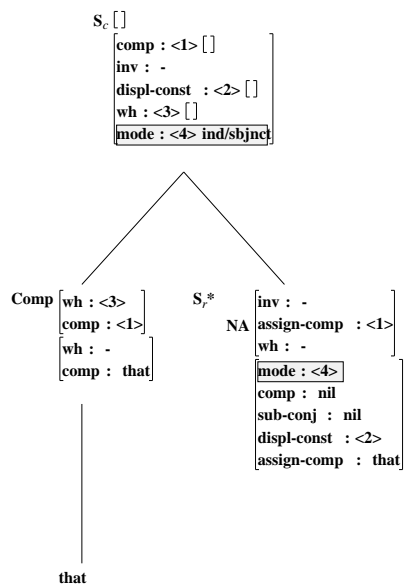,height=8.2cm}
\caption{Tree $\beta$COMPs, anchored by {\it that}}
\label{comp-tree}
\end{figure}

The grammar handles the following complementizers: {\it that\/}, {\it
whether\/}, {\it if\/}, {\it for\/}, and no complementizer, and the
clause types: indicative, infinitival, gerundive, past participial,
subjunctive and small clause ({\bf nom/prep}).  The {\bf
$<$comp$>$} feature in a clausal tree reflects the value of the
complementizer if one has adjoined to the clause. 

The {\bf $<$comp$>$} and {\bf $<$wh$>$} features receive their root
node values from the particular complementizer which anchors the tree.
The $\beta$COMPs tree adjoins to an S node with the feature {\bf
$<$comp$>$=nil}; this feature indicates that the tree does not already
{\bf have} a complementizer adjoined to it.\footnote{ Because root S's
cannot have complementizers, the parser checks that the root S has {\bf
$<$comp$>$=nil} at the end of the derivation, when the S is also checked for
a tensed verb.} We ensure that there are no stacked complementizers by
requiring the foot node of $\beta$COMPs to have {\bf $<$comp$>$=nil}.

\section{Case assignment, {\it for\/} and the two {\it to\/}'s}
\label{for-complementizer}

The {\bf $<$assign-comp$>$} feature is used to represent the
requirements of particular types of clauses for particular
complementizers.  So while the {\bf $<$comp$>$} feature represents
constraints originating from the VP dominating the clause, the {\bf
$<$assign-comp$>$} feature represents constraints originating from the
highest VP in the clause. {\bf $<$assign-comp$>$} is used to control the

the appearance of subjects in infinitival clauses (see discussion of
ECM constructions in \ref{ecm-verbs}), to block bare indicative
sentential subjects (bare infinitival subjects are allowed), and to
block `that-trace' violations.

Examples (\ex{2}), (\ex{3}) and (\ex{4}) show that an accusative
case subject is obligatory in an infinitive clause if the
complementizer {\it for\/} is present. The infinitive clauses in
(\ex{1}) is analyzed in the English XTAG grammar as
having a PRO subject.

\enumsentence{Christy wants to pass the exam.}
\enumsentence{Mike wants for her to pass the exam.}
\enumsentence{$\ast$Mike wants for she to pass the exam.}
\enumsentence{$\ast$Christy wants for to pass the exam.}
 
The {\it for-to\/} construction is particularly illustrative of the
difficulties and benefits faced in using a lexicalized grammar.  It is
commonly accepted that {\it for\/} behaves as a case-assigning
complementizer in this construction, assigning accusative case to the
`subject' of the clause since the infinitival verb does not assign
case to its subject position.  However, in our featurized grammar, the
absence of a feature licenses anything, so we must have overt null
case assigned by infinitives to ensure the correct distribution of PRO
subjects. (See section~\ref{case-assignment} for more discussion of
case assignment.)  This null case assignment clashes with accusative
case assignment if we simply add {\it for\/} as a standard
complementizer, since NP's (including PRO) are drawn from the lexicon
already marked for case.  Thus, we must use the {\bf
$<$assign-comp$>$} feature to pass information about the verb up to
the root of the embedded sentence.  To capture these facts, two
infinitive {\it to}'s are posited. One infinitive {\it to\/} has {\bf
$<$assign-case$>$=none} which forces a PRO subject, and {\bf
$<$assign-comp$>$=inf\_nil} which prevents {\it for\/} from
adjoining. The other infinitive {\it to\/} has no value at all for
{\bf $<$assign-case$>$} and has {\bf $<$assign-comp$>$=for/ecm}, so that
it can only occur either with the complementizer {\it for\/} or with
ECM constructions. In those
instances either {\it for} or the ECM verb
supplies the {\bf $<$assign-case$>$} value, assigning
 accusative case to the overt subject.
 
\section{Sentential Complements of Verbs}
\label{sent-complements}
{\bf Tree families}: Tnx0Vs1, Tnx0Vnx1s2, TItVnx1s2, TItVpnx1s2, TItVad1s2.

Verbs that select sentential complements restrict the {\bf $<$mode$>$}
and {\bf $<$comp$>$} values for those complements. Since with very few
exceptions\footnote{For example, long distance extraction is not
possible from the S complement in it-clefts.} long distance extraction
is possible from sentential complements, the S complement nodes are
adjunction nodes. Figure \ref{think} shows the declarative tree
for sentential complements, anchored by {\it think}.  

\begin{figure}[hbt]
\centering
\hspace{0.0in}
\psfig{figure=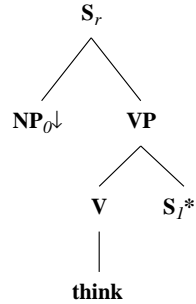,height=1.7in}
\caption{Sentential complement tree: $\beta$nx0Vs1}
\label{think}
\label{2;1,10}
\end{figure}

The need for an adjunction node rather than a substitution node at S$_{1}$ may
not be obvious until one considers the derivation of sentences with long
distance extractions.  For example, the declarative in (\ex{1}) is derived by
adjoining the tree in Figure~\ref{aard-emu}(b) to the S$_{1}$ node of the tree
in Figure~\ref{aard-emu}(a).  Since there are no bottom features on S$_{1}$,
the same final result could have been achieved with a substitution node at
S$_{1}$.

\enumsentence{The emu thinks that the aardvark smells terrible.}

\begin{figure}[htb]
\centering
\begin{tabular}{ccc}
\psfig{figure=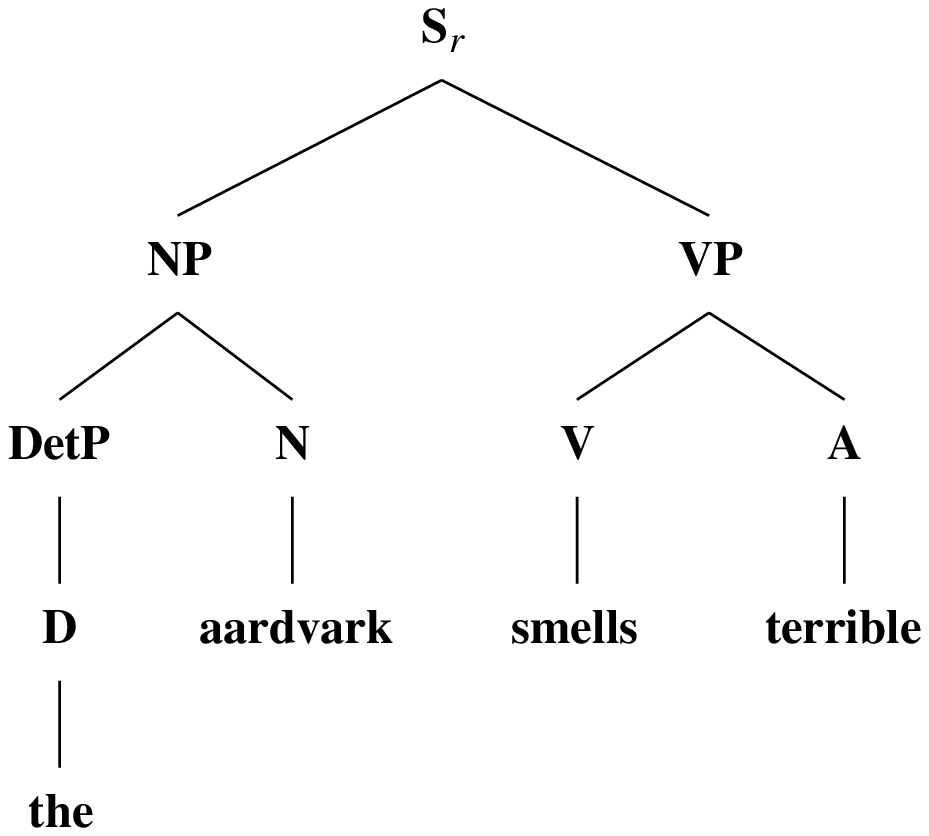,height=2.1in}&
\hspace{0.3in}&
\psfig{figure=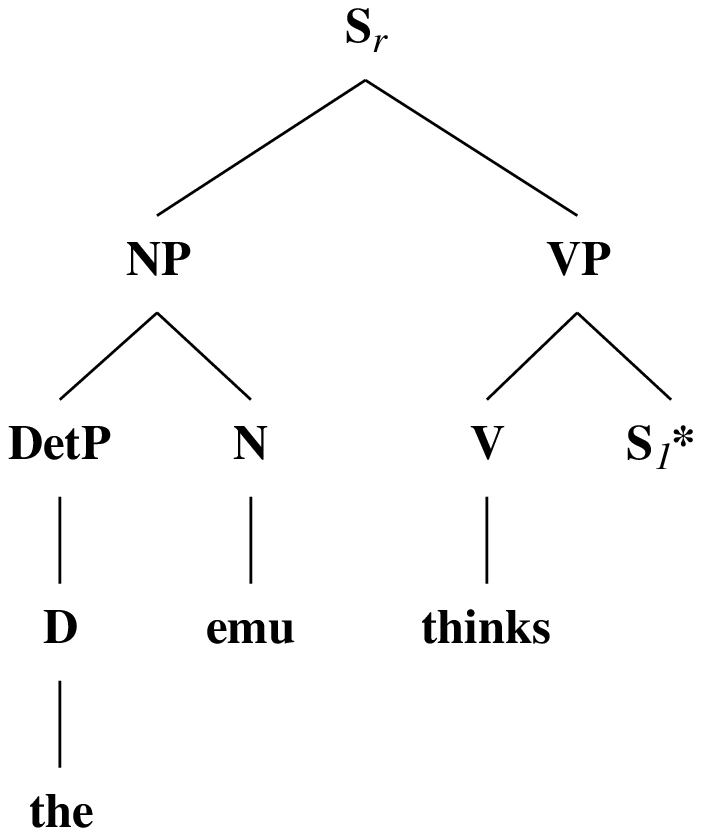,height=2.1in}\\
(a)&&(b)\\
\end{tabular}
\caption{Trees for {\it The emu thinks that the aardvark smells terrible.}}  
\label{aard-emu}
\label{1;4,4}
\end{figure}

However, adjunction is crucial in deriving sentences with
long distance extraction, as in sentences (\ex{1}) and (\ex{2}).  

\enumsentence{Who does the emu think smells terrible?}
\enumsentence{Who did the elephant think the panda heard the emu say
smells terrible?} 

The example in (\ex{-1}) is derived from the trees for {\it who smells
terrible?}  shown in Figure ~\ref{who-smells} and {\it the emu thinks} S shown
in Figure~\ref{aard-emu}(b), by adjoining the latter at the S$_r$ node of the
former.\footnote{See Chapter~\ref{auxiliaries} for a discussion of do-support.}
This process is recursive, allowing sentences like (\ex{0}). Such a
representation has been shown by \cite{kj85} to be well-suited for describing
unbounded dependencies.

\begin{figure}[thb]
\centering
\hspace{0.0in}
\psfig{figure=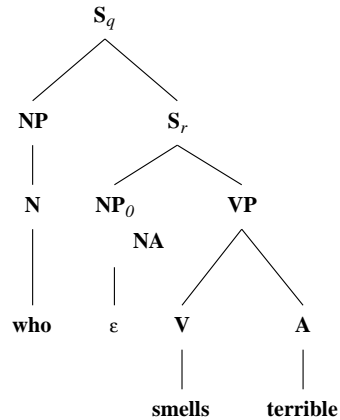,height=2.3in}
\caption{Tree for {\it Who smells terrible?}}
\label{who-smells}
\label{1;4,14}
\end{figure}

In English, a complementizer may not appear on a complement with an extracted
subject (the `that-trace' configuration). This phenomenon
is illustrated in (\ex{1})-(\ex{3}):

\enumsentence{Which animal did the giraffe say that he likes?}
\enumsentence{$\ast$Which animal did the giraffe say that likes him?}
\enumsentence{Which animal did the giraffe say likes him?}

These sentences are derived in XTAG by adjoining the tree for {\it did the
giraffe say} S at the S$_r$ node of the tree for either {\it which animal likes
him} (to yield sentence~(\ex{0})) or {\it which animal he likes} (to yield
sentence~(\ex{-2})).  That-trace violations are blocked by the presence of the
feature {\bf $<$assign-comp$>$=inf\underline{~}nil/ind\underline{~}nil/ecm}
feature on the bottom of the S$_r$ node of trees with extracted subjects (W0),
i.e. those used in sentences such as (\ex{-1}) and (\ex{0}).  
If a complementizer tree, $\beta$COMPs, adjoins to a subject
extraction tree at $S_r$, its {\bf $<$assign-comp$>$ =
that/whether/for/if} feature will clash and the derivation will
fail. If there is no complementizer, there is no feature clash, and this will
permit the derivation of sentences like (\ex{0}), or of ECM constructions, in
which case the ECM verb will have {\bf $<$assign-comp$>$=ecm} (see
section~\ref{ecm-verbs} for more discussion of the ECM case).
Complementizers may adjoin normally to object extraction trees such as those
used in sentence~(\ex{-2}), and so object extraction trees have no value 
for the {\bf $<$assign-comp$>$} feature.

In the case of indirect questions, subjacency follows from the
principle that a given tree cannot contain more than one
wh-element. Extraction out of an indirect question is ruled out
because a sentence like:

\enumsentence{$\ast$ Who$_{i}$ do you wonder who$_{j}$ e$_{j}$ loves e$_{i}$ ?}

\noindent would have to be derived from the adjunction of {\it do you
wonder} into {\it who$_{i}$ who$_{j}$ e$_{j}$ loves e$_{i}$}, which is an
ill-formed elementary tree.\footnote{This does not mean that elementary trees
with more than one gap should be ruled out across the grammar. Such trees might
be required for dealing with parasitic gaps or gaps in coordinated structures.}

\subsection{Exceptional Case Marking Verbs}
\label{ecm-verbs}

{\bf Tree family}: TXnx0Vs1
Exceptional Case Marking verbs are those which assign accusative case to the
subject of the sentential complement. This is in contrast to verbs
in the Tnx0Vnx1s2 family (section~\ref{nx0Vnx1s2-family}), which assign 
accusative case to an NP which is not part of the sentential complement.  

The subject of an ECM infinitive
complement is assigned accusative case is a manner
analogous to that of a subject in a {\it for-to\/} construction, as described
in section~\ref{for-complementizer}.  As in the {\it for-to\/} case, the
ECM verb assigns accusative case into the subject of the lower infinitive, and
so the infinitive uses the {\it to} which has no value for
{\bf $<$assign-case$>$} and has {\bf $<$assign-comp$>$=for/ecm}.  The ECM verb
has {\bf $<$assign-comp$>$=ecm} and {\bf $<$assign-case$>$=acc} on its
foot.  The former allows the {\bf $<$assign-comp$>$} features of the ECM
verb and the {\it to} tree to unify, and so be used together, and the latter
assigns the accusative case to the lower subject.  

Figure~\ref{expects-decl} shows the 
declarative tree for the 
tree for the TXnx0Vs1 family, in this case anchored by {\it expects}.
Figure~\ref{van-expects} shows a parse for {\it Van expects Bob to talk}

\begin{figure}[hbt]
\centering
\hspace{0.0in}
\psfig{figure=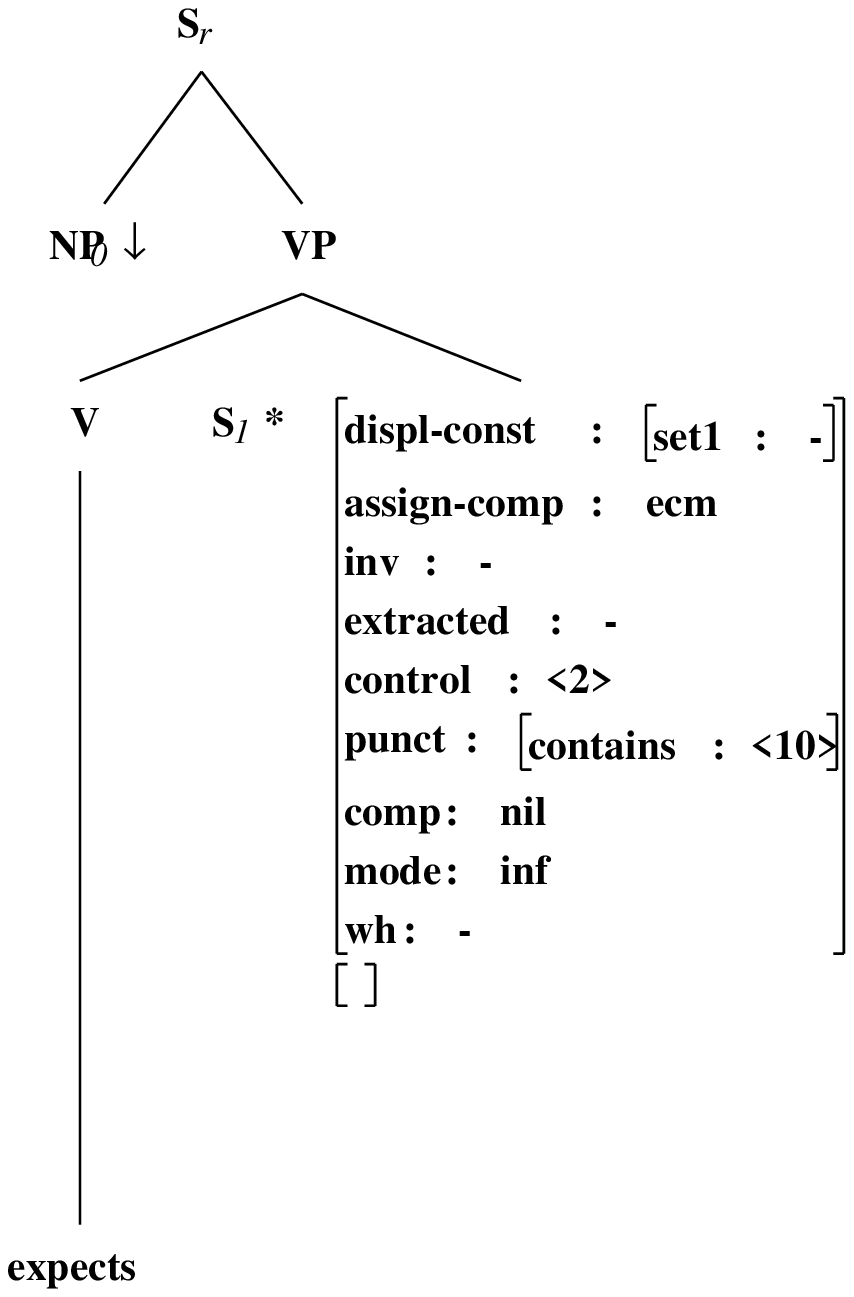,height=3.3in}
\caption{ECM tree: $\beta$Xnx0Vs1}
\label{expects-decl}
\label{3;1,15}
\end{figure}

\begin{figure}[hbt]
\centering
\hspace{0.0in}
\psfig{figure=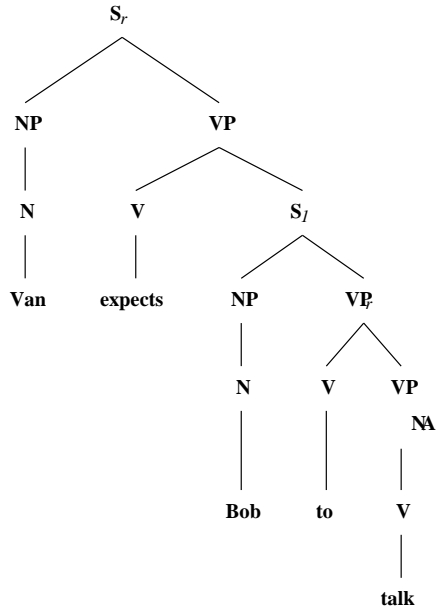,height=3.3in}
\caption{Sample ECM parse}
\label{van-expects}
\end{figure}

The ECM and {\it for-to\/} cases are analogous in how they are used together
with the correct infinitival {\it to} to assign accusative case to the 
subject of the lower infinitive.  However, they are different in that
{\it for} is blocked along with other complementizers in subject extraction
contexts, as discussed in section~\ref{sent-complements}, as in
(\ex{1}), while subject extraction is compatible with ECM cases, 
as in (\ex{2}).

\enumsentence{$\ast$What child did the giraffe ask for to leave?}
\enumsentence{Who did Bill expect to eat beans?}

Sentence (\ex{-1}) is ruled out by the {\bf $<$assign-comp$>$=
inf\underline{~}nil/ind\underline{~}nil/ecm} feature
on the subject extraction tree for {\it ask}, since the
{\bf $<$assign-comp$>$=for} feature from the {\it for} tree will fail to 
unify.  However, 
(\ex{0}) will be allowed since {\bf $<$assign-comp$>$=ecm} feature on the
{\it expect} tree will unify with the foot of the ECM verb tree.  
The use of features allows the ECM and
{\it for-to\/} constructions to act the same for exceptional case assignment,
while also being distinguished for that-trace violations.

Verbs that take bare infinitives, as in (\ex{1}), are 
also treated as ECM verbs,
the only difference being that their foot feature has
{\bf $<$mode$>$=base} instead of {\bf $<$mode$>$=inf}.  Since the complement
does not have {\it to}, there is no question of using the {\it to} tree
for allowing accusative case to be assigned.  Instead, verbs with
{\bf $<$mode$>$=base} allow either accusative or nominative case to be 
assigned to the subject, and the foot of the ECM bare infinitive tree 
forces it to be accusative by its {\bf $<$assign-case$>$=acc} value at its
foot node unifies with the {\bf $<$assign-case$>$=nom/acc} value of the 
bare infinitive clause.

\enumsentence{Bob sees the harmonica fall.}

The trees in the TXnx0Vs1 family are generally parallel to those in the
Tnx0Vs1 family, except for the {\bf $<$assign-case$>$} and
{\bf $<$assign-comp$>$} values on the foot nodes.  However, the TXnx0Vs1
family also includes a tree for the passive, which of course is not
included in the Tnx0Vs1 family.  Unlike all the other trees in the 
TXnx0Vs1 family, the passive tree is not rooted in S, and is instead a 
VP auxiliary tree.  Since the
subject of the infinitive is not thematically selected by the ECM verb,
it is not part of the ECM verb's tree, and so it cannot be part of the
passive tree.   Therefore, the passive acts as a raising verb (see 
section~\ref{sm-clause-xtag-analysis}).
For example, to derive (\ex{2}), the tree in Figure~\ref{expects-passive}
would adjoin into a 
derivation for {\it Bob to talk} at the VP node  (and the 
{\bf $<$mode$>$=passive} feature, not shown, forces the auxiliary to
adjoin in, as for other passives, as described in chapter~\ref{passives}).

\enumsentence{Van expects Bob to talk.}
\enumsentence{Bob was expected to talk.}

\begin{figure}[hbt]
\centering
\hspace{0.0in}
\psfig{figure=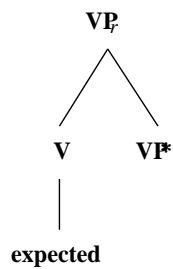,height=1.5in}
\caption{ECM passive}
\label{expects-passive}
\label{3;2,15}
\end{figure}

It has long been noted that passives of both full and bare infinitive 
ECM constructions are full infinitives, as in (\ex{0}) and (\ex{2}).

\enumsentence{Bob sees the harmonica fall.}
\enumsentence{The harmonica was seen to fall.}
\enumsentence{$\ast$The harmonica was seen fall.}

Under the TAG ECM analysis, this fact is easy to implement.  The foot
node of the ECM passive tree is simply set to have {\bf $<$mode$>$=inf},
which prevents the derivation of (\ex{0}).  Therefore, for all the other
trees in the family, to foot nodes are set to have {\bf $<$mode$>$=base} 
or {\bf $<$mode$>$=inf} depending on whether it is a bare infinitive or not.
These foot nodes are all S nodes.  The VP foot node of the passive tree, 
however, has {\bf $<$mode$>$=inf} regardless.

\section{Sentential Subjects}
\label{sent-subjs}

{\bf Tree families}: Ts0Vnx1, Ts0Ax1, Ts0N1, Ts0Pnx1, Ts0ARBPnx1, 
Ts0PPnx1, Ts0PNaPnx1, Ts0V, Ts0Vtonx1, Ts0NPnx1, Ts0APnx1, Ts0A1s1.

Verbs that select sentential subjects anchor trees that have an S node
in the subject position rather than an NP node.  Since extraction is
not possible from sentential subjects, they are implemented as
substitution nodes in the English XTAG grammar.  Restrictions on
sentential subjects, such as the required {\it that} complementizer for
indicatives, are enforced by feature values specified on the S
substitution node in the elementary tree.  

Sentential subjects behave essentially like sentential complements, with a few
exceptions.  In general, all verbs which license sentential subjects license
the same set of clause types. Thus, unlike sentential complement verbs which
select particular complementizers and clause types, the matrix verbs licensing
sentential subjects merely license the S argument. Information about the
complementizer or embedded verb is located in the tree features, rather than in
the features of each verb selecting that tree.  Thus, all sentential subject
trees have the same {\bf $<$mode$>$}, {\bf $<$comp$>$} and {\bf
$<$assign-comp$>$} values shown in Figure~\ref{comparison}(a).

\begin{figure}[htb]
\centering
\begin{tabular}{ccc}
\psfig{figure=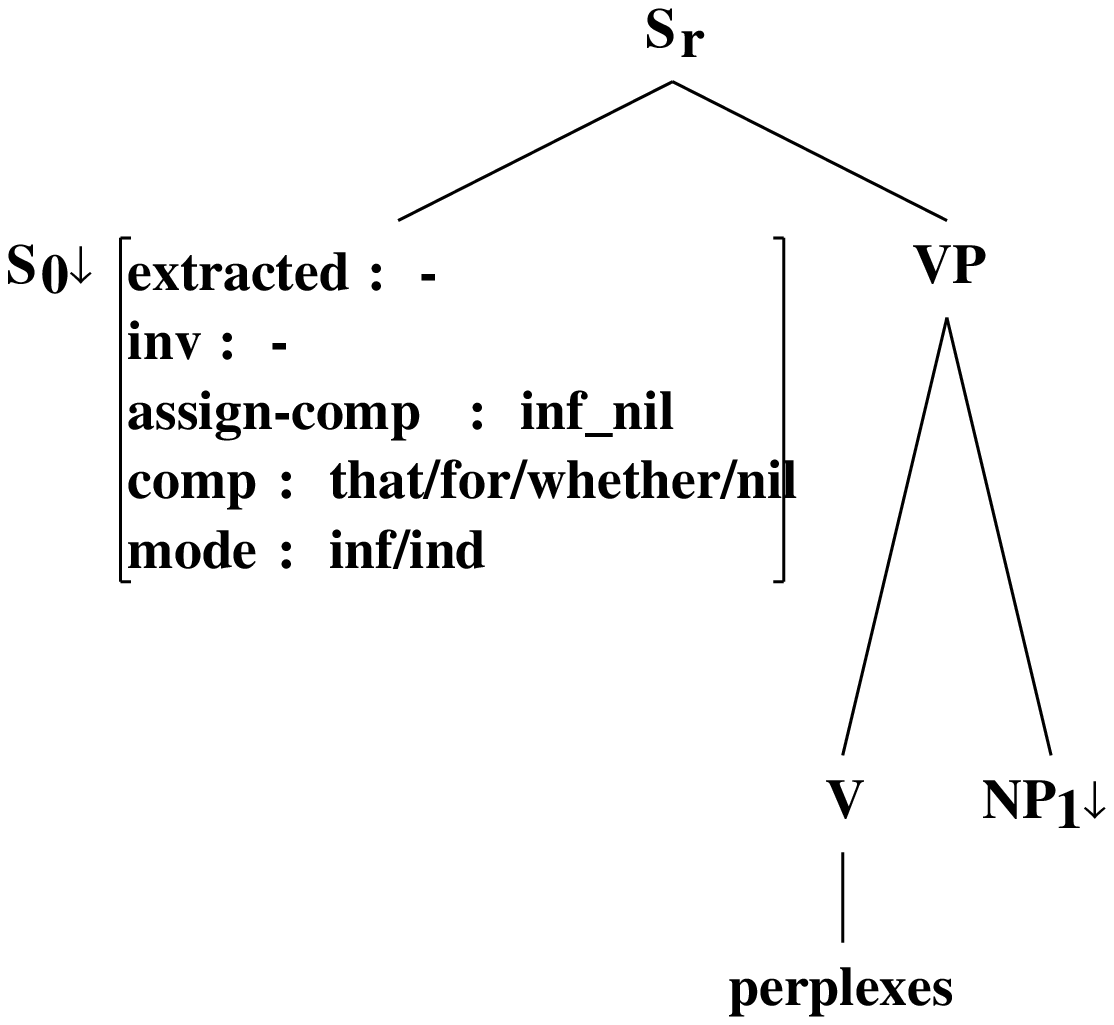,height=2.2in}&
\hspace{0.5in}&
\psfig{figure=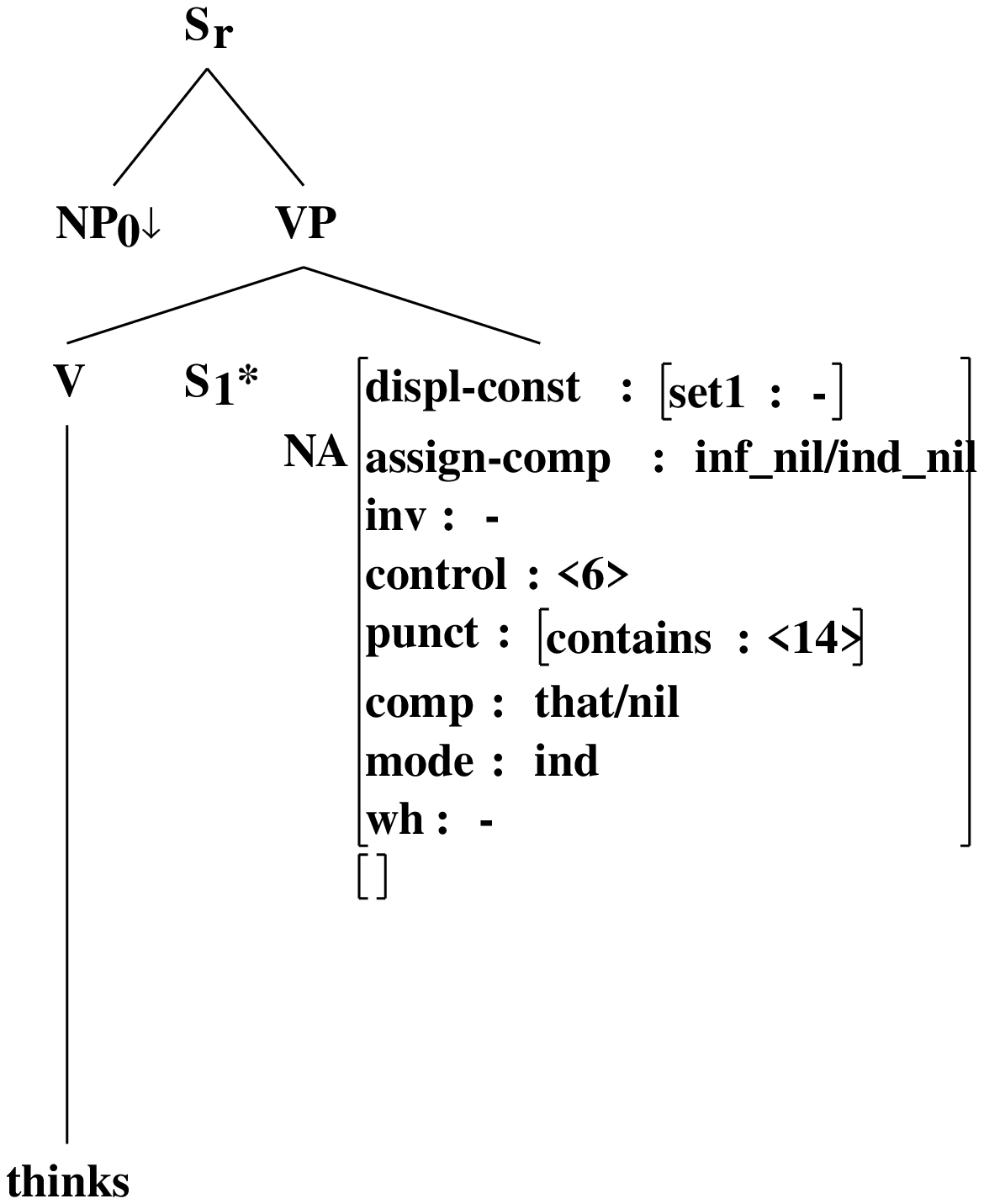,height=2.6in}\\
(a)&&(b)\\
\end{tabular}
\caption{Comparison of {\bf $<$assign-comp$>$} values for sentential
subjects: $\alpha$s0Vnx1 (a) and sentential complements: $\beta$nx0Vs1 (b)}
\label{comparison}
\label{1;1,16}
\end{figure}

The major difference in clause types licensed by S-subjs and S-comps is that
indicative S-subjs obligatorily have a complementizer (see examples in
section~\ref{data}). The {\bf $<$assign-comp$>$} feature is used here to
license a null complementizer for infinitival but not indicative clauses. {\bf
$<$assign-comp$>$} has the same possible values as {\bf $<$comp$>$}, with the
exception that the {\bf nil} value is `split' into {\bf ind\_nil} and {\bf
inf\_nil}.  This difference in feature values is illustrated in
Figure~\ref{comparison}.

Another minor difference is that {\it whether\/} but not {\it if\/} is
grammatical with S-subjs.\footnote{Some speakers also find {\it if\/}
  as a complementizer only marginally grammatical in S-comps.} Thus,
{\it if} is not among the {\bf $<$comp$>$} values allowed in S-subjs.
The final difference from S-comps is that there are no S-subjs with
{\bf $<$mode$>$=ger}. As noted in footnote~\ref{gerund-footnote} of
this chapter, gerundive complements are only allowed when there is no
corresponding NP parse. In the case of gerundive S-subjs, there is
always an NP parse available.

\section{Nouns and Prepositions taking Sentential Complements}
\label{NPA}

{\bf Trees}: $\alpha$NXNs, $\beta$vxPs, $\beta$Pss, $\beta$nxPs,
Tnx0N1s1, Tnx0A1s1.

\begin{figure}[thb]
\centering
\begin{tabular}{ccc}
\psfig{figure=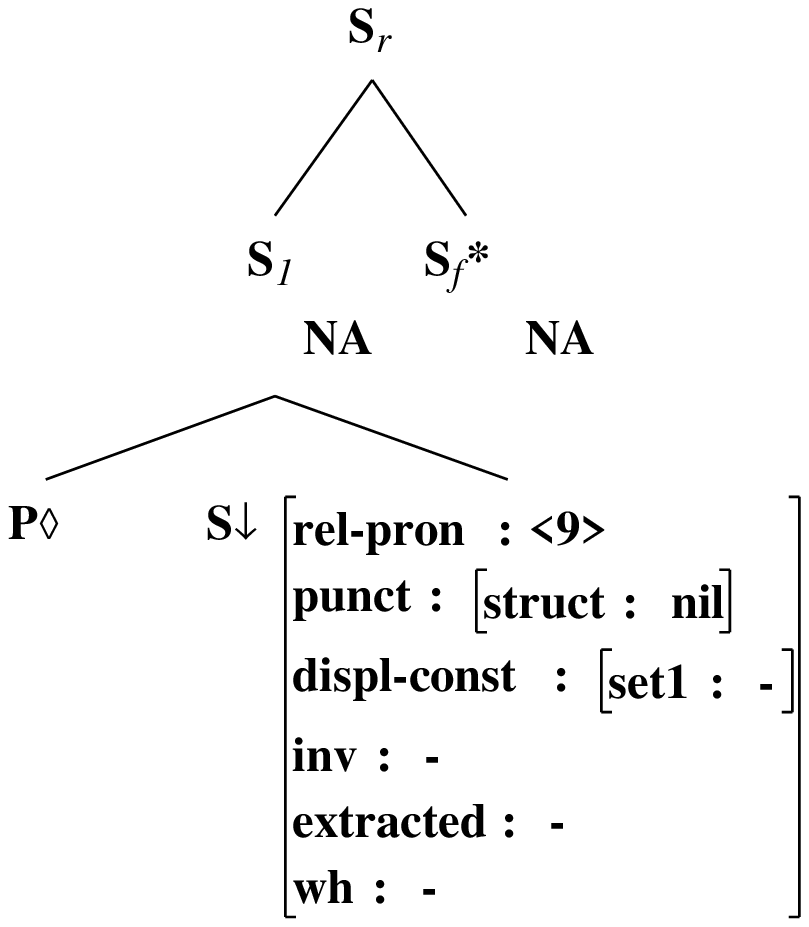,height=5.6cm}&
\hspace{0.3in}&
\psfig{figure=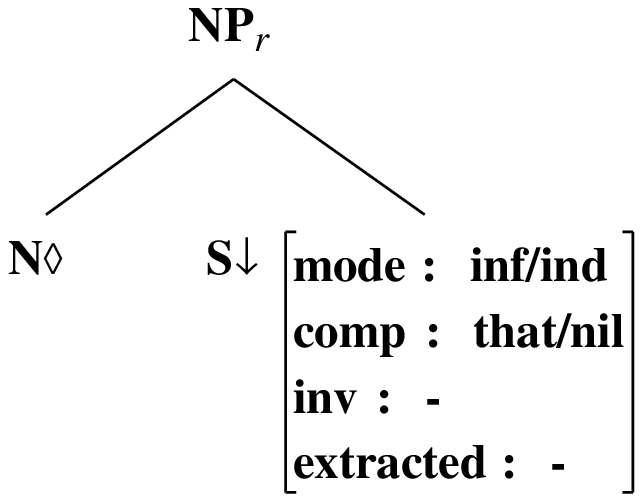,height=4cm}
\\
(a) && (b)\\
\end{tabular}
\caption{Sample trees for preposition: $\beta$Pss (a) and noun: $\alpha$NXNs (b) taking
sentential complements}
\label{nounprep}
\end{figure}

Prepositions and nouns can also select sentential complements, using
the trees listed above.  These trees use the {\bf $<$mode$>$} and {\bf
$<$comp$>$} features as shown in Figure~\ref{nounprep}.  For example,
the noun {\it claim} takes only indicative complements with {\it
that}, while the preposition {\it with} takes small clause
complements, as seen in sentences (\ex{1})-(\ex{4}).

\enumsentence{Beth's claim that Clove was a smart dog....}
\enumsentence{$\ast$Beth's claim that Clove a smart dog....}
\enumsentence{Dania wasn't getting any sleep with Doug sick.}
\enumsentence{$\ast$Dania wasn't getting any sleep with Doug was sick.}

\section{PRO control}
\label{PRO-control}

\subsection{Types of control}

In the literature on control, two types are often distinguished: obligatory
control, as in sentences~(\ex{1}), (\ex{2}), (\ex{3}), and (\ex{4}) and optional 
control, as in sentence~(\ex{5}).

\enumsentence{Srini$_i$ promised Mickey$_{i}$ [PRO$_i$ to leave].}
\enumsentence{Srini persuaded Mickey$_{i}$ [PRO$_i$ to leave].}
\enumsentence{Srini$_{i}$ wanted [PRO$_i$ to leave].}
\enumsentence{Christy$_{i}$ left the party early [PRO$_i$ to go to the airport].}
\enumsentence{[PRO$_{arb/i}$ to dance] is important for Bill$_{i}$.}

At present, an analysis for obligatory control into complement clauses
(as in sentences~(\ex{-4}), (\ex{-3}), and (\ex{-2})) has been implemented. An
analysis for cases of obligatory control into adjunct clauses and optional
control exists and can be found in \cite{bhatt94}.

\subsection{A feature-based analysis of PRO control}
The analysis for obligatory control involves co-indexation of the control feature
of the NP anchored by PRO to the control feature of the controller.
A feature equation in the tree anchored by the control verb 
co-indexes the control feature of the controlling NP with the foot
node of the tree.  All sentential trees have a co-indexed
control feature from the root S to the subject NP. 

When the tree containing the controller adjoins
onto the complement clause tree containing the PRO, 
the features of the foot node of the
auxiliary tree are unified with the bottom features of the root node of the 
complement clause
tree containing the PRO. This leads to the control feature of the controller
being co-indexed with the control feature of the PRO.

Depending on the choice of the controlling verb, the control
propagation paths in the auxiliary trees are different.  In the case of subject
control (as in sentence~(\ex{-3})), the subject NP and the foot node are
have co-indexed control features, while for object control
(e.g. sentence~(\ex{-4}), the object NP and the foot node are 
co-indexed for control. Among verbs that belong to the Tnx0Vnx1s2 family,
i.e. verbs that take an NP object and a clausal complement, subject-control
verbs form a distinct minority, {\em promise} being the only commonly used
verb in this class.

Consider the derivation of sentence~(\ex{-3}). The auxiliary tree for
{\em persuade}, shown in Figure \ref{persuade-tree}, has the following
feature equation~(\ex{1}).
\enumsentence{  NP$_{1}$:{\bf $<$control$>$} = S$_{2}$.t:{\bf $<$control$>$} }
The auxiliary tree adjoins into the tree for {\em leave}, shown in
Figure \ref{leave-tree}, which 
has the following feature equation~(\ex{1}).
\enumsentence{S$_{r}$.b:{\bf $<$control$>$} = NP$_{0}$.t:{\bf $<$control$>$}}
Since the adjunction takes place at the root node (S$_{r}$) of the
{\em leave} tree, after unification, NP$_{1}$ of the {\em persuade}
tree and NP$_{0}$ of the {\em leave} tree share a control feature. The
resulting derived and derivation trees are shown in Figures
\ref{derived-tree} and \ref{derivation-tree}.

\begin{figure}[hbt]
\centering
\hspace{0.0in}
\psfig{figure=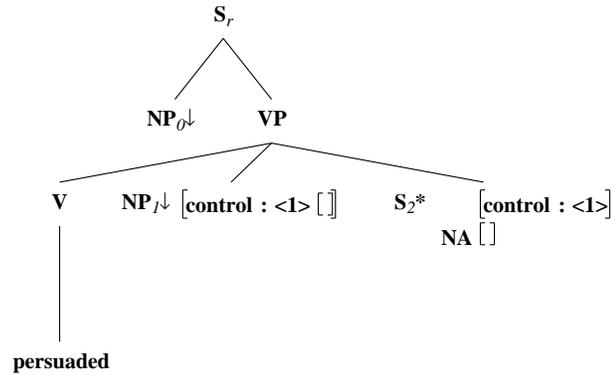,height=5.2cm}
\caption{Tree for {\it persuaded}}
\label{persuade-tree}
\end{figure}

\begin{figure}[hbt]
\centering
\hspace{0.0in}
\psfig{figure=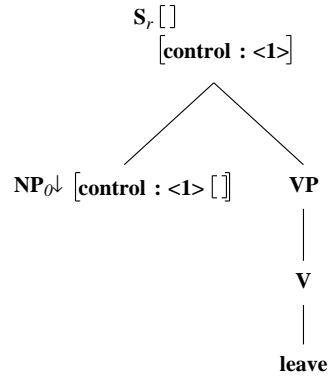,height=5.2cm}
\caption{Tree for {\it leave}}
\label{leave-tree}
\end{figure}

\begin{figure}[hbt]
\centering
\hspace{0.0in}
\psfig{figure=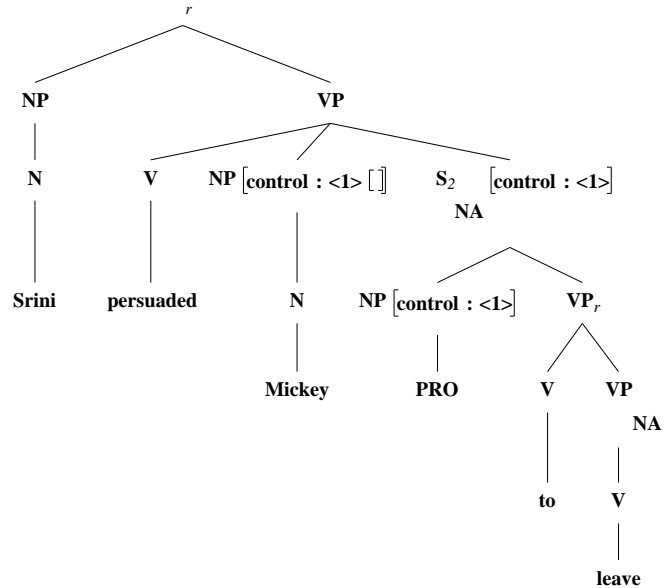,height=8.2cm}
\caption{Derived tree for {\it Srini persuaded Mickey to leave}}
\label{derived-tree}
\end{figure}

\begin{figure}[hbt]
\centering
\hspace{0.0in}
\psfig{figure=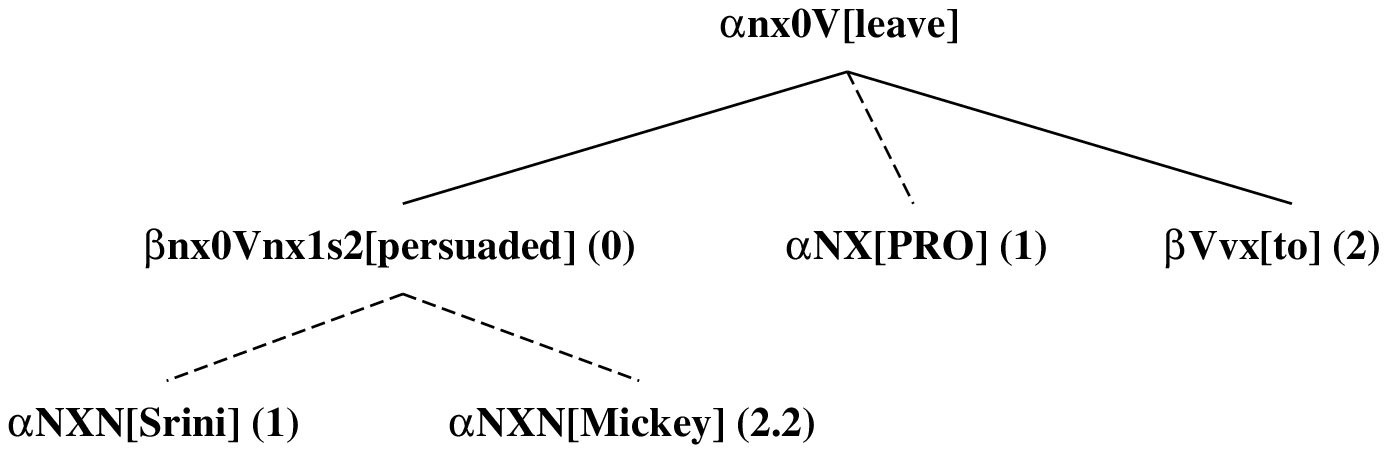,height=4.2cm}
\caption{Derivation tree for {\it Srini persuaded Mickey to leave}}
\label{derivation-tree}
\end{figure}

\subsection{The nature of the control feature}
The control feature does not have any value and is used only for
co-indexing purposes. If two NPs have their control features
co-indexed, it means that they are participating in a relationship
of control; the c-commanding NP controls the c-commanded NP. 

\subsection{Long-distance transmission of control features}
Cases involving embedded infinitival complements with PRO subjects such as
(\ex{1}) can also be handled.

\enumsentence{ John$_i$ wants [PRO$_i$ to want [PRO$_i$ to dance]].}

The control feature of `John' and the two PRO's all get co-indexed.
This treatment might appear to lead to a problem. Consider (\ex{1}):

\enumsentence{ John$_{*i}$ wants [Mary$_i$ to want [PRO$_i$ to dance]].}

If both the `want' trees have the control feature of their subject
co-indexed to their foot nodes, we would have a situtation where
the PRO is co-indexed for control feature with `John', as well as with `Mary'. 
Note that the higher `want' in (\ex{-1}) is {\em want$_{ECM}$} 
- it assigns case
to the subject of the lower clause while the lower `want' in (\ex{-1}) is 
not. Subject control is restricted
to non-ECM (Exceptional Case Marking) verbs that take infinitival 
complements. Since the two `want's in (\ex{-1}) are different with
respect to their control (and other) properties, the control feature
of PRO stops at `Mary' and is not transmitted to the higher clause.

\subsection{Locality constraints on control}
PRO control obeys locality constraints. The controller for PRO has to be
in the immediately higher clause. Consider the ungrammatical sentence~(\ex{1})
((\ex{1}) is ungrammatical only with the co-indexing indicated below).
\enumsentence{* John$_i$ wants [PRO$_i$ to persuade Mary$_i$ [PRO$_i$ to dance]]}
However, such a derivation is ruled out automatically by the 
mechanisms of a TAG derivation and feature unification. 
Suppose it was possible to first compose the {\em want} tree with the
{\em dance} tree and then insert the {\em persuade} tree. (This is not
possible in the XTAG grammar because of the convention that
auxiliary trees have NA (Null Adjunction) constraints on their foot nodes.)
Even then, at the end of the derivation the control feature of the 
subject of {\em want} would end up co-indexed with the PRO subject of
{\em persuade} and the control feature of {\em Mary} would be co-indexed with the
PRO subject of {\em dance} as desired. There is no way to generate the illegal
co-indexing in (\ex{-1}). Thus the locality constraints on PRO control 
fall out from the mechanics of TAG derivation and feature unification.

\section{Reported speech}

Reported speech is handled in the XTAG grammar by having the reporting
clause adjoin into the quote. Thus, the reporting clause is an
auxiliary tree, anchored by the reporting verb. See \cite{doran-diss}
for details of the analysis. There are trees in both the Tnx0Vs1 and
Tnx0nx1s2 families to handle reporting clauses which precede, follow
and come in the middle of the quote.

\chapter{The English Copula, Raising Verbs, and Small Clauses}
\label{small-clauses}

The English copula, raising verbs, and small clauses are all handled in XTAG by
a common analysis based on sentential clauses headed by non-verbal elements.
Since there are a number of different analyses in the literature of how these
phenomena are related (or not), we will present first the data for all three
phenomena, then various analyses from the literature, finishing with the
analysis used in the English XTAG grammar.\footnote{This chapter is strongly
based on \cite{heycock91}.  Sections \ref{sm-clause-data} and
\ref{sm-clause-other-analyses} are greatly condensed from her paper, while the 
description of the XTAG analysis in section \ref{sm-clause-xtag-analysis} is an
updated and expanded version.}

\section{Usages of the copula, raising verbs, and small clauses}
\label{sm-clause-data}

\subsection{Copula}
\label{copula-data}

The verb {\it be} as used in sentences ({\ex{1}})-({\ex{3}}) is often
referred to as the \xtagdef{copula}.  It can be followed by a noun, adjective, or
prepositional phrase.

\enumsentence{Carl is a jerk .}
\enumsentence{Carl is upset .}
\enumsentence{Carl is in a foul mood .}

Although the copula may look like a main verb at first glance, its syntactic
behavior follows the auxiliary verbs rather than main verbs.  In particular,

\begin{itemize}
\item Copula {\it be} inverts with the subject.
\enumsentence{is Beth writing her dissertation ?\\
		is Beth upset ?\\
		$\ast$wrote Beth her dissertation ?}

\item Copula {\it be} occurs to the left of the negative marker {\it
not}.
\enumsentence{Beth is not writing her dissertation .\\
		Beth is not upset .\\
		$\ast$Beth wrote not her dissertation .}

\item Copula {\it be} can contract with the negative marker {\it not}.
\enumsentence{Beth isn't writing her dissertation .\\
		Beth isn't upset .\\
		$\ast$Beth wroten't her dissertation .}

\item Copula {\it be} can contract with pronominal subjects.
\enumsentence{She's writing her dissertation .\\
		She's upset .\\
		$\ast$She'ote her dissertation .}

\item Copula {\it be} occurs to the left of adverbs in the unmarked order.
\enumsentence{Beth is often writing her dissertation .\\
		Beth is often upset .\\
		$\ast$Beth wrote often her dissertation .}
\end{itemize}

Unlike all the other auxiliaries, however, copula {\it be} is not followed by
a verbal category (by definition) and therefore must be the rightmost verb.  In
this respect, it is like a main verb.

The semantic behavior of the copula is also unlike main verbs.  In particular,
any semantic restrictions or roles placed on the subject come from the
complement phrase (NP, AP, PP) rather than from the verb, as illustrated in
sentences ({\ex{1}}) and ({\ex{2}}).  Because the complement phrases predicate
over the subject, these types of sentences are often called
\xtagdef{predicative} sentences.

\enumsentence{The bartender was garrulous .}
\enumsentence{?The cliff was garrulous .}

\subsection{Raising Verbs}
\label{raising-verbs}

Raising verbs are the class of verbs that share with the copula the property
that the complement, rather than the verb, places semantic constraints on
the subject.  

\enumsentence{Carl seems a jerk .\\
		Carl seems upset .\\
		Carl seems in a foul mood .}

\enumsentence{Carl appears a jerk .\\
		Carl appears upset .\\
		Carl appears in a foul mood .}

The raising verbs are similar to auxiliaries in that they order with other
verbs, but they are unique in that they can appear to the left of the
infinitive, as seen in the sentences in ({\ex{1}}).  They cannot, however,
invert or contract like other auxiliaries ({\ex{2}}), and they appear to the
right of adverbs ({\ex{3}}).

\enumsentence{Carl seems to be a jerk .\\
		Carl seems to be upset .\\
		Carl seems to be in a foul mood .}

\enumsentence{$\ast$seems Carl to be a jerk ?\\
		$\ast$Carl seemn't to be upset .\\
		$\ast$Carl`ems to be in a foul mood .}

\enumsentence{Carl often seems to be upset .\\
		$\ast$Carl seems often to be upset .}

\subsection{Small Clauses}

One way of describing small clauses is as predicative sentences without the
copula.  Since matrix clauses require tense, these clausal structures appear
only as embedded sentences.  They occur as complements of certain verbs, each
of which may allow certain types of small clauses but not others, depending on its
lexical idiosyncrasies.

\enumsentence{I consider [Carl a jerk] .\\
		I consider [Carl upset] .\\
		?I consider [Carl in a foul mood] .}

\enumsentence{I prefer [Carl in a foul mood] .\\
		??I prefer [Carl upset] .}

\subsection{Raising Adjectives}
\label{raising-adjs}

Raising adjectives are the class of adjectives that 
share with the copula and raising verbs the property
that the complement, rather than the verb, places semantic constraints on
the subject.  

They appear with the copula in a matrix clause, as in ({\ex{1}}).  However,
in other cases, such as that of small clauses ({\ex{2}}), they do not
have to appear with the copula.

\enumsentence{Carl is likely to be a jerk .\\
		Carl is likely to be upset .\\
		Carl is likely to be in a foul mood .\\
                Carl is likely to perjure himself .}

\enumsentence{I consider Carl likely to perjure himself .}

\section{Various Analyses}
\label{sm-clause-other-analyses}

\subsection{Main Verb Raising to INFL + Small Clause}

In \cite{pollack89} the copula is generated as the head of a VP, like any main
verb such as {\it sing} or {\it buy}. Unlike all other main verbs\footnote{with
the exception of {\it have} in British English. See
footnote~\ref{have-footnote} in Chapter~\ref{auxiliaries}.}, however, {\it be}
moves out of the VP and into Infl in a tensed sentence.  This analysis aims to
account for the behavior of {\it be} as an auxiliary in terms of inversion,
negative placement and adverb placement, while retaining a sentential structure
in which {\it be} heads the main VP at D-Structure and can thus be the only
verb in the clause.

Pollock claims that the predicative phrase is not an argument of {\it be},
which instead he assumes to take a small clause complement, consisting of a
node dominating an NP and a predicative AP, NP or PP. The subject NP of the
small clause then raises to become the subject of the sentence.  This accounts
for the failure of the copula to impose any selectional restrictions on the
subject.  Raising verbs such as {\it seem} and {\it appear}, presumably, take the
same type of small clause complement.

\subsection{Auxiliary + Null Copula}
\label{la}

In \cite{lapointe80} the copula is treated as an auxiliary verb that takes as its
complement a VP headed by a passive verb, a present participle, or a null verb
(the true copula). This verb may then take AP, NP or PP complements.  The
author points out that there are many languages that have been analyzed as
having a null copula, but that English has the peculiarity that its
null copula requires the co-presence of the auxiliary {\it be}.

\subsection{Auxiliary + Predicative Phrase}
\label{gpsg}

In GPSG (\cite{gazdar85}, \cite{sag85}) the copula is treated as an auxiliary
verb that takes an X$^{2}$ category with a + value for the head feature
[PRD] (predicative). AP, NP, PP and VP can all be [+PRD], but a
Feature Co-occurrence Restriction guarantees that a [+PRD] VP will be
headed by a verb that is either passive or a present participle.

GPSG follows \cite{chomsky70} in adopting the binary valued features [V] and
[N] for decomposing the verb, noun, adjective and preposition categories.  In
that analysis, verbs are [+V,$-$N], nouns are [$-$V,+N], adjectives [+V,+N] and
prepositions [$-$V,$-$N].  NP and AP predicative complements generally pattern
together; a fact that can be stated economically using this category
decomposition.  In neither \cite{sag85} nor \cite{chomsky70} is there any
discussion of how to handle the complete range of complements to a verb like
{\it seem}, which takes AP, NP and PP complements, as well as infinitives.  The
solution would appear to be to associate the verb with two sets of rules for
small clauses, leaving aside the use of the verb with an expletive subject and
sentential complement.

\subsection{Auxiliary + Small Clause}

\label{mo}
In \cite{moro90} the copula is treated as a special functional category - a
lexicalization of tense, which is considered to head its own projection. It
takes as a complement the projection of another functional category, Agr
(agreement). This projection corresponds roughly to a small clause, and is
considered to be the domain within which predication takes place.  An NP must
then raise out of this projection to become the subject of the sentence: it may
be the subject of the AgrP, or, if the predicate of the AgrP is an NP, this may
raise instead.  In addition to occurring as the complement of {\it be}, AgrP is
selected by certain verbs such as {\it consider}. It follows from this analysis
that when the complement to {\it consider} is a simple AgrP, it will always
consist of a subject followed by a predicate, whereas if the complement
contains the verb {\it be}, the predicate of the AgrP may raise to the left of
{\it be}, leaving the subject of the AgrP to the right.

\enumsentence{John$_{i}$ is [$_{AgrP}$ $t_{i}$ the culprit ] .}
\enumsentence{The culprit$_{i}$ is [$_{AgrP}$ John $t_{i}$ ] .}
\enumsentence{I consider [$_{AgrP}$ John the culprit] .}
\enumsentence{I consider [John$_{i}$ to be [$_{AgrP}$ $t_{i}$ the culprit ]] .}
\enumsentence{I consider [the culprit$_{i}$ to be [$_{AgrP}$ John $t_{i}$ ]] .}

Moro does not discuss a number of aspects of his analysis, including the
nature of Agr and the implied existence of sentences without VP's. 

\section{XTAG analysis}
\label{sm-clause-xtag-analysis}

\begin{figure}[htbp]
\centering
\begin{tabular}{ccccc}
{\psfig{figure=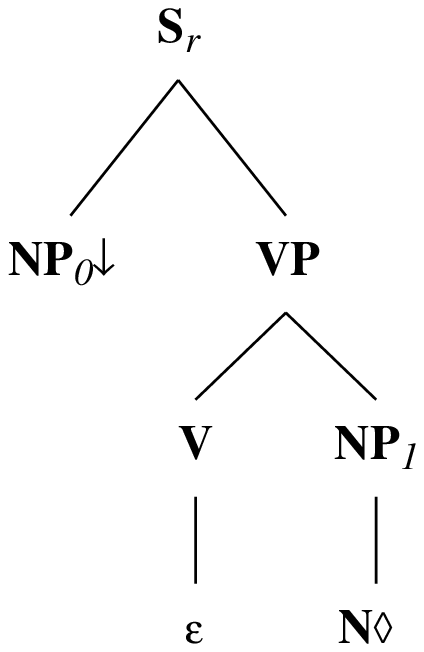,height=2.3in}} &
\hspace{0.5in} &
{\psfig{figure=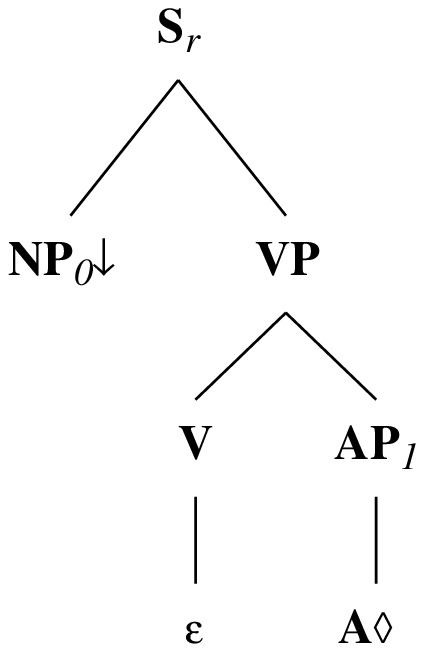,height=2.4in}} &
\hspace{0.5in} &
{\psfig{figure=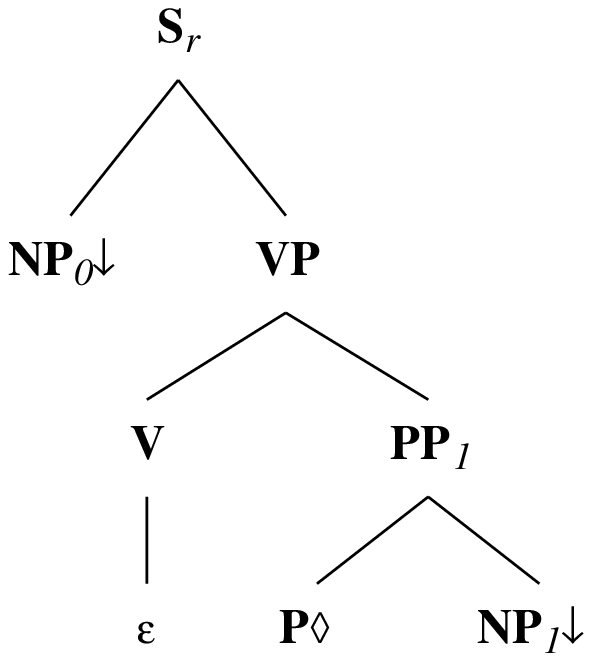,height=2.4in}} \\
(a)&&(b)&&(c)\\
\end{tabular}
\caption{Predicative trees: $\alpha$nx0N1 (a), $\alpha$nx0Ax1 (b) and $\alpha$nx0Pnx1 (c)}
\label{predicative-trees}
\label{1;1,7}
\label{1;1,9}
\end{figure}

The XTAG grammar provides a uniform analysis for the copula, raising verbs and
small clauses by treating the maximal projections of lexical items that can be
predicated as predicative clauses, rather than simply noun, adjective and
prepositional phrases.  The copula adjoins in for matrix clauses, as do the
raising verbs.  Certain other verbs (such as {\it consider}) can take the
predicative clause as a complement, without the adjunction of the copula, to
form the embedded small clause.

The structure of a predicative clause, then, is roughly as seen in
({\ex{1}})-({\ex{3}}) for NP's, AP's and PP's.  The XTAG trees corresponding
to these structures\footnote{There are actually two other predicative trees in
the XTAG grammar.  Another predicative noun phrase tree is needed for noun
phrases without determiners, as in the sentence {\it They are firemen}, and
another prepositional phrase tree is needed for exhaustive prepositional
phrases, such as {\it The workers are below}.} are shown in
Figures~\ref{predicative-trees}(a),
\ref{predicative-trees}(b), and \ref{predicative-trees}(c), 
respectively.

\enumsentence{[$_{S}$ NP [$_{VP}$  N \ldots ]]}
\enumsentence{[$_{S}$ NP [$_{VP}$  A \ldots ]]}
\enumsentence{[$_{S}$ NP [$_{VP}$  P \ldots ]]}

The copula {\it be} and raising verbs all get the basic auxiliary tree as
explained in the section on auxiliary verbs (section \ref{aux-non-inverted}).
Unlike the raising verbs, the copula also selects the inverted auxiliary tree
set.  Figure~\ref{Vvx-with-nomprep} shows the basic auxiliary tree anchored by
the copula {\it be}.  The {\bf $<$mode$>$} feature is used to distinguish the
predicative constructions so that only the copula and raising verbs adjoin onto
the predicative trees.  

\begin{figure}[htb]
\centering
\begin{tabular}{c}
{\psfig{figure=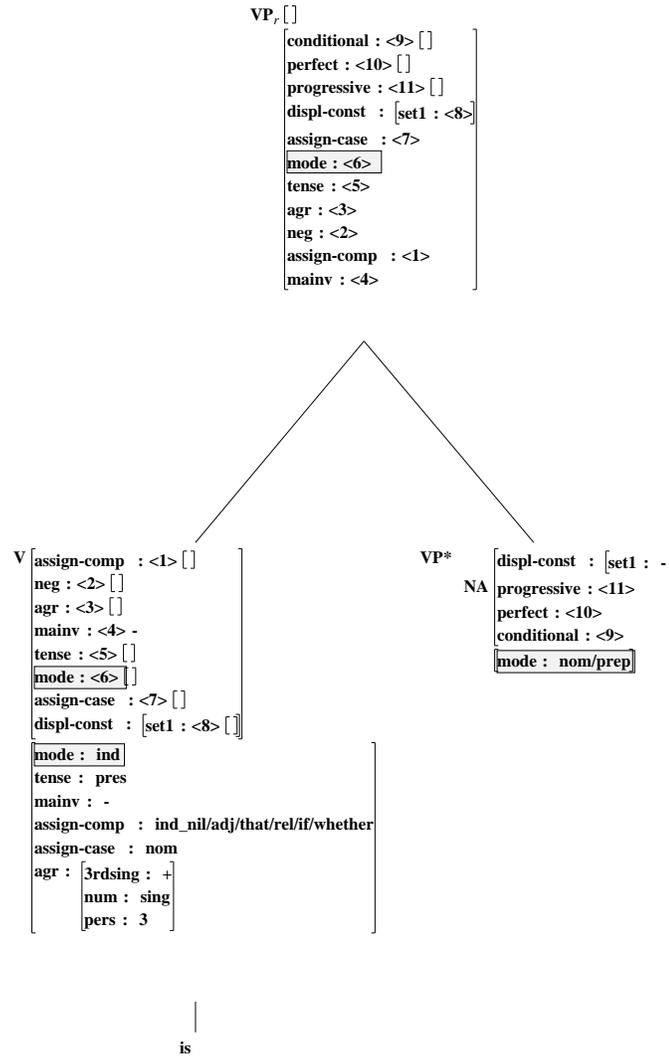,height=5.7in}} \\
\end{tabular}
\caption{Copula auxiliary tree: $\beta$Vvx}
\label{Vvx-with-nomprep}
\end{figure}

There are two possible values of {\bf $<$mode$>$} that correspond to the
predicative trees, {\bf nom} and {\bf prep}.  They correspond to a modified
version of the four-valued [N,V] feature described in section \ref{gpsg}.  The
{\bf nom} value corresponds to [N+], selecting the NP and AP predicative
clauses.  As mentioned earlier, they often pattern together with respect to
constructions using predicative clauses.  The remaining prepositional phrase
predicative clauses, then, correspond to the {\bf prep} mode.

Figure~\ref{upset-with-features} shows the predicative adjective tree from
Figure~\ref{predicative-trees}(b) now anchored by {\it upset} and with the
features visible.  As mentioned, {\bf $<$mode$>$=nom} on the VP node prevents
auxiliaries other than the copula or raising verbs from adjoining into this
tree.  In addition, it prevents the predicative tree from occurring as a matrix
clause.  Since all matrix clauses in XTAG must be mode indicative ({\bf ind})
or imperative ({\bf imp}), a tree with {\bf $<$mode$>$=nom} or {\bf
$<$mode$>$=prep} must have an auxiliary verb (the copula or a raising verb)
adjoin in to make it {\bf $<$mode$>$=ind}.

\begin{figure}[htb]
\centering
\begin{tabular}{c}
{\psfig{figure=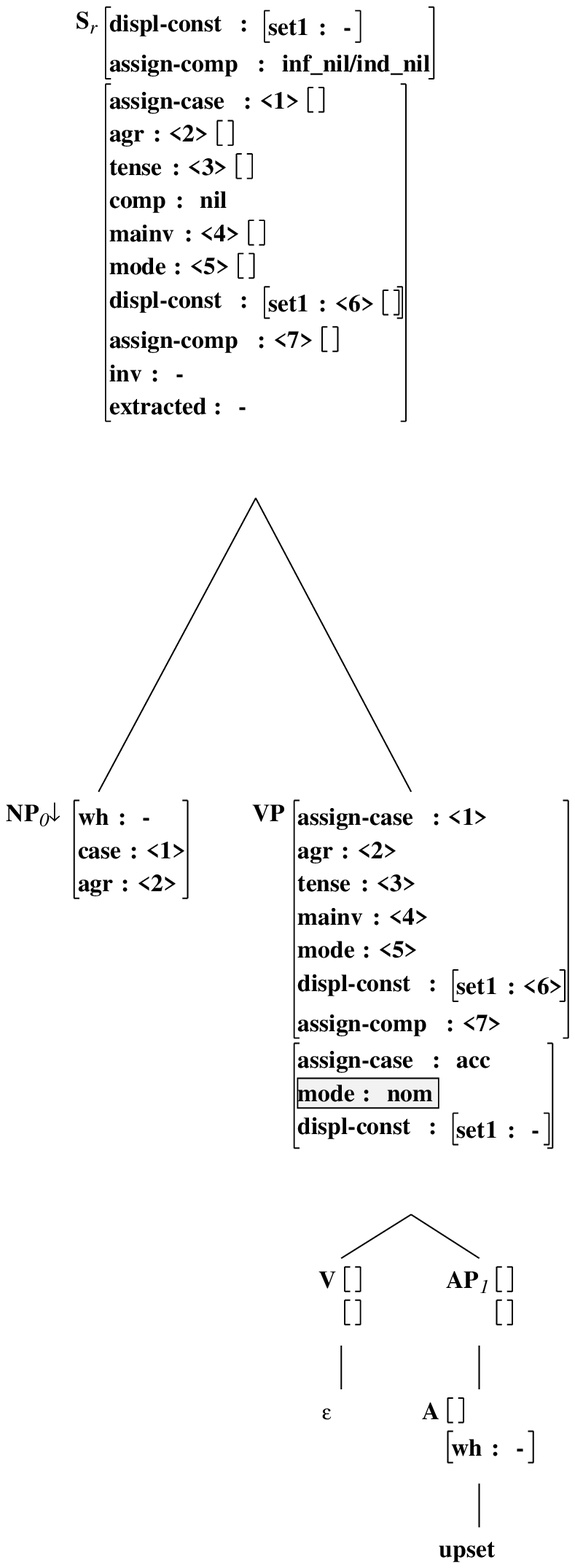,height=6.3in}} \\
\end{tabular}
\caption{Predicative AP tree with features: $\alpha$nx0Ax1}
\label{upset-with-features}
\label{1;1,4}
\end{figure}

The distribution of small clauses as embedded complements to some verbs is also
managed through the mode feature.  Verbs such as {\it consider} and {\it
prefer} select trees that take a sentential complement, and then restrict that
complement to be {\bf $<$mode$>$=nom} and/or {\bf $<$mode$>$=prep},
depending on the lexical idiosyncrasies of that particular verb.  Many verbs
that don't take small clause complements do take sentential complements that
are {\bf $<$mode$>$=ind}, which includes small clauses with the copula
already adjoined.  Hence, as seen in sentence sets ({\ex{1}})-({\ex{3}}),
{\it consider} takes only small clause complements, {\it prefer} takes both
{\bf prep} (but not {\bf nom}) small clauses and indicative clauses, while {\it
feel} takes only indicative clauses.

\enumsentence{She considers Carl a jerk .\\
		?She considers Carl in a foul mood .\\
		$\ast$She considers that Carl is a jerk .}

\enumsentence{$\ast$She prefers Carl a jerk .\\
		She prefers Carl in a foul mood .\\
		She prefers that Carl is a jerk .}

\enumsentence{$\ast$She feels Carl a jerk .\\
		$\ast$She feels Carl in a foul mood .\\
		She feels that Carl is a jerk .}

\noindent
Figure \ref{consider-with-features} shows the tree anchored by {\it consider}
that takes the predicative small clauses.

\begin{figure}[htb]
\centering
\begin{tabular}{c}
{\psfig{figure=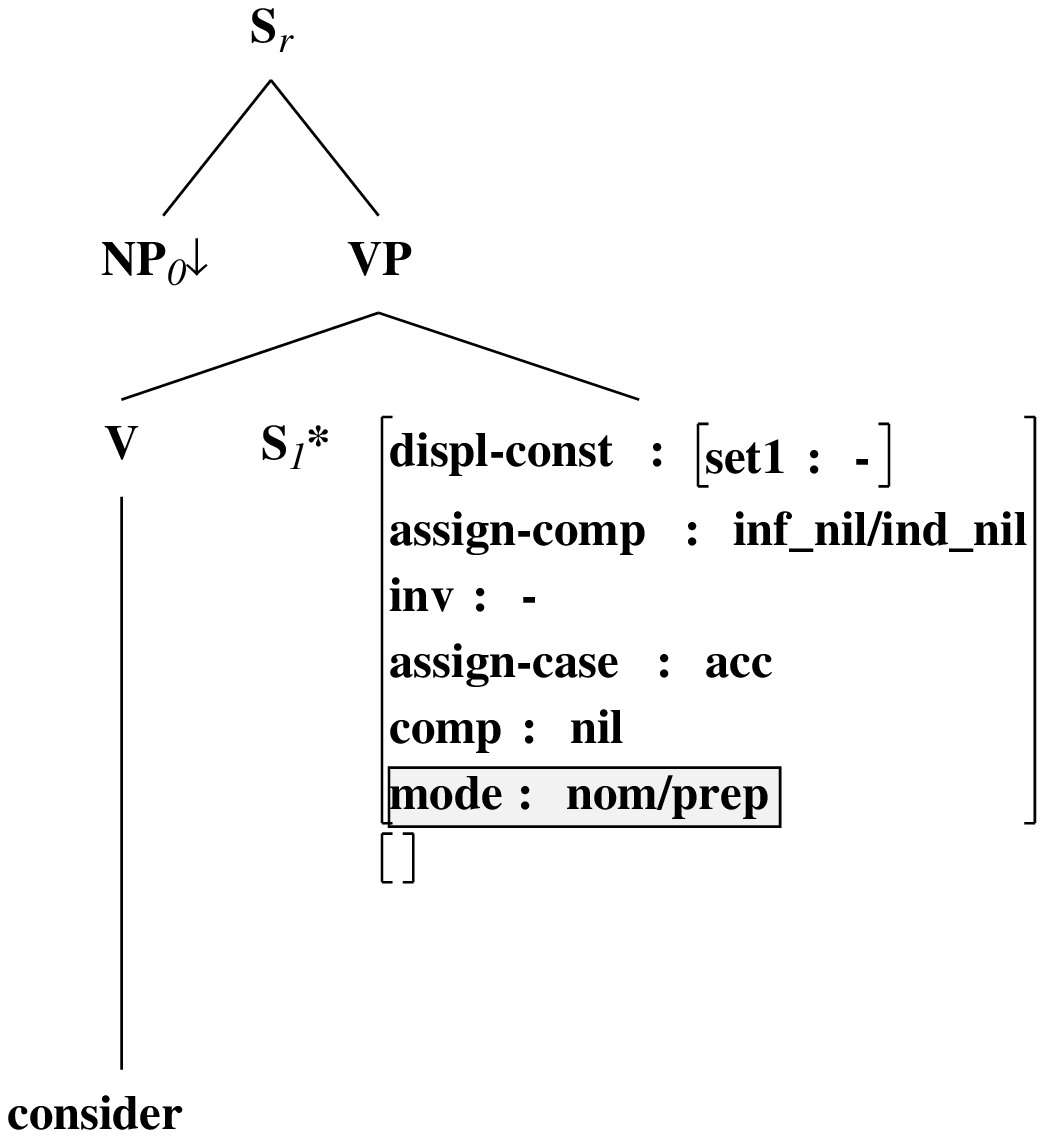,height=2.3in}} \\
\end{tabular}
\caption{{\it Consider} tree for embedded small clauses}
\label{consider-with-features}
\end{figure}

Raising verbs such as {\it seems} work essentially the same as the
auxiliaries, in that they also select the basic auxiliary tree, as in
Figure~\ref{Vvx-with-nomprep}.  The only difference is that 
the value of {\bf $<$mode$>$} 
on the VP foot node might be different, depending on what types of
complements the raising verb takes.  Also, two of the raising verbs take
an additional tree, $\beta$Vpxvx, shown in Figure~\ref{Vpxvx}, which
allows for an experiencer argument, as in {\it John seems to me
to be happy}.  

\begin{figure}[htb]
\centering
\begin{tabular}{c}
{\psfig{figure=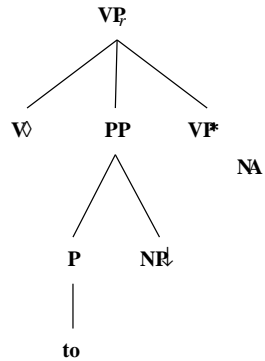,height=2.0in}} \\
\end{tabular}
\caption{Raising verb with experiencer tree: $\beta$Vpxvx}
\label{Vpxvx}
\end{figure}

Raising adjectives, such as {\it likely}, take the tree shown in
Figure~\ref{Vvx-adj}.  This tree combines aspects of the auxiliary
tree $\beta$Vvx and the adjectival predicative tree shown in
Figure~\ref{predicative-trees}(b).  As with $\beta$Vvx, it adjoins
in as a VP auxiliary tree.  However, since it is anchored by an
adjective, not a verb, it is similar to the adjectival predicative
tree in that it has an $\epsilon$ at the V node, and a feature value
of {\bf $<$mode$>$=nom} which is passed up to the VP root indicates
that it is an adjectival predication.  This serves the same purpose
as in the 
case of the tree in Figure~\ref{upset-with-features}, and forces another
auxiliary verb, such as the copula, to adjoin in to make it
{\bf $<$mode$>$=ind}.

\begin{figure}[htb]
\centering
\begin{tabular}{c}
{\psfig{figure=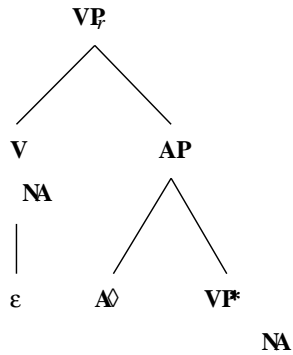,height=2.0in}} \\
\end{tabular}
\caption{Raising adjective tree: $\beta$Vvx-adj}
\label{Vvx-adj}
\end{figure}

\section{Non-predicative {\it BE}}
\label{equative-be-xtag-analysis}

The examples with the copula that we have given seem to indicate that {\it be}
is always followed by a predicative phrase of some sort.  This is not the case,
however, as seen in sentences such as ({\ex{1}})-({\ex{6}}).  The noun phrases in
these sentences are not predicative.  They do not take raising verbs, and they
do not occur in embedded small clause constructions.

\enumsentence{my teacher is Mrs. Wayman .}
\enumsentence{Doug is the man with the glasses .}

\enumsentence{$\ast$My teacher seems Mrs. Wayman .}
\enumsentence{$\ast$Doug appears the man with the glasses .}

\enumsentence{$\ast$I consider [my teacher Mrs. Wayman] .}
\enumsentence{$\ast$I prefer [Doug the man with the glasses] .}

In addition, the subject and complement can exchange positions in these type of
examples but not in sentences with predicative {\it be}.  Sentence ({\ex{1}})
has the same interpretation as sentence ({\ex{-4}}) and differs only in the
positions of the subject and complement NP's. Similar sentences, with a
predicative {\it be}, are shown in ({\ex{2}}) and ({\ex{3}}).  In this case,
the sentence with the exchanged NP's ({\ex{3}}) is ungrammatical.

\enumsentence{The man with the glasses is Doug .}
\enumsentence{Doug is a programmer .}
\enumsentence{$\ast$A programmer is Doug .}

The non-predicative {\it be} in ({\ex{-8}}) and ({\ex{-7}}), also called
\xtagdef{equative be}, patterns differently, both syntactically and
semantically, from the predicative usage of {\it be}.  Since these sentences
are clearly not predicative, it is not desirable to have a tree structure that
is anchored by the NP, AP, or PP, as we have in the predicative sentences.  In
addition to the conceptual problem, we would also need a mechanism to block
raising verbs from adjoining into these sentences (while allowing them for true
predicative phrases), and prevent these types of sentence from being embedded
(again, while allowing them for true predicative phrases).

\begin{figure}[htb]
\centering
\begin{tabular}{ccc}
{\psfig{figure=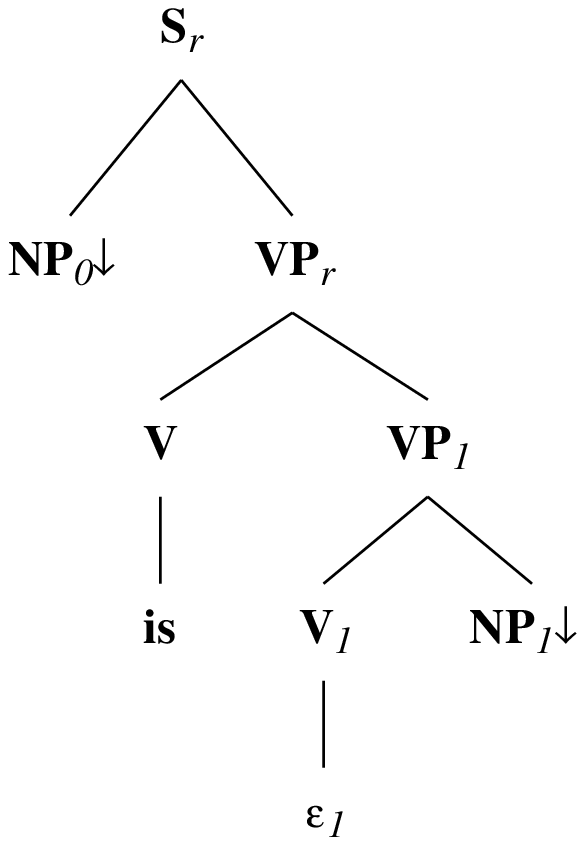,height=1.9in}} &
\hspace{1.0in}&
{\psfig{figure=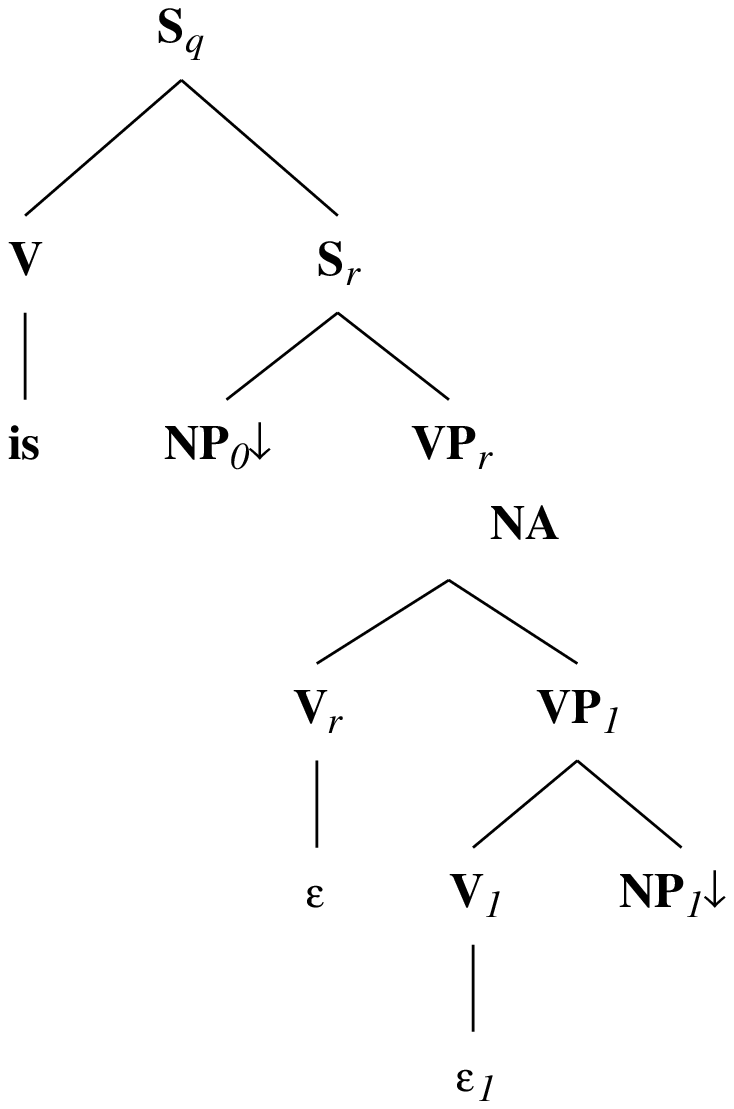,height=2.5in}} \\
(a)&&(b)\\
\end{tabular}
\caption{Equative {\it BE} trees: $\alpha$nx0BEnx1 (a) and $\alpha$Invnx0BEnx1 (b)}
\label{equative-be}
\label{1;1,6}
\end{figure}

Although non-predicative {\it be} is not a raising verb, it does exhibit the
auxiliary verb behavior set out in section \ref{copula-data}.  It inverts,
contracts, and so forth, as seen in sentences ({\ex{1}}) and ({\ex{2}}), and
therefore can not be associated with any existing tree family for main verbs.
It requires a separate tree family that includes the tree for inversion.
Figures~\ref{equative-be}(a) and \ref{equative-be}(b) show the declarative and
inverted trees, respectively, for equative {\it be}.

\enumsentence{is my teacher Mrs. Wayman ?}
\enumsentence{Doug isn't the man with the glasses .}

\chapter{Ditransitive constructions and dative shift}
\label{double-objs}

Verbs such as {\it give\/} and {\it put\/} that require two objects, as
shown in examples (\ex{1})-(\ex{4}), are termed ditransitive.

\enumsentence{Christy gave a cannoli to Beth Ann .}
\enumsentence{$\ast$Christy gave Beth Ann .}
\enumsentence{Christy put a cannoli in the refrigerator .} 
\enumsentence{$\ast$Christy put a cannoli .}

The indirect objects {\it Beth Ann\/} and {\it refrigerator\/} appear in
these examples in the form of PP's.  Within the set of ditransitive
verbs there is a subset that also allow two NP's as in (\ex{1}). As can
be seen from (\ex{1}) and (\ex{2}) this two NP, or double-object,
construction is grammatical for {\it give\/} but not for {\it put}.  

\enumsentence{Christy gave Beth Ann a cannoli .}
\enumsentence{$\ast$Christy put the refrigerator the cannoli .}

The alternation between (\ex{-5}) and (\ex{-1}) is known as dative
shift.\footnote{In languages similar to English that have overt case marking
indirect objects would be marked with dative case. It has also been suggested
that for English the preposition {\it to} serves as a dative case marker.} In
order to account for verbs with dative shift the English XTAG grammar includes
structures for both variants in the tree family Tnx0Vnx1Pnx2.  The declarative
trees for the shifted and non-shifted alternations are shown in
Figure~\ref{dative-alt}.

\begin{figure}[htb]
\centering
\begin{tabular}{ccc}
{\psfig{figure=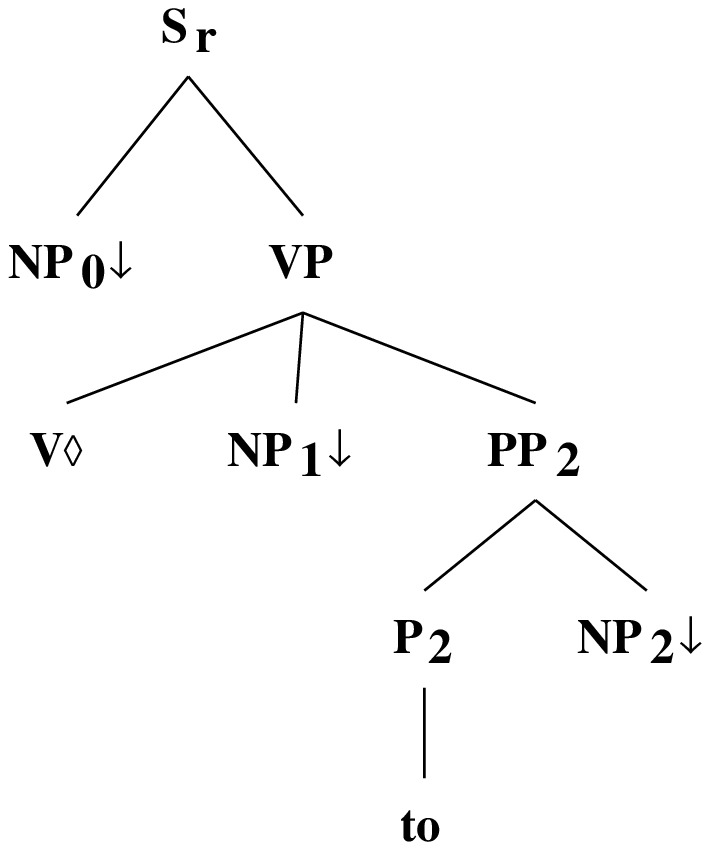,height=2.0in}}&
\hspace*{0.5in} &
{\psfig{figure=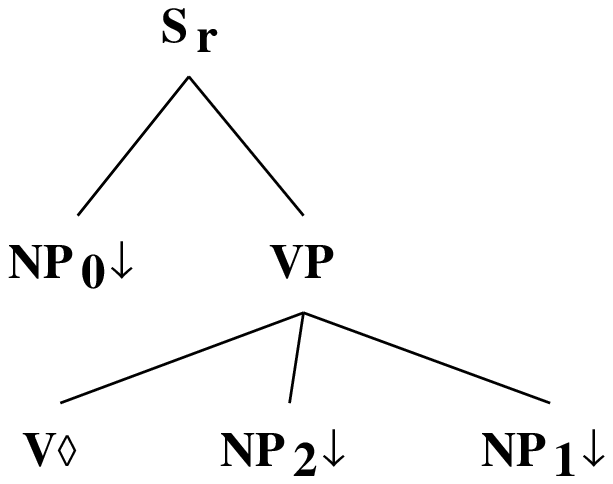,height=1.1in}}
\\
(a)&\hspace*{0.5in}&(b)\\
\end{tabular}
\caption{Dative shift trees: $\alpha$nx0Vnx1Pnx2 (a) and $\alpha$nx0Vnx2nx1 (b)}
\label{dative-alt}
\label{2;1,2}
\end{figure}

The indexing of nodes in these two trees represents the fact that the semantic
role of the indirect object (NP$_2$) in Figure~\ref{dative-alt}(a) is the same
as that of the direct object (NP$_2$) in Figure~\ref{dative-alt}(b) (and vice
versa).  This use of indexing is consistent with our treatment of other
constructions such as passive and ergative.

Verbs that do not show this alternation and have only the NP PP structure
(e.g. {\it put\/}) select the tree family Tnx0Vnx1pnx2.  Unlike the
Tnx0Vnx1Pnx2 family, the Tnx0Vnx1pnx2 tree family does not contain trees for
the NP NP structure. Other verbs such as {\it ask} allow only the NP NP
structure as shown in (\ex{1}) and (\ex{2}).

\enumsentence{Beth Ann asked Srini a question .}
\enumsentence{$\ast$Beth Ann asked a question to Srini .}

Verbs that only allow the NP NP structure select the tree family
Tnx0Vnx1nx2. This tree family does not have the trees for the NP PP
structure. 

Notice that in Figure~\ref{dative-alt}(a) the preposition {\it to\/} is
built into the tree.  There are other apparent cases of dative shift
with {\it for}, such as in (\ex{1}) and (\ex{2}), that we have taken to
be structurally distinct from the cases with {\it to}.  

\enumsentence{Beth Ann baked Dusty a biscuit .}
\enumsentence{Beth Ann baked a biscuit for Dusty .}

\cite{mccawley88} notes:

\begin{quote}
A ``{\it for-dative}'' expression in underlying structure is external
to the V with which it is combined, in view of the fact that the
latter behaves as a unit with regard to all relevant syntactic
phenomena.
\end{quote}

In other words, the {\it for} PP's that appear to undergo dative shift are
actually adjuncts, not complements. Examples (\ex{1}) and (\ex{2}) demonstrate
that PP's with {\it for} are optional while ditransitive {\it to} PP's are not.

\enumsentence{Beth Ann made dinner .}
\enumsentence{$\ast$Beth Ann gave dinner .}

Consequently, in the XTAG grammar the apparent dative shift with {\it
  for} PP's is treated as Tnx0Vnx1nx2 for the NP NP structure, and as
a transitive plus an adjoined adjunct PP for the NP PP structure.  To
account for the ditransitive {\it to} PP's, the preposition {\it to}
is built into the tree family Tnx0Vnx1tonx2. This accounts for the
fact that {\it to} is the only preposition allowed in dative shift
constructions.

\cite{mccawley88} also notes that the {\it to} and {\it for} cases
differ with respect to passivization; the indirect objects with {\it
  to} may be the subjects of corresponding passives while the alleged
indirect objects with {\it for} cannot, as in
sentences~(\ex{1})-(\ex{4}). Note that the passivisation examples are
for NP NP structures of verbs that take {\it to} or {\it for} PP's.

\enumsentence{Beth Ann gave Clove dinner .}
\enumsentence{Clove was given dinner (by Beth Ann) .}
\enumsentence{Beth Ann made Clove dinner .}
\enumsentence{?Clove was made dinner (by Beth Ann) .} 

However, we believe that this to be incorrect, and that the indirect objects in
the {\it for} case are allowed to be the subjects of passives, as in
sentences~(\ex{1})-(\ex{2}).  The apparent strangeness of sentence~(\ex{0}) is
caused by interference from other interpretations of {\it Clove was made
dinner .}

\enumsentence{Dania baked Doug a cake .}
\enumsentence{Doug was baked a cake by Dania .}

\chapter{It-clefts}
\label{it-clefts}

There are several varieties of it-clefts in English.  All the
it-clefts have four major components:

\begin{itemize}
\item {\bf the dummy subject:}  {\it it},
\item {\bf the main verb:}  {\it be},
\item {\bf the clefted element:}  A constituent (XP) compatible with
any gap in the clause,
\item {\bf the clause:}  A clause (e.g. S) with or without a gap.
\end{itemize}

\noindent
Examples of it-clefts are shown in (\ex{1})-(\ex{4}).

\enumsentence{it was [$_{XP}$ here $_{XP}$] [$_{S}$ that the ENIAC was
created . $_{S}$]}
\enumsentence{it was [$_{XP}$ at MIT $_{XP}$]  [$_{S}$ that colorless green
ideas slept furiously . $_{S}$]}
\enumsentence{it is [$_{XP}$ happily $_{XP}$]  [$_{S}$ that Seth quit Reality . $_{S}$]}
\enumsentence{it was  [$_{XP}$ there $_{XP}$]  [$_{S}$ that she would
have to enact her renunciation . $_{S}$]}

The clefted element can be of a number of categories, for example NP, PP or
adverb. The clause can also be of several types. The English XTAG grammar
currently has a separate analysis for only a subset of the `specificational'
it-clefts\footnote{See e.g. \cite{Ball91},
\cite{Delin89} and \cite{Delahunty84} for more detailed discussion of
types of it-clefts.}, in particular the ones without gaps in the clause
(e.g. (\ex{-1}) and (\ex{-0})).  It-clefts that have gaps in the clause, such
as (\ex{-3}) and (\ex{-2}) are currently handled as relative clauses. Although
arguments have been made against treating the clefted element and the clause as
a constituent (\cite{Delahunty84}), the relative clause approach does capture
the restriction that the clefted element must fill the gap in the clause, and
does not require any additional trees.

In the `specificational' it-cleft without gaps in the clause, the
clefted element has the role of an adjunct with respect to the clause.
For these cases the English XTAG grammar requires additional trees.
These it-cleft trees are in separate tree families because, although
some researchers (e.g. \cite{Akmajian70}) derived it-clefts through
movement from other sentence types, most current researchers
(e.g. \cite{Delahunty84}, \cite{Knowles86}, \cite{gazdar85},
\cite{Delin89} and \cite{Sornicola88}) favor base-generation of the
various cleft sentences.  Placing the it-cleft trees in their own tree
families is consistent with the current preference for base
generation, since in the XTAG English grammar, structures that would
be related by transformation in a movement-based account will appear
in the same tree family. Like the base-generated approaches, the
placement of it-clefts in separate tree families makes the claim that
there is no derivational relation between it-clefts and other sentence
types.

The three it-cleft tree families are virtually identical except for the
category label of the clefted element.  Figure~\ref{pp-it-clefts} shows the
declarative tree and an inverted tree for the PP It-cleft tree family.

\begin{figure}[htb]
\centering
\begin{tabular}{ccc}
{\psfig{figure=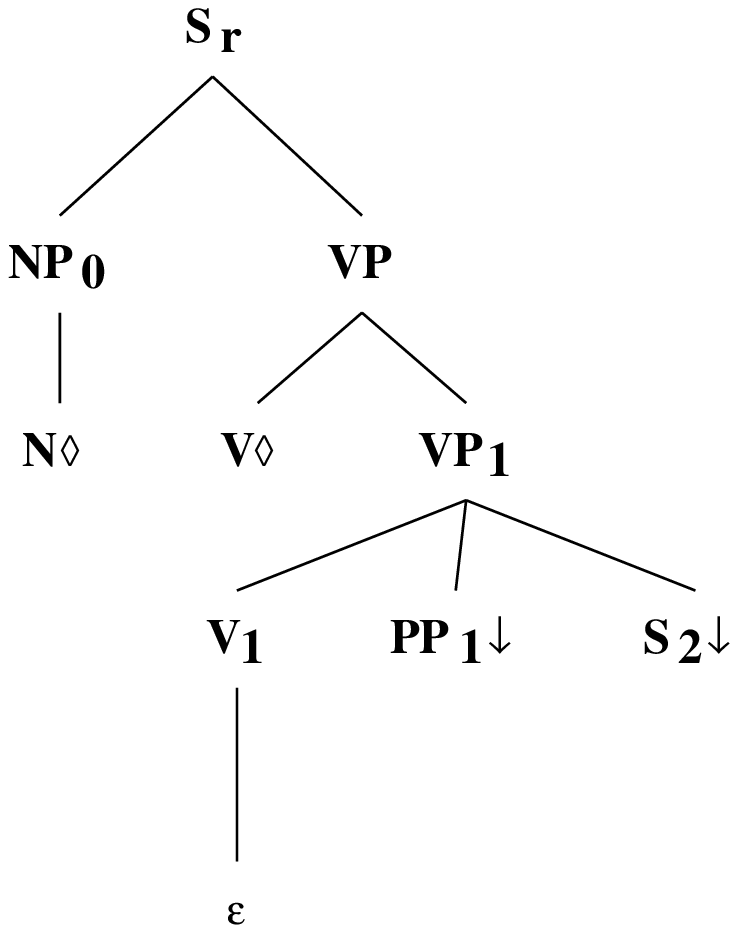,height=2.0in}} &
\hspace*{0.5in} &
{\psfig{figure=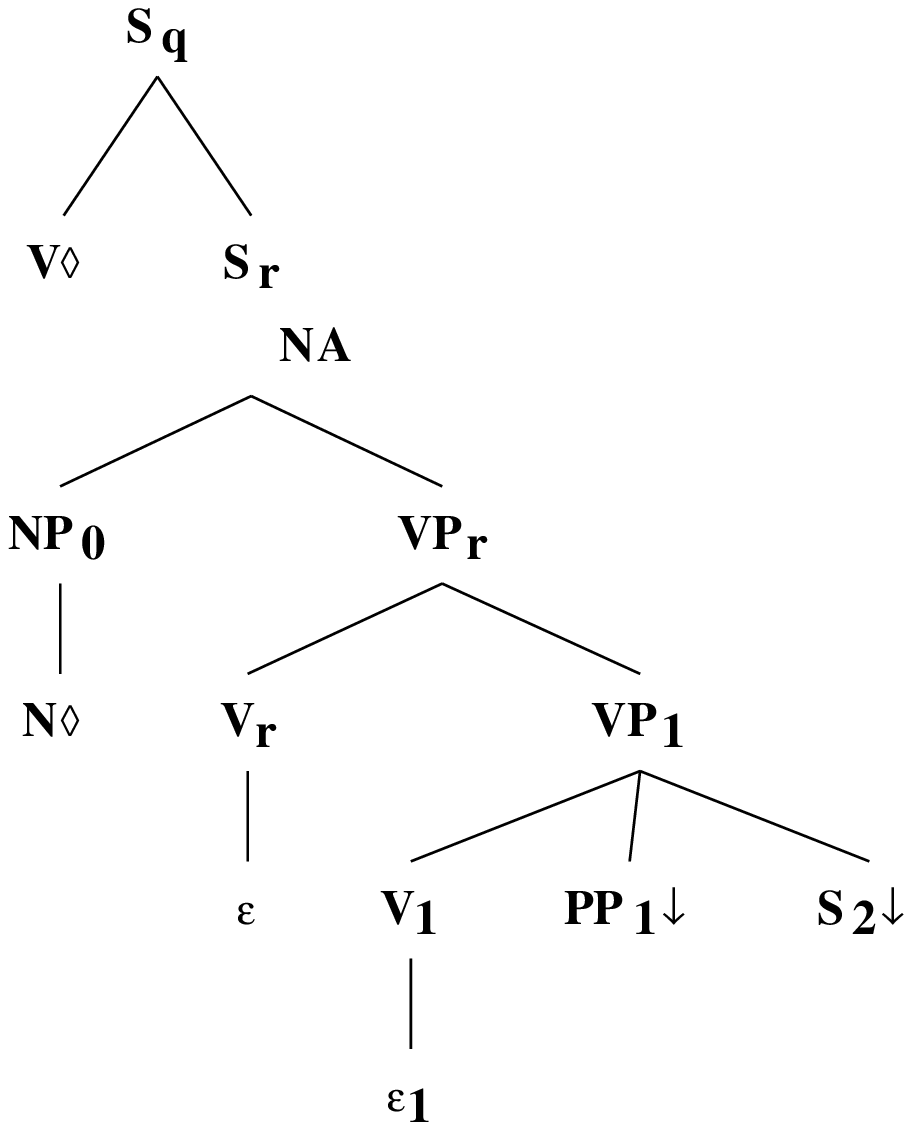,height=2.5in}} \\
(a)&\hspace*{0.5in}&(b)\\
\end{tabular}
\caption{It-cleft with PP clefted element: $\alpha$ItVpnx1s2 (a) and
$\alpha$InvItVpnx1s2 (b)}
\label{pp-it-clefts}
\label{1;1,3}
\label{1;3,3}
\end{figure}

The extra layer of tree structure in the VP represents that, while {\it be} is
a main verb rather than an auxiliary in these cases, it retains some auxiliary
properties. The VP structure for the equative/it-cleft-{\it be} is identical to
that obtained after adjunction of predicative-{\it be} into
small-clauses.\footnote{For additional discussion of equative or
predicative-{\it be} see Chapter~\ref{small-clauses}.}  The inverted tree in
Figure~\ref{pp-it-clefts}(b) is necessary because of {\it be}'s auxiliary-like
behavior.  Although {\it be} is the main verb in it-clefts, it inverts like an
auxiliary.  Main verb inversion cannot be accomplished by adjunction as is done
with auxiliaries and therefore must be built into the tree family. The tree in
Figure~\ref{pp-it-clefts}(b) is used for yes/no questions such as (\ex{1}).

\enumsentence{was it in the forest that the wolf talked to the little girl ?}

\part{Sentence Types}
\chapter{Passives}
\label{passives}
In passive constructions such as (\ex{1}), the subject NP is
interpreted as having the same role as the direct object NP in the
corresponding active declarative (\ex{2}).

\enumsentence{{\bf An airline buy-out bill} was approved by the House. (WSJ)}
\enumsentence{The House approved {\bf an airline buy-out bill}.}

\begin{figure}[hbt]
\centering
\begin{tabular}{ccccc}
\psfig{figure=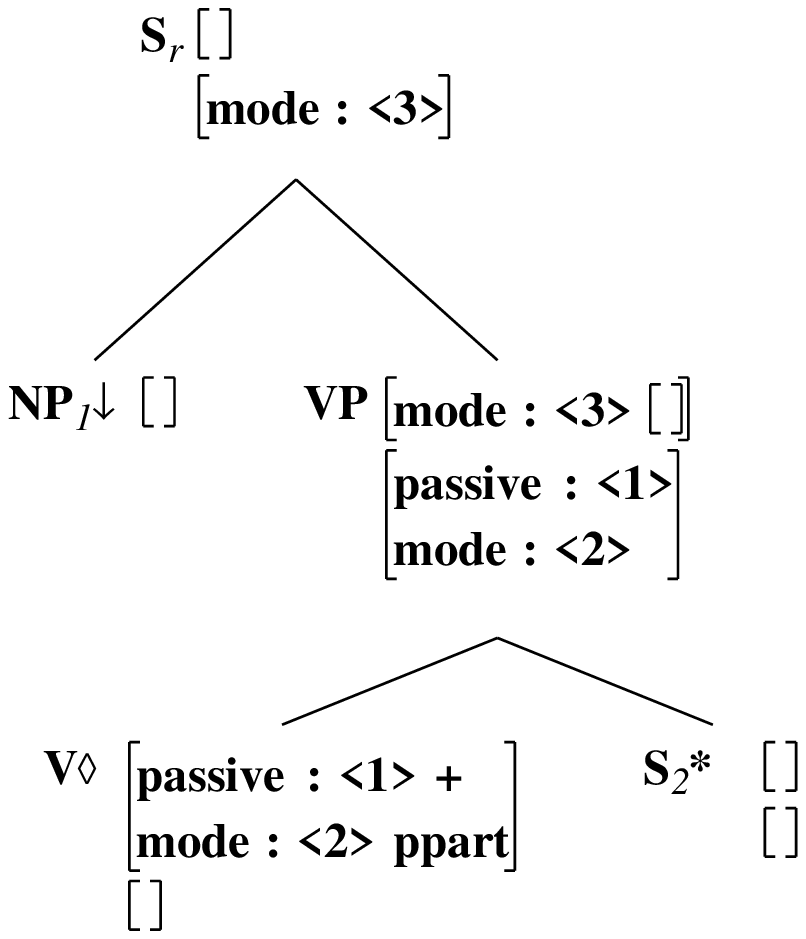,height=6.5cm}&
\hspace{1.0in}&
\psfig{figure=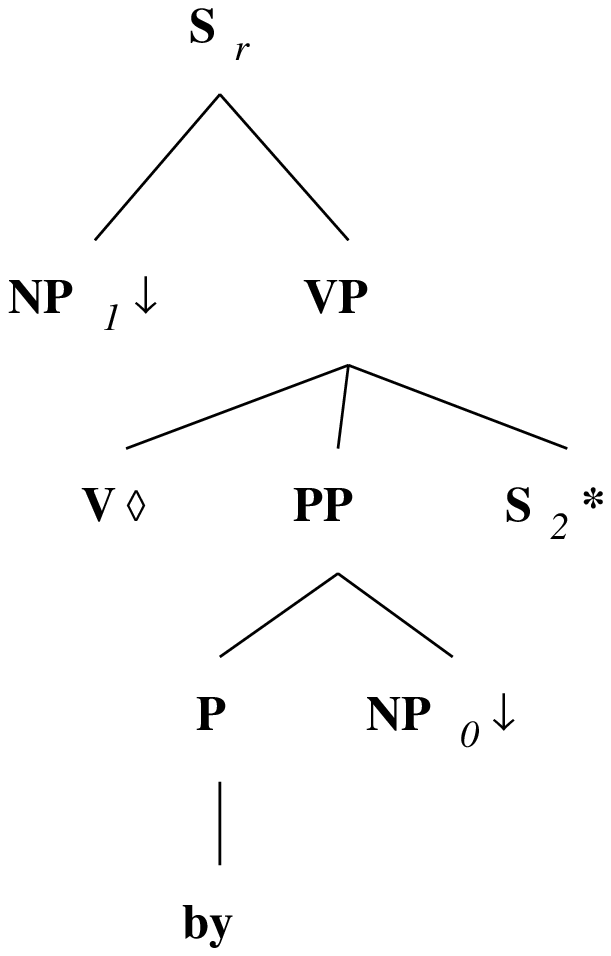,height=6.5cm}&
\hspace{1.0in}&
\psfig{figure=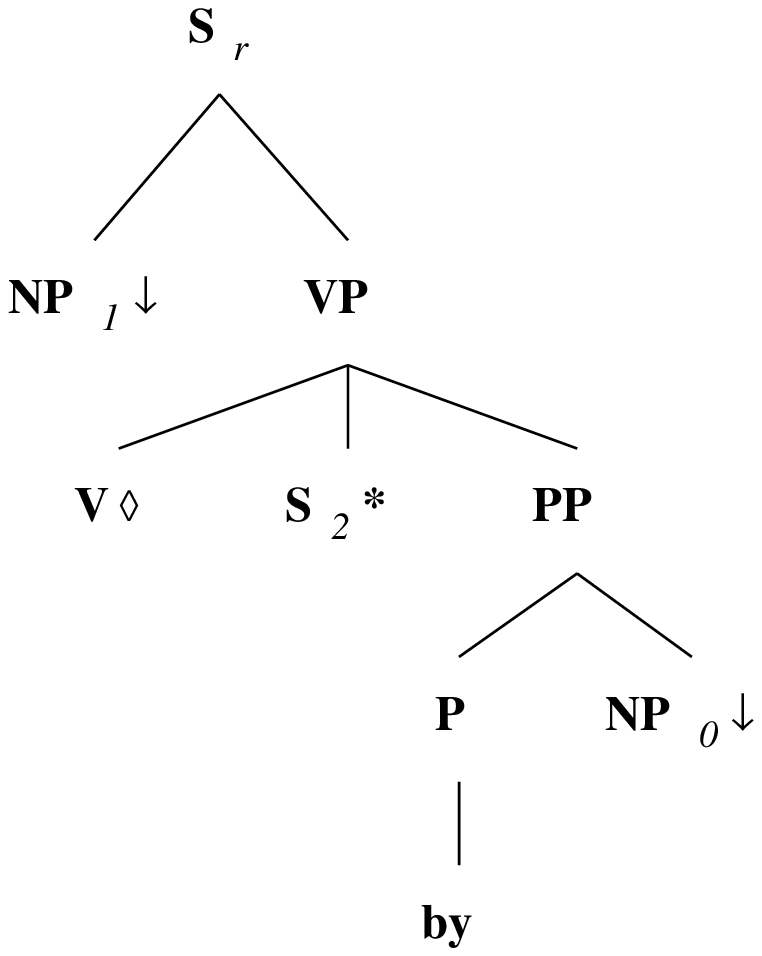,height=6.5cm}\\
(a)&&(b)&&(c)
\end{tabular}
\caption{Passive trees in the Sentential Complement with NP tree family:
$\beta$nx1Vs2 (a), $\beta$nx1Vbynx0s2 (b) and $\beta$nx1Vs2bynx0 (c)}
\label{passive-trees}
\label{2;2,5}
\end{figure}

In a movement analysis, the direct object is said to have moved to the subject
position.  The original declarative subject is either absent in the passive or
is in a {\it by} headed PP ({\it by} phrase). In the English XTAG grammar,
passive constructions are handled by having separate trees within the
appropriate tree families.  Passive trees are found in most tree families that
have a direct object in the declarative tree (the light verb tree families, for
instance, do not contain passive trees).  Passive trees occur in pairs - one
tree with the {\it by} phrase, and another without it.  Variations in the
location of the {\it by} phrase are possible if a subcategorization includes
other arguments such as a PP or an indirect object. Additional trees are
required for these variations.  For example, the Sentential Complement with NP
tree family has three passive trees, shown in Figure~\ref{passive-trees}: one
without the {\it by}-phrase (Figure~\ref{passive-trees}(a)), one with the {\it
by} phrase before the sentential complement (Figure~\ref{passive-trees}(b)),
and one with the {\it by} phrase after the sentential complement
(Figure~\ref{passive-trees}(c)).

Figure~\ref{passive-trees}(a) also shows the feature restrictions imposed on
the anchor\footnote{A reduced set of features are shown for readability.}. Only
verbs with {\bf $<$mode$>$=ppart} (i.e. verbs with passive morphology) can
anchor this tree.  The {\bf $<$mode$>$} feature is also responsible for
requiring that passive {\it be} adjoin into the tree to create a matrix
sentence.  Since a requirement is imposed that all matrix sentences must have
{\bf $<$mode$>$=ind/imp}, an auxiliary verb that selects {\bf
$<$mode$>$=ppart} and {\bf $<$passive$>$=+} (such as {\it was}) must adjoin
(see Chapter~\ref{auxiliaries} for more information on the auxiliary verb
system).

\chapter{Extraction}
\label{extraction}

The discussion in this chapter covers constructions that are analyzed
as having wh-movement in GB, in particular, wh-questions and
topicalization. Relative clauses, which could also be considered
extractions, are discussed in Chapter~\ref{rel_clauses}.

Extraction involves a constituent appearing in a linear position to the left of
the clause with which it is interpreted. One clause argument position is
empty. For example, the position filled by {\it frisbee} in the declarative in
sentence~(\ex{1}) is empty in sentence~(\ex{2}). The wh-item {\it what} in
sentence~(\ex{2}) is of the same syntactic category as {\it frisbee} in
sentence~(\ex{1}) and fills the same role with respect to the
subcategorization.

\enumsentence{Clove caught a frisbee.}
\enumsentence{What$_{i}$ did Clove catch $\epsilon_{i}$?}

The English XTAG grammar represents the connection between the extracted
element and the empty position with co-indexing (as does GB).  The {\bf
$<$trace$>$} feature is used to implement the co-indexing.  In extraction trees
in XTAG, the `empty' position is filled with an {\it $\epsilon$}.  The
extracted item always appears in these trees as a sister to the S$_{r}$
tree, with both dominated by a S$_{q}$ root node.  The S$_{r}$ subtrees in
extraction trees have the same structure as the declarative tree in the same
tree family.  The additional structure in extraction trees of the S$_{q}$ and
NP nodes roughly corresponds to the CP and Spec of CP positions in GB.

All sentential trees with extracted components (this does not include relative
clause trees) are marked {\bf $<$extracted$>$=+} at the top S node, while
sentential trees with no extracted components are marked {\bf
$<$extracted$>$=--}.  Items that take embedded sentences, such as nouns, verbs
and some prepositions can place restrictions on whether the embedded sentence
is allowed to be extracted or not.  For instance, sentential subjects and
sentential complements of nouns and prepositions are not allowed to be
extracted, while certain verbs may allow extracted sentential complements and
others may not (e.g. sentences (\ex{1})-(\ex{4})).

\enumsentence{The jury wondered [who killed Nicole].}
\enumsentence{The jury wondered [who Simpson killed].}
\enumsentence{The jury thought [Simpson killed Nicole].}
\enumsentence{$\ast$The jury thought [who did Simpson kill]?}
The {\bf $<$extracted$>$} feature is also used to block embedded topicalization
in infinitival complement clauses as exemplified in (\ex{1}). 
\enumsentence{* John wants [ Bill$_{i}$ [PRO to see t$_{i}$]]}
Verbs such as {\em want} that take non-{\em wh} infinitival complements
specify that the {\bf $<$extracted$>$} feature of their complement clause
(i.e. of the foot S node)
is {\bf --}. Clauses that involve topicalization have {\bf +} as the value
of their {\bf $<$extracted$>$} feature (i.e. of the root S node). 
Sentences like (\ex{0}) are thus ruled out.  

\begin{figure}[htb]
\centering
\mbox{}
\psfig{figure=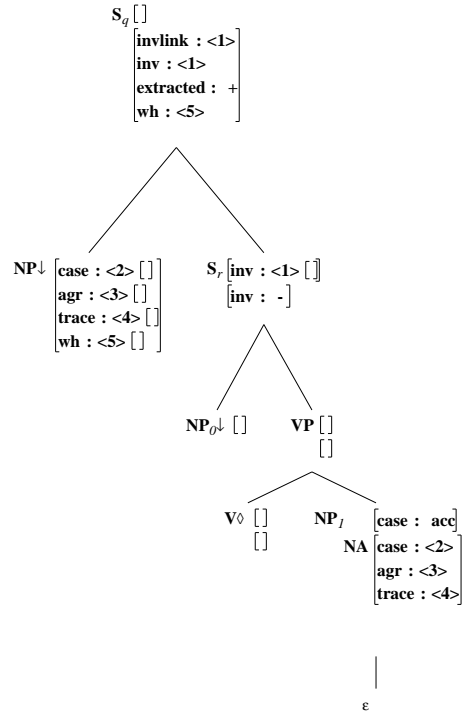,height=10.0cm}
\caption{Transitive tree with object extraction: $\alpha$W1nx0Vnx1}
\label{alphaW1nx0Vnx1}
\label{2;5,1}
\end{figure}

The tree that is used to derive the embedded sentence in (\ex{-2}) in
the English XTAG grammar is shown in
Figure~\ref{alphaW1nx0Vnx1}\footnote{Features not pertaining to this
  discussion have been taken out to improve readability.}.  The
important features of extracted trees are:

\begin{itemize}
\item The subtree that has S$_{r}$ as its root is identical to the
  declarative tree or a non-extracted passive tree, except for having
  one NP position in the VP filled by $\epsilon$.

\item The root S node is S$_{q}$, which dominates NP and S$_{r}$.
  
\item The {\bf $<$trace$>$} feature of the $\epsilon$ filled NP is
  co-indexed with the {\bf $<$trace$>$} feature of the NP daughter of
  S$_{q}$.

\item The {\bf $<$case$>$} and {\bf $<$agr$>$} features are passed
  from the empty NP to the extracted NP.  This is particularly
  important for extractions from subject NP's, since {\bf $<$case$>$}
  can continue to be assigned from the verb to the subject NP
  position, and from there be passed to the extracted NP.
  
\item The {\bf $<$inv$>$} feature of S$_{r}$ is co-indexed to the {\bf
    $<$wh$>$} feature of NP through the use of the {\bf $<$invlink$>$}
  feature in order to force subject-auxiliary inversion where needed
  (see section~\ref{topicalization} for more discussion of the {\bf
    $<$inv$>$}/{\bf$<$wh$>$} co-indexing and the use of these trees
  for topicalization).

\end{itemize}

\section{Topicalization and the value of the {\bf $<$inv$>$} feature}
\label{topicalization}

Our analysis of topicalization uses the same trees as wh-extraction.  For any
NP complement position a single tree is used for both wh-questions and for
topicalization from that position. Wh-questions have subject-auxiliary
inversion and topicalizations do not.  This difference between the
constructions is captured by equating the values of the S$_{r}$'s {\bf
$<$inv$>$} feature and the extracted NP's {\bf $<$wh$>$} feature.  This means
that if the extracted item is a wh-expression, as in wh-questions, the value of
{\bf $<$inv$>$} will be {\bf +} and an inverted auxiliary will be forced to
adjoin. If the extracted item is a non-wh, {\bf $<$inv$>$} will be {\bf --}
and no auxiliary adjunction will occur. An additional complication is that
inversion only occurs in matrix clauses, so the values of {\bf $<$inv$>$} and
{\bf $<$wh$>$} should only be equated in matrix clauses and not in embedded
clauses.  In the English XTAG grammar, appropriate equating of the {\bf
$<$inv$>$} and {\bf $<$wh$>$} features is accomplished using the {\bf
$<$invlink$>$} feature and a restriction imposed on the root S of a
derivation. In particular, in extraction trees that are used for both
wh-questions and topicalizations, the value of the {\bf $<$inv$>$} feature for
the top of the S$_{r}$ node is co-indexed to the value of the {\bf $<$inv$>$}
feature on the bottom of the S$_{q}$ node.  On the bottom of the S$_{q}$ node
the {\bf $<$inv$>$} feature is co-indexed to the {\bf $<$invlink$>$} feature.
The {\bf $<$wh$>$} feature of the extracted NP node is co-indexed to the value
of the {\bf $<$wh$>$} feature on the bottom of S$_{q}$. The linking between the
value of the S$_{q}$ {\bf $<$wh$>$} and the {\bf $<$invlink$>$} features is
imposed by a condition on the final root node of a derivation (i.e. the top S
node of a matrix clause) requires that {\bf $<$invlink$>$=$<$wh$>$}.  For
example, the tree in Figure~\ref{alphaW1nx0Vnx1} is used to
derive both (\ex{1}) and (\ex{2}).

\enumsentence{John, I like.}
\enumsentence{Who do you like?}

For the question in (\ex{0}), the extracted item {\it who} has the feature
value {\bf $<$wh$>$=+}, so the value of the {\bf $<$inv$>$} feature on VP is
also $+$ and an auxiliary, in this case {\it do}, is forced to adjoin.  For the
topicalization (\ex{-1}) the values for {\it John}'s {\bf $<$wh$>$} feature and
for S$_{q}$'s {\bf $<$inv$>$} feature are both {\bf --} and no auxiliary
adjoins.

\section{Extracted subjects}
\label{subject-extraction}

The extracted subject trees provide for sentences like (\ex{1})-(\ex{3}),
depending on the tree family with which it is associated.

\enumsentence{Who left?}
\enumsentence{Who wrote the paper?}
\enumsentence{Who was happy?}

Wh-questions on subjects differ from other argument extractions in
not having subject-auxiliary inversion.  This means that in subject
wh-questions the linear order of the constituents is the same as in
declaratives so it is difficult to tell whether the subject has moved
out of position or not (see \cite{heycock/kroch93gagl} for arguments
for and against moved subject). 

The English XTAG treatment of subject extractions assumes the
following:

\begin{itemize}
\item Syntactic subject topicalizations don't exist; and 
\item Subjects in wh-questions are extracted rather than in situ.
\end{itemize}

The assumption that there is no syntactic subject topicalization is reasonable
in English since there is no convincing syntactic evidence and since the
interpretability of subjects as topics seems to be mainly affected by discourse
and intonational factors rather than syntactic structure. As for the assumption
that wh-question subjects are extracted, these questions seem to have more
similarities to other extractions than to the two cases in English that have
been considered in situ wh: multiple wh questions and echo questions. In
multiple wh questions such as sentence~(\ex{1}), one of the wh-items is blocked
from moving sentence initially because the first wh-item already occupies the
location to which it would move.

\enumsentence{Who ate what?}

This type of `blocking' account is not applicable to
subject wh-questions because there is no obvious candidate to do the
blocking.  Similarity between subject wh-questions and echo questions
is also lacking.  At least one account of echo questions
(\cite{hockey94}) argues that echo questions are not ordinary
wh-questions at all, but rather focus constructions in which the
wh-item is the focus. Clearly, this is not applicable to subject
wh-questions. So it seems that treating subject wh-questions similarly
to other wh-extractions is more justified than an in situ treatment. 

Given these assumptions, there must be separate trees in each tree family for
subject extractions. The declarative tree cannot be used even though the linear
order is the same because the structure is different. Since topicalizations are
not allowed, the {\bf $<$wh$>$} feature for the extracted NP node is set in
these trees to {\bf +}.  The lack of subject-auxiliary inversion is handled
by the absence of the {\bf $<$invlink$>$} feature.  Without the presence of
this feature, the {\bf $<$wh$>$} and {\bf $<$inv$>$} are never linked, so
inversion can not occur.  Like other wh-extractions, the S$_{q}$ node is marked
{\bf $<$extracted$>$=+} to constrain the occurrence of these trees in
embedded sentences. The tree in Figure~\ref{alphaW0nx0V} is an example of a
subject wh-question tree.

\begin{figure}[htb]
\centering
\begin{tabular}{c}
\psfig{figure=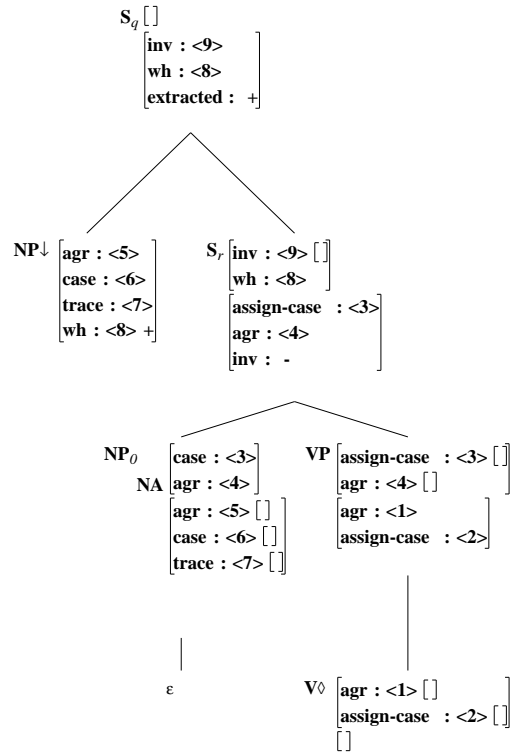,height=10.3cm}
\end{tabular}
\caption{Intransitive tree with subject extraction: $\alpha$W0nx0V}
\label{alphaW0nx0V}
\label{1;4,13} 
\end{figure}

\section{Wh-moved NP complement}
\label{NP-extr}

Wh-questions can be formed on every NP object or indirect object that appears
in the declarative tree or in the passive trees, as seen in sentences
(\ex{1})-(\ex{6}).  A tree family will contain one tree for
each of these possible NP complement positions.
Figure~\ref{ditrans-extractions} shows the two extraction trees from the
ditransitive tree family for the extraction of the direct
(Figure~\ref{ditrans-extractions}(a)) and indirect object
(Figure~\ref{ditrans-extractions}(b)).

\enumsentence{Dania asked Beth a question.}
\enumsentence{Who$_{i}$ did Dania ask $\epsilon_{i}$ a question?}
\enumsentence{What$_{i}$ did Dania ask Beth $\epsilon_{i}$?}
\enumsentence{Beth was asked a question by Dania.}
\enumsentence{Who$_{i}$ was Beth asked a question by $\epsilon_{i}$??}
\enumsentence{What$_{i}$ was Beth asked $\epsilon_{i}$? by Dania?}

\begin{figure}[htb]
\centering
\begin{tabular}{ccc}
\psfig{figure=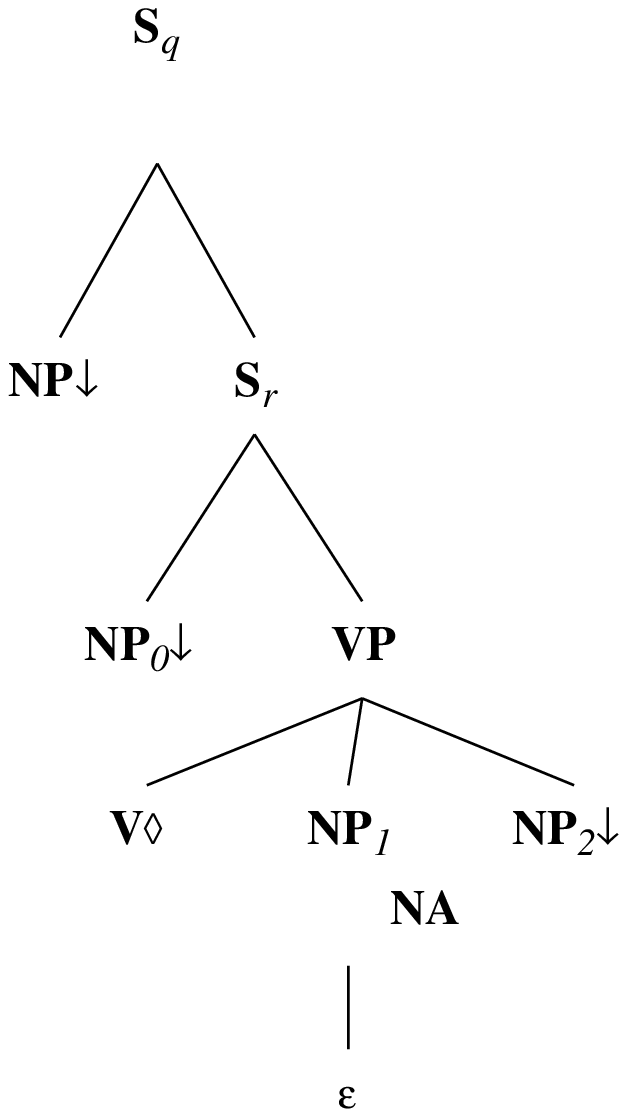,height=6.0cm}&
\hspace{1.0in}&
\psfig{figure=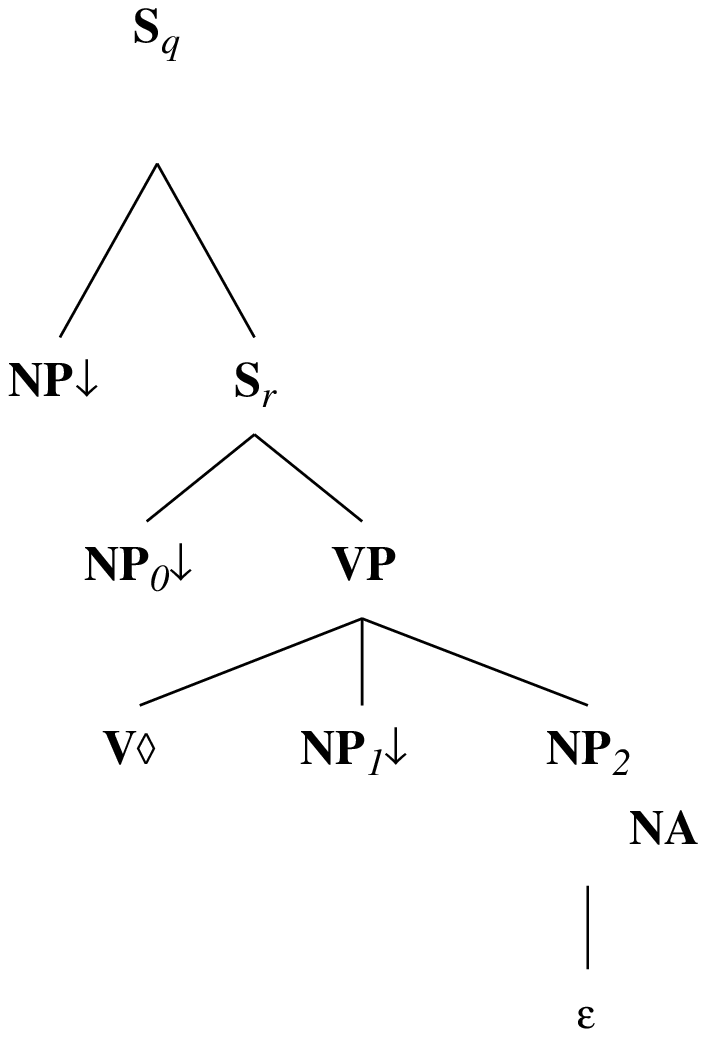,height=6.0cm}\\
(a)&&(b)
\end{tabular}
\caption{Ditransitive trees with direct object: $\alpha$W1nx0Vnx1nx2 (a) and
indirect object extraction: $\alpha$W2nx0Vnx1nx2 (b)}
\label{ditrans-extractions}
\label{2;5,3}
\end{figure}

\section{Wh-moved object of a P}
Wh-questions can be formed on the NP object of a complement PP as in
sentence~(\ex{1}).

\enumsentence{$[$Which dog$]_{i}$ did Beth Ann give a bone to $\epsilon_{i}$?}

The {\it by} phrases of passives behave like complements and can undergo the
same type of extraction, as in (\ex{1}).

\enumsentence{$[$Which dog$]_{i}$ was the frisbee caught by $\epsilon_{i}$?}

Tree structures for this type of sentence are very similar to those for the
wh-extraction of NP complements discussed in section~\ref{NP-extr} and have the
identical important features related to tree structure and trace and inversion
features.  The tree in Figure~\ref{alphaW2nx0Vnx1pnx2} is an example of this
type of tree.  Topicalization of NP objects of prepositions is handled the same
way as topicalization of complement NP's.

\begin{figure}[htb]
\centering
\mbox{}
\psfig{figure=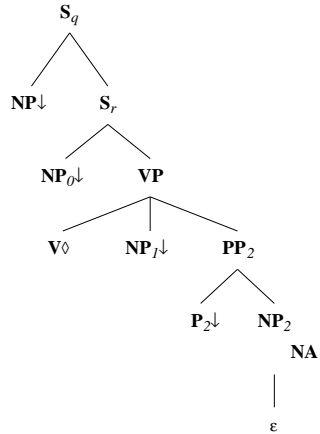,height=6.0cm}
\caption{Ditransitive with PP tree with the object of the PP extracted: $\alpha$W2nx0Vnx1pnx2}
\label{alphaW2nx0Vnx1pnx2}
\label{2;8,4}
\end{figure}

\section{Wh-moved PP}
Like NP complements, PP complements can be extracted to form
wh-questions, as in sentence (\ex{1}).

\enumsentence{[To which dog]$_{i}$ did Beth Ann throw the frisbee $\epsilon_{i}$?}

As can be seen in the tree in Figure~\ref{alphapW2nx0Vnx1pnx2}, extraction of
PP complements is very similar to extraction of NP complements from the same
positions.

\begin{figure}[htb]
\centering
\mbox{}
\psfig{figure=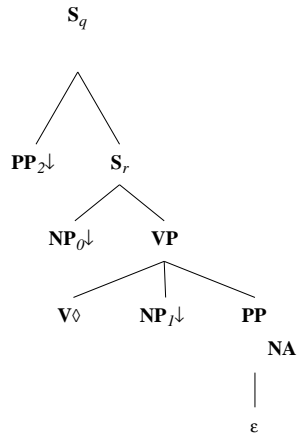,height=6.0cm}
\caption{Ditransitive with PP with PP extraction tree: $\alpha$pW2nx0Vnx1pnx2}
\label{alphapW2nx0Vnx1pnx2} 
\label{2;9,4}
\end{figure}

The PP extraction trees differ from NP extraction trees in having a PP
rather than an NP left daughter node under S$_{q}$ and in having the
$\epsilon$ fill a PP rather than an NP position in the VP. In other
respects these PP extraction structures behave like the NP extractions,
including being used for topicalization.

\section{Wh-moved S complement}

Except for the node label on the extracted position, the trees for wh-questions
on S complements look exactly like the trees for wh-questions on NP's in the
same positions.  This is because there is no separate wh-lexical item for
clauses in English, so the item {\it what} is ambiguous between representing a
clause or an NP.  To illustrate this ambiguity notice that the question in
(\ex{1}) could be answered by either a clause as in (\ex{2}) or an NP as in
(\ex{3}).  The extracted NP in these trees is constrained to be {\bf
$<$wh$>$=+}, since sentential complements can not be topicalized.

\enumsentence{What does Clove want?}
\enumsentence{for Beth Ann to play frisbee with her}
\enumsentence{a biscuit}

\section{Wh-moved Adjective complement}
In subcategorizations that select an adjective complement, that
complement can be questioned in a wh-question, as in sentence~(\ex{1}).

\enumsentence{How$_{i}$ did he feel $\epsilon_{i}$?}

\begin{figure}[htb]
\centering
\mbox{}
\psfig{figure=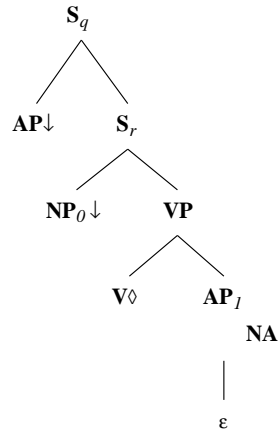,height=6.0cm}
\caption{Predicative Adjective tree with extracted adjective: $\alpha$WA1nx0Vax1}
\label{wh-adj-extr} 
\label{1;7,14}
\end{figure}

The tree families with adjective complements include trees for such adjective
extractions that are very similar to the wh-extraction trees for other
categories of complements.  The adjective position in the VP is filled by an
{\it $\epsilon$} and the trace feature of the adjective complement and of the
adjective daughter of S$_{q}$ are co-indexed.  The extracted adjective is
required to be {\bf $<$wh$>$=+}\footnote{{\it How} is the only {\bf
$<$wh$>$=+} adjective currently in the XTAG English grammar.}, so no
topicalizations are allowed.  An example of this type of tree is shown in
Figure~\ref{wh-adj-extr}.

\chapter{Relative Clauses}
\label{rel_clauses}

Relative clauses are NP modifiers, which involve extraction of
an argument or an adjunct. The NP head (the
portion of the NP being modified by the relative clause) is 
not directly related to the extracted element. 
For example in (\ex{1}), {\it the person} is the head NP
and is modified by the relative clause {\it whose mother $\epsilon$ 
likes Chris}. {\em The person} is not interpreted as the subject of the
relative clause which is missing an overt subject. In other cases, such
as (\ex{2}), the relationship between the head NP {\em export exhibitions}
may seem to be more direct but even there we assume that there are two
independent relationships: one between the entire relative clause
and the NP it modifies, and another between the extracted element
and its trace. The extracted element may be an overt {\em wh}-phrase
as in (\ex{1}) or a covert element as in (\ex{2}). 

\enumsentence{the person whose mother likes Chris}
\enumsentence{export exhibitions that included high-tech items}

Relative clauses are represented in the English XTAG grammar by auxiliary trees
that adjoin to NP's. These trees are anchored by the verb in the clause and
appear in the appropriate tree families for the various verb
subcategorizations. Within a tree family there will be groups of relative
clause trees based on the declarative tree and each passive tree. Within each
of these groups, there is a separate relative clause tree corresponding to each
possible argument that can be extracted from the clause. There is
no  relationship between the extracted position and the head NP.
The relationship between the relative clause and the head NP is treated
as a semantic relationship which will be provided by any reasonable
compositional theory. The relationship between the extracted element
(which can be covert) is captured by co-indexing the
{\bf $<$trace$>$} features of the extracted NP and the NP$_{w}$ node in the
relative clause tree. If for example, it is {\bf NP$_{0}$} that is extracted,
we have the following feature equations:\\
{\bf NP$_{w}$.t:$\langle$ trace $\rangle =$NP$_{0}$.t:$\langle$ trace $\rangle$}\\
{\bf NP$_{w}$.t:$\langle$ case $\rangle =$NP$_{0}$.t:$\langle$ case $\rangle$}\\
{\bf NP$_{w}$.t:$\langle$ agr $\rangle =$NP$_{0}$.t:$\langle$ agr $\rangle$}
\footnote{
No adjunct traces are represented in the XTAG analysis of adjunct extraction.
Relative clauses on adjuncts do not have traces and consequently feature
equations of the kind shown here are not present.}

Representative examples from the transitive tree family
are shown with a relevant subset of their features in
Figures~\ref{trans-rel-clause-trees}(a) and \ref{trans-rel-clause-trees}(b).
Figure~\ref{trans-rel-clause-trees}(a) involves a relative clause with a 
covert extracted element, while figure~\ref{trans-rel-clause-trees}(b)
involves a relative clause with an overt {\em wh}-phrase.\footnote{
The convention followed in naming relative clause trees is outlined
in Appendix~\ref{tree-naming}.}

\begin{figure}[htb]
\begin{tabular}{cc}
\psfig{figure=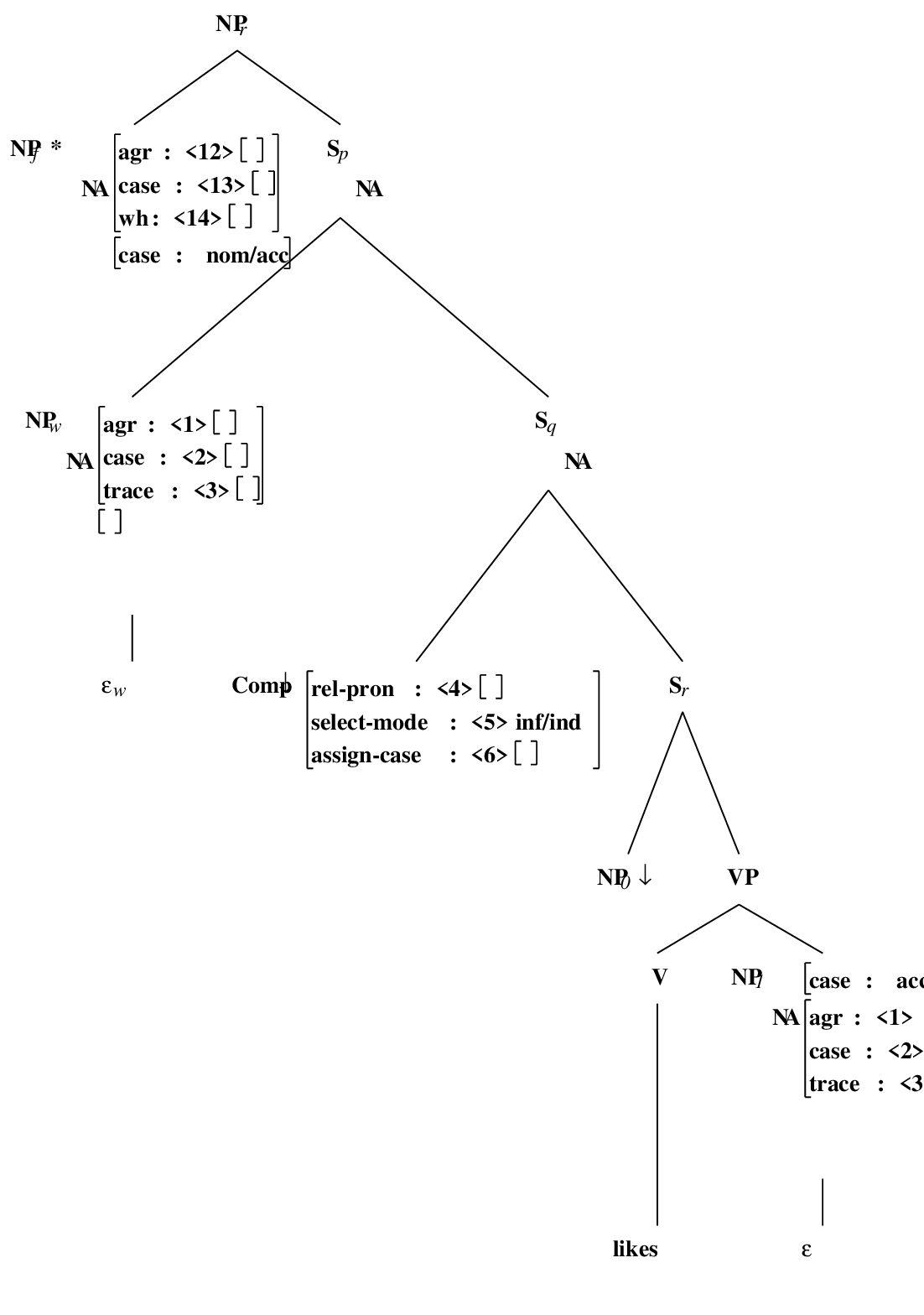,height=10.0cm}&
\psfig{figure=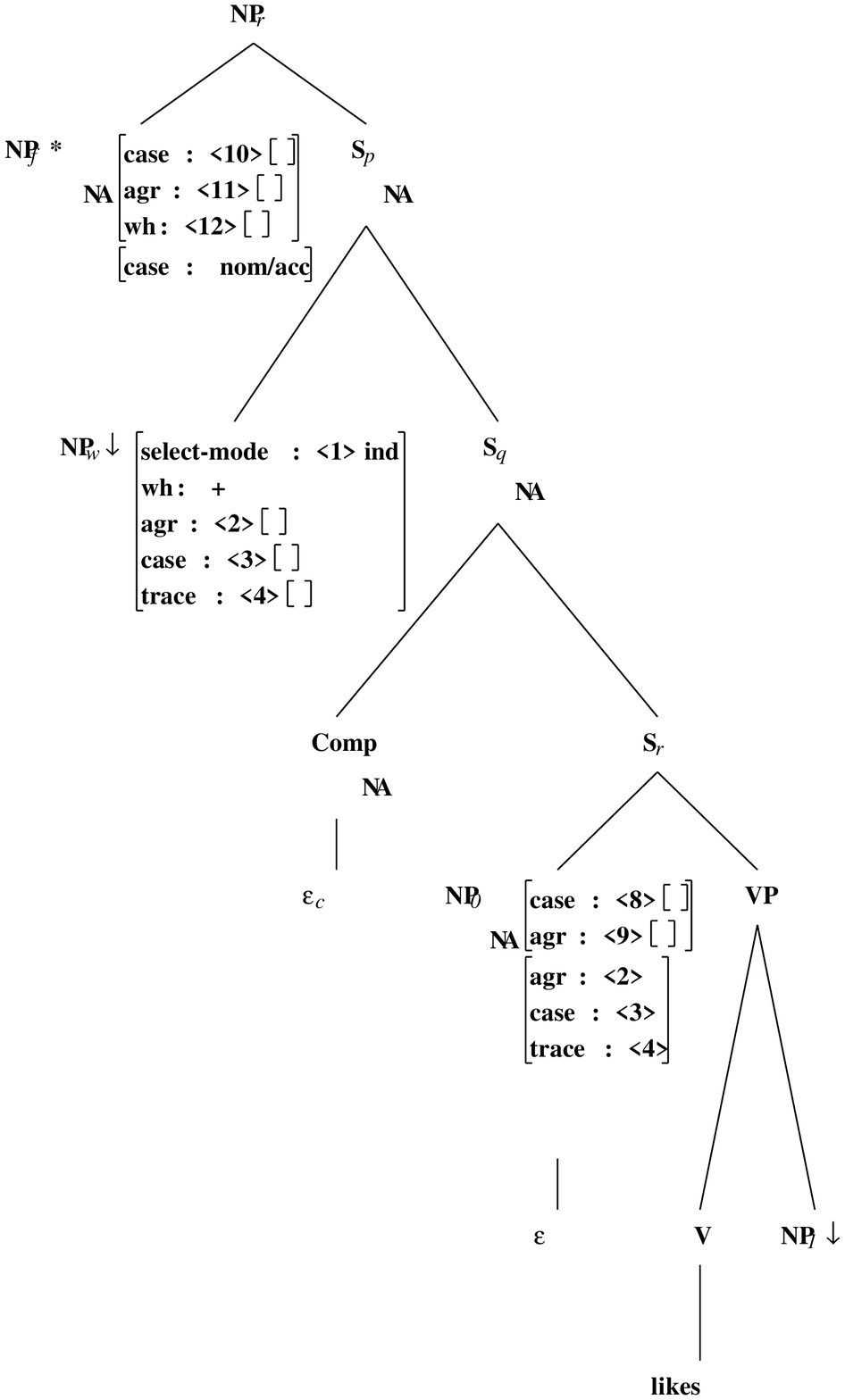,height=10.0cm}\\
(a)&(b)
\end{tabular}
\caption{Relative clause trees in the transitive tree family: $\beta$Nc1nx0Vnx1
(a) and $\beta$N0nx0Vnx1 (b)}
\label{trans-rel-clause-trees}
\label{2;16,1}
\label{2;15,1}
\end{figure}

The above analysis is essentially identical to the GB analysis of 
relative clauses. One aspect of its implementation is that 
an covert {\bf $+<$wh$>$} NP and a covert Comp have to be introduced.
See  (\ex{1}) and (\ex{2}) for example.

\enumsentence{export exhibitions [ [$_{NP_{w}}$$\epsilon$]$_{i}$ [ that [ $\epsilon$$_{i}$ included high-tech items]]]}
\enumsentence{the export exhibition [ [$_{NP_{w}}$$\epsilon$]$_{i}$ [ $\epsilon$$_{C}$ [Muriel planned  $\epsilon$$_{i}$]]]}

The lexicalized nature of XTAG makes it problematic to have trees headed by
null strings. Of the two null trees, NP$_{w}$ and Comp, that we could postulate,
the former is definitely more undesirable because it would lead to 
massive overgeneration, as can be seen in (\ex{1}) and (\ex{2}).

\enumsentence{* [$_{NP_{w}}$$\epsilon$] did John eat the apple? (as a {\em wh}-question)}
\enumsentence{* I wonder [[$_{NP_{w}}$$\epsilon$] Mary likes John](as an indirect question)}

The presence of an initial headed by a null Comp does not lead to 
problems of overgeneration because relative clauses are the only 
environment with a Comp substitution node. \footnote{Complementizers
in clausal complementation are introduced by adjunction. See
section \ref{comp-distr}.}

Consequently. our treatment of relative clauses has different 
trees to handle relative clauses with an overt extracted {\em wh}-NP
and relative clauses with a covert extracted {\em wh}-NP. Relative
clauses with an overt extracted {\em wh}-NP involve substitution
of a $+${\bf $<$wh$>$} NP into the NP$_{w}$ node
\footnote{The feature equation used is
{\bf NP$_{w}$.t:$<$wh$> = +$}. Examples of NPs that could substitute under
NP$_{w}$ are {\em whose mother}, {\em who}, {\em whom}, and also 
{\em which} but not {\em when} and {\em where} which are treated as exhaustive 
$+${\em wh} PPs.
}
and have a Comp node headed 
by $\epsilon$$_{C}$ built in. Relative clauses with a covert extracted 
{\em wh}-NP have a NP$_{w}$ node headed by $\epsilon$$_{w}$ built in and
involve substitution into the Comp node. The Comp node that is introduced
by substitution can be the $\epsilon$$_{C}$ (null complementizer), {\em that},
and {\em for}. 

For example, the tree shown in
Figure~\ref{trans-rel-clause-trees}(b) is used for the relative
clauses shown in sentences (\ex{1})-(\ex{2}), while the tree shown
in Figure~\ref{trans-rel-clause-trees}(a) is used for the relative
clauses in sentences (\ex{3})-(\ex{6}).

\enumsentence{the man who Muriel likes}
\enumsentence{the man whose mother Muriel likes}

\enumsentence{the man Muriel likes}
\enumsentence{the book for Muriel to read}
\enumsentence{the man that Muriel likes}
\enumsentence{the book Muriel is reading}

Cases of PP pied-piping (cf. \ex{1}) are handled in a similar fashion
 by building in a PP$_{w}$ node.
\enumsentence{the demon by whom Muriel was chased}
See the tree in Figure~\ref{trans-rel-clause-trees2}. 

\begin{figure}[htb]
\begin{tabular}{cc}
\centerline{\psfig{figure=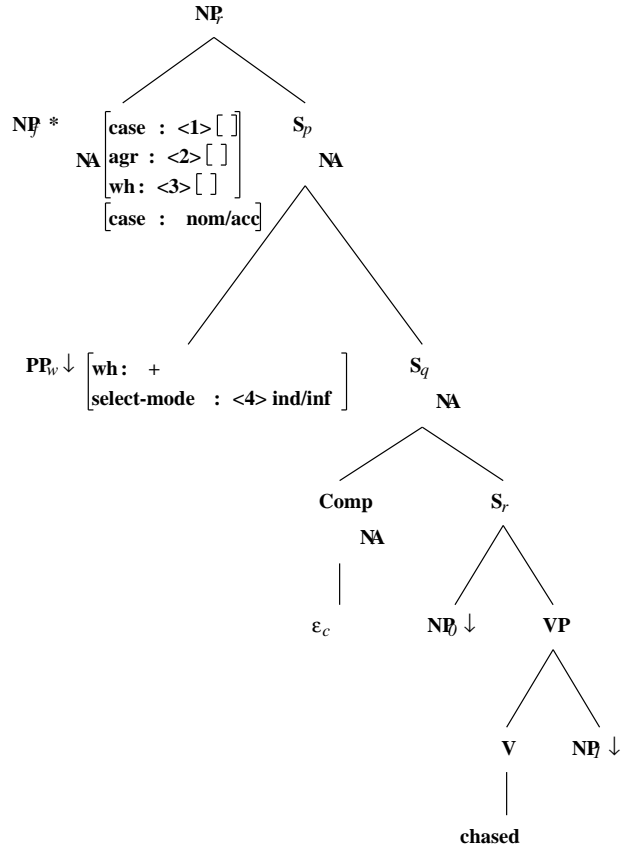,height=12.0cm}}
\end{tabular}
\caption{Adjunct relative clause tree with PP-pied-piping in the transitive tree family: 
$\beta$Npxnx0Vnx1}
\label{trans-rel-clause-trees2}
\label{2;Npxnx0Vnx1}
\end{figure}

\section{Complementizers and clauses}
The co-occurrence constraints that exist between various Comps
and the clause type of the clause they occur with are 
implemented through combinations of different
clause types using the {\bf $<$mode$>$} feature, the {\bf $<$select-mode$>$}
feature, and the {\bf $<$rel-pron$>$} feature. 

Clauses are specified for the {\bf $<$mode$>$} feature which indicates
the clause type of that clause. Possible values for the {\bf $<$mode$>$}
feature are {\bf ind, inf, ppart, ger} etc. 

Comps are lexically specified for a feature named {\bf $<$select-mode$>$}.
In addition, the {\bf $<$select-mode$>$} feature of the Comp is 
equated with the  {\bf $<$mode$>$} feature of its complement S by the following equation:\\
{\bf S$_{r}$.t:$\langle$mode$\rangle =$ Comp.t:$\langle$select-mode$\rangle$}

The lexical specifications of the Comps are shown below:
\begin{itemize}
\item $\epsilon$$_{C}$, {\bf Comp.t:$\langle$select-mode$\rangle 
=$ind/inf/ger/ppart}
\item {\em that}, {\bf Comp.t:$\langle$select-mode$\rangle =$ind}
\item {\em for}, {\bf Comp.t:$\langle$select-mode$\rangle =$inf}
\end{itemize}

The following examples display the co-occurence constraints which 
the {\bf $<$select-mode$>$} specifications assigned above implement.

For $\epsilon$$_{C}$:
\enumsentence{the book Muriel likes ({\bf S.t:$<$mode$> =$ ind})}
\enumsentence{a book to like ({\bf S.t:$<$mode$> =$ inf})}
\enumsentence{the girl reading the book ({\bf S.t:$<$mode$> =$ ger})}
\enumsentence{the book read by Muriel ({\bf S.t:$<$mode$> =$ ppart})}

For {\em for}:
\enumsentence{*the book for Muriel likes ({\bf S.t:$<$mode$> =$ ind})}
\enumsentence{a book for Mary to like ({\bf S.t:$<$mode$> =$ inf})}
\enumsentence{*the girl for reading the book ({\bf S.t:$<$mode$> =$ ger})}
\enumsentence{*the book for read by Muriel ({\bf S.t:$<$mode$> =$ ppart})}

For {\em that}:
\enumsentence{the book that Muriel likes ({\bf S.t:$<$mode$> =$ ind})}
\enumsentence{*a book that (Muriel) to like ({\bf S.t:$<$mode$> =$ inf})}
\enumsentence{*the girl that reading the book ({\bf S.t:$<$mode$> =$ ger})}
\enumsentence{*the book that read by Muriel ({\bf S.t:$<$mode$> =$ ppart})}

Relative clause trees that have substitution of {\bf NP$_{w}$} have
the following feature equations:\\
{\bf S$_{r}$.t:$\langle$mode$\rangle =$ NP$_{w}$.t:$\langle$select-mode$\rangle$}\\
{\bf NP$_{w}$.t:$\langle$select-mode$\rangle =$ind}

The examples that follow are intended to provide the rationale for 
the above setting of features.
\enumsentence{
the boy whose mother chased the cat ({\bf S$_{r}$.t:$\langle$mode$\rangle =$ind})}
\enumsentence{
*the boy whose mother to chase the cat ({\bf S$_{r}$.t:$\langle$mode$\rangle =
$inf})}
\enumsentence{
*the boy whose mother eaten the cake ({\bf S$_{r}$.t:$\langle$mode$\rangle 
=$ppart})}
\enumsentence{
*the boy whose mother chasing the cat ({\bf S$_{r}$.t:$\langle$mode$\rangle =$
ger})}
\enumsentence{
the boy [whose mother]$_{i}$ Bill believes $\epsilon$$_{i}$ to chase the cat\\ ({\bf S$_{r}$.t:
$\langle$mode$\rangle =$ind})}

The feature equations that appear in trees which have substitution of 
{\bf PP$_{w}$} are:\\
{\bf S$_{r}$.t:$\langle$mode$\rangle =$ PP$_{w}$.t:$\langle$select-mode$\rangle$}\\
{\bf PP$_{w}$.t:$\langle$mode$\rangle =$ind/inf} \footnote{As is the case for
{\bf NP$_{w}$} substitution, any  $+${\bf wh}-PP can substitute under PP$_{w}$.
This is implemented by the following equation:\\
{\bf PP$_{w}$.t:$\langle$wh$\rangle = +$}

Not all cases of pied-piping involve substitution of {\bf PP$_{w}$}.
In some cases, the P may be built in. In cases where part of the pied-piped
PP is part of the anchor, it continues to function as an anchor even after
pied-piping i.e. the P node and the {\bf NP$_{w}$} nodes are represented
separately.
}

Examples that justify the above feature setting follow.
\enumsentence{
the person [by whom] this machine was invented ({\bf S$_{r}$.t:$\langle$mode$\rangle =$ind})}
\enumsentence{
a baker [in whom]$_{i}$ PRO to trust $\epsilon$$_{i}$ ({\bf S$_{r}$.t:$\langle$mode$\rangle =$
inf})}
\enumsentence{
*the fork [with which] (Geoffrey) eaten the pudding ({\bf S$_{r}$.t:$\langle$
mode$\rangle =$ppart})}
\enumsentence{
*the person [by whom] (this machine) inventing ({\bf S$_{r}$.t:$\langle$mode
$\rangle =$ger})}

\subsection{Further constraints on the null Comp $\epsilon$$_{C}$}
There are additional constraints on where the null Comp $\epsilon$$_{C}$
can occur. The null Comp is not permitted in cases of subject
extraction unless there is an intervening clause or or
the relative clause is a reduced relative ({\bf mode = ppart/ger}).
This can be seen in (\ex{1}-\ex{4}). 

\enumsentence{
*the toy [$\epsilon$$_{i}$ [$\epsilon$$_{C}$ [ $\epsilon$$_{i}$ likes Dafna]]]}
\enumsentence{
the toy [$\epsilon$$_{i}$ [$\epsilon$$_{C}$ Fred thinks [ $\epsilon$$_{i}$ likes Dafna]]]}
\enumsentence{
the boy [$\epsilon$$_{i}$ [$\epsilon$$_{C}$ [ $\epsilon$$_{i}$ eating the guava]]]}
\enumsentence{
the guava [$\epsilon$$_{i}$ [$\epsilon$$_{C}$ [ $\epsilon$$_{i}$ eaten by the boy]]]}

To model this paradigm, the feature {\bf $\langle$rel-pron$\rangle$} is used in
conjunction with the following equations:

\begin{itemize}
\item {\bf S$_{r}$.t:$\langle$rel-pron$\rangle =$ Comp.t:$\langle$rel-pron$\rangle$}
\item {\bf S$_{r}$.b:$\langle$rel-pron$\rangle =$ S$_{r}$.b:$\langle$mode$\rangle$}
\item {\bf Comp.b:$\langle$rel-pron$\rangle =$ppart/ger/adj-clause}
(for $\epsilon$$_{C}$)
\end{itemize}

The full set of the equations shown above is only present in Comp
substitution trees involving subject extraction. So (\ex{1}) will
not be ruled out.

\enumsentence{
the toy [$\epsilon$$_{i}$ [$\epsilon$$_{C}$ [ Dafna likes $\epsilon$$_{i}$ ]]]}

The feature mismatch induced by the above equations 
is not remedied by adjunction of just any S-adjunct
because all other S-adjuncts
are transparent to the {\bf $\langle$rel-pron$\rangle$} feature
because of the following equation:\\
{\bf S$_{m}$.b:$\langle$rel-pron$\rangle =$ S$_{f}$.t:$\langle$rel-pron$\rangle$}

\section{Reduced Relatives}
Reduced relatives are permitted only in cases of subject-extraction.
Past participial reduced relatives are only permitted on passive
clauses.
See (\ex{1}-\ex{8}).

\enumsentence{
the toy [$\epsilon$$_{i}$ [$\epsilon$$_{C}$ [ $\epsilon$$_{i}$ playing the banjo]]]
}
\enumsentence{
*the instrument [$\epsilon$$_{i}$ [$\epsilon$$_{C}$ [ Amis playing $\epsilon$$_{i}$ ]]]
}
\enumsentence{
*the day [$\epsilon$$_{w}$ [$\epsilon$$_{C}$ [ Amis playing the banjo]]]
}
\enumsentence{
the apple [$\epsilon$$_{i}$ [$\epsilon$$_{C}$ [ $\epsilon$$_{i}$ eaten by Dafna]]]
}
\enumsentence{
*the child [$\epsilon$$_{i}$ [$\epsilon$$_{C}$ [ the apple eaten by $\epsilon$$_{i}$ ]]]
}
\enumsentence{
*the day [$\epsilon$$_{w}$ [$\epsilon$$_{C}$ [ Amis eaten the apple]]]
}
\enumsentence{
*the apple [$\epsilon$$_{i}$ [$\epsilon$$_{C}$ [ Dafna eaten $\epsilon$$_{i}$ ]]]
}
\enumsentence{
*the child [$\epsilon$$_{i}$ [$\epsilon$$_{C}$ [ $\epsilon$$_{i}$ eaten the apple ]]]
}

These restrictions are built into the {\bf $<$mode$>$} specifications
of {\bf S.t}. So non-passive cases of subject extraction have the following
feature equation:\\
{\bf S$_{r}$.t:$\langle$mode$\rangle =$ ind/ger/inf}

Passive cases of subject extraction have the following
feature equation:\\
{\bf S$_{r}$.t:$\langle$mode$\rangle =$ ind/ger/ppart/inf}

Finally, all cases of non-subject extraction have the following
feature equation:\\
{\bf S$_{r}$.t:$\langle$mode$\rangle =$ ind/inf}\\

\subsection{Restrictive vs. Non-restrictive relatives}

The English XTAG grammar does not contain any  syntactic distinction between
restrictive and non-restrictive relatives because we believe this to
be a semantic and/or pragmatic difference.

\section{External syntax}
A relative clause can combine with the NP it modifies in at least 
the following two ways:

\enumsentence{\ [the [toy [$\epsilon$$_{i}$ [$\epsilon$$_{C}$ [Dafna likes $\epsilon$$_{i}$ ]]]]]
}
\label{n-attach-ex}
\enumsentence{\ [[the toy] [$\epsilon$$_{i}$ [$\epsilon$$_{C}$ [Dafna likes $\epsilon$$_{i}$ ]]]]
}
\label{np-attach-ex}

Based on cases like (\ex{1}) and (\ex{2}), which are problematic for the
structure in (\ref{n-attach-ex}), the structure in (\ref{np-attach-ex}) is adopted.

\enumsentence{ [[the man and the woman] [who met on the bus]]}
\enumsentence{ [[the man and the woman] [who like each other]]} 

As it stands, the RC analysis sketched so far will combine in two
ways with the Determiner tree shown in Figure~(\ref{trans-rel-clause-trees3}),
\footnote{The determiner tree shown has the {\bf $<$rel-clause$>$} 
feature built in. The RC analysis would give two
parses in the absence of this feature.}
giving us both the possiblities shown in (\ref{n-attach-ex}) and (\ref{np-attach-ex}). In order
to block the structure exemplified in (\ref{n-attach-ex}), the feature 
{\bf $\langle$rel-clause$\rangle$} is used in combination with the following
equations.

\begin{figure}[htb]
\begin{tabular}{cc}
\centerline{\psfig{figure=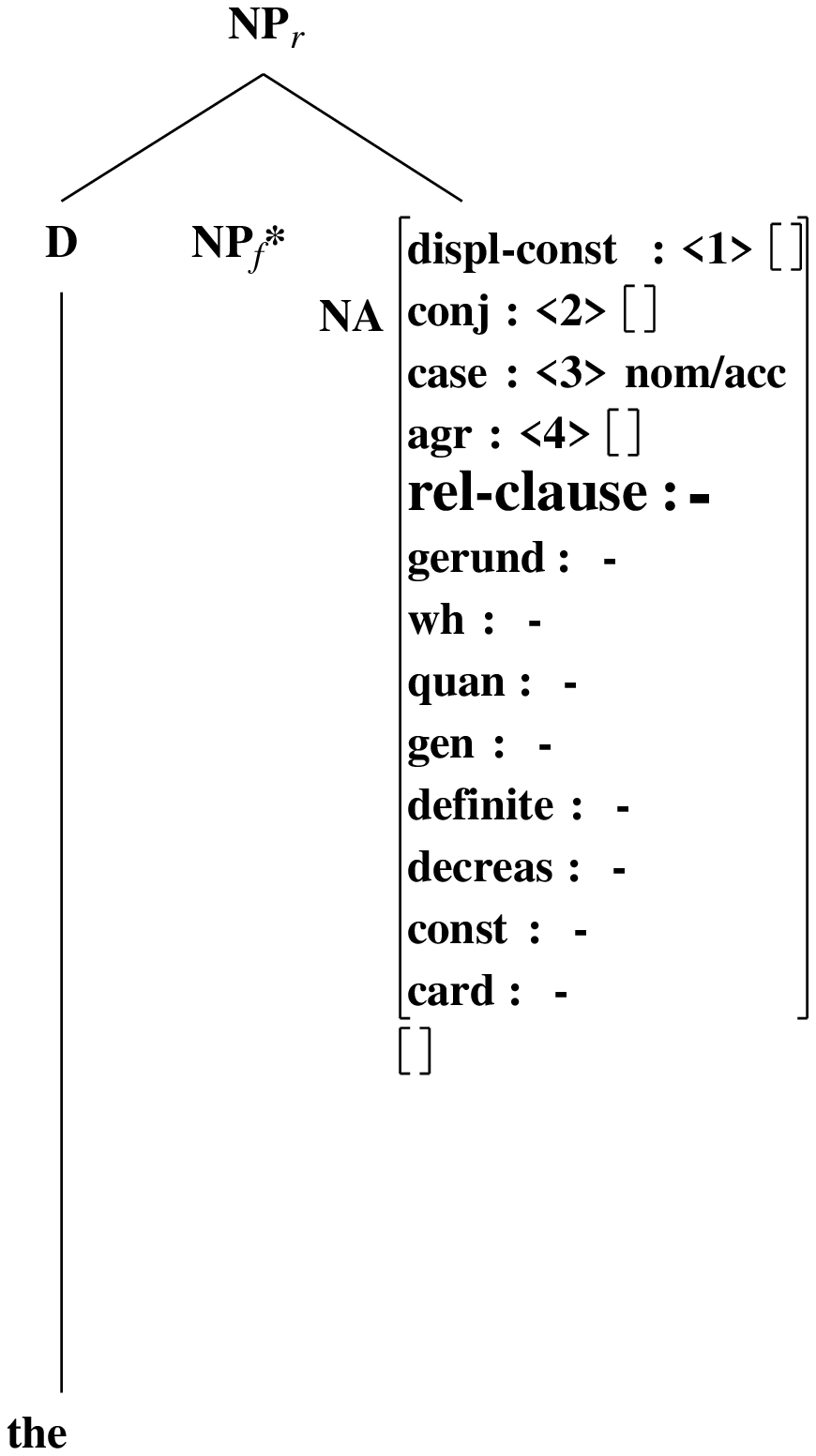,height=10.0cm}}
\end{tabular}
\label{trans-rel-clause-trees3}
\caption{Determiner tree with {\bf $<$rel-clause$>$} feature: $\beta$Dnx}
\end{figure}

On the RC:\\
{\bf NP$_{r}$.b:$\langle$rel-clause$\rangle = +$}

On the Determiner tree:\\
{\bf NP$_{f}$.t:$\langle$rel-clause$\rangle = -$}

Together, these equations block
introduction of the determiner above the relative clause.

\section{Other Issues}

\subsection{Interaction with adjoined Comps}
The XTAG analysis now has two different ways of introducing a 
complementizer like {\em that} or {\em for}, depending upon whether
it occurs in a relative clause or in sentential complementation. 
Relative clause complementizers substitute in (using the
tree $\alpha$Comp), while sentential complementizers adjoin in
(using the tree $\beta$COMPs). Cases like (\ex{1}) where 
both kinds of complementizers illicitly occur together are blocked.

\enumsentence{*the book [$\epsilon$$_{w_{i}}$ [that [that [Muriel wrote 
$\epsilon$$_{i}$]]]]} 

This is accomplished by setting the {\bf S$_{r}$.t:$<$comp$>$} feature
in the relative clause tree to {\bf nil}. The {\bf S$_{r}$.t:$<$comp$>$} 
feature of the auxiliary tree that introduces 
(the sentential complementation) {\em that} is set to
{\bf that}. This leads to a feature clash ruling out (\ex{0}). On the
other hand, if a sentential complement taking verb is adjoined
in at S$_{r}$, this feature clash goes away (cf. \ex{1}).

\enumsentence{the book [$\epsilon$$_{w_{i}}$ [that Beth thinks [that [Muriel wrote
$\epsilon$$_{i}$]]]]}

\subsection{Adjunction on PRO}
Adjunction on PRO, which would yield the ungrammatical (\ex{1}) is blocked.

\enumsentence{*I want [[PRO [who Muriel likes] to read a book]].}
This is done by specifying the {\bf $<$case$>$} feature of {\bf NP$_{f}$} to be
{\bf nom/acc}. The {\bf $<$case$>$} feature of PRO is {\bf null}. This
leads to a feature clash and blocks adjunction of relative clauses on to
PRO.

\subsection{Adjunct relative clauses}
Two types of trees to handle adjunct relative clauses exist in the 
XTAG grammar: one in which there is {\bf PP$_{w}$} substitution with 
a null {\bf Comp} built in and one in which there is a null {\bf NP$_{w}$}
built in and a {\bf Comp} substitutes in. There is no {\bf NP$_{w}$}
substitution tree with a null {\bf Comp} built in. This is because of
the contrast between (\ex{1}) and (\ex{2}).
\enumsentence{the day [[on whose predecessor] [$\epsilon$$_{C}$ [Muriel left]]]}
\enumsentence{*the day [[whose predecessor] [$\epsilon$$_{C}$ [Muriel left]]]}
In general, adjunct relatives are not possible with an overt {\bf NP$_{w}$}. 
We do not consider (\ex{1}) and (\ex{2}) to be counterexamples to 
the above statements because we consider {\em where} and {\em when}
to be exhaustive {\bf PP}s that head a {\bf PP} initial tree.

\enumsentence{the place [where [$\epsilon$$_{C}$ [Muriel wrote her first book]]]}
\enumsentence{the time [when [$\epsilon$$_{C}$ [Muriel lived in Bryn Mawr]]]}

\subsection{ECM}
Cases where {\em for} assigns exceptional case (cf. \ex{1}, \ex{2}) are handled.

\enumsentence{a book [$\epsilon$$_{w_{i}}$ [for [Muriel to read $\epsilon$$_{i}$]]]}
\enumsentence{the time [$\epsilon$$_{w_{i}}$ [for [Muriel to leave Haverford]]]}

The assignment of case by {\em for} is implemented by a combination of the
following equations.\\
{\bf Comp.t:$\langle$assign-case$\rangle =$acc}\\
{\bf S$_{r}$.t:$\langle$assign-case$\rangle =$Comp.t:$\langle$assign-case$\rangle$}\\
{\bf S$_{r}$.b:$\langle$assign-case$\rangle =$NP$_{0}$.t:$\langle$case$\rangle$}

\section{Cases not handled}
\subsection{Partial treatment of free-relatives}
Free relatives are only partially handled. All free relatives on non-subject
positions and some free relatives on subject positions 
are handled. The structure assigned 
to free relatives treats the extracted {\em wh}-NP as the head NP of
the relative clause. The remaining relative clause modifies this
extracted {\em wh}-NP (cf. \ex{1}-\ex{3}).

\enumsentence{what(ever) [$\epsilon$$_{w_{i}}$ [$\epsilon$$_{C}$ 
[Mary likes $\epsilon$$_{i}$]]]}
\enumsentence{where(ever) [$\epsilon$$_{w}$ [$\epsilon$$_{C}$ 
[Mary lives]]]}
\enumsentence{who(ever) [$\epsilon$$_{w_{i}}$ [$\epsilon$$_{C}$ 
[Muriel thinks [$\epsilon$$_{i}$ likes Mary]]]]}

However, simple subject extractions without further emebedding are not
handled (cf. \ex{1}).

\enumsentence{who(ever) [$\epsilon$$_{w_{i}}$ [$\epsilon$$_{C}$ [$\epsilon$$_{i}$ likes Bill]]]}
This is because (\ex{-1}) is treated exactly like the ungrammatical (\ex{1}).
\enumsentence{*the person [ $\epsilon$$_{w_{i}}$ [$\epsilon$$_{C}$
[$\epsilon$$_{i}$ likes Bill]]]}

\subsection{Adjunct P-stranding}
The following cases of adjunct preposition stranding are not handled 
(cf. \ex{1}, \ex{2}).

\enumsentence{the pen Muriel wrote this letter with}
\enumsentence{the street Muriel lives on}

Adjuncts are not built into elementary trees in XTAG. So there is no
clean way to represent adjunct preposition stranding. A better
solution is, probably , available if we make use of multi-component
adjunction. 

\subsection{Overgeneration}
The following ungrammatical sentences are currently being 
accepted by the XTAG grammar. This is because no clean 
and conceptually attractive way of ruling them out
is obvious to us.

\subsubsection{{\em how} as {\em wh}-NP}
In standard American English, {\em how} is not acceptable as a 
relative pronoun (cf. \ex{1}).

\enumsentence{*the way [how [$\epsilon$$_{C}$ [PRO to solve this problem]]]}

However, (\ex{0}) is accepted by the current grammar.
The only way to rule (\ex{0}) out would be to introduce a special feature
devoted to this purpose. This is unappealing. Further, there exist
speech registers/dialects of English, where (\ex{0}) is acceptable. 

\subsubsection{{\em for}-trace effects}
(\ex{1}) is ungrammatical, being an instance of a violation of the
{\em for}-trace filter of early transformational grammar.

\enumsentence{the person [$\epsilon$$_{w_{i}}$ [for 
[$\epsilon$$_{i}$ to read the book]]]}

The XTAG grammar currently accepts (\ex{0}).\footnote{It may be of
some interest that (\ex{0}) is acceptable in certain dialects of Belfast
English.}

\subsubsection{Internal head constraint}
Relative clauses in English (and in an overwhelming number of languages)
obey a `no internal head' constraint. This constraint is exemplified in
the contrast between (\ex{1}) and (\ex{2}).

\enumsentence{the person [who$_{i}$ [$\epsilon$$_{C}$ 
Muriel likes $\epsilon$$_{i}$]]}
\enumsentence{*the person [[which person]$_{i}$ [$\epsilon$$_{C}$ 
Muriel likes $\epsilon$$_{i}$]]}

We know of no good way to rule (\ex{0}) out, while still ruling (\ex{1}) in.
\enumsentence{the person [[whose mother]$_{i}$ [$\epsilon$$_{C}$
Muriel likes $\epsilon$$_{i}$]]}

Dayal (1996) suggests that `full' NPs such as {\em which person} and
{\em whose mother} are R-expressions while {\em who} and {\em whose}
are pronouns. R-expressions, unlike pronouns, are subject to Condition C.
(\ex{-2}) is, then, ruled out as a violation of Condition C since {\em 
the person} and {\em which person} are co-indexed and {\em the person}
c-commands {\em which person}. If we accept Dayal's argument, we 
have a principled reason for allowing overgeneration of relative clauses
that violate the internal head constraint, the reason being that 
the XTAG grammar does generate binding theory violations.

\subsubsection{Overt Comp constraint on stacked relatives}
Stacked relatives of the kind in (\ex{1}) are handled.

\enumsentence{ [[the book [that Bill likes]] [which Mary wrote]]}

There is a constraint on stacked relatives: all but the relative clause
closest to the head-NP must have either an overt {\bf Comp} or 
an overt {\bf NP$_{w}$}. Thus (\ex{1}) is ungrammatical.

\enumsentence{*[[the book [that Bill likes]] [Mary wrote]]}

Again, no good way of handling this constraint is known to us 
currently. 

\chapter{Adjunct Clauses}
\label{adjunct-cls}
\label{sub-conj}

Adjunct clauses include subordinate clauses (i.e. those with overt
subordinating conjunctions), purpose clauses and participial adjuncts.

Subordinating conjunctions each select four trees, allowing them to
appear in four different positions relative to the matrix clause.  The
positions are (1) before the matrix clause, (2) after the matrix
clause, (3) before the VP, surrounded by two punctuation marks, and
(4) after the matrix clause, separated by a punctuation mark. Each of
these trees is shown in Figure \ref{sub-conj-trees}.

\begin{figure}[htb]
\centering
\begin{tabular}{cccc}
\psfig{figure=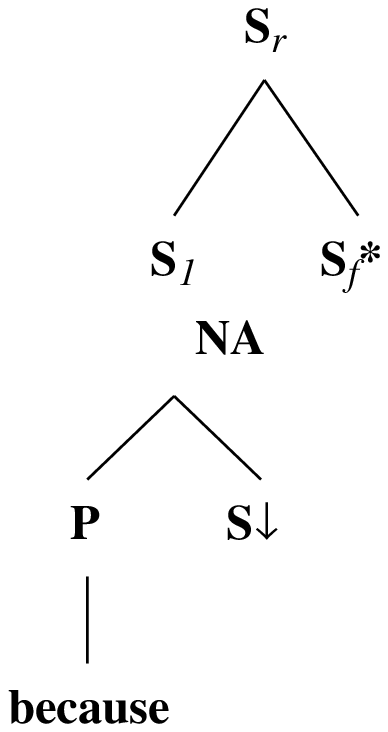,height=2.1in}&
\psfig{figure=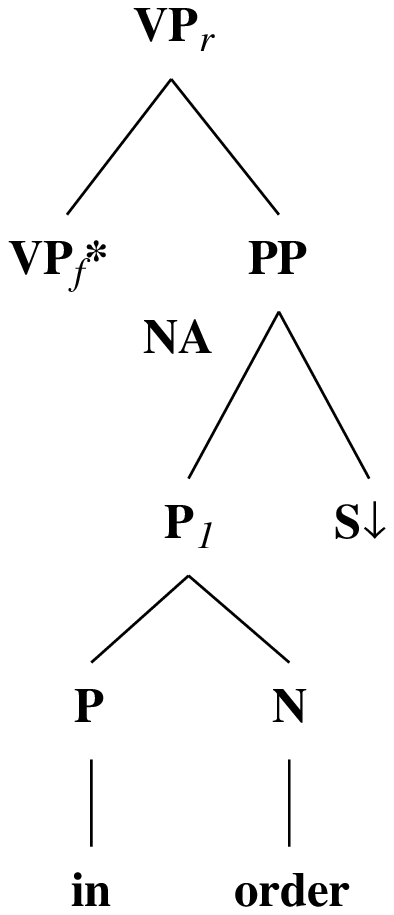,height=2.1in}&
\psfig{figure=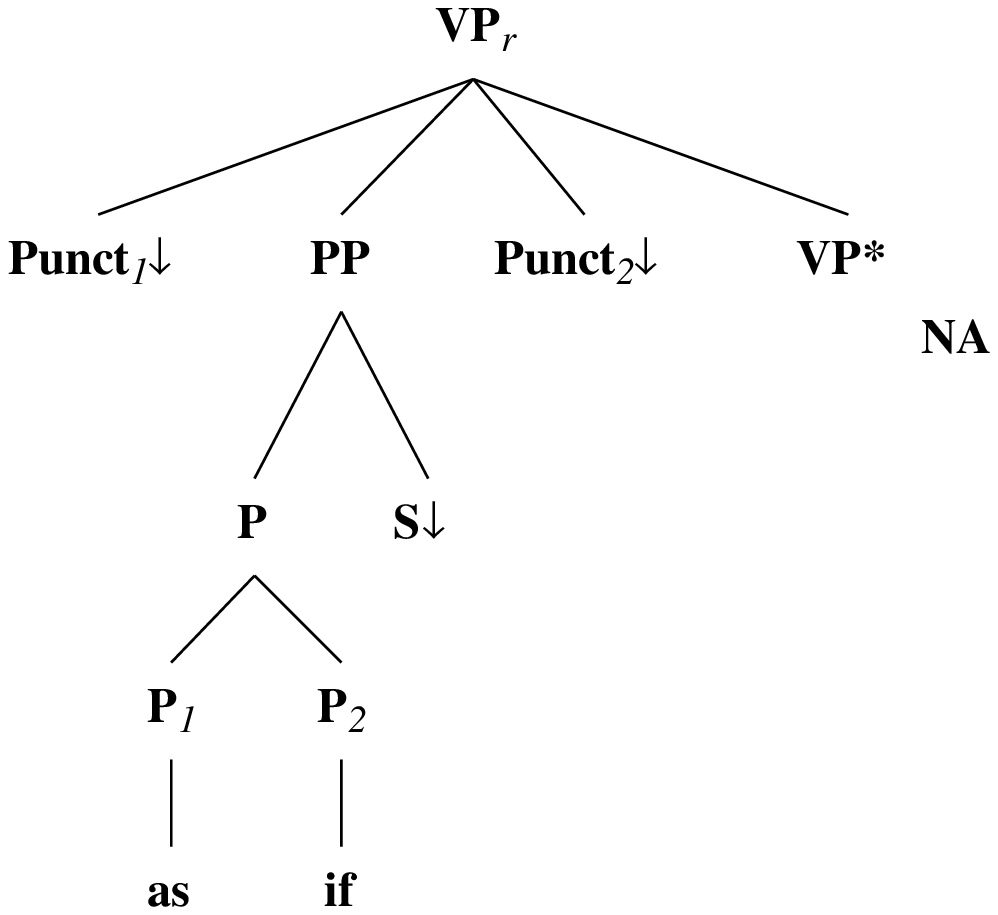,height=2.1in}&
\psfig{figure=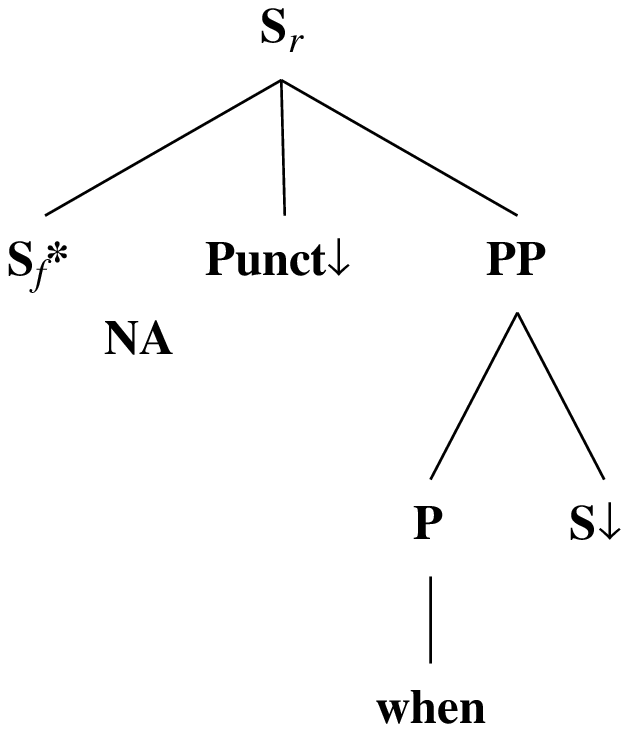,height=2in}\\
(1) $\beta$Pss & (2) $\beta$vxPNs & (3) $\beta$puPPspuvx & (4) $\beta$spuPs \\
\end{tabular}
\caption{Auxiliary Trees for Subordinating Conjunctions}
\label{sub-conj-trees}
\end{figure}

Sentence-initial adjuncts adjoin at the root S of the matrix clause,
while sentence-final adjuncts adjoin at a VP node. In this, the XTAG
analysis follows the findings on the attachment sites of adjunct
clauses for conditional clauses (\cite{iatridou91}) and for
infinitival clauses (\cite{Browning87}). One compelling argument is
based on Binding Condition C effects.  As can be seen from examples
(\ex{1})-(\ex{3}) below, no Binding Condition violation occurs when
the adjunct is sentence initial, but the subject of the matrix clause
clearly governs the adjunct clause when it is in sentence final
position and co-indexation of the pronoun with the subject of the
adjunct clause is impossible.

\enumsentence{Unless she$_i$ hurries, Mary$_i$ will be late for the meeting.}
\enumsentence{$\ast$She$_i$ will be late for the meeting unless Mary$_i$ hurries.}
\enumsentence{Mary$_i$ will be late for the meeting unless she$_i$ hurries.}

We had previously treated subordinating conjunctions as a subclass of
{\em conjunction}, but are now assigning them the POS {\em
preposition}, as there is such clear overlap between words that
function as prepositions (taking NP complements) and subordinating
conjunctions (taking clausal complements). While there are some
prepositions which only take NP complements and some which only take
clausal complements, many take both as shown in examples
(\ex{1})-(\ex{4}), and it seems to be artificial to assign them two
different parts-of-speech.

\enumsentence{Helen left before the party.}
\enumsentence{Helen left before the party began.}
\enumsentence{Since the election, Bill has been elated.}
\enumsentence{Since winning the election, Bill has been elated.}

Each subordinating conjunction selects the values of the {\bf
$<$mode$>$} and {\bf $<$comp$>$} features of the subordinated S. The
{\bf $<$mode$>$} value constrains the types of clauses the
subordinating conjunction may appear with and the {\bf $<$comp$>$}
value constrains the complementizers which may adjoin to that
clause. For instance, indicative subordinate clauses may appear with
the complementizer {\it that} as in (\ex{1}), while participial
clauses may not have any complementizers (\ex{2}).

\enumsentence{Midge left that car so that Sam could drive to work.}
\enumsentence{*Since that seeing the new VW, Midge could think of
nothing else.}

\subsection{Multi-word Subordinating Conjunctions}

We extracted a list of multi-word conjunctions, such as {\it as if},
{\it in order}, and {\it for all (that)}, from \cite{quirk85}. For the
most part, the components of the complex are all anchors, as shown in
Figures~\ref{conjs}(a). In one case, {\it as ADV as}, there is a great
deal of latitude in the choice of adverb, so this is a substitution
site (Figures~\ref{conjs}(b)). This multi-anchor treatment is very
similar to that proposed for idioms in \cite{AS89}, and the analysis
of light verbs in the XTAG grammar (see section~\ref{nx0lVN1-family}).

\begin{figure}[htb]
\centering
\begin{tabular}{ccc}
\psfig{figure=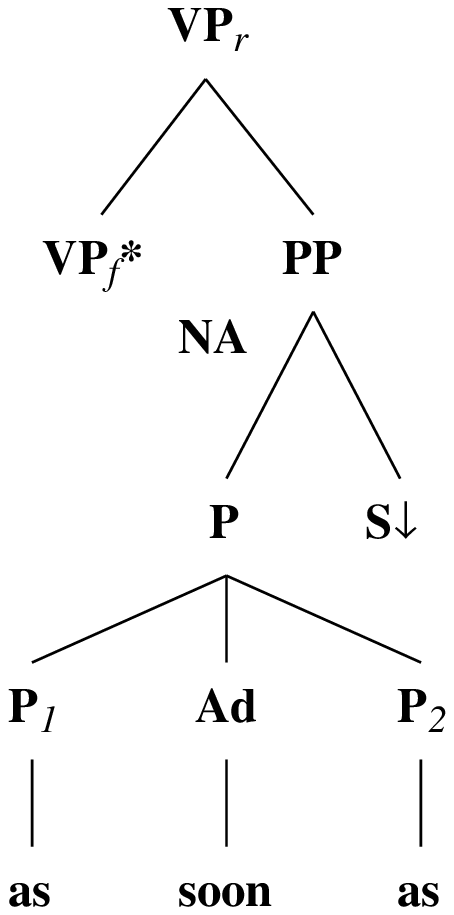,height=2.7in}&
\hspace*{0.5in} &
\psfig{figure=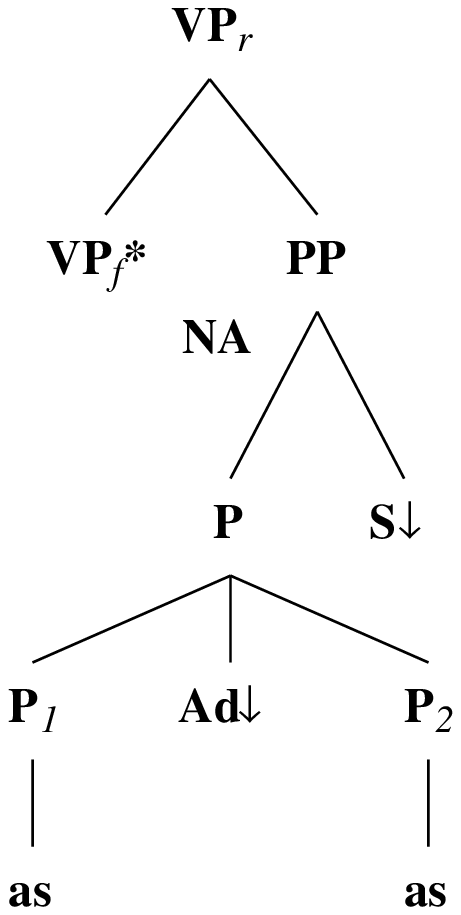,height=2.7in}\\
(a)&\hspace*{0.5in} &(b)\\
\end{tabular}
\caption{Trees Anchored by Subordinating Conjunctions:  $\beta$vxPARBPs and $\beta$vxParbPs}
\label{conjs}
\end{figure}

\section{``Bare'' Adjunct Clauses}

``Bare'' adjunct clauses do not have an overt subordinating
conjunction, but are typically parallel in meaning to clauses with
subordinating conjunctions. For this reason, we have elected to handle
them using the same trees shown above, but with null anchors. They are
selected at the same time and in the same way the {\it PRO} tree is,
as they all have {\it PRO} subjects.  Three values of {\bf $<$mode$>$}
are licensed: {\bf inf} (infinitive), {\bf ger} (gerundive) and {\bf
ppart} (past participal).\footnote{We considered allowing bare
indicative clauses, such as {\it He died that others may live}, but
these were considered too archaic to be worth the additional ambiguity
they would add to the grammar.} They interact with complementizers as
follows:

\begin{itemize}
\item Participial complements do not license any
complementizers:\footnote{While these sound a bit like extraposed
relative clauses (see \cite{kj87}), those move only to the right and
adjoin to S; as these clauses are equally grammatical both
sentence-initially and sentence-finally, we are analyzing them as
adjunct clauses.}

\enumsentence{[Destroyed by the fire], the building still stood.}
\enumsentence{The fire raged for days [destroying the building].}
\enumsentence{$\ast$[That destroyed by the fire], the building
still stood.}

\begin{figure}[htb]
\begin{tabular}{cc}
\psfig{figure=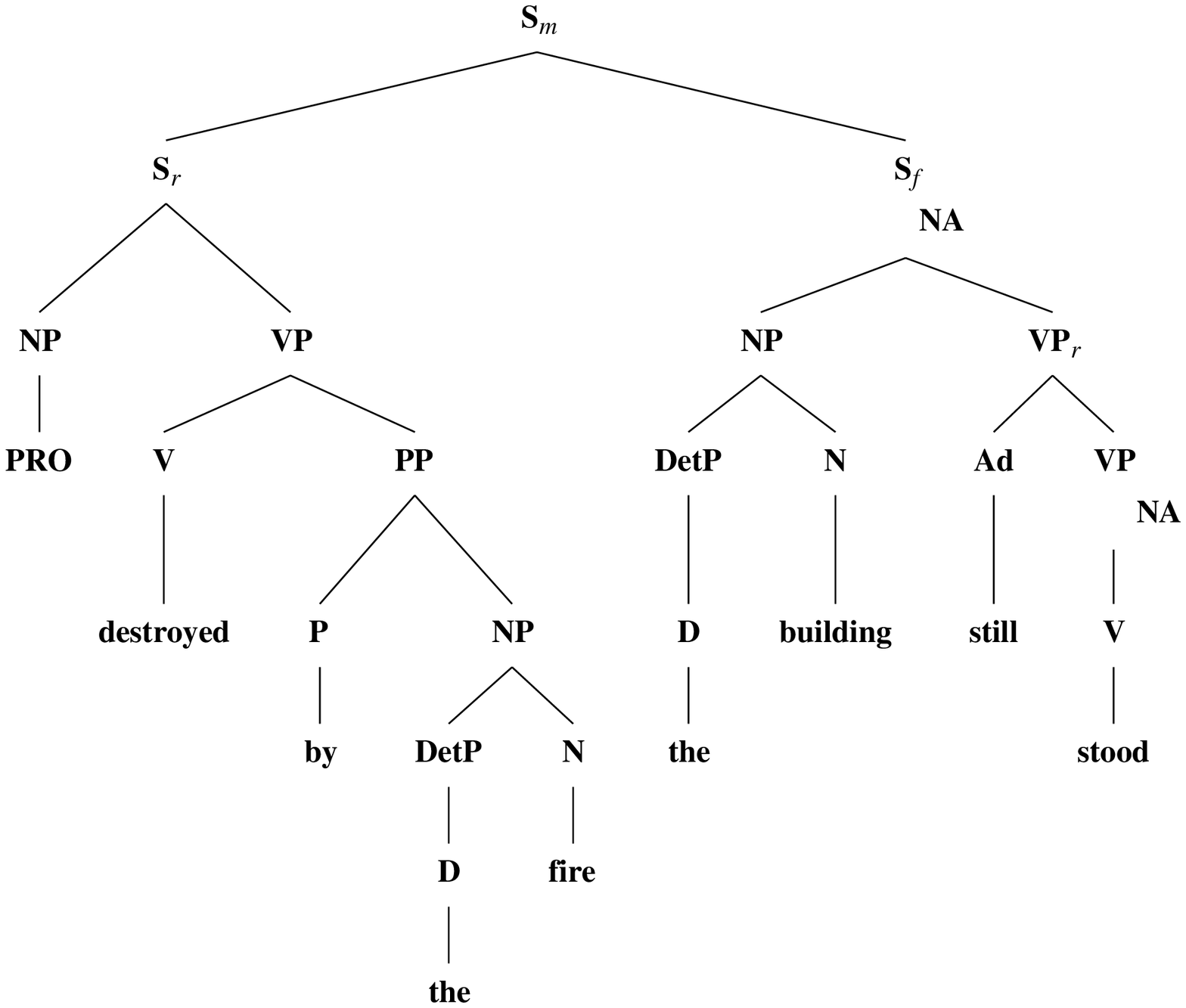,height=2.7in}&
\psfig{figure=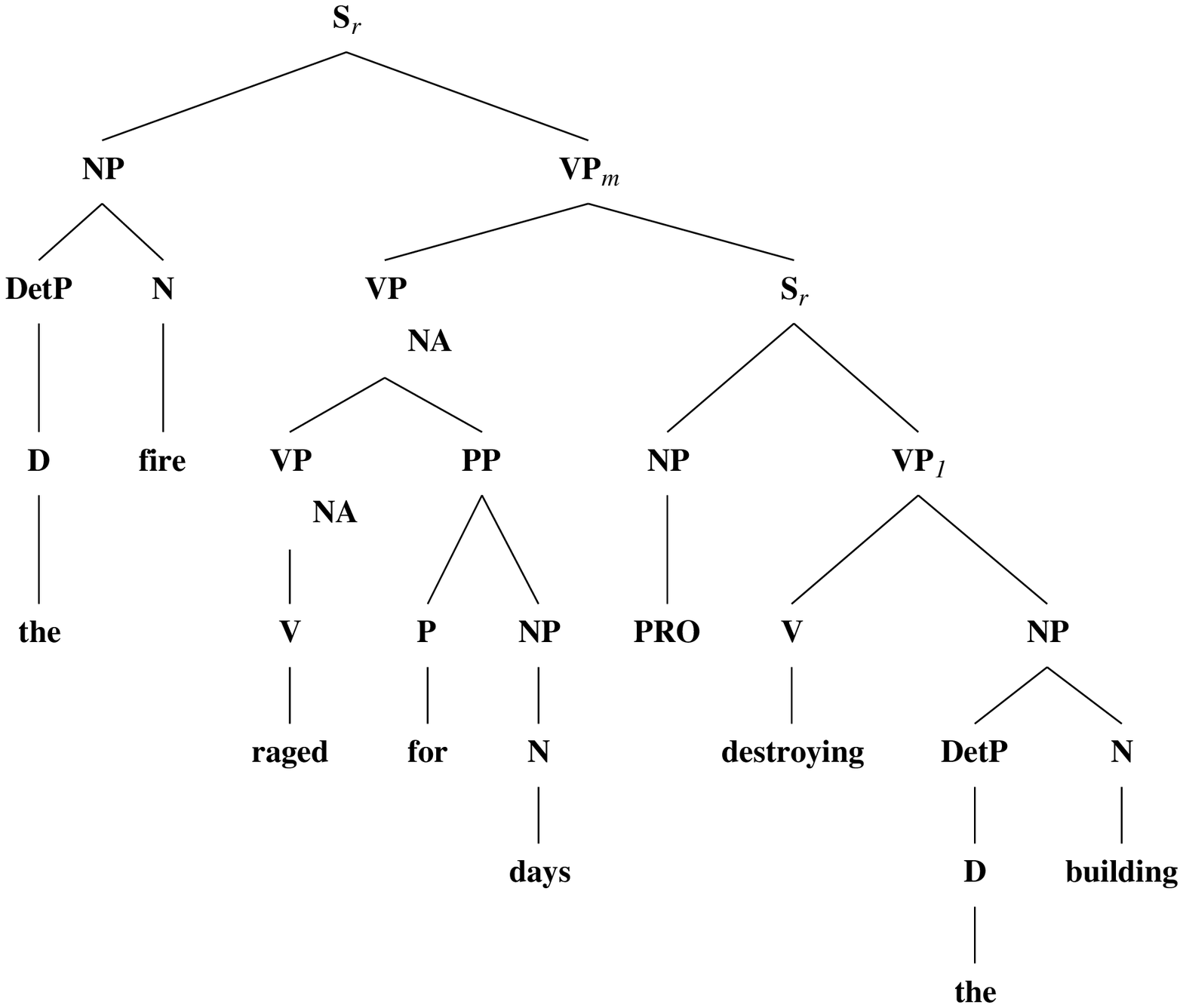,height=2.7in}\\
(a)&(b)
\end{tabular}
\caption{Sample Participial Adjuncts}
\label{destroyed}
\end{figure}

\item Infinitival adjuncts, including purpose clauses, are licensed both with and without the complementizer
{\it for}.
\enumsentence{Harriet bought a Mustang [to impress Eugene].}
\enumsentence{[To impress Harriet], Eugene dyed his hair.}
\enumsentence{Traffic stopped [for Harriet to cross the street].}
\end{itemize}

\section{Discourse Conjunction}

The CONJs auxiliary tree is used to handle `discourse' conjunction,
as in sentence (\ex{1}).  Only the coordinating conjunctions ({\it
and, or} and {\it but}) are allowed to adjoin to the roots of
matrix sentences. Discourse conjunction with {\it and} is shown in the
derived tree in Figure~\ref{seuss-sentence}.

\enumsentence{And Truffula trees are what everyone needs! \cite{seuss71}}

\begin{figure}[htbp]
\centering
\hspace{0in}
\psfig{figure=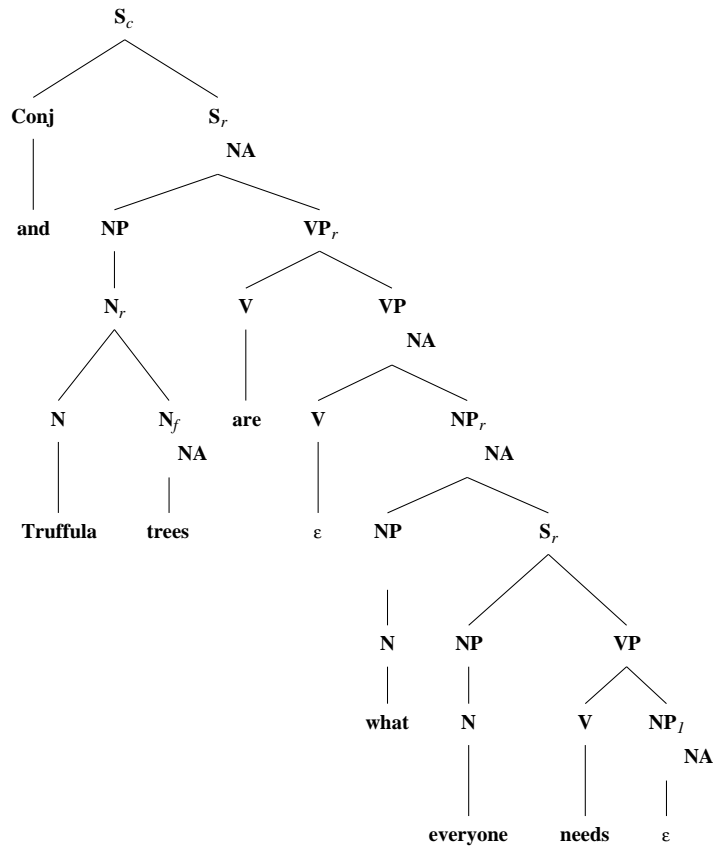,height=4.5in}
\caption{Example of discourse conjunction, from Seuss' {\it The
Lorax}\protect\nocite{seuss71}}
\label{seuss-sentence}
\end{figure}

\chapter{Imperatives}
\label{imperatives}

Imperatives in English do not require overt subjects.  The subject in
imperatives is second person, i.e.\ {\it you}, whether it is overt or
not, as is clear from the verbal agreement and the interpretation.
Imperatives with overt subjects can be  parsed using the trees already
needed for declaratives.  The imperative cases in which the subject is
not overt are handled by the imperative trees discussed in this section.

The imperative trees in English XTAG grammar are identical to the declarative
tree except that the NP$_{0}$ subject position is filled by an $\epsilon$, the
NP$_{0}$ {\bf $<$agr~pers$>$} feature is set in the tree to the value {\bf 2nd}
and the {\bf $<$mode$>$} feature on the root node has the value {\bf imp}.  The
value for {\bf $<$agr~pers$>$} is hardwired into the epsilon node and insures
the proper verbal agreement for an imperative.  The {\bf $<$mode$>$} value of
{\bf imp} on the root node is recognized as a valid mode for a matrix clause.
The {\bf imp} value for {\bf $<$mode$>$} also allows imperatives to be blocked
from appearing as embedded clauses.  Figure \ref{alphaInx0Vnx1} is the
imperative tree for the transitive tree family.

\begin{figure}[htbp]
\centering{
\begin{tabular}{c}
\psfig{figure=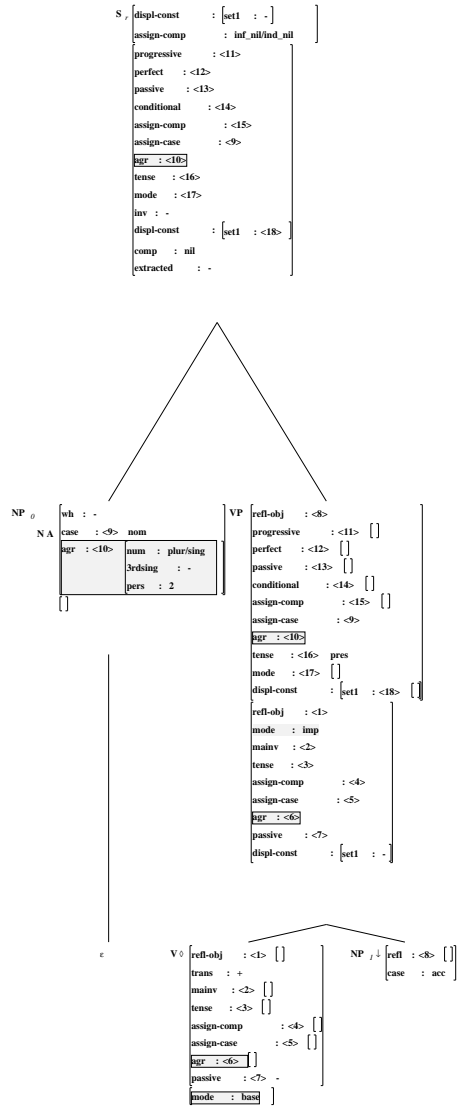,height=6in}
\end{tabular}
}
\caption{Transitive imperative tree: $\alpha$Inx0Vnx1}
\label{alphaInx0Vnx1}
\label{2;11,1}
\end{figure}

\chapter{Gerund NP's}
\label{gerunds-chapter}

There are two types of gerunds identified in the linguistics
literature. One is the class of {\it derived nominalizations} (also 
called {\it nominal gerundives} or {\it action nominalizations}) 
exemplified in (\ex{1}), which instantiates the direct object within an
{\it of} PP.
The other is the class of so-called {\it sentential} or
{\it VP gerundives} exemplified in (\ex{2}). In the English XTAG grammar,
the derived nominalizations are termed {\bf determiner gerunds}, and the
sentential or VP gerunds are termed {\bf NP gerunds}.

\enumsentence{Some think that {\bf the selling of bonds} is beneficial.}

\enumsentence{Are private markets approving of {\bf Washington bashing Wall
Street}?}

Both types of gerunds exhibit a similar distribution, appearing in most
places where NP's are allowed.\footnote{an exception being the NP positions
in ``equative BE'' sentences, such as, {\it John is my father}.}  The bold
face portions of sentences (\ex{1})--(\ex{3}) show examples of gerunds as a
subject and as the object of a preposition.

\enumsentence{{\bf Avoiding such losses} will take a monumental effort.}
\enumsentence{{\bf Mr. Nolen's wandering} doesn't make him a weirdo.}
\enumsentence{Are private markets approving of {\bf Washington bashing Wall
Street}?}

The motivation for splitting the gerunds into two classes is semantic as
well as structural in nature. Semantically, the two gerunds are in sharp
contrast with each other. NP gerunds refer to an action, i.e., a way of
doing something, whereas determiner gerunds refer to a fact. Structurally,
there are a number of properties (extensively discussed in \cite{Lees60})
that show that NP gerunds have the syntax of verbs, whereas determiner
gerunds have the syntax of basic nouns.  Firstly, the fact that the direct
object of the determiner gerund can only appear within an {\it of} PP
suggests that the determiner gerund, like nouns, is not a case assigner and
needs insertion of the preposition {\it of} for assignment of case to the
direct object. NP gerunds, like verbs, need no such insertion and can
assign case to their direct object.  Secondly, like nouns, only determiner
gerunds can appear with articles (cf. example (\ex{1}) and
(\ex{2})). Thirdly, determiner gerunds, like nouns, can be modified by
adjectives (cf. example (\ex{3})), whereas NP gerunds, like verbs, resist
such modification (cf. example (\ex{4})). Fourthly, nouns, unlike verbs,
cannot co-occur with aspect (cf. example (\ex{5}) and (\ex{6})). Finally,
only NP gerunds, like verbs, can take adverbial modification (cf. example
(\ex{7}) and (\ex{8})).

\enumsentence{\ldots the proving of the theorem\ldots. \hspace{1.0in} (det
ger with article)}
\enumsentence{* \ldots the proving the theorem\ldots. \hspace{1.0in} (NP ger
with article)}
\enumsentence{John's rapid writing of the book\ldots. \hspace{1.0in} (det
ger with Adj)}
\enumsentence{* John's rapid writing the book\ldots. \hspace{1.0in} (NP ger
with Adj)}
\enumsentence{* John's having written of the book\ldots. \hspace{1.0in}
(det ger with aspect)}
\enumsentence{John having written the book\ldots. \hspace{1.0in} (NP ger
with aspect)}
\enumsentence{* His writing of the book rapidly\ldots. \hspace{1.0in} (det
ger with Adverb)}
\enumsentence{His writing the book rapidly\ldots. \hspace{1.0in} (NP ger
with Adverb)}

In English XTAG, the two types of gerunds are assigned separate
trees within each tree family, but in order to capture their similar
distributional behavior, both are assigned NP as the category label of
their top node. The feature {\bf gerund = +/--} distinguishes gerund NP's from
regular NP's where needed.\footnote{This feature is also needed to restrict
the selection of gerunds in NP positions. For example, the subject and
object NP's in the ``equative BE'' tree (Tnx0BEnx1) cannot be filled by
gerunds, and are therefore assigned the feature {\bf
gerund = --}, which prevents gerunds (which have the feature {\bf gerund =
+}) from substituting into these NP positions.} The determiner gerund 
and the NP gerund trees are discussed in section~(\ref{detger-sec}) and
~(\ref{NPger-sec}) respectively.

\section{Determiner Gerunds}
\label{detger-sec}
The determiner gerund tree in Figure~\ref{detgerund-tree} is anchored by a
V, capturing the fact that the gerund is derived from a verb. The verb
projects an N and instantiates the direct object as an {\it of} PP. The
nominal category projected by the verb can now display all the syntactic
properties of basic nouns, as discussed above. For example, it can be
straightforwardly modified by adjectives; it cannot co-occur with aspect;
and it can appear with articles. The only difference of the determiner
gerund nominal with the basic nominals is that the former cannot occur
without the determiner, whereas the latter can. The determiner gerund 
tree therefore has an initial D modifying the N.\footnote{Note that
the determiner can adjoin to the gerund only from {\it within} the gerund
tree. Adjunction of determiners to the gerund root node is prevented by
constraining determiners to only select NP's with the feature {\bf gerund = --}.
This rules out sentences like {\it Private markets approved of (*the) [the
selling of bonds]}.} It is used for gerunds such as the ones in bold face
in sentences (\ex{1}), (\ex{2}) and (\ex{3}).

\begin{figure}[htb]
\centering
\begin{tabular}{c}
{\psfig{figure=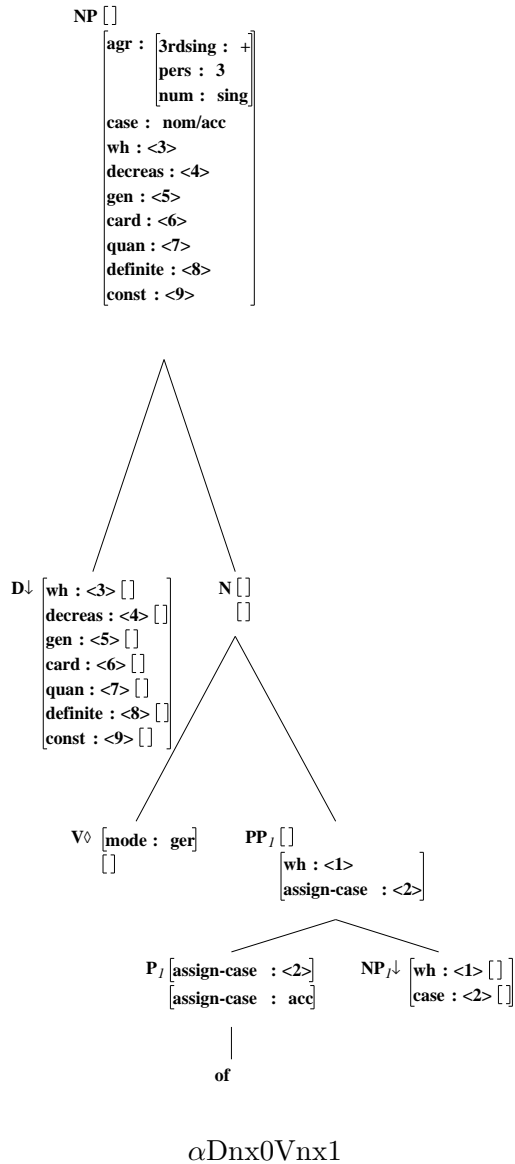,height=5.9in}}\\
$\alpha$Dnx0Vnx1\\
\end{tabular}
\caption{Determiner Gerund tree from the transitive tree family: $\alpha$Dnx0Vnx1}
\label{detgerund-tree}
\label{2;12,1}

\end{figure}

The D node can take a simple determiner (cf. example (\ex{1})), a
genitive pronoun (cf. example (\ex{2})), or a genitive NP (cf. example
(\ex{3})).\footnote{The trees for genitive pronouns and genitive NP's
have the root node labelled as D because they seem to function as
determiners and as such, sequence with the rest of the
determiners. See Chapter~\ref{det-comparitives} for discussion on
determiner trees.}

\enumsentence{Some think that {\bf the selling of bonds} is beneficial.}
\enumsentence{{\bf His painting of Mona Lisa} is highly acclaimed.}
\enumsentence{Are private markets approving of {\bf Washington's bashing of Wall Street}?}

\section{NP Gerunds}
\label{NPger-sec}
NP gerunds show a number of structural peculiarities, the crucial one being
that they have the internal properties of sentences. In the English XTAG
grammar, we adopt a position similar to that of \cite{Rosenbaum67} and
\cite{Emonds70} -- that these gerunds are NP's exhaustively dominating a
clause. Consequently, the tree assigned to the transitive NP gerund tree
(cf. Figure~\ref{NPgerund-tree}) looks exactly like the declarative
transitive tree for clauses except for the root node label and the feature
values. The anchoring verb projects a VP. Auxiliary adjunction is allowed,
subject to one constraint -- that the highest verb in the verbal sequence
be in gerundive form, regardless of whether it is a main or auxiliary verb.
This constraint is implemented by requiring the topmost VP node to be {\bf
$<$mode$>$ = ger}. In the absence of any adjunction, the anchoring verb
itself is forced to be gerundive. But if the verbal sequence has more than
one verb, then the sequence and form of the verbs is limited by the
restrictions that each verb in the sequence imposes on the succeeding
verb. The nature of these restrictions for sentential clauses, and the
manner in which they are implemented in XTAG, are both discussed in
Chapter~\ref{auxiliaries}. The analysis and implementation discussed there
differs from that required for gerunds only in one respect -- that the
highest verb in the verbal sequence is required to be {\bf $<$mode$>$ =
ger}.

\begin{figure}[htb]
\centering
\begin{tabular}{cc}
{\psfig{figure=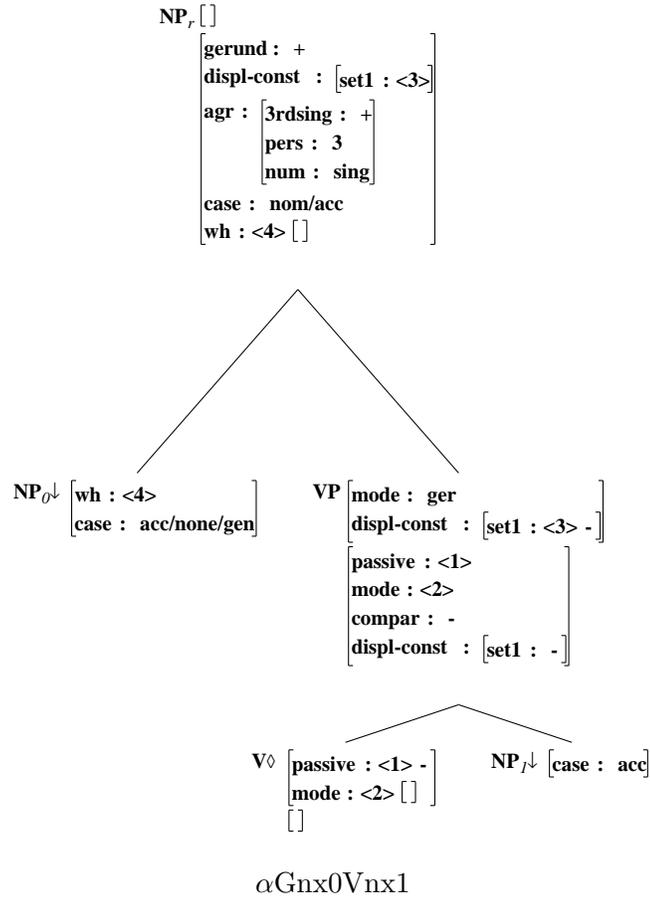,height=4.5in}}\\
$\alpha$Gnx0Vnx1\\
\end{tabular}
\caption{NP Gerund tree from the transitive tree family: $\alpha$Gnx0Vnx1}
\label{NPgerund-tree}
\label{2;13,1}
\end{figure}

Additionally, the subject in the NP gerund tree is required to have {\bf
$<$case$>$=acc/none/gen}, i.e., it can be either a PRO (cf. example
\ex{1}), a genitive NP (cf. example \ex{2}), or an accusative NP
(cf. example \ex{3}). The whole NP formed by the gerund can occur in either
nominative or accusative positions.

\enumsentence{\ldots John does not like {\bf wearing a hat}.}
\enumsentence{Are private markets approving of {\bf Washington's bashing Wall
Street}?}
\enumsentence{Mother disapproved of {\bf me wearing such casual clothes}.}

One question that arises with respect to gerunds is whether there is anything
special about their distribution as compared to other types of NP's.  In fact,
it appears that gerund NP's can occur in any NP position.  Some verbs might not
seem to be very accepting of gerund NP arguments, as in (\ex{1}) below, but we
believe this to be a semantic incompatibility rather than a syntactic problem
since the same structures are possible with other lexical items.

\enumsentence{? [$_{NP}$John's tinkering$_{NP}$] ran.}
\enumsentence{[$_{NP}$John's tinkering$_{NP}$] worked.}

By having the root node of gerund trees be NP, the gerunds have the same
distribution as any other NP in the English XTAG grammar without doing
anything exceptional. The clause structure is captured by the form of the trees
and by inclusion in the tree families.

\section{Gerund Passives}

It was mentioned above that the NP gerunds display certain clausal
properties. It is therefore not surprising that they too have their own set
of transformationally related structures. For example, NP gerunds allow
passivization just like their sentential counterparts (cf. examples
(\ex{1}) and (\ex{2})).

\enumsentence{The lawyers objected to {\bf the slanderous book being
written by John}.}
\enumsentence{Susan could not forget {\bf having
been embarrassed by the vicar}.}

In the English XTAG grammar, gerund passives are treated in an almost
exactly similar fashion to sentential passives, and are assigned separate
trees within the appropriate tree families. The passives occur in pairs,
one with the {\it by} phrase, and another without it. There are two feature
restrictions imposed on the passive trees: (a) only verbs with {\bf
$<$mode$>$ = ppart} (i.e., verbs with passive morphology) can be the
anchors, and (b) the highest verb in the verb sequence is required to be
{\bf $<$mode$>$ = ger}. The two restrictions, together, ensure the
selection of only those sequences of auxiliary verb(s) that select {\bf
$<$mode$>$ = ppart} and {\bf $<$passive$>$ = +} (such as {\it being} or
{\it having been} but NOT {\it having}). The passive trees are assumed to
be related to only the NP gerund trees (and not the determiner gerund
trees), since passive structures involve movement of some object to the
subject position (in a movement analysis). Also, like the sentential
passives, gerund passives are found in most tree families that have a
direct object in the declarative tree. Figure~\ref{pass-trees} shows
the gerund passive trees for the transitive tree family.

\begin{figure}[htb]
\centering
\begin{tabular}{cc}
{\psfig{figure=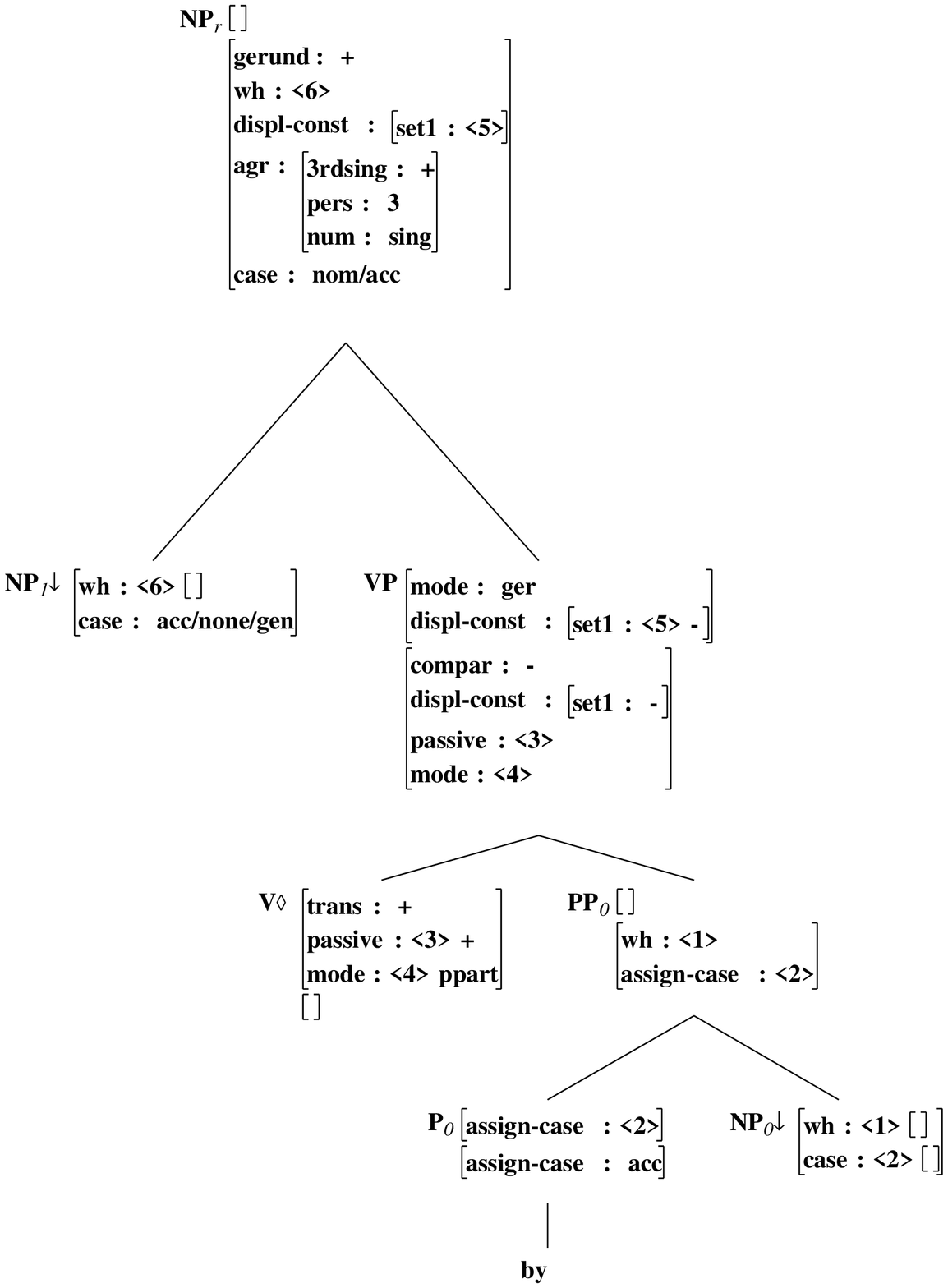,height=4.0in}}&
{\psfig{figure=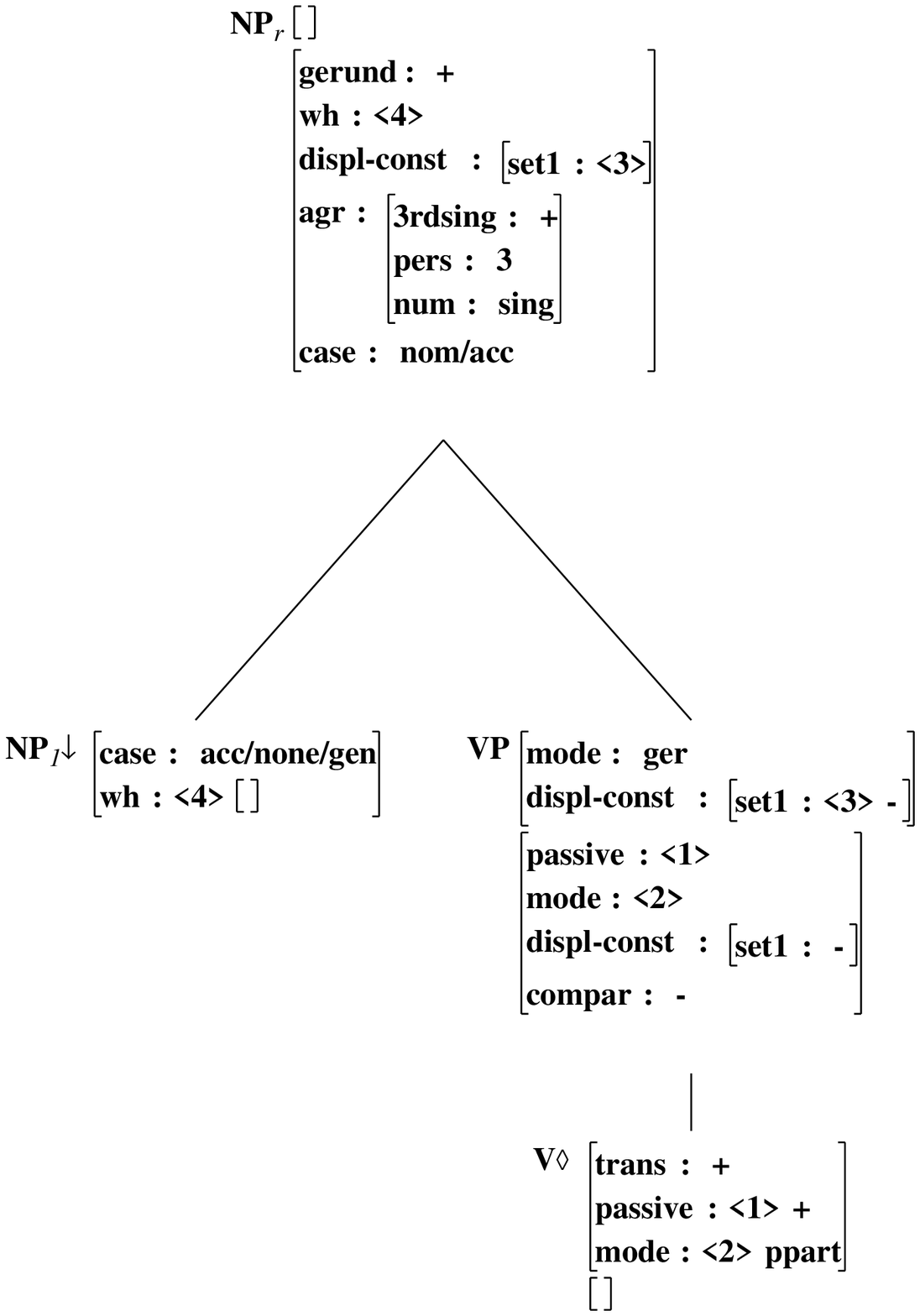,height=4.0in}}
\\
(a) $\alpha$Gnx1Vbynx0&(b) $\alpha$Gnx1V\\
\end{tabular}
\caption{Passive Gerund trees from the transitive tree family: $\alpha$Gnx1Vbynx0 (a) and
$\alpha$Gnx1V (b)}
\label{pass-trees}
\end{figure}

\part{Other Constructions}
\chapter{Determiners and Noun Phrases}
\label{det-comparitives}

In our English XTAG grammar,\footnote{A more detailed discussion of
this analysis can be found in \cite{ircs:det98}.} all nouns select the
noun phrase (NP) tree structure shown in Figure~\ref{np-tree}.  Common
nouns do not require determiners in order to form grammatical
NPs. Rather than being ungrammatical, singular countable nouns without
determiners are restricted in interpretation and can only be
interpreted as mass nouns.  Allowing all nouns to head determinerless
NPs correctly treats the individuation in countable NPs as a property
of determiners. Common nouns have negative(``-'') values for
determiner features in the lexicon in our analysis and can only
acquire a positive(``+'') value for those features if determiners
adjoin to them.  Other types of NPs such as pronouns and proper nouns
have been argued by Abney \cite{Abney87} to either be determiners or
to move to the determiner position because they exhibit
determiner-like behavior. We can capture this insight in our system by
giving pronouns and proper nouns positive values for determiner
features. For example pronouns and proper nouns would be marked as
definite, a value that NPs containing common nouns can only obtain by
having a definite determiner adjoin. In addition to the determiner
features, nouns also have values for features such as reflexive
({\bf refl}), case, pronoun ({\bf pron}) and conjunction ({\bf conj}).

\begin{figure}[ht]
\centering
\begin{tabular}{c}

{\psfig{figure=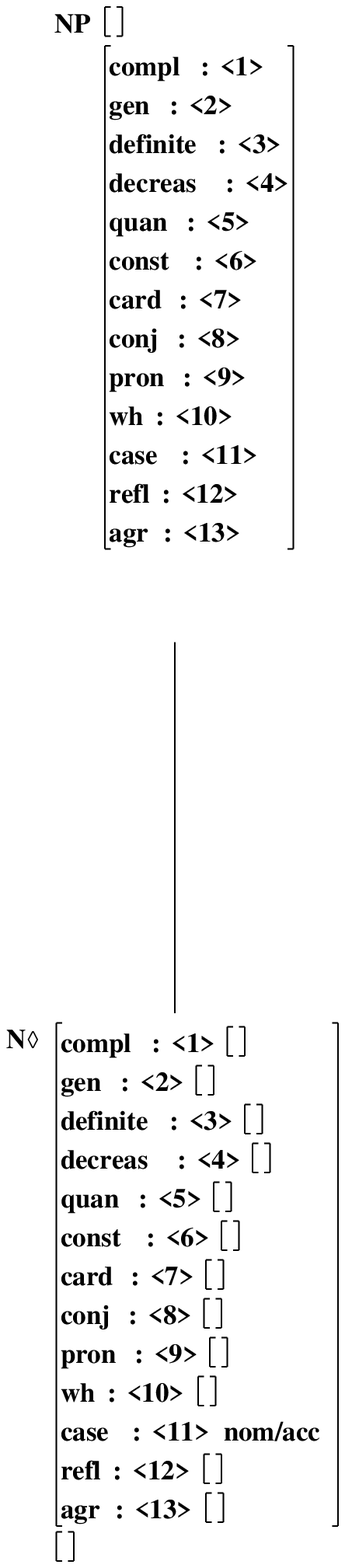,height=16.0cm}}\\
\end{tabular}

\caption{NP Tree}
\label{np-tree}

\end{figure}

  A single tree structure is selected by simple determiners, an
auxiliary tree which adjoins to NP. An example of this determiner tree
anchored by the determiner {\it these\/} is shown in
Figure~\ref{det-trees}. In addition to the determiner features the
tree in Figure~\ref{det-trees} has noun features such as {\bf case}
(see section 4.4.2), the {\bf conj} feature to control conjunction
(see Chapter \ref{conjunction}), {\bf rel-clause$-$} (see Chapter
\ref{rel_clauses}) and {\bf gerund$-$} (see Chapter
\ref{gerunds-chapter}) which prevent determiners from adjoining on top
of relative clauses and gerund NPs respectively, and the {\bf
displ-const} feature which is used to simulate multi-component
adjunction.

Complex determiners such as genitives and partitives also anchor tree
structures that adjoin to NP. They differ from the simple determiners
in their internal complexity. Details of our treatment of these more
complex constructions appear in Sections \ref{genitives} and
\ref{partitives}.  Sequences of determiners, as in the NPs {\it all
her dogs\/} or {\it those five dogs\/} are derived by multiple
adjunctions of the determiner tree, with each tree anchored by one of
the determiners in the sequence. The order in which the determiner
trees can adjoin is controlled by features.

\begin{figure}[ht]
\centering
\begin{tabular}{c}

{\psfig{figure=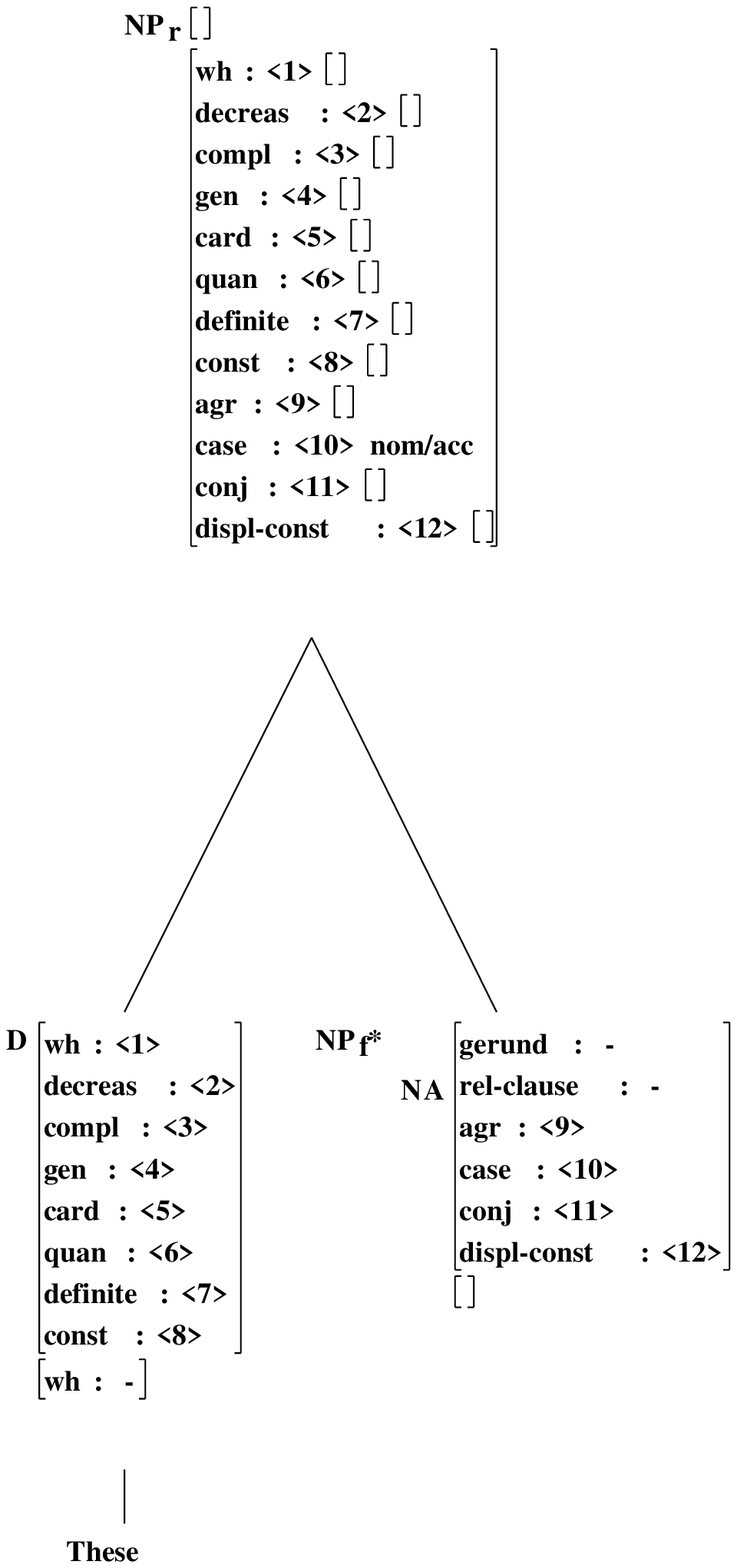,height=14cm}}
\end{tabular}
\caption{Determiner Trees with Features}
\label{det-trees}

\end{figure}

This treatment of determiners as adjoining onto NPs is similar to that
of \cite{Abeille90:TAG}, and allows us to capture one of the insights of the DP
hypothesis, namely that determiners select NPs as complements. In
Figure~\ref{det-trees} the determiner and its NP complement appear in
the configuration that is typically used in LTAG to represent
selectional relationships. That is, the head serves as the anchor of
the tree and it's complement is a sister node in the same elementary tree.

The XTAG treatment of determiners uses nine features for representing
their properties: definiteness ({\bf definite}), quantity
({\bf quan}), cardinality ({\bf card}), genitive ({\bf gen}), 
decreasing ({\bf decreas}), constancy ({\bf const}), {\bf wh}, agreement ({\bf
agr}), and complement ({\bf compl}). Seven of these
features were developed by semanticists for their accounts of semantic
phenomena (\cite{KeenanStavi86:LP}, \cite{BarwiseCooper81:LP},
\cite{Partee90:BK}), another was developed for a semantic
account of determiner negation by one of the authors of this
determiner analysis (\cite{Mateyak97}), and the last is the familiar
agreement feature. When used together these features also account for
a substantial portion of the complex patterns of English determiner
sequencing. Although we do not claim to have exhaustively covered the
sequencing of determiners in English, we do cover a large subset, both
in terms of the phenomena handled and in terms of corpus coverage. The
XTAG grammar has also been extended to include complex determiner
constructions such as genitives and partitives using these determiner
features.

Each determiner carries with it a set of values for these features
that represents its own properties, and a set of values for the
properties of NPs to which can adjoin. The features are crucial to
ordering determiners correctly. The semantic definitions
underlying the features are given below.

\begin{description}

\item[Definiteness:] Possible Values [+/--]. \\
A function f is definite iff f is non-trivial and whenever
f(s)~$\neq~\emptyset$ then it is always the intersection of one or
more individuals.  \cite{KeenanStavi86:LP}

\item[Quantity:]  Possible Values [+/--]. \\
If A and B are sets denoting an NP and associated predicate, respectively; E is
a domain in a model M, and F is a bijection from M$_{1}$ to M$_{2}$, then we
say that a determiner satisfies the constraint of quantity if
Det$_{E_{1}}$AB~$\leftrightarrow$~Det$_{E_{2}}$F(A)F(B). \cite{Partee90:BK}

\item[Cardinality:]  Possible Values [+/--]. \\
A determiner D is cardinal iff D $\in$ cardinal numbers~$\geq$~1.

\item[Genitive:]  Possible Values [+/--]. \\
Possessive pronouns and the possessive morpheme ({\it 's}) are marked {\bf
gen$+$}; all other nouns are {\bf gen$-$}.

\item[Decreasing:]  Possible Values [+/--]. \\
A set of Q properties is decreasing iff whenever s$\leq$t and t$\in$Q then
s$\in$Q. A function f is decreasing iff for all properties f(s) is a decreasing
set.

A non-trivial NP (one with a Det) is decreasing iff its denotation in any model
is decreasing. \cite{KeenanStavi86:LP}

\item[Constancy:] Possible Values [+/--]. \\
If A and B are sets denoting an NP and associated predicate, respectively, and
E is a domain, then we say that a determiner displays constancy if
(A$\cup$B)~$\subseteq$~E~$\subseteq$~E$^{\prime}$ then
Det$_{E}$AB~$\leftrightarrow$~Det$_{E^{\prime}}$AB. Modified from
\cite{Partee90:BK}

\item[Complement:] Possible Values [+/--]. \\
A determiner Q is positive complement if and only if for every set X, there
exists a continuous set of possible values for the size of the negated
determined set, NOT(QX), and the cardinality of QX is the only aspect of QX
that can be negated. (adapted from \cite{Mateyak97})

\end{description}

The {\bf wh}-feature has been discussed in the linguistics literature mainly in relation to wh-movement and with respect to NPs and nouns as well as determiners. We give a shallow but useful working definition of the {\bf wh}-feature below:

\begin{description}

\item[Wh:]  Possible Values [+/--]. \\
Interrogative determiners are {\bf wh$+$}; all other determiners are
{\bf wh$-$}. 
\end{description}

The {\bf agr} feature is inherently a noun feature.  While determiners
are not morphologically marked for agreement in English many of them
are sensitive to number.  Many determiners are semantically either
singular or plural and must adjoin to nouns which are the same. For
example, {\it a\/} can only adjoin to singular nouns ({\it a dog\/} vs
{\it $\ast$a dogs\/} while {\it many\/} must have plurals ({\it many
dogs\/} vs {\it $\ast$many dog\/}). Other determiners such as {\it some} are
unspecified for agreement in our analysis because they are compatible
with either singulars or plurals ({\it some dog}, {\it some
dogs}). The possible values of agreement for determiners are: [3sg, 3pl, 3].

The determiner tree in Figure~\ref{det-trees} shows the appropriate
feature values for the determiner {\it these}, while Table \ref{det-values}
shows the corresponding feature values of several other common determiners.

\begin{table}
\centering
\begin{tabular}{|l||c|c|c|c|c|c|c|c|c|}
\hline
Det&definite&quan&card&gen&wh&decreas&const&agr&compl\\
\hline
\hline
all&$-$&$+$&$-$&$-$&$-$&$-$&$+$&3pl&$+$\\
both&$+$&$-$&$-$&$-$&$-$&$-$&$+$&3pl&$+$\\
this&$+$&$-$&$-$&$-$&$-$&$-$&$+$&3sg&$-$\\
these&$+$&$-$&$-$&$-$&$-$&$-$&$+$&3pl&$-$\\
that&$+$&$-$&$-$&$-$&$-$&$-$&$+$&3sg&$-$\\
those&$+$&$-$&$-$&$-$&$-$&$-$&$+$&3pl&$-$\\
what&$-$&$-$&$-$&$-$&$+$&$-$&$+$&3&$-$\\
whatever&$-$&$-$&$-$&$-$&$-$&$-$&$+$&3&$-$\\
which&$-$&$-$&$-$&$-$&$+$&$-$&$+$&3&$-$\\
whichever&$-$&$-$&$-$&$-$&$-$&$-$&$+$&3&$-$\\
the&$+$&$-$&$-$&$-$&$-$&$-$&$+$&3&$-$\\
each&$-$&$+$&$-$&$-$&$-$&$-$&$+$&3sg&$-$\\
every&$-$&$+$&$-$&$-$&$-$&$-$&$+$&3sg&$+$\\
a/an&$-$&$+$&$-$&$-$&$-$&$-$&$+$&3sg&$+$\\
some$_{1}$&$-$&$+$&$-$&$-$&$-$&$-$&$+$&3&$-$\\
some$_{2}$&$-$&$+$&$-$&$-$&$-$&$-$&$-$&3pl&$-$\\
any&$-$&$+$&$-$&$-$&$-$&$-$&$+$&3sg&$+$\\
another&$-$&$+$&$-$&$-$&$-$&$-$&$+$&3sg&$+$\\
few&$-$&$+$&$-$&$-$&$-$&$+$&$-$&3pl&$-$\\
a few&$-$&$+$&$-$&$-$&$-$&$-$&$+$&3pl&$-$\\
many&$-$&$+$&$-$&$-$&$-$&$-$&$-$&3pl&$+$\\
many a/an&$-$&$+$&$-$&$-$&$-$&$-$&$-$&3sg&$+$\\
several&$-$&$+$&$-$&$-$&$-$&$-$&$+$&3pl&$-$\\
various&$-$&$-$&$-$&$-$&$-$&$-$&$+$&3pl&$-$\\
sundry&$-$&$-$&$-$&$-$&$-$&$-$&$+$&3pl&$-$\\
no&$-$&$+$&$-$&$-$&$-$&$+$&$+$&3&$-$\\
neither&$-$&$-$&$-$&$-$&$-$&$+$&$+$&3&$-$\\
either&$-$&$-$&$-$&$-$&$-$&$-$&$+$&3&$-$\\
\hline
\hline
GENITIVE&$+$&$-$&$-$&$+$&$-$&$-$&$+$&UN\footnotemark&$-$\\
CARDINAL&$-$&$+$&$+$&$-$&$-$&$-$&$+$&3pl\footnotemark\ &$-$\footnotemark\ \\
PARTITIVE&$-$&+/-\footnotemark\ &$-$&$-$&$-$&$-$&$+$&UN&+/-\\
\hline
\end{tabular}
 \caption{Determiner Features associated with D anchors}
\label{det-values}
\end{table}
\addtocounter{footnote}{-3}
\footnotetext{We use the symbol UN to represent the fact that the selectional
restrictions for a given feature are unspecified, meaning the noun phrase that
the determiner selects can be either positive or negative for this
feature.}
\stepcounter{footnote}
\footnotetext{Except {\it one} which is 3sg.}
\stepcounter{footnote}
\footnotetext{Except {\it one} which is {\bf compl+}.}
\stepcounter{footnote}
\footnotetext{A partitive can be either {\bf quan+} or {\bf quan-}, depending
upon the nature of the noun that anchors the partitive.  If the anchor noun is
modified, then the quantity feature is determined by the modifier's quantity
value.}

In addition to the features that represent their own properties, determiners
also have features to represent the selectional
restrictions they impose on the NPs they take as complements.  The
selectional restriction features of a determiner appear on the NP footnode of
the auxiliary tree that the determiner anchors.  The NP$_{f}$ node in Figure~\ref{det-trees} shows the selectional feature
restriction imposed by {\it these}\footnote{In addition to this tree, {\it
these} would also anchor another auxiliary tree that adjoins onto {\bf card+}
determiners.}, while Table~\ref{det-ordering} shows the corresponding
selectional feature restrictions of several other determiners.
\small
\begin{table}
\centering
\begin{tabular}{|l||c|c|c|c|c|c|c|c|c||l|}
\hline
Det&defin&quan&card&gen&wh&decreas&const&agr&compl&{\it e.g.}\\
\hline
\hline
&$-$&$-$&$-$&$-$&$-$&$-$&$-$&3pl&$-$&{\it dogs}\\
all&$+$&$-$&$-$&UN&$-$&UN&UN&3pl&$-$&{\it these dogs}\\
&UN&UN&$+$&UN&UN&UN&UN&3pl&UN&{\it five dogs}\\
\hline
&$-$&$-$&$-$&$-$&$-$&$-$&$-$&3pl&$-$&{\it dogs}\\
{both}&$+$&$-$&$-$&UN&$-$&UN&UN&3pl&$-$&{\it these dogs}\\
\hline
&$-$&$-$&$-$&$-$&$-$&$-$&$-$&3sg&$-$&{\it dog}\\
&$-$&$+$&UN&UN&$-$&$+$&$-$&3&UN&{\it few dogs}\\
{this/that}&$-$&$+$&UN&UN&$-$&$-$&$-$&3pl&$+$&{\it many dogs}\\
&UN&UN&$+$&UN&UN&UN&UN&3sg&UN&{\it five dogs}\\
\hline
&$-$&$-$&$-$&$-$&$-$&$-$&$-$&3pl&$-$&{\it dogs}\\
these/those&$-$&$+$&UN&UN&$-$&$+$&$-$&3pl&UN&{\it few dogs}\\
&UN&UN&$+$&UN&UN&UN&UN&3pl&UN&{\it five dogs}\\
\hline
what/which&$-$&$-$&$-$&$-$&$-$&$-$&$-$&3&$-$&{\it dog(s)}\\
whatever&$-$&$+$&UN&UN&$-$&$+$&$-$&3&UN&{\it few dogs}\\
whichever&UN&UN&$+$&UN&UN&UN&UN&3&UN&{\it many dogs}\\
\hline
&$-$&$-$&$-$&$-$&$-$&$-$&$-$&3&$-$&{\it dog(s)}\\
the&$-$&$+$&UN&UN&$-$&$+$&$-$&3&UN&{\it few dogs}\\
&$+$&$-$&$-$&$-$&$-$&$-$&$-$&UN&$-$&{\it the me}\\
&$-$&$+$&UN&UN&$-$&$-$&$-$&3pl&$+$&{\it many dogs}\\
&UN&UN&$+$&UN&UN&UN&UN&3&UN&{\it five dogs}\\
\hline
&$-$&$-$&$-$&$-$&$-$&$-$&$-$&3sg&$-$&{\it dog}\\
every/each&$-$&$+$&UN&UN&$-$&$+$&$-$&3&UN&{\it few dogs}\\
&UN&UN&$+$&UN&UN&UN&UN&3&UN&{\it five dogs}\\
\hline
a/an&$-$&$-$&$-$&$-$&$-$&$-$&$-$&3sg&$-$&{\it dog}\\
\hline
some$_{1,2}$&$-$&$-$&$-$&$-$&$-$&$-$&$-$&3&$-$&{\it dog(s)}\\
some$_{1}$&UN&UN&$+$&UN&UN&UN&UN&3pl&UN&{\it dogs}\\
\hline
&$-$&$-$&$-$&$-$&$-$&$-$&$-$&3sg&$-$&{\it dog}\\
any&$-$&$+$&UN&UN&$-$&$+$&$-$&3&UN&{\it few dogs}\\
&UN&UN&$+$&UN&UN&UN&UN&3&UN&{\it five dogs}\\
\hline
&$-$&$-$&$-$&$-$&$-$&$-$&$-$&3sg&$-$&{\it dog}\\
another&$-$&$+$&UN&UN&$-$&$+$&$-$&3&UN&{\it few dogs}\\
&UN&UN&$+$&UN&UN&UN&UN&3&UN&{\it five dogs}\\
\hline
few&$-$&$-$&$-$&$-$&$-$&$-$&$-$&3pl&$-$&{\it dogs}\\
\hline
a few&$-$&$-$&$-$&$-$&$-$&$-$&$-$&3pl&$-$&{\it dogs}\\
\hline
many&$-$&$-$&$-$&$-$&$-$&$-$&$-$&3pl&$-$&{\it dogs}\\
\hline
many a/an&$-$&$-$&$-$&$-$&$-$&$-$&$-$&3sg&$-$&{\it dog}\\
\hline
several&$-$&$-$&$-$&$-$&$-$&$-$&$-$&3pl&$-$&{\it dogs}\\
\hline
various&$-$&$-$&$-$&$-$&$-$&$-$&$-$&3pl&$-$&{\it dogs}\\
\hline
sundry&$-$&$-$&$-$&$-$&$-$&$-$&$-$&3pl&$-$&{\it dogs}\\
\hline
no&$-$&$-$&$-$&$-$&$-$&$-$&$-$&3&$-$&{\it dog(s)}\\
\hline
neither&$-$&$-$&$-$&$-$&$-$&$-$&$-$&3sg&$-$&{\it dog}\\
\hline
either&$-$&$-$&$-$&$-$&$-$&$-$&$-$&3sg&$-$&{\it dog}\\
\hline
\end{tabular}
\caption{Selectional Restrictions Imposed by Determiners on the NP
foot node}
\label{det-ordering}
\end{table}

\begin{table}[htb]
\centering
\begin{tabular}{|l||c|c|c|c|c|c|c|c|c|}
\hline\hline
Det&definite&quan&card&gen&wh&decreas&const&agr&compl\\
\hline
\hline
&$-$&$-$&$-$&$-$&$-$&$-$&$-$&3&$-$\\
&$-$&$+$&UN&UN&$-$&$+$&$-$&3&UN\\
GENITIVE&$-$&$+$&UN&UN&$-$&$-$&$-$&3pl&$+$\\
&UN&UN&$+$&UN&UN&UN&UN&3&UN\\
&$-$&$+$&$-$&$-$&$-$&$-$&$+$&3pl&$-$\\
&$-$&$-$&$-$&$-$&$-$&$-$&$+$&3pl&$-$\\
\hline
CARDINAL&$-$&$-$&$-$&$-$&$-$&$-$&$-$&3pl\footnotemark&$-$\\
\hline
PARTITIVE&UN&UN&UN&UN&$-$&UN&UN&UN&UN\\
\hline
\end{tabular}
\caption{Selectional Restrictions Imposed by Groups of
Determiners/Determiner Constructions}
\label{det-ordering2}
\end{table}
\footnotetext{{\it one} differs from the rest of CARD in selecting
singular nouns}

\normalsize

\section{The Wh-Feature}
\label{agr-section}
A determiner with a {\bf wh+} feature is always the left-most
determiner in linear order since no determiners have selectional
restrictions that allow them to adjoin onto an NP with a +wh feature
value.  The presence of a wh+ determiner makes the entire NP wh+, and
this is correctly represented by the coindexation of the determiner
and root NP nodes' values for the wh-feature. Wh+ determiners'
selectional restrictions on the NP foot node of their tree only allows them
adjoin onto NPs that are {\bf wh-} or unspecified for the
wh-feature. Therefore ungrammatical sequences such as {\it $\ast$which what
dog} are impossible.  The adjunction of {\bf wh +} determiners onto
{\bf wh+} pronouns is also prevented by the same mechanism.

\section{Multi-word Determiners}
The system recognizes the multi-word determiners {\it a few} and {\it many a}.
The features for a multi-word determiner are located on the parent node of its 
two components (see Figure~\ref{multi-det-tree}).  We chose to represent these 
determiners as multi-word constituents because neither determiner retains the 
same set of features as either of its parts.  For example, the determiner 
{\it a} is 3sg and {\it few} is decreasing, while {\it a few} is 3pl and 
increasing.  Additionally, {\it many} is 3pl and {\it a} displays constancy, 
but {\it many a} is 3sg and does not display constancy.  Example sentences 
appear in (\ex{1})-(\ex{2}).

\begin{itemize}
\item{Multi-word Determiners}
\enumsentence{{\bf a few} teaspoons of sugar should be adequate .}
\enumsentence{{\bf many a} man has attempted that stunt, but none have
succeeded .}

\end{itemize}

\begin{figure}[htb]
\centering
\begin{tabular}{cc}
{\psfig{figure=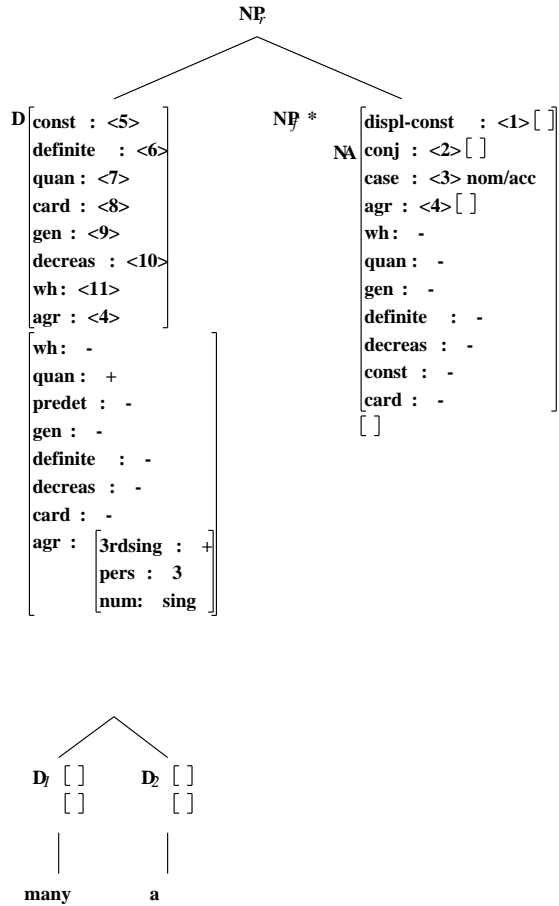,height=5.0in}}
\end{tabular}\\
\caption{Multi-word Determiner tree:  $\beta$DDnx}
\label{multi-det-tree}
\end{figure}

\section{Genitive Constructions}
\label{genitives}        

There are two kinds of genitive constructions: genitive pronouns, and genitive
NP's (which have an explicit genitive marker, {\it 's}, associated with them).
It is clear from examples such as {\it her dog returned home\/} and
{\it her five dogs returned home} vs {\it
$\ast$dog returned home\/} that genitive pronouns function as determiners and as
such, they sequence with the rest of the determiners.  The features for the
genitives are the same as for other determiners.  Genitives are not required to agree with
either the determiners or the nouns in the NPs that they modify. The
value of the {\bf agr} feature for an NP with a genitive determiner
depends on the NP to which the genitive determiner adjoins. While it
might seem to make sense to take {\it their} as 3pl, {\it my} as 1sg,
and {\it Alfonso's} as 3sg, this number and person information only
effects the genitive NP itself and bears no relationship to the number
and person of the NPs with these items as determiners. Consequently,
we have represented {\bf agr} as unspecified for genitives in Table
\ref{det-values}.

Genitive NP's are particularly interesting because they are potentially
recursive structures.  Complex NP's can easily be embedded within a determiner.

\enumsentence{[[[John]'s friend from high school]'s uncle]'s mother came to town.}

There are two things to note in the above example.  One is that in embedded
NPs, the genitive morpheme comes at the end of the NP phrase, even if the head
of the NP is at the beginning of the phrase.  The other is that the determiner
of an embedded NP can also be a genitive NP, hence the possibility of recursive
structures.

In the XTAG grammar, the genitive marker, {\it 's}, is separated from the
lexical item that it is attached to and given its own category (G).  In
this way, we can allow the full complexity of NP's to come from the
existing NP system, including any recursive structures.  As with the simple
determiners, there is one auxiliary tree structure for genitives which
adjoins to NPs. As can be seen in \ref{gen-trees}, this tree is anchored by
the genitive marker {\it 's} and has a branching D node which accomodates
the additional internal structure of genitive determiners. Also, like simple
determiners, there is one initial tree structure
(Figure~\ref{subst-genNP-tree}) available for substitution where needed, as
in, for example, the Determiner Gerund NP tree (see
Chapter~\ref{gerunds-chapter} for discussion on determiners for gerund
NP's).

\begin{figure}[ht]
\centering
\begin{tabular}{c}

{\psfig{figure=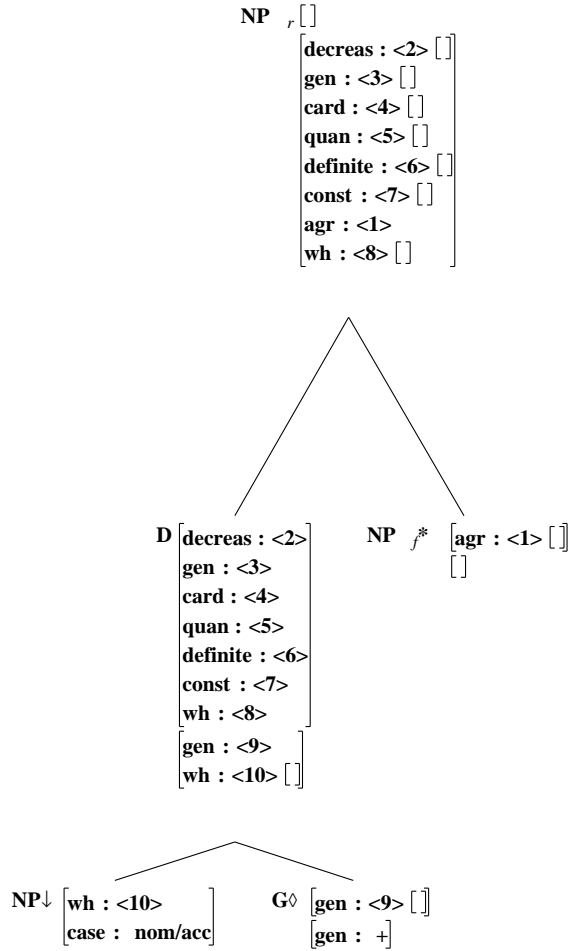,height=13.0cm}}\\
\end{tabular}
\caption{Genitive Determiner Tree}
\label{gen-trees}

\end{figure}

\begin{figure}[htb]
\centering
\begin{tabular}{c}
{\psfig{figure=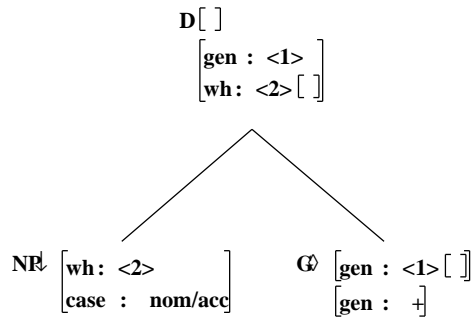,height=1.8in}}\\
\end{tabular}
\caption{Genitive NP tree for substitution: $\alpha$DnxG}
\label{subst-genNP-tree}
\end{figure}

Since the NP node which is sister
to the G node could also have a genitive determiner in it, the type of
genitive recursion shown in (\ex{0}) is quite naturally accounted for
by the genitive tree structure used in our analysis.

\section{Partitive Constructions}        
\label{partitives}                        

The deciding factor for including an analysis of partitive constructions(e.g.\ {\it some kind
of}, {\it all of\/}) as a complex determiner constructions was the
behavior of the agreement features.  If partitive constructions are analyzed as
an NP with an adjoined PP, then we would expect to get agreement with the head
of the NP (as in ({\ex{1}})).  If, on the other hand, we analyze them
as a determiner construction, then we would expect to get agreement with the
noun that the determiner sequence modifies (as we do in ({\ex{2}})).

\enumsentence{a {\it kind} [of these machines] {\it is} prone to failure.}
\enumsentence{[a kind of] these {\it machines are} prone to failure.}

In other words, for partitive constructions, the semantic head of the NP is the second rather than the first noun in linear order. That the agreement shown in ({\ex{0}}) is possible suggests that the second noun in linear order in these constructions should also be treated as the syntactic head. Note that both the partitive and PP readings are usually possible for a particular NP. In the cases where either the partitive or the PP reading is preferred, we take it to be just that, a preference, most appropriately modeled not in the grammar but in a component such as the heuristics used with the XTAG parser for reducing the analyses produced to the most likely. 

In our analysis the partitive tree in Figure~\ref{part-tree} is anchored
by one of a limited group of nouns that can appear in the determiner
portion of a partitive construction. A rough semantic characterization
of these nouns is that they either represent quantity (e.g. {\it part, half,
most, pot, cup, pound} etc.) or classification (e.g. {\it type, variety,
kind, version} etc.).  In the absence of a more implementable
characterization we use a list of such nouns compiled from a
descriptive grammar \cite{quirk85}, a thesaurus, and from online
corpora. In our grammar the nouns on the list are the only ones that
select the partitive determiner tree.

\begin{figure}[ht]
\centering
\begin{tabular}{c}

{\psfig{figure=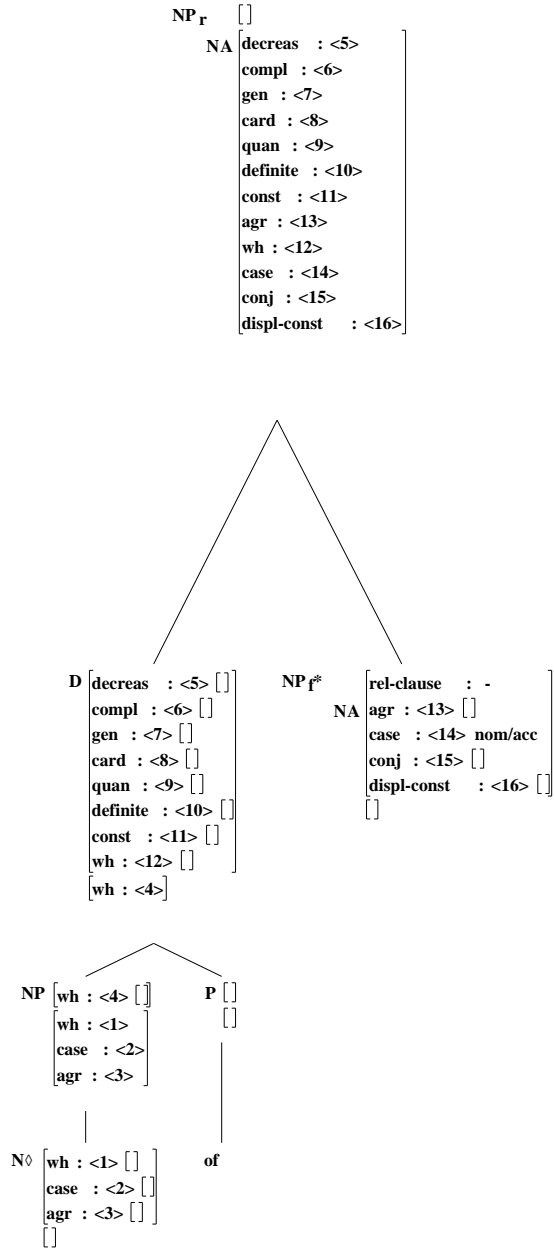,height=17.0cm}}\\
\end{tabular}
\caption{Partitive Determiner Tree}
\label{part-tree}

\end{figure}

Like other determiners, partitives can adjoin to an NP consisting of just
a noun ({\it `[a certain kind of] machine'}), or adjoin to NPs
that already have determiners ({\it `[some parts of] these
machines'}. Notice that just as for the genitives, the complexity and
the recursion are contained below the D node and rest of the structure
is the same as for simple determiners.

\section{Adverbs, Noun Phrases, and Determiners}
\label{adverbial-section}

Many adverbs interact with the noun phrase and determiner system in English.
For example, consider sentences (\ref{approx})-(\ref{double}) below.

\enumsentence{\label{approx}{\bf Approximately} thirty people came to the lecture.}
\enumsentence{\label{practically}{\bf Practically} every person in the theater was laughing
hysterically during that scene.}
\enumsentence{\label{only}{\bf Only} John's crazy mother can make stuffing that tastes so good.}
\enumsentence{\label{relatively}{\bf Relatively} few programmers remember how to program in
COBOL.}
\enumsentence{\label{not}{\bf Not} every martian would postulate that all humans speak a
universal language.}
\enumsentence{\label{enough}{\bf Enough} money was gathered to pay off the group gift.}
\enumsentence{\label{quite}{\bf Quite} a few burglaries occurred in that neighborhood last
year.}
\enumsentence{\label{double}I wanted to be paid {\bf double} the amount they offered.}

Although there is some debate in the literature as to whether these should be
classified as determiners or adverbs, we believe that these items that
interact with the NP and determiner system are in fact adverbs.   These items
exhibit a broader distribution than either determiners or adjectives in that
they can modify many other phrasal categories, including adjectives, verb
phrases, prepositional phrases, and other adverbs.

Using the determiner feature system, we can obtain a close approximation to an
accurate characterization of the behavior of the adverbs that interact with
noun phrases and determiners.  Adverbs can adjoin to either a determiner or a
noun phrase (see figure~\ref{det-adv-trees}), with the adverbs restricting what
types of NPs or determiners they can modify by imposing feature requirements on
the foot D or NP node.  For
example, the adverb {\it approximately}, seen in (\ref{approx})
above, selects for determiners that are {\bf card+}.  The adverb {\it enough}
in (\ref{enough}) is an example of an adverb that selects for a noun phrase,
specifically a noun phrase that is not modified by a determiner.

\begin{figure}[ht]
\centering
\begin{tabular}{ccc}
{\psfig{figure=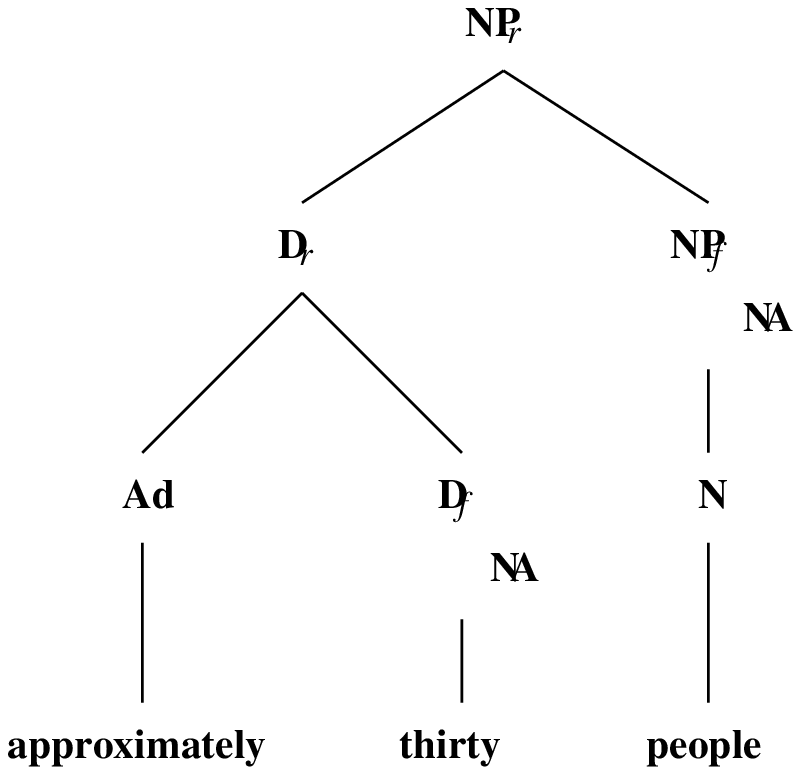,height=5.0cm}}&&
{\psfig{figure=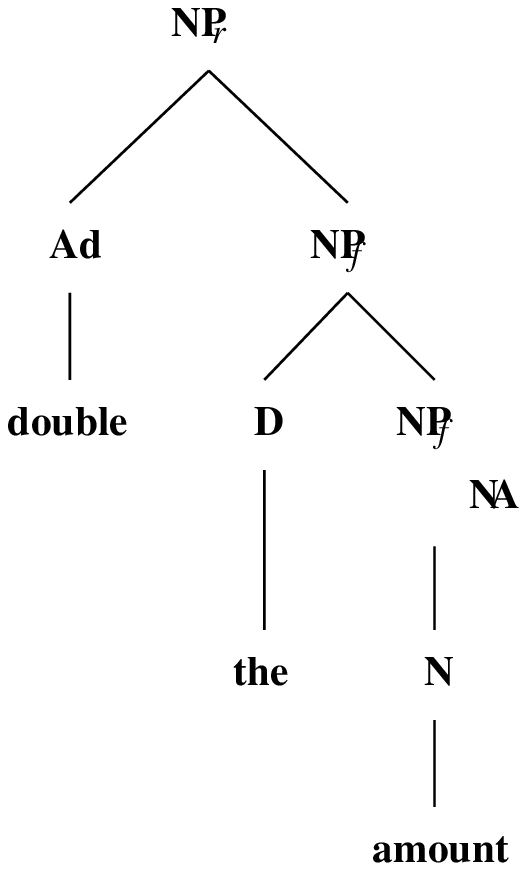,height=5.0cm}}\\
(a)&&(b)
\end{tabular}
\caption{(a) Adverb modifying a determiner; (b) Adverb modifying a noun phrase}
\label{det-adv-trees}

\end{figure}

Most of the adverbs that modify determiners and NPs divide into six classes,
with some minor variation within classes, based on the pattern of these
restrictions.  Three of the classes are adverbs that modify determiners, while
the other three modify NPs.

The largest of the five classes is the class of adverbs that modify cardinal
determiners.  This class includes, among others, the adverbs {\it about}, {\it
at most}, {\it exactly}, {\it nearly}, and {\it only}.  These adverbs have the
single restriction that they must adjoin to determiners that are {\bf card+}.
Another class of adverbs consists of those that can modify the determiners {\it
every}, {\it all}, {\it any}, and {\it no}.  The adverbs in this class are {\it
almost}, {\it nearly}, and {\it practically}.  Closely related to this class
are the adverbs {\it mostly} and {\it roughly}, which are restricted to
modifying {\it every} and {\it all}, and {\it hardly}, which can only modify
{\it any}.  To select for {\it every}, {\it all}, and {\it any}, these adverbs
select for determiners that are [{\bf quan+}, {\bf card-}, {\bf const+}, {\bf
compl+}], and to select for {\it no}, the adverbs choose a determiner that is
[{\bf quan+}, {\bf decreas+}, {\bf const+}].  The third class of adverbs that
modify determiners are those that modify the determiners {\it few} and {\it
many}, representable by the feature sequences [{\bf quan+}, {\bf decreas+},
{\bf const-}] and [{\bf quan+}, {\bf decreas-}, {\bf const-}, {\bf 3pl}, {\bf
compl+}], respectively.  Examples of these adverbs are {\it awfully}, {\it
fairly}, {\it relatively}, and {\it very}.

Of the three classes of adverbs that modify noun phrases, one actually consists
of a single adverb {\it not}, that only modifies determiners that are {\bf
compl+}.  Another class consists of the focus adverbs, {\it at least}, {\it
even}, {\it only}, and {\it just}.  These adverbs select NPs that are {\bf wh-}
and {\bf card-}.  For the NPs that are {\bf card+}, the focus adverbs actually
modify the cardinal determiner, and so these adverbs are also included in the
first class of adverbs mentioned in the previous paragraph.  The last major
class that modify NPs consist of the adverbs {\it double} and {\it twice},
which select NPs that are [{\bf definite+}] (i.e., {\it the}, {\it
this/that/those/these}, and the genitives).

Although these restrictions succeed in recognizing the correct
determiner/adverb sequences, a few unacceptable sequences slip through.  For
example, in handling the second class of adverbs mentioned above, {\it every},
{\it all}, and {\it any} share the features [{\bf quan+}, {\bf card-}, {\bf
const+}, {\bf compl+}] with {\it a} and {\it another}, and so {\it
$\ast$nearly a man} is acceptable in this system.  In addition to this
over-generation within a major class, the adverb {\it quite} selects for
determiners and NPs in what seems to be a purely idiosyncratic fashion.
Consider the following examples.

\eenumsentence{\label{quite2}\item[a.] {\bf Quite} a few members of the audience had to
leave.
	\item[b.] There were {\bf quite} many new participants at this
year's conference.
	\item[c.] {\bf Quite} few triple jumpers have jumped that far.
	\item[d.] Taking the day off was {\bf quite} the right thing to do.
	\item[e.] The recent negotiation fiasco is {\bf quite} another issue.
	\item[f.] Pandora is {\bf quite} a cat!}

In examples (\ref{quite2}a)-(\ref{quite2}c), {\it quite} modifies the determiner, while in (\ref{quite2}d)-(\ref{quite2}f),
{\it quite} modifies the entire noun phrase.  Clearly, it functions in a
different manner in the two sets of sentences; in (\ref{quite2}a)-(\ref{quite2}c), {\it quite}
intensifies the amount implied by the determiner, whereas in (\ref{quite2}d)-(\ref{quite2}f), it
singles out an individual from the larger set to which it belongs.  To capture
the selectional restrictions needed for (\ref{quite2}a)-(\ref{quite2}c), we utilize the two sets of
features mentioned previously for selecting {\it few} and {\it many}.  However,
{\it a few} cannot be singled out so easily; using the sequence [{\bf quan+},
{\bf card-}, {\bf decreas-}, {\bf const+}, {\bf 3pl}, {\bf compl-}], we also
accept the ungrammatical NPs {\it $\ast$quite several members} and {\it
$\ast$quite some members} (where {\it quite} modifies {\it some}).  In
selecting {\it the} as in (d) with the features [{\bf definite+}, {\bf gen-}, {\bf
3sg}], {\it quite} also selects {\it this} and {\it that}, which are
ungrammatical in this position.  Examples (\ref{quite2}e) and (\ref{quite2}f) present yet another
obstacle in that in selecting {\it another} and {\it a}, {\it quite}
erroneously selects {\it every} and {\it any}.

It may be that there is an undiscovered semantic feature that would
alleviate these difficulties.  However, on the whole, the determiner feature system we have proposed can be
used as a surprisingly efficient method of characterizing the interaction of
adverbs with determiners and noun phrases.

\chapter{Modifiers}
\label{modifiers}

This chapter covers various types of modifiers: adverbs, prepositions,
adjectives, and noun modifiers in noun-noun compounds.\footnote{Relative
clauses are discussed in Chapter~\ref{rel_clauses}.}  These categories
optionally modify other lexical items and phrases by adjoining onto them.  In
their modifier function these items are adjuncts; they are not part of the
subcategorization frame of the items they modify.  Examples of some of these
modifiers are shown in (\ex{1})-(\ex{3}).

\enumsentence{[$_{ADV}$ certainly $_{ADV}$], the October 13 sell-off
didn't settle any stomachs . (WSJ)}

\enumsentence{Mr. Bakes [$_{ADV}$ previously $_{ADV}$] had a turn at running
Continental . (WSJ)}

\enumsentence{most [$_{ADJ}$ foreign $_{ADJ}$] [$_{N}$ government
$_{N}$] [$_{N}$ bond $_{N}$] [prices] rose [$_{PP}$ during the week
$_{PP}$]. }

The trees used for the various modifiers are quite similar in form.
The modifier anchors the tree and the root and foot nodes of the tree
are of the category that the particular anchor modifies. Some
modifiers, e.g. prepositions, select for their own arguments and those
are also included in the tree.  The foot node may be to the right or
the left of the anchoring modifier (and its arguments) depending on
whether that modifier occurs before or after the category it
modifies. For example, almost all adjectives appear to the left of the
nouns they modify, while prepositions appear to the right when
modifying nouns.

\section{Adjectives}
\label{adj-modifier}

In addition to being modifiers, adjectives in the XTAG English grammar can be
also anchor clauses (see Adjective Small Clauses in
Chapter~\ref{small-clauses}).  There is also one tree family, Intransitive with
Adjective (Tnx0Vax1), that has an adjective as an argument and is used for
sentences such as {\it Seth felt happy}. In that tree family the adjective
substitutes into the tree rather than adjoining as is the case for modifiers.

\begin{figure}[htb]
\centering
\begin{tabular}{cc}
{\psfig{figure=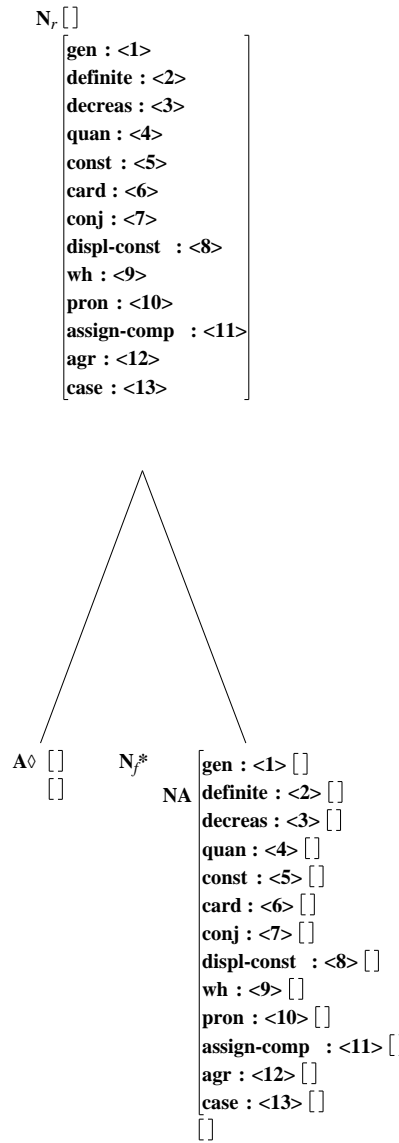,height=6.5in}}
\end{tabular}\\
\caption {Standard Tree for Adjective modifying a Noun: $\beta$An}
\label {An-tree}
\end{figure}

As modifiers, adjectives anchor the tree shown in
Figure~\ref{An-tree}.  The features of the N node onto which the
$\beta$An tree adjoins are passed through to the top node of the
resulting N.  The null adjunction marker (NA) on the N foot node
imposes right binary branching such that each subsequent adjective
must adjoin on top of the leftmost adjective that has already
adjoined.  Due to the NA constraint, a sequence of adjectives will
have only one derivation in the XTAG grammar. The adjective's
morphological features such as superlative or comparative are
instantiated by the morphological analyzer. See
Chapter~\ref{compars-chapter} for a description of how we handle
comparatives.  At this point, the treatment of adjectives in the XTAG
English grammar does not include selectional or ordering
restrictions. Consequently, any adjective can adjoin onto any noun and
on top of any other adjective already modifying a noun. All of the
modified noun phrases shown in (\ex{1})-(\ex{4}) currently parse with
the same structure shown for {\it colorless green ideas\/} in Figure
\ref{colorless-green-adj}.

\enumsentence{big green bugs}
\enumsentence{big green ideas}
\enumsentence{colorless green ideas}
\enumsentence{?green big ideas}

\begin{figure}[htb]
\centering
\begin{tabular}{cc}
{\psfig{figure=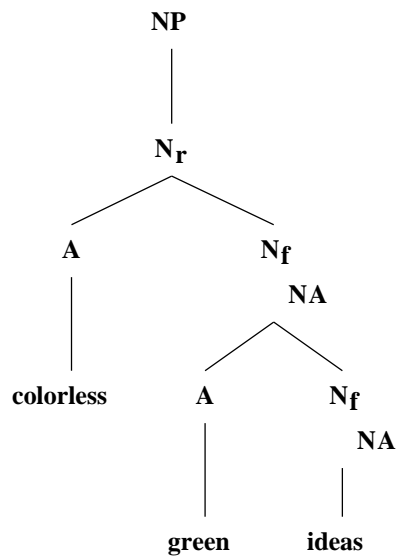,height=3in}}
\end{tabular}\\
\caption {Multiple adjectives modifying a noun}
\label {colorless-green-adj}
\end{figure}

While (\ex{-2})-(\ex{0}) are all semantically anomalous, (\ex{0}) also
suffers from an ordering problem that makes it seem ungrammatical in
the absence of any licensing context. One could argue that the grammar
should accept (\ex{-3})-(\ex{-1}) but not (\ex{0}).  One of the future
goals for the grammar is to develop a treatment of adjective ordering
similar to that developed by
\cite{ircs:det98} for determiners\footnote{See
Chapter~\ref{det-comparitives} or \cite{ircs:det98} for details of the
determiner analysis.}. An adequate implementation of ordering
restrictions for adjectives would rule out (\ex{0}).

\section{Noun-Noun Modifiers}
\label{noun-modifier}

Noun-noun compounding in the English XTAG grammar is very similar to
adjective-noun modification.  The noun modifier tree, shown in
Figure~\ref{noun-compound-tree}, has essentially the same structure as the
adjective modifier tree in Figure~\ref{An-tree}, except for the syntactic
category label of the anchor.  

\begin{figure}[htb]
\centering
\begin{tabular}{c}
{\psfig{figure=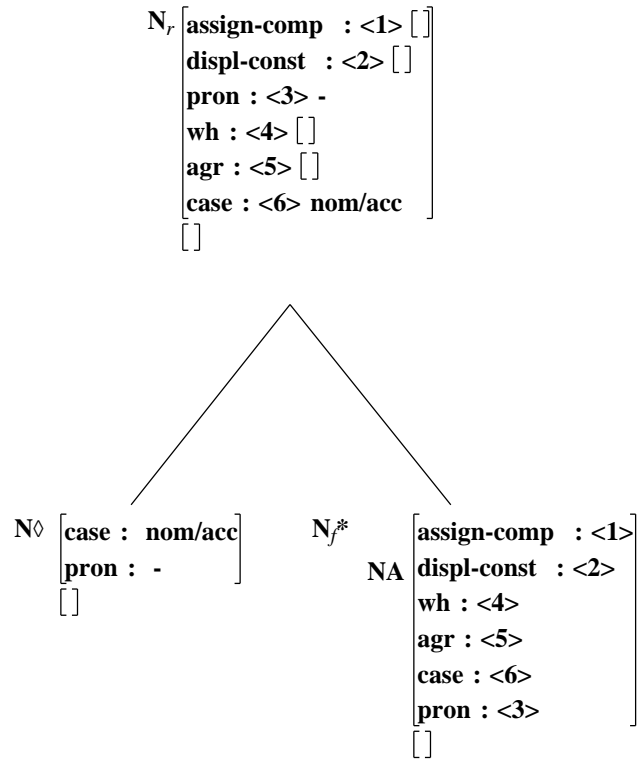,height=4.5in}}
\end{tabular}
\caption {Noun-noun compounding tree: $\beta$Nn (not all features displayed)}
\label {noun-compound-tree}
\end{figure}

Noun compounds have a variety of scope possibilities not available to
adjectives, as illustrated by the single bracketing possibility in (\ex{1}) and
the two possibilities for (\ex{2}).  This ambiguity is manifested in the XTAG
grammar by the two possible adjunction sites in the noun-noun compound tree
itself.  Subsequent modifying nouns can adjoin either onto the N$_r$ node or
onto the N anchor node of that tree, which results in exactly the two
bracketing possibilities shown in (\ex{2}).  This inherent structural ambiguity
results in noun-noun compounds regularly having multiple derivations. However,
the multiple derivations are not a defect in the grammar because they are
necessary to correctly represent the genuine ambiguity of these phrases.

\enumsentence{[$_{N}$ big [$_{N}$ green design $_{N}$]$_{N}$]}

\enumsentence{[$_{N}$ computer [$_{N}$ furniture design $_{N}$]$_{N}$]\\
\/~~[$_{N}$ [$_{N}$ computer furniture $_{N}$] design $_{N}$]}

Noun-noun compounds have no restriction on number.  XTAG allows nouns to be either singular or plural as in (\ex{1})-(\ex{3}).
\enumsentence{Hyun is taking an algorithms course .}
\enumsentence{waffles are in the frozen foods section .}
\enumsentence{I enjoy the dog shows .}

\section{Time Noun Phrases}
\label{timenps}

Although in general NPs cannot modify clauses or other NPs, there is a
class of NPs, with meanings that relate to time, that can do
so.\footnote{ There may be other classes of NPs, such as directional
phrases, such as {\em north, south} etc., which behave similarly. We
have not yet analyzed these phrases.} We call this class of NPs
``Time~NPs''.  Time~NPs behave essentially like PPs. Like PPs,
time~NPs can adjoin at four places: to the right of an NP, to the
right and left of a VP, and to the left of an~S.

Time~NPs may include determiners, as in {\em this month} in example
(\ex{1}), or may be single lexical items as in {\em today} in example
(\ex{2}).  Like other NPs, time~NPs can also include adjectives, as in
example (\ex{6}).

\enumsentence{Elvis left the building \underline{this week}}
\enumsentence{Elvis left the building \underline{today}}
\enumsentence{It has no bearing on our work force \underline{today} (WSJ)}
\enumsentence{The fire \underline{yesterday} claimed two lives}

\enumsentence{\underline{Today} it has no bearing on our work force}
\enumsentence{Michael \underline{late yesterday} announced a buy-back program}

The XTAG analysis for time~NPs is fairly simple, requiring only the
creation of proper NP auxiliary trees.  Only nouns that can be part of
time~NPs will select the relevant auxiliary trees, and so only these
type of NPs will behave like PPs under the XTAG analysis.
Currently, about 60 words select Time~NP trees, but since these
words can form NPs that include determiners and adjectives, a large
number of phrases are covered by this class of modifiers.

Corresponding to the four positions listed above, time~NPs
can select one of the four trees shown in Figure~\ref{timenp-trees}.

\begin{figure}[htb]
\centering
\begin{tabular}{ccccccc}
{\psfig{figure=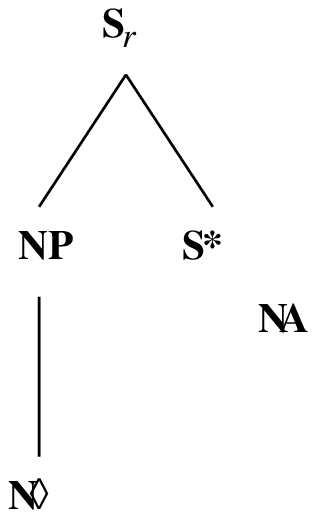,height=1.5in}}
& \hspace{.5in} &
{\psfig{figure=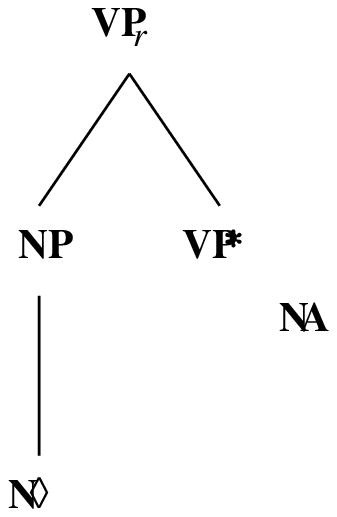,height=1.5in}}
&  \hspace{.5in} &
{\psfig{figure=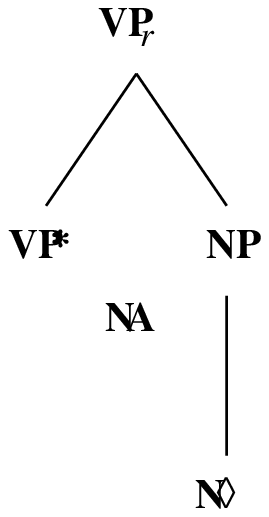,height=1.5in}}
&  \hspace{.5in} &
{\psfig{figure=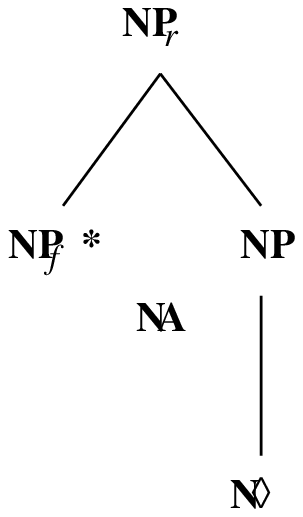,height=1.5in}}\\
$\beta$Ns&&$\beta$Nvx&&$\beta$vxN&&$\beta$nxN\\
\end{tabular}
\caption{Time Phrase Modifier trees: $\beta$Ns, $\beta$Nvx, $\beta$vxN, $\beta$nxN}
\label{timenp-trees}
\end{figure}

Determiners can be added to time~NPs by adjunction in
the same way that they are added to NPs in other
positions. The trees in Figure~\ref{everymonth} show that the
structures of examples (\ex{-5}) and (\ex{-4}) differ only in the
adjunction of  {\em this} to the time~NP in example (\ex{-5}).

\begin{figure}[htb] 
\centering 
\begin{tabular}{ccc}
\psfig{figure=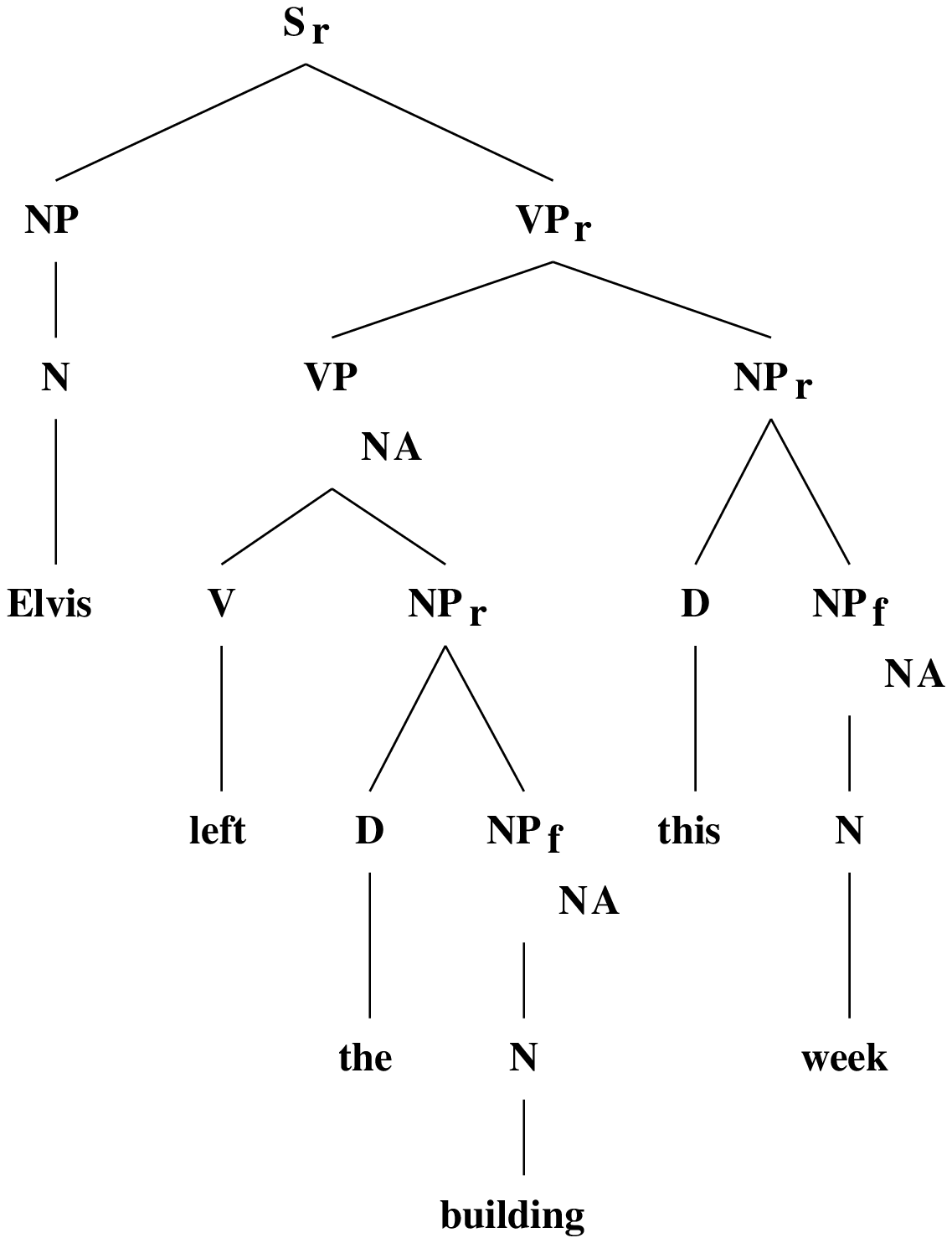,height=3.5in}
& \hspace{.5in} &
\psfig{figure=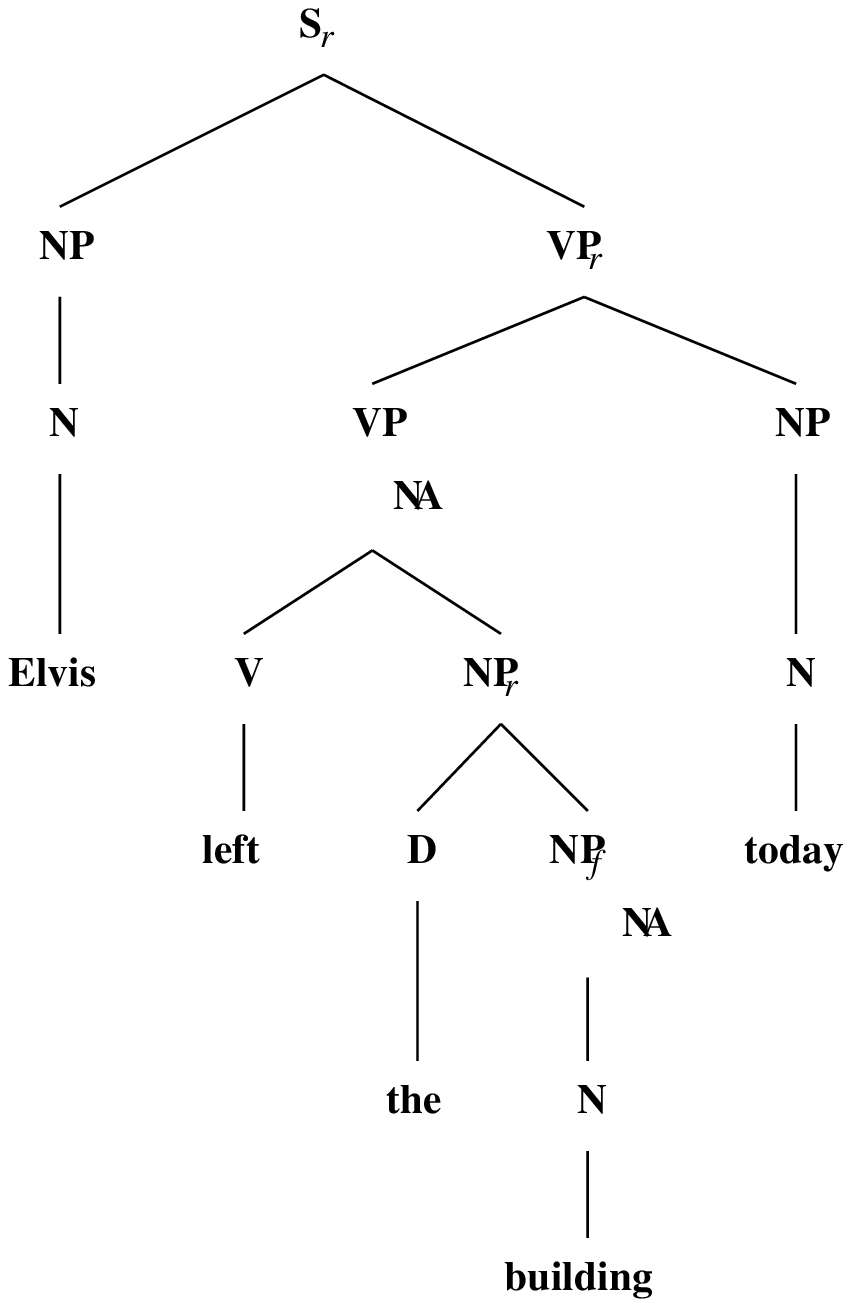,height=3.5in} \\
\end{tabular}\\
\caption{Time~NPs with and without a determiner} 
\label{everymonth}
\end{figure}

The sentence 
\enumsentence{Esso said the Whiting field started production Tuesday (WSJ)} 
has (at least) two different interpretations, depending on whether
{\em Tuesday} attaches to {\em said} or to {\em started}. 
Valid time~NP analyses are available for both these interpretations and 
are shown in Figure~\ref{esso}.

\begin{figure}[htb] \centering \begin{tabular}{ccc}
{\psfig{figure=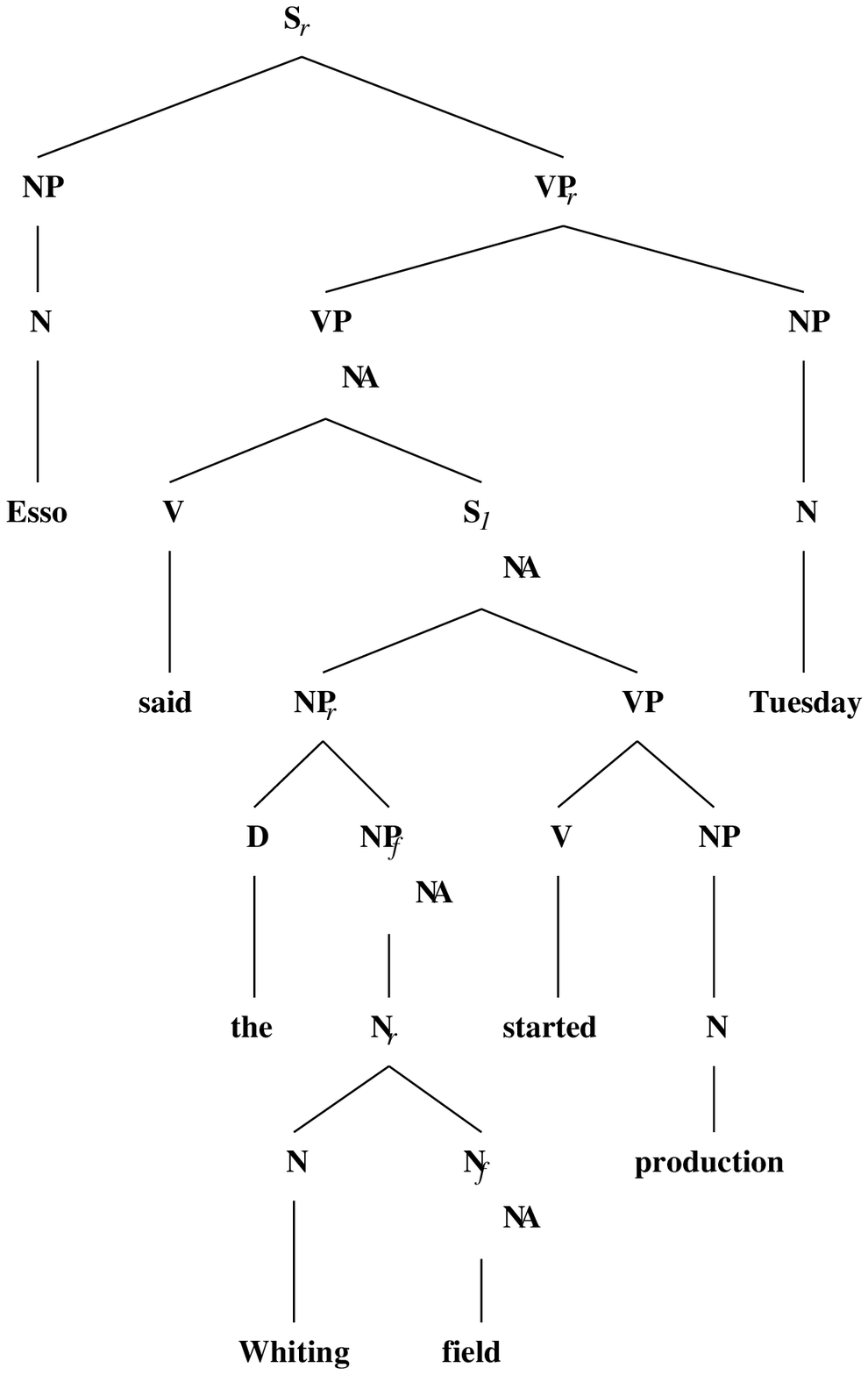,height=3.5in}} & \hspace{.5in} &
{\psfig{figure=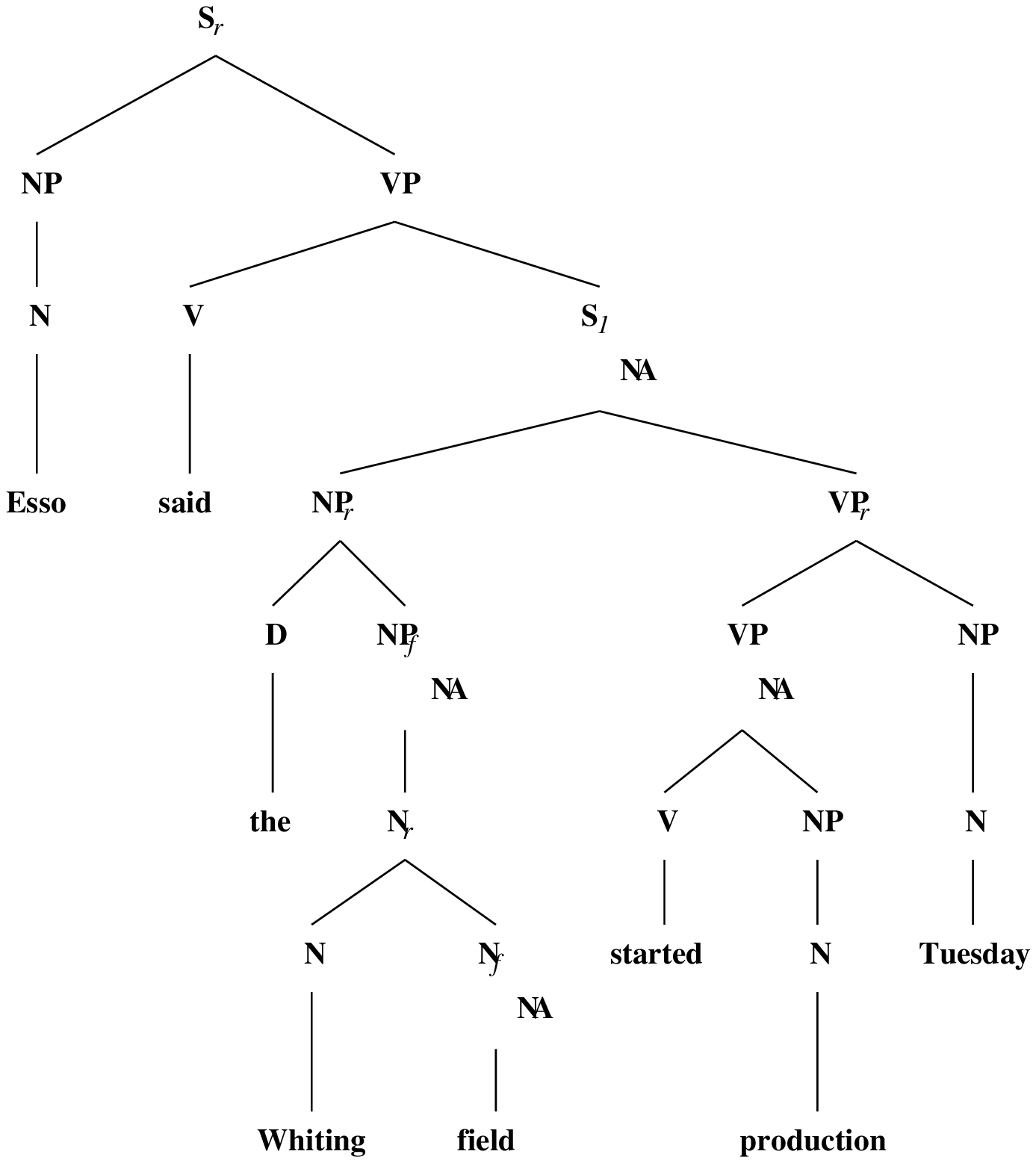,height=3.5in}} \\ \end{tabular}\\
\caption {Time~NP trees: Two different attachments} \label{esso}
\end{figure}

Derived tree structures for examples (\ex{-4}) -- (\ex{-1}), which
show the four possible time~NP positions are shown in
Figures~\ref{bearingtrees} and \ref{lateyesterday}.  The derivation
tree for example (\ex{-1}) is also shown in
Figure~\ref{lateyesterday}.

\begin{figure}[htb] \centering \begin{tabular}{ccc}
{\psfig{figure=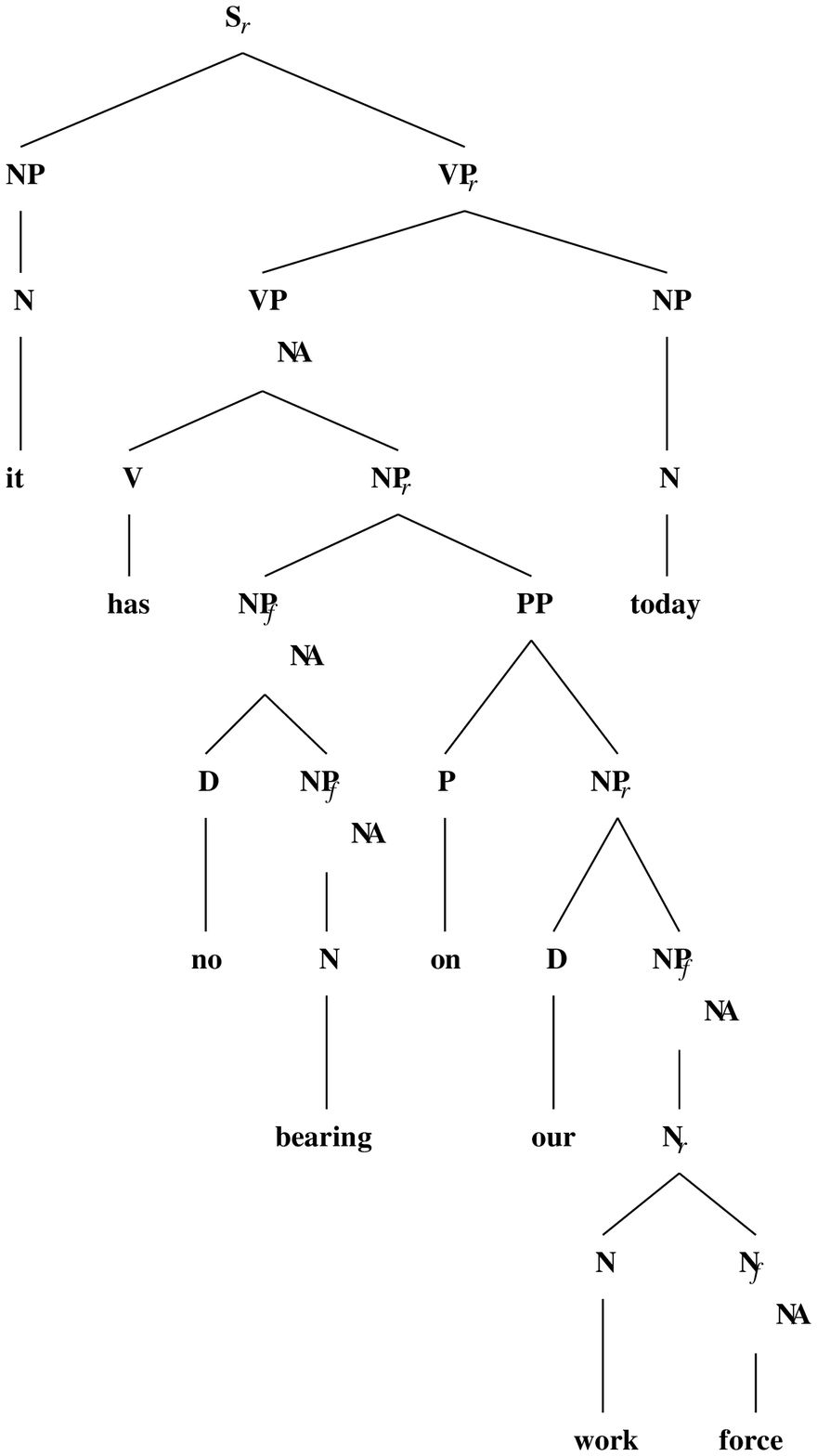,height=3.5in}} & 
{\psfig{figure=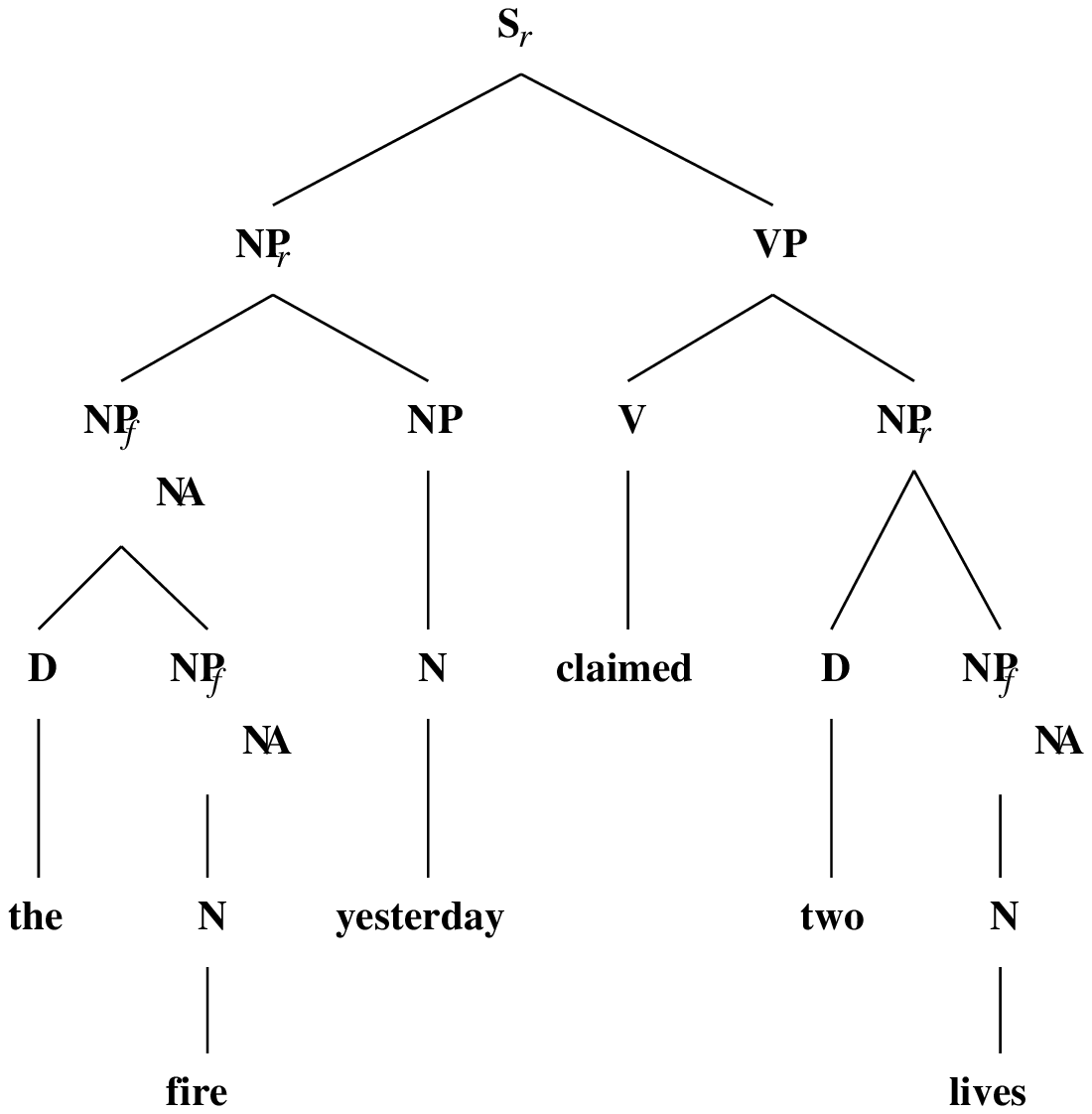,height=2.7in}} &  

\hspace*{-.55in} {\psfig{figure=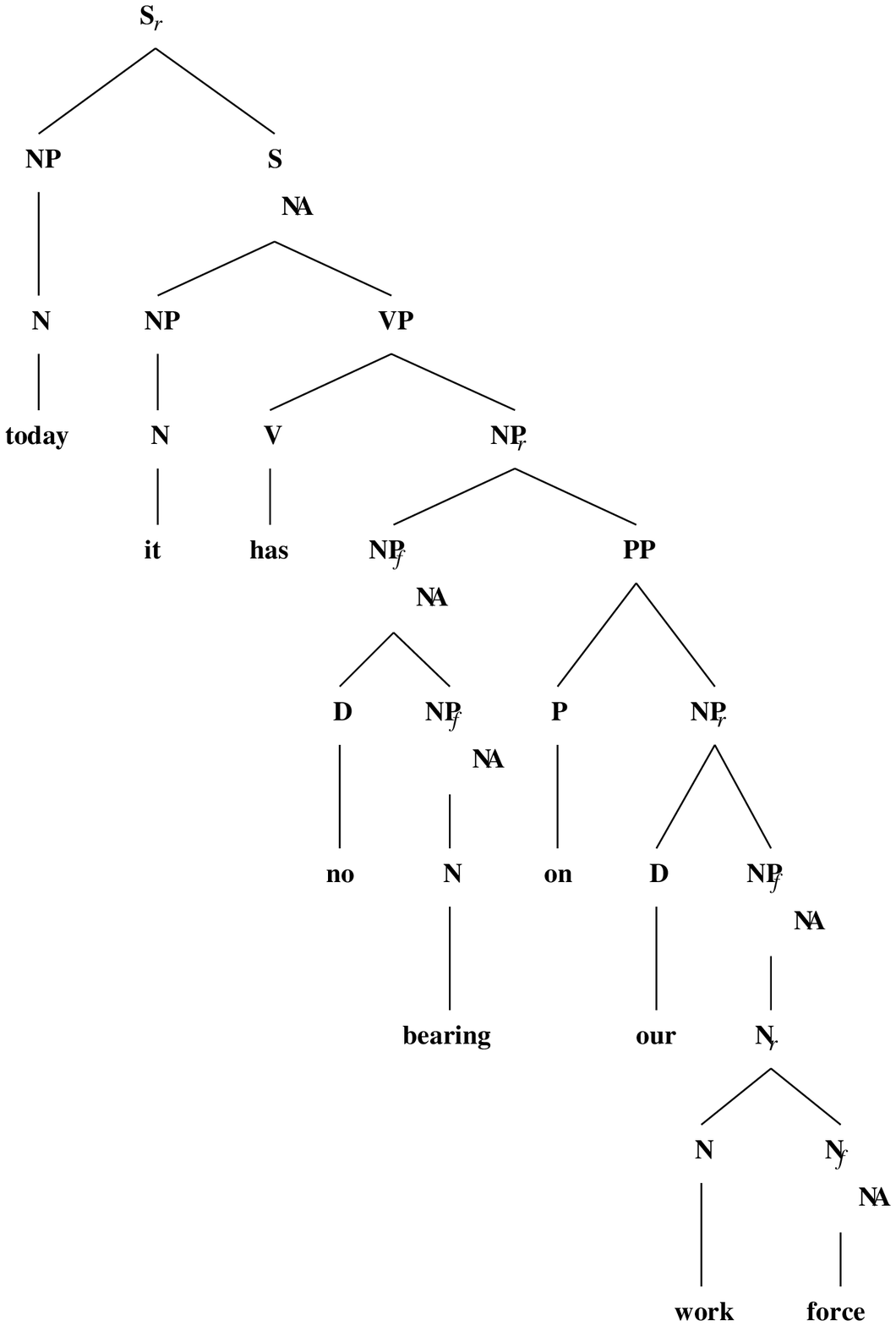,height=3.5in}} \\ \end{tabular}\\
\caption {Time~NPs in different positions
($\beta$vxN, $\beta$nxN and $\beta$Ns)} \label {bearingtrees}
\end{figure}

\begin{figure}[htb] 
\centering
\begin{tabular}{cc}
\psfig{figure=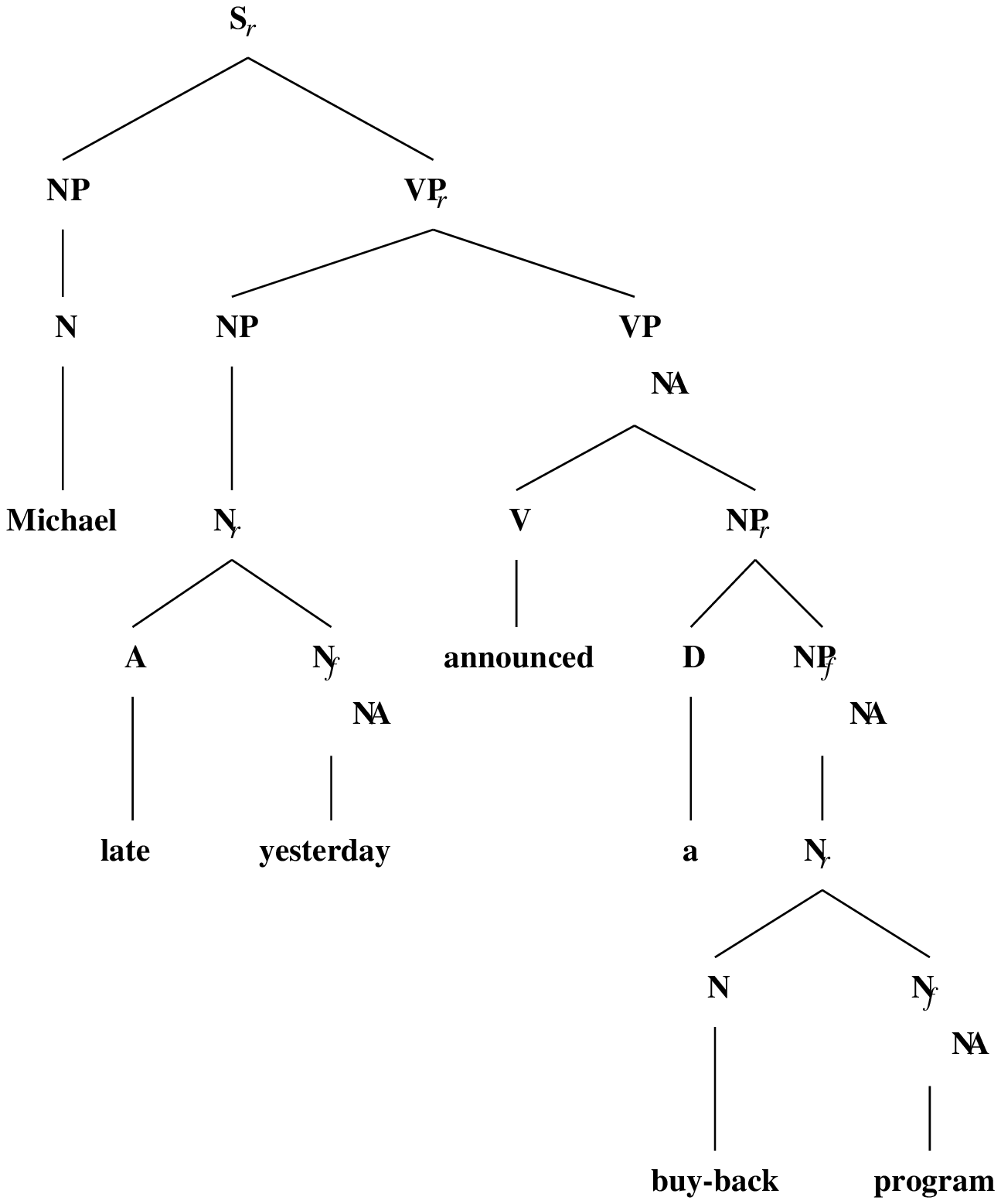,height=4in} &
\hspace{-1in} \raisebox{2.5in}{\psfig{figure=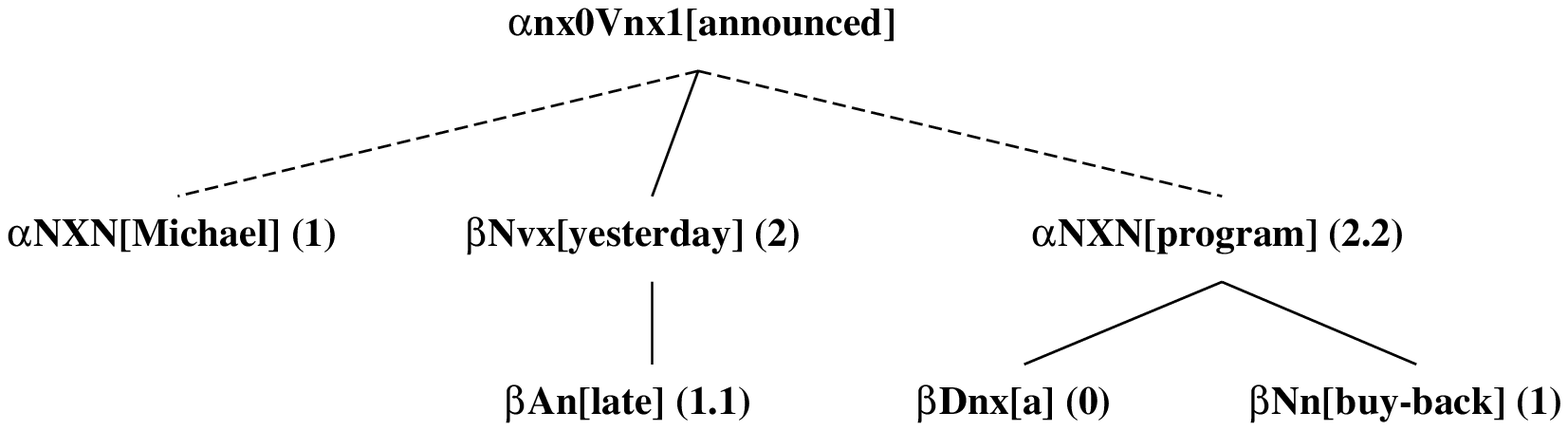,height=1.25in}} \\
\end{tabular}
\caption {Time~NPs: Derived tree and Derivation ($\beta$Nvx position)} 
\label{lateyesterday} 
\end{figure}

\section{Prepositions}
\label{prep-modifier}

There are three basic types of prepositional phrases, and three places
at which they can adjoin.  The three types of prepositional phrases
are: Preposition with NP Complement, Preposition with Sentential
Complement, and Exhaustive Preposition.  The three places are to the
right of an NP, to the right of a VP, and to the left of an S.  Each
of the three types of PP can adjoin at each of these three places, for
a total of nine PP modifier trees. Table \ref{prep-summary} gives the
tree family names for the various combinations of type and
location. (See Section \ref{post-PP} for discussion of the
$\beta$spuPnx, which handles post-sentential comma-separated PPs.)

\begin{table}[htb]
\centering
\begin{tabular}{|l||c|c|c|}
\hline
\multicolumn{1}{|c||}{}&\multicolumn{3}{c|}{position and category modified}\\
\cline{2-4}
\multicolumn{1}{|c||}{}&pre-sentential&post-NP&post-VP\\
\multicolumn{1}{|c||}{Complement type}&S modifier&NP modifier&VP modifier\\
\hline
\hline
S-complement&$\beta$Pss&$\beta$nxPs&$\beta$vxPs\\
\hline
NP-complement&$\beta$Pnxs&$\beta$nxPnx&$\beta$vxPnx\\
\hline
no complement&$\beta$Ps&$\beta$nxP&$\beta$vxP\\
(exhaustive)&&&\\
\hline
\end{tabular}
\caption{Preposition Anchored Modifiers}
\label{prep-summary}
\end{table}

The subset of preposition anchored modifier trees in Figure~\ref{prep-trees}
illustrates the locations and the four PP types.  Example sentences using the 
trees in Figure \ref{prep-trees} are shown in (\ex{1})-(\ex{4}). There are also
more trees with multi-word prepositions as anchors. Examples of these are: 
{\it ahead of}, {\it contrary to}, {\it at variance with} and {\it as recently
as}.

\begin{figure}[htb]
\centering
\begin{tabular}{ccccccc}
{\psfig{figure=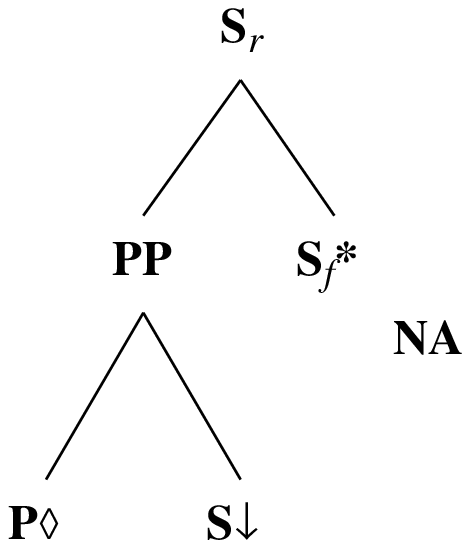,height=1.5in}}
& \hspace{.5in} &
{\psfig{figure=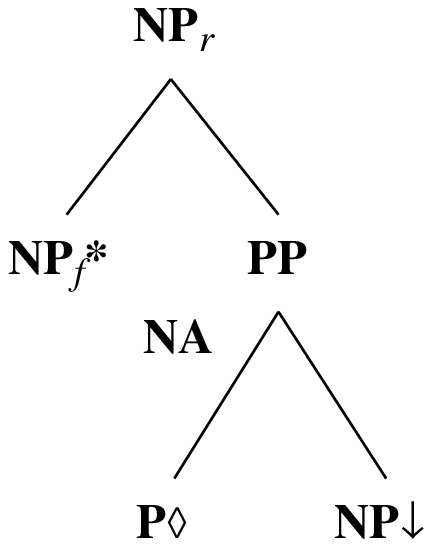,height=1.5in}}
&  \hspace{.5in} &
{\psfig{figure=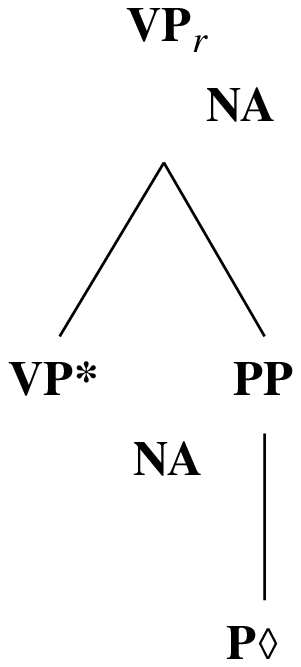,height=1.5in}}
&  \hspace{.5in} &
{\psfig{figure=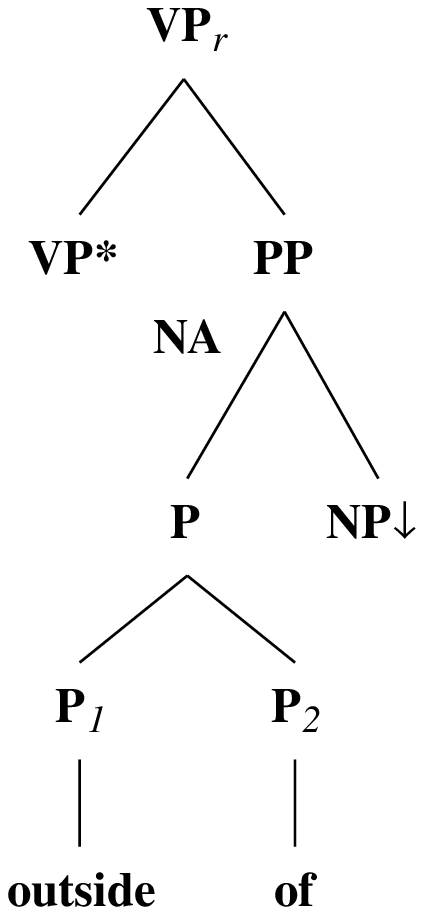,height=1.75in}}
\\
$\beta$Pss&&$\beta$nxPnx&&$\beta$vxP&&$\beta$vxPPnx\\
\end{tabular}\\
\caption {Selected Prepositional Phrase Modifier trees:
$\beta$Pss, $\beta$nxPnx, $\beta$vxP and $\beta$vxPPnx}
\label {prep-trees}
\end{figure}

\enumsentence{[$_{PP}$ with Clove healthy $_{PP}$], the veterinarian's
bill will be more affordable . ($\beta$Pss\footnote{{\it Clove healthy} is an adjective small clause})}
\enumsentence{The frisbee [$_{PP}$ in the brambles $_{PP}$] was hidden .
($\beta$nxPnx)}
\enumsentence{Clove played frisbee [$_{PP}$ outside $_{PP}$] . ($\beta$vxP)}
\enumsentence{Clove played frisbee [$_{PP}$ outside of the house
$_{PP}$] . ($\beta$vxPPnx)}

Prepositions that take NP complements assign accusative case to those
complements (see section~\ref{prep-case} for details).  Most prepositions take
NP complements.  Subordinating conjunctions are analyzed in XTAG as Preps 
(see Section~\ref{adjunct-cls} for details). Additionally, a few non-conjunction 
prepositions take S complements (see Section~\ref{NPA} for details).

\section{Adverbs}
\label{adv-modifier}

In the English XTAG grammar, VP and S-modifying adverbs anchor the
auxiliary trees $\beta$ARBs, $\beta$sARB, $\beta$vxARB and
$\beta$ARBvx,\footnote{In the naming conventions for the XTAG trees,
ARB is used for {\underline a}dve{\underline {rb}}s.  Because the
letters in A, Ad, and Adv are all used for other parts of speech
({\underline a}djective, {\underline d}eterminer and {\underline
v}erb), ARB was chosen to eliminate ambiguity.
Appendix~\ref{tree-naming} contains a full explanation of naming
conventions.}  allowing pre and post modification of S's and VP's.
Besides the VP and S-modifying adverbs, the grammar includes adverbs
that modify other categories. Examples of adverbs modifying an
adjective, an adverb, a PP, an NP, and a determiner are shown in
(\ex{1})-(\ex{8}). (See Sections \ref{par-adverb} and
\ref{post-adverb} for discussion of the $\beta$puARBpuvx and
$\beta$spuARB, which handle pre-verbal parenthetical adverbs and
post-sentential comma-separated adverbs.)

\begin{itemize}
\item{Modifying an adjective}
\enumsentence{{\bf extremely} good}
\enumsentence{{\bf rather} tall}
\enumsentence{rich {\bf enough}}

\item{Modifying an adverb}
\enumsentence{oddly {\bf enough}} 
\enumsentence{{\bf very} well}

\item{Modifying a PP}
\enumsentence{{\bf right} through the wall}

\item{Modifying a NP}
\enumsentence{{\bf quite} some time}

\item{Modifying a determiner}
\enumsentence{{\bf exactly} five men}

\end{itemize}

XTAG has separate trees for each of the modified categories and for pre and
post modification where needed.  The kind of treatment given to adverbs here is
very much in line with the base-generation approach proposed by \cite{Ernst84},
which assumes all positions where an adverb can occur to be base-generated, and
that the semantics of the adverb specifies a range of possible positions
occupied by each adverb. While the relevant semantic features of the adverbs
are not currently implemented, implementation of semantic features is scheduled
for future work.  The trees for adverb anchored modifiers are very similar in
form to the adjective anchored modifier trees.  Examples of two of the basic
adverb modifier trees are shown in Figure~\ref{adv-trees}.

\begin{figure}[hb]
\centering
\begin{tabular}{ccc}
{\psfig{figure=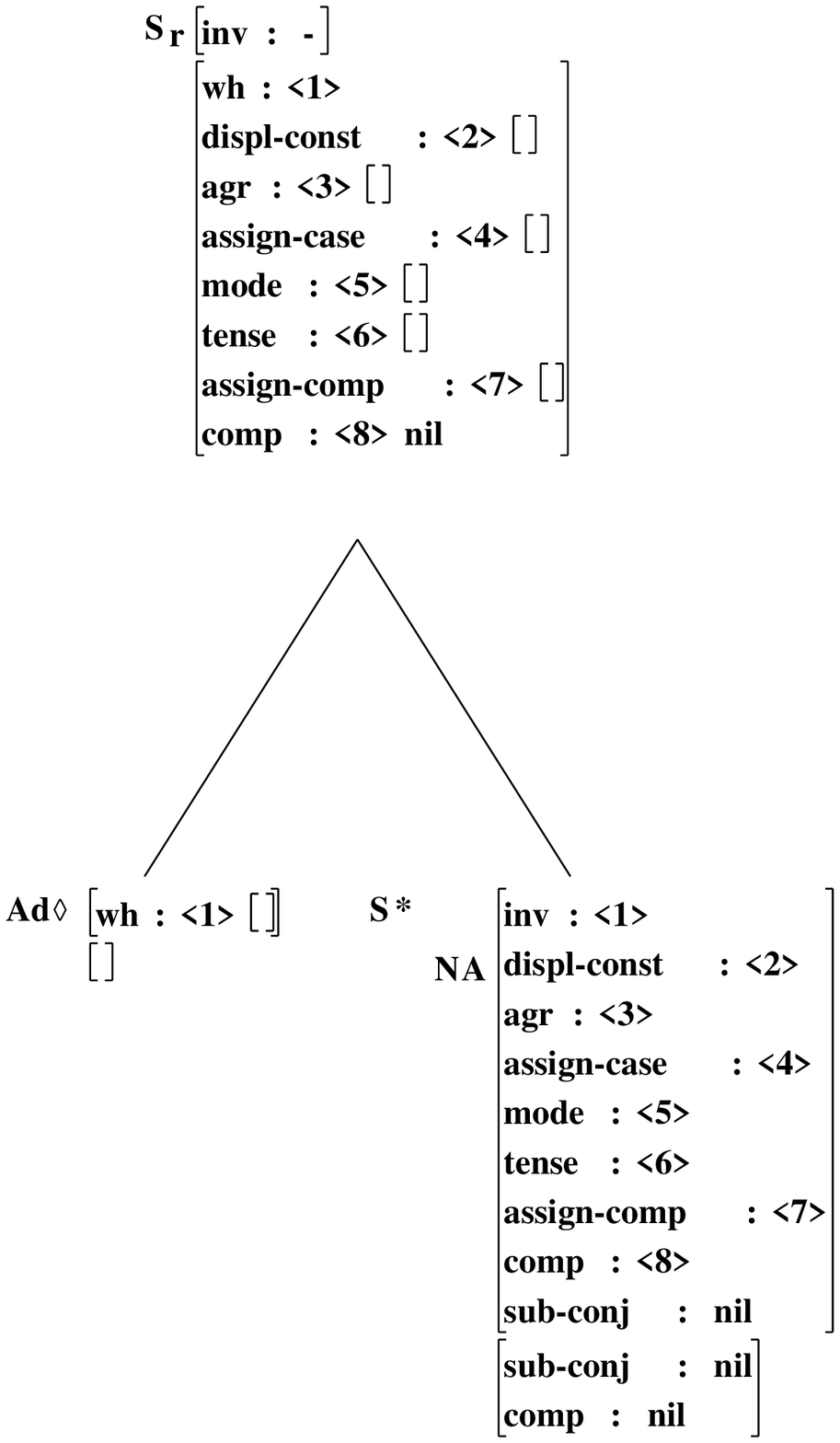,height=5in}}&
\hspace*{1.0in}&
{\psfig{figure=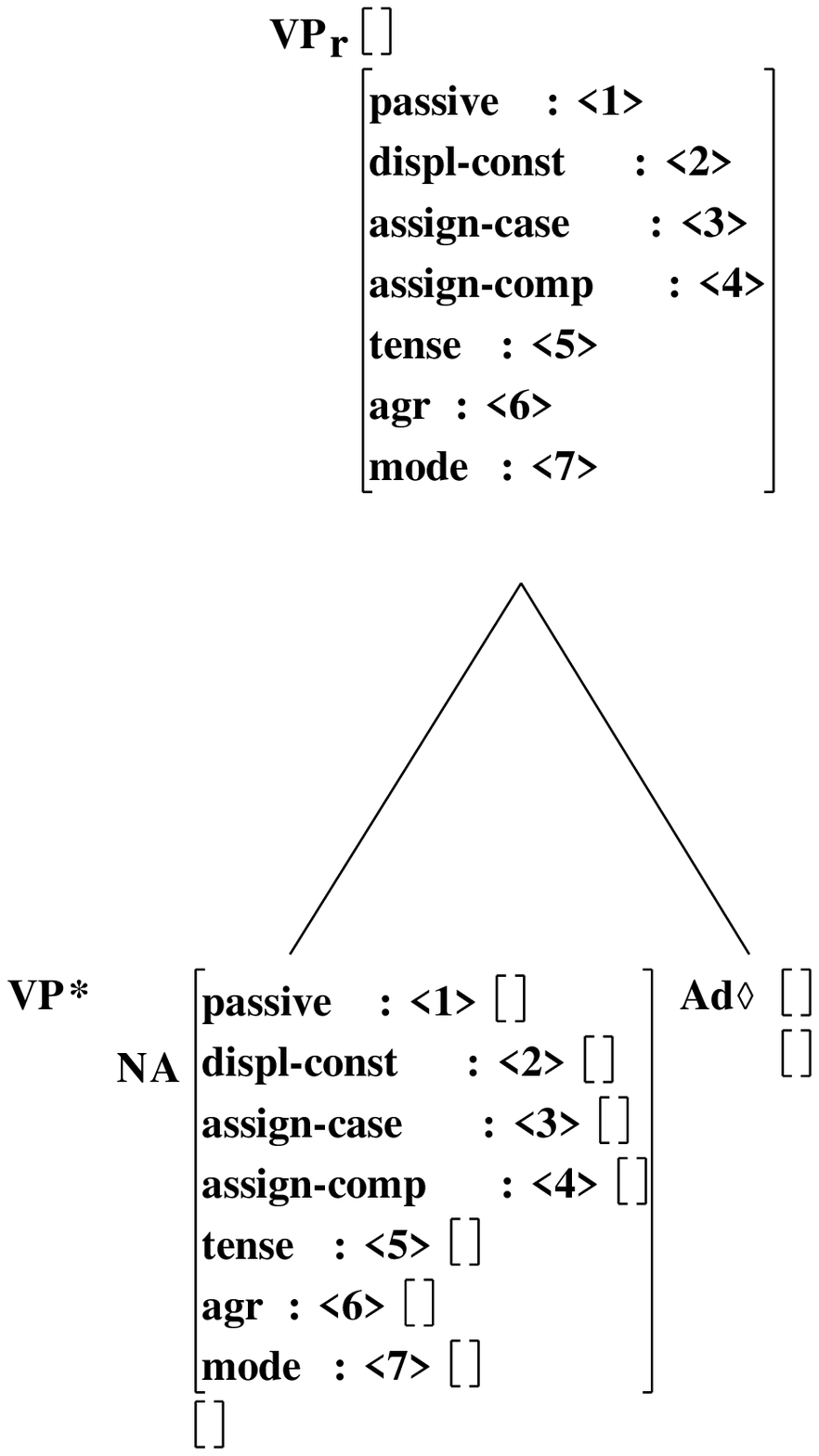,height=4.5in}}\\
(a)&&(b)\\
\end{tabular}
\caption {Adverb Trees for pre-modification of S: $\beta$ARBs (a) and
post-modification of a VP: $\beta$vxARB (b)}
\label{adv-trees}
\end{figure}

\newpage

Like the adjective anchored trees, these trees also have the NA
constraint on the foot node to restrict the number of derivations
produced for a sequence of adverbs.  Features of the modified category
are passed from the foot node to the root node, reflecting correctly
that these types of properties are unaffected by the adjunction of an
adverb.  A summary of the categories modified and the position of
adverbs is given in Table \ref{adv-summary}.

\begin{table}[h]
\centering
\begin{tabular}{|c||c|c|}
\hline
&\multicolumn{2}{c|}{Position with respect to item modified}\\
\cline{2-3}
Category Modified&Pre&Post\\
\hline
\hline
S&$\beta$ARBs&$\beta$sARB\\
\hline
VP&$\beta$ARBvx,$\beta$puARBpuvx&$\beta$vxARB\\
\hline
A&$\beta$ARBa&$\beta$aARB\\
\hline
PP&$\beta$ARBpx&$\beta$pxARB\\
\hline
ADV&$\beta$ARBarb&$\beta$arbARB\\
\hline
NP&$\beta$ARBnx&\\
\hline
Det&$\beta$ARBd&\\
\hline
\end{tabular}
\caption{Simple Adverb Anchored Modifiers}
\label{adv-summary}
\end{table}

In the English XTAG grammar, no traces are posited for wh-adverbs, in-line with
the base-generation approach (\cite{Ernst84}) for various positions of
adverbs. Since convincing arguments have been made against traces for adjuncts
of other types (e.g. \cite{Baltin}), and since the reasons for wanting traces
do not seem to apply to adjuncts, we make the general assumption in our grammar
that adjuncts do not leave traces.  Sentence initial wh-adverbs select the same
auxiliary tree used for other sentence initial adverbs ($\beta$ARBs) with the
feature {\bf $<$wh$>$=+}.  Under this treatment, the derived tree for the
sentence {\it How did you fall?} is as in Figure (\ref{how-did-you-fall}), with
no trace for the adverb.

\begin{figure}[htbp]
\centering
\begin{tabular}{c}
{\psfig{figure=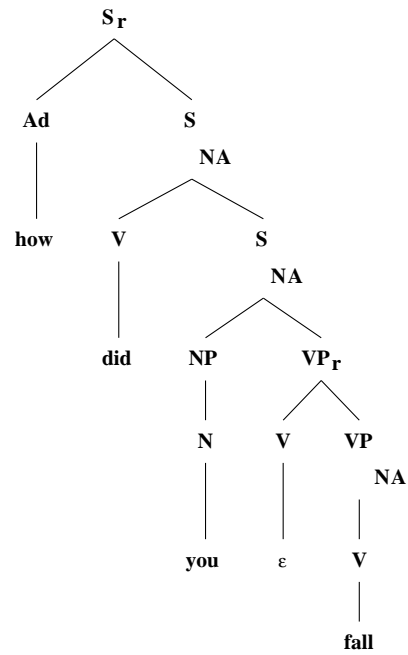,height=3.5in}}
\end{tabular}
\caption {Derived tree for {\it How did you fall?}}
\label {how-did-you-fall}
\end{figure}

\begin{figure}[htbp]
\centering
\begin{tabular}{c}
{\psfig{figure=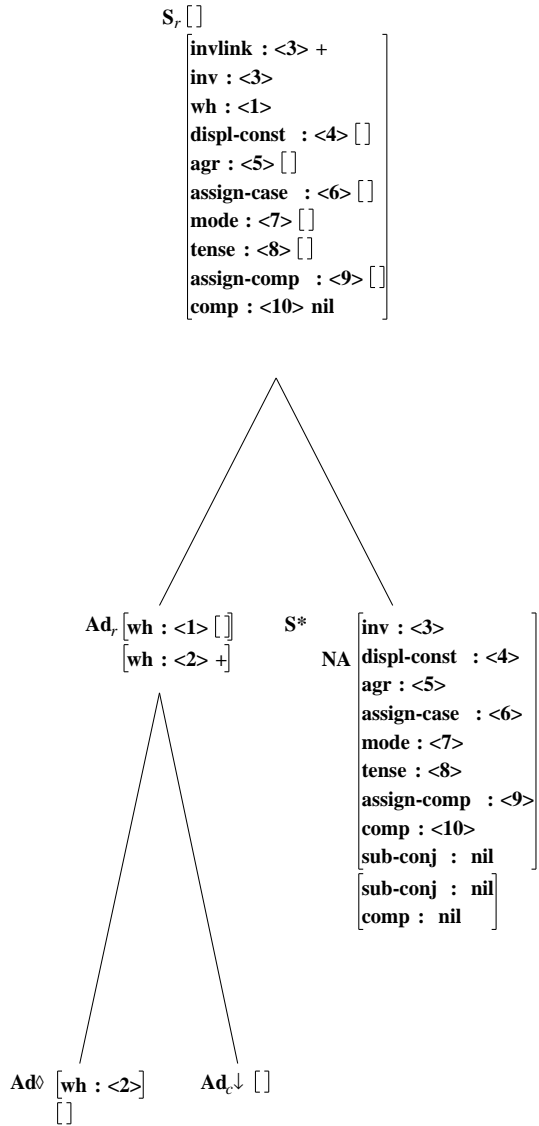,height=6.0in}}
\end{tabular}
\caption {Complex adverb phrase modifier: $\beta$ARBarbs}
\label{weird-adv-tree}
\end{figure}

There is one more adverb modifier tree in the grammar which is not included in
Table \ref{adv-summary}.  This tree, shown in Figure~\ref{weird-adv-tree}, has
a complex adverb phrase and is used for wh+ two-adverb phrases that occur
sentence initially, such as in sentence (\ex{1}).  Since {\it how} is the only
wh+ adverb, it is the only adverb that can anchor this tree.

\enumsentence{how quickly did Srini fix the problem ?}

Focus adverbs such as {\it only}, {\it even}, {\it just} and {\it at least} 
are also handled by the system.  Since the syntax allows focus adverbs to 
appear in practically any position, these adverbs select most of the trees 
listed in Table \ref{adv-summary}.  It is left up to the semantics or 
pragmatics to decide the correct scope for the focus adverb for a given 
instance.  In terms of the ability of the focus adverbs to modify at different
levels of a noun phrase, the focus adverbs can modify either cardinal 
determiners or noun-cardinal noun phrases, and cannot modify at the level of 
noun.  The tree for adverbial modification of noun phrases is in shown
Figure~\ref{other-adv-trees}(a). 

In addition to {\it at least}, the system handles the other two-word
adverbs, {\it at most} and {\it up to}, and the three-word {\it as-as}
adverb constructions, where an adjective substitutes between the two
occurrences of {\it as}.  An example of a three-word {\it as-as} adverb
is {\it as little as}.  Except for the ability of {\it at least} to
modify many different types of constituents as noted above, the
multi-word adverbs are restricted to modifying cardinal determiners.
Example sentences using the trees in Figure~\ref{other-adv-trees} are
shown in (\ex{1})-(\ex{5}).

\begin{itemize}
\item{Focus Adverb modifying an NP}
\enumsentence{{\bf only} a member of our crazy family could pull off that kind
of a stunt .}
\enumsentence{{\bf even} a flying saucer sighting would seem interesting in
comparison \\ with your story .}
\enumsentence{The report includes a proposal for {\bf at least} a partial 
impasse in negotiations .}

\item{Multi-word adverbs modifying cardinal determiners}
\enumsentence{{\bf at most} ten people came to the party .} 
\enumsentence{They gave monetary gifts of {\bf as little as} five dollars .}

\end{itemize}

\begin{figure}[htb]
\centering
\begin{tabular}{ccccccc}
{\psfig{figure=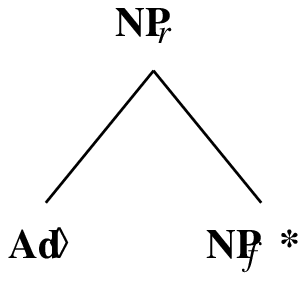,height=1.1in}}
&  \hspace{.5in} &
{\psfig{figure=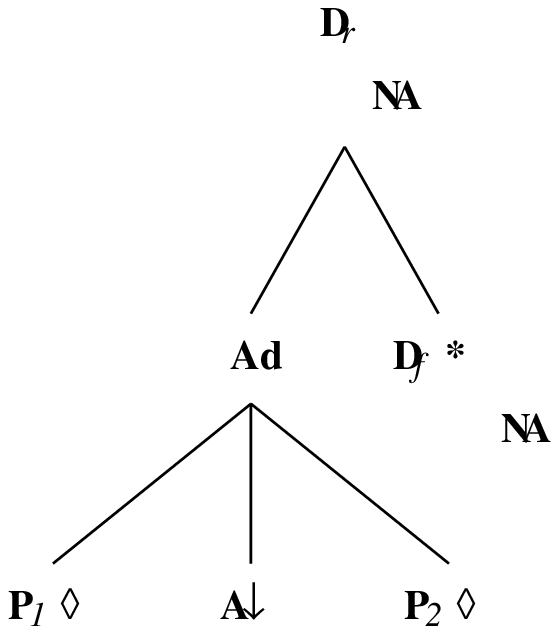,height=1.75in}}
& \hspace{.5in} &
{\psfig{figure=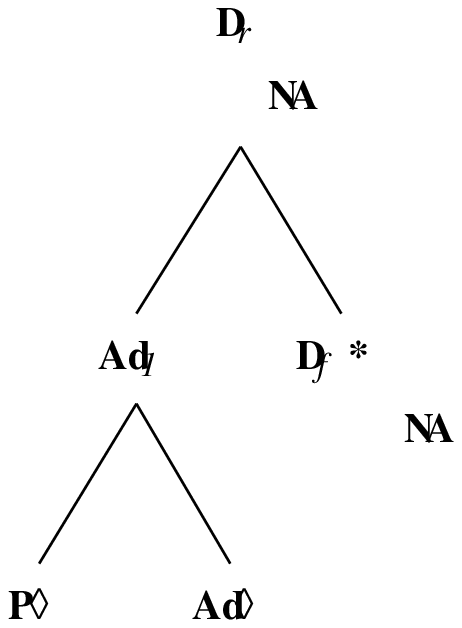,height=1.75in}}\\

$\beta$ARBnx&&$\beta$PaPd&&$\beta$PARBd&&\\
(a)&&(b)&&(c)\\
\end{tabular}\\
\caption {Selected Focus and Multi-word Adverb Modifier trees:
$\beta$ARBnx, $\beta$PARBd and $\beta$PaPd}
\label {other-adv-trees}
\end{figure}

The grammar also includes auxiliary trees anchored by multi-word adverbs
like {\it a little}, {\it a bit}, {\it a mite}, {\it sort of}, {\it kind
of}, etc.. 

Multi-word adverbs like {\it sort of} and {\it kind of} can pre- modify
almost any non-clausal category. The only strict constraint on their
occurrence is that they can't modify nouns (in which case an adjectival
interpretation would obtain)\footnote{Note that there are semantic/lexical
constraints even for the categories that these adverbs {\it can} modify,
and no doubt invite a more in-depth analysis.}. The category which they
scope over can be directly determined from their position, except for when
they occur sentence finally in which case they are assumed to modify
VP's. The complete list of auxiliary trees anchored by these adverbs are as
follows: $\beta$NPax, $\beta$NPpx, $\beta$NPnx, $\beta$NPvx, $\beta$vxNP,
$\beta$NParb. Selected trees are shown in Figure~\ref{sortof-adv-tree}, and
some examples are given in (\ex{1})-(\ex{4}).

\begin{figure}[htb]
\centering
\begin{tabular}{ccccccc}
{\psfig{figure=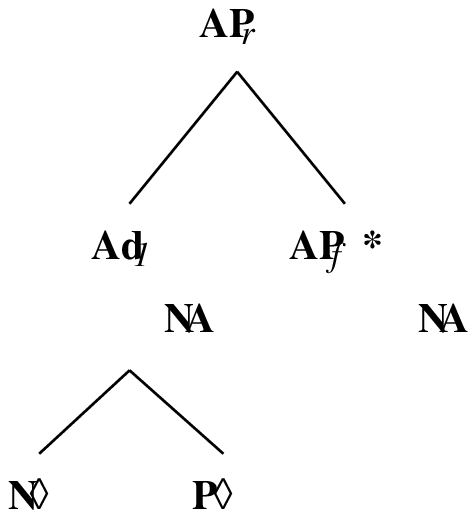,height=1.5in}}
& \hspace{.5in} & 
{\psfig{figure=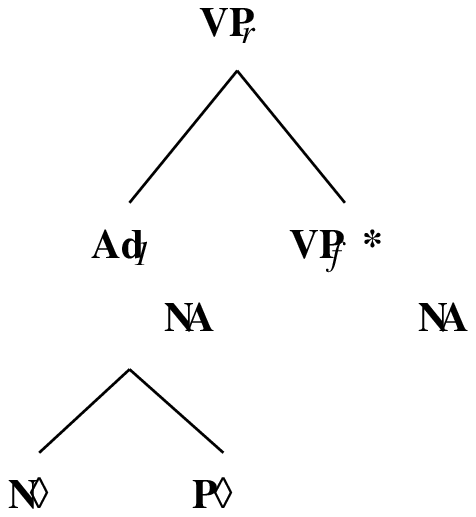,height=1.5in}}
& \hspace{.5in} &
{\psfig{figure=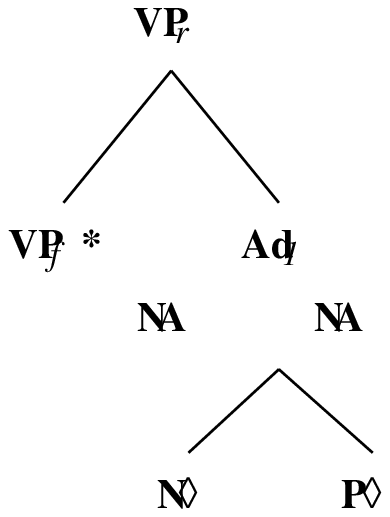,height=1.5in}}
\\
$\beta$NPax&&$\beta$NPvx&&$\beta$vxNP&&\\
(a)&&(b)&&(c)&&\\
\end{tabular}\\
\caption{Selected Multi-word Adverb Modifier trees (for adverbs like {\it
sort of}, {\it kind of}): $\beta$NPax, $\beta$NPvx, $\beta$vxNP.}
\label{sortof-adv-tree}
\end{figure}

\enumsentence{John is {\bf sort of} [$_{AP}$ tired].}

\enumsentence{John is {\bf sort of} [$_{PP}$ to the right].}
 
\enumsentence{John could have been {\bf sort of} [$_{VP}$ eating the cake].}
 
\enumsentence{John has been eating his cake {\bf sort of} [$_{ADV}$ slowly].}

There are some multi-word adverbs that are, however, not so free in their
distribution. Adverbs like {\it a little}, {\it a bit}, {\it a mite} modify
AP's in predicative constructions (sentences with the copula and small
clauses, AP complements in sentences with raising verbs, and AP's when they
are subcategorized for by certain verbs (e.g., {\it John felt angry}). They
can also post-modify VP's and PP's, though not as freely as
AP's\footnote{They can also appear before NP's, as in, ``John wants {\it a
little} sugar''. However, here they function as multi-word determiners and
should not be analyzed as adverbs.}. Finally, they also function as
downtoners for almost all adverbials\footnote{It is to be noted that this
analysis, which allows these multiword adverbs to modify adjectival phrases
as well as adverbials, will yield (not necessarily desirable) multiple
derivations for a sentence like {\it John is a little unecessarily
stupid}. In one derivation, {\it a little} modifies the AP and in the other
case, it modifies the adverb.}. Some examples are provided in
(\ex{1})-(\ex{4}).

\enumsentence{Mickey is {\bf a little} [$_{AP}$ tired].}

\enumsentence{The medicine [$_{VP}$ has eased John's pain] {\bf a little}.}

\enumsentence{John is {\bf a little} [$_{PP}$ to the right].}

\enumsentence{John has been reading his book {\bf a little} [$_{ADV}$ loudly].}

Following their behavior as described above, the auxiliary trees they
anchor are $\beta$DAax, $\beta$DApx, $\beta$vxDA, $\beta$DAarb,
$\beta$DNax, $\beta$DNpx, $\beta$vxDN, $\beta$DNarb. Selected trees are
shown in Figure~\ref{alittle-adv-tree}).

\begin{figure}[htb]
\centering
\begin{tabular}{ccccccc}
{\psfig{figure=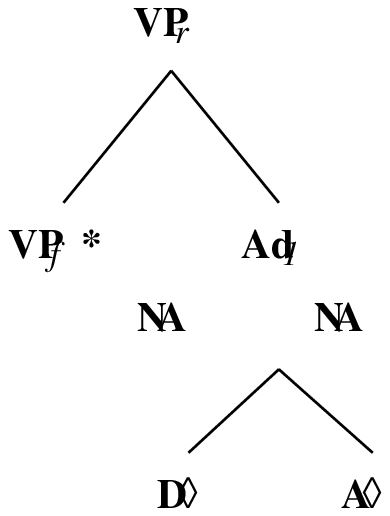,height=1.5in}}
& \hspace{.5in} & 
{\psfig{figure=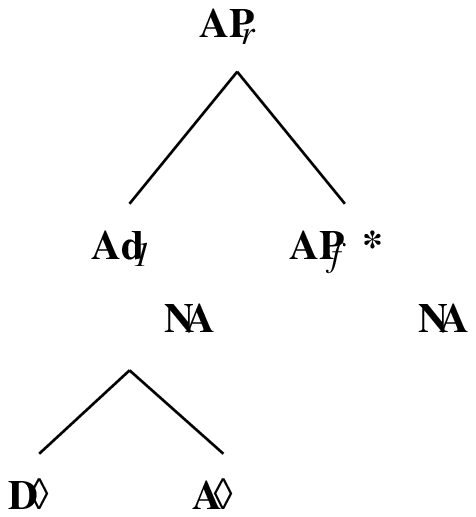,height=1.5in}}
& \hspace{.5in} &
{\psfig{figure=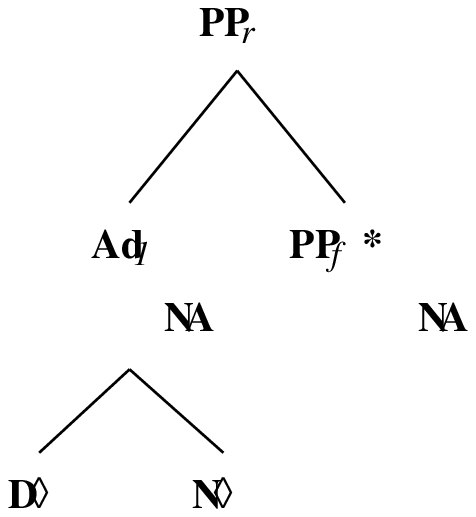,height=1.5in}}
\\
$\beta$vxDA&&$\beta$DAax&&$\beta$DNpx&&\\
(a)&&(b)&&(c)&&\\
\end{tabular}\\
\caption{Selected Multi-word Adverb Modifier trees (for adverbs like {\it
a little}, {\it a bit}): $\beta$vxDA, $\beta$DAax, $\beta$DNpx.}
\label{alittle-adv-tree}
\end{figure}

\section{Locative Adverbial Phrases}
\label{locatives}

Locative adverbial phrases are multi-word adverbial modifiers whose 
meanings relate to spatial location. Locatives consist of a locative adverb 
(such as {\it ahead} or {\it downstream}) preceded by an NP, an adverb, or 
nothing, as in Examples (\ex{1})--(\ex{3}) respectively. The modifier as a 
whole describes a position relative to one previously 
specified in the discourse. The nature of the relation, which is usually
a direction, is specified by the anchoring locative adverb({\em behind, 
east}). If an NP or a second adverb is present in the phrase, it specifies the 
degree of the relation (for example: {\it three city blocks, many meters,} 
and {\it far}).

\enumsentence{The accident {\it three blocks ahead} stopped traffic}
\enumsentence{The ship sank {\it far offshore}}
\enumsentence{The trouble {\it ahead} distresses me}

Locatives can modify NPs, VPs and Ss. They modify NPs only by right-adjoining 
post-positively, as in Example (\ex{-2}). Post-positive is also the more 
common position when a locative modifies either of the other categories. 
Locatives pre-modify VPs only when separated by balanced punctuation 
(commas or dashes). The trees locatives select when modifying NPs are shown 
in Figure~\ref{loc-np-trees}.

\begin{figure}[htb]
\centering
\begin{tabular}{ccc}
\psfig{figure=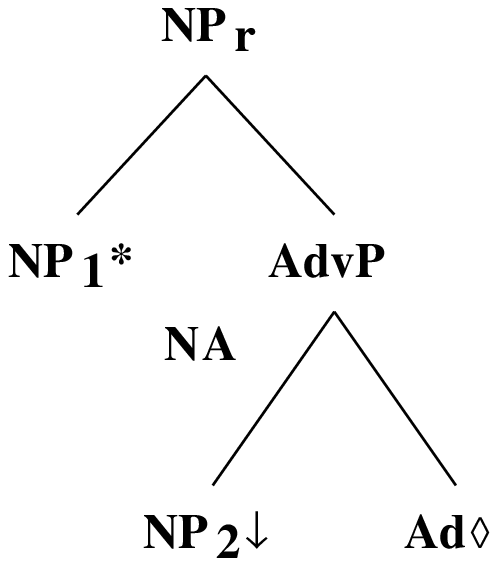,height=4.0cm}
& \hspace{0.5in} &
\psfig{figure=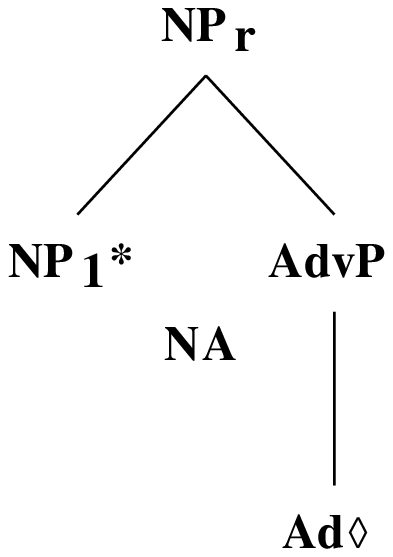,height=4.0cm}
\end{tabular}
\caption{Locative Modifier Trees: $\beta$nxnxARB, $\beta$nxARB}
\label{loc-np-trees}
\end{figure}
  
When the locative phrase consists of only the anchoring locative adverb, as in
Example (\ex{-1}), it uses the $\beta$nxARB tree, shown in 
Figure~\ref{loc-np-trees}, and its VP analogue, $\beta$vxARB. In 
addition, these are the trees selected when the locative anchor is 
modified by an adverb expressing degree, as in Example \ex{-1}. The 
degree adverb adjoins on to the anchor using the $\beta$ARBarb tree, 
which is described in Section~\ref{adv-modifier}. Figure~\ref{toupees} 
shows an example of these trees in action. Though there is a tree 
for a pre-sentential locative phrase, $\beta$nxARBs, there is no 
corresponding post-sentential tree, as it is highly debatable whether 
the post-sentential version actually has the entire sentence or just 
the preceding verb phrase as its scope. Thus, in accordance with XTAG 
practice, which considers ambiguous post-sentential modifiers to be 
VP-modifiers rather than S-modifiers, there is only a $\beta$vxnxARB 
tree, as shown in Figure~\ref{toupees}. 

\begin{figure}[htb]
\centering
\begin{tabular}{ccc}
\psfig{figure=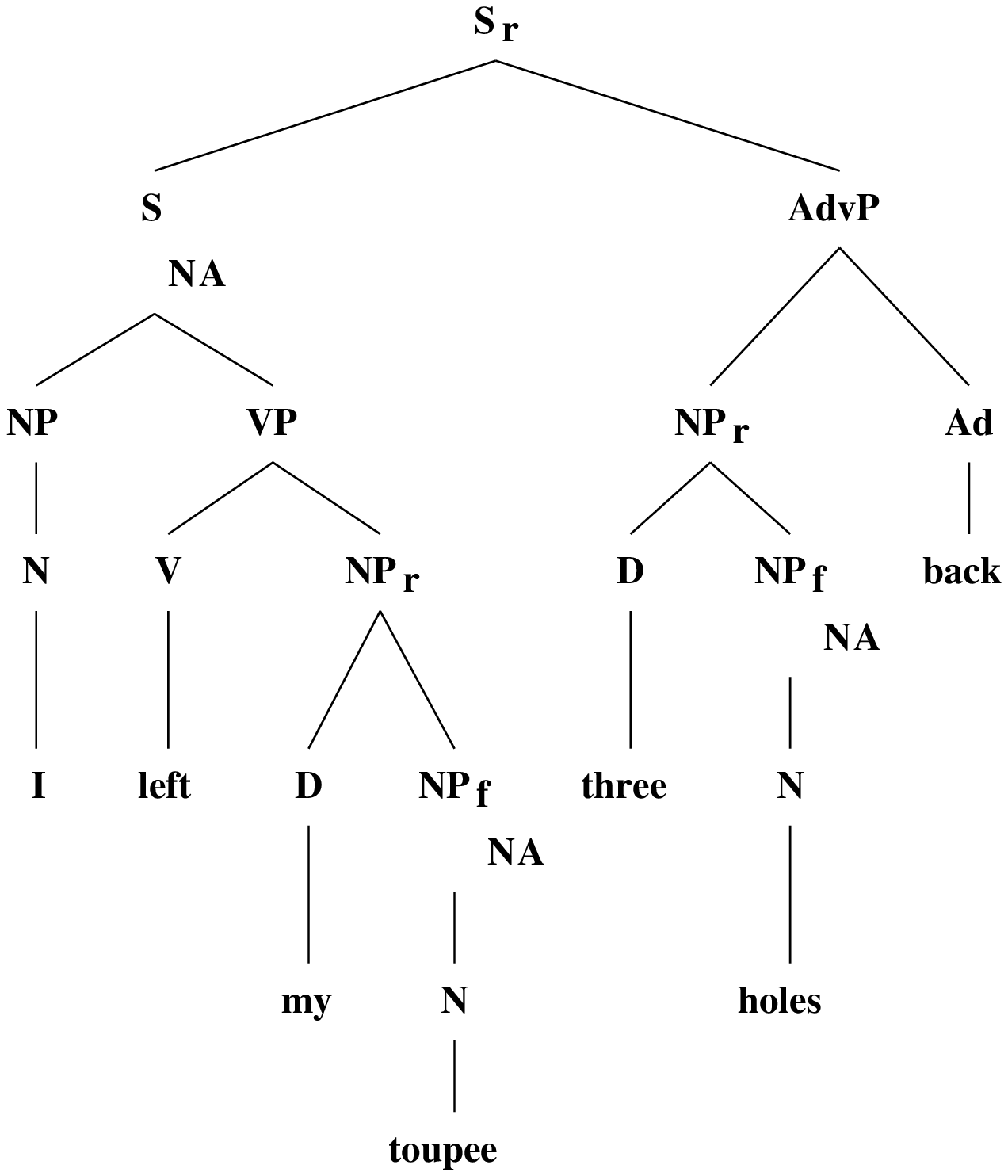,height=7.0cm}
& \hspace{0.5in} &
\psfig{figure=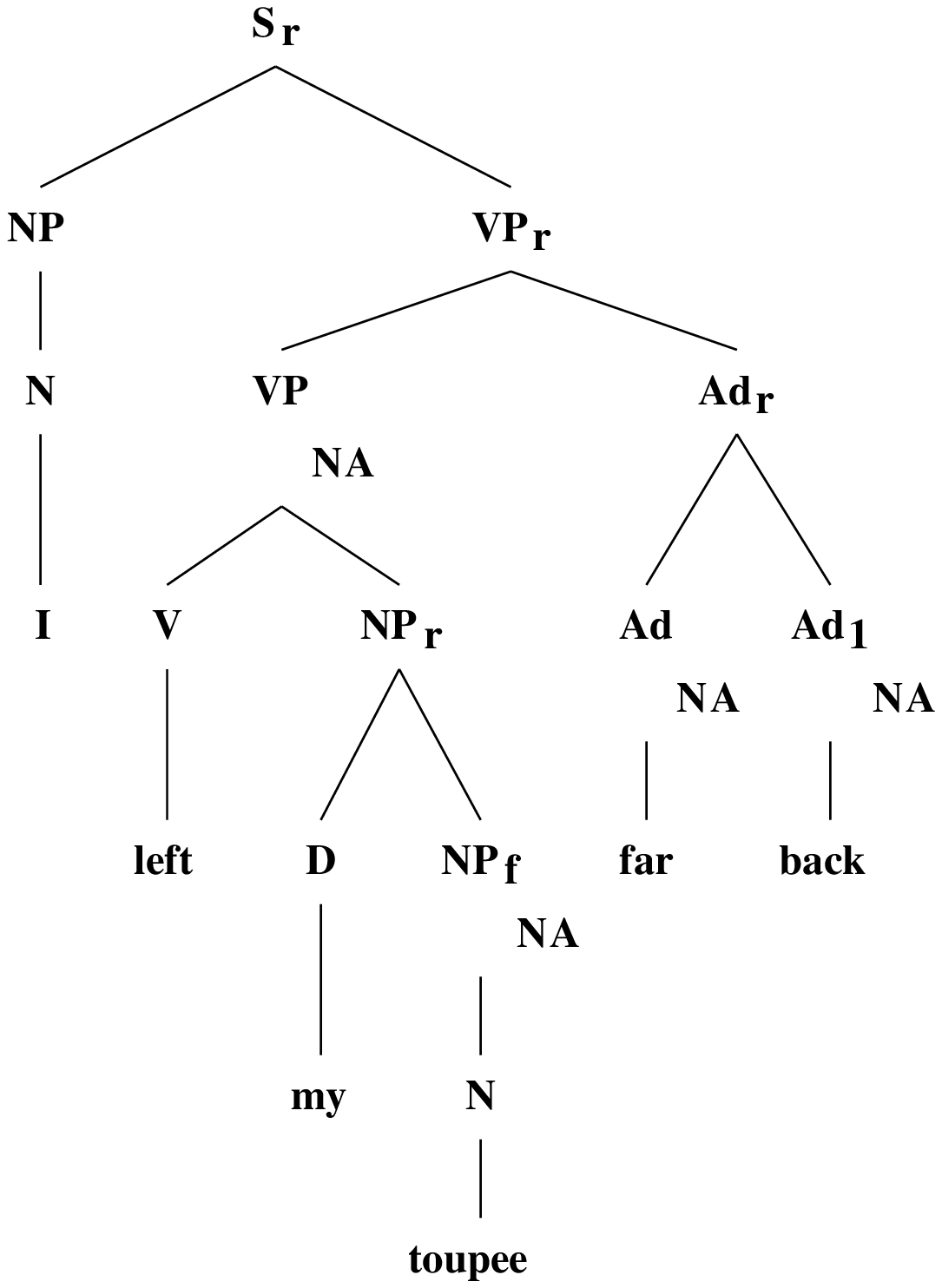,height=7.0cm}
\end{tabular}
\caption{Locative Phrases featuring NP and Adverb Degree Specifications}
\label{toupees}
\end{figure}

One possible analysis of locative phrases with NPs might maintain that 
the NP is the head, with the locative adverb modifying the NP. This is 
initially attractive because of the similarity to time NPs, which also 
feature NPs that can modify clauses. This analysis seems insufficient, 
however, in light of the fact that virtually any NP can occur in 
locative phrases, as in example (\ex{1}). Therefore, in the XTAG analysis 
the locative adverb anchors the locative phrase trees. A complete summary 
of all trees selected by locatives is contained in Table~\ref{loc-summary}. 
26\footnote{Though nearly all of these adverbs are spatial in nature, 
this number also includes a few temporal adverbs, such as {\it ago}, 
that also select these trees.} adverbs select the locative trees.

\enumsentence{I left my toupee and putter {\it three holes back}}

\begin{table}[htb]
\centering
\begin{tabular}{|l||c|c|}
\hline
\multicolumn{1}{|c||}{}&
\multicolumn{2}{|c|}{Degree Phrase Type}\\
\cline{2-3}
\multicolumn{1}{|c||}{Category Modified}&NP&Ad/None\\
\hline
\hline
NP&$\beta$nxnxARB&$\beta$nxARB\\
\hline
VP (post)&$\beta$vxnxARB&$\beta$vxARB\\
\hline
VP (pre, punct-separated)&$\beta$punxARBpuvx&$\beta$puARBpuvx\\
\hline
S&$\beta$nxARBs&$\beta$ARBs\\
\hline
\end{tabular}
\caption{Locative Modifiers}
\label{loc-summary}
\end{table}

\chapter{Auxiliaries}
\label{auxiliaries}

Although there has been some debate about the lexical category of auxiliaries,
the English XTAG grammar follows \cite{mccawley88}, \cite{haegeman91}, and
others in classifying auxiliaries as verbs. The category of verbs can therefore
be divided into two sets, main or lexical verbs, and auxiliary verbs, which can
co-occur in a verbal sequence.  Only the highest verb in a verbal sequence is
marked for tense and agreement regardless of whether it is a main or auxiliary
verb.  Some auxiliaries ({\it be}, {\it do}, and {\it have}) share with main
verbs the property of having overt morphological marking for tense and
agreement, while the modal auxiliaries do not.  However, all auxiliary verbs
differ from main verbs in several crucial ways.

\begin{itemize}

\item Multiple auxiliaries can occur in a single sentence, while a matrix
sentence may have at most one main verb. 

\item Auxiliary verbs cannot occur as the sole verb in the sentence, but must
be followed by a main verb.

\item All auxiliaries precede the main verb in verbal sequences.

\item Auxiliaries do not subcategorize for any arguments.

\item Auxiliaries impose requirements on the morphological form of the verbs
that immediately follow them.

\item Only auxiliary verbs invert in questions (with the sole exception in 
American English of main verb {\it be}\footnote{Some dialects, particularly
British English, can also invert main verb {\it have} in yes/no questions
(e.g. {\it have you any Grey Poupon ?}).  This is usually attributed to the
influence of auxiliary {\it have}, coupled with the historic fact that English
once allowed this movement for all verbs.\label{have-footnote}}).

\item An auxiliary verb must precede sentential negation (e.g. $\ast${\it John not goes}).

\item Auxiliaries can form contractions with subjects and negation (e.g. {\it
he'll}, {\it won't}).

\end{itemize}

\noindent The restrictions that an auxiliary verb imposes on the succeeding verb limits
the sequence of verbs that can occur.  In English, sequences of up to five
verbs are allowed, as in sentence (\ex{1}).

\enumsentence{The music should have been being played [for the president] .}

\noindent 
The required ordering of verb forms when all five verbs are present is:

\begin{quote}
\begin{tabular}{ccl}
& & {\bf modal base perfective progressive passive}
\end{tabular}
\end{quote}

\noindent
The rightmost verb is the main verb of the sentence.  While a main verb
subcategorizes for the arguments that appear in the sentence, the auxiliary
verbs select the particular morphological forms of the verb to follow each of
them.  The auxiliaries included in the English XTAG grammar are listed in Table
\ref{aux-table} by type.  The third column of Table \ref{aux-table} lists the
verb forms that are required to follow each type of auxiliary verb.

\vspace*{0.2in}

\begin{table}[ht]
\centering
\begin{tabular}{|l|c|c|}  
\hline
TYPE&LEX ITEMS&SELECTS FOR\\     
\hline
modals & {\it can}, {\it could}, {\it may}, {\it might}, {\it will}, & base form\footnotemark
\\ & {\it would}, {\it ought}, {\it shall}, {\it should} & (e.g. {\it will
go}, {\it might come})\\ & {\it need} &\\
\hline
perfective & {\it have} & past participle\\
& & (e.g. {\it has gone})\\  
\hline
progressive & {\it be} & gerund\\
& & (e.g. {\it is going}, {\it was coming})\\  
\hline
passive & {\it be} & past participle\\
& & (e.g. {\it was helped by Jane})\\  
\hline
do support & {\it do} &base form\\
& & (e.g. {\it did go}, {\it does come})\\  
\hline
infinitive to & {\it to} & base form\\
& & (e.g. {\it to go}, {\it to come})\\  
\hline
\end{tabular}
\caption{Auxiliary Verb Properties}
\label{aux-table}
\end{table}

\vspace*{0.2in}

\footnotetext{There are American dialects, particularly in the South, which
allow double modals such as {\it might could} and {\it might should}. These
constructions are not allowed in the XTAG English grammar.}

\section{Non-inverted sentences}
\label{aux-non-inverted}

This section and the sections that follow describe how the English XTAG grammar
accounts for properties of the auxiliary system described above.

In our grammar, auxiliary trees are added to the main verb tree by adjunction.
Figure~\ref{Vvx} shows the adjunction tree for non-inverted
sentences.\footnote{We saw this tree briefly in section~\ref{case-for-verbs},
but with most of its features missing.  The full tree is presented here.}

\begin{figure}[htb]
\centering
\begin{tabular}{c}
\psfig{figure=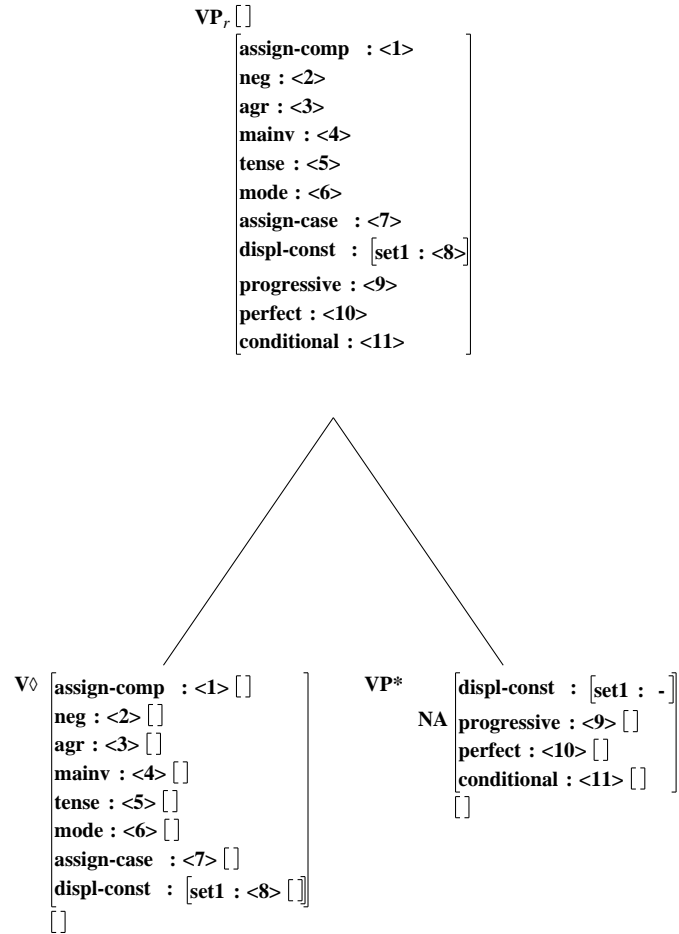,height=5.2in}
\end{tabular}
\caption{Auxiliary verb tree for non-inverted sentences: $\beta$Vvx }
\label{Vvx} 
\end{figure}

\begin{figure}[htbp]
\centering
\begin{tabular}{ccc}
{\psfig{figure=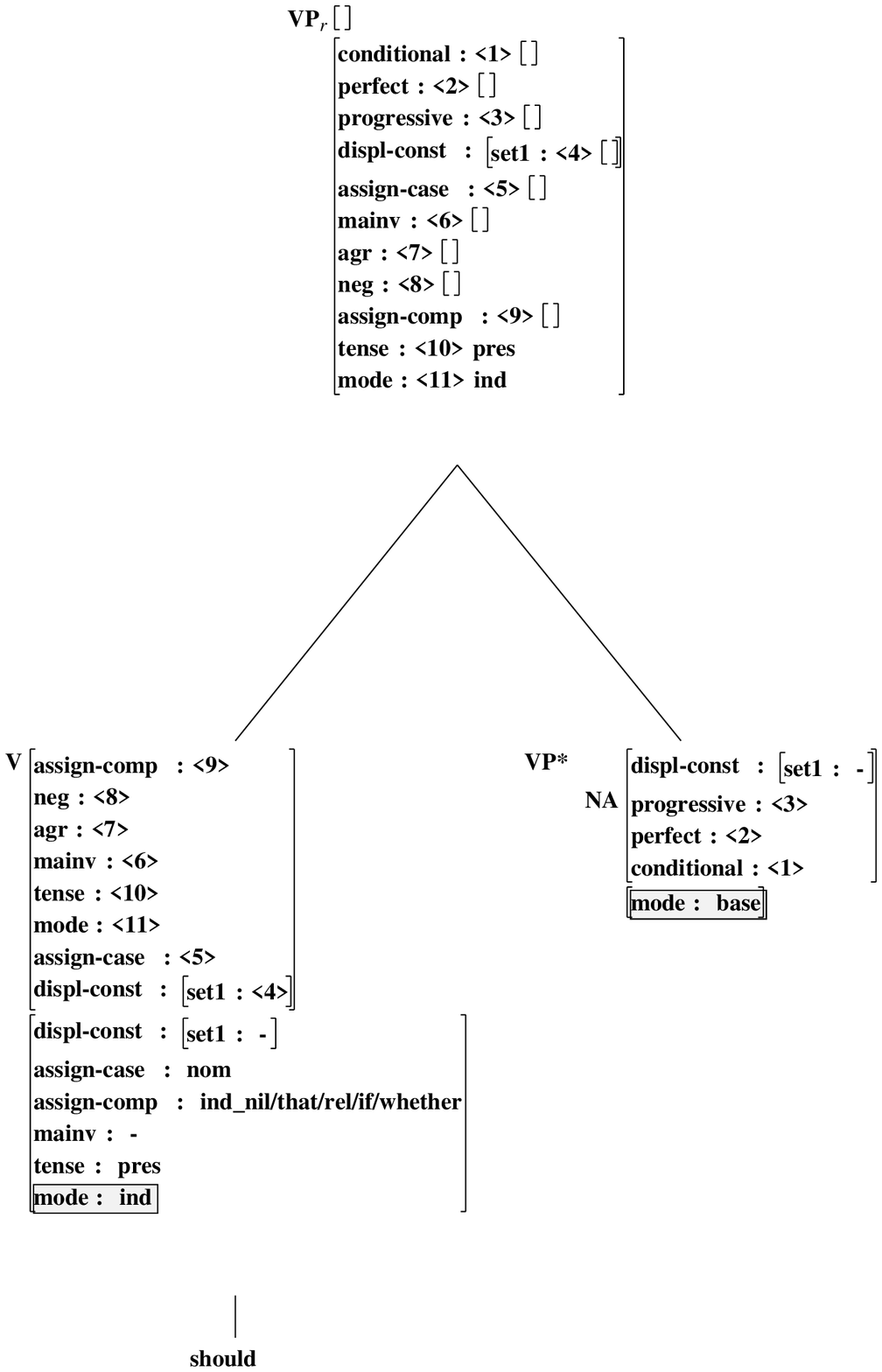,height=3.9in}} &
\hspace*{1in}&
{\psfig{figure=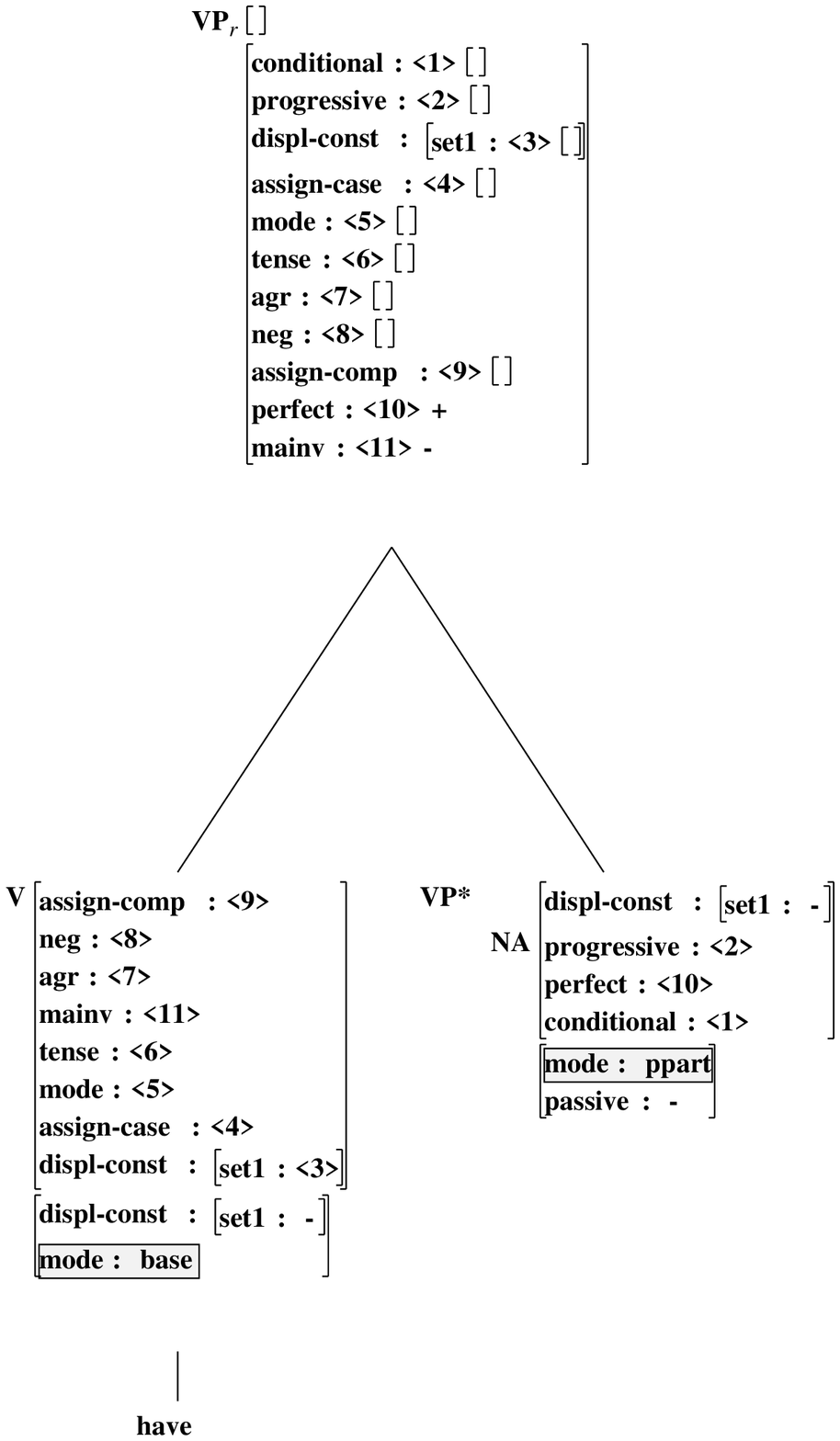,height=3.9in}} \\
\\
{\psfig{figure=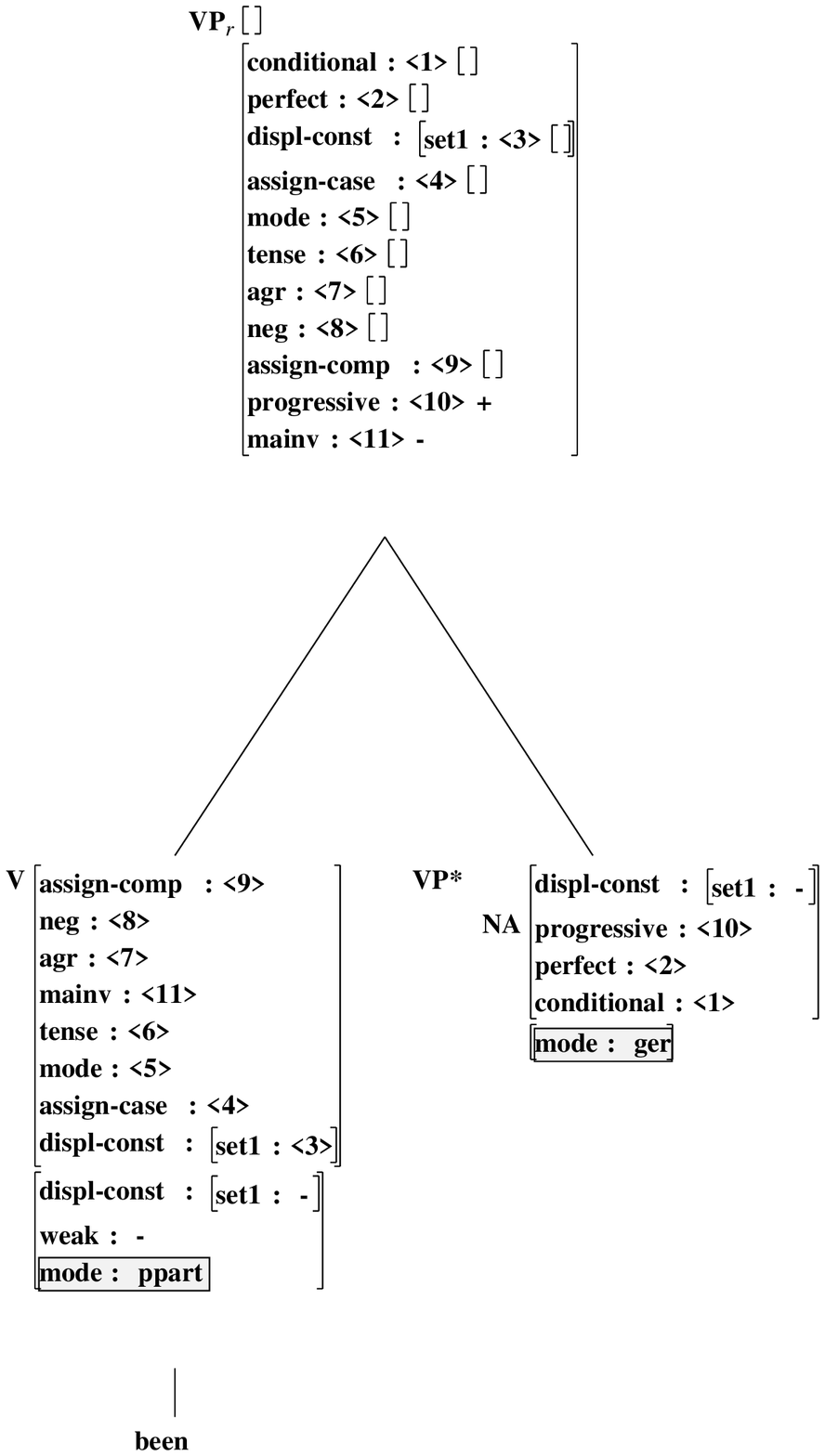,height=3.9in}} &
\hspace*{1in}&
{\psfig{figure=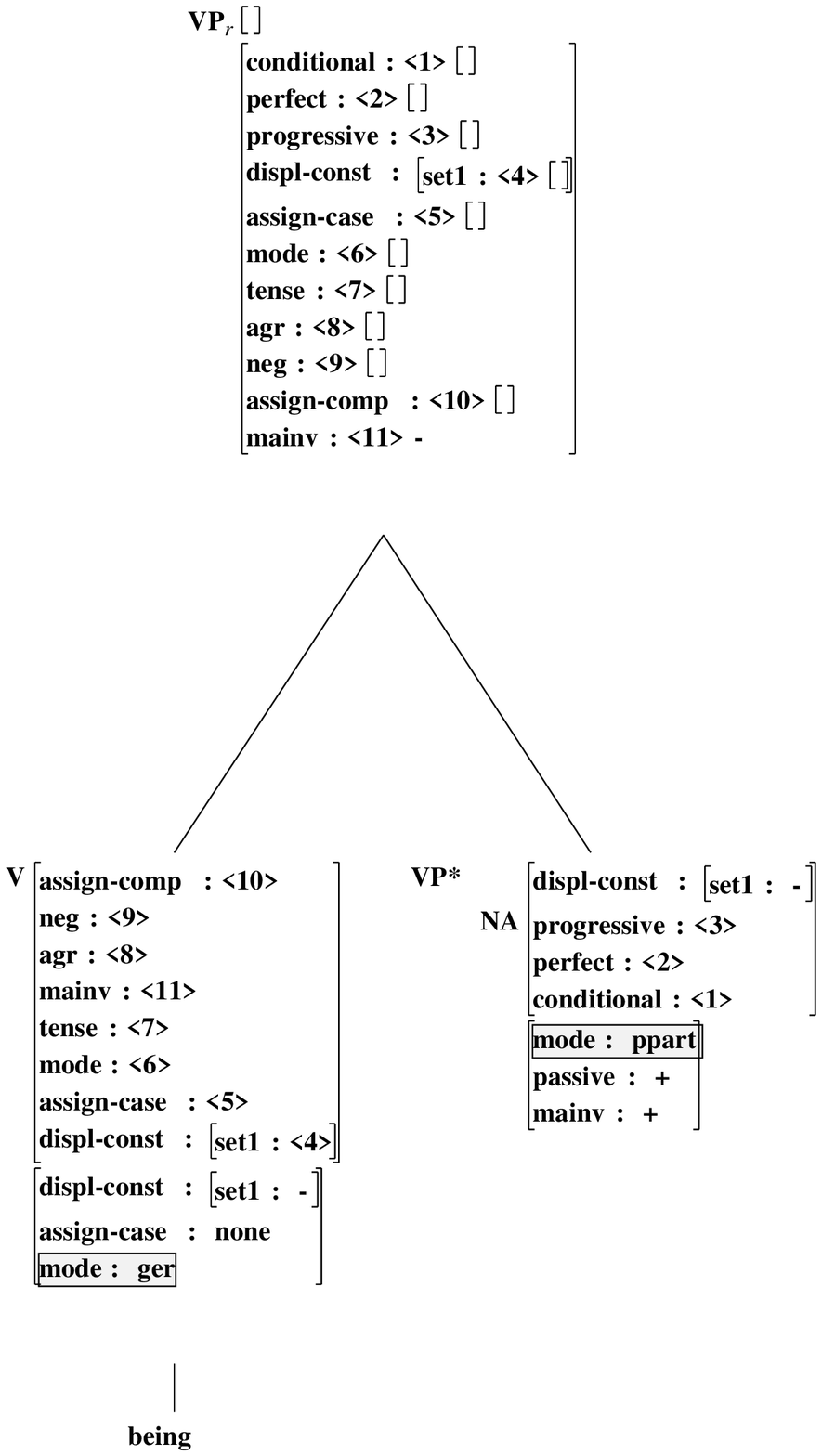,height=3.9in}} \\
\end{tabular}
\caption{Auxiliary trees for {\it The music should have been being played .}}
\label{anchored-aux-trees}
\end{figure}

The restrictions outlined in column 3 of Table \ref{aux-table} are
implemented through the features {\bf $<$mode$>$}, {\bf
$<$perfect$>$}, {\bf $<$progressive$>$} and {\bf $<$passive$>$}.
The syntactic lexicon entries for the auxiliaries gives values for
these features on the foot node~(VP$^{*}$) in Figure~\ref{Vvx}.  Since
the top features of the foot node must eventually unify with the
bottom features of the node it adjoins onto for the sentence to be
valid, this enforces the restrictions made by the auxiliary node.  In
addition to these feature values, each auxiliary also gives values to
the anchoring node~(V$\diamond$), to be passed up the tree to the root
VP~(VP$_{r}$) node; there they will become the new features for the
top VP node of the sentential tree.  Another auxiliary may now adjoin
on top of it, and so forth.  These feature values thereby ensure the
proper auxiliary sequencing.  Figure~\ref{anchored-aux-trees} shows the auxiliary trees anchored by the four 
auxiliary verbs in sentence (\ex{0}).  Figure~\ref{non-inverted-sentence} shows
the final tree created for this sentence.

\begin{figure}[htb]
\centering
\begin{tabular}{c}
{\psfig{figure=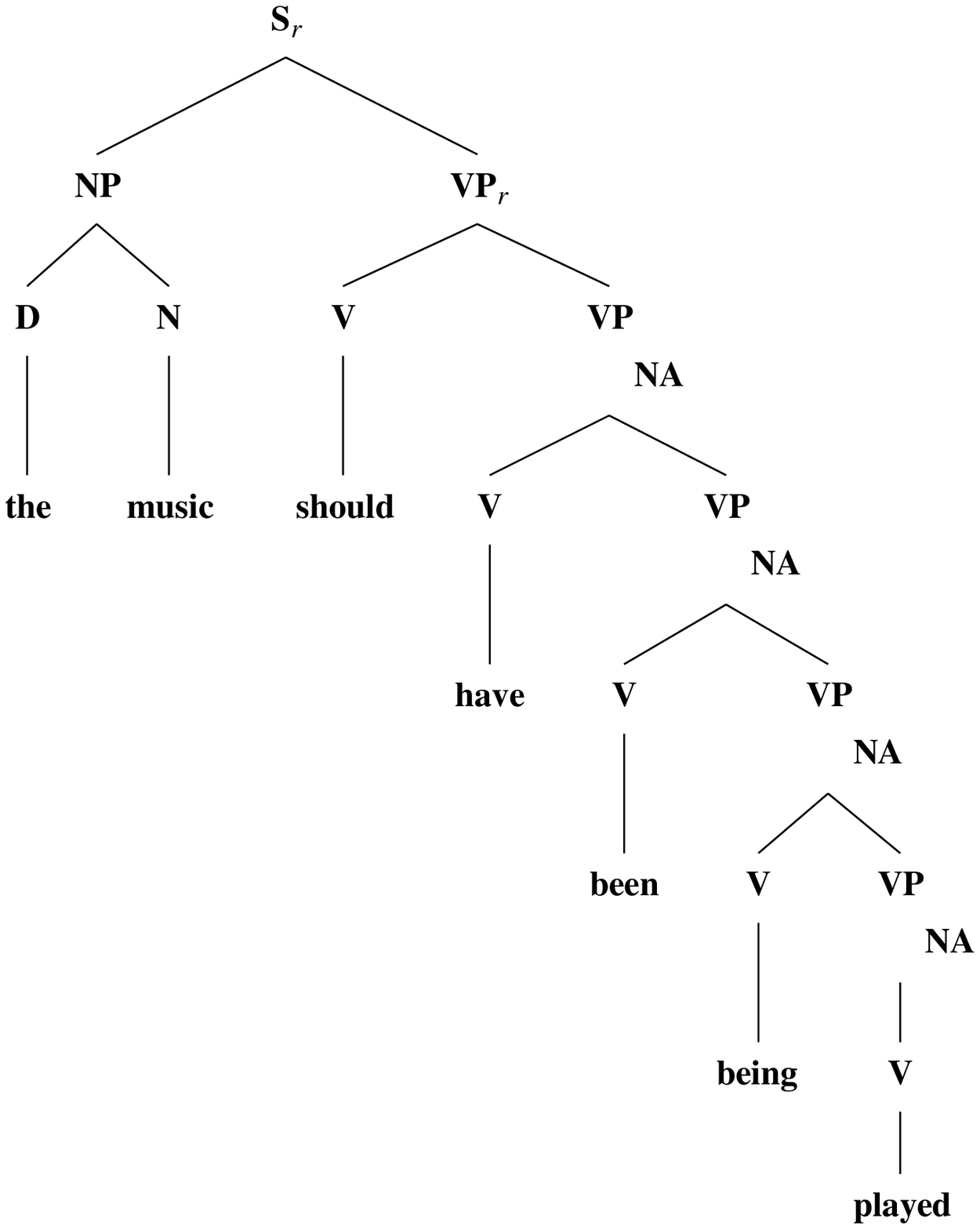,height=5.1in}}
\end{tabular}
\caption{{\it The music should have been being played .}}
\label{non-inverted-sentence}
\end{figure}

The general English restriction that matrix clauses must have tense
(or be imperatives) is enforced by requiring the top S-node of a
sentence to have {\bf $<$mode$>$=ind/imp} (indicative or imperative).
Since only the indicative and imperative sentences have tense,
non-tensed clauses are restricted to occurring in embedded
environments.

Noun-verb contractions are labeled NVC in their part-of-speech field
in the morphological database and then undergo special processing to
split them apart into the noun and the reduced verb before
parsing. The noun then selects its trees in the normal fashion. The
contraction, say {\it 'll} or {\it 'd}, likewise selects the normal
auxiliary verb tree, $\beta$Vvx. However, since the contracted form,
rather than the verb stem, is given in the morphology, the contracted
form must also be listed as a separate syntactic entry. These entries
have all the same features of the full form of the auxiliary verbs,
with tense constraints coming from the morphological entry (e.g. {\it
it's} is listed as {\sc it 's NVC 3sg PRES}). The ambiguous
contractions {\it 'd} ({\it had/would}) and {\it `s} ({\it has/is})
behave like other ambiguous lexical items; there are simply multiple
entries for those lexical items in the lexicon, each with different
features. In the resulting parse, the contracted form is shown with
features appropriate to the full auxiliary it represents.

\section{Inverted Sentences}

In inverted sentences, the two trees shown in Figure~\ref{inverted-trees}
adjoin to an S tree anchored by a main verb.  The tree in
Figure~\ref{inverted-trees}(a) is anchored by the auxiliary verb and adjoins to
the S node, while the tree in Figure~\ref{inverted-trees}(b) is anchored by an
empty element and adjoins at the VP node.  Figure~\ref{yes/no-question} shows
these trees (anchored by {\it will}) adjoined to the declarative transitive
tree\footnote{The declarative transitive tree was seen in
section~\ref{nx0Vnx1-family}.} (anchored by main verb {\it buy}).

\begin{figure}[htbp]
\centering
\begin{tabular}{ccc}
{\psfig{figure=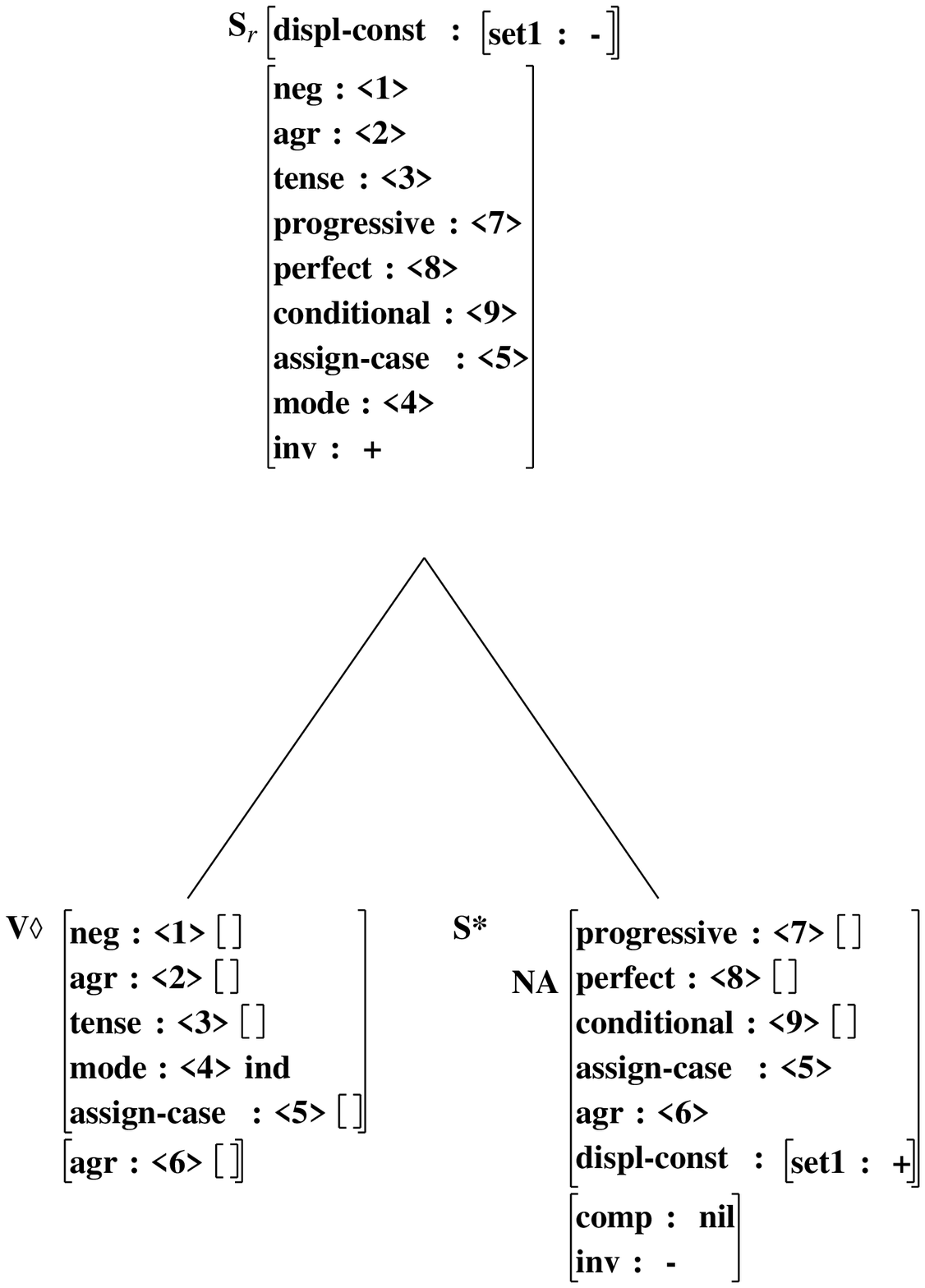,height=4.5in}} &
\hspace*{0.5in} &
{\psfig{figure=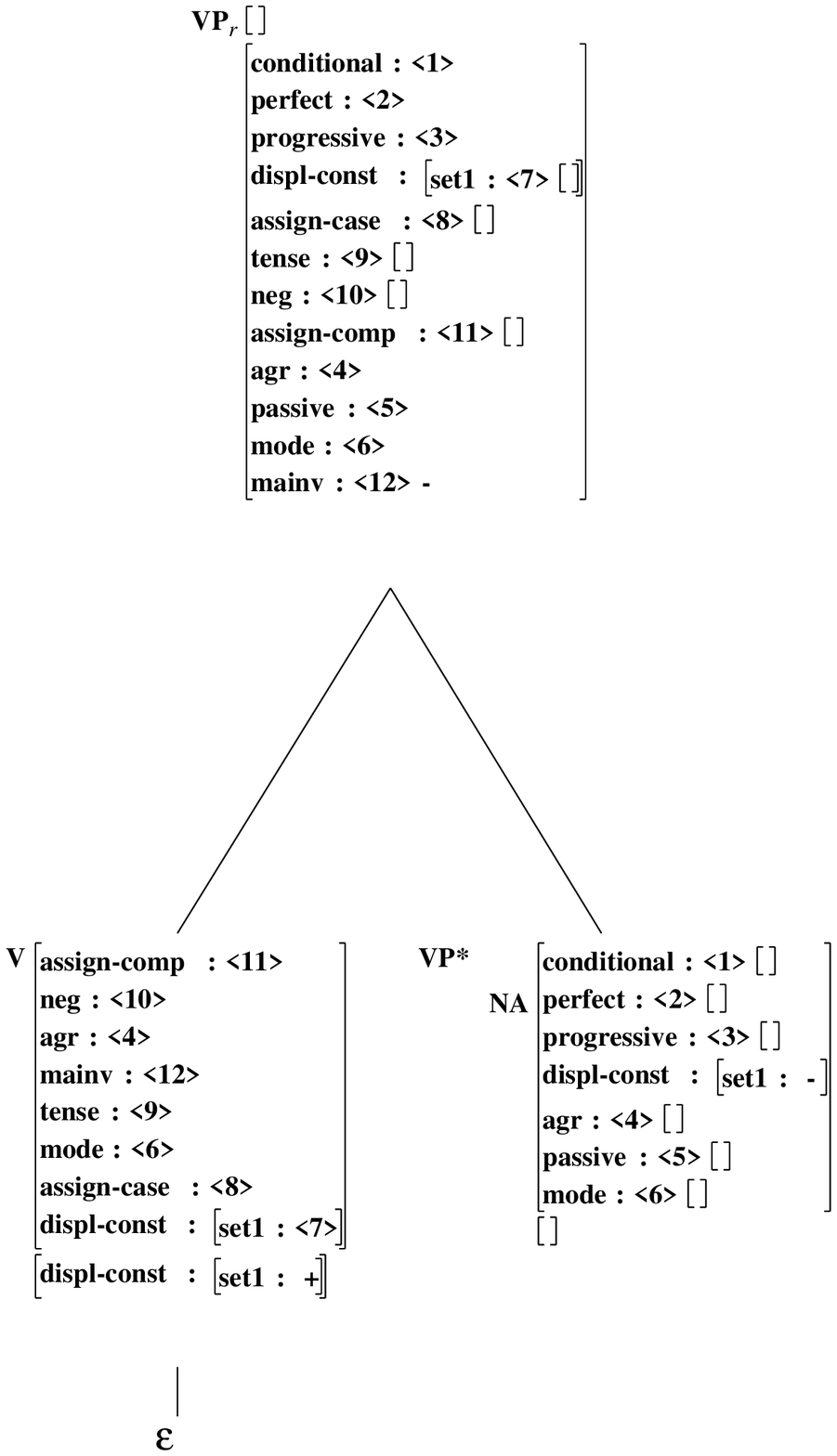,height=5in}} \\
(a) &&(b) \\ 
\end{tabular}
\caption{Trees for auxiliary verb inversion: $\beta$Vs (a) and $\beta$Vvx (b)}
\label{inverted-trees}
\end{figure}

\begin{figure}[htb]
\centering
\begin{tabular}{c}
{\psfig{figure=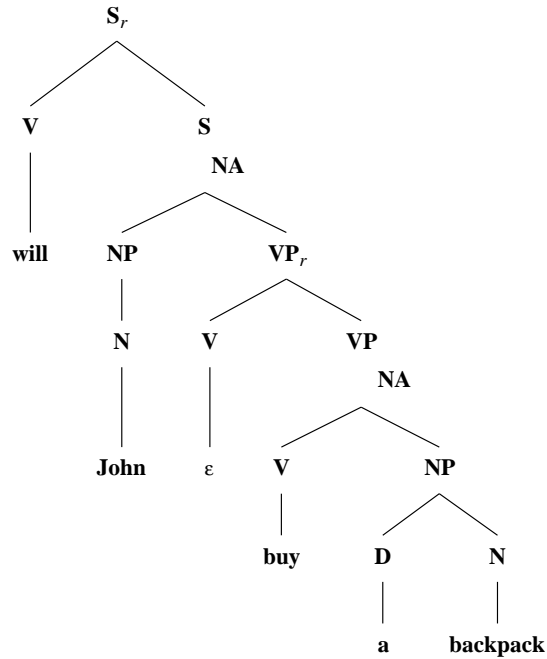,height=4.0in}} \\
\end{tabular}
\caption{{\it will John buy a backpack ?}}
\label{yes/no-question}
\end{figure}

The feature {\bf $<$displ-const$>$} ensures that both of the trees in
Figure~\ref{inverted-trees} must adjoin to an elementary tree whenever one of
them does. For more discussion on this mechanism, which simulates tree local
multi-component adjunction, see \cite{hockeysrini93}.  The tree in
Figure~\ref{inverted-trees}(b), anchored by $\epsilon$, represents the
originating position of the inverted auxiliary. Its adjunction blocks the {\bf
$<$assign-case$>$} values of the VP it dominates from being co-indexed with the
{\bf $<$case$>$} value of the subject. Since {\bf $<$assign-case$>$} values
from the VP are blocked, the {\bf $<$case$>$} value of the subject can only be
co-indexed with the {\bf $<$assign-case$>$} value of the inverted auxiliary
(Figure~\ref{inverted-trees}(a)).  Consequently, the inverted auxiliary
functions as the case-assigner for the subject in these inverted structures.
This is in contrast to the situation in uninverted structures where the anchor
of the highest (leftmost) VP assigns case to the subject (see
section~\ref{case-for-verbs} for more on case assignment).  The XTAG analysis
is similar to GB accounts where the inverted auxiliary plus the
$\epsilon$-anchored tree are taken as representing I to C movement.

\section{Do-Support}

It is well-known that English requires a mechanism called `do-support' for
negated sentences and for inverted yes-no questions without auxiliaries.

\enumsentence {John does not want a car .}
\enumsentence {$\ast$John not wants a car .}
\enumsentence {John will not want a car .}
\enumsentence {Do you want to leave home ?}
\enumsentence {$\ast$want you to leave home ?}
\enumsentence {will you want to leave home ?}

\subsection{In negated sentences}
\label{do-support-negatives}

The GB analysis of do-support in negated sentences hinges on the separation of
the INFL and VP nodes (see \cite{chomsky65}, \cite{jackendoff72} and
\cite{chomsky86}).  The claim is that the presence of the negative morpheme
blocks the main verb from getting tense from the INFL node, thereby forcing the
addition of a verbal lexeme to carry the inflectional elements.  If an
auxiliary verb is present, then it carries tense, but if not, periphrastic or
`dummy', {\it do} is required.  This seems to indicate that {\it do} and other
auxiliary verbs would not co-occur, and indeed this is the case (see sentences
(\ex{1})-(\ex{2})).  Auxiliary {\it do} is allowed in English when no
negative morpheme is present, but this usage is marked as emphatic.  Emphatic
{\it do} is also not allowed to co-occur with auxiliary verbs (sentences
(\ex{3})-(\ex{6})).

\enumsentence {$\ast$We will have do bought a sleeping bag .}
\enumsentence {$\ast$We do will have bought a sleeping bag .}
\enumsentence {You {\bf do} have a backpack, don't you ?}
\enumsentence {I {\bf do} want to go !}
\enumsentence {$\ast$You {\bf do} can have a backpack, don't you ?}
\enumsentence {$\ast$I {\bf did} have had a backpack !}

At present, the XTAG grammar does not have analyses for emphatic {\it do}.

In the XTAG grammar, {\it do} is prevented from co-occurring with other
auxiliary verbs by a requirement that it adjoin only onto main verbs
({\bf $<$mainv$>$ = $+$}).  It has
indicative mode, so no other auxiliaries can adjoin above it.\footnote{Earlier,
we said that indicative mode carries tense with it.  Since only the topmost
auxiliary carries the tense, any subsequent verbs must {\bf not} have
indicative mode.}  The lexical item {\it not} is only allowed to adjoin onto a
non-indicative (and therefore non-tensed) verb.  Since all matrix clauses must
be indicative (or imperative), a negated sentence will fail unless an auxiliary
verb, either {\it do} or another auxiliary, adjoins somewhere above the
negative morpheme, {\it not}. In addition to forcing adjunction of an
auxiliary, this analysis of {\it not} allows it freedom to move around in the
auxiliaries, as seen in the sentences (\ex{1})-(\ex{4}).

\enumsentence {John will have had a backpack .}
\enumsentence {$\ast$John not will have had a backpack .}
\enumsentence {John will not have had a backpack .}
\enumsentence {John will have not had a backpack .}

\subsection{In inverted yes/no questions}

In inverted yes/no questions, {\it do} is required if there is no auxiliary
verb to invert, as seen in sentences (\ex{-12})-(\ex{-10}), replicated here
as (\ex{1})-(\ex{3}).

\enumsentence {do you want to leave home ?}
\enumsentence {$\ast$want you to leave home ?}
\enumsentence {will you want to leave home ?}
\enumsentence {$\ast$do you will want to leave home ?}

In English, unlike other Germanic languages, the main verb cannot move to the
beginning of a clause, with the exception of main verb {\it be}.\footnote{The
inversion of main verb {\it have} in British English was previously noted.}  In
a GB account of inverted yes/no questions, the tense feature is said to be in
C$^{0}$ at the front of the sentence.  Since main verbs cannot move, they
cannot pick up the tense feature, and do-support is again required if there is
no auxiliary verb to perform the role.  Sentence (\ex{0}) shows that {\it do}
does not interact with other auxiliary verbs, even when in the inverted
position.

In XTAG, trees anchored by a main verb that lacks tense are required to have an
auxiliary verb adjoin onto them, whether at the VP node to form a declarative
sentence, or at the S node to form an inverted question.  {\it Do} selects the
inverted auxiliary trees given in Figure~\ref{inverted-trees}, just as other
auxiliaries do, so it is available to adjoin onto a tree at the S node to form
a yes/no question.  The mechanism described in
section~\ref{do-support-negatives} prohibits {\it do} from co-occurring with
other auxiliary verbs, even in the inverted position.

\section{Infinitives}

The infinitive {\it to} is considered an auxiliary verb in the XTAG system, and
selects the auxiliary tree in Figure~\ref{Vvx}.  {\it To}, like {\it do}, does
not interact with the other auxiliary verbs, adjoining only to main verb base
forms, and carrying infinitive mode.  It is used in embedded clauses, both with
and without a complementizer, as in sentences (\ex{1})-(\ex{3}).  Since it
cannot be inverted, it simply does not select the trees in
Figure~\ref{inverted-trees}.

\enumsentence {John wants to have a backpack .}
\enumsentence {John wants Mary to have a backpack .}
\enumsentence {John wants for Mary to have a backpack .}

The usage of infinitival {\em to} interacts closely with the distribution of
null subjects (PRO), and is described in more detail in
section~\ref{for-complementizer}.

\section{Semi-Auxiliaries}

Under the category of semi-auxiliaries, we have placed several verbs that do
not seem to closely follow the behavior of auxiliaries.  One of these 
auxiliaries, {\it dare}, mainly behaves as a modal and selects for the base 
form of the verb.  The other semi-auxiliaries all select for the infinitival 
form of the verb.  Examples of this second type of semi-auxiliary are {\it 
used to}, {\it ought to}, {\it get to}, {\it have to}, and {\it BE to}.  

\subsection{Marginal Modal {\it dare}}

The auxiliary {\it dare} is unique among modals in that it both allows
DO-support and exhibits a past tense form.  It clearly falls in modal position
since no other auxiliary (except {\it do}) may precede it in linear
order\footnote{Some speakers accept {\it dare} preceded by a modal, as in {\it
I might dare finish this report today}.  In the XTAG analysis, this particular
double modal usage is accounted for.  Other cases of double modal occurrence
exist in some dialects of American English,  although these are not accounted
for in the system, as was mentioned earlier.\label{dare-footnote}}.  Examples appear below.

\enumsentence{she {\bf dare} not have been seen .}
\enumsentence{she does not {\bf dare} succeed .}
\enumsentence{Jerry {\bf dared} not look left or right .}
\enumsentence{only models {\bf dare} wear such extraordinary outfits .}
\enumsentence{{\bf dare} Dale tell her the secret ?}
\enumsentence{$\ast$Louise had dared not tell a soul .}

As mentioned above, auxiliaries are prevented from having DO-support within the
XTAG system.  To allow for DO-support in this case, we had to create a lexical
entry for {\it dare} that allowed it to have the feature {\bf
mainv+} and to have {\bf base} mode (this measure is
what also allows {\it dare} to occur in double-modal sequences).  A second
lexical entry was added to handle the regular modal occurrence of {\it dare}.
Additionally, all other modals are classified as being present tense, while
{\it dare} has both present and past forms.  To handle this behavior, {\it
dare} was given similar features to the other modals in the morphology minus
the specification for tense. 

\subsection{Other semi-auxiliaries}

The other semi-auxiliaries all select for the infinitival form of the verb.
Many of these auxiliaries allow for DO-support and can appear in both base and
past participle forms, in addition to being able to stand alone (indicative 
mode).  Examples of this type appear below.

\enumsentence{Alex {\bf used} to attend karate workshops .}
\enumsentence{Angelina might have {\bf used} to believe in fate .}
\enumsentence{Rich did not {\bf used} to want to be a physical therapist .}
\enumsentence{Mick might not {\bf have} to play the game tonight .}
\enumsentence{Singer {\bf had} to have been there .}
\enumsentence{Heather has {\bf got} to finish that project before she goes
insane .}

The auxiliaries {\it ought to} and {\it BE to} may not be preceded by any other
auxiliary.  

\enumsentence{Biff {\bf ought} to have been working harder .}
\enumsentence{$\ast$Carson does {\bf ought} to have been working harder .}
\enumsentence{the party {\bf is} to take place this evening .}
\enumsentence{$\ast$the party had {\bf been} to take place this evening .}

The trickiest element in this group of auxiliaries is {\it used to}.  While the
other verbs behave according to standard inflection for auxiliaries, {\it used
to} has the same form whether it is in mode base, past participle, or
indicative forms.  The only connection {\it used to} maintains with the
infinitival form {\it use} is that occasionally, the bare form {\it use} will
appear with DO-support.  Since the three modes mentioned above are mutually
exclusive in terms of both the morphology and the lexicon, {\it used} has three
entries in each.  

\subsection{Other Issues}

There is a lingering problem with the auxiliaries that stems from the fact that
there currently is no way to distinguish between the main verb and auxiliary verb
behaviors for a given letter string within the morphology.  This situation
results in many unacceptable sentences being successfully parsed by the system.
Examples of the unacceptable sentences are given below.

\enumsentence{the miller {\bf cans} tell a good story .  (vs  the farmer {\bf
cans} peaches in July .)}
\enumsentence{David {\bf wills} have finished by noon .  (vs  the old man {\bf
wills} his fortune to me .)}
\enumsentence{Sarah {\bf needs} not leave .  (vs  Sarah {\bf needs} to leave .)}
\enumsentence{Jennifer {\bf dares} not be seen .  (vs  the young woman {\bf
dares} him to do the stunt .)}
\enumsentence{Lila {\bf does use} to like beans .  (vs  Lila {\bf does use} her
new cookware .)}

\chapter{Conjunction}
\label{conjunction}

\section{Introduction}

The XTAG system can handle sentences with conjunction of two
constituents of the same syntactic category. The coordinating
conjunctions which select the conjunction trees are {\it and}, {\it
or} and {\it but}.\footnote{We believe that the restriction of {\it
but} to conjoining only two items is a pragmatic one, and our grammars
accepts sequences of any number of elements conjoined by {\it but}.}
There are also multi-word conjunction trees, anchored by {\it either-or}, 
{\it neither-nor}, {\it both-and}, and {\it as well as}.  There are eight
syntactic categories that can be coordinated, and in each case an
auxiliary tree is used to implement the conjunction.  These eight
categories can be considered as four different cases, as described in
the following sections.  In all cases the two constituents are
required to be of the same syntactic category, but there may also be
some additional constraints, as described below.

\section{Adjective, Adverb, Preposition and PP Conjunction}

Each of these four categories has an auxiliary tree that is used for
conjunction of two constituents of that category.  The auxiliary tree
adjoins into the left-hand-side component, and the right-hand-side
component substitutes into the auxiliary tree.  

\begin{figure}[htb]
\centering
\begin{tabular}{ccc}
{\psfig{figure=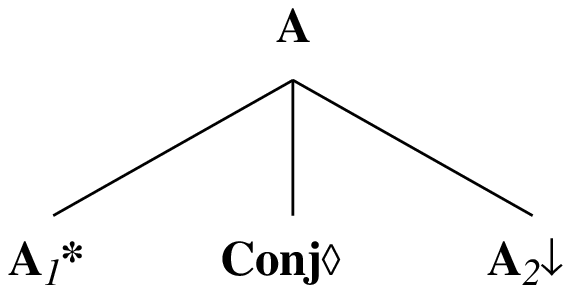,height=1in}}&
\hspace*{0.5in}&
{\psfig{figure=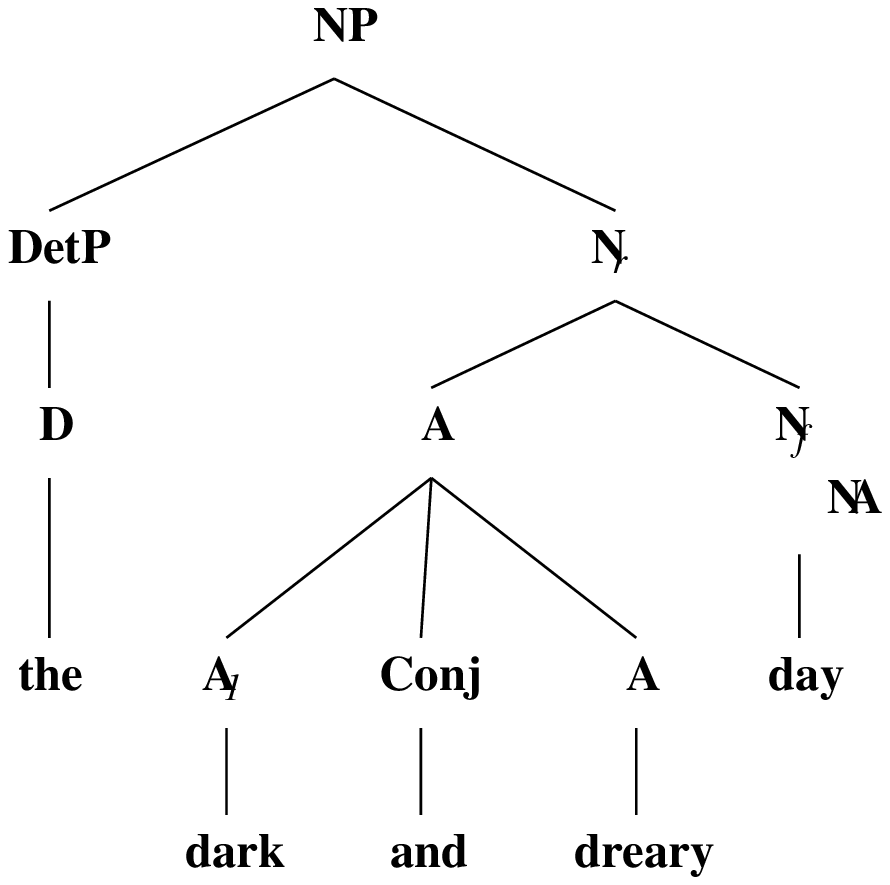,height=2.8in}}\\
(a) & \hspace*{0.5in}& (b)\\
\end{tabular}
\caption{Tree for adjective conjunction: $\beta$a1CONJa2 and a resulting parse tree}
\label{A1conjA2}
\end{figure}

Figure~\ref{A1conjA2}(a) shows the auxiliary tree for adjective conjunction,
and is used, for example, in the derivation of the parse tree for the noun
phrase {\it the dark and dreary day}, as shown in Figure~\ref{A1conjA2}(b).
The auxiliary tree adjoins onto the node for the left adjective, and the
right adjective substitutes into the right hand side node of the auxiliary
tree. The analysis for adverb, preposition and PP conjunction is exactly the
same and there is a corresponding auxiliary tree for each of these that is
identical to that of Figure~\ref{A1conjA2}(a) except, of course, for the node
labels.

\section{Noun Phrase and Noun Conjunction}

The tree for NP conjunction, shown in Figure~\ref{NP1conjNP2}(a), has
the same basic analysis as in the previous section except that the
{\bf $<$wh$>$} and {\bf $<$case$>$} features are used to force the two
noun phrases to have the same {\bf $<$wh$>$} and {\bf $<$case$>$}
values.  This allows, for example, {\it he and she wrote the book
together} while disallowing {\it $\ast$he and her wrote the book
together.}  Agreement is lexicalized, since the various conjunctions
behave differently. With {\it and}, the root {\bf $<$agr num$>$} value
is {\bf $<$plural$>$}, no matter what the number of the two
conjuncts. With {\it or}, however, the root {\bf $<$agr num$>$} is
co-indexed with the {\bf $<$agr num$>$} feature of the right
conjunct. This ensures that the entire conjunct will bear the number
of both conjuncts if they agree (Figure~\ref{NP1conjNP2}(b)), or of
the most ``recent'' one if they differ ({\it Either the boys or John
is going to help you.}). There is no rule per se on what the
agreement should be here, but people tend to make the verb agree with
the last conjunct (cf. \cite{quirk85}, section 10.41
for discussion). The tree for N conjunction is identical to that for
the NP tree except for the node labels. (The multi-word conjunctions
do not select the N conjunction tree - {\it $^*$the both dogs and
cats}).

\begin{figure}[htb]
\centering
\begin{tabular}{cc}
{\psfig{figure=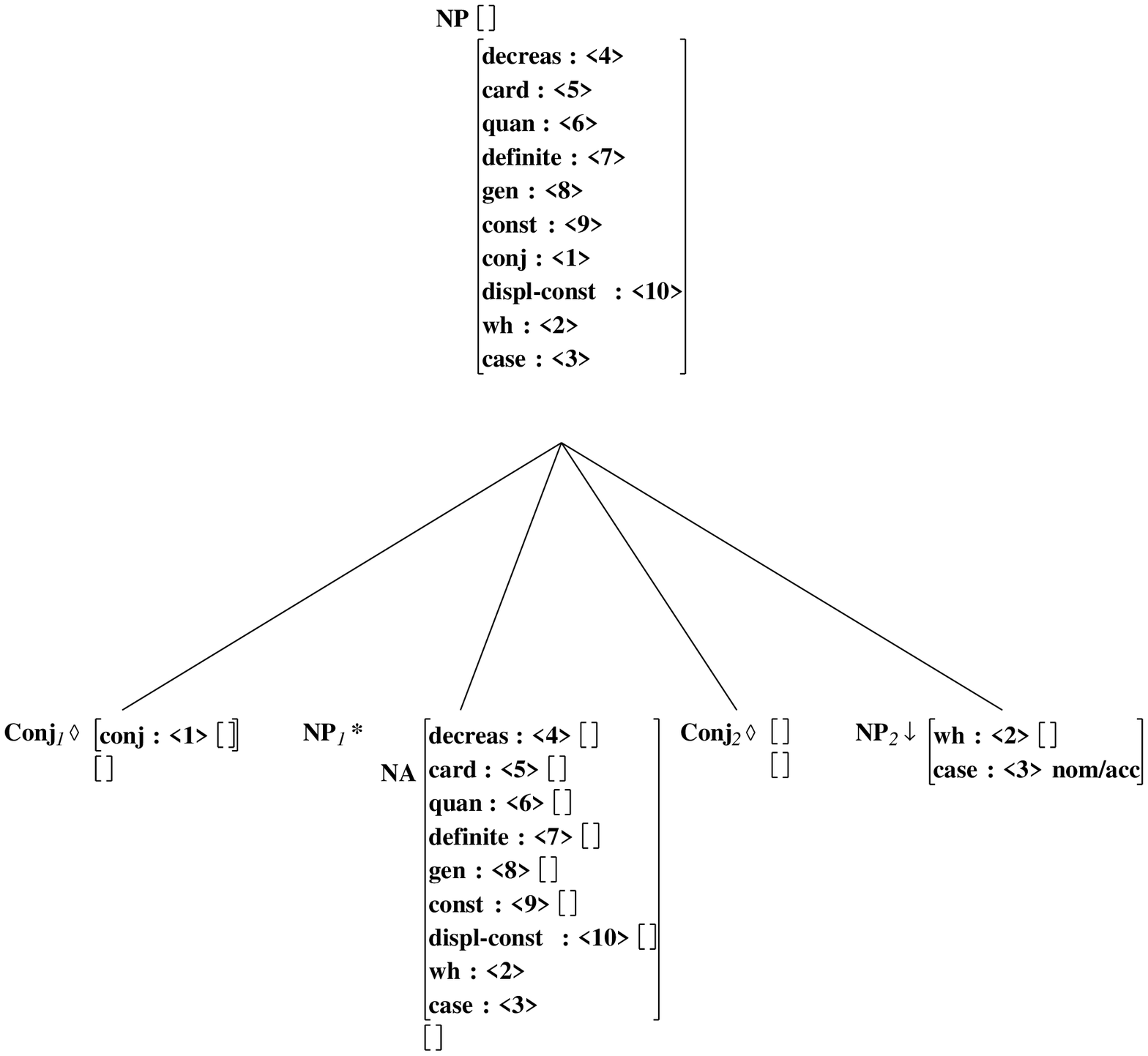,height=4in}}
\hspace{0.5cm} &
{\psfig{figure=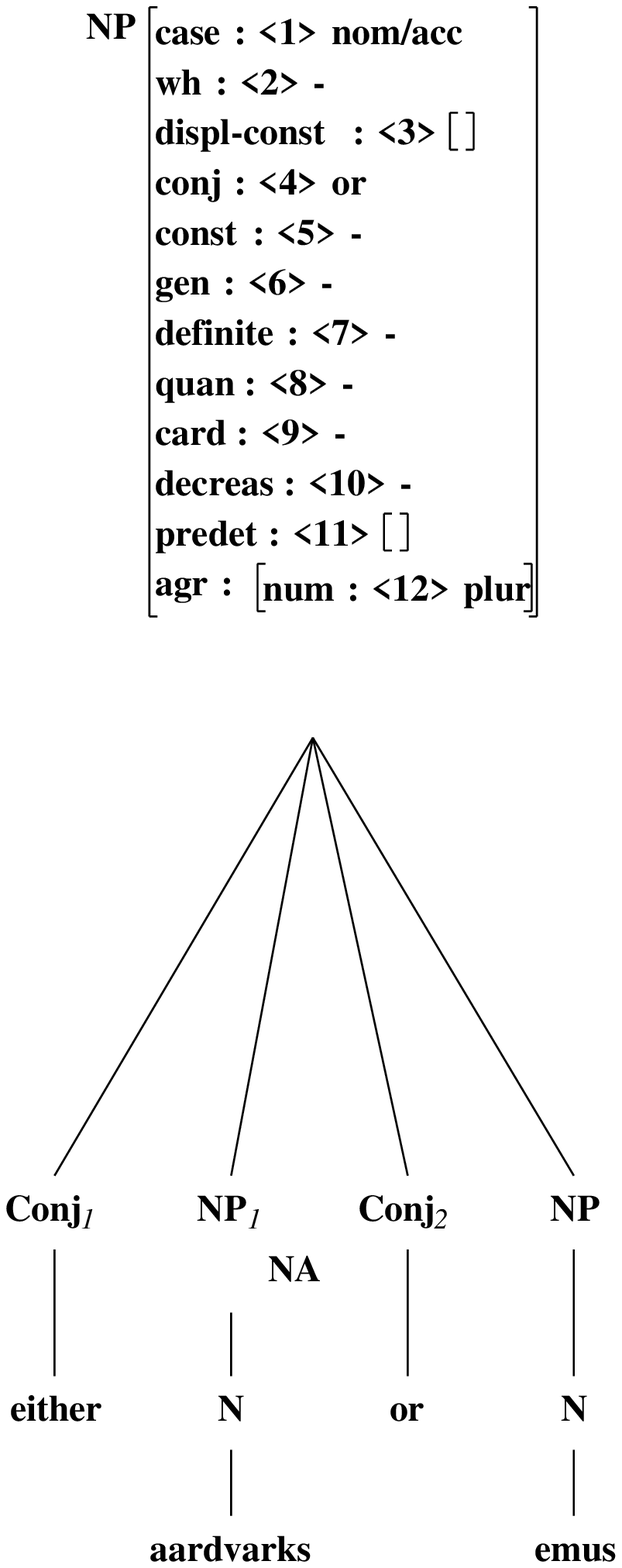,height=4in}}\\
(a) &  (b)\\
\end{tabular}
\caption{Tree for NP conjunction: $\beta$CONJnx1CONJnx2 and a resulting
parse tree}
\label{NP1conjNP2}
\end{figure}

\section{Determiner Conjunction}

In determiner coordination, all of the determiner feature values are
taken from the left determiner, and the only requirement is that the
{\bf $<$wh$>$} feature is the same, while the other features, such as
{\bf $<$card$>$}, are unconstrained.  For example, {\it which and
what} and {\it all but one} are both acceptable determiner
conjunctions, but {\it $\ast$which and all} is not.

\enumsentence{how many and which people camp frequently ?}
\enumsentence{$^*$some or which people enjoy nature .}

\begin{figure}[htbp]
\centering
\begin{tabular}{c}
\psfig{figure=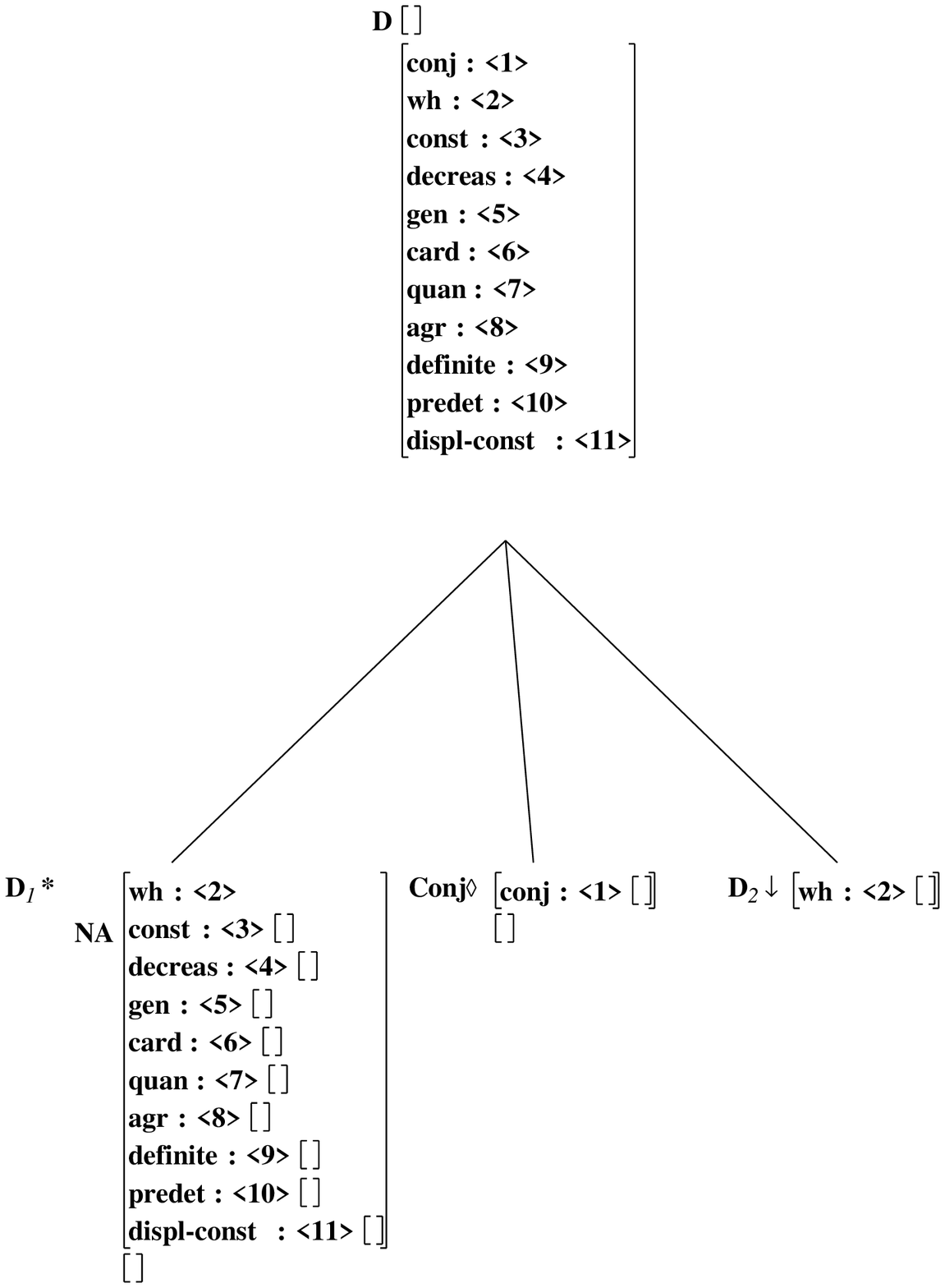,height=5.3in}
\end{tabular}
\vspace{-0.25in}
\caption{Tree for determiner conjunction: $\beta$d1CONJd2.ps}
\label{DX1conjDX2}
\end{figure}

\section{Sentential Conjunction}

The tree for sentential conjunction, shown in Figure~\ref{S1conjS2},
is based on the same analysis as the conjunctions in the previous two
sections, with a slight difference in features.  The {\bf $<$mode$>$}
feature\footnote{See section~\ref{s-features} for an explanation of
the {\bf $<$mode$>$} feature.}  is used to constrain the two sentences
being conjoined to have the same mode so that {\it the day is dark and
the phone never rang} is acceptable, but {\it $\ast$the day dark and
the phone never rang} is not. Similarly, the two sentences must agree
in their {\bf $<$wh$>$}, {\bf $<$comp$>$} and {\bf $<$extracted$>$}
features.  Co-indexation of the {\bf $<$comp$>$} feature ensures that
either both conjuncts have the same complementizer, or there is a
single complementizer adjoined to the complete conjoined S.  The {\bf
$<$assign-comp$>$} feature\footnote{See
section~\ref{for-complementizer} for an explanation of the {\bf
$<$assign-comp$>$} feature.} feature is used to allow conjunction of
infinitival sentences, such as {\it to read and to sleep is a good
life}.

\begin{figure}[htb]
\centering
\begin{tabular}{c}
\psfig{figure=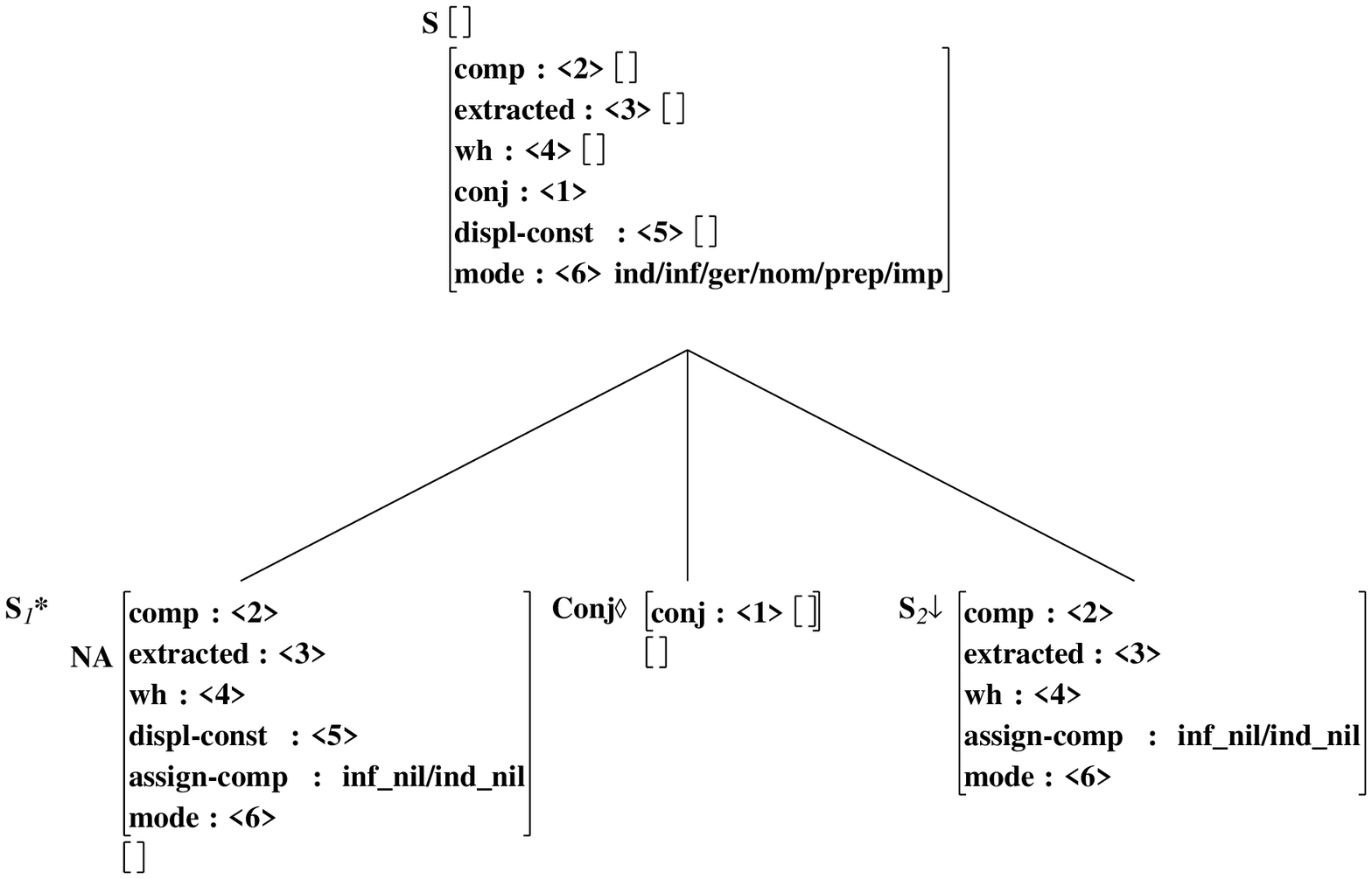,height=3.5in}
\end{tabular}
\caption{Tree for sentential conjunction: $\beta$s1CONJs2}
\label{S1conjS2}
\end{figure}

\section{Comma as a conjunction}

We treat comma as a conjunction in conjoined lists. It anchors the
same trees as the lexical conjunctions, but is considerably more
restricted in how it combines with them. The trees anchored by commas
are prohibited from adjoining to anything but another comma conjoined
element or a non-coordinate element. (All scope possibilities are
allowed for elements coordinated with lexical conjunctions.) Thus,
structures such as Tree
\ref{Comma-conj}(a) are permitted, with each element stacking
sequentially on top of the first element of the conjunct, while
structures such as Tree \ref{Comma-conj}(b) are blocked. 

\begin{figure}[htb]
\centering
\begin{tabular}{ccc}
{\psfig{figure=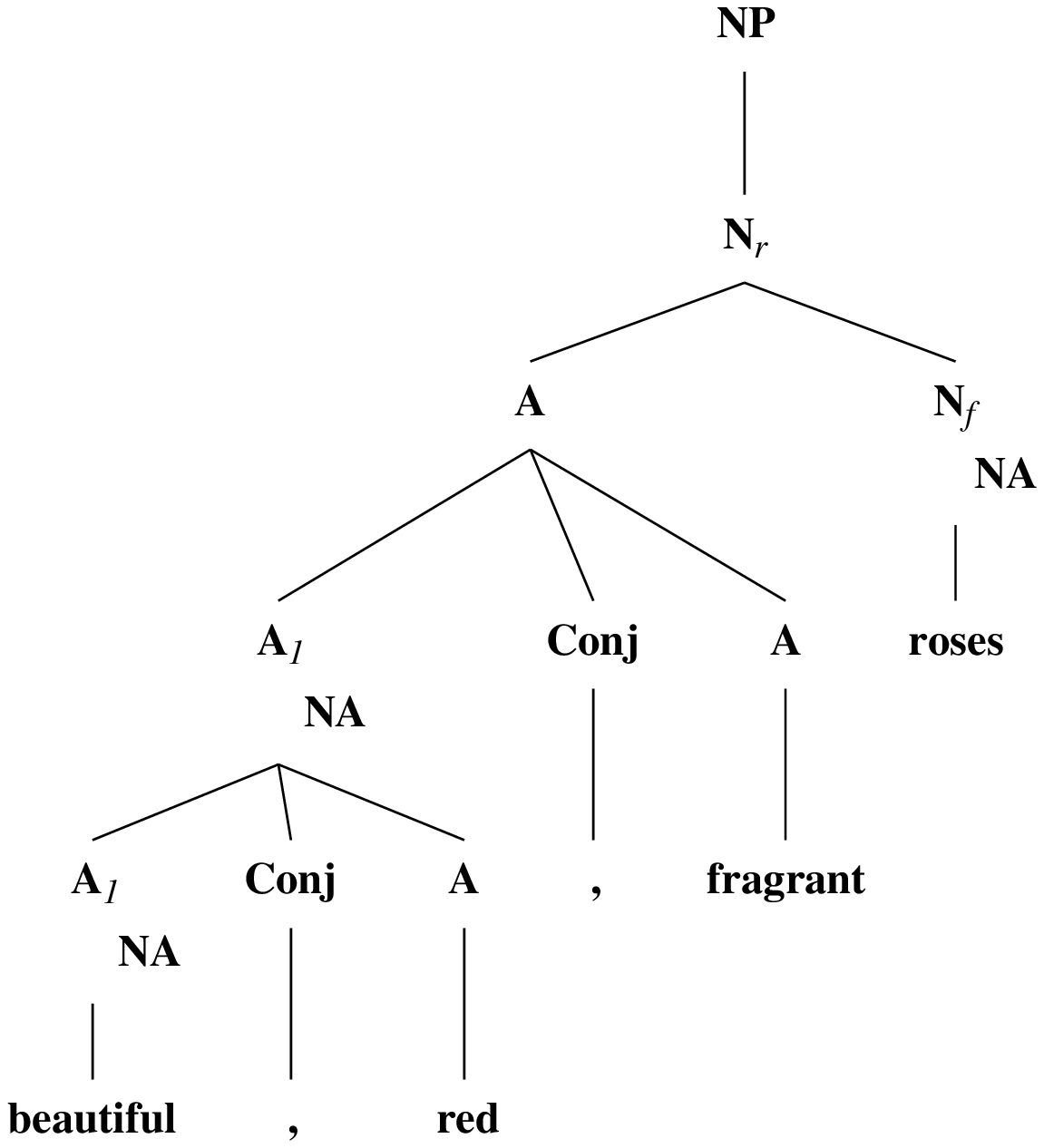,height=2.75in}}&
\hspace*{0.5in}&
{\psfig{figure=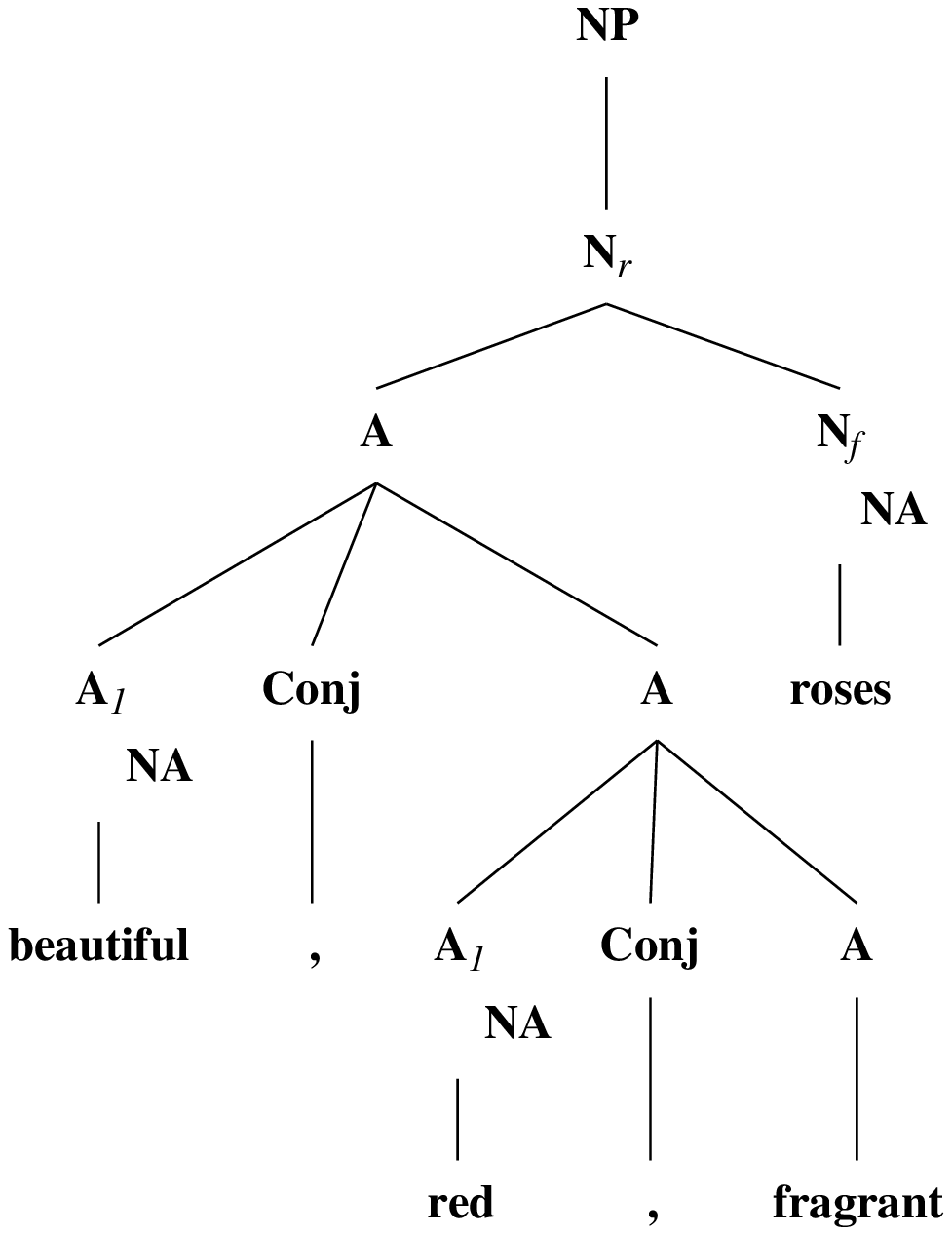,height=2.75in}}\\
(a) Valid tree with comma conjunction & \hspace*{0.5in}& (b) Invalid tree\\
\end{tabular}
\caption{}
\label{Comma-conj}
\end{figure}

This is accomplished by using the {\bf $<$conj$>$} feature, which has the
values {\bf and/or/but} and {\bf comma} to differentiate the lexical
conjunctions from commas. The {\bf $<$conj$>$} values for a comma-anchored
tree and {\it and}-anchored tree are shown in Figure
\ref{conj-contrast}. The feature {\bf $<$conj$>$ = comma/none} on
A$_1$ in (a) only allows comma conjoined or non-conjoined elements as
the left-adjunct, and {\bf $<$conj$>$ = none} on A in (a) allows
only a non-conjoined element as the right conjunct. We also need the
feature {\bf $<$conj$>$ = and/or/but/none} on the right conjunct of
the trees anchored by lexical conjunctions like (b), to block
comma-conjoined elements from substituting there. Without this
restriction, we would get multiple parses of the NP in Tree
\ref{Comma-conj}; with the restrictions we only get the derivation
with the correct scoping, shown as (a).

Since comma-conjoined lists can appear without a lexical conjunction
between the final two elements, as shown in example (\ex{1}), we cannot
force all comma-conjoined sequences to end with a lexical conjunction.

\enumsentence{So it is too with many other spirits which we all know: the
spirit of Nazism or Communism, school spirit , the spirit of a street
corner gang or a football team, the spirit of Rotary or the Ku Klux
Klan. \hfill [Brown cd01]}

\begin{figure}[htb]
\centering
\begin{tabular}{cc}
{\psfig{figure=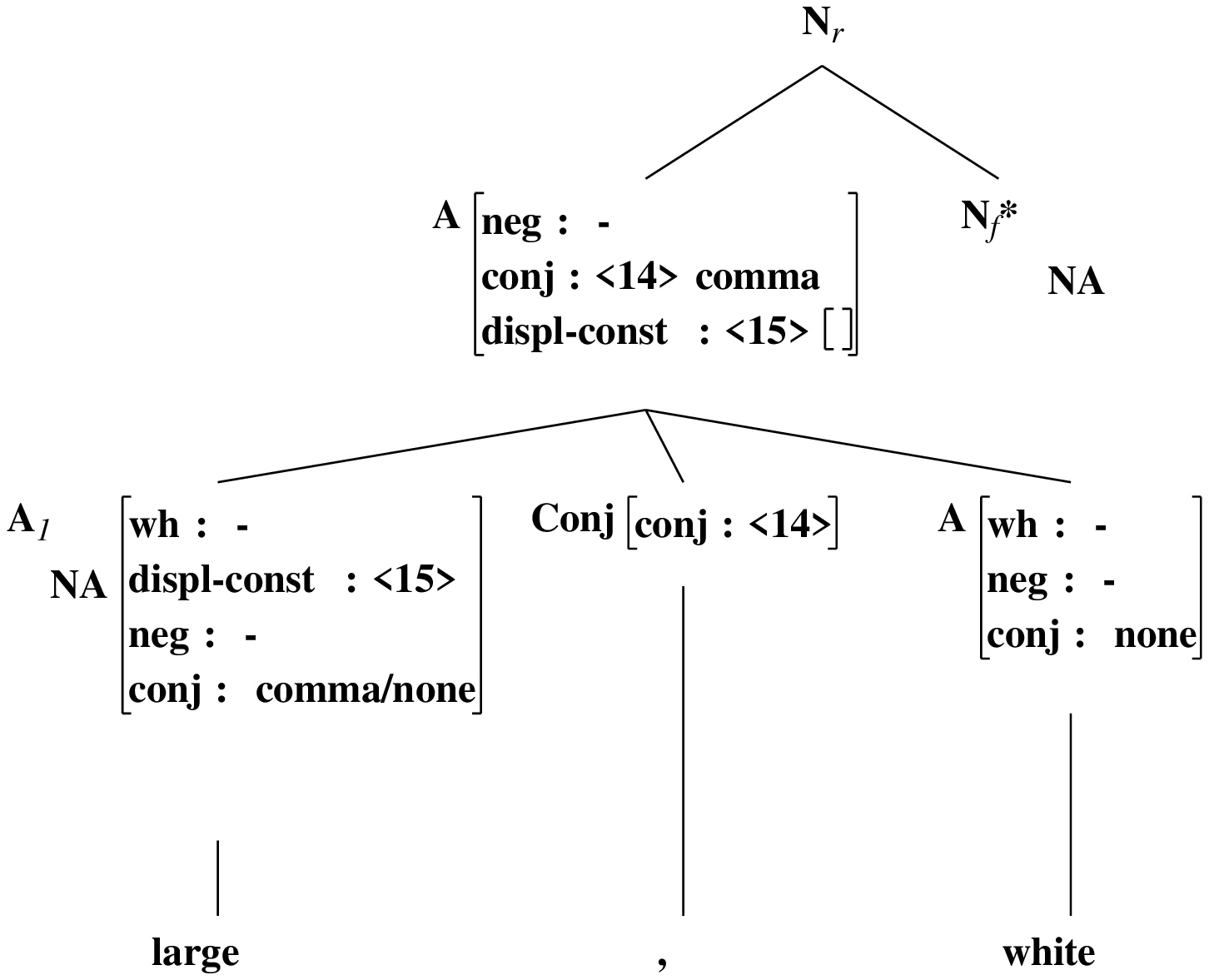,height=2.5in}}&
{\psfig{figure=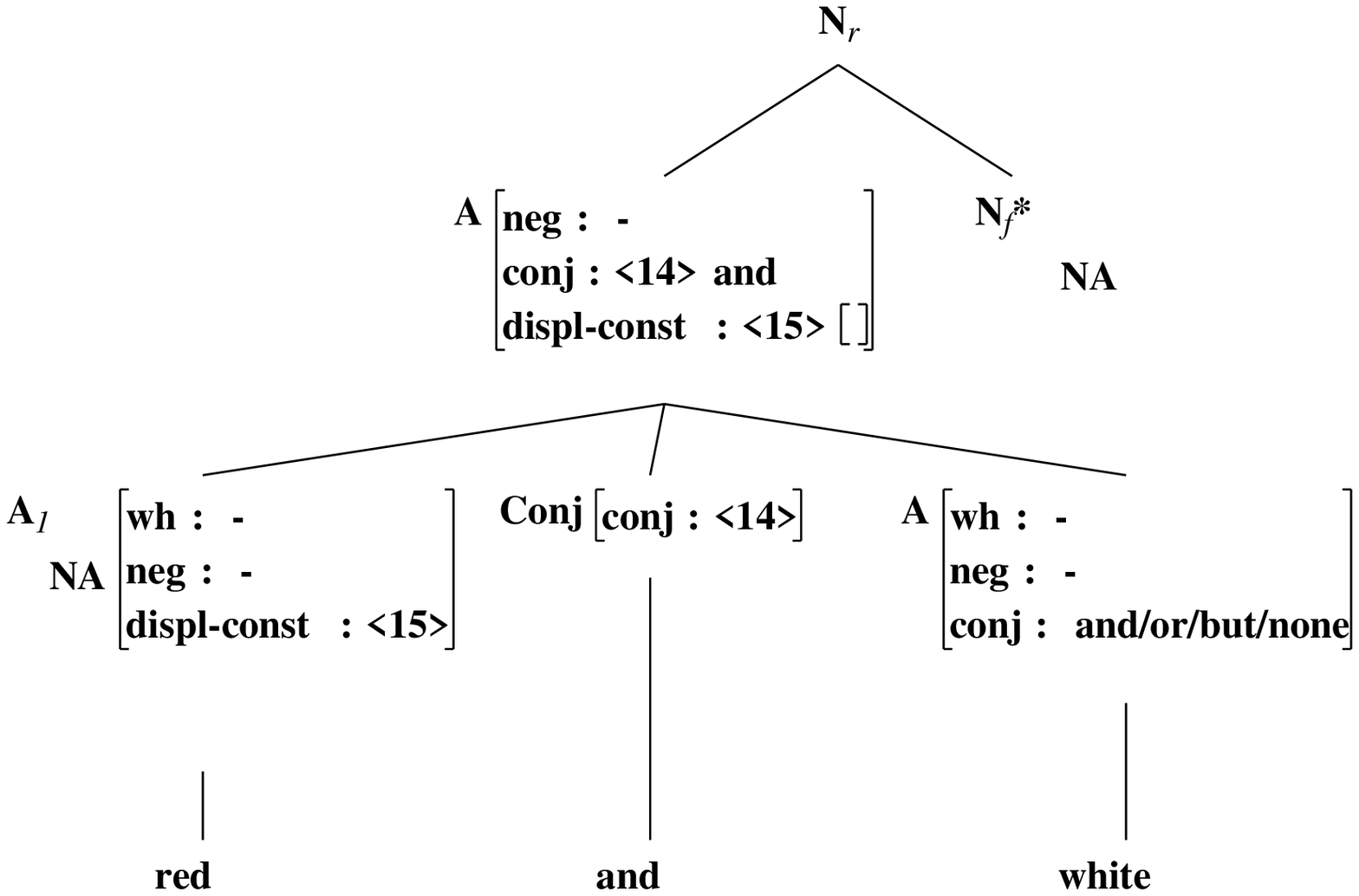,height=2.5in}}\\
\end{tabular}
\caption{$\beta$a1CONJa2 (a) anchored by comma and (b) anchored by {\it and}}
\label{conj-contrast}
\end{figure}

\section{{\it But-not}, {\it not-but}, {\it and-not} and  {\it
$\epsilon$-not}}

We are analyzing conjoined structures such as {\it The women but not
the men} with a multi-anchor conjunction tree anchored by the
conjunction plus the adverb {\it not}. The alternative is to allow
{\it not} to adjoin to any constituent. However, this is the only
construction where {\it not} can freely occur onto a constituent other
than a VP or adjective (cf. $\beta$NEGvx and $\beta$NEGa trees). It
can also adjoin to some determiners, as discussed in Section
\ref{det-comparitives}. We want to allow sentences like (\ex{1}) and
rule out those like (\ex{2}). The tree for the good example is shown
in Figure \ref{but-not}. There are similar trees for {\it and-not} and
{\it $\epsilon$-not}, where $\epsilon$ is interpretable as either {\it
and} or {\it but}, and a tree with {\it not} on the first conjunct for
{\it not-but}.

\enumsentence{Beth grows basil in the house (but) not in the garden .}
\enumsentence{$^*$Beth grows basil (but) not in the garden .}

\begin{figure}[htb]
\centering
\begin{tabular}{c}
\psfig{figure=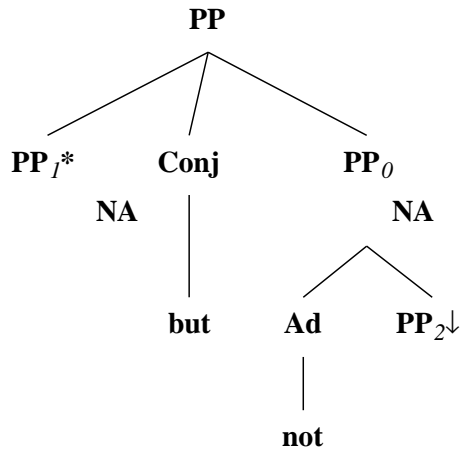,height=2.5in}
\end{tabular}
\caption{Tree for conjunction with but-not: $\beta$px1CONJARBpx2}
\label{but-not}
\end{figure}

Although these constructions sound a bit odd when the two conjuncts do
not have the same number, they are sometimes possible. The agreement
information for such NPs is always that of the non-negated conjunct:
{\it his sons, and not Bill, are in charge of doing the laundry} or
{\it not Bill, but his sons, are in charge of doing the laundry}
(Some people insist on having the commas here, but they are frequently
absent in corpus data.) The agreement feature from the non-negated
conjunct in passed to the root NP, as shown in Figure
\ref{not-but}. Aside from agreement, these constructions behave just
like their non-negated counterparts.

\begin{figure}[htb]
\centering
\begin{tabular}{c}
\psfig{figure=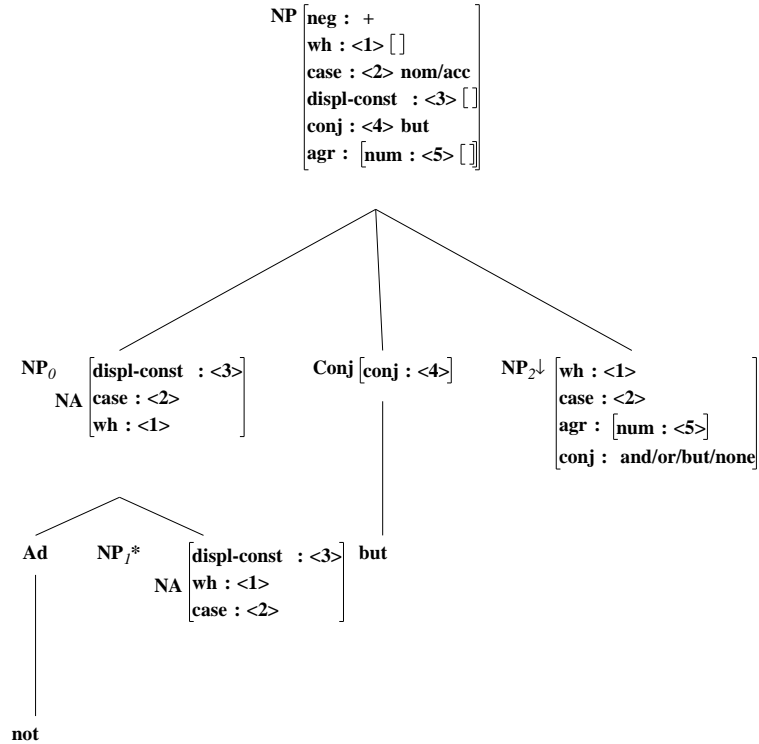,height=4in}
\end{tabular}
\caption{Tree for conjunction with not-but: $\beta$ARBnx1CONJnx2} 
\label{not-but}
\end{figure}

\section{{\it To} as a Conjunction}

{\it To} can be used as a conjunction for adjectives
(Fig. \ref{to-conj}) and determiners, when they denote points on a
scale:

\enumsentence{two to three degrees}
\enumsentence{high to very high temperatures}

As far as we can tell, when the conjuncts are determiners they must be
cardinal.

\begin{figure}[htb]
\centering
\begin{tabular}{c}
\psfig{figure=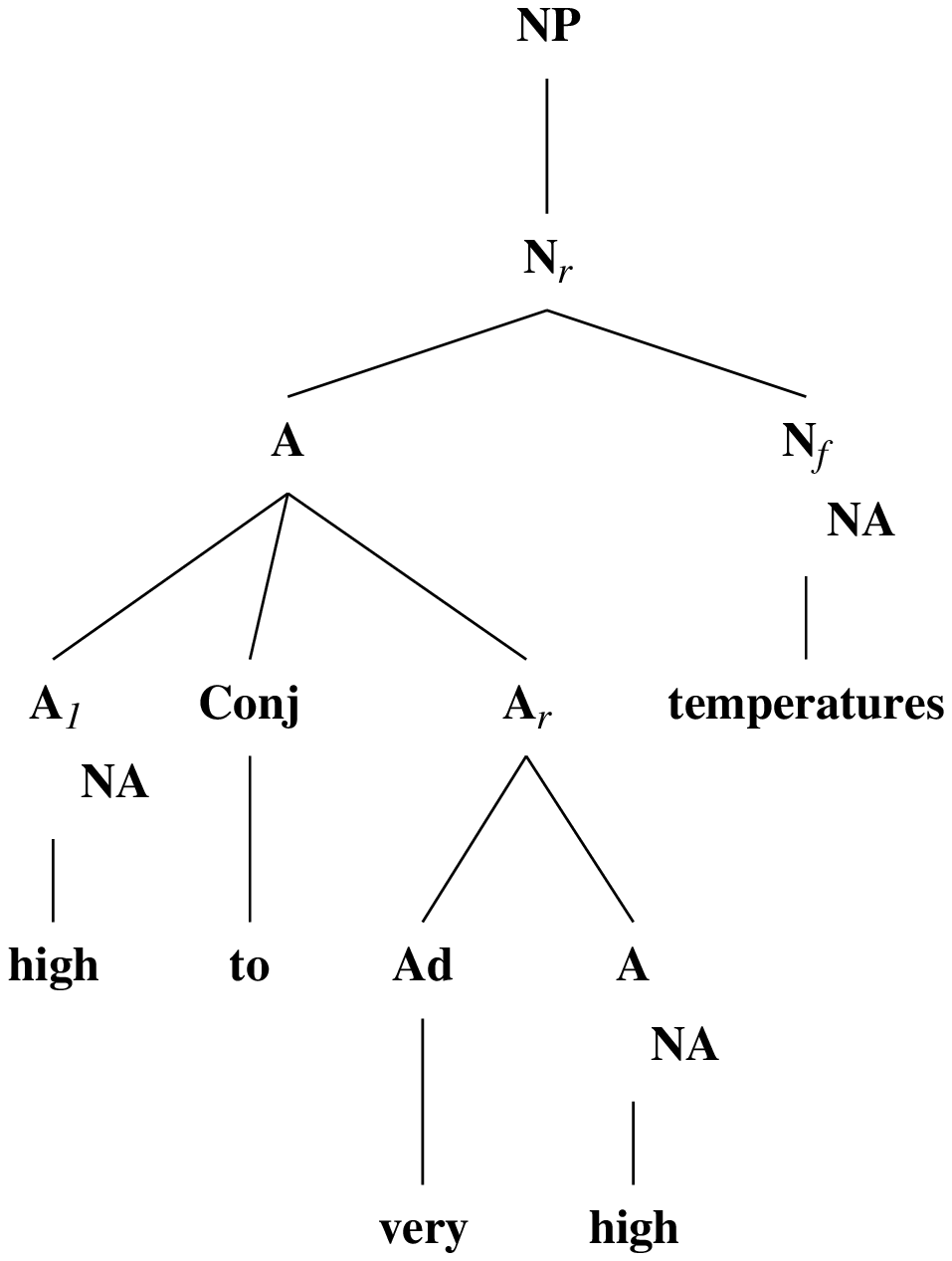,height=3.5in}
\end{tabular}
\caption{Example of conjunction with {\it to}} 
\label{to-conj}
\end{figure}

\section{Predicative Coordination}

This section describes the method for predicative coordination
(including VP coordination of various kinds) used in XTAG. The
description is derived from work described in (\cite{anoopjoshi96}).
It is important to say that this implementation of predicative
coordination is not part of the XTAG release at the moment due massive
parsing ambiguities. This is partly because of the current
implementation and also the inherent ambiguities due to VP
coordination that cause a combinatorial explosion for the parser. We
are trying to remedy both of these limitations using a probability
model for coordination attachments which will be included as part of a
later XTAG release.

This extended domain of locality in a lexicalized Tree Adjoining
Grammar causes problems when we consider the coordination of such
predicates. Consider~(\ex{1}) for instance, the NP {\em the beans that
  I bought from Alice} in the Right-Node Raising (RNR) construction
has to be shared by the two elementary trees (which are anchored by
{\em cooked} and {\em ate} respectively).

\enumsentence{(((Harry cooked) and (Mary ate)) the beans that I bought
  from Alice)}

We use the standard notion of coordination which is shown in
Figure~\ref{fig:conj} which maps two constituents of {\em like type},
but with different interpretations, into a constituent of the same
type.

\begin{figure}[htbp]
  \begin{center}
    \leavevmode
    \psfig{figure=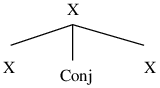,scale=110}
  \end{center}
  \caption{Coordination schema}
  \label{fig:conj}
\end{figure}

We add a new operation to the LTAG formalism (in addition to
substitution and adjunction) called {\em conjoin} (later we discuss an
alternative which replaces this operation by the traditional
operations of substitution and adjunction). While substitution and
adjunction take two trees to give a derived tree, {\em conjoin\/}
takes three trees and composes them to give a derived tree.  One of
the trees is always the tree obtained by specializing the schema in
Figure~\ref{fig:conj} for a particular category. The tree obtained
will be a lexicalized tree, with the lexical anchor as the
conjunction: {\em and}, {\em but}, etc.

The conjoin operation then creates a {\em contraction\/} between nodes
in the contraction sets of the trees being coordinated.  The term {\em
  contraction\/} is taken from the graph-theoretic notion of edge
contraction. In a graph, when an edge joining two vertices is
contracted, the nodes are merged and the new vertex retains edges to
the union of the neighbors of the merged vertices. The conjoin
operation supplies a new edge between each corresponding node in the
contraction set and then contracts that edge.

For example, applying {\em conjoin\/} to the trees {\em Conj(and)},
$\alpha(eats)$ and $\alpha(drinks)$ gives us the derivation tree and
derived structure for the constituent in \ex{1} shown in
Figure~\ref{fig:vpc}.

\enumsentence{$\ldots$ eats cookies and drinks beer}

\begin{figure}[htbp]
  \begin{center}
    \leavevmode
    \psfig{figure=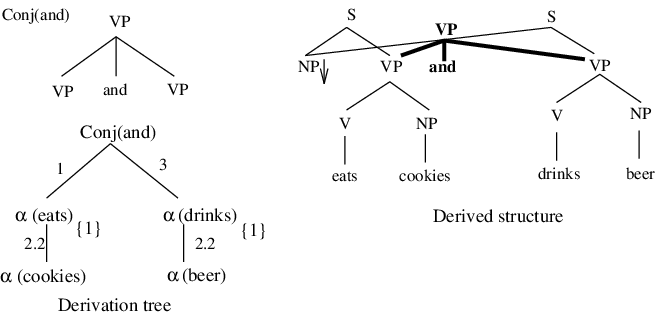,scale=110}
  \end{center}
  \caption{An example of the {\em conjoin\/} operation. $\{1\}$
    denotes a shared dependency.}
  \label{fig:vpc}
\end{figure}

Another way of viewing the conjoin operation is as the construction of
an auxiliary structure from an elementary tree. For example, from the
elementary tree $\alpha(drinks)$, the conjoin operation would create
the auxiliary structure $\beta(drinks)$ shown in
Figure~\ref{fig:aux-conj}. The adjunction operation would now be
responsible for creating contractions between nodes in the contraction
sets of the two trees supplied to it. Such an approach is attractive
for two reasons. First, it uses only the traditional operations of
substitution and adjunction. Secondly, it treats {\em conj X} as a
kind of ``modifier'' on the left conjunct {\em X}. This approach
reduces some of the parsing ambiguities introduced by the predicative
coordination trees and forms the basis of the XTAG implementation.

\begin{figure}[htbp]
  \begin{center}
    \leavevmode
    \psfig{figure=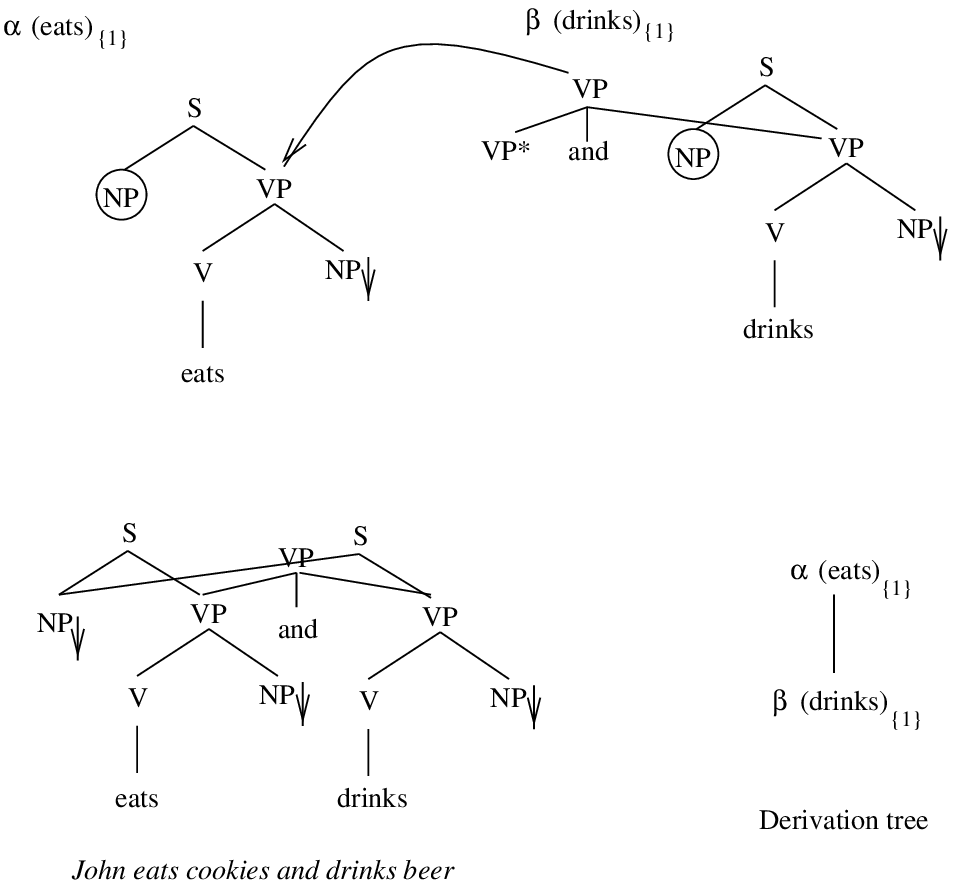,scale=110}
  \end{center}
  \caption{Coordination as adjunction.}
  \label{fig:aux-conj}
\end{figure}

More information about predicative coordination can be found in
(\cite{anoopjoshi96}), including an extension to handle gapping constructions.

\section{Pseudo-coordination}

The XTAG grammar does handle one sort of verb pseudo-coordination. 
Semi-idiomatic phrases such as 'try and' and 'up and' (as in 'they might 
try and come today') are handled as multi-anchor modifiers 
rather than as true coordination. These items adjoin to a V node, using 
the $\beta$VCONJv tree. This tree adjoins only to verbs in their base 
morphological (non-inflected) form. The verb anchor of the $\beta$VCONJv 
must also be in its base form, as shown in examples 
(\ex{1})-(\ex{3}). This blocks 3rd-person singular derivations, 
which are the only person morphologically marked in the present, except when 
an auxiliary verb is present or the verb is in the infinitive.

\enumsentence{$\ast$He tried and came yesterday.}
\enumsentence{They try and exercise three times a week.}
\enumsentence{He wants to try and sell the puppies.}

\chapter{Comparatives}
\label{compars-chapter}

\section{Introduction}

Comparatives in English can manifest themselves in many ways, acting
on many different grammatical categories and often involving ellipsis.
A distinction must be made at the outset between two very different
sorts of comparatives---those which make a comparison between two
propositions and those which compare the extent to which an entity has
one property to a greater or lesser extent than another property.  The
former, which we will refer to as {\it propositional} comparatives, is
exemplified in (\ex{1}), while the latter, which we will call {\it
metalinguistic} comparatives (following Hellan 1981), is seen in
(\ex{2}):

\enumsentence{Ronaldo is more angry than Romario.}
\enumsentence{Ronaldo is more angry than upset.}

\noindent In (\ex{-1}), the extent to which Ronaldo is angry is greater
than the extent to which Romario is angry.  Sentence (\ex{0})
indicates that the extent to which Ronaldo is angry is greater than
the extent to which he is upset.

Apart from certain of the elliptical cases, both kinds of comparatives
can be handled straightforwardly in the XTAG system.  Elliptical cases
which are not presently covered include those exemplified by the
following sentences, which would presumably be handled in the same way
as other sorts of VP ellipsis would.

\enumsentence{Ronaldo is more angry than Romario is.}
\enumsentence{Bill eats more broccoli than George eats.}
\enumsentence{Bill eats more broccoli than George does.}

We turn to the analysis of metalinguistic comparatives first.

\section{Metalinguistic Comparatives}

A metalinguistic comparison can be performed on basically all of the
predicational categories---adjectives, verb phrases, prepositional
phrases, and nouns---as in the following examples:

\enumsentence{The table is more long than wide. (AP)}
\enumsentence{Clark more makes the rules than follows them. (VP)}
\enumsentence{Calvin is more in the living room than in the kitchen. (PP)}
\enumsentence{That unindentified amphibian in the bush is more frog
than toad, I would say. (NP)}

\noindent At present, we only deal with the adjectival metalinguistic
comparatives as in (\ex{-3}).  The analysis given here for these can
be easily extended to prepositional phrases and nominal comparatives of
the metalinguistic sort, but, as with coordination in XTAG, verb
phrases will prove more difficult.

Adjectival comparatives appear to distribute with simple adjectives,
as in the following examples:

\enumsentence{Herbert is more livid than angry.}
\enumsentence{Herbert is more livid and furious than angry.}
\enumsentence{The more innovative than conventional medication cured
everyone in the sick ward.}
\enumsentence{The elephant, more wobbly than steady, fell from the
circus ball.}

\begin{figure}[htb]
\centering
\begin{tabular}{cc}
{\psfig{figure=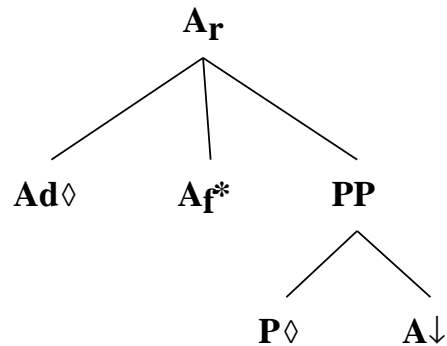,height=2.0in}}
\end{tabular}\\
\caption {Tree for Metalinguistic Adjective Comparative: $\beta$ARBaPa}
\label {ARBaPa-tree}
\end{figure}

This patterning indicates that we can give these comparatives a tree
that adjoins quite freely onto adjectives, as in
Figure~\ref{ARBaPa-tree}.  This tree is anchored by {\it more/less -
than}.  To avoid grammatically incorrect comparisons such as {\it more
brighter than dark}, the feature {\bf compar} is used to block this
tree from adjoining onto morphologically comparative adjectives.  The
foot node is {\bf compar-}, while {\it brighter} and its comparative
siblings are {\bf compar+}\footnote {The analysis given later for
adjectival propositional comparatives produces aggregated {\bf
compar+} adjectives such as {\it more bright}, which will also be
incompatible (as desired) with $\beta$ARBaPa.}.  We also wish to block
strings like {\it more brightest than dark}, which is accomplished
with the feature {\bf super}, indicating superlatives.  This feature
is negative at the foot node so that $\beta$ARBaPa cannot adjoin to
superlatives like {\it nicest}, which are specified as {\bf super+}
from the morphology.  Furthermore, the root node is {\bf
super+} so that $\beta$ARBaPa cannot adjoin onto itself and produce
monstrosities such as (\ex{1}):

\enumsentence{*Herbert is more less livid than angry than furious.}

\noindent Thus, the use of the {\bf super} feature is less to indicate
superlativeness specifically, but rather to indicate that the subtree
below a {\bf super+} node contains a full-fleshed comparison.  In the
case of lexical superlatives, the comparison is against everything,
implicitly.

A benefit of the multiple-anchor approach here is that we will never
allow sentences such as (\ex{1}), which would be permissible if we
split the comparative component and the {\it than} component of
metalinguistic comparatives into two separate trees.

\enumsentence{*Ronaldo is angrier than upset.}

We also see another variety of adjectival comparatives of the form {\it
more/less than X}, which indicates some property which is more or less
extreme than the property {\it X}.  In a sentence such as (\ex{1}),
some property is being said to hold of Francis such that it is of a
kind with {\it stupid} and that it exceeds {\it stupid} on some scale
(intelligence, for example).  Quirk et al. also note that these
constructions remark on the inadequacy of the lexical item.  Thus, in
(\ex{0}), it could be that {\it stupid} is a starting point from which
the speaker makes an approximation for some property which the speaker
feels is beyond the range of the English lexicon, but which expresses
the supreme lack of intellect of the individual it is predicated of.

\enumsentence{Francis is more than stupid.}
\enumsentence{Romario is more than just upset.}

Taking our inspiration from $\beta$ARBaPa, we can handle these
comparatives, which have the same distribution but contain an empty
adjective, by using the tree shown in Figure~\ref{ARBPa-tree}.

\begin{figure}[htb]
\centering
\begin{tabular}{cc}
{\psfig{figure=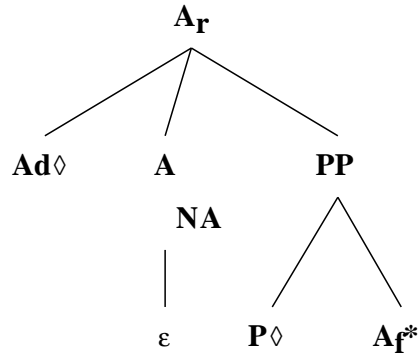,height=2.0in}}
\end{tabular}\\
\caption {Tree for Adjective-Extreme Comparative: $\beta$ARBPa}
\label {ARBPa-tree}
\end{figure}

This sort of metalinguistic comparative also occurs with the verb
phrase, prepositional phrase, and noun varieties.

\enumsentence{Clark more than makes the rules. (VP)}
\enumsentence{Calvin's hands are more than near the cookie jar. (PP)}
\enumsentence{That stuff on her face is more than mud. (NP)}

\noindent Presumably, the analysis for these would parallel that for
adjectives, though it has not yet been implemented.

\section{Propositional Comparatives}

\subsection{Nominal Comparatives}\label{nom-comparatives-section}

Nominal comparatives are considered here to be those which compare the
cardinality of two sets of entities denoted by nominal phrases.  The
following data lay out a basic distribution of these comparatives.

\enumsentence{More vikings than mongols eat spam.}
\enumsentence{*More the vikings than mongols eat spam.}
\enumsentence{Vikings eat less spaghetti than spam.}
\enumsentence{More men that walk to the store than women who despise
spam enjoyed the football game.}
\enumsentence{More men than James like scotch on the rocks.}
\enumsentence{Elmer knows fewer martians than rabbits.}

Looking at these examples, we are tempted to produce a tree for this
construction that is similar to $\beta$ARBaPa.  However, it is quite
common for the {\it than} portion of these comparatives to be left
out, as in the following sentences:

\enumsentence{More vikings eat spam.}
\enumsentence{Mongols eat less spam.}

\noindent Furthermore, {\it than NP} cannot occur without {\it more}.
These facts indicate that we can and should build up nominal
comparatives with two separate trees.  The first, which allows a
comparative adverb to adjoin to a noun, is given in
Figure~\ref{nom-compar}(a). The second is the noun-phrase modifying
prepositional tree.  The tree $\beta$CARBn is anchored by {\it
more/less/fewer} and $\beta$CnxPnx is anchored by {\it than}.  The
feature {\bf compar} is used to ensure that only one $\beta$CARBn tree
can adjoin to any given noun---its foot node is {\bf compar-} and the
root node is {\bf compar+}.  All nouns are {\bf compar-}, and the {\bf
compar} value is passed up through all trees which adjoin to N or NP.
In order to ensure that we do not allow sentences like *{\it Vikings
than mongols eat spam}, the {\bf compar} feature is used.  The NP foot
node of $\beta$CnxPnx is {\bf compar+}; thus, $\beta$CnxPnx will
adjoin only to NP's which have been already modified by $\beta$CARBn
(and thereby comparativized).  In this way, we capture sentences like
(\ex{-1}) en route to deriving sentences like (\ex{-7}), in a
principled and simple manner.

\begin{figure}[htbp]
\centering
\begin{tabular}{ccc}
{\psfig{figure=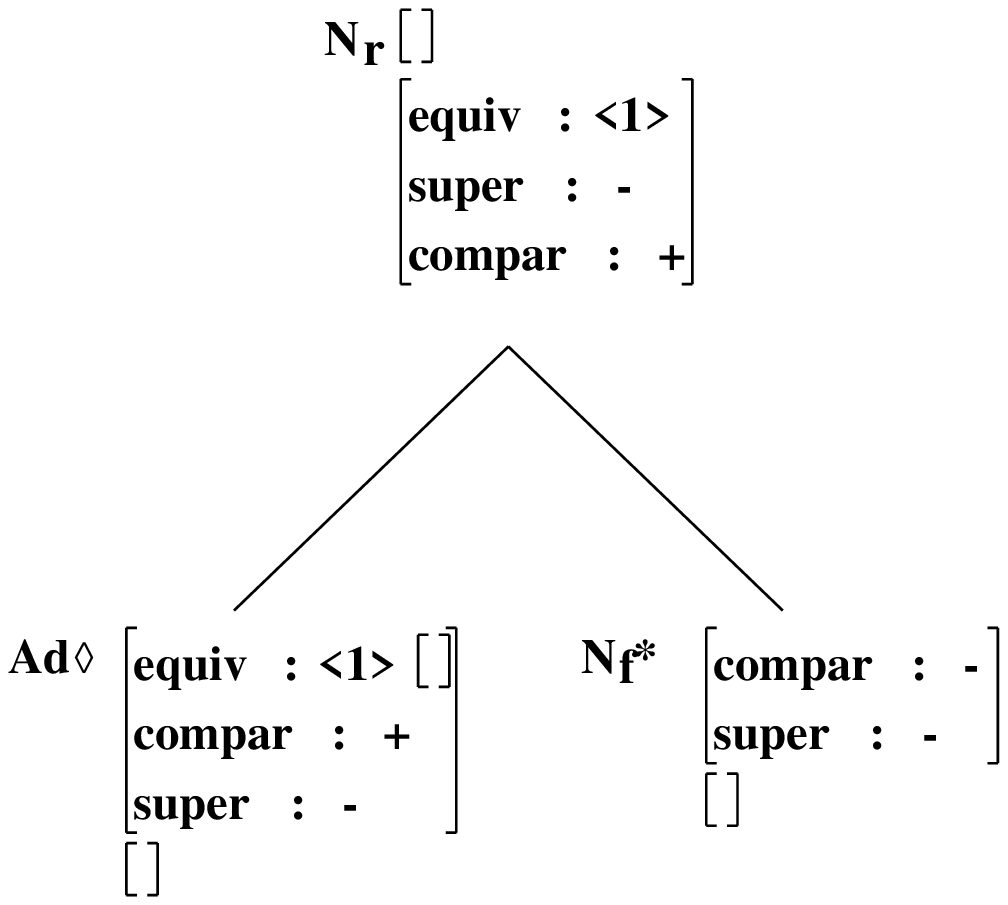,height=2.1in}}  &
\hspace{0.6in}
{\psfig{figure=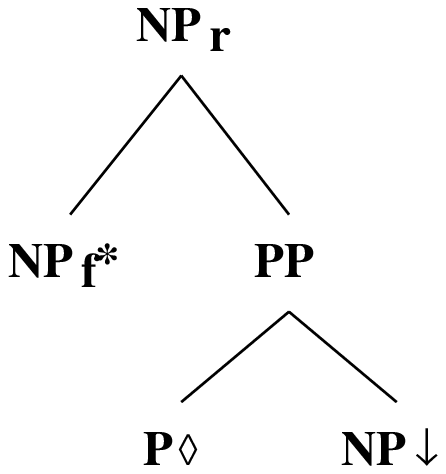,height=1.7in}} \\
(a) $\beta$CARBn tree& \qquad(b) $\beta$CnxPnx tree \\
\end{tabular}\\
\caption {Nominal comparative trees}
\label {nom-compar}
\end{figure}

Further evidence for this approach comes from comparative clauses
which are missing the noun phrase which is being compared against
something, as in the following:

\enumsentence{The vikings ate more.\footnote{We ignore here the
interpretation in which the comparison covers the eating event,
focussing only on the one which the comparison involves the stuff
being eaten.}}
\enumsentence{The vikings ate more than a boar.\footnote{This
sentence differs from the metalinguistic comparison {\it That stuff on
her face is more than mud} in that it involves a comment on the
quantity and/or type of the compared NP, whereas the other expresses
that the property denoted by the compared noun is an inadequate
characterization of the thing being described.}}

\noindent Sometimes the missing noun refers to an entity or set
available in the prior discourse, while at other times it is a
reference to some anonymous, unspecified set.  The former is
exemplified in a mini-discourse such as the following: \\

\noindent Calvin: ``The mongols ate spam.''\\
\noindent Hobbes: ``The vikings ate more.''  \\

\noindent The latter can be seen in the following example: \\

\noindent Calvin: ``The vikings ate a a boar.''\\
\noindent Hobbes: ``Indeed. But in fact, the vikings ate more than a boar.'' \\

Since the lone comparatives {\it more/less/fewer} have the same basic
distribution as noun phrases, the tree in Figure~\ref{lone-compar} is
employed to capture this fact. The root node of $\alpha$CARB is {\bf
compar+}.  Not only does this accord with our intuitions about what
the {\bf compar} feature is supposed to indicate, it also permits
$\beta$nxPnx to adjoin, giving us strings such as {\it more than NP}
for free.

\begin{figure}[htb]
\centering
\begin{tabular}{cc}
{\psfig{figure=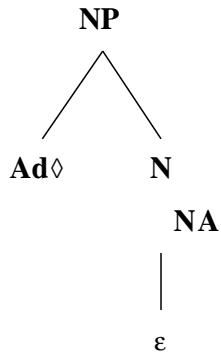,height=2.0in}}
\end{tabular}\\
\caption {Tree for Lone Comparatives: $\alpha$CARB}
\label {lone-compar}
\end{figure}

Thus, by splitting nominal comparatives into multiple trees, we make
correct predictions about their distribution with a minimal number of
simple trees.  Furthermore, we now also get certain comparative
coordinations for free, once we place the requirement that nouns and
noun phrases must match for {\bf compar} if they are to be
coordinated.  This yields strings such as the following:

\enumsentence{Julius eats more grapes and fewer boars than avocados.}
\enumsentence{Were there more or less than fifty people (at the party)?}

\noindent The structures are given in Figure~\ref{comparconjs}. Also, 
it will block strings like {\it more men and women than
children} under the (impossible) interpretation that there are more
men than children but the comparison of the quantity of women to
children is not performed.  Unfortunately, it will permit comparative
clauses such as {\it more grapes and fewer than avocados} under the
interpretation in which there are more grapes than avocados and fewer
of some unspecified thing than avocados (see Figure~\ref{badcomparconj}).

\begin{figure}[htb]
\centering
\begin{tabular}{cc}
{\psfig{figure=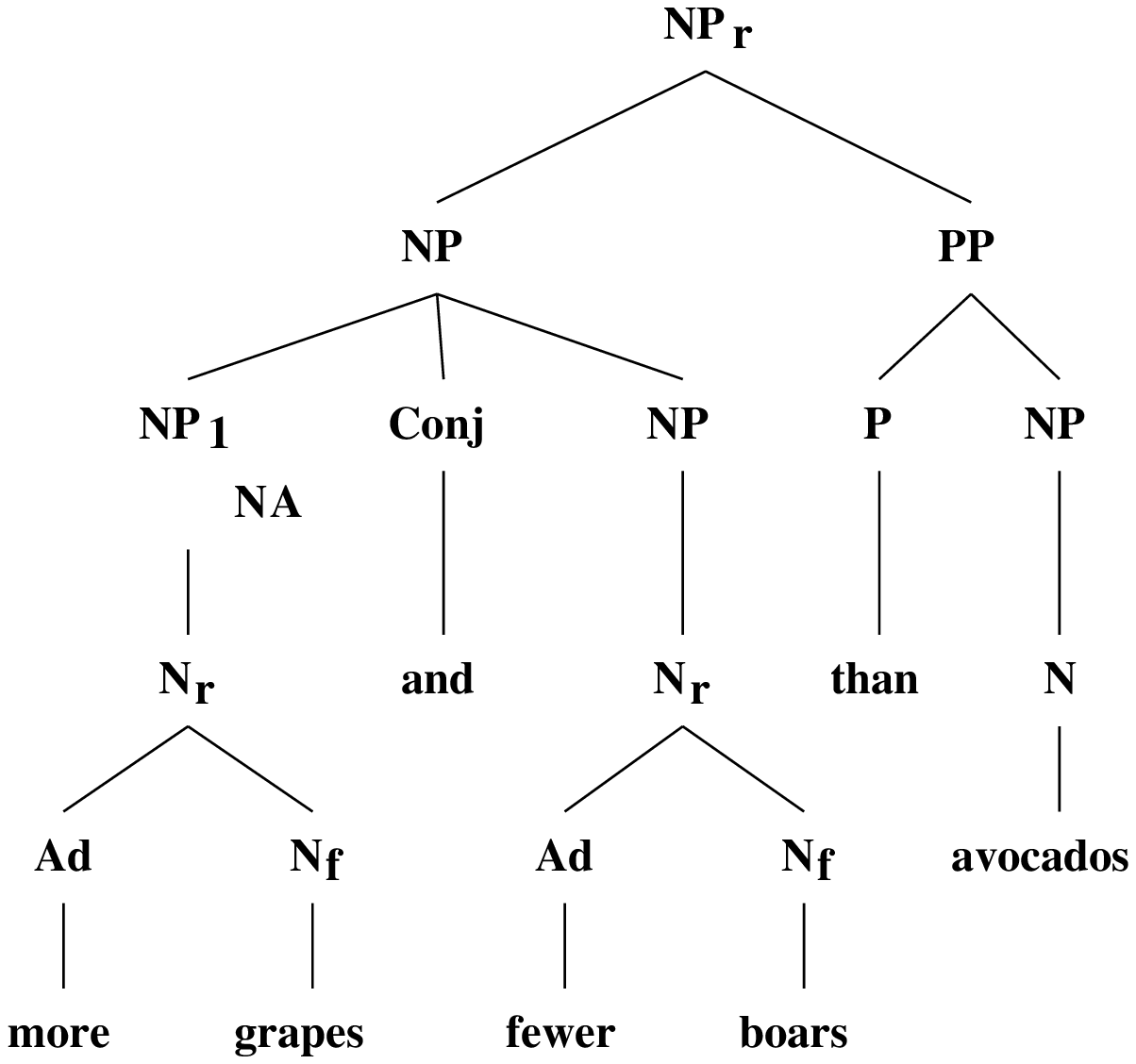,height=3.0in}}\\
{\psfig{figure=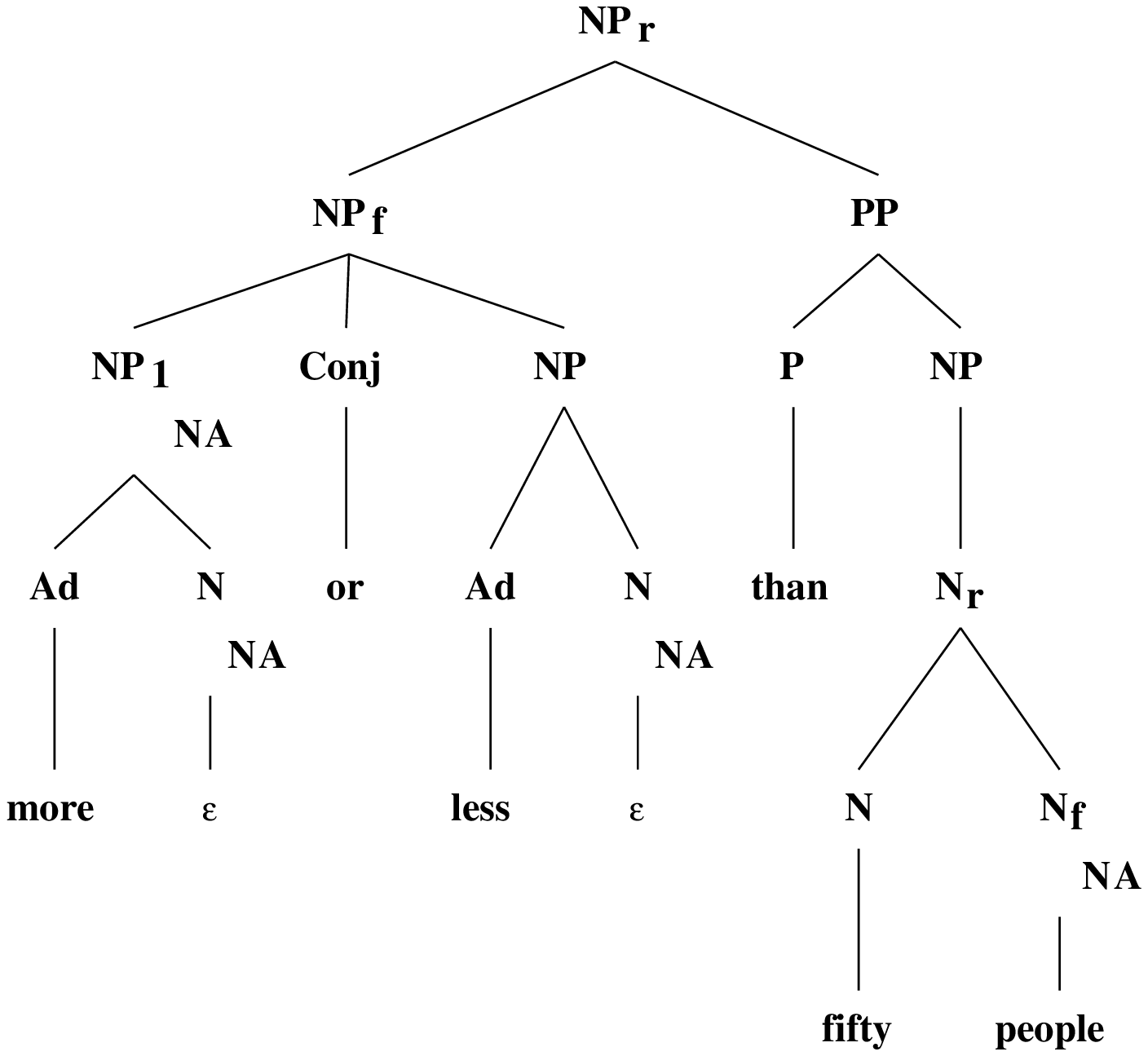,height=3.0in}}
\end{tabular}\\
\caption {Comparative conjunctions.}
\label{comparconjs}
\end{figure}

\begin{figure}[htb]
\centering
\begin{tabular}{cc}
{\psfig{figure=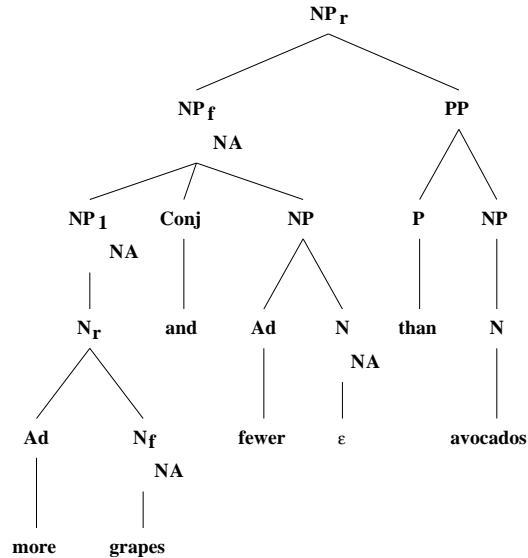,height=3.0in}}
\end{tabular}\\
\caption {Comparative conjunctions.}
\label{badcomparconj}
\end{figure}

One aspect of this analysis is that it handles the elliptical
comparatives such as the following:

\enumsentence{Arnold kills more bad guys than Steven.}

\noindent In a sense, this is actually only simulating the ellipsis of
these constructions indirectly.  However, consider the following
sentences:

\enumsentence{Arnold kills more bad guys than I do.}
\enumsentence{Arnold kills more bad guys than I.}
\enumsentence{Arnold kills more bad guys than me.}

\noindent The first of these has a {\it pro}-verb phrase which has a
nominative subject.  If we totally drop the second verb phrase, we
find that the second NP can be in either the nominative or the
accusative case.  Prescriptive grammars disallow accusative case, but
it actually is more common to find accusative case---use of the
nominative in conversation tends to sound rather stiff and unnatural.
This accords with the present analysis in which the second noun phrase
in these comparatives is the complement of {\it than} in $\beta$nxPnx,
and receives its case-marking from {\it than}.  This does mean that
the grammar will not currently accept (\ex{-1}), and indeed such
sentences will only be covered by an analysis which really deals with
the ellipsis.  Yet the fact that most speakers produce (\ex{0})
indicates that some sort of restructuring has occured that results in
the kind of structure the present analysis offers.

There is yet another distributional fact which falls out of this
analysis.  When comparative or comparativized adjectives modify a noun
phrase, they can stand alone or occur with a {\it than} phrase;
furthermore, they are obligatory when a {\it than}-phrase is present.

\enumsentence{Hobbes is a better teacher.}
\enumsentence{Hobbes is a better teacher than Bill.}
\enumsentence{A more exquisite horse launched onto the racetrack.} 
\enumsentence{A more exquisite horse than Black Beauty launched onto
the racetrack.} 
\enumsentence{*Hobbes is a teacher than Bill.}

\noindent Comparative adjectives such as {\it better} 
come from the lexicon as {\bf compar+}.  By having trees such as
$\beta$An transmit the {\bf compar} value of the A node to the root
N node, we can signal to $\beta$CnxPnx that it may adjoin when a
comparative adjective has adjoined.  An example of such an adjunction
is given in Figure~\ref{better-teacher-than-Bill}. Of course, if no
comparative element is present in the lower part of the noun phrase,
$\beta$nxPnx will not be able to adjoin since nouns themselves are
{\bf compar-}.  In order to capture the fact that a comparative
element blocks further modification to N, $\beta$An must only adjoin
to N nodes which are {\bf compar-} in their lower feature matrix.

\begin{figure}[htb]
\centering
\begin{tabular}{cc}
{\psfig{figure=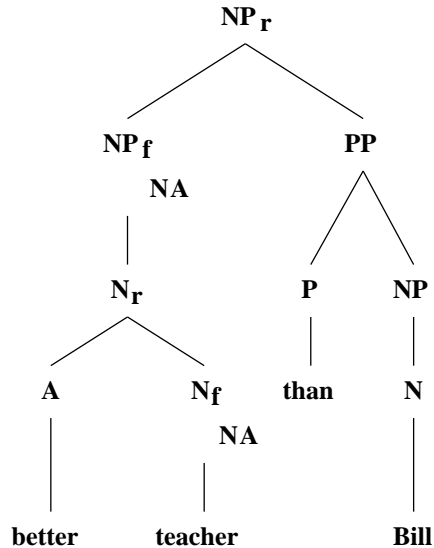,height=3.0in}}
\end{tabular}\\
\caption {Adjunction of $\beta$nxPnx to NP modified by comparative adjective.}
\label {better-teacher-than-Bill}
\end{figure}

In order to obtain this result for phrases like {\it more exquisite
horse}, we need to provide a way for {\it more} and {\em less} to
modify adjectives without a {\it than}-clause as we have with
$\beta$ARBaPa.  Actually, we need this ability independently for
comparative adjectival phrases, as discussed in the next section.

\subsection{Adjectival Comparatives}

With nominal comparatives, we saw that a single analysis was amenable
to both ``pure'' comparatives and elliptical comparatives.  This is
not possible for adjectival comparatives, as the following examples
demonstrate: 

\enumsentence{The dog is less patient.}
\enumsentence{The dog is less patient than the cat.}
\enumsentence{The dog is as patient.}
\enumsentence{The dog is as patient as the cat.}
\enumsentence{The less patient dog waited eagerly for its master.}
\enumsentence{*The less patient than the cat dog waited eagerly for
its master.}

\noindent The last example shows that comparative adjectival phrases cannot
distribute quite as freely as comparative nominals.

The analysis of elliptical comparative adjectives follows closely to
that of comparative nominals.  We build them up by first adjoining the
comparative element to the A node, which then signals to the AP node,
via the {\bf compar} feature, that it may allow a {\it than}-clause to
adjoin.  The relevant trees are given in
Figure~\ref{ellip-adj-compar}.  $\beta$CARBa is anchored by {\it more,
less} and {\it as}, and $\beta$axPnx is anchored by both {\it than}
and {\it as}.

\begin{figure}[htbp]
\centering
\begin{tabular}{ccc}
{\psfig{figure=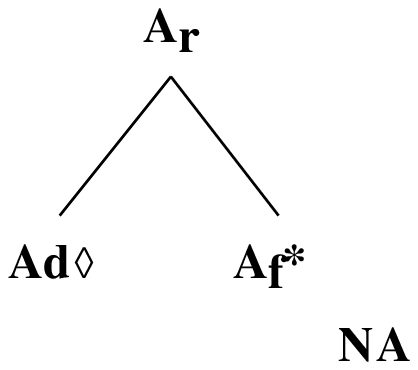,height=1.2in}}  &
\hspace{0.6in}
{\psfig{figure=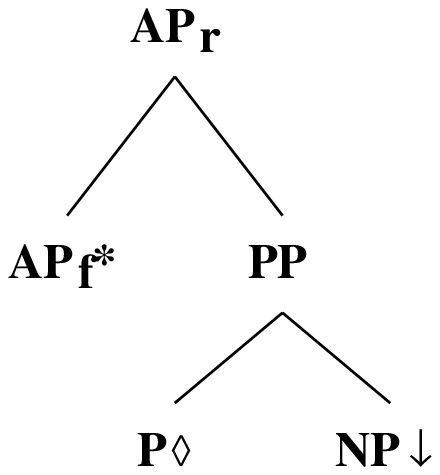,height=2.0in}} \\
(a) $\beta$CARBa tree& \qquad(b) $\beta$axPnx tree \\
\end{tabular}\\
\caption {Elliptical adjectival comparative trees}
\label {ellip-adj-compar}
\end{figure}

The advantages of this analysis are many.  We capture the
distribution exhibited in the examples given in (\ex{-5}) - (\ex{0}).
With $\beta$CARBa, comparative elements may modify adjectives wherever
they occur.  However, {\it than} clauses for adjectives have a more
restricted distribution which coincides nicely with the distribution
of AP's in the XTAG grammar.  Thus, by making them adjoin to AP rather
than A, ill-formed sentences like (\ex{0}) are not allowed.

\begin{figure}[htb]
\centering
\begin{tabular}{cc}
{\psfig{figure=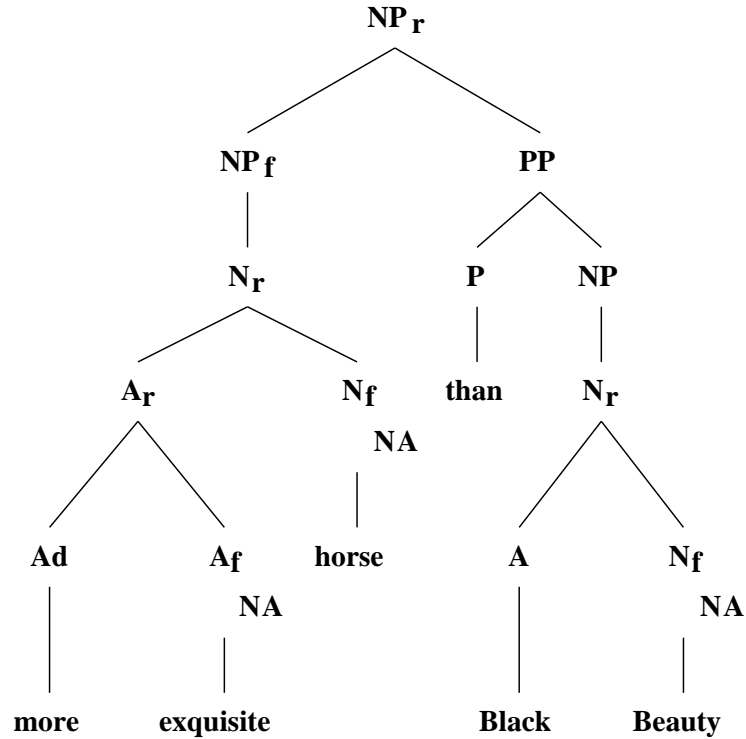,height=4.0in}}
\end{tabular}\\
\caption {Comparativized adjective triggering $\beta$CnxPnx.}
\label {black_beauty}
\end{figure}

There are two further advantages to this analysis.  One is that
$\beta$CARBa interacts with $\beta$nxPnx to produce sequences like
{\it more exquisite horse than Black Beauty}, a result alluded to at
the end of Section~\ref{nom-comparatives-section}.  We achieve this by
ensuring that the comparativeness of an adjective is controlled by a
comparative adverb which adjoins to it.  A sample derivation is given
in Figure~\ref{black_beauty}.  The second advantage is that we get
sentences such as (\ex{1}) for free.

\enumsentence{Hobbes is better than Bill.}

\noindent Since {\it better} comes from the lexicon as {\bf compar+}
and this value is passed up to the AP node, $\beta$axPnx can adjoin as
desired, giving us the derivation given in
Figure~\ref{better-than-Bill}.

\begin{figure}[htb]
\centering
\begin{tabular}{cc}
{\psfig{figure=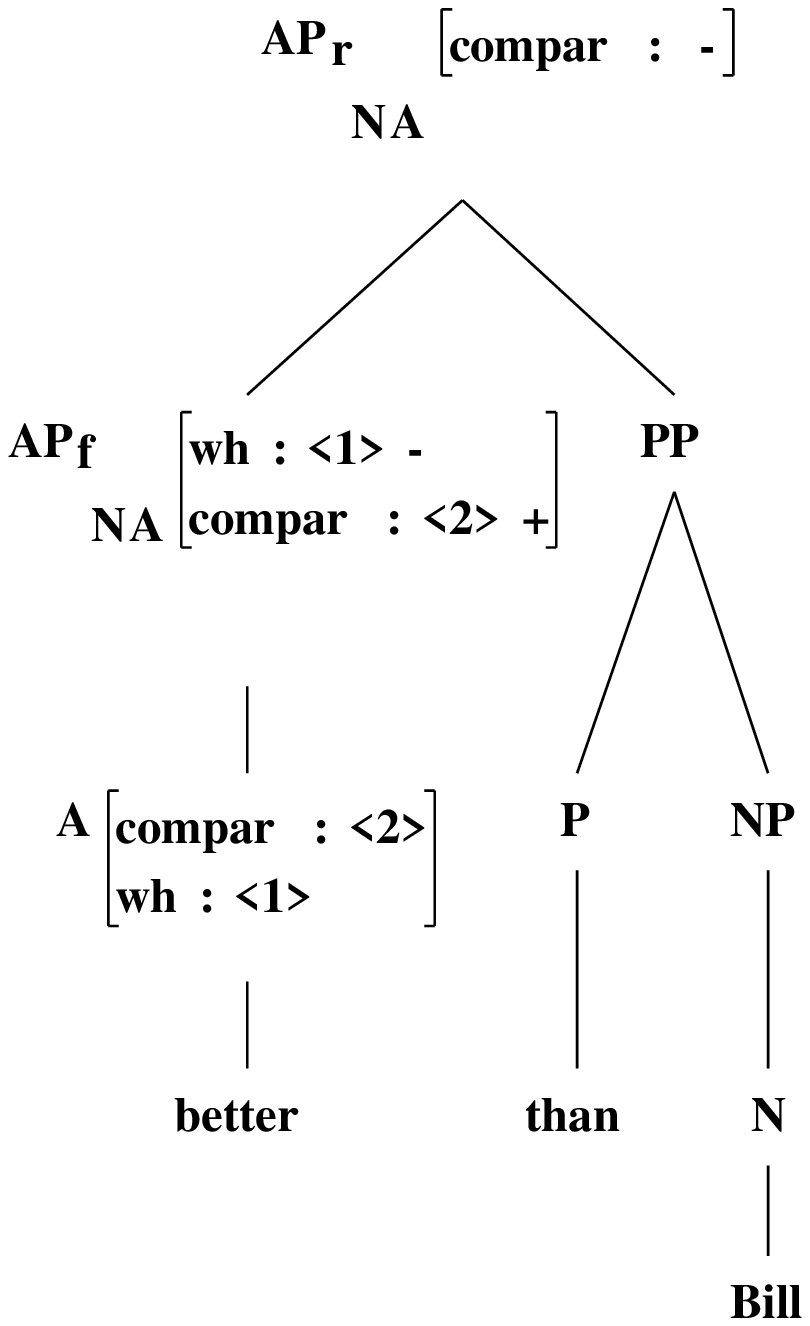,height=4.0in}}
\end{tabular}\\
\caption {Adjunction of $\beta$axPnx to comparative adjective.}
\label {better-than-Bill}
\end{figure}

Notice that the root AP node of Figure~\ref{better-than-Bill} is {\bf
compar-}, so we are basically saying that strings such as {\it better
than Bill} are not ``comparative.''  This accords with our use of the
{\bf compar} feature---a positive value for {\bf compar} signals that
the clause beneath it is to {\bf be} compared against something else.
In the case of {\it better than Bill}, the comparison has been
fulfilled, so we do not want it to signal for further comparisons.  A
nice result which follows is that $\beta$axPnx cannot adjoin more than
once to any given AP spine, and we have no need for the NA constraint
on the tree's root node.  Also, this treatment of the comparativeness
of various strings proves important in getting the coordination of
comparative constructions to work properly.

A note needs to be made about the analysis regarding the interaction
of the equivalence comparative construction {\it as ... as} and the
inequivalence comparative construction {\it more/less ... than}.  In
the grammar, {\it more, less}, and {\it as} all anchor $\beta$CARBa, and
both {\it than} and {\it as} anchor $\beta$axPnx.  Without further
modifications, this of course will give us sentences such as the
following:

\enumsentence{*?Hobbes is as patient than Bill.}
\enumsentence{*?Hobbes is more patient as Bill.}

\noindent Such cases are blocked with the feature {\bf equiv}:  {\it
more, less, fewer} and {\it than} are {\bf equiv-} while {\it as} (in
both adverbial and prepositional uses) is {\bf equiv+}.  The
prepositional trees then require that their P node and the node to
which they are adjoining match for {\bf equiv}.

An interesting phenomena in which comparisons seem to be paired with
an inappropriate {\it as/than}-clause is exhibited in (\ex{1}) and
(\ex{2}).

\enumsentence{Hobbes is as patient or more patient than Bill.}
\enumsentence{Hobbes is more patient or as patient as Bill.}

\noindent Though prescriptive grammars disfavor these sentences, these
are perfectly acceptable.  We can capture the fact that the {\it
as/than}-clause shares the {\bf equiv} value with the latter of the
comparison phrases by passing the {\bf equiv} value for the second
element to the root of the coordination tree.

\subsection{Adverbial Comparatives}

The analysis of adverbial comparatives encouragingly parallels the
analysis for nominal and elliptical adjectival comparatives---with,
however, some interesting differences.  Some examples of adverbial
comparatives and their distribution are given in the following:

\enumsentence{Albert works more quickly.}
\enumsentence{Albert works more quickly than Richard.}
\enumsentence{Albert works more.}
\enumsentence{*Albert more works.}
\enumsentence{Albert works more than Richard.}
\enumsentence{Hobbes eats his supper more quickly than Calvin.}
\enumsentence{Hobbes more quickly eats his supper than Calvin.}
\enumsentence{*Hobbes more quickly than Calvin eats his supper.}

\noindent When {\it more} is used alone as an adverb, it must also
occur after the verb phrase. Also, it appears that adverbs modified by
{\it more} and {\it less} have the same distribution as when they are
not modified.  However, the {\it than} portion of an adverbial
comparative is restricted to post verb phrase positions.

The first observation can be captured by having {\it more} and {\it
less} select only $\beta$vxARB from the set of adverb trees.
Comparativization of adverbs looks very similar to that of other
categories, and we follow this trend by giving the tree in
Figure~\ref{more-adv-mod}(a), which parallels the adjectival and
nominal trees, for these instances.  This handles the quite free
distribution of adverbs which have been comparativized, while the tree
in Figure~\ref{more-adv-mod}(b), $\beta$vxPnx, allows the {\it than}
portion of an adverbial comparative to occur only after the verb
phrase, blocking examples such as (\ex{0}). 

\begin{figure}[htbp]
\centering
\begin{tabular}{ccc}
{\psfig{figure=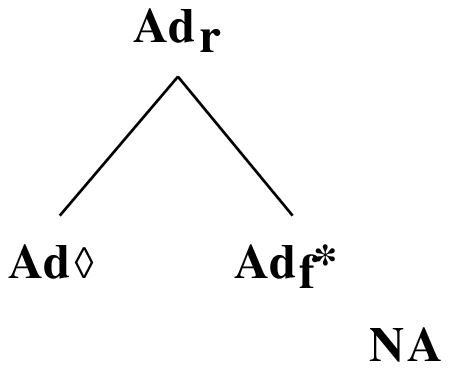,height=1.2in}}  &
\hspace{0.6in}
{\psfig{figure=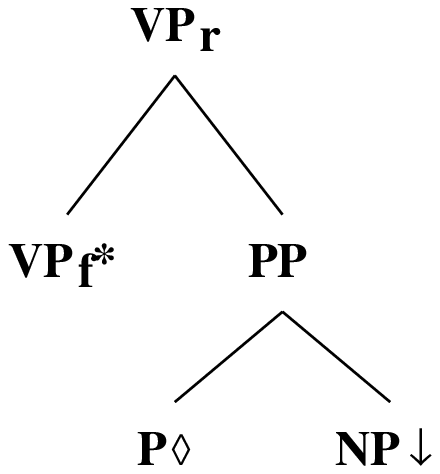,height=2.0in}} \\
(a) $\beta$CARBarb tree& \qquad(b) $\beta$vxPnx tree \\
\end{tabular}\\
\caption {Adverbial comparative trees}
\label {more-adv-mod}
\end{figure}

The usage of the {\bf compar} feature parallels that of the adjectives
and nominals; however, trees which adjoin to VP are {\bf compar-} on
their root VP node.  In this way, $\beta$vxPnx anchored by {\it than}
or {\it as} (which must adjoin to a {\bf compar+} VP) can only adjoin
immediately above a comparative or comparativized adverb.  This avoids
extra parses in which the comparative adverb adjoins at a VP node
lower than the {\it than}-clause.

A final note is that {\it as} may anchor $\beta$vxPnx
non-comparatively, as in sentence (\ex{1}). This means that
there will be two parses for sentences such as (\ex{2}).

\enumsentence{John works as a carpenter.}
\enumsentence{John works as quickly as a carpenter.}

\noindent This appears to be a legitimate ambiguity.  One is that John
works as quickly as a carpenter (works quickly), and the other is that
John works quickly when he is acting as a carpenter (but maybe he is
slow when he acting as a plumber).

\section{Future Work}

\begin{itemize}
\item Interaction with determiner sequencing (e.g., {\it several more
men than women} but not {\it *every more men than women}).

\item Handle sentential complement comparisons (e.g., {\it Bill eats
more pasta than Angus drinks beer}).

\item Add partitives.

\item Deal with constructions like {\it as many} and {\it as much}.

\item Look at {\it so...as} construction.

\end{itemize}

\chapter{Punctuation Marks}
\label{punct-chapt}

Many parsers require that punctuation be stripped out of the
input. Since punctuation is often optional, this sometimes has no
effect. However, there are a number of constructions which must
obligatorily contain punctuation and adding analyses of these to the
grammar without the punctuation would lead to severe
overgeneration. An especially common example is noun
appositives. Without access to punctuation, one would have to allow
every combinatorial possibility of NPs in noun sequences, which is
clearly undesirable (especially since there is already unavoidable
noun-noun compounding ambiguity). Aside from coverage issues, it is
also preferable to take input ``as is'' and do as little editing as
possible. With the addition of punctuation to the XTAG grammar, we
need only do/assume the conversion of certain sequences of punctuation
into the ``British'' order (this is discussed in more detail below in
Section \ref{bal}).

The XTAG POS tagger currently tags every punctuation mark as
itself. These tags are all converted to the POS tag {\it Punct} before
parsing. This allows us to treat the punctuation marks as a single POS
class. They then have features which distinguish amongst them.
Wherever possible we have the punctuation marks as anchors, to
facilitate early filtering.

The full set of punctuation marks is separated into three classes:
balanced, separating and terminal. The balanced punctuation marks are
quotes and parentheses, separating are commas, dashes, semi-colons and
colons, and terminal are periods, exclamation points and question
marks. Thus, the {\bf $<$punct$>$} feature is complex (like the {\bf
$<$agr$>$} feature), yielding feature equations like {\bf $<$Punct bal
= paren$>$} or {\bf $<$Punct term = excl$>$}. Separating and terminal
punctuation marks do not occur adjacent to other members of the same
class, but may occasionally occur adjacent to members of the other
class, e.g. a question mark on a clause which is separated by a dash
from a second clause. Balanced punctuation marks are sometimes adjacent
to one another, e.g. quotes immediately inside of parentheses. The
{\bf $<$punct$>$} feature allows us to control these local
interactions.

We also need to control non-local interaction of punctuation
marks. Two cases of this are so-called quote alternation, wherein
embedded quotation marks must alternate between single and double, and
the impossibility of embedding an item containing a colon inside of
another item containing a colon. Thus, we have a fourth value for {\bf
$<$punct$>$}, {\bf $<$contains colon/dquote/etc. +/-$>$}, which
indicates whether or not a constituent contains a particular
punctuation mark. This feature is percolated through all auxiliary
trees.  Things which may not embed are: colons under colons,
semi-colons, dashes or commas; semi-colons under semi-colon or commas.
Although it is rare, parentheses may appear inside of parentheses, say
with a bibliographic reference inside a parenthesized sentence.

\section{Appositives, parentheticals and vocatives}

These trees handle constructions where additional lexical material is
only licensed in conjunction with particular punctuation marks. Since
the lexical material is unconstrained (virtually any noun can occur as
an appositive), the punctuation marks are anchors and the other nodes
are substitution sites. There are cases where the lexical material is
restricted, as with parenthetical adverbs like {\it however}, and in
those cases we have the adverb as the anchor and the punctuation marks
as substitution sites.

When these constructions can appear inside of clauses
(non-peripherally), they must be separated by punctuation marks on
both sides. However, when they occur peripherally they have either a
preceding or following punctuation mark. We handle this by having
both peripheral and non-peripheral trees for the relevant
constructions. The alternative is to insert the second (following)
punctuation mark in the tokenization process (i.e. insert a comma
before the period when an appositive appears on the last NP of a
sentence). However, this is very difficult to do accurately.

\subsection{$\beta$nxPUnxPU}

The symmetric (non-peripheral) tree for NP appositives, anchored by:
comma, dash or parentheses. It is shown in Figure \ref{nxPUnxPU} anchored by
parentheses. 

\enumsentence{The music here , Russell Smith's ``Tetrameron '' ,
sounded good . [Brown:cc09]}
\enumsentence{...cost 2 million pounds (3 million dollars)}
\enumsentence{Sen. David Boren (D., Okla.)...}

\enumsentence{
...some analysts believe the two recent natural
disasters -- Hurricane Hugo and the San Francisco earthquake -- will
carry economic ramifications.... [WSJ]
}

\begin{figure}[hbt]
\centering
\hspace{0.0in}
\psfig{figure=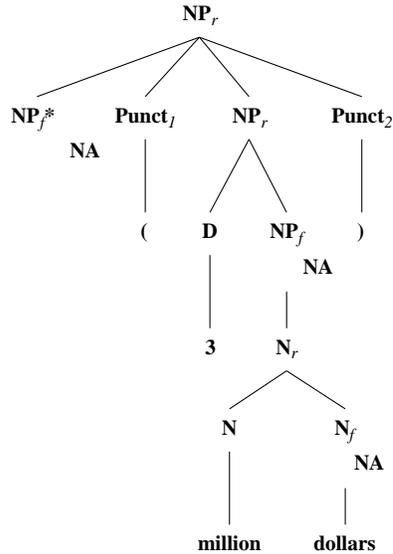,height=3.0in}
\caption{The $\beta$nxPUnxPU tree, anchored by parentheses}
\label{nxPUnxPU}
\end{figure}

The punctuation marks are the anchors and the appositive NP is
substituted. The appositive can be conjoined, but only with a lexical
conjunction (not with a comma). Appositives with commas or dashes
cannot be pronouns, although they may be conjuncts containing
pronouns.  When used with
parentheses this tree actually presents an alternative rather than an
appositive, so a pronoun is possible. Finally, the appositive position
is restricted to having nominative or accusative case to block PRO
from appearing here.

Appositives can be embedded, as in (\ex{1}), but do not seem to be
able to stack on a single NP. In this they are more like restrictive
relatives than appositive relatives, which typically can stack.

\enumsentence{...noted Simon Briscoe, UK economist for Midland
Montagu, a unit of Midland Bank PLC.}

\subsection{$\beta$nPUnxPU}

The symmetric (non-peripheral) tree for N-level NP appositives, is
anchored by comma. The modifier is typically an address.  It is clear
from examples such as (\ex{1}) that these are attached at N, rather
than NP. {\it Carrier } is not an appositive on {\it Menlo Park}, as it
would be if these were simply stacked appositives. Rather, {\it
Calif.} modifies {\it Menlo Park}, and that entire complex is
compounded with {\it carrier}, as shown in the correct derivation in
Figure \ref{nPUnx}. Because this distinction is less clear when the
modifier is peripheral (e.g. ends the sentence), and it would be
difficult to distinguish between NP and N attachment, we do not
currently allow a peripheral N-level attachment.

\enumsentence{An official at Consolidated Freightways Inc., a Menlo
Park, Calif., less-than-truckload carrier , said...}
\enumsentence{Rep. Ronnie Flippo (D., Ala.), of the delegation,
says...}

\begin{figure}[hbt]
\centering
\hspace{0.0in}
\psfig{figure=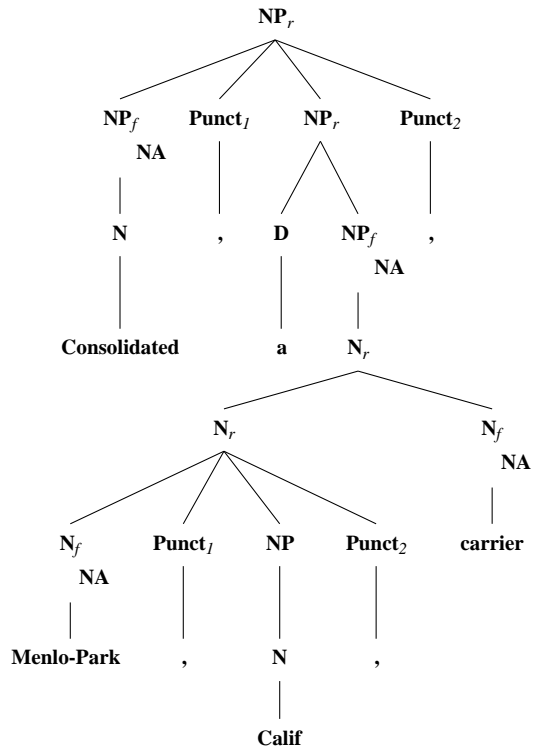,height=4.0in}
\caption{An N-level modifier, using the $\beta$nPUnx tree}
\label{nPUnx}
\end{figure}

\subsection{$\beta$nxPUnx}

This tree, which can be anchored by a comma, dash or colon, handles
asymmetric (peripheral) NP appositives and NP colon expansions of
NPs. Figure \ref{nxPUnx} shows this tree anchored by a dash and a
colon. Like the symmetric appositive tree, $\beta$nxPUnxpu, the
asymmetric appositive cannot be a pronoun, while the colon expansion
can. Thus, this constraint comes from the syntactic entry in both
cases rather than being built into the tree.

\begin{figure}[hbt]
\centering
\hspace{0.0in}
\begin{tabular}{cc}
\psfig{figure=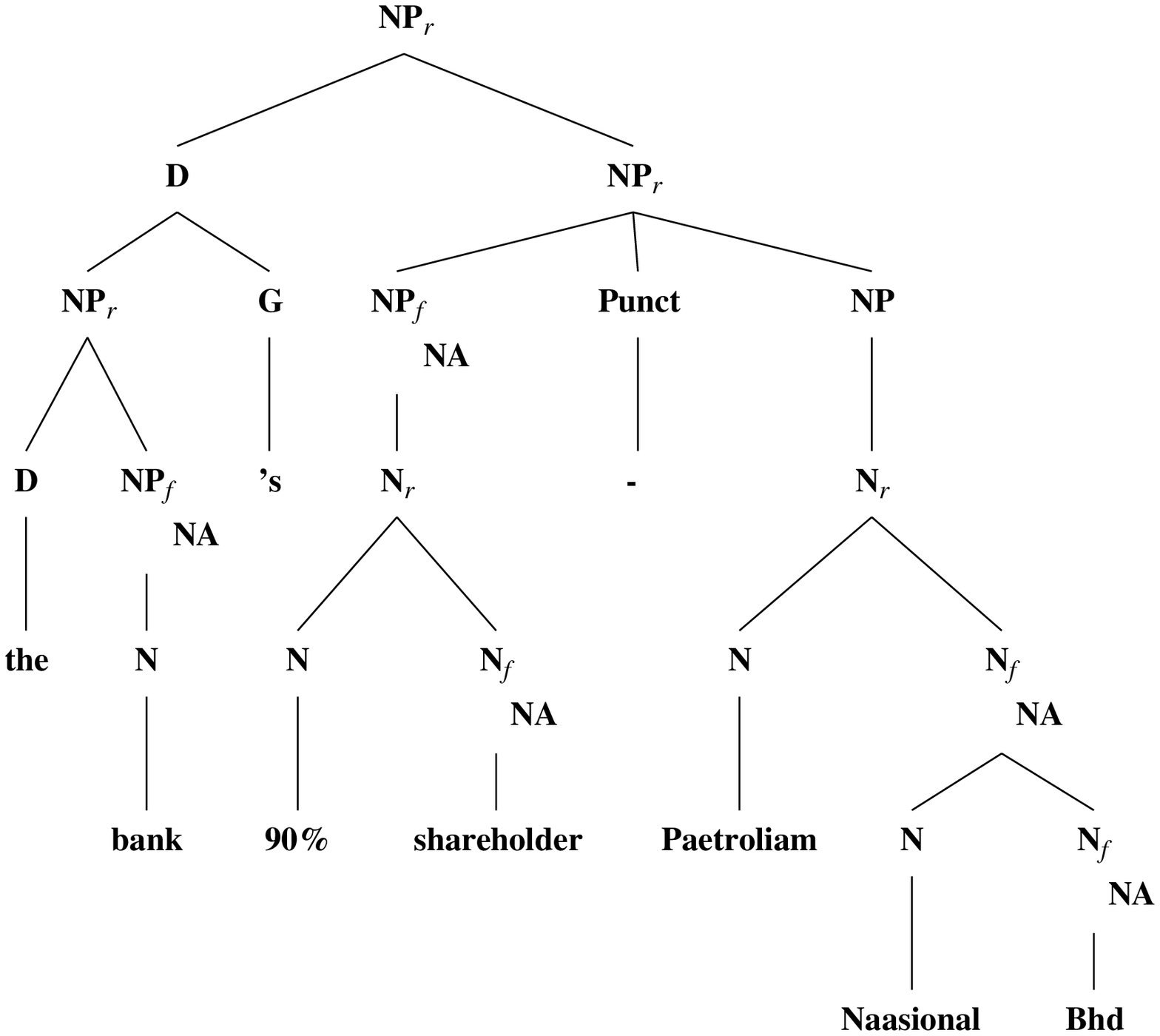,height=3.0in}
& \psfig{figure=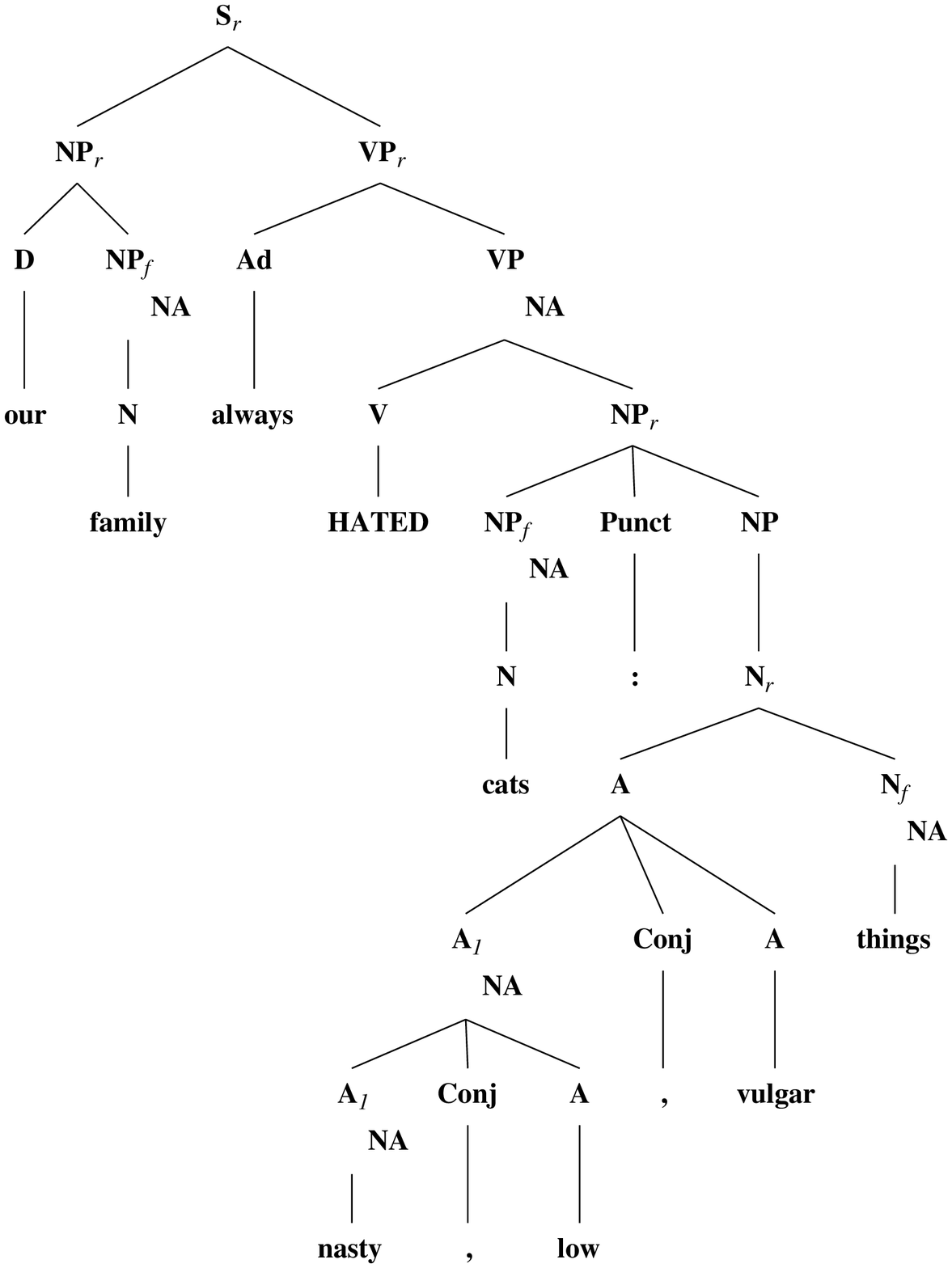,height=4.5in}
\\
(a) & (b) \\
\end{tabular}
\caption{The derived trees for an NP with (a) a peripheral, dash-separated
appositive and (b) an NP colon expansion (uttered by the Mouse in
\protect{\it Alice's Adventures in Wonderland})}
\label{nxPUnx}
\end{figure}

\enumsentence{the bank's 90\% shareholder -- Petroliam Nasional Bhd. [Brown]}

\enumsentence{...said Chris Dillow, senior U.K. economist at Nomura
Research Institute .}

\enumsentence{...qualities that are seldom found in one work: Scrupulous
scholarship, a fund of personal experience,... [Brown:cc06]}
\enumsentence{I had eyes for only one person : him .}

The colon expansion cannot itself contain a colon, so the foot $S$ has the
feature NP.t:$<punct contains colon> = -$. 

\subsection{$\beta$PUpxPUvx}

Tree for pre-VP parenthetical PP, anchored by commas or dashes - 

\enumsentence{John , in a fit of anger , broke the vase}
\enumsentence{Mary , just within the last year , has totalled two cars}

\noindent
These are clearly not NP modifiers. 

Figures \ref{betaPUpxPUvx} and \ref{PUpxPUvx-anger} show this tree
alone and as part of the parse for (\ex{-1}).

\begin{figure}[hbt]
\centering
\hspace{0.0in}
\psfig{figure=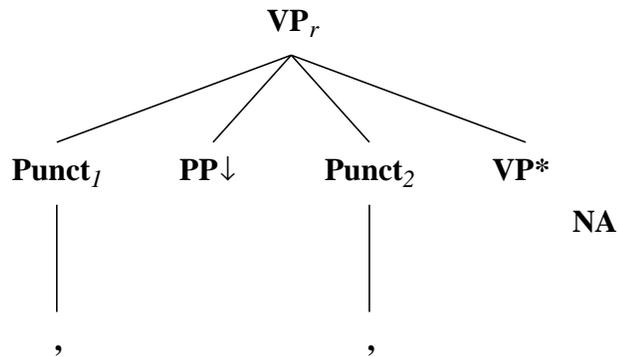,height=2.0in}
\caption{The $\beta$PUpxPUvx tree, anchored by commas}
\label{betaPUpxPUvx}
\end{figure}

\begin{figure}[hbt]
\centering
\hspace{0.0in}
\psfig{figure=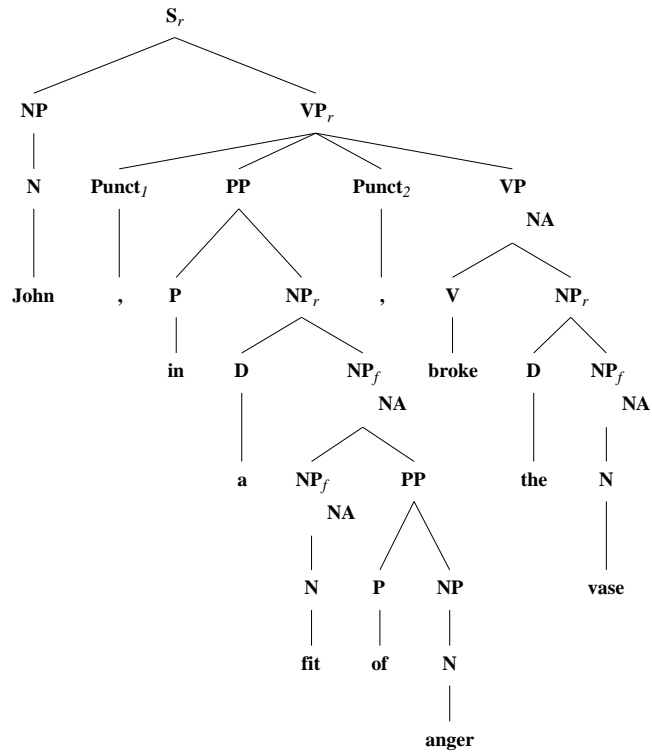,height=4.0in}
\caption{Tree illustrating the use of $\beta$PUpxPUvx}
\label{PUpxPUvx-anger}
\end{figure}

\subsection{$\beta$puARBpuvx}
\label{par-adverb}

Parenthetical adverbs - {\it however}, {\it though}, etc. Since the
class of adverbs is highly restricted, this tree is anchored by the
adverb and the punctuation marks substitute.  The punctuation marks
may be either commas or dashes.  Like the parenthetical PP above,
these are clearly not NP modifiers.

\enumsentence{The new argument over the notification
guideline , however , could sour any atmosphere of cooperation that
existed . \hfill [WSJ]}

\subsection{$\beta$sPUnx}

Sentence final vocative, anchored by comma:  
 
\enumsentence{You were there , Stanley/my boy .}

Also, when anchored by colon, NP expansion on S. These often appear to
be extraposed modifiers of some internal NP. The NP must be quite
heavy, and is usually a list:

\enumsentence{Of the major expansions in 1960, three were financed
under the R. I. Industrial Building Authority's 100\% guaranteed
mortgage plan: Collyer Wire, Leesona Corporation, and American Tube \&
Controls.}

A simplified version of this sentence is shown in figure
\ref{sPUnx}. The NP cannot be a pronoun in either of these cases.
Both vocatives and colon expansions are restricted to appear on tensed
clauses (indicative or imperative).

\begin{figure}[hbt]
\centering
\hspace{0.0in}
\psfig{figure=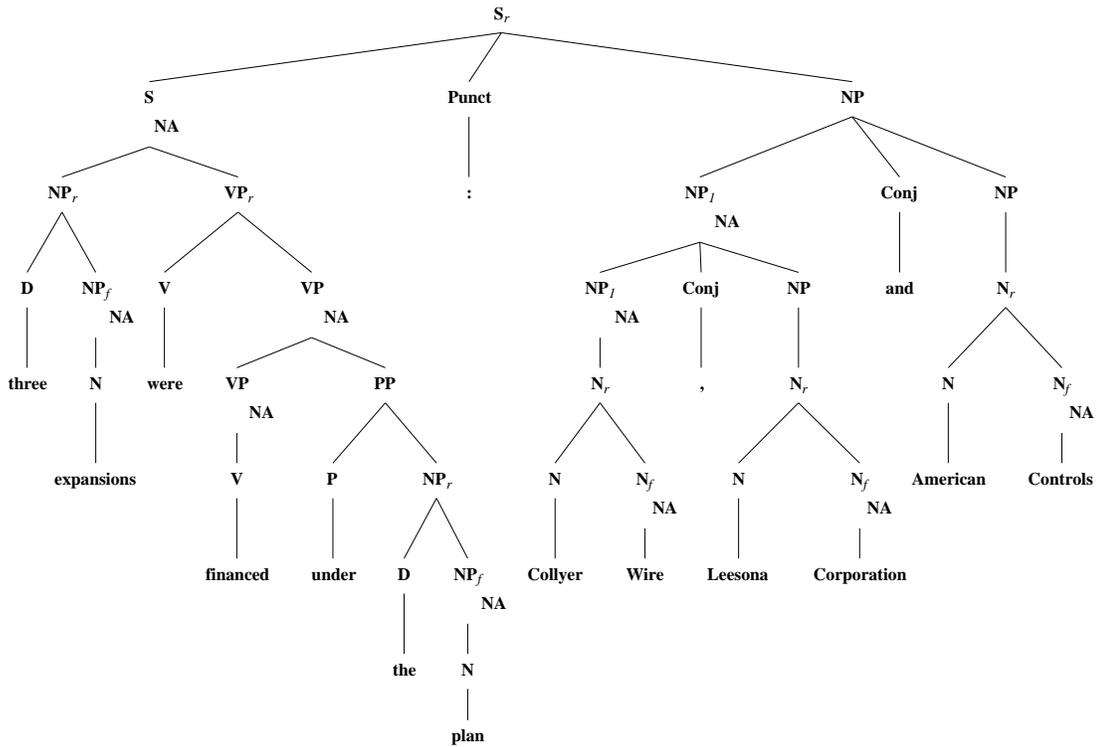,height=4.0in}
\caption{A tree illustrating the use of sPUnx for a colon expansion
attached at S.}
\label{sPUnx}
\end{figure}

\subsection{$\beta$nxPUs}

Tree for sentence initial vocatives, anchored by a comma: 

\enumsentence{Stanley/my boy , you were there .}

The noun phrase may be anything but a pronoun, although it is most
commonly a proper noun. The clause adjoined to must be indicative or
imperative.

\section{Bracketing punctuation}
\label{bal}

\subsection{Simple bracketing}

Trees: $\beta$PUsPU, $\beta$PUnxPU, $\beta$PUnPU, $\beta$PUvxPU, $\beta$PUvPU,
$\beta$PUarbPU, $\beta$PUaPU, $\beta$PUdPU, $\beta$PUpxPU,
$\beta$PUpPU

\noindent
These trees are selected by parentheses and quotes and can adjoin onto
any node type, whether a head or a phrasal constituent.  This handles
things in parentheses or quotes which are syntactically integrated
into the surrounding context. Figure \ref{bal-trees} shows the $\beta$PUsPU
anchored by parentheses, and this tree along with $\beta$PUnxPU in a
derived tree.

\enumsentence{Dick Carroll and his accordion (which we now refer to
as ``Freida'') held over at Bahia Cabana where ``Sir'' Judson
Smith brings in his calypso capers Oct. 13 .\hfill [Brown:ca31]}

\enumsentence{...noted that the term ``teacher-employee'' (as
opposed to, e.g., ``maintenance employee'') was a not inapt
description. \hfill [Brown:ca35]}

\begin{figure}[hbt]
\centering
\hspace{0.0in}
\begin{tabular}{cc}
\psfig{figure=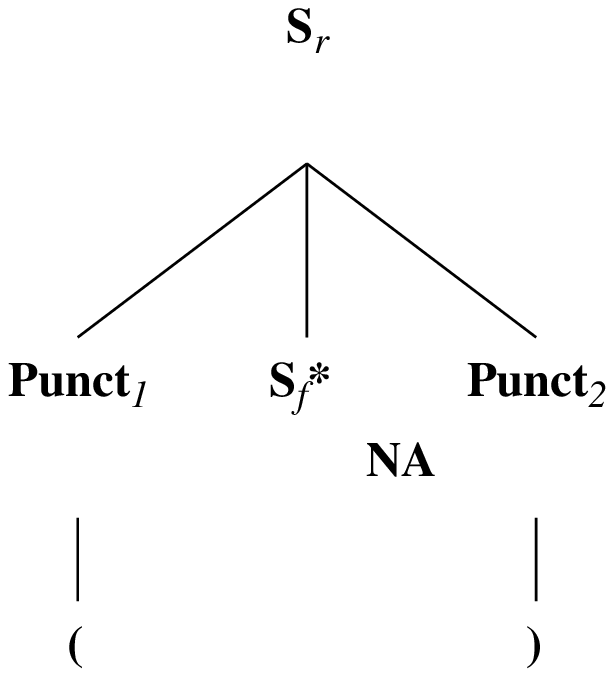,height=2.0in}
& %\hspace{0.5in}
\psfig{figure=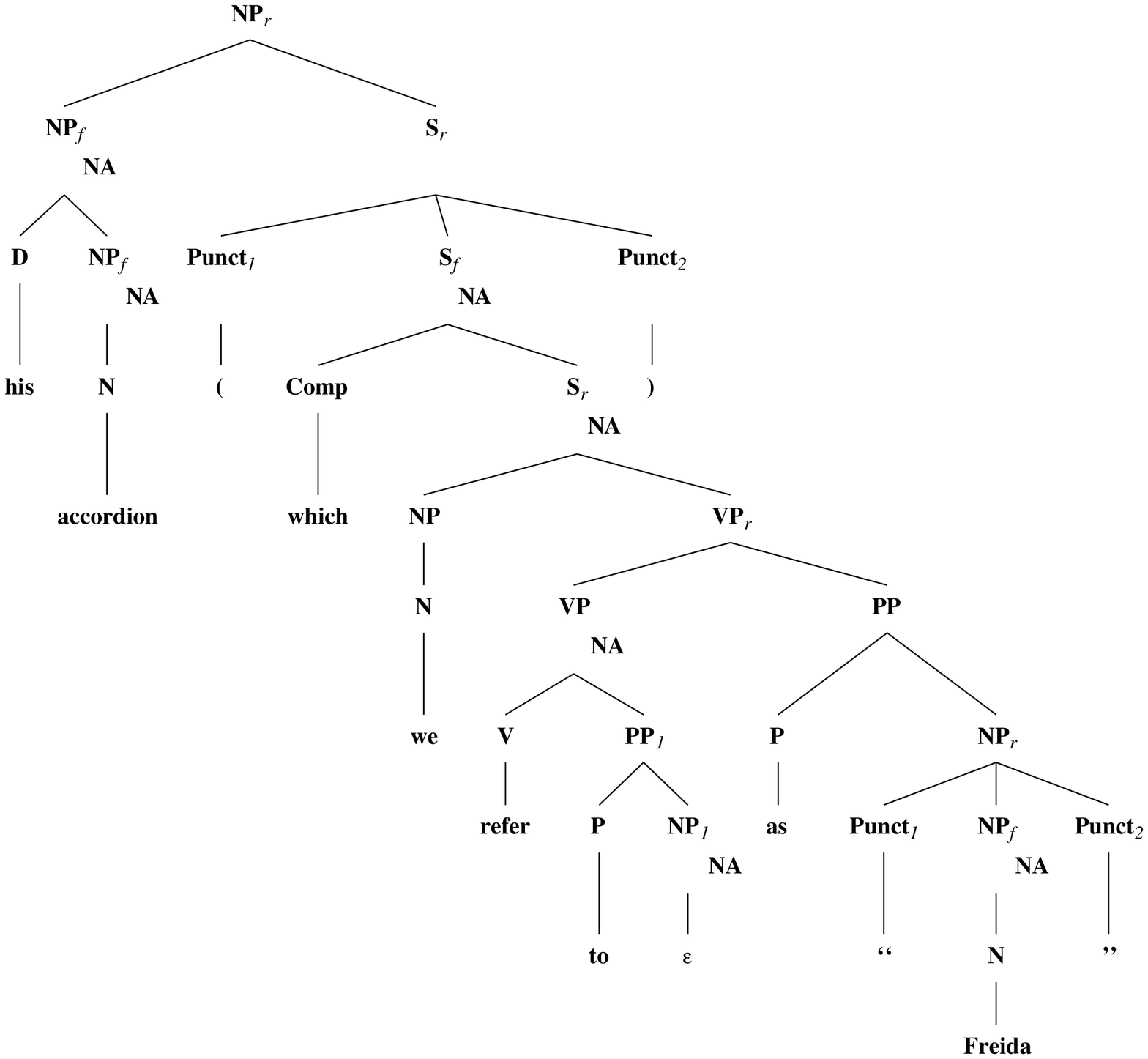,height=4in}
\\
(a) & (b) \\
\end{tabular}
\caption{$\beta$PUsPU anchored by parentheses, and in a derivation,
along with  $\beta$PUnxPU}
\label{bal-trees}
\end{figure}

There is a convention in English that quotes embedded in quotes
alternate between single and double; in American English the outermost
are double quotes, while in British English they are single.  The {\bf
contains} feature is used to control this alternation. The trees
anchored by double quotation marks have the feature {\bf punct
contains dquote = -} on the foot node and the feature {\bf punct
contains dquote = +} on the root. All adjunction trees are transparent
to the {\bf contains} feature, so if any tree below the double
quote is itself enclosed in double quotes the derivation will
fail. Likewise with the trees anchored by single quotes. The quote
trees in effect ``toggle'' the {\bf contains Xquote}
feature. Immediate proximity is handled by the {\bf punct balanced}
feature, which allows quotes inside of parentheses, but not vice-versa.

In addition, American English typically places/moves periods (and
commas) inside of quotation marks when they would logically occur
outside, as in example
\ex{1}. The comma in the first part of the quote is not part of the
quote, but rather part of the parenthetical quoting clause. However,
by convention it is shifted inside the quote, as is the final
period. British English does not do this. We assume here that the
input has already been tokenized into the ``British'' format.

\enumsentence{``You can't do this to us ,'' Diane screamed .  ``We are
Americans.''}

The $\beta$PUsPU can handle quotation marks around multiple sentences,
since the sPUs tree allows us to join two sentences with a period,
exclamation point or question mark. Currently, however, we cannot
handle the style where only an open quote appears at the beginning of
a paragraph when the quotation extends over multiple
paragraphs. We could allow a lone open quote to select the $\beta$PUs
tree, if this is deemed desirable.

Also, the $\beta$PUsPU is selected by a pair of commas to handle
non-peripheral appositive relative clauses, such as in example
(\ex{1}). Restrictive and appositive relative clauses are not
syntactically differentiated in the XTAG grammar
(cf. Chapter \ref{rel_clauses}).

\enumsentence{This news , announced by Jerome Toobin , the orchestra's
administrative director , brought applause ...  [Brown:cc09]}

The trees discussed in this section will only allow balanced
punctuation marks to adjoin to constituents. We will not get them
around non-constituents, as in (\ex{1}).

\enumsentence{Mary asked him to leave (and  he left)}

\subsection{$\beta$sPUsPU}

This tree allows a parenthesized clause to adjoin onto a
non-parenthesized clause.

\enumsentence{Innumerable motels from Tucson to New York boast
swimming pools ( `` swim at your own risk '' is the hospitable sign
poised at the brink of most pools ) . \hfill [Brown:ca17]}

\section{Punctuation trees containing no lexical material}

\subsection{$\alpha$PU}

This is the elementary tree for substitution of punctuation
marks. This tree is used in the quoted speech trees, where including
the punctuation mark as an anchor along with the verb of saying would
require a new entry for every tree selecting the relevant tree
families. It is also used in the tree for parenthetical adverbs
($\beta$puARBpuvx), and for S-adjoined PPs and adverbs ($\beta$spuARB
and $\beta$spuPnx).

\subsection{$\beta$PUs}

Anchored by comma: allows comma-separated clause initial adjuncts,
(\ex{1}-\ex{2}).

\enumsentence{Here , as in ``Journal'' , Mr. Louis has given himself
the lion's share of the dancing... \hfill [Brown:cc09]}

\enumsentence{Choreographed by Mr. Nagrin, the work filled the second
half of a program}

To keep this tree from appearing on root Ss (i.e. {\it , sentence}),
we have a root constraint that {\bf $<$punct struct = nil$>$} (similar
to the requirement that root Ss be tensed, i.e. {\bf $<$mode~=~ind/imp$>$}).  The {\bf $<$punct struct$>$ = nil} feature on the foot
blocks stacking of multiple punctuation marks. This feature is shown
in the tree in Figure \ref{PUs}.

\begin{figure}[hbt]
\centering
\hspace{0.0in}
\psfig{figure=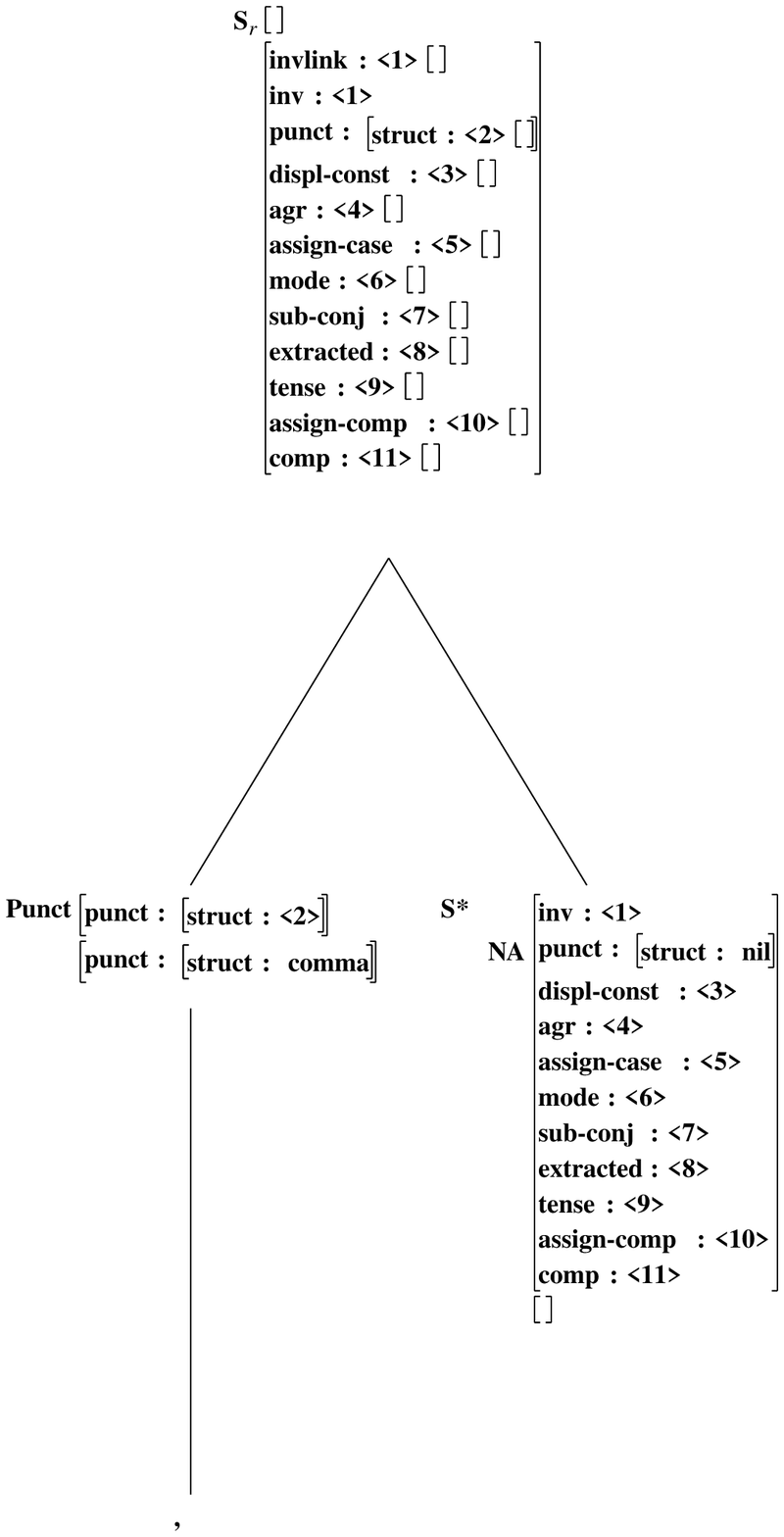,height=5.5in}
\caption{$\beta$PUs, with features displayed}
\label{PUs}
\end{figure}

This tree can be also used by adjuncts on embedded clauses:

\enumsentence{One might expect that in a poetic career of seventy-odd
years, some changes in style and method would have occurred, some
development taken place. \hfill [Brown:cj65]}

These adjuncts sometimes have commas on both sides of the adjunct, or,
like (\ex{0}), only have them at the end of the adjunct.

Finally, this tree is also used for peripheral appositive relative
clauses.

\enumsentence{Interest may remain limited into tomorrow's U.K. trade figures,
which the market will be watching closely to see if there is any
improvement after disappointing numbers in the previous two months.}

\subsection{$\beta$sPUs}
\label{sPUs}

This tree handles clausal ``coordination'' with comma, dash, colon,
semi-colon or any of the terminal punctuation marks. The first clause
must be either indicative or imperative. The second may also be
infinitival with the separating punctuation marks, but must be
indicative or imperative with the terminal marks; with a comma, it may
only be indicative. The two clauses need not share the same mode. NB:
Allowing the terminal punctuation marks to anchor this tree allows us
to parse sequences of multiple sentences. This is not the usual mode
of parsing; if it were, this sort of sequencing might be better
handled by a higher level of processing.

\enumsentence{For critics , Hardy has had no poetic periods -- one does
not speak of early Hardy or late Hardy , or of the London or Max Gate
period....}

\enumsentence{Then there was exercise , boating and hiking , which was
not only good for you but also made you more virile : the thought of
strenuous activity left him exhausted.}

This construction is one of the few where two non-bracketing
punctuation marks can be adjacent. It is possible (if rare) for the
first clause to end with a question mark or exclamation point, when
the two clauses are conjoined with a semi-colon, colon or
dash. Features on the foot node, as shown in Figure \ref{sPUs-tree}, control
this interaction.

\begin{figure}[hbt]
\centering
\hspace{0.0in}
\psfig{figure=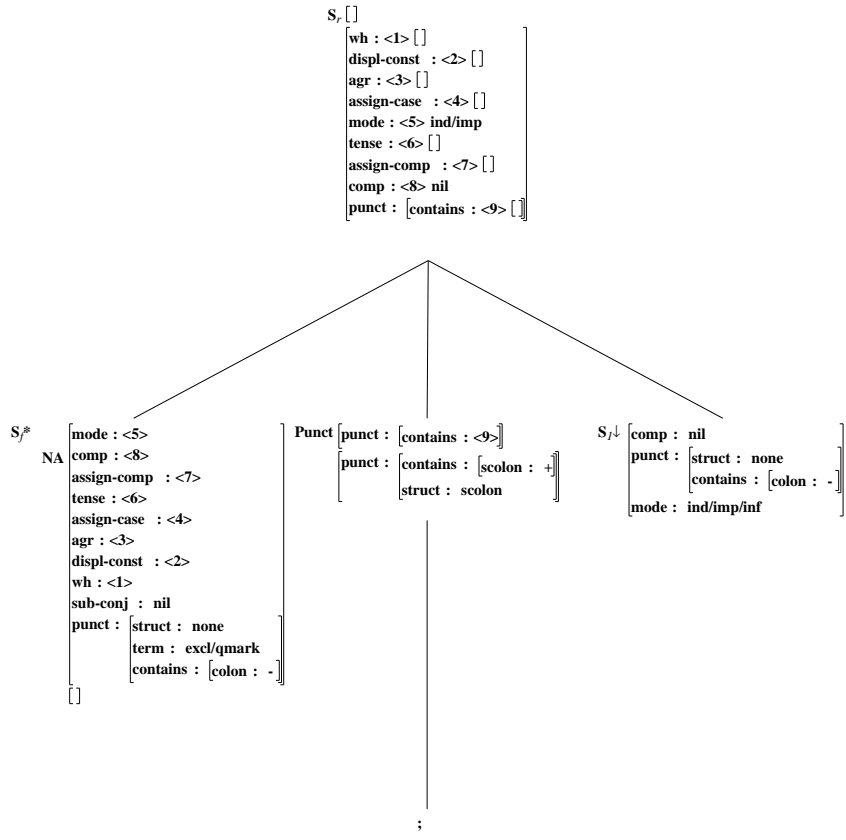,height=4.5in}
\caption{$\beta$sPUs, with features displayed}
\label{sPUs-tree}
\end{figure}

Complementizers are not permitted on either conjunct. Subordinating
conjunctions sometimes appear on the right conjunct, but seem to be
impossible on the left:
        
\enumsentence{Killpath would just have to go out and drag Gun back by
the heels once an hour ; because he'd be damned if he was going to
be a mid-watch pencil-pusher . [Brown:cl17]}

\enumsentence{The best rule of thumb for detecting corked wine
(provided the eye has not already spotted it) is to smell the wet end
of the cork after pulling it : if it smells of wine , the bottle is
probably all right ; if it smells of cork , one has grounds for
suspicion.      [Brown:cf27]}

\subsection{$\beta$sPU}

This tree handles the sentence final punctuation marks when selected
by a question mark, exclamation point or period. One could also
require a final punctuation mark for all clauses, but such an approach
would not allow non-periods to occur internally, for instance before a
semi-colon or dash as noted above in Section \ref{sPUs}. This tree
currently only adjoins to indicative or imperative (root) clauses.

\enumsentence{He left !}
\enumsentence{Get lost .}
\enumsentence{Get lost ?}

The feature {\bf punct bal= nil} on the foot node ensures that this
tree only adjoins inside of parentheses or quotes completely enclosing
a sentence (\ex{1}), but does not restrict it from adjoining to clause
which ends with balanced punctuation if only the end of the clause is
contained in the parentheses or quotes (\ex{2}).

\enumsentence{(John then left .)} 
\enumsentence{(John then left) .}
\enumsentence{Mary asked him to leave (immediately) .}

This tree is also selected by the colon to handle a colon expansion
after adjunct clause --

\enumsentence{Expressed differently : if the price for becoming a
faithful follower... \hfill [Brown:cd02]}

\enumsentence{Expressing it differently : if the price for becoming a
faithful follower... }

\enumsentence{To express it differently : if the price for becoming a
faithful follower... \hfill [Brown:cd02]}

This tree is only used after adjunct (untensed) clauses, which adjoin
to the tensed clause using the adjunct clause trees (cf Section
\ref{adjunct-cls} ); the {\bf mode} of the complete clause is that of
the matrix rather than the adjunct. Indicative or imperative
(i.e. root) clauses separated by a colon use the $\beta$sPUs tree
(Section \ref{sPUs}).

\subsection{$\beta$vPU}

This tree is anchored by a colon or a dash, and occurs between a verb
and its complement. These typically are lists.

\enumsentence{Printed material Available , on request , from U.S. Department of
Agriculture , Washington 25 , D.C. , are : Cooperative Farm Credit Can
Assist......\hfill [Brown:ch01]}

\subsection{$\beta$pPU}

This tree is anchored by a colon or a dash, and occurs between a
preposition and its complement. It typically occurs with a sequence
of complements. As with the tree above, this typically occurs with a
conjoined complement.

\enumsentence{...and utilization such as : (A) the protection of forage...}
\enumsentence{...can be represented as : Af.}

\section{Other trees}

\subsection{$\beta$spuARB}
\label{post-adverb}

In general, we attach post-clausal modifiers at the VP node, as you
typically get scope ambiguity effects with negation ({\it John didn't
leave today} -- did he leave or not?). However, with post-sentential,
comma-separated adverbs, there is no ambiguity - in {\it John didn't
leave, today} he definitely did not leave. Since this tree is only
selected by a subset of the adverbs (namely, those which can appear
pre-sententially, without a punctuation mark), it is anchored by the
adverb.

\enumsentence{The names of some of these products don't suggest the 
risk involved in buying them , either . \hfill [WSJ]}

\subsection{$\beta$spuPnx}
\label{post-PP}

Clause-final PP separated by a comma. Like the adverbs described
above, these differ from VP adjoined PPs in taking widest scope.

\enumsentence{...gold for current delivery settled at \$367.30 an
ounce , up 20 cents .} 

\enumsentence{It increases employee commitment to the company , with
all that means for efficiency and quality control .}

\subsection{$\beta$nxPUa}

Anchored by colon or dash, allows for post-modification of NPs by
adjectives.
        
\enumsentence{Make no mistake , this Gorky Studio drama is a
respectable import -- aptly grave , carefully written , performed and
directed . }

\part{Appendices}
\appendix
\chapter{Future Work}
\label{future-work}

\section{Adjective ordering}

At this point, the treatment of adjectives in the XTAG English grammar does not
include selectional or ordering restrictions.\footnote{This section is a repeat
of information found in section~\ref{adj-modifier}.} Consequently, any
adjective can adjoin onto any noun and on top of any other adjective already
modifying a noun. All of the modified noun phrases shown in (\ex{1})-(\ex{4})
currently parse.

\enumsentence{big green bugs}
\enumsentence{big green ideas}
\enumsentence{colorless green ideas}
\enumsentence{$\ast$green big ideas}

While (\ex{-2})-(\ex{0}) are all semantically anomalous, (\ex{0}) also suffers
from an ordering problem that makes it seem ungrammatical as well.  Since the
XTAG grammar focuses on syntactic constructions, it should accept
(\ex{-3})-(\ex{-1}) but not (\ex{0}).  Both the auxiliary and determiner
ordering systems are structured on the idea that certain types of lexical items
(specified by features) can adjoin onto some types of lexical items, but not
others.  We believe that an analysis of adjectival ordering would follow the
same type of mechanism.

\section{More work on Determiners}

In addition to the analysis described in Chapter~\ref{det-comparitives}, there
remains work to be done to complete the analysis of determiner constructions in
English.\footnote{This section is from \cite{ircs:det98}.}  Although
constructions such as determiner coordination are easily handled if
overgeneration is allowed, blocking sequences such as {\it one and some} while
allowing sequences such as {\it five or ten} still remains to be worked out.
There are still a handful of determiners that are not currently handled by our
system.  We do not have an analysis to handle {\it most}, {\it such}, {\it
certain}, {\it other} and {\it own}\footnote{The behavior of {\it own} is
sufficiently unlike other determiners that it most likely needs a tree of its
own, adjoining onto the right-hand side of genitive determiners.}.  In
addition, there is a set of lexical items that we consider adjectives ({\it
enough}, {\it less}, {\it more} and {\it much}) that have the property that
they cannot cooccur with determiners.  We feel that a complete analysis of
determiners should be able to account for this phenomenon, as well.

\section{{\it -ing} adjectives}

An analysis has already been provided for past participal ({\it -ed})
adjectives (as in sentence~ (\ex{1})), which are restricted to the
Transitive Verb family.\footnote{This analysis may need to be extended
to the Transitive Verb particle family as well.}  A similar analysis
needs to take place for the present participle~({\it -ing}) used as a
pre-nominal modifier.  This type of adjective, however, does not seem
to be as restricted as the~{\it -ed} adjectives, since verbs in other
tree families seem to exhibit this alternation as well
(e.g. sentences~(\ex{2}) and (\ex{3})).

\enumsentence{The murdered man was a doctoral student at UPenn .}
\enumsentence{The man died .}
\enumsentence{The dying man pleaded for his life .}

\section{Verb selectional restrictions}

Although we explicitly do not want to model semantics in the XTAG grammar,
there is some work along the syntax/semantics interface that would help reduce
syntactic ambiguity and thus decrease the number of semantically anomalous
parses.  In particular, verb selectional restrictions, particularly for PP
arguments and adjuncts, would be quite useful.  With the exception of the
required {\it to} in the Ditransitive with PP Shift tree family (Tnx0Vnx1Pnx2),
any preposition is allowed in the tree families that have prepositions as their
arguments.  In addition, there are no restrictions as to which prepositions are
allowed to adjoin onto a given verb.  The sentences in (\ex{1})-(\ex{3}) are
all currently accepted by the XTAG grammar.  Their violations are stronger than
would be expected from purely semantic violations, however, and the presence of
verb selectional restrictions on PP's would keep these sentences from being
accepted.

\enumsentence{\#survivors walked of the street .}
\enumsentence{\#The man about the earthquake survived .}
\enumsentence{\#The president arranged on a meeting .}

\section{Thematic Roles}

Elementary trees in TAGs capture several notions of locality, with the most
primary of these being locality of $\theta$-role assignment.  Each elementary
tree has associated with it the $\theta$-roles assigned by the anchor of that
elementary tree.  In the current XTAG system, while the notion of locality of
$\theta$-role assignment within an elementary tree has been implicit, the
$\theta$-roles assigned by a head have not been explicitly represented in the
elementary tree. Incorporating $\theta$-role information will make the
elementary trees more informative and will enable efficient pruning of spurious
derivations when embedded into a specific context.  In the case of a
Synchronous TAG, $\theta$-roles can also be used to automatically establish
links between two elementary trees, one in the object language and one in the
target language.

\chapter{Metarules}
\label{metarules}

\section{Introduction}

XTAG has now a collection of functions accessible from the user interface that
helps the user in the construction and maintenance of a tag tree-grammar.
This subsystem is based on the idea of metarules (\cite{becker93}).
Here our primary
purpose is to describe the facilities implemented under this 
metarule-based subsystem.
For a discussion of the metarules as a method for compact representation 
of the Lexicon see \cite{becker93} and \cite{srini94}.

The basic idea of using metarules is to take profit of the similarities of the
relations involving related pairs of XTAG elementary trees. 
For example, in the English grammar described in this technical report,
comparing the XTAG trees for the basic form and the 
wh-subject moved form, the relation between this two trees for transitive verbs
($\alpha nx_0Vnx_1$, $\alpha W_0nx_0Vnx_1$) is similar to the relation for the
intransitive verbs ($\alpha nx_0V$, $\alpha W_0nx_0V$) and also to the 
relation for the ditransitives ($\alpha nx_0Vnx_1nx_2$, 
$\alpha W_0nx_0Vnx_1nx_2$). Hence, instead of generating by hand the six trees 
mentioned above, a more natural and robust way would be generating by hand 
only the basic trees for the intransitive, transitive and ditransitive cases, 
and letting the wh-subject moved trees to be automatically generated by the
application of a unique transformation rule that would account exactly for
the identical relation involved in each of the three pairs above.

Notice that the degree of generalization can be much higher than it might be
thought in principle from the above paragraph. For example, once a rule for
passivization is applied to the tree different basic trees above, the 
wh-subject moved rule could be again applied to generate the wh-moved subject
versions for the passive form. Depending on the degree of regularity that one
can find in the grammar being built, the reduction in the number of original
trees can be exponential. 

We still make here a point that the reduction of effort in grammar construction
is not the only advantage of the approach. Robustness, 
reliability and maintainability of the grammar achieved by the use of  
metarules are equally or even more important.

In the next section we define a metarule in XTAG. 
Section 3 gives some
linguistically motivated examples
of metarule for the English grammar described in this
technical report and their application. Section 4 describes the access through 
the user interface.

\section{The definition of a metarule in XTAG}

A metarule specifies a rule for transforming grammar rules into grammar rules. 
In XTAG the grammar 
rules are lexicalized trees. Hence an XTAG metarule {\bf mr} 
is a pair {\bf (lhs, rhs)} of XTAG trees, where:

\begin{itemize}
\item {\bf lhs}, the {\it left-hand side} of the metarule, is a pattern tree,
        i.e., it is intended to present a specific pattern of tree to look for
        in the trees submitted to the application of the metarule.

\item When a metarule {\bf mr} is applied to an input tree {\bf inp}, the first
        step is to verify if the input tree matches the pattern specified by
        the {\bf lhs}. If there is no match, the application {\it fails}.

\item {\bf rhs}, the {\it right-hand side} of the metarule, specifies (together
        with {\bf lhs}) the transformation that will be done in {\bf inp},
        in case of successful matching, thus generating the output tree of
        the metarule application\footnote{actually more than one output tree 
        can be generated from the successful application of a rule to an 
        input tree, as will be seen soon}.
\end{itemize}
 
\subsection{Node names, variable instantiation, and matches}

We will use the terms ({\bf lhs}, {\bf rhs} and {\bf inp}) as introduced above
to refer to the parts of a generic metarule being applied to an input tree. 

The nodes at {\bf lhs} can take three 
different forms: a constant node, a typed variable node, and a non-typed 
variable node. The naming conventions for these different classes of nodes is 
given below.

\begin{itemize}

\item {\bf Constant Node:} Its name must not initiate by a question mark
        (`?' character). They are like we expect for names to be in normal
        XTAG trees; for instance, {\bf inp} is expected to have only constant
        nodes. Some examples of constant nodes are $NP$, $V$, $NP_0$, $NP_1$,
        $S_r$. We will call the two parts that compose such names
        the {\it stem} and the {\it subscript}.
        In the examples above  $NP$, $V$ and $S$ are stems and 
        $0$, $1$, $r$ are subscripts. Notice that the
        subscript part can also be empty as in two of the above examples.

\item {\bf Non-Typed Variable Node:} Its name initiates by a question 
        mark (`?'), followed by a sequence of digits (i.e. a number) which 
        uniquely identifies the variable. Examples: ?1, ?3, 
        ?3452\footnote{Notice
        however that having the sole purpose of distinguishing between 
        variables, a number like the one in the last example 
        is not very likely
        to occur, and a metarule with more than 
        three thousand variables can give
        you a place in the Guinness TagBook of Records.}. We assume that
        there is no stem and no subscript in this names, i.e., `?' is just
        a meta-character to introduce a variable, and the number is the 
        variable identifier. 

\item {\bf Typed Variable Node:} Its name initiates by a question mark (`?')
        followed by a sequence of digits, but is additionally followed by
        a {\it type specifiers definition}. A {\it type specifiers definition}
        is a sequence of one or more {\it type specifier} separated by a slash
        (`/'). A {\it type specifier} has the same form of a regular XTAG node
        name (like the constant nodes), except that the subscript can be also
        a question mark. Examples of typed variables are:
        $?1VP$ (a single type specifier with stem $VP$ and no subscript), 
        $?3NP_1/PP$ (two type specifiers, $NP_1$ and $PP$),
        $?1NP_?$ (one type specifier, $NP_?$ with undetermined subscript). 
        We'll see ahead that each type specifier represents an alternative
        for matching, and the presence of `?' in subscript position of a
        type specifier means that matching will only check for the stem
        \footnote{This is different from not having a subscript which is 
        interpreted as checking that the input tree have no subscript 
        for matching}.
\end{itemize}

During the process of matching, variables are associated (we use the
term {\it instantiated}) with `tree material'.  According to its class
a variable can be instantiated with different kinds of tree material:

\begin{itemize}
\item   A typed variable will be instantiated with exactly one node of
        the input tree, which is in accordance to one of its type specifiers
        (The full rule is in the following subsection). 

\item   A non-typed variable will be instantiated with a range of subtrees.
        These subtrees will be taken from one of the nodes of the input tree
        {\bf inp}. Hence, there will a node $n$ in {\bf inp}, with subtrees
        $n.t_1$, $n.t_2$, ..., $n.t_k$, in this order, where the variable
        will be instantiated with some subsequence of these subtrees 
        (e.g., $n.t_2$, $n.t_3$, $n.t_4$). Note however, that some of these
        subtrees, may be incomplete, i.e., they may not go all the way to the 
        bottom leaves. Entire subtrees may be removed. Actually for each
        child of the non-typed variable node, one subtree that matches this
        child subtree will be removed from some of the $n.t_i$(maybe an entire
        $n.t_i$), leaving in place a mark for inserting material during the
        substitution of occurences at {\bf rhs}.\\
        Notice still that the variable can
        be instantiated with a single tree and even with no tree. 

\end{itemize} 

We define a {\it match} to be a complete instantiation of all variables 
appearing in the metarule. In the process of matching, there may be several
possible ways of instantiating the set of variables of the metarule, i.e.,
several possible matches. This is due to the presence of non-typed variables.

Now, we are ready to define what we mean by a successful matching. The process
of matching is {\it successful} 
if the number of possible matches is greater then 0.
When there is no possible match the process is said to {\it fail}.
In addition to return success or failure, the process also return the set of
all possible {\it matches}, which will be used for generating the output.

\subsection{Structural Matching}

The process of matching {\bf lhs} and {\bf inp} can be seen as a recursive 
procedure for matching trees, starting at their roots and proceeding in a 
top-down style along with their subtrees. 
In the explanation of this process that 
follows we have used the term {\bf lhs} not only to refer to the whole tree 
that contains the pattern 
but to any of its subtrees that is being considered in a 
given recursive step. The same applies to {\bf inp}. 
By now we ignore feature equations,
which will be accounted for in the next subsection.

The process described below returns 
at the end the set of matches (where an empty set means the same 
as failure). We first give one auxiliary definition, of valid Mapping, and
one recursive function Match, that matches lists of trees instead of trees,
and then define the process of matching two trees as a special case of
call to Match.

Given a list $list_{lhs}=[lhs_1, lhs_2, ..., lhs_l]$ of nodes of {\bf lhs}
and a list $list_{inp}=[inp_1, inp_2, ..., inp_i]$ of nodes of {\bf inp},
we define a {\it mapping} from $list_{lhs}$ to $list_{inp}$ to be a function
$Mapping$,
that for each element of $list_{lhs}$ assigns a list of elements of 
$list_{inp}$, defined by the following condition:
$$concatenation\ (Mapping(lhs_1),\ Mapping(lhs_2),\ ...,\ Mapping(lhs_l))\ =\ 
        list_{inp}$$,
i.e., the elements of $list_{inp}$ are split into sublists and assigned in 
order of appearance in the list to the elements of $list_{lhs}$.

We say that a mapping is a {\it valid mapping} if for all $j$, $1\leq j \leq l$
(where $l$ is the length of $list_{lhs}$), the following restrictions apply:

\begin{enumerate}

\item if $lhs_j$ is a constant node, then $Mapping(lhs_j)$ must have a 
        single element, say, $rhs_{g(j)}$, and the two nodes must have the same
        name and agree on the markers (foot, substitution, head and NA), i.e.,
        if $lhs_j$ is NA, then $rhs_{g(j)}$ must be NA, 
        if $lhs_j$ has no markers, then $rhs_{g(j)}$ must have no markers, etc.

\item if $lhs_j$ is a type variable node, then $Mapping(lhs_j)$ must have a 
        single element, say, $rhs_{g(j)}$, and $rhs_{g(j)}$ must be 
        {\it marker-compatible} and 
        {\it type-compatible} with $lhs_j$. \\
        $rhs_{g(j)}$ is 
        {\it marker-compatible} with $lhs_j$ if any marker
         (foot, substitution, head and NA) present in $lhs_j$ is also
        present in $rhs_{g(j)}$\footnote{Notice that, unlike the case for the
        constant node, the inverse is not required, 
        i.e., if $lhs_j$ has no marker, $rhs_{g(j)}$ is still 
        allowed to have some.}.\\
        $rhs_{g(j)}$ is {\it type-compatible} with $lhs_j$ 
        if there is at least one of the alternative 
        type specifiers for the typed variable that satisfies 
        the conditions below. 

\begin{itemize}
\item   $rhs_{g(j)}$ has the stem defined in the type specifier.
\item   if the type specifier doesn't have subscript, then 
        $rhs_{g(j)}$ must have no subscript.
\item   if the type specifier has a subscript different of `?', then 
        $rhs_{g(j)}$ must have the same subscript as in the type specifier
        \footnote{If the type specifier has a `?' subscript, there is no
        restriction, and that is exactly its function: to allow for the 
        matching to be independent of the subscript}.
\end{itemize}

\item if $lhs_j$ is a non-typed variable node, then there's actually no
        requirement: $Mapping(lhs_j)$ may have any length and even be 
        empty.
\end{enumerate}
        
The following algorithm, Match, takes as input a list of nodes of {\bf lhs}
and a list of nodes of {\bf inp}, and returns the set of possible matches
generated in the attempt of match this two lists. If the result is an empty
set, this means that the matching failed.

\begin{tabbing}
012\=0123\=0123\=0123\=0123\=0123\=0123\=0123\=0123\=0123\=0123\=0123\=012\kill

\> Function Match ($list_{lhs}$, $list_{rhs}$) \\
\>\> Let $MAPPINGS$ be the list of all valid mappings from $list_{lhs}$ to 
        $list_{rhs}$ \\

\>\> Make $MATCHES=\emptyset$ \\

\>\> For each mapping $Mapping\in MAPPINGS$ do: \\

\>\>\> Make $Matches=\{\emptyset \}$ \\

\>\>\> For each $j$, $1 \leq j \leq l$, where $l=length(list_{lhs})$, do: \\

\>\>\>\> if $lhs_j$ is a constant node, then \\

\>\>\>\>\> let \>$children_{lhs}$ be the list of children of $lhs_j$ \\
\>\>\>\>\>     \>$lhr_{g(j)}$ be the single element in $Mapping(lhs_j)$ \\
\>\>\>\>\>     \>$children_{rhs}$ be the list of children of $lhr_{g(j)}$ \\

\>\>\>\>\> Make $Matches=\{m\cup m_j\ \mid\  m\in Matches$ \\
\>\>\>\>\>\>\>\>\> $and\ m_j\in$ Match$(children_{lhs},\ children_{rhs})\}$ \\ 

\>\>\>\> if $lhs_j$ is a typed variable node, then \\

\>\>\>\>\> let \>$children_{lhs}$ be the list of children of $lhs_j$ \\
\>\>\>\>\>     \>$lhr_{g(j)}$ be the single element in $Mapping(lhs_j)$ \\
\>\>\>\>\>     \>$children_{rhs}$ be the list of children of $lhr_{g(j)}$ \\

\>\>\>\>\> Make $Matches=\{\{(lhs_j,lhr_{g(j)})\}\cup
        m\cup m_j\ \mid\  m\in Matches$ \\
\>\>\>\>\>\>\>\>\> $and\ m_j\in$ Match$(children_{lhs},\ children_{rhs})\}$ \\ 

\>\>\>\> if $lhs_j$ is a non-typed variable node, then \\

\>\>\>\>\> let \>$children_{lhs}$ be the list of children of $lhs_j$ \\
\>\>\>\>\> \> $sl$ be the number of nodes in $children_{lhs}$ \\

\>\>\>\>\>\> $DESC_s$ be the set of s-size lists given by: \\
\>\>\>\>\>\>\>\> $DESC_s=\{[dr_1,dr_2,...,dr_s]\ \mid\ $ \\
\>\>\>\>\>\>\>\>\>\> for every $1 \leq k \leq s$, $dr_k$ is a descendant \\
\>\>\>\>\>\>\>\>\>\>\>\> of
                        some node in $Mapping(lhs_j)\}$\footnote{it's not 
                necessary to be a proper descendant,
                i.e., $dr_k$ may be a node in $Mapping(lhs_j)$}\\
\>\>\>\>\>\>\>\>\>\> for every $1 < k \leq s$, $dr_l$ is {\it to the right of}
                        $dr_{k-1}$\footnote{recall that a node 
$n$ is to the right of a node $m$, if $n$ and $m$ are not descendant of each
other, and all the leaves dominated by $n$ are to the right of the leaves
dominated by $m$.}.\\

\>\>\>\>\>\> For every list $Desc=[dr_1,dr_2,...,dr_s] \in DESC_s$ do: \\

\>\>\>\>\>\>\> Let Tree-Material be the list of subtrees dominated \\
\>\>\>\>\>\>\>\>\> by the nodes in $Mapping(lhs_j)$, but, with the \\
\>\>\>\>\>\>\>\>\> subtrees dominated by the nodes in $DESC_s$ \\
\>\>\>\>\>\>\>\>\> cut out from these trees \\

\>\>\>\>\>\>\> Make $Matches=\{\{(lhs_j,\ Tree-Struct)\}\ \cup 
        m\cup m_j\ \mid$ \\
\>\>\>\>\>\>\>\>\> $m\in Matches\ and\ 
        m_j\in$ Match$(children_{lhs},\ Desc)\}$ \\ 

\>\>\> Make $MATCHES\ =\ MATCHES\ \cup\ Matches$ \\

\>\> Return $MATCHES$

\end{tabbing}

Finally we can define the process of structurally matching {\bf lhs} to
{\bf inp} as the evaluation of Match([root({\bf lhs})], [root({\bf inp})].
If the result is an empty set, the matching failed, otherwise the resulting
set is the set of possible matches that will be used for generating the
new trees (after being pruned by the feature equation matching).

\subsection{Output Generation}
\label{output-gen}

Although nothing has yet been said about the feature
equations, which is the subject of the next subsection, we assume that only
matches that meet the additional constraints imposed by feature equations
are considered for output. If no structural match survives feature equations
checking, that matching has failed.

If the process of matching {\bf lhs} to {\bf inp} fails, there are two 
alternative behaviors according to the value of a parameter\footnote{the
parameter is accessible at the Lisp interface by the name 
{\it XTAG::*metarules-copy-unmatched-trees*}}. 
If the parameter is set to false, which is the {\it default} value, 
no output is generated. 
On the other hand, if 
it is set to true, then the own {\bf inp} tree is copied to the output.

If the process of matching succeeds, as many trees will be generated in the
output as the number of possible matches obtained in the process. For a 
given match, the output tree is generated by substituting in the {\bf rhs} tree
of the metarule the occurrences of variables by the material to which they have
been instantiated in the match. The case of the typed-variable is simple. 
The name of the variable is just substituted by the name of the node to which
it has been instantiated from {\bf inp}. A very important detail is that the
marker (foot, substitution, head, NA, or none) at the output tree node comes 
from what is specified in the {\bf rhs} node, which can be different of the
marker at the variable node in {\bf inp} and of the associated node from 
{\bf inp}.

The case of the non-typed variable, not surpringly, is not so simple. 
In the output tree, this node
will be substituted by the subtree list that was associated to this node,
in the same other, attaching to the parent of this non-typed variable node.
But remember, that some subtrees may have been removed from some of the trees
in this list, maybe entire elements of this list, due to the effect of the
children of the metavariable in {\bf lhs}. 
It is a 
requirement that any occurence of a non-typed variable node at the {\bf rhs}
tree has exactly the same number of children than the unique occurence of
this non-typed variable node in {\bf lhs}. Hence, when generating the output
tree, the subtrees at {\bf rhs}
will be inserted exactly at the points where subtrees were removed during 
matching, in a positional, one to one correspondance.

For feature equations in the output trees see the
next subsection. The comments at the output are the comments at the {\bf lhs}
tree of the metarule followed by the coments at {\bf inp}, both parts 
introduced by appropriate headers, allowing the user to have a complete
history of each tree.

\subsection{Feature Matching}

In the previous subsections we have considered only the aspects of a metarule
involving the structural part of the XTAG trees. In a feature based grammar
as XTAG is, accounting for features is essential. A metarule is not really
worth if it doesn't account for the proper change of feature 
equations\footnote{Notice that what is really important is not the features
themselves, but the feature equations that relate the feature values of nodes
of the same tree} from the input to the output tree. 
The aspects that have to be considered here are:

\begin{itemize}
\item   Which feature equations should be required to be present in {\bf inp}
        in order for the match to succeed.

\item   Which feature equations should be generated in the output tree as a 
        function of the feature equations in the input tree.
\end{itemize}

Based on the possible combinations of these requirements we partition the 
feature equations into the following five classes\footnote{This 
classification is really a partition, i.e., no equation may be conceptually
in more than one class at the same time.}: 

\begin{itemize}
\item   {\it Require \& Retain:} Feature equations in this class
        are required to be in {\bf inp} in order for matching to succeed.
        Upon matching, these equations will be copied to the output tree.
        To achieve this behaviour, the equation must be placed in 
        the {\bf lhs} tree of the metarule preceded by a plus character
        (e.g. $+V.t:<trans>=+$)\footnote{Commutativity of equations is 
        accounted for in the system. Hence an equation $x=y$ can also be
        specified as $y=x$. Associativity is not accounted for and its need by
        an user is viewed as indicating misspecification at the input trees.}
        
\item   {\it Require \& Don't Copy:} The equation is required to be in 
        {\bf inp}
        for matching, but should not be copied to the output tree.
        Those equations must be in {\bf lhs} preceded by minus character
        (e.g. $-NP_1:<case>=acc$).

\item   {\it Optional \& Don't Copy:} 
        The equation is not required for matching,
        but we have to make sure not to copy it to the output tree set of
        equations, regardless of it being present or not in {\bf inp}.
        Those equations must be in {\bf lhs} in raw form, i.e. neither preceded
        by a plus nor minus character
        (e.g. $S_r.b:<perfect>=VP.t:<perfect>$).

\item   {\it Optional \& Retain:} 
        The equation is not required for matching but,
        in case it is found in {\bf inp} it must be copied to the output tree.
        This is the {\it default} case, and hence these equations should not be
        present in the metarule specification.

\item   {\it Add:} The equation is not required for matching but we want it to
        be put in the output tree anyway.
        These equations are placed in raw form in the {\bf rhs} (notice in this
        case it is the right hand side).
\end{itemize}

Typed variables can be used in feature equations in both {\bf lhs} and 
{\bf rhs}. They are intended to represent the nodes of the input tree to which
they have been instantiated. For each resulting match 
from the structural matching process the following is done:

\begin{itemize}
\item   The (typed) variables in the equations at {\bf lhs} and {\bf rhs} are 
        substituted by the names of the nodes they have been instantiated to.

\item   The requirements concerning feature equations are checked, according
        to the above rules.

\item   If the match survives feature equation checking, the proper output tree
        is generated, according to Section~\ref{output-gen} and to the 
        rules described above for the feature equations.
\end{itemize}

Finally, a new kind of metavariable, which is not used at the nodes, can be
introduced in the feature equations part. They have the same form of the
non-typed variables, i.e. quotation mark, followed by a number, and are used
in the place of feature values and feature names. Hence, if the equation
$?NP_?.b:<?2> = ?3$ appears in {\bf lhs}, this means, that all feature
equations of {\bf inp} that match a bottom attribute of some $NP$ to any 
feature value (but not to a feature path) will not be copied to the output.

\setcounter{topnumber}{4}
\setcounter{bottomnumber}{4}
\setcounter{totalnumber}{4}

\section{Examples}

Figure~\ref{wh-subj} shows a metarule for wh-movement of the subject. Among
the trees to which it have been applied are the basic trees of intransitive, 
transitive and ditransitive families (including prepositional complements),
passive trees of the same families, and ergative.

\begin{figure}[htb]
\begin{center}
\begin{tabular}{c@{\hspace{2em}}c}
\framebox{\psfig{figure=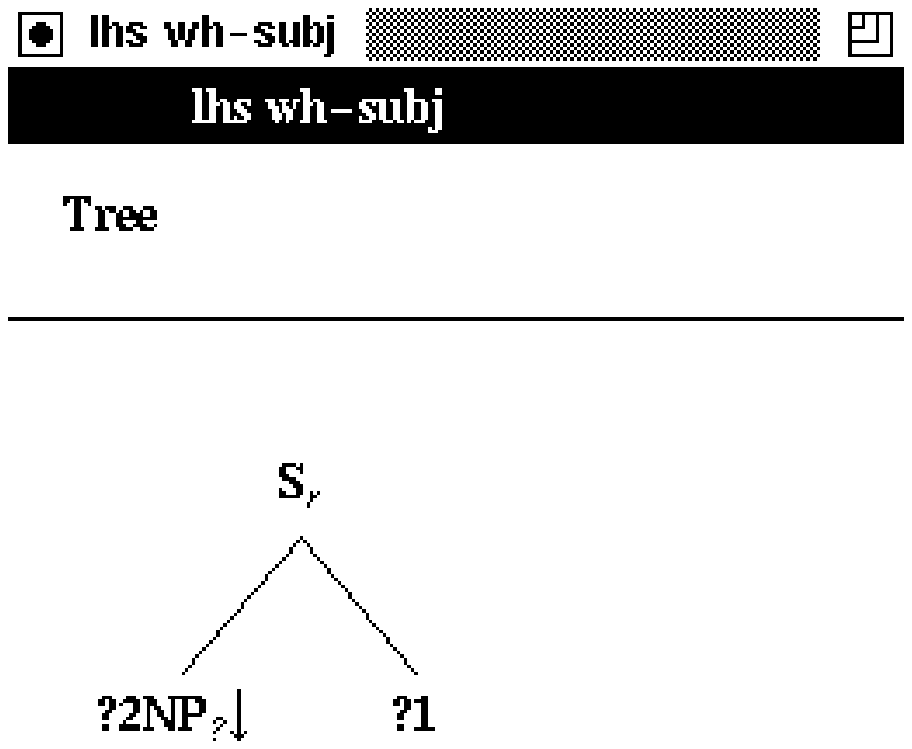,height=4cm}} &
\framebox{\psfig{figure=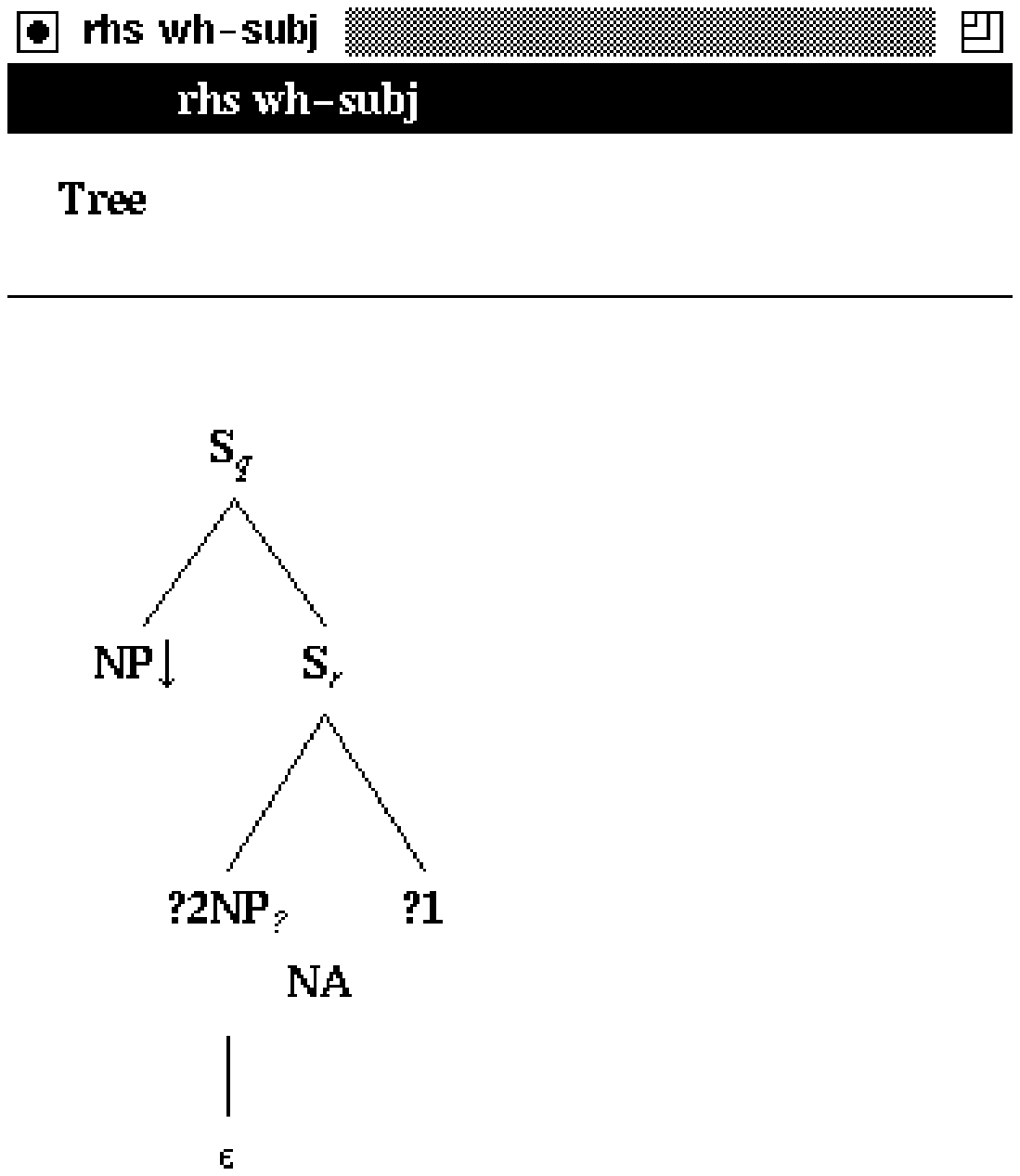,height=4cm}} \\
{lhs} & {rhs} \\
\end{tabular}
\end{center}
\caption{Metarule for wh-movement of subject}
\label{wh-subj}
\end{figure}

Figure~\ref{wh-obj} shows a metarule for wh-movement of an NP in object
position. Among
the trees to which it have been applied are the basic and passive trees of  
transitive and ditransitive families.

\begin{figure}[!htb]
\begin{center}
\begin{tabular}{c@{\hspace{2em}}c}
\framebox{\psfig{figure=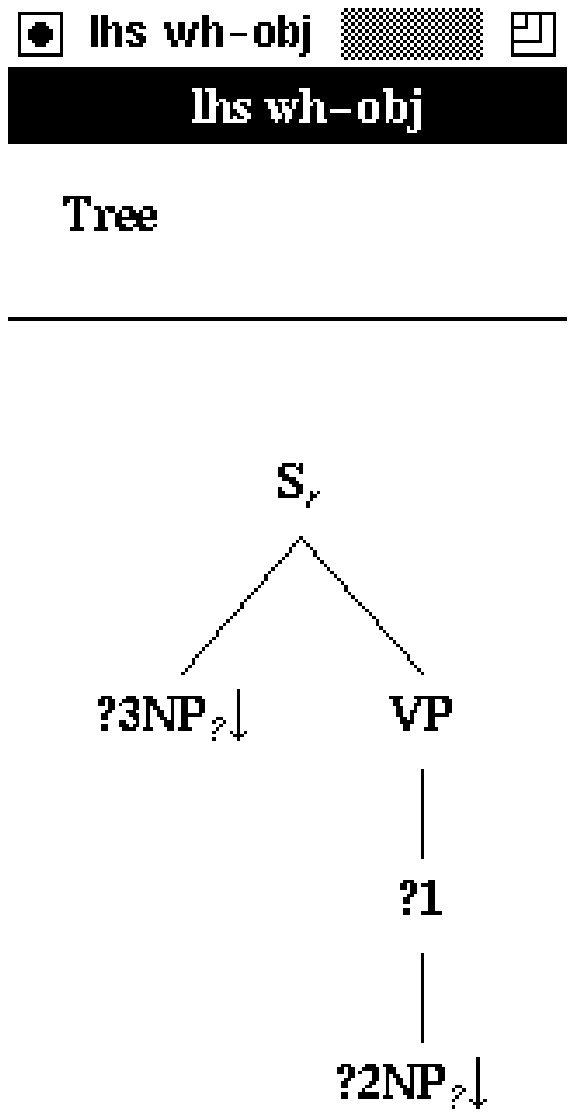,height=4cm}} &
\framebox{\psfig{figure=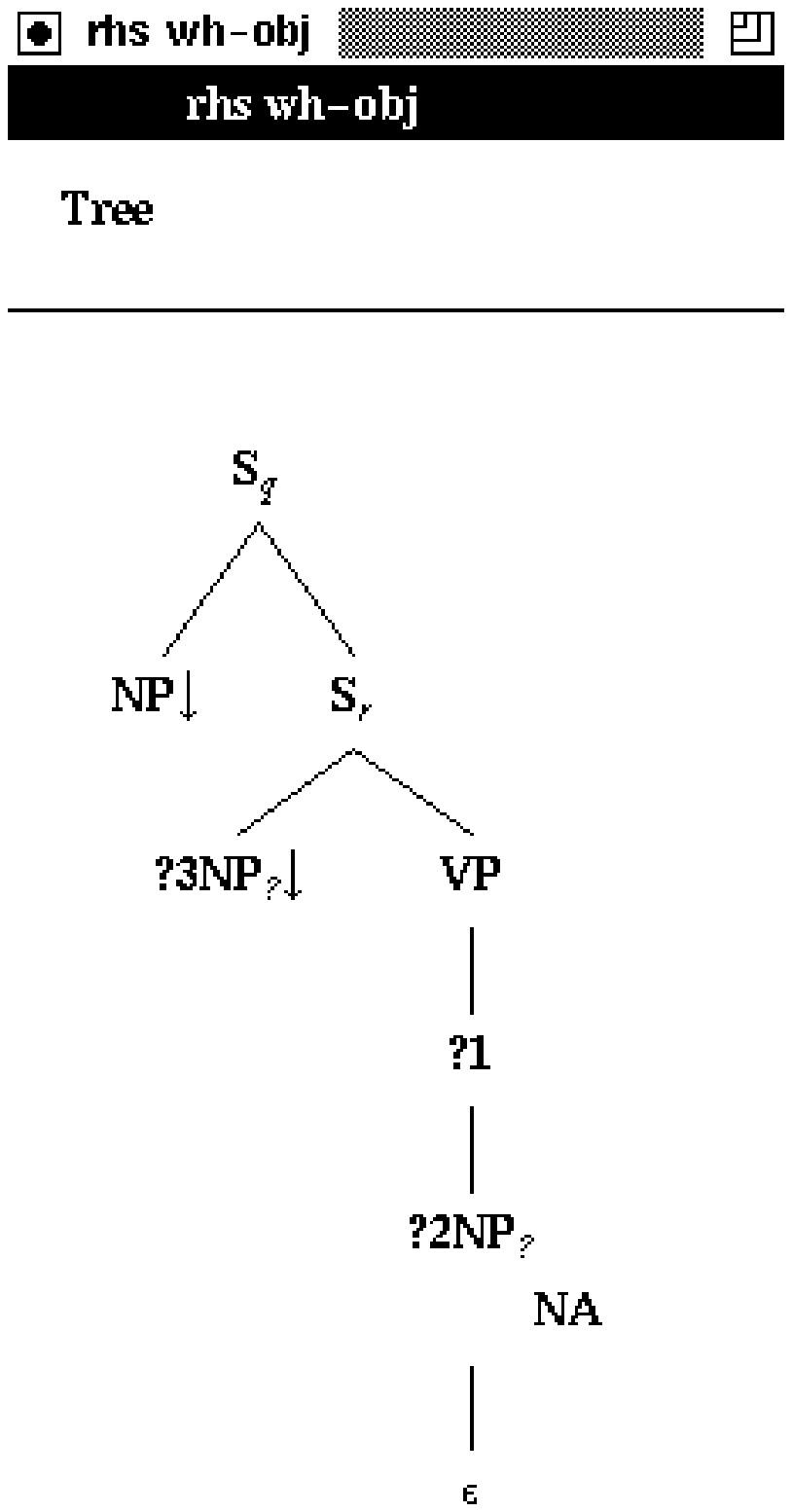,height=4cm}} \\
{lhs} & {rhs} \\
\end{tabular}
\end{center}
\caption{Metarule for wh-movement of object}
\label{wh-obj}
\end{figure}

Figure~\ref{wh} shows a metarule for general wh-movement of an NP. 
It can be applied to
generate trees with either subject or object NP moved. We show in 
Figure~\ref{prep}, the basic tree for the family Tnx0Vnx1Pnx2 and the tree
wh-trees generated by the application of the rule.

\begin{figure}[!htb]
\begin{center}
\begin{tabular}{c@{\hspace{2em}}c}
\framebox{\psfig{figure=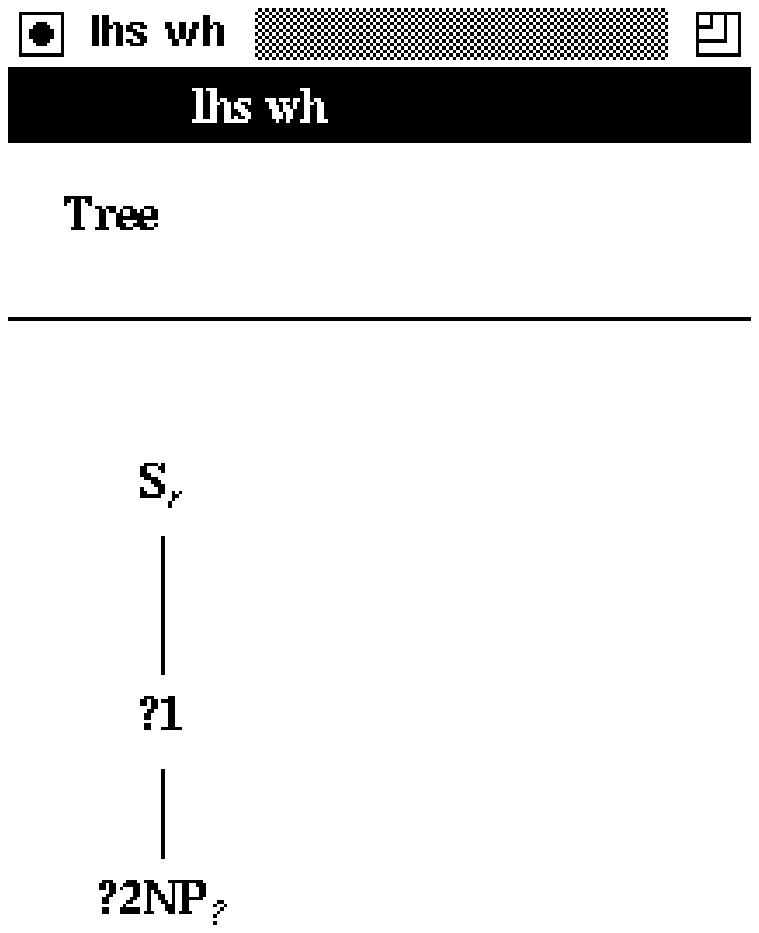,height=4cm}} &
\framebox{\psfig{figure=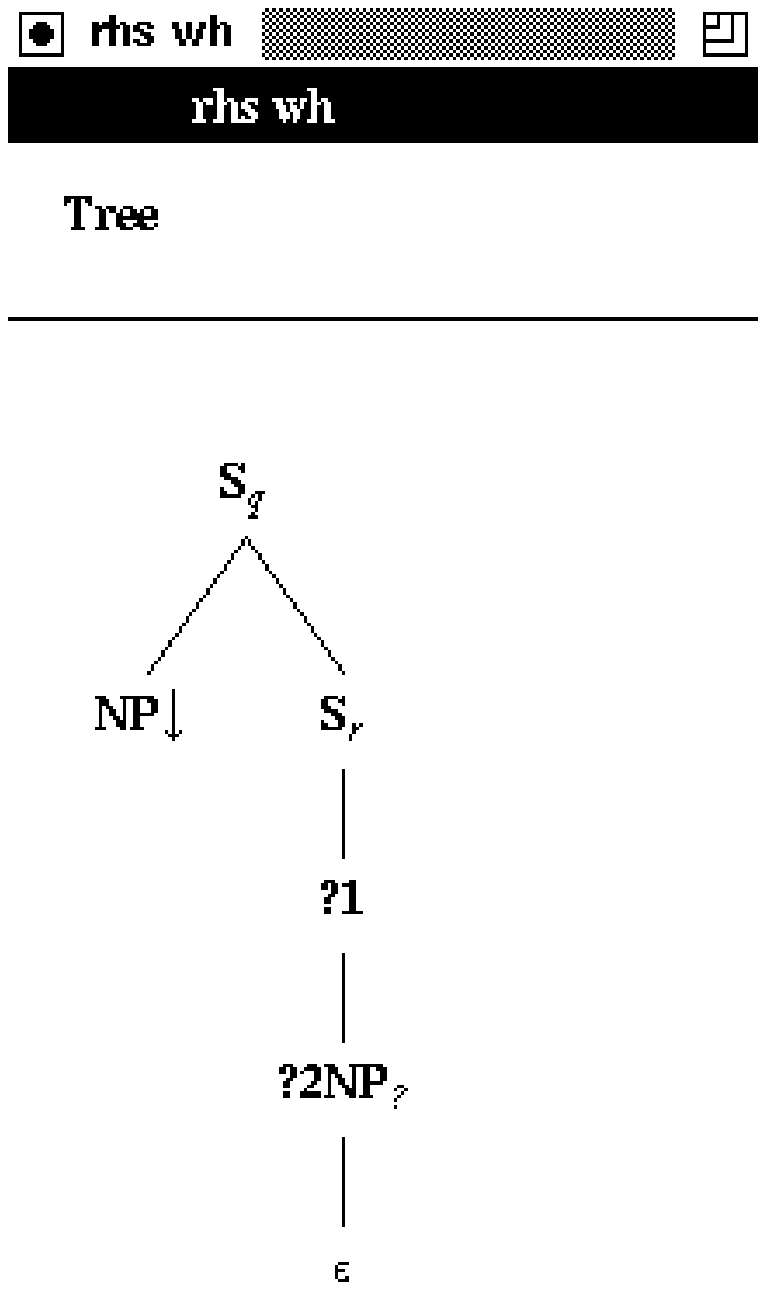,height=4cm}} \\
{lhs} & {rhs} \\
\end{tabular}
\end{center}
\caption{Metarule for general wh movement of an NP}
\label{wh}
\end{figure}

\begin{figure}[!htb]
\begin{center}
\begin{tabular}{c@{\hspace{2em}}c}
\framebox{\psfig{figure=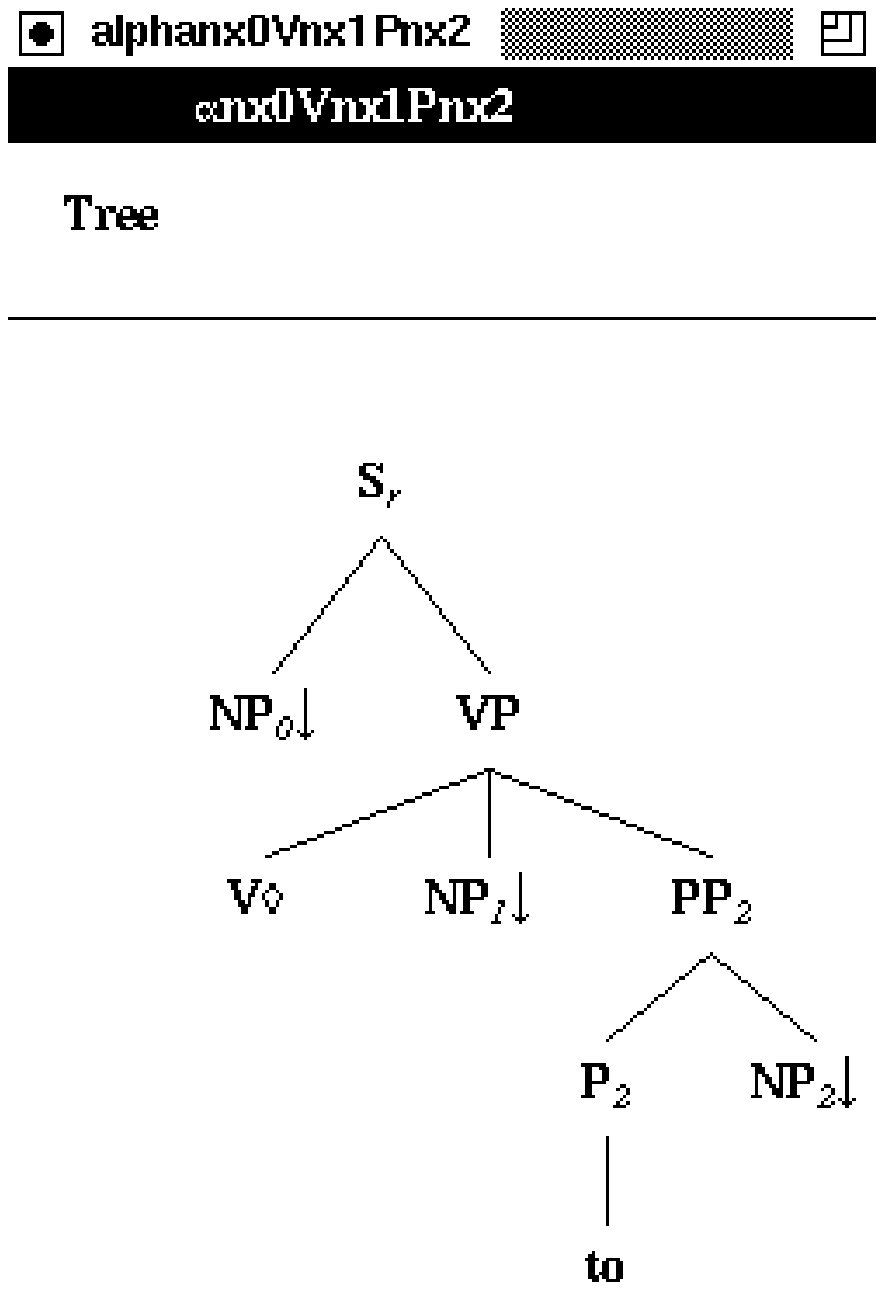,height=4cm}} &
\framebox{\psfig{figure=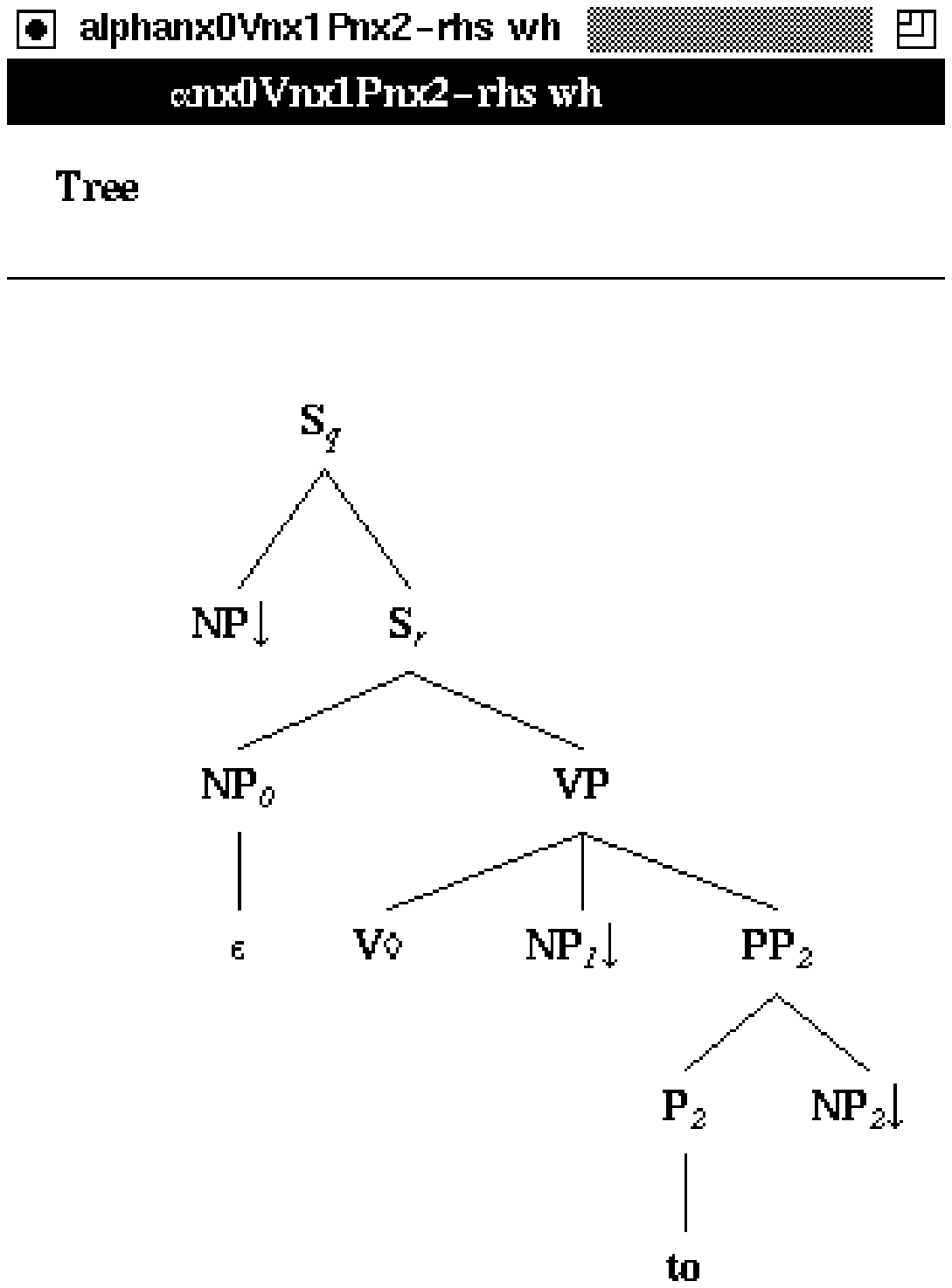,height=4cm}} \\
{Tnx0Vnx1Pnx2} & {subject moved} \\
\\
\framebox{\psfig{figure=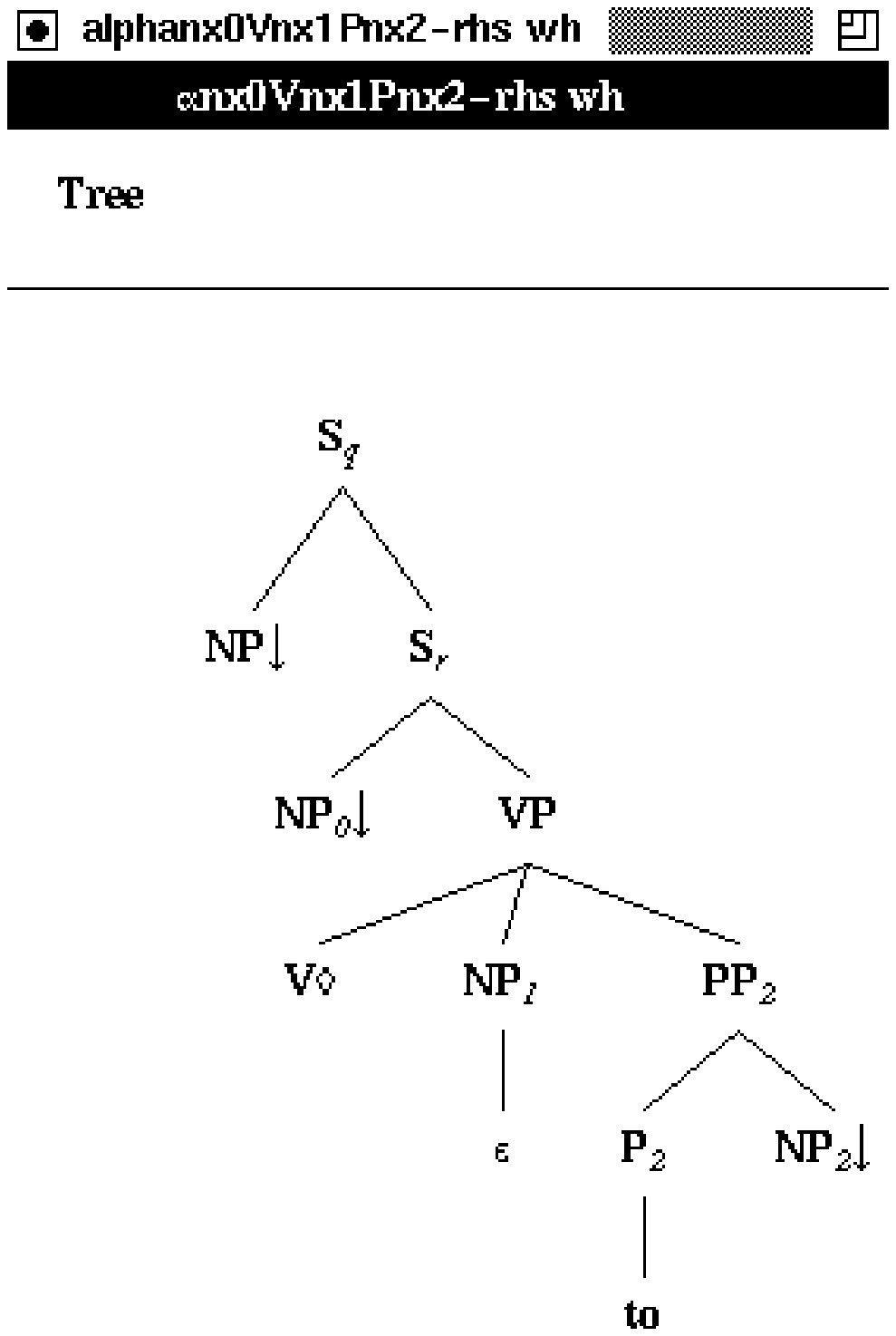,height=4cm}} &
\framebox{\psfig{figure=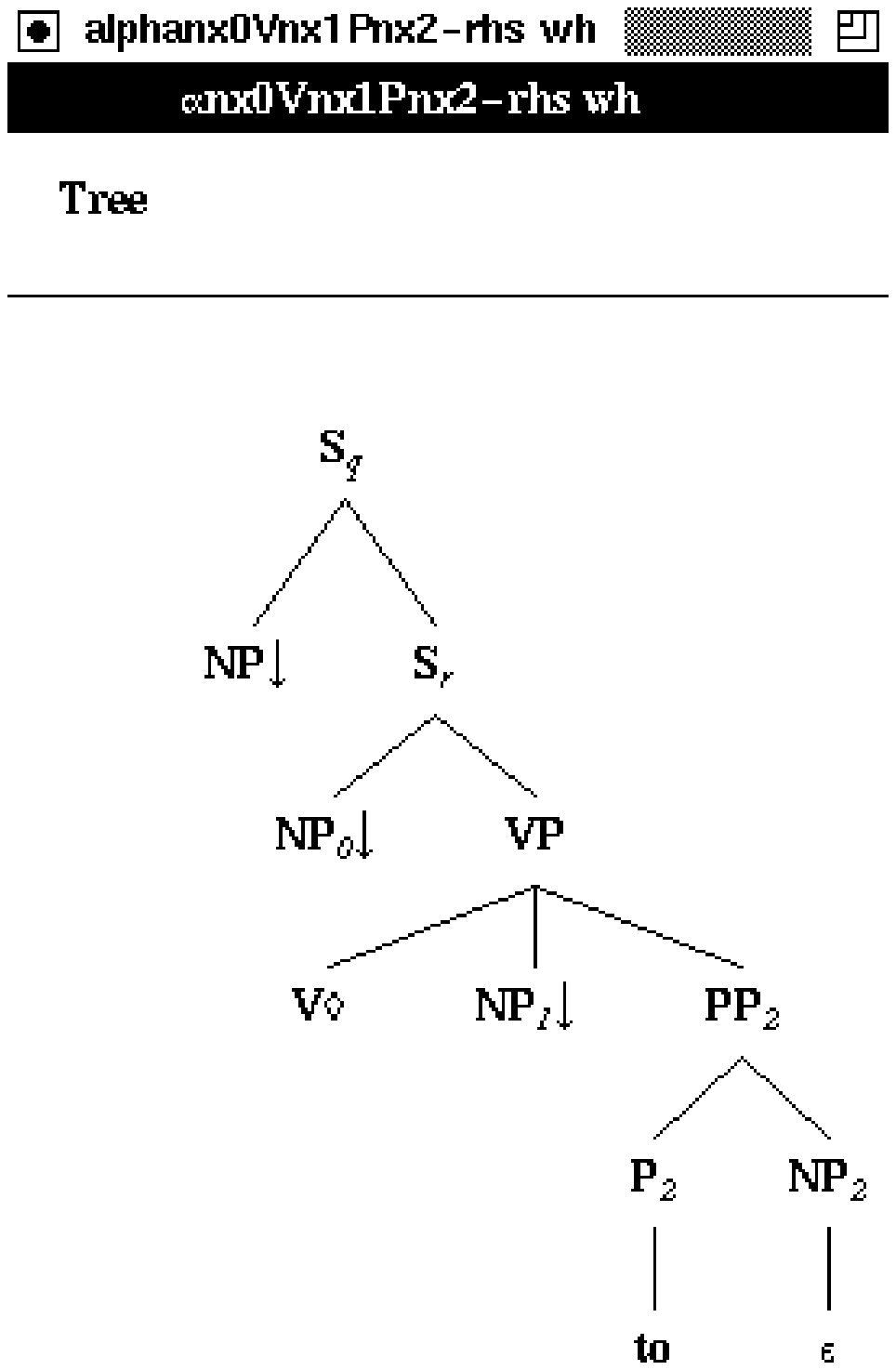,height=4cm}} \\
{NP object moved} & {NP object moved from PP} \\
\end{tabular}
\end{center}
\caption{Application of wh-movement rule to Tnx0Vnx1Pnx2}
\label{prep}
\end{figure}

\section{The Access to the Metarules through the XTAG Interface}

We first describe the access to the metarules subsystem using buffers with
single metarule applications. Then we proceed by describing the application of 
multiple metarules in what we call the parallel, sequential, and cumulative
modes to input tree files.

We have defined conceptually a metarule as an ordered pair of trees. 
In the implementation
of the metarule subsystem it works the same: a metarule is a buffer with two
trees. The name of the metarule is the name of the buffer. The first tree that
appear in the main window under the metarule 
buffer is the {\it left hand side},
the next appearing below is the {\it right hand side}\footnote{Although a 
buffer is intended to implement the concept of a set (not a sequence) of trees
we take profit of the actual organization of the system to realize the
concept of (ordered) tree pair in the implementation.}. 
The positional approach allows us
to have naming freedom: the tree names
are irrelevant\footnote{so that even if we want to have mnemonic names 
resembling their distinct character - left or right hand side, - we have some
naming flexibility to call them e.g. {\it lhs23} or {\it lhs-passive}, ...}.
Since we can save buffers into text files, we can talk also about metarule
files.

The available options for applying a metarule which is in a buffer are:

\begin{itemize}
\item   For applying it to a single input tree, click in the name of the
        tree in the main window, and choose the option 
        {\it apply metarule to tree}. 
        You will be prompted for the name of the metarule
        to apply to the tree
        which should be, as we mentioned before, the name of the buffer that
        contains the metarule trees. The output trees will be generated
        at the end of the buffer that contains the input tree. The names of
        the trees depend of a LISP parameter {\it *metarules-change-name* }.
        If the value of the parameter is {\bf false} --- the {\it default} 
        value --- then the new trees will have the same name as the input, 
        otherwise, the name of the input tree followed by a dash (`-') and
        the name of the right hand side of the tree\footnote{the reason why 
        we do not use the name of the metarule, i.e. the name of the buffer,
        is because in some forms of application the metarules do not
        carry individual names, as we'll see soon is the case when a set of
        metarules from a file is applied.}.

        The value of the parameter can be changed by choosing {\it Tools} 
        at the menu bar and then either {\it name mr output trees =
        input}  or {\it append rhs name to mr output trees}.

\item   For applying it to all the trees of a buffer, click in the name of the 
        buffer that contains the trees and proceed as above. The output will
        be a new buffer with all the output trees. The name of the new buffer
        will be the same as the input buffer prefixed by "MR-". The names of
        the trees follow the conventions above.

\end{itemize}

The other options concern application to files (instead of buffers). 
Lets first define
the concepts of parallel, sequential and cumulative application of metarules. 
One metarule 
file can contain more than one metarule. The first two trees, i.e., the first
tree pair, form one metarule - lets call it $mr_0$. Subsequent pairs in the
sequence of trees define additional metarules --- 
$mr_1$, $mr_2$, ..., $mr_n$.

\begin{itemize}
\item We say that a metarule file is applied in parallel to a tree 
(see Figure~\ref{parallel})
if each of the 
metarules is applied independently to the input generating its particular 
output trees\footnote{remember a metarule application generates as many 
output trees as the number of matches}. We generalize the concept to the
application in parallel of a metarule file to a tree file (with possibly more 
than one tree), generating all the trees
as if each metarule in the metarule file was applied to each tree in the
input file.

\begin{figure}[htb]
\centerline{\psfig{figure=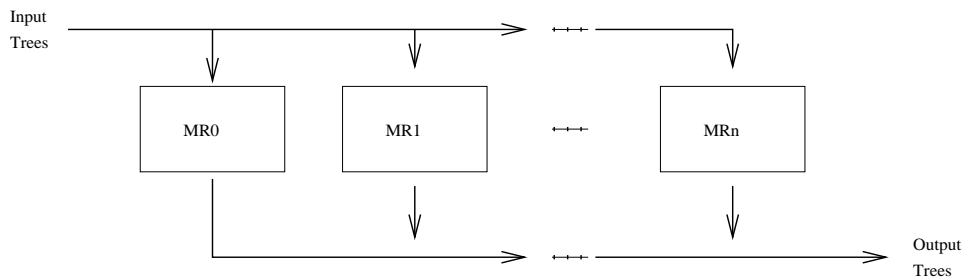,width=5in}}

\caption{Parallel application of metarules}
\label{parallel}
\end{figure}

\item We say that a metarule file $mr_0, mr_1, mr_2, ...,mr_n$ is applied in 
sequence to a input tree file 
(see Figure~\ref{sequential})
if we apply $mr_0$ to the trees of the input file, and
for each $0<i\leq n$ apply metarule $mr_i$ to the trees generated as a 
result of the application of $mr_{i-1}$.

\begin{figure}[htb]
\centerline{\psfig{figure=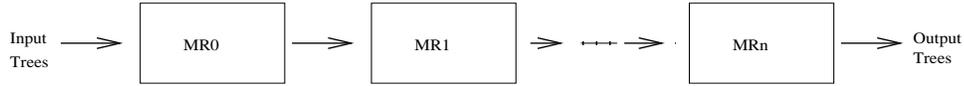,width=5in}}

\caption{Sequential application of metarules}
\label{sequential}
\end{figure}

\item Finally, the cumulative application is similar to the sequential, 
except that the input trees at each stage are by-passed to the output together
with the newly generated ones (see Figure~\ref{cumulative}).

\begin{figure}[htb]
\centerline{\psfig{figure=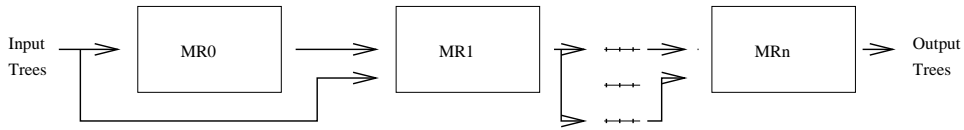,width=5in}}

\caption{Cumulative application of metarules}
\label{cumulative}
\end{figure}

\end{itemize}

Remember that in case of matching failure the output result is decided as 
explained in subsection ~\ref{output-gen} either to be empty or to be the 
input tree. The reflex here of having the parameter set for copying the input
is that for the parallel application the output will have as many copies of
the input as matching failures. For the sequential case the decision apply
at each level, and setting the parameter for copying, in a certain sense, 
guarantees for the 'pipe' not to break.
Due to its nature and unlike the two other modes, the cumulative 
application is not affected by this parameter.

The options for application of metarules to files are available by clicking
at the menu item {\it Tools} and then choosing the appropriate function
among:

\begin{itemize}
\item   {\it Apply metarule to files:}  
        You'll be prompted for the metarule file name
        which should contain one metarule\footnote{if it contains more than 
        2 trees, the additional trees are ignored}, and for input file names.
        Each input file name {\bf inpfile} will be independently submitted to 
        the application of the metarule generating an output file with the 
        name    {\bf MR-inpfile}.

\item   {\it Apply metarules in parallel to files:}
        You'll be prompted for the metarules file name with one or more 
        metarules and for input file names.
        Each input file name {\bf inpfile} will be independently submitted to 
        the application of the metarules in parallel. For each parallel 
        application to a file {\bf inpfile} an output file with the 
        name    {\bf MRP-inpfile} will be generated.

\item   {\it Apply metarules in sequence to files:}  
        The interaction is as described for the application in parallel, 
        except that
        the application of the metarules are in sequence and that 
        the output files are prefixed by {\bf MRS-} instead of {\bf MRP-}.

\item   {\it Apply metarules cumulatively to files:}  
        The interaction is as described for the applications in parallel
        and in sequence, except that the mode of application is cumulative
        and that the output files are prefixed by {\bf MRC-}.
\end{itemize}

Finally still under the {\it Tools} menu we can change the setting of the 
parameter that controls the output result on matching failure 
(see Subsection~\ref{output-gen})
by choosing
either {\it copy input on mr matching failure} or 
{\it no output on mr matching failure}.

\chapter{Lexical Organization}
\label{lexorg}

%\input{psfig} 

%\begin{document}

%\bibliographystyle{acl}

\section{Introduction}

% problem

An important characteristic of an
FB-LTAG is that it is lexicalized, i.e., each lexical item is anchored to a
tree structure that encodes subcategorization information.  Trees with the same
canonical subcategorizations are grouped into tree families.  The reuse of tree
substructures, such as wh-movement,
 in many different trees 
creates redundancy, which poses a problem for grammar development
and maintenance \cite{vijay-schabes92}.  To consistently implement a change in
some general aspect of the design of the grammar, all the relevant trees
currently must be inspected and edited.  Vijay Shanker and Schabes suggested
the use of hierarchical organization and of tree descriptions to specify
substructures that would be present in several elementary trees of a grammar.
Since then, in addition to ourselves, 
Becker, \cite{becker94}, Evans et al. \cite{Evans95}, and 
Candito\cite{Candito96} have
developed systems for organizing trees of a TAG which could be used for
developing and maintaining grammars.

 Our system is based
on the ideas expressed in Vijay-Shanker and Schabes, \cite{vijay-schabes92}, to
use partial-tree descriptions in specifying a grammar by separately defining
pieces of tree structures to encode independent syntactic principles. Various
individual specifications are then combined to form the elementary trees of the
grammar. The chapter begins with a description of 
our grammar development system, and its implementation. We will then
show the main results of using this tool to generate 
the Penn English grammar as well as a Chinese
TAG.  We describe the significant properties of both grammars, pointing out the
major differences between them, and the methods by which our system is informed
about these language-specific properties.  The chapter ends with the conclusion
and future work.

\section{System Overview}

In our approach,  
three types of components -- subcategorization frames, blocks and
lexical redistribution rules -- are used to describe lexical and
syntactic information.
%Lexical Redistribution Rule (LRR).
Actual trees are generated automatically from these abstract descriptions,
as shown in Figure \ref{system-overview}.
In maintaining the grammar only the abstract
descriptions need ever be manipulated; the tree descriptions and the
actual trees which they subsume are computed deterministically from these
high-level descriptions.

\begin{figure}[htb]
\centerline{\psfig{figure=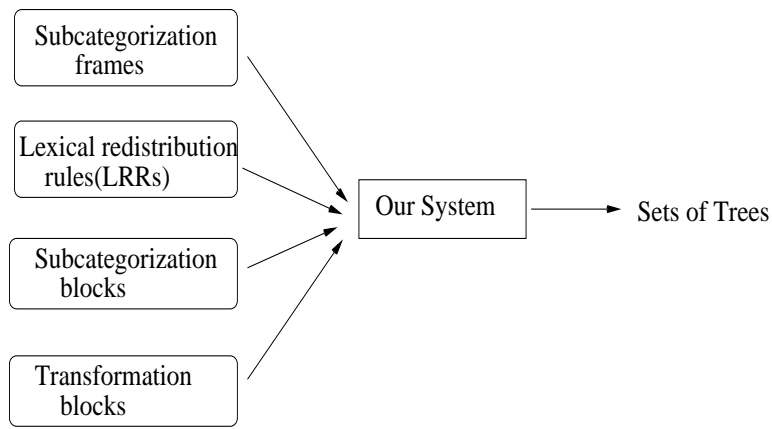,height=2.2in,width=4in,angle=0}}
\caption{Lexical Organization: System Overview}
\label{system-overview}
\end{figure}

\subsection{Subcategorization frames}
Subcategorization frames specify
the category of the main anchor, 
the number of  arguments,
each argument's category and position with respect to the anchor,
and other information
such as feature equations or node expansions.
Each tree family has one canonical subcategorization frame.

\subsection{Blocks}

Blocks are used to represent the tree substructures that are reused
in different trees, i.e. blocks 
subsume classes of trees. Each block includes
a set of nodes, dominance relation, parent relation, 
precedence relation between nodes,
and feature equations. This follows the definition of the tree 
descriptions specified in a logical language patterned after Rogers and
Vijay-Shanker\cite{rogers-vijay94}. 

Blocks are divided into two types according to their functions: 
subcategorization  blocks
and transformation blocks. 
The former describes structural configurations incorporating
the various information in a subcategorization frame. 
For example, some of the subcategorization blocks
used in the
development of the English grammar are shown in Figure 
\ref{subcat-blocks}.\footnote{
In order to focus on the use of tree descriptions and to 
make the figures less cumbersome, we show only
the structural aspects and do not show the feature value specification.
The parent, (immediate dominance), relationship is illustrated by
a plain line and the dominance relationship by a dotted line.
The arc between nodes shows the precedence order of
the nodes are unspecified. The nodes' categories
are enclosed in parentheses.}

When the subcategorization frame for a  verb is given
 by the grammar
developer, the system will automatically create a new block (of code) by
essentially selecting the appropriate primitive subcategorization blocks
corresponding to the argument information specified in that verb
frame. 

The transformation blocks are used for various transformations such as
wh-movement. These transformation blocks do not encode
rules for modifying trees, but rather describe
 the properties of a particular syntactic construction.
 Figure \ref{trans-blocks} depicts
our representation of phrasal extraction.
This can be specialized to give the
blocks for wh-movement, topicalization, relative clause formation, etc.  
For example, the
wh-movement block is defined by further specifying that the ExtractionRoot is
labeled S, the NewSite
has a +wh feature, and so on.

\begin{figure}[htb]
\centerline{\psfig{figure=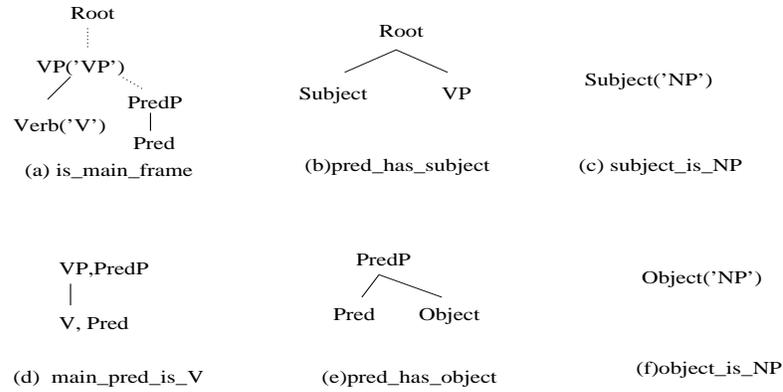,height=2in,width=4in}}
\caption{Some subcategorization blocks}
\label{subcat-blocks}
\end{figure}

\begin{figure}[htb]
\centerline{\psfig{figure=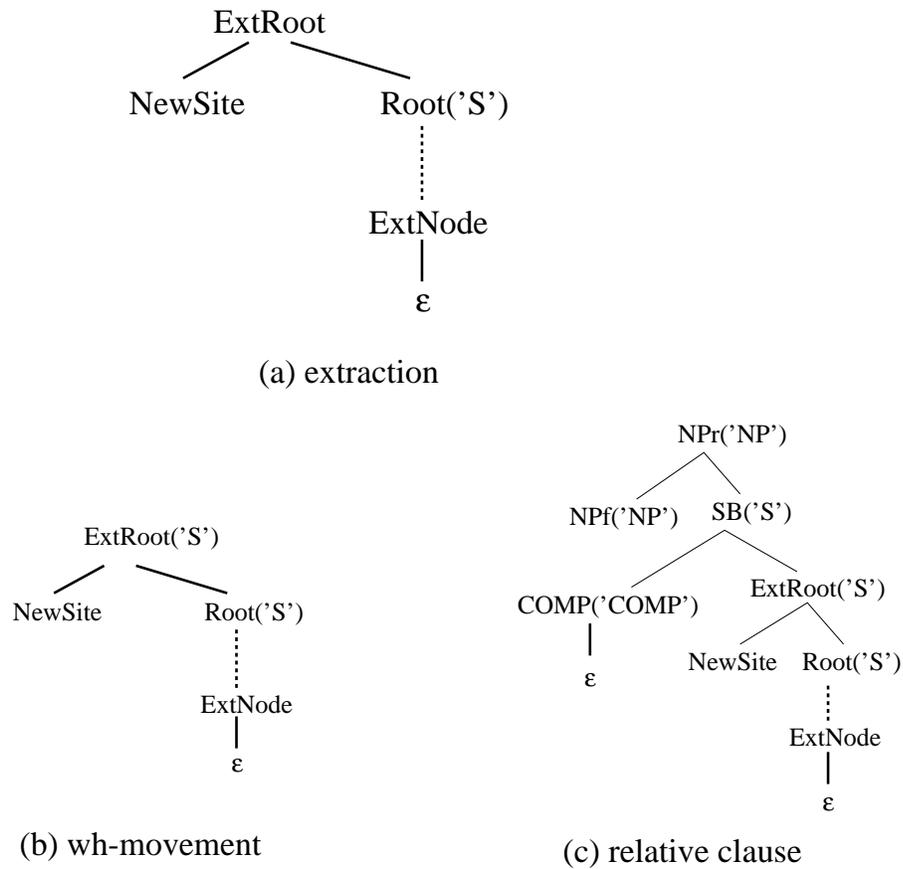,height=4.5in}}
\caption{Transformation blocks for extraction}
\label{trans-blocks}
\end{figure}

\subsection{Lexical Redistribution Rules (LRRs)}
The third type of machinery available for a grammar developer is the
Lexical Redistribution Rule (LRR). An LRR is a pair 
($r_{l}$, $r_{r}$) of subcategorization frames, 
which produces a new frame when applied to a subcategorization frame s,
by first {\it matching}\footnote{Matching occurs successfully when frame s
is compatible with $r_{l}$ in the type of anchors, the number of arguments,
their positions, categories and features. In other words, incompatible
features etc. will block certain LRRs from being applied.}
 the
left frame $r_{l}$ of r to s, then 
combining information in $r_{r}$ and s.
LRRs are introduced to incorporate
the connection between subcategorization frames. For example,
most transitive verbs have a frame for active(a subject and an object)
and another frame for passive, where the object in the former frame becomes
the subject in the latter. An LRR, denoted as  passive LRR,
is built to produce the passive subcategorization frame from
the active one.
Similarly, applying dative-shift LRR to the frame with
one NP subject and two NP objects will produce 
a frame with an NP subject and an PP object.

Besides the distinct content, LRRs and blocks also differ in several 
aspects:
\begin{itemize}
  \item They have different functionalities: Blocks represent
    the substructures that are reused in different trees. They
    are used to reduce the redundancy among trees; LRRs are introduced
    to incorporate the connections between the closely
     related subcategorization
    frames. 
  \item Blocks are strictly additive and can be added in any order.
       LRRs, on the other hand,
    produce different results depending on the order they are applied in,
    and
    are allowed to be non-additive, i.e., to remove information from 
    the subcategorization frame they are being applied to, as in
    the procedure of passive from active.
\end{itemize}

\begin{figure}[htb]
\centerline{\psfig{figure=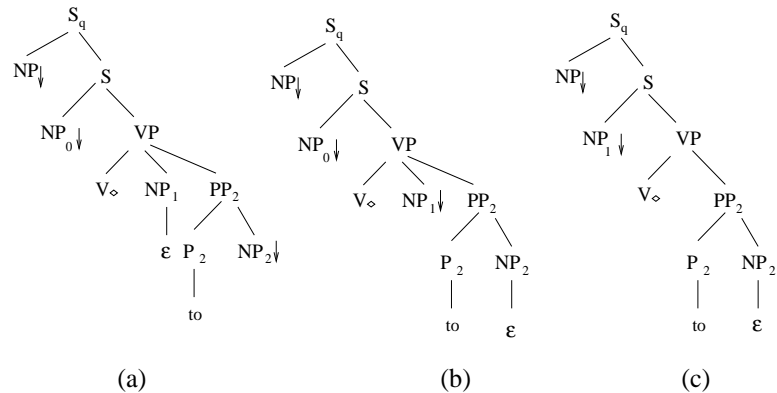,height=2in,width=4in,angle=0}}
\caption{Elementary trees generated from combining blocks}
\label{elem}
\end{figure}

\subsection{Tree generation}
To generate elementary trees,
 we begin with a canonical subcategorization frame. 
The system will first generate 
related subcategorization frames 
by applying LRRs, then select
subcategorization blocks corresponding to the information in
the subcategorization frames, next 
the combinations of these blocks
are further combined with the blocks corresponding to
 various transformations, finally,  a set of trees are generated
from those combined blocks, and they are 
the tree family for this subcategorization frame. 
Figure \ref{elem} shows some
of the trees produced in this way.  For instance, the last tree is obtained by
incorporating information from the ditransitive verb subcategorization frame,
applying the dative-shift and passive LRRs, and then combining them with the
wh-non-subject extraction block. 
Besides, in our system the hierarchy for subcategorization frames is implicit
as shown in Figure \ref{lattice}.

\begin{figure}[htb]
\centerline{\psfig{figure=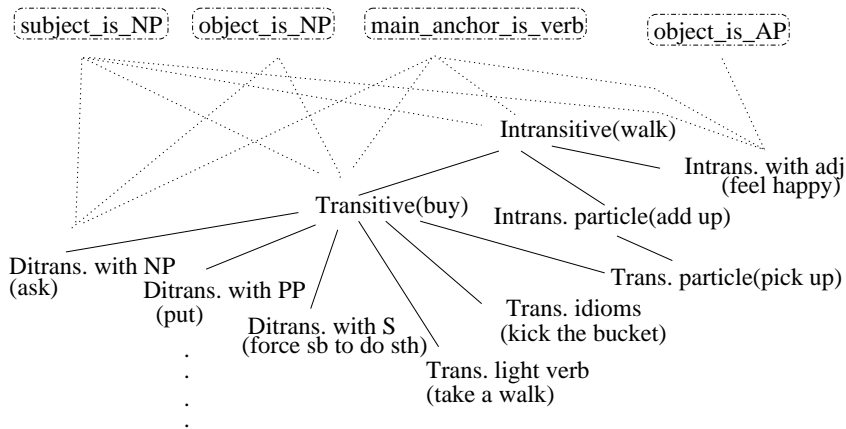,height=2.2in,angle=0}}
\caption{Partial inheritance lattice in English}
\label{lattice}
\end{figure}

\section{Implementation}

The input of our system is the description of the language, which includes 
the subcategorization frame list, LRR list, subcategorization block
list and transformation lists. The output  is a list of
trees generated automatically by the system, as shown in Figure
\ref{impl}. The tree generation module is written in Prolog, and 
the rest part is in C. We also have a graphic interface to input 
the language description. Figure \ref{interface-firstlevel} and
\ref{interface-block} are two snapshots of the interface.

\begin{figure}[htb]
\centerline{\psfig{figure=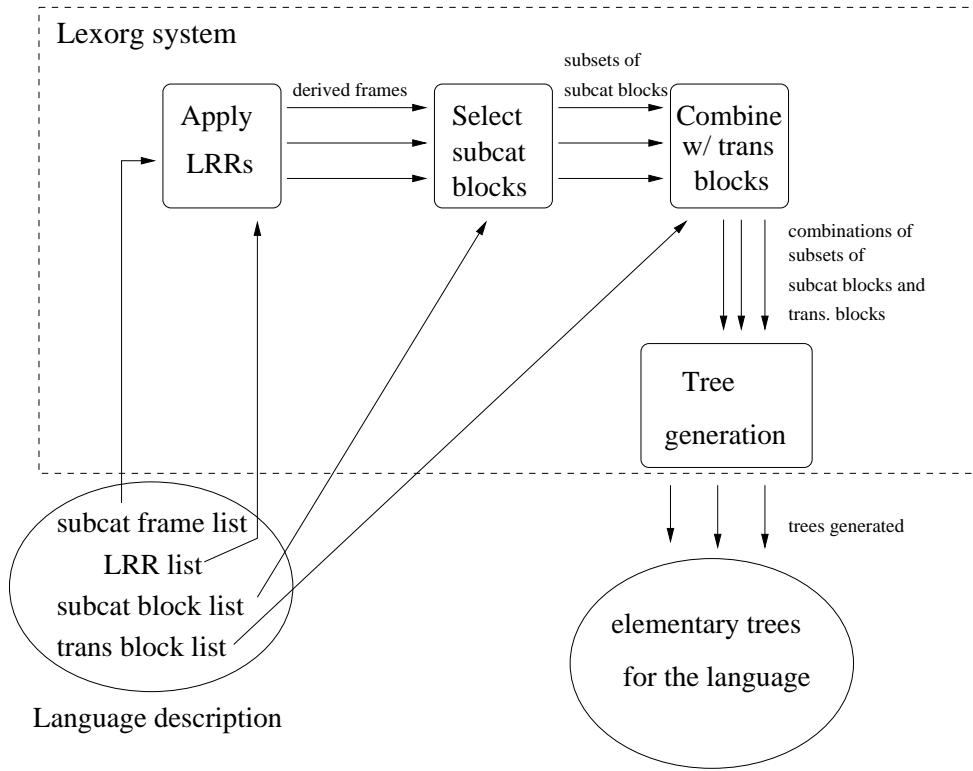,height=4in}}
\caption{Implementation of the system}
\label{impl}
\end{figure}

\begin{figure}[htb]
\centerline{\psfig{figure=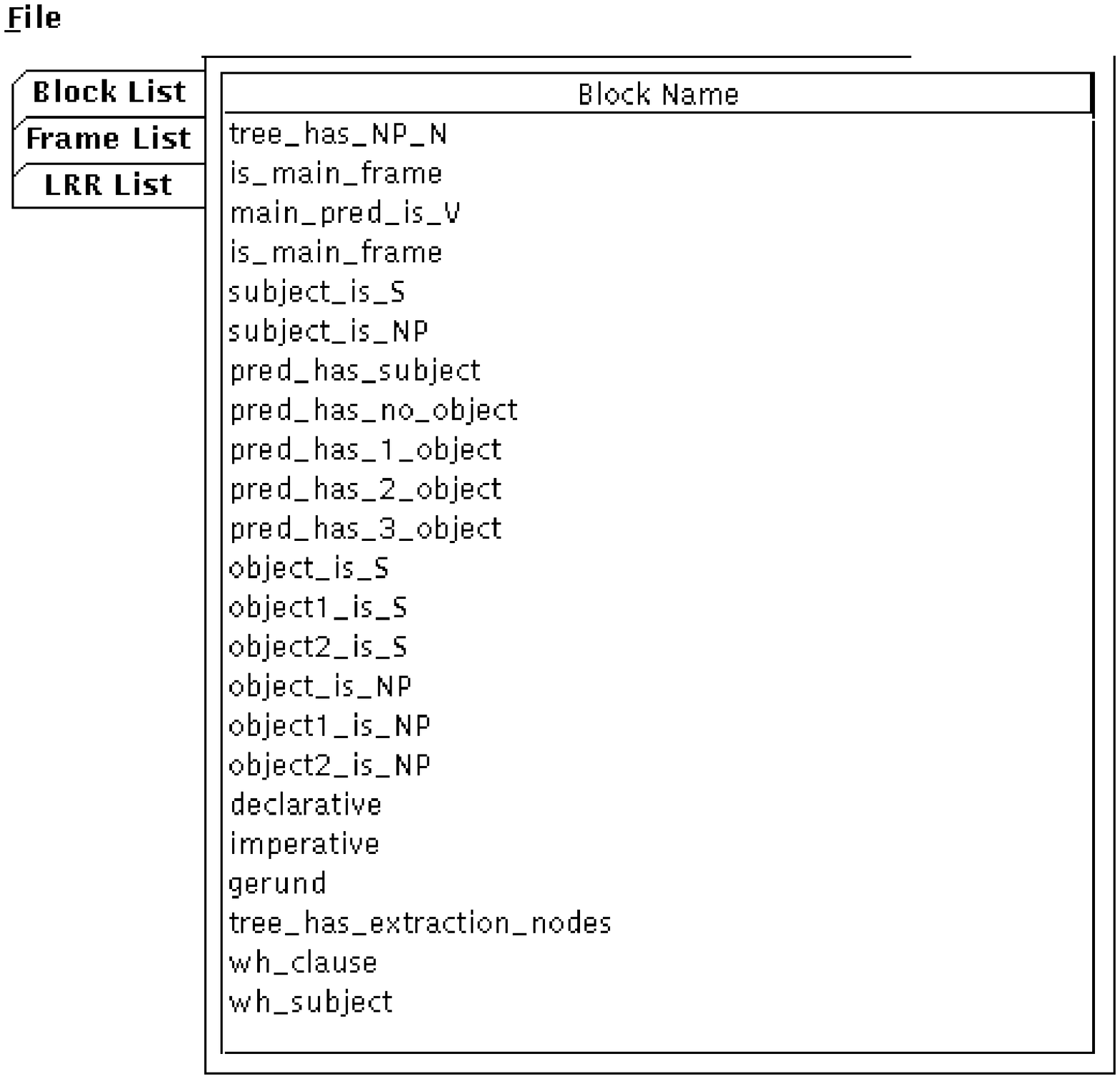,height=3in,angle=90}}
\caption{Interface for creating a grammar}
\label{interface-firstlevel}
\end{figure}

\begin{figure}[htb]
\centerline{\psfig{figure=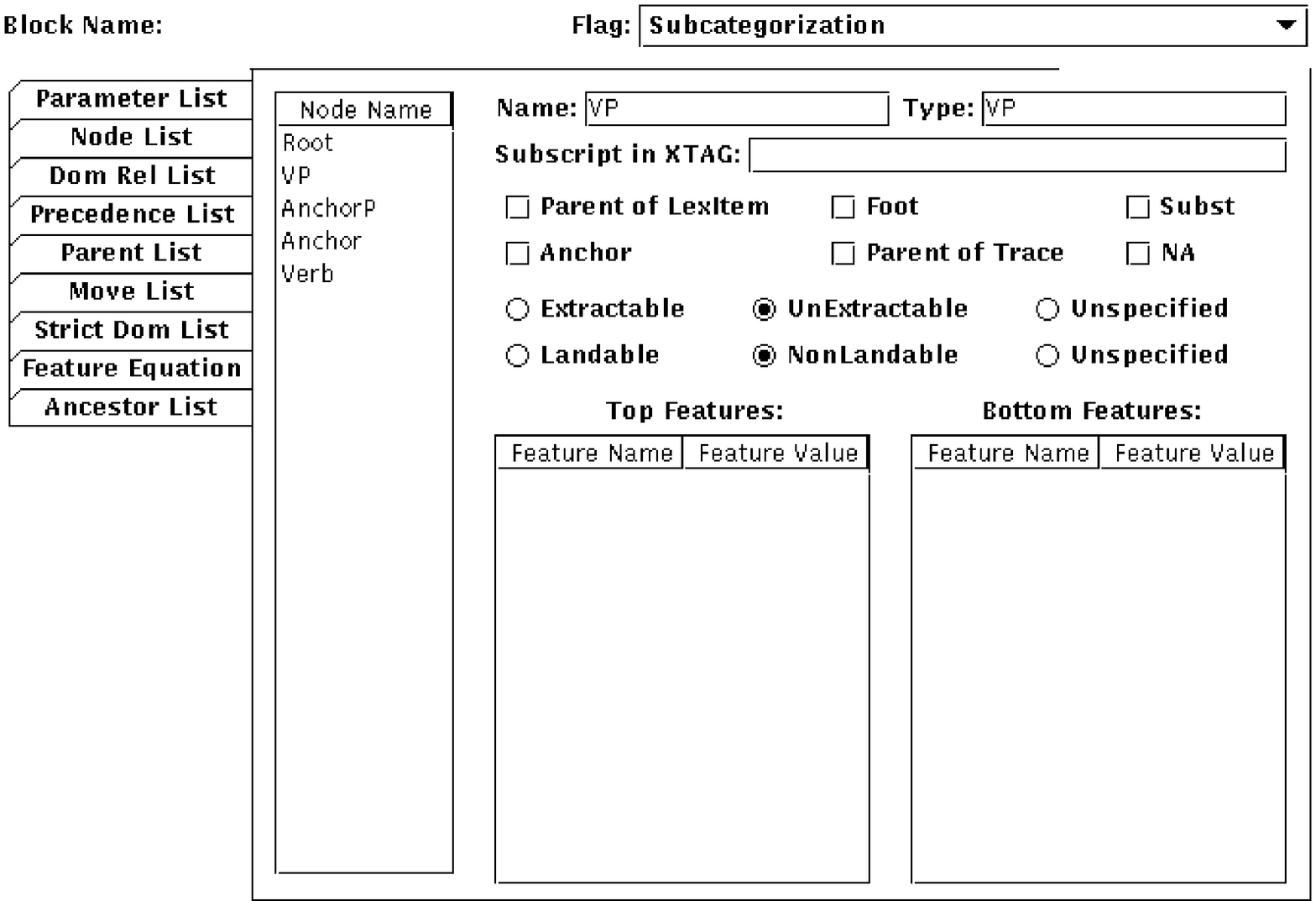,height=3in,angle=90}}
\vspace{0.4in}
\caption{Part of the Interface for creating blocks}
\label{interface-block}
\end{figure}

\section{Generating grammars}

We have used our tool to specify a grammar for English in order to produce
the trees used in the current English XTAG grammar.
We have also used our tool
to generate a large grammar for Chinese.
In designing these
grammars, we have tried to specify the grammars to reflect the similarities and
the differences between the languages. The major features of our specification
of these two grammars\footnote{Both grammars are still under development,
so the contents of these two tables might change a lot in the future
according to
the analyses we choose for certain phenomenon. For example,
the majority of work on Chinese grammar treat ba-construction as some kind
of object-fronting where the character 
{\it ba} is either an object marker or a preposition. According to this
analysis, an LRR rule for ba-construction is used in our grammar to generate
the preverbal-object frame from the postverbal frame. However, there has
been some argument for treating {\it ba} as a verb. If we later choose
that analysis, the main verbs in the patterns ``NP0 VP'' and ``NP0 ba NP1 VP''
will be different, therefore no LRR will be needed for it. As a result, the
numbers of LRRs, subcat frames and tree generated will change accordingly.}
 are summarized in Table \ref{table} and \ref{table2}.

\begin{table}[ht]
\centering
\begin{tabular}{|l|l|l|}     \hline
         & English & Chinese \\ \hline
examples     & passive       & bei-construction  \\
of LRRs         & dative-shift  & object fronting  \\ 
         & ergative  & ba-construction   \\ \hline
examples  & wh-question  & topicalization  \\
of transformation         & relativization & relativization \\
blocks         & declarative & argument-drop \\ \hline
\#  LRRs &   6     & 12  \\ \hline
\#  subcat blocks & 34 & 24 \\ \hline
\#  trans  blocks & 8  & 15 \\ \hline
\#  subcat frames  & 43 & 23 \\ \hline
\#  trees generated   & 638 & 280 \\ \hline
\end{tabular}  \\
\caption{Major features of English and Chinese grammars}
\label{table}
\end{table}

\begin{table}[ht]
\centering
\begin{tabular}{|l|l|l|l|}     \hline
         & both grammars & English & Chinese \\ \hline
         & causative & long passive &  VO-inversion \\
LRRs     & short passive & ergative  & ba-const \\
         &               & dative-shift  &       \\  \hline
         & topicalization  &  &   \\
trans blocks    & relativization & gerund & argument-drop \\
        & declarative &  &  \\ \hline
        & NP/S subject &  &  zero-subject \\
 subcat blocks & S/NP/PP object & PL object & preverbal object \\
               & V predicate &  prep predicate & \\ \hline
\end{tabular}  \\
\caption{Comparison of the two grammars}
\label{table2}
\end{table}

By focusing on the specification of individual grammatical information, we
have been able to generate nearly all of the trees from the tree families 
used in the current English grammar developed at Penn\footnote{We have not yet
attempted to extend our coverage to include punctuation, it-clefts, and a few
idiosyncratic analyses.}. 
 Our approach, has also exposed certain gaps in the Penn grammar.
%in \ref{envgram97}.  
We are encouraged with the utility of our tool and the ease with which this
large-scale grammar was developed.

We are currently working on expanding the contents of
subcategorization frame to include 
trees for other categories of words. For example, 
a frame which has no specifier and one NP complement
and whose predicate is a preposition will correspond to
PP $\rightarrow$ P NP tree. We'll also introduce a modifier field
and semantic features, so that the head features will propagate
from modifiee to modified node, while non-head features from
the predicate as the head of the modifier will be passed to
the modified node.

% insert figures here

\section{Summary}

We have described a tool for grammar development in which tree descriptions are
used to provide an abstract specification of the linguistic phenomena relevant
to a particular language.  In grammar development and maintenance, only the
abstract specifications need to be edited, and any changes or corrections will
automatically be proliferated throughout the grammar.  In addition to
lightening the more tedious aspects of grammar maintenance, this approach also
allows a unique perspective on the general characteristics of a language.
Defining hierarchical blocks for the grammar both necessitates and facilitates
an examination of the linguistic assumptions that have been made with regard to
feature specification and tree-family definition. This can be very useful for
gaining an overview of the theory that is being implemented and exposing gaps
that remain unmotivated and need to be investigated.  The type of gaps that can
be exposed could include a missing subcategorization frame that might arise
from the automatic combination of blocks and which would correspond to an
entire tree family, a missing tree which would represent a particular type of
transformation for a subcategorization frame, or inconsistent feature
equations.  By focusing on syntactic properties at a higher level, our
approach allows new opportunities for the investigation of how languages relate
to themselves and to each other.

%\bibliography{/mnt/linc/home/fxia/paper/mpaper/ACL98/ACL98}

%\end{document}

\chapter{Tree Naming conventions}
\label{tree-naming}

The various trees within the XTAG grammar are named more or less according to
the following tree naming conventions.  Although these naming conventions are
generally followed, there are occasional trees that do not strictly follow
these conventions.

\section{Tree Families}
Tree families are named according to the basic declarative tree structure in
the tree family (see section~\ref{family-trees}), but with a T as the first
character instead of an $\alpha$ or $\beta$.

\section{Trees within tree families}
\label{family-trees}

Each tree begins with either an $\alpha$ (alpha) or a $\beta$ (beta) symbol,
indicating whether it is an initial or auxiliary tree, respectively.  Following
an $\alpha$ or a $\beta$ the name may additionally contain one of:

\begin{description}
\item\begin{tabular}{ll}
I&imperative\\
E&ergative\\
N{0,1,2}&relative clause\{position\}\\
G&NP gerund\\
D&Determiner gerund\\
pW{0,1,2}&wh-PP extraction\{position\}\\
W{0,1,2}&wh-NP extraction\{position\}\\
X&ECM (eXceptional case marking)\\
\end{tabular}
\end{description}

\noindent Numbers are assigned according to the position of the argument in the
declarative tree, as follows:

\begin{description}
\item\begin{tabular}{ll}
0&subject position\\
1&first argument (e.g. direct object)\\
2&second argument (e.g. indirect object)\\
\end{tabular}
\end{description}

\noindent The body of the name consists of a string of the following 
components, which corresponds to the leaves of the tree.  The anchor(s) of the
trees is(are) indicated by capitalizing the part of speech corresponding to the
anchor.

\begin{description}
\item\begin{tabular}{ll}
s&sentence\\
a&adjective\\
arb&adverb\\
be&{\it be}\\
c&relative complementizer\\
x&phrasal category\\
d&determiner\\
v&verb\\
lv&light verb\\
conj&conjunction\\
comp&complementizer\\
it&{\it it}\\
n&noun\\
p&preposition\\
to&{\it to}\\
pl&particle\\
by&{\it by}\\
neg&negation\\
\end{tabular}
\end{description}

\noindent As an example, the transitive declarative tree consists of a subject
NP, followed by a verb (which is the anchor), followed by the object NP.  This
translates into $\alpha$nx0Vnx1.  If the subject NP had been extracted, then
the tree would be $\alpha$W0nx0Vnx1.  A passive tree with the {\it by} phrase
in the same tree family would be $\alpha$nx1Vbynx0.  Note that even though the
object NP has moved to the subject position, it retains the object encoding
(nx1). 

\section{Assorted Initial Trees}

Trees that are not part of the tree families are generally gathered into
several files for convenience.  The various initial trees are located in {\tt
lex.trees}.  All the trees in this file should begin with an $\alpha$,
indicating that they are initial trees.  This is followed by the root category
which follows the naming conventions in the previous section (e.g. n for noun,
x for phrasal category).  The root category is in all capital letters.  After
the root category, the node leaves are named, beginning from the left, with the
anchor of the tree also being capitalized.  As an example, the $\alpha$NXN
tree is rooted by an NP node (NX) and anchored by a noun (N).

\section{Assorted Auxiliary Trees}

The auxiliary trees are mostly located in the buffers {\tt
prepositions.trees}, {\tt conjunctions.trees}, {\tt
determiners.trees}, {\tt advs-adjs.trees}, and {\tt modifiers.trees},
although a couple of other files also contain auxiliary trees.  The
auxiliary trees follow a slightly different naming convention from the
initial trees.  Since the root and foot nodes must be the same for the
auxiliary trees, the root nodes are not explicitly mentioned in the
names of auxiliary trees.  The trees are named according to the leaf
nodes, starting from the left, and capitalizing the anchor node.  All
auxiliary trees begin with a $\beta$, of course.  For example,
$\beta$ARBs, indicates a tree anchored by an adverb (ARB), that
adjoins onto the left of an S node (Note that S must be the foot node,
and therefore also the root node).

\subsection{Relative Clause Trees}
For relative clause trees, the following naming conventions have been
adopted: if the {\em wh}-moved NP is overt, it is not explicitly
represented. Instead the index of the site of movement
(0 for subject, 1 for object, 2 for indirect object) is appended to the
N. So $\beta$N0nx0Vnx1 is a subject
extraction relative clause with {\bf NP$_{w}$} substitution
and $\beta$N1nx0Vnx1 is an object extraction
relative clause. If the {\em wh}-moved NP is covert and Comp substitutes
in, the Comp node is represented by {\em c} in the tree name and the
index of the extraction site follows {\em c}. Thus
$\beta$Nc0nx0Vnx1 is a subject extraction
relative clause with Comp substitution. Adjunct trees are similar, except
that since the extracted material is not co-indexed to a trace, no index
is specified (cf. $\beta$Npxnx0Vnx1, which is an adjunct relative clause with
PP pied-piping, and $\beta$Ncnx0Vnx1, which is an adjunct relative clause
with Comp substitution). Cases of pied-piping, in which the pied-piped
material is part of the anchor have the anchor capitalized or spelled-out
(cf. $\beta$Nbynx0nx1Vbynx0 which is a relative clause with {\em by}-phrase
pied-piping and {\bf NP$_{w}$} substitution.).

\chapter{Features}
\label{features}

Table~\ref{feature-table} contains a comprehensive list of the features in the
XTAG grammar and their possible values.

This section consists of short `biographical' sketches of the various features
currently in use in the XTAG English grammar.

\footnotesize
\begin{table}[htbp]
\centering
\begin{tabular}{|l|l|}
\hline
Feature&Value\\
\hline
\hline
$<$agr 3rdsing$>$&$+,-$\\
$<$agr num$>$&plur,sing\\
$<$agr pers$>$&1,2,3\\
$<$agr gen$>$&fem,masc,neuter\\
$<$assign-case$>$&nom,acc,none\\
$<$assign-comp$>$&that,whether,if,for,ecm,rel,inf\_nil,ind\_nil,ppart\_nil,none\\
$<$card$>$&$+,-$\\
$<$case$>$&nom,acc,gen,none\\
$<$comp$>$&that,whether,if,for,rel,inf\_nil,ind\_nil,nil\\
$<$compar$>$&$+,-$\\
$<$compl$>$&$+,-$\\
$<$conditional$>$&$+,-$\\
$<$conj$>$&and,or,but,comma,scolon,to,disc,nil\\
$<$const$>$&$+,-$\\
$<$contr$>$&$+,-$\\
$<$control$>$&no value, indexing only\\
$<$decreas$>$&$+,-$\\
$<$definite$>$&$+,-$\\
$<$displ-const$>$&$+,-$\\
$<$equiv$>$&$+,-$\\
$<$extracted$>$&$+,-$\\
$<$gen$>$&$+,-$\\
$<$gerund$>$&$+,-$\\
$<$inv$>$&$+,-$\\
$<$invlink$>$&no value, indexing only\\
$<$irrealis$>$&$+,-$\\
$<$mainv$>$&$+,-$\\
$<$mode$>$&base,ger,ind,inf,imp,nom,ppart,prep,sbjunt\\
$<$neg$>$&$+,-$\\
$<$passive$>$&$+,-$\\
$<$perfect$>$&$+,-$\\
$<$pred$>$&$+,-$\\
$<$progressive$>$&$+,-$\\
$<$pron$>$&$+,-$\\
$<$punct bal$>$&dquote,squote,paren,nil\\
$<$punct contains colon$>$&$+,-$\\
$<$punct contains dash$>$&$+,-$\\
$<$punct contains dquote$>$&$+,-$\\
$<$punct contains scolon$>$&$+,-$\\
$<$punct contains squote$>$&$+,-$\\
$<$punct struct$>$&comma,dash,colon,scolon,nil\\
$<$punct term$>$&per,qmark,excl,nil\\
$<$quan$>$&$+,-$\\
$<$refl$>$&$+,-$\\
$<$rel-clause$>$&$+,-$\\
$<$rel-pron$>$&ppart,ger,adj-clause\\
$<$select-mode$>$&ind,inf,ppart,ger\\
$<$super$>$&$+,-$\\
$<$tense$>$&pres,past\\
$<$trace$>$&no value, indexing only\\
$<$trans$>$&$+,-$\\
$<$weak$>$&$+,-$\\
$<$wh$>$&$+,-$\\
\hline
\end{tabular}
\caption{List of features and their possible values}
\label{feature-table}
\end{table}

\normalsize

\section{Agreement}
{\bf $\langle$agr$\rangle$} is a complex feature.
It can have as its subfeatures:\\
{\bf $\langle$agr 3rdsing$\rangle$}, possible values: {\bf $+/-$ }\\
{\bf $\langle$agr num$\rangle$}, possible values: {\bf $plur,sing$ }\\
{\bf $\langle$agr pers$\rangle$}, possible values: {\bf $1,2,3$ }\\
{\bf $\langle$agr gen$\rangle$}, possible values: {\bf $masc,fem,neut$ }

These features are used to ensure agreement between a verb and its subject.

Where does it occur:\\ Nouns comes specified from the lexicon with
their {\bf $\langle$agr$\rangle$} features. e.g. {\em books} is {\bf
$\langle$agr 3rdsing$\rangle$:~--}, {\bf $\langle$agr num$\rangle$:~plur}, and {\bf $\langle$agr pers$\rangle$:~3}. Only pronouns use the
{\bf $<$gen$>$} (gender) feature.

The {\bf $\langle$agr$\rangle$} features of a noun are transmitted up the 
NP tree by the following equation:\\
{\bf NP.b:$\langle$agr$\rangle =$ N.t:$\langle$agr$\rangle$}

Agreement between a verb and its subject is mediated by the following feature
equations:

\enumsentence{ {\bf NP$_{subj}$:$\langle$agr$\rangle =$ VP.t:$\langle$agr$\rangle$}}

\enumsentence{ {\bf VP.b:$\langle$agr$\rangle =$ V.t:$\langle$agr$\rangle$}}

Agreement has to be done as a two step process because whether the
verb agrees with the subject or not depends upon whether some auxiliary verb
adjoins in and upon what the {\bf $\langle$agr$\rangle$} specification of 
the verb is. 

Verbs also come specified from the lexicon with their {\bf
$\langle$agr$\rangle$} features, e.g. the {\bf $\langle$agr$\rangle$}
features of the verb {\em sings} are {\bf $\langle$agr
3rdsing$\rangle$:~+}, {\bf $\langle$agr num$\rangle$:~sing}, and {\bf
$\langle$agr pers$\rangle$:~3}; Non-finite forms of the verb {\em
sing} e.g. {\em singing} do not come with an {\bf
$\langle$agr$\rangle$} feature specification.

\subsection{Agreement and Movement}
The {\bf $\langle$agr$\rangle$} features of a moved NP and its trace 
are co-indexed. This captures the fact that movement does not disrupt 
a pre-existing agreement relationship between an NP and a verb.

\enumsentence{ \ [Which boys]$_{i}$ does John think [t$_{i}$ are/*is intelligent]?}

\section{Case}

There are two features responsible for case-assignment:\\
{\bf $\langle$case$\rangle$}, possible values: {\bf nom, acc, gen, none}\\
{\bf $\langle$assign-case$\rangle$}, possible values: {\bf nom, acc, none}

Case assigners (prepositions and verbs) as well as the VP, S and PP
nodes that dominate them have an {\bf $\langle$assign-case$\rangle$}
case feature. Phrases and lexical items that have case i.e. Ns and NPs
have a {\bf $\langle$case$\rangle$} feature.

Case assignment by prepositions involves the following equations:

\enumsentence{ {\bf PP.b:$\langle$assign-case$\rangle =$ P.t:$\langle$case$\rangle$}}

\enumsentence{ {\bf NP.t:$\langle$case$\rangle =$ P.t:$\langle$case$\rangle$}}

Prepositions come specified from the lexicon with their {\bf $\langle$assign-case$\rangle$}
feature.

\enumsentence{ {\bf P.b:$\langle$assign-case$\rangle =$ acc}}

Case assignment by verbs has two parts: assignment of case to the
object(s) and assignment of case to the subject. Assignment of case to
the object is simpler.  English verbs always assign accusative case to
their NP objects (direct or indirect).  Hence this is built into the
tree and not put into the lexical entry of each individual verb.

\enumsentence{ {\bf NP$_{object}$.t:$\langle$case$\rangle =$ acc}}

Assignment of case to the subject involves the following two equations.

\enumsentence{ {\bf NP$_{subj}$:$\langle$case$\rangle =$ VP.t:$\langle$assign-case$\rangle$}}

\enumsentence{ {\bf VP.b:$\langle$assign-case$\rangle =$ V.t:$\langle$assign-case$\rangle$}}

This is a two step process -- the final case assigned to the subject
depends upon the {\bf $\langle$assign-case$\rangle$} feature of the
verb as well as whether an auxiliary verb adjoins in.

Finite verbs like {\em sings} have {\bf nom} as the value of their
{\bf $\langle$assign-case$\rangle$} feature. Non-finite verbs have
{\bf none} as the value of their {\bf $\langle$assign-case$\rangle$}
feature. So if no auxiliary adjoins in, the only subject they can have
is {\bf PRO} which is the only NP with {\bf none} as the value its
{\bf $\langle$case$\rangle$} feature.

\subsection{ECM}
Certain verbs e.g. {\em want, believe, consider} etc. and one complementizer
{\em for} are able to assign case to the subject of their complement clause. 

The complementizer {\em for}, like the preposition {\em for}, has the
{\bf $\langle$assign-case$\rangle$} feature of its complement set to
{\bf acc}. Since the {\bf $\langle$assign-case$\rangle$} feature of
the root S$_{r}$ of the complement tree and the {\bf
$\langle$case$\rangle$} feature of its NP subject are co-indexed, this
leads to the subject being assigned accusative case.

ECM verbs have the {\bf $\langle$assign-case$\rangle$}  feature of their
foot S node set to {\bf acc}. The co-indexation between the 
{\bf $\langle$assign-case$\rangle$} feature of
the root S$_{r}$ and the {\bf $\langle$case$\rangle$} feature of the NP subject
leads to the subject being assigned accusative case.

\subsection{Agreement and Case}
The {\bf $\langle$case$\rangle$} features of a moved NP and its trace 
are co-indexed. This captures the fact that movement does not disrupt 
a pre-existing relationship of case-assignment between a verb and an NP.

\enumsentence{ Her$_{i}$/*She$_{i}$, I think that Odo like t$_{i}$.}

\section{Extraction and Inversion}
{\bf $\langle$extracted$\rangle$}, possible vales are {\bf $+/-$}

All sentential trees with extracted components, with the exception of
relative clauses are marked {\bf S.b$\langle$extracted$\rangle = +$}
at their top S node. The extracted element may be a {\em wh}-NP or a
topicalized NP. The {\bf $\langle$extracted$\rangle$} feature 
is currently used to block embedded topicalizations as exemplified
by the following example.
\enumsentence{ * John wants [Bill$_{i}$ [PRO to leave t$_{i}$]] }

\noindent
{\bf $\langle$trace$\rangle$}: this feature is not assigned any value and
is used to co-index moved NPs and their traces which are marked by
$\epsilon$.

\noindent
{\bf $\langle$wh$\rangle$}: possible values are {\bf $+/-$}\\ NPs like
{\em who}, {\em what} etc. come marked from the lexicon with a value
of {\bf $+$} for the feature {\bf $\langle$wh$\rangle$}.  Non {\em
wh}-NPs have {\bf $-$} as the value of their {\bf
$\langle$wh$\rangle$} feature. Note that {\bf $\langle$wh$\rangle$ = +
} NPs are not restricted to occurring in extracted positions, to allow
for the correct treatment of echo questions.

The {\bf $\langle$wh$\rangle$} feature is propagated up by possessives
-- e.g. the $+$ {\bf $\langle$wh$\rangle$} feature of the determiner
{\em which} in {\em which boy} is propagated up to the level of the NP
so that the value of the {\bf $\langle$wh$\rangle$} feature of the
entire NP is $+${\bf $\langle$wh$\rangle$}. This process is recursive
e.g. {\em which boy's mother}, {\em which boy's mother's sister}.

The {\bf $\langle$wh$\rangle$} feature
is also propagated up PPs. Thus the PP {\em to whom} has $+$ as the value of its 
{\bf $\langle$wh$\rangle$} feature. 

In trees with extracted NPs, the {\bf $\langle$wh$\rangle$} feature of the
root node S node is equated with the {\bf $\langle$wh$\rangle$} feature
of the extracted NPs. 

The {\bf $\langle$wh$\rangle$} feature is used to impose
subcategorizational constraints.
Certain verbs like {\em wonder} can
only take interrogative complements, other verbs such as {\em know}
can take both interrogative and non-interrogative complements, and yet
other verbs like {\em think} can only take non-interrogative
complements (cf. the {\bf $\langle$extracted$\rangle$} and {\bf
$\langle$mode$\rangle$} features also play a role in imposing 
subcategorizational constraints).

The {\bf $\langle$wh$\rangle$} feature is also used to get the correct
inversion patterns.

\subsection{Inversion, Part 1}
The following three features are used to ensure the correct pattern of
inversion:\\
{\bf $\langle$wh$\rangle$}: possible values are {\bf $+/-$}\\
{\bf $\langle$inv$\rangle$}: possible values are {\bf $+/-$}\\
{\bf $\langle$invlink$\rangle$}: possible values are {\bf $+/-$}

Facts to be captured:\\
1. No inversion with topicalization\\
2. No inversion with matrix extracted subject {\em wh}-questions\\
3. Inversion with matrix extracted object {\em wh}-questions\\
4. Inversion with all matrix {\em wh}-questions involving extraction from an
embedded clause\\
5. No inversion in embedded questions \\
6. No matrix subject topicalizations.

Consider a tree with object extraction, where NP is extracted. 
The following feature equations are used:\\

\enumsentence{ {\bf S$_{q}$.b:$\langle$wh$\rangle =$ NP.t:$\langle$wh$\rangle$}\label{inv1}}
\enumsentence{ {\bf S$_{q}$.b:$\langle$invlink$\rangle =$  S$_{q}$.b:$\langle$inv$\rangle$}\label{inv2}}
\enumsentence{ {\bf S$_{q}$.b:$\langle$inv$\rangle =$  S$_{r}$.t:$\langle$inv$\rangle$}\label{inv3}}
\enumsentence{ {\bf S$_{r}$.b:$\langle$inv$\rangle = -$}\label{inv4}}

\noindent
{\bf Root restriction}: A restriction is imposed on the final root
node of any XTAG derivation of a tensed sentence which equates the
{\bf $\langle$wh$\rangle$} feature and the {\bf
$\langle$invlink$\rangle$} feature of the final root node.

If the extracted NP is not a {\em wh}-word i.e. its {\bf
$\langle$wh$\rangle$} feature has the value $-$, at the end of the
derivation, {\bf S$_{q}$.b:$\langle$wh$\rangle$} will also have the
value $-$. Because of the root constraint {\bf
S$_{q}$.b:$\langle$wh$\rangle$} will be equated to {\bf
S$_{q}$.b:$\langle$invlink$\rangle$} which will also come to have the
value $-$. Then, by (\ref{inv3}), {\bf
S$_{r}$.t:$\langle$inv$\rangle$} will acquire the value $-$. This will
unify with {\bf S$_{r}$.b:$\langle$inv$\rangle$} which has the value
$-$ (cf. \ref{inv4}). Consequently, no auxiliary verb adjunction will
be forced. Hence, there will never be inversion in topicalization.

If the extracted NP is a {\em wh}-word i.e. its {\bf $\langle$wh$\rangle$} 
feature has the value $+$, at the end of the derivation, 
{\bf S$_{q}$.b:$\langle$wh$\rangle$} will also have the value $+$. Because of
the root constraint {\bf S$_{q}$.b:$\langle$wh$\rangle$} will be equated 
to {\bf S$_{q}$.b:$\langle$invlink$\rangle$} which will also come to have
the value $+$. Then, by (\ref{inv3}), {\bf S$_{r}$.t:$\langle$inv$\rangle$} 
will acquire the value $+$. This will not unify with {\bf S$_{r}$.b:$\langle$inv$\rangle$}
which has the value $+$ (cf. \ref{inv4}). Consequently, the adjunction
of an inverted auxiliary verb is required for the derivation to succeed.

Inversion will still take place even if the extraction is from an embedded
clause.

\enumsentence{ Who$_{i}$ does Loida think [Miguel likes t$_{i}$]}

This is because the adjoined tree's root node will also have its 
{\bf S$_{r}$.b:$\langle$inv$\rangle$} set to $-$.

Note that inversion is only forced upon us because S$_{q}$ is the
final root node and the {\bf Root restriction} applies. In embedded
environments, the root restriction would not apply and the feature
clash that forces adjunction would not take place.

The {\bf $\langle$invlink$\rangle$} feature is not present in subject
extractions.  Consequently there is no inversion in subject questions.

Subject topicalizations are blocked by setting the 
{\bf $\langle$wh$\rangle$} feature of the extracted NP to $+$ i.e. only
{\em wh}-phrases can go in this location. 

\subsection{Inversion, Part 2}

{\bf $\langle$displ-const$\rangle$}:\\ Possible values: {\bf [set1:
+], [set1: --]}\\ In the previous section, we saw how inversion is
triggered using the {\bf $\langle$invlink$\rangle$}, {\bf
$\langle$inv$\rangle$}, {\bf $\langle$wh$\rangle$} features. Inversion
involves movement of the verb from V to C. This movement process is
represented using the {\bf $\langle$displ-const$\rangle$} feature
which is used to simulate Multi-Component TAGs.\footnote{The {\bf
$\langle$displ-const$\rangle$} feature is also used in the ECM
analysis.} The sub-value {\bf set1} indicates the inversion
multi-component set; while there are not currently any other uses of
this mechanism, it could be expanded with other sets receiving
different {\bf set} values.

The {\bf $\langle$displ-const$\rangle$} feature is used to ensure
adjunction of two trees, which in this case are the auxiliary
tree corresponding to the moved verb (S adjunct) and the auxiliary tree
corresponding to the trace of the moved verb (VP adjunct). The following
equations are used:

\enumsentence{  {\bf S$_{r}$.b:$\langle$displ-const set1$\rangle = -$}\label{dis1}}
\enumsentence{  {\bf S.t:$\langle$displ-const set1$\rangle = +$}\label{dis2}}
\enumsentence{  {\bf VP.b:$\langle$displ-const set1$\rangle =$
          V.t:$\langle$displ-const set1$\rangle$}\label{dis3}}
\enumsentence{  {\bf V.b:$\langle$displ-const set1$\rangle = +$}\label{dis4}}
\enumsentence{  {\bf S$_{r}$.b:$\langle$displ-const set1$\rangle =$ 
          VP.t:$\langle$displ-const set1$\rangle$}\label{dis5}}

\section{Clause Type}
There are several features that mark clause type.\footnote{We have
already seen one instance of a feature that marks clause-type: {\bf
$\langle$extracted$\rangle$}, which marks whether a certain S involves
extraction or not.} They are:\\ {\bf $\langle$mode$\rangle$}\\ 
{\bf $\langle$passive$\rangle$}: possible values are {\bf +/--}

\noindent
{\bf $\langle$mode$\rangle$}: possible values are 
{\bf base, ger, ind, inf, imp, nom, ppart, prep, sbjnct}\\
The {\bf $\langle$mode$\rangle$} feature of a verb in its root form is
{\bf base}. The {\bf $\langle$mode$\rangle$} feature of a verb in its past 
participial form is {\bf ppart}, the {\bf $\langle$mode$\rangle$} feature of a 
verb in its progressive/gerundive form is {\bf ger}, 
the {\bf $\langle$mode$\rangle$} feature of a tensed verb is {\bf ind},
and the {\bf $\langle$mode$\rangle$} feature of a verb in the imperative 
is {\bf imp}. 

\noindent
{\bf nom} is the {\bf $\langle$mode$\rangle$} value of AP/NP
predicative trees headed by a null copula.  {\bf prep} is the {\bf
$\langle$mode$\rangle$} value of PP predicative trees headed by a null
copula.  Only the copula auxiliary tree, some sentential complement
verbs (such as {\it consider} and raising verb auxiliary trees have
{\bf nom/prep} as the {\bf $\langle$mode$\rangle$} feature
specification of their foot node. This allow them, and only them, to
adjoin onto AP/NP/PP predicative trees with null copulas.

\subsection{Auxiliary Selection}
The {\bf $\langle$mode$\rangle$} feature is also used to state the
subcategorizational constraints between an auxiliary verb and its
complement. We model the following constraints:\\
{\em have} takes past participial complements\\
passive {\em be} takes past participial complements\\
active {\em be} takes progressive complements\\
modal verbs, {\em do}, and {\em to} take VPs headed by verbs in their
base form as their complements. 

An auxiliary verb transmits its own mode to its root and imposes its
subcategorizational restrictions on its complement i.e. on its foot node.
e.g. the auxiliary {\em have} in its infinitival form involves the
following equations:

\enumsentence{ {\bf VP$_{r}$.b:$\langle$mode$\rangle =$ 
          V.t:$\langle$mode$\rangle$}\label{mode1}}
\enumsentence{ {\bf  V.t:$\langle$mode$\rangle =$ base}\label{mode2}}
\enumsentence{ {\bf VP.b:$\langle$mode$\rangle =$ ppart}\label{mode3}}

\noindent
{\bf $\langle$passive$\rangle$}: This feature is used to ensure that
passives only have {\em be} as their auxiliary. Passive trees start
out with their {\bf $\langle$passive$\rangle$} feature as {\bf +}.
This feature starts out at the level of the verb and is percolated up
to the level of the VP. This ensures that only auxiliary verbs whose
foot node has {\bf +} as their {\bf $\langle$passive$\rangle$} feature
can adjoin on a passive. Passive trees have {\bf ppart} as the value
of their {\bf $\langle$mode$\rangle$} feature. So the only auxiliary
trees that we really have to worry about blocking are trees whose foot
nodes have {\bf ppart} as the value of their {\bf
$\langle$mode$\rangle$} feature. There are two such trees -- the {\em
be} tree and the {\em have} tree. The {\em be} tree is fine because
its foot node has {\bf +} as its {\bf $\langle$passive$\rangle$}
feature, so both the {\bf $\langle$passive$\rangle$} and {\bf
$\langle$mode$\rangle$} values unify; the {\em have} tree is blocked
because its foot node has {\bf --} as its {\bf
$\langle$passive$\rangle$} feature.

\section{Relative Clauses}
Features that are peculiar to the relative clause system are:\\
{\bf $\langle$select-mode$\rangle$}, possible values are {\bf ind, inf, ppart, ger}\\
{\bf $\langle$rel-pron$\rangle$}, possible values are {\bf ppart, ger, adj-clause}\\
{\bf $\langle$rel-clause$\rangle$}, possible values are {\bf +/--}

\noindent
{\bf $\langle$select-mode$\rangle$}:\\
Comps are lexically specified for {\bf $\langle$select-mode$\rangle$}.
In addition, the {\bf $\langle$select-mode$\rangle$} feature of a Comp
is equated to the {\bf $\langle$mode$\rangle$} feature of its
sister S node by the following equation:

\enumsentence{ {\bf Comp.t:$\langle$select-mode$\rangle =$ S$_{t}$.t:$\langle$mode$\rangle$}}

The lexical specifications of the Comps are shown below:
\begin{itemize}
\item $\epsilon$$_{C}$, {\bf Comp.t:$\langle$select-mode$\rangle
=$ind/inf/ger/ppart}
\item {\em that}, {\bf Comp.t:$\langle$select-mode$\rangle =$ind}
\item {\em for}, {\bf Comp.t:$\langle$select-mode$\rangle =$inf}
\end{itemize}

\noindent
{\bf $\langle$rel-pron$\rangle$}:\\
There are additional constraints on where the null Comp $\epsilon$$_{C}$
can occur. The null Comp is not permitted in cases of subject
extraction unless there is an intervening clause or or
the relative clause is a reduced relative ({\bf mode = ppart/ger}).

To model this paradigm, the feature {\bf $\langle$rel-pron$\rangle$} is used in
conjunction with the following equations.

\enumsentence{
{\bf S$_{r}$.t:$\langle$rel-pron$\rangle =$ Comp.t:$\langle$rel-pron$\rangle$}}
\enumsentence{
{\bf S$_{r}$.b:$\langle$rel-pron$\rangle =$ S$_{r}$.b:$\langle$mode$\rangle$}}
\enumsentence{
{\bf Comp.b:$\langle$rel-pron$\rangle =$ppart/ger/adj-clause}
(for $\epsilon$$_{C}$)}

The full set of the equations above is only present in Comp
substitution trees involving subject extraction. So the following will
not be ruled out.

\enumsentence{
the toy [$\epsilon$$_{i}$ [$\epsilon$$_{C}$ [ Dafna likes t$_{i}$ ]]] }

The feature mismatch induced by the above equations
is not remedied by adjunction of just any S-adjunct
because all other S-adjuncts
are transparent to the {\bf $\langle$rel-pron$\rangle$} feature
because of the following equation:

\enumsentence{
{\bf S$_{m}$.b:$\langle$rel-pron$\rangle =$ S$_{f}$.t:$\langle$rel-pron$\rangle$}}

\noindent
{\bf $\langle$rel-clause$\rangle$}:\\ The XTAG analysis forces the
adjunction of the determiner below the relative clause. This is done
by using the {\bf $\langle$rel-clause$\rangle$} feature. The relevant
equations are:

\enumsentence{ On the root of the RC: {\bf NP$_{r}$.b:$\langle$rel-clause$\rangle = +$}}
\enumsentence{ On the foot node of the 
Determiner tree: {\bf NP$_{f}$.t:$\langle$rel-clause$\rangle = -$}}

\section{Complementizer Selection}
The following features are used to ensure the appropriate distribution
of complementizers:
\\
{\bf $\langle$comp$\rangle$}, possible values: {\bf that, if, whether,
for, rel, inf\_nil, ind\_nil, nil}\\
{\bf $\langle$assign-comp$\rangle$}, possible values: {\bf that, if,
whether, for, ecm, rel, ind\_nil, inf\_nil, none}\\
{\bf $\langle$mode$\rangle$}, possible values: {\bf ind, inf, sbjnct, ger, base, ppart, 
nom, prep}\\
{\bf $\langle$wh$\rangle$}, possible values: {\bf +, --}

The value of the {\bf $\langle$comp$\rangle$} feature tells us what complementizer we 
are dealing with. The trees which introduce complementizers come 
specified from the lexicon with their 
{\bf $\langle$comp$\rangle$} feature and {\bf $\langle$assign-comp$\rangle$} 
feature. The {\bf $\langle$comp$\rangle$} of the Comp tree regulates 
what kind of tree goes above the Comp tree, while the 
{\bf $\langle$assign-comp$\rangle$} feature regulates what kind of tree
goes below.
e.g.
the following equations are used for {\em that}:

\enumsentence{ {\bf S$_{c}$.b:$\langle$comp$\rangle =$ Comp.t:$\langle$comp$\rangle$} }
\enumsentence{ {\bf S$_{c}$.b:$\langle$wh$\rangle =$ Comp.t:$\langle$wh$\rangle$}}
\enumsentence{ {\bf S$_{c}$.b:$\langle$mode$\rangle =$ ind/sbjnct}}
\enumsentence{ {\bf S$_{r}$.t:$\langle$assign-comp$\rangle =$ Comp.t:$\langle$comp$\rangle$}}
\enumsentence{ {\bf S$_{r}$.b:$\langle$comp$\rangle =$ nil}}

By specifying {\bf S$_{r}$.b:$\langle$comp$\rangle =$ nil}, we ensure that
complementizers do not adjoin onto other complementizers. The root node
of a complementizer tree always has its {\bf $\langle$comp$\rangle$} feature
set to a value other than {\bf nil}.

Trees that take clausal complements specify with the {\bf $\langle$comp$\rangle$} feature
on their foot node what kind of complementizer(s) they can take. 
The {\bf $\langle$assign-comp$\rangle$} feature of an S node is determined 
by the highest VP below the S node and the syntactic configuration
the S node is in. 

\subsection{Verbs with object sentential complements}
Finite sentential complements:

\enumsentence{ {\bf S$_{1}$.t:$\langle$comp$\rangle =$ that/whether/if/nil}}
\enumsentence{{\bf S$_{1}$.t:$\langle$mode$\rangle =$ ind/sbjnct} or {\bf S$_{1}$.t:$\langle$mode$\rangle =$ ind}}
\enumsentence{ {\bf S$_{1}$.t:$\langle$assign-comp$\rangle =$ ind\_nil/inf\_nil}}

The presence of an overt complementizer is optional.

Non-finite sentential complements, do not permit {\em for}:

\enumsentence{ {\bf S$_{1}$.t:$\langle$comp$\rangle =$ nil}}
\enumsentence{ {\bf S$_{1}$.t:$\langle$mode$\rangle =$ inf}}
\enumsentence{ {\bf S$_{1}$.t:$\langle$assign-comp$\rangle =$ ind\_nil/inf\_nil}
}

Non-finite sentential complements, permit {\em for}:

\enumsentence{ {\bf S$_{1}$.t:$\langle$comp$\rangle =$ for/nil}}
\enumsentence{ {\bf S$_{1}$.t:$\langle$mode$\rangle =$ inf}}
\enumsentence{ {\bf S$_{1}$.t:$\langle$assign-comp$\rangle =$ ind\_nil/inf\_nil}}

Cases like `*I want for to win' are independently ruled out due to a 
case feature clash between the {\bf acc} assigned by {\em for} and the
intrinsic case feature {\bf none} on the PRO.

Non-finite sentential complements, ECM:

\enumsentence{ {\bf S$_{1}$.t:$\langle$comp$\rangle =$ nil}}
\enumsentence{ {\bf S$_{1}$.t:$\langle$mode$\rangle =$ inf}}
\enumsentence{ {\bf S$_{1}$.t:$\langle$assign-comp$\rangle =$ ecm}}

\subsection{Verbs with sentential subjects}
The following contrast involving complementizers surfaces with sentential
subjects:

\enumsentence{ *(That) John is crazy is likely.}

Indicative sentential subjects obligatorily have complementizers while
infinitival sentential subjects may or may not have a complementizer. 
Also {\em if} is possible as the complementizer of an object clause
but not as the complementizer of a sentential subject. 

\enumsentence{ {\bf S$_{0}$.t:$\langle$comp$\rangle =$ that/whether/for/nil}}
\enumsentence{ {\bf S$_{0}$.t:$\langle$mode$\rangle =$ inf/ind}}
\enumsentence{ {\bf S$_{0}$.t:$\langle$assign-comp$\rangle =$ inf\_nil}}

If the sentential subject is finite and a complementizer does
not adjoin in, the {\bf $\langle$assign-comp$\rangle$} feature of the 
S$_{0}$ node of the embedding clause and the root node of the
embedded clause will fail to unify. If a complementizer adjoins in,
there will be no feature-mismatch because the root of the
complementizer tree is not specified for the {\bf $\langle$assign-comp$\rangle$} feature.

The {\bf $\langle$comp$\rangle$} feature {\bf nil} is split into two
{\bf $\langle$assign-comp$\rangle$} features {\bf ind\_nil} and
{\bf inf\_nil} to capture the fact that there are certain configurations in
which it is acceptable for an infinitival clause to lack a complementizer
but not acceptable for an indicative clause to lack a complementizer. 

\subsection{{\em That}-trace and {\em for}-trace effects}

\enumsentence{ Who$_{i}$ do you think (*that) t$_{i}$ ate the apple?}

{\em That} trace violations are blocked by the presence of the following
equation:

\enumsentence{ {\bf S$_{r}$.b:$\langle$assign-comp$\rangle =$ inf\_nil/ind\_nil/ecm}}

on the bottom of the S$_{r}$ nodes of trees with extracted subjects (W0). 
The {\bf ind\_nil} feature specification permits the above example
while the {\bf inf\_nil/ecm} feature specification allows the
following examples to be derived:

\enumsentence{ Who$_{i}$ do you want [ t$_{i}$ to win the World Cup]?}
\enumsentence{ Who$_{i}$ do you consider [ t$_{i}$ intelligent]?}

The feature equation that ruled out the {\em that}-trace filter violations
will also serve to rule out the {\em for}-trace violations above.

\section{Determiner ordering}
{\bf $\langle$card$\rangle$}, possible values are {\bf +, --}\\
{\bf $\langle$compl$\rangle$}, possible values are {\bf +, --}\\
{\bf $\langle$const$\rangle$}, possible values are {\bf +, --}\\
{\bf $\langle$decreas$\rangle$}, possible values are {\bf +, --}\\
{\bf $\langle$definite$\rangle$}, possible values are {\bf +, --}\\
{\bf $\langle$gen$\rangle$}, possible values are {\bf +, --}\\
{\bf $\langle$quan$\rangle$}, possible values are {\bf +, --}

For detailed discussion see Chapter \ref{det-comparitives}.

\section{Punctuation}
{\bf $\langle$punct$\rangle$} is a complex feature. It has the following
as its subfeatures:\\
{\bf $\langle$punct bal$\rangle$}, possible values are {\bf dquote,
squote, paren, nil}\\
{\bf $\langle$punct contains colon$\rangle$}, possible values are {\bf +, --}\\
{\bf $\langle$punct contains dash$\rangle$}, possible values are {\bf +, --}\\
{\bf $\langle$punct contains dquote$\rangle$}, possible values are {\bf +, --}\\
{\bf $\langle$punct contains scolon$\rangle$}, possible values are {\bf +, --}\\
{\bf $\langle$punct contains squote$\rangle$}, possible values are {\bf +, --}\\
{\bf $\langle$punct struct$\rangle$}, possible values are {\bf comma,
dash, colon, scolon, none, nil}\\
{\bf $\langle$punct term$\rangle$}, possible values are {\bf per, qmark, excl, 
none, nil}

For detailed discussion see Chapter~\ref{punct-chapt}.

\section{Conjunction}
{\bf $\langle$conj$\rangle$}, possible values are {\bf but, and, or,
comma, scolon, to, disc, nil}\\
The {\bf $\langle$conj$\rangle$} feature is specified in the lexicon
for each conjunction and is passed up to the root node 
of the conjunction tree. If the conjunction is {\em and}, the 
root {\bf $\langle$agr num$\rangle$} is {\bf $\langle$plural$\rangle$}, no
matter what the number of the two conjuncts. With {\em or}, the
the root {\bf $\langle$agr num$\rangle$} is equated to the
{\bf $\langle$agr num$\rangle$} feature of the right conjunct.

The {\bf $\langle$conj$\rangle$=disc} feature is only used at the root
of  the
$\beta$CONJs tree.  It blocks the adjunction of one $\beta$CONJs tree
on another.  The following equations are used, where S$_{r}$ is
the substitution node and S$_{c}$ is the root node:
\enumsentence{ S$_{r}$.t:$\langle$conj$\rangle$ = disc}
\enumsentence{ S$_{c}$.b:$\langle$conj$\rangle$ = and/or/but/nil}

\section{Comparatives}
{\bf $\langle$compar$\rangle$}, possible values are {\bf +, --}\\
{\bf $\langle$equiv$\rangle$}, possible values are {\bf +, --}\\
{\bf $\langle$super$\rangle$}, possible values are {\bf +, --}

For detailed discussion see Chapter~\ref{compars-chapter}.

\section{Control}
{\bf $\langle$control$\rangle$} has no value and is used only for indexing
purposes.  The root node of every clausal tree has its {\bf
$\langle$control$\rangle$} feature coindexed with the control feature of
its subject.  This allows adjunct control to take place. In addition,
clauses that take infinitival clausal complements have the control feature
of their subject/object coindexed with the control feature of their
complement clause S, depending upon whether they are subject control verbs
or object control verbs respectively.

\section{Other Features}
{\bf $\langle$neg$\rangle$}, possible values are {\bf +, --}\\
Used for controlling the interaction of negation and auxiliary verbs.

\noindent
{\bf $\langle$pred$\rangle$}, possible values are {\bf +, --}\\
The {\bf $\langle$pred$\rangle$} feature is used in the following tree
families: Tnx0N1.trees and Tnx0nx1ARB.trees.
In the Tnx0N1.trees family, the following equations are used:\\
for $\alpha$W1nx0N1:

\enumsentence{ NP$_{1}$.t:$\langle$pred$\rangle$ = +}
\enumsentence{ NP$_{1}$.b:$\langle$pred$\rangle$ = +}
\enumsentence{ NP.t:$\langle$pred$\rangle$ = +}
\enumsentence{ N.t:$\langle$pred$\rangle$ = NP.b:$\langle$pred$\rangle$}

This is the only tree in this tree family to use the 
{\bf $\langle$pred$\rangle$} feature.

The other tree family where the {\bf $\langle$pred$\rangle$} feature is
used is Tnx0nx1ARB.trees.  Within this family, this feature (and the
following equations) are used only in the $\alpha$W1nx0nx1ARB tree.

\enumsentence{ AdvP$_{1}$.t:$\langle$pred$\rangle$ = +}
\enumsentence{ AdvP$_{1}$.b:$\langle$pred$\rangle$ = +}
\enumsentence{ NP.t:$\langle$pred$\rangle$ = +}
\enumsentence{ AdvP.b:$\langle$pred$\rangle$ = NP.t:$\langle$pred$\rangle$}

\noindent
{\bf $\langle$pron$\rangle$}, possible values are {\bf +, --}\\
This feature indicates whether a particular NP is a pronoun or not. 
Certain constructions which do not permit pronouns use this 
feature to block pronouns.

\noindent
{\bf $\langle$tense$\rangle$}, possible values are {\bf pres, past}\\
It does not seem to be the case that the {\bf $\langle$tense$\rangle$}
feature interacts with other features/syntactic processes. It 
comes from the lexicon with the verb and is transmitted up the
tree in such a way that the root S node ends up with the
tense feature of the highest verb in the tree. The equations
used for this purpose are:

\enumsentence{ {\bf S$_{r}$.b:$\langle$tense$\rangle$ = VP.t:$\langle$tense$\rangle$}}
\enumsentence{ {\bf VP.b:$\langle$tense$\rangle$ = V.t:$\langle$tense$\rangle$}}

{\bf $\langle$trans$\rangle$}, possible values are {\bf +, --}\\
Many but not all English verbs can anchor both transitive and intransitive trees.

\enumsentence{ The sun melted the ice cream.}
\enumsentence{ The ice cream melted.}
\enumsentence{ Elmo borrowed a book.}
\enumsentence{ * A book borrowed.}

Transitive trees have the {\bf $\langle$trans$\rangle$} feature of their
anchor set to {\ +} and intransitive trees have the 
{\bf $\langle$trans$\rangle$} feature of their
anchor set to {\ --}. Verbs such as {\em melt} which can occur 
in both transitive and intransitive trees come unspecified for the 
{\bf $\langle$trans$\rangle$} feature from the lexicon. Verbs which 
can only occur in transitive trees e.g. {\em borrow} have their
{\bf $\langle$trans$\rangle$} feature 
specified in the lexicon as {\bf +} thus blocking their anchoring of 
an intransitive tree.

\chapter{Evaluation and Results}
\label{evaluation}

In this appendix we describe various evaluations done of the XTAG
grammar. Some of these evaluations were done on an earlier version of
the XTAG grammar (the 1995 release), while other were done more
recently. We will try to indicate in each section which version was
used. 

\section{Parsing Corpora}

In the XTAG project, we have used corpus analysis in two main ways:
(1) to measure the performance of the English grammar on a given genre
and (2) to identify gaps in the grammar. 

The second type of evaluation involves performing detailed error
analysis on the sentences rejected by the parser, and we have done
this several times on WSJ and Brown data.

Based on the results of such analysis, we prioritize upcoming grammar
development efforts. The results of a recent error analysis are shown
in Table \ref{errors}.  The table does not show errors in parsing due
to mistakes made by the POS tagger which contributed the largest
number of errors: 32. At this point, we have added a treatment of
punctuation to handle \#1, an analysis of time NPs (\#2), a large
number of multi-word prepositions (part of \#3), gapless relative
clauses (\#7), bare infinitives (\#14) and have added the missing
subcategorization (\#3) and missing lexical entry (\#12).  We are in
the process of extending the parser to handle VP coordination (\#9)
(See Section~\ref{conjunction} on recent work to handle VP and other
predicative coordination). We find that this method of error analysis
is very useful in focusing grammar development in a productive
direction.

\begin{table}[htb]
\centering
\begin{tabular}{|l|l|l|} \hline
Rank & No of errors & Category of error \\ \hline
\#1  & 11  &    Parentheticals and appositives \\ \hline
\#2  & 8     &  Time NP \\ \hline
\#3  & 8  &     Missing subcat \\ \hline
\#4  & 7 &      Multi-word construction \\ \hline
\#5  & 6 &       Ellipsis \\ \hline
\#6  & 6  &      Not sentences \\ \hline
\#7  & 3  &      Relative clause with no gap \\ \hline
\#8  & 2  &      Funny coordination \\ \hline
\#9  & 2  &      VP coordination \\ \hline
\#10  & 2  &      Inverted predication \\ \hline
\#11  & 2  &      Who knows \\ \hline
\#12  & 1  &      Missing entry \\ \hline
\#13  & 1   &     Comparative? \\ \hline
\#14  & 1    &    Bare infinitive \\ \hline
\end{tabular}
\caption{Results of Corpus Based Error Analysis}
\label{errors}
\end{table}

To ensure that we are not losing coverage of certain phenomena as we
extend the grammar, we have a benchmark set of grammatical and
ungrammatical sentences from this technical report. We parse these
sentences periodically to ensure that in adding new features and
constructions to the grammar, we are not blocking previous analyses.
There are approximately $590$ example sentences in this set.

\section{TSNLP}

In addition to corpus-based evaluation, we have also run the English
Grammar on the Test Suites for Natural Language Processing (TSNLP)
English corpus \cite{Lehmann96}. The corpus is intended to be a
systematic collection of English grammatical phenomena, including
complementation, agreement, modification, diathesis, modality, tense
and aspect, sentence and clause types, coordination, and negation. It
contains 1409 grammatical sentences and phrases and 3036 ungrammatical
ones.

\begin{table*}[htb]
\centering
\begin{tabular}{|l|c|c|}
\hline
Error Class & \% & Example \\ \hline
POS Tag &  19.7\% & She adds  to/V it , He noises/N him abroad \\ \hline
Missing lex item & 43.3\% & {\it used} as an auxiliary V, {\it calm NP down} \\ \hline
Missing tree & 21.2\% & {\it should've}, {\it bet NP NP S}, {\it
regard NP as Adj} \\ \hline
Feature clashes & 3\% & {\it My every firm}, {\it All money} \\ \hline
Rest&12.8\% & {\it approx}, {\it e.g.} \\
\hline
\end{tabular}
\caption{Breakdown of TSNLP Errors}
\label{tsnlp-table}
\end{table*}

There were 42 examples which we judged ungrammatical, and removed from
the test corpus. These were sentences with conjoined subject pronouns,
where one or both were accusative, e.g. {\it Her and him succeed.}
Overall, we parsed 61.4\% of the 1367 remaining sentences and
phrases. The errors were of various types, broken down in
Table~\ref{tsnlp-table}. As with the error analysis described above,
we used this information to help direct our grammar development
efforts. It also highlighted the fact
that our grammar is heavily slanted toward American English---our
grammar did not handle {\it dare} or {\it need} as auxiliary verbs,
and there were a number of very British particle constructions,
e.g. {\it She misses him out}. 

One general problem with the test-suite is that it uses a very
restricted lexicon, and if there is one problematic lexical item it is
likely to appear a large number of times and cause a disproportionate
amount of grief. {\it Used to} appears 33 times and we got all 33
wrong. However, it must be noted that the XTAG grammar has analyses
for syntactic phenomena that were not represented in the TSNLP test
suite such as sentential subjects and subordinating clauses among
others. This effort was, therefore, useful in highlighting some
deficiencies in our grammar, but did not provide the same sort of
general evaluation as parsing corpus data.

\section{Chunking and Dependencies in XTAG Derivations}

We evaluated the XTAG parser for the text chunking
task~\cite{abney91}. In particular, we compared NP chunks and verb
group (VG) chunks\footnote{We treat a sequence of verbs and verbal
  modifiers, including auxiliaries, adverbs, modals as constituting a
  verb group.}  produced by the XTAG parser with the NP and VG chunks
from the Penn Treebank~\cite{marcus93}. The test involved $940$
sentences of length $15$ words or less from sections $17$ to $23$ of
the Penn Treebank, parsed using the XTAG English grammar. The results
are given in Table~\ref{chunking-results}.

\begin{table*}[htb]
\centering
\begin{tabular}{|l|c|c|}
\hline
& NP Chunking & VG Chunking \\ \hline
Recall & 82.15\% & 74.51\% \\ \hline
Precision & 83.94\%  & 76.43\% \\ \hline
\end{tabular}
\caption{Text Chunking performance of the XTAG parser}
\label{chunking-results}
\end{table*} 

\begin{table*}[htb]
\centering
\begin{tabular}{|c|c|c|c|} \hline
System & Training Size & Recall & Precision  \\ \hline \hline
Ramshaw \& Marcus & Baseline & 81.9\% & 78.2\% \\ \hline
Ramshaw \& Marcus & 200,000 & 90.7\% & 90.5\% \\ 
(without lexical information) & & & \\ \hline 
Ramshaw \& Marcus & 200,000 & 92.3\% & 91.8\% \\ 
(with lexical information) & & & \\ \hline \hline
Supertags & Baseline & 74.0\% & 58.4\% \\ \hline
Supertags & 200,000 & 93.0\% & 91.8\% \\ \hline
Supertags & 1,000,000 & 93.8\% & 92.5\% \\ \hline
\end{tabular}
\caption{Performance comparison of the transformation based noun
chunker and the supertag based noun chunker}
\label{nc-compare}
\end{table*}

As described earlier, the results cannot be directly compared with
other results in chunking such as in~\cite{lance&mitch95} since we do
not train from the Treebank before testing. However, in earlier work,
text chunking was done using a technique called
supertagging~\cite{srini97iwpt} (which uses the XTAG English grammar)
which can be used to train from the Treebank.  The comparative results
of text chunking between supertagging and other methods of chunking is
shown in Figure~\ref{nc-compare}.\footnote{It is important to note in
  this comparison that the supertagger uses lexical information on a
  per word basis only to pick an initial set of supertags for a given
  word.}

We also performed experiments to determine the accuracy of the
derivation structures produced by XTAG on WSJ text, where the
derivation tree produced after parsing XTAG is interpreted as a
dependency parse. We took sentences that were $15$ words or less from
the Penn Treebank~\cite{marcus93}. The sentences were collected from
sections $17$--$23$ of the Treebank. $9891$ of these sentences were
given at least one parse by the XTAG system. Since XTAG typically
produces several derivations for each sentence we simply picked a
single derivation from the list for this evaluation. Better results
might be achieved by ranking the output of the parser using the sort
of approach described in~\cite{srinietal95}.

There were some striking differences in the dependencies implicit in
the Treebank and those given by XTAG derivations. For instance, often
a subject NP in the Treebank is linked with the first auxiliary verb
in the tree, either a modal or a copular verb, whereas in the XTAG
derivation, the same NP will be linked to the main verb. Also XTAG
produces some dependencies within an NP, while a large number of words
in NPs in the Treebank are directly dependent on the verb. To
normalize for these facts, we took the output of the NP and VG chunker
described above and accepted as correct any dependencies that were
completely contained within a single chunk.

For example, for the sentence {\em Borrowed shares on the Amex rose to
another record}, the XTAG and Treebank chunks are shown below.

\begin{verbatim}
XTAG chunks:     
 [Borrowed shares] [on the Amex] [rose] 
    [to another record] 
Treebank chunks: 
 [Borrowed shares on the Amex] [rose] 
    [to another record] 
\end{verbatim}

Using these chunks, we can normalize for the fact that in the
dependencies produced by XTAG {\em borrowed} is dependent on {\em
shares} (i.e. in the same chunk) while in the Treebank {\em borrowed}
is directly dependent on the verb {\em rose}. That is to say, we are
looking at links between \underline{chunks}, not between
\underline{words}. The dependencies for the sentence are given below.

\begin{verbatim}
XTAG dependency    Treebank dependency

Borrowed::shares   Borrowed::rose 
shares::rose       shares::rose 
on::shares         on::shares 
the::Amex          the::Amex 
Amex::on           Amex::on 
rose::NIL          rose::NIL
to::rose           to::rose 
another::record    another::record 
record::to         record::to 
\end{verbatim}

After this normalization, testing simply consisted of counting how
many of the dependency links produced by XTAG matched the Treebank
dependency links. Due to some tokenization and subsequent alignment
problems we could only test on $835$ of the original $9891$ parsed
sentences. There were a total of $6135$ dependency links extracted
from the Treebank. The XTAG parses also produced $6135$ dependency
links for the same sentences. Of the dependencies produced by the XTAG
parser, $5165$ were correct giving us an accuracy of $84.2\%$.

\section{Comparison with IBM}

The evaluation in this section was done with the earlier 1995 release
of the grammar. This section describes an experiment to measure the
crossing bracket accuracy of the XTAG-parsed IBM-manual sentences.  In
this experiment, XTAG parses of 1100 IBM-manual sentences have been
ranked using certain heuristics. The ranked parses have been
compared\footnote{We used the parseval program written by Phil Harison
  (phil@atc.boeing.com).}  against the bracketing given in the
Lancaster Treebank of IBM-manual sentences\footnote{The Treebank was
  obtained through Salim Roukos (roukos@watson.ibm.com) at IBM.}.
Table~\ref{ibm-results} shows the results of XTAG obtained in this
experiment, which used the highest ranked parse for each system. It
also shows the results of the latest IBM statistical grammar
(\cite{jelineketal94}) on the same genre of sentences. Only the
highest-ranked parse of both systems was used for this evaluation.
Crossing Brackets is the percentage of sentences with no pairs of
brackets crossing the Treebank bracketing (i.e.  (~(~a~b~)~c~) has a
crossing bracket measure of one if compared to (~a~(~b~c~)~)~). Recall
is the ratio of the number of constituents in the XTAG parse to the
number of constituents in the corresponding Treebank sentence.
Precision is the ratio of the number of correct constituents to the
total number of constituents in the XTAG parse.

\begin{table}[ht]
\centering
\begin{tabular}{|l|c|c|c|c|} \hline 
System & \# of & Crossing Bracket & Recall & Precision \\
& sentences & Accuracy & & \\ \hline
XTAG & 1100 & 81.29\% & 82.34\% & 55.37\% \\ \hline
IBM Statistical & 1100 & 86.20\% & 86.00\% & 85.00\% \\
grammar & &  &  &\\ \hline
\end{tabular}

\vspace{0.1in}

\caption{Performance of XTAG on IBM-manual sentences}
\label{ibm-results} 

\end{table}

As can be seen from Table~\ref{ibm-results}, the precision figure for
the XTAG system is considerably lower than that for IBM. For the
purposes of comparative evaluation against other systems, we had to
use the same crossing-brackets metric though we believe that the
crossing-brackets measure is inadequate for evaluating a grammar like
XTAG. There are two reasons for the inadequacy. First, the parse
generated by XTAG is much richer in its representation of the internal
structure of certain phrases than those present in manually created
treebanks (e.g. IBM: [$_N$ your personal computer], XTAG: [$_{NP}$
[$_G$ your] [$_N$ [$_N$ personal] [$_N$ computer]]]). This is
reflected in the number of constituents per sentence, shown in the
last column of Table~\ref{const-no}.\footnote{We are aware of the fact
  that increasing the number of constituents also increases the recall
  percentage. However we believe that this a legitimate gain.}

\begin{table}[ht]
\centering
\begin{tabular}{|l|c|c|c|c|} \hline
System & Sent. & \# of & Av. \# of & Av. \# of \\
& Length & sent & words/sent & Constituents/sent \\ \hline
XTAG & 1-10 & 654 & 7.45 & 22.03  \\ \cline{2-5}
& 1-15 & 978 & 9.13 & 30.56 \\ \hline
IBM Stat. & 1-10 & 447 & 7.50 & 4.60 \\ \cline{2-5}
Grammar & 1-15 & 883 & 10.30 & 6.40 \\ \hline
\end{tabular}
\caption{Constituents in XTAG parse and IBM parse}
\label{const-no}
\end{table}

A second reason for considering the crossing bracket measure
inadequate for evaluating XTAG is that the primary structure in XTAG
is the derivation tree from which the bracketed tree is derived. Two
identical bracketings for a sentence can have completely different
derivation trees (e.g. {\it kick the bucket} as an idiom vs. a
compositional use). A more direct measure of the performance of XTAG
would evaluate the derivation structure, which captures the
dependencies between words.

\section{Comparison with Alvey}

The evaluation in this section was done with the earlier 1995 release
of the grammar. This section compares XTAG to the Alvey Natural
Language Tools (ANLT) Grammar. We parsed the set of LDOCE Noun Phrases
presented in Appendix B of the technical report (\cite{Carroll93})
using XTAG.  Table~\ref{Alvey-xtag} summarizes the results of this
experiment.  A total of 143 noun phrases were parsed. The NPs which
did not have a correct parse in the top three derivations were
considered failures for either system. The maximum and average number
of derivations columns show the highest and the average number of
derivations produced for the NPs that have a correct derivation in the
top three.  We show the performance of XTAG both with and without the
tagger since the performance of the POS tagger is significantly
degraded on the NPs because the NPs are usually shorter than the
sentences on which it was trained. It would be interesting to see if
the two systems performed similarly on a wider range of data.

\begin{table}[ht]
\centering
\begin{tabular}{|l|c|c|c|c|c|}  \hline

System & \# of & \# parsed & \% parsed & Maximum & Average \\
& NPs &&& derivations & derivations \\ \hline
ANLT Parser & 143 & 127 & 88.81\% & 32 & 4.57 \\ \hline
XTAG Parser with & 143 & 93 & 65.03\% & 28 & 3.45 \\
POS tagger & & & & & \\ \hline
XTAG Parser without & 143 & 120 & 83.91\% & 28 & 4.14\\
POS tagger & & & & & \\ \hline
\end{tabular} \\

\vspace{0.1in}

\caption{Comparison of XTAG and ANLT Parser}
\label{Alvey-xtag}
\end{table}

\section{Comparison with CLARE}

The evaluation in this section was done with the earlier 1995 release
of the grammar. This section compares the performance of XTAG against
that of the CLARE-2 system (\cite{clare-report92}) on the ATIS corpus.
Table~\ref{clare-results} shows the performance results. The
percentage parsed column for both systems represents the percentage of
sentences that produced any parse.  It must be noted that the
performance result shown for CLARE-2 is without any tuning of the
grammar for the ATIS domain. The performance of CLARE-3, a later
version of the CLARE system, is estimated to be 10\% higher than that
of the CLARE-2 system.\footnote{When CLARE-3 is tuned to the ATIS
  domain, performance increases to 90\%. However XTAG has not been
  tuned to the ATIS domain.}

\begin{table}[ht]
\centering
\begin{tabular}{|l|c|c|}  \hline

System & Mean length & \% parsed \\ \hline
CLARE-2  & 6.53 & 68.50\% \\ \hline
XTAG  & 7.62 & 88.35\% \\ \hline
\end{tabular}
\caption{Performance of CLARE-2 and XTAG on the ATIS corpus}
\label{clare-results}
\end{table}

In an attempt to compare the performance of the two systems on a wider
range of sentences (from similar genres), we provide in
Table~\ref{clare-results1} the performance of CLARE-2 on LOB corpus
and the performance of XTAG on the WSJ corpus. The performance was
measured on sentences of up to 10 words for both systems.
\begin{table}[ht]
\centering
\begin{tabular}{|c|c|c|c|}  \hline

System & Corpus & Mean length & \% parsed \\ \hline
CLARE-2 & LOB & 5.95 & 53.40\% \\ \hline
XTAG & WSJ & 6.00 & 55.58\% \\ \hline
\end{tabular}
\caption{Performance of CLARE-2 and XTAG on LOB and WSJ corpus
respectively}
\label{clare-results1}
\end{table}

\bibliographystyle{aaai-named}
\bibliography{xtag}

\end{document}